%% file: GaugeCNNTheory.tex
\documentclass{article}
\usepackage{arxiv} 

\usepackage[T1]{fontenc}
\usepackage[utf8]{inputenc}

\usepackage{amsmath}
\usepackage{amsfonts}
\usepackage{amssymb}
\usepackage{dsfont} 

\usepackage[
    scr=dutchcal, scrscaled=1.05,
  ]{mathalfa} 
\usepackage{bm}
\usepackage{phoenician}

\usepackage{anyfontsize}

\usepackage{mathtools}

\usepackage{graphicx}

\usepackage{float}

\usepackage{marginnote}

\usepackage{wrapfig}

\usepackage{sidecap}

\usepackage[font=small]{caption}
\usepackage[font=small]{subcaption}

\usepackage{booktabs}
\usepackage{makecell} 
\usepackage{diagbox}
\usepackage{multirow}
\newcounter{magicrownumbers} 
\newcommand\rownumber{\stepcounter{magicrownumbers}\arabic{magicrownumbers}}

\usepackage{pdflscape} 

\usepackage{enumitem}
\setlist[description]{
    labelwidth=0pt,
    leftmargin=35pt, 
    itemindent=-20pt, 
    font=\normalfont\bfseries\rule[2.2pt]{2pt}{2pt}\space
}

\usepackage[table]{xcolor} 

\usepackage{tikz}
\usetikzlibrary{arrows}
\usetikzlibrary{calc}
\usetikzlibrary{positioning}

\usepackage{tikz-cd}
\tikzcdset{every label/.append style = {font = \small}}

\usepackage{nicefrac}

\usepackage{accents}

\usepackage{afterpage}

\usepackage{hyperref}

\newcommand{\nocontentsline}[3]{}
\newcommand{\tocless}[2]{\bgroup\let\addcontentsline=\nocontentsline#1{#2}\egroup}
\newcommand{\toclesslab}[3]{\bgroup\let\addcontentsline=\nocontentsline#1{#2\label{#3}}\egroup}

\usepackage{etoc} 

\usepackage[numbers]{natbib}

\usepackage[amsmath,thmmarks,thref,hyperref]{ntheorem}

\makeatletter
\def\thm@@thmline@name#1#2#3#4{%
        \@dottedtocline{-2}{0em}{2.3em}%
                   {\makebox[\widesttheorem][l]{#1 \protect\numberline{#2}}#3}%
                   {#4}}
\@ifpackageloaded{hyperref}{
\def\thm@@thmline@name#1#2#3#4#5{%
    \ifx\\#5\\%
        \@dottedtocline{-2}{0em}{2.3em}%
            {\makebox[\widesttheorem][l]{#1 \protect\numberline{#2}}#3}%
            {#4}
    \else
        \ifHy@linktocpage\relax\relax
            \@dottedtocline{-2}{0em}{2.3em}%
                {\makebox[\widesttheorem][l]{#1 \protect\numberline{#2}}#3}%
                {\hyper@linkstart{link}{#5}{#4}\hyper@linkend}%
        \else
            \@dottedtocline{-2}{0em}{2.3em}%
                {\hyper@linkstart{link}{#5}%
                  {\makebox[\widesttheorem][l]{#1 \protect\numberline{#2}}#3}\hyper@linkend}%
                    {#4}%
        \fi
    \fi}
}
\makeatother
\newlength\widesttheorem
\AtBeginDocument{
  \settowidth{\widesttheorem}{Proposition 10\quad}
}

\theoremstyle{plain}
\theoremheaderfont{\normalfont\bfseries\hspace{-\theoremindent}}
\theoremseparator{.}
\theorembodyfont{\itshape}
\theoremstyle{plain}
\newtheorem{pro}{Proposition}[section]

\newtheorem{thm}[pro]{Theorem}
\newtheorem{dfn}[pro]{Definition}

\newtheorem*{rem*}{Remark}

\theoremheaderfont{\sc}\theorembodyfont{\upshape}
\theoremstyle{nonumberplain}
\theoremheaderfont{\normalfont\it\hspace{-\theoremindent}}
\theoremseparator{:}
\theoremsymbol{$\Box$}
\newtheorem{proof}{Proof}

\usepackage{custom_commands}

\usepackage{sectsty}
\partfont{\fontsize{16.2}{19.2}\bfseries\textsl}

\renewcommand\part{%
    \markboth{}{}\secdef\@part\@spart
}

\title{Coordinate Independent Convolutional Networks}
\newcommand{\subtitle}{\\[1.25ex]
    \fontsize{11.5}{13.8}\selectfont
    Isometry and Gauge Equivariant Convolutions on Riemannian Manifolds
}

\date{}

\author{%
    Maurice Weiler \\
    University of Amsterdam \\
    \texttt{m.weiler.ml@gmail.com} \\
    \And
    Patrick Forré \\
    University of Amsterdam \\
    \texttt{p.d.forre@uva.nl} \\
    \AND
    Erik Verlinde \\
    University of Amsterdam \\
    \texttt{e.p.verlinde@uva.nl} \\
    \And
    Max Welling \\
    University of Amsterdam, \\
    Qualcomm AI Research \\
    \texttt{m.welling@uva.nl} \\
}

\begin{document}

\maketitle

\vspace*{-5ex}
\input{chapters/00_abstract.tex}
\newpage

\input{chapters/10_intro.tex}
\input{chapters/11_outline.tex}

\input{chapters/20_visual_intro.tex}

\input{chapters/P1_intro.tex}

\input{chapters/30_local_intro.tex}
\input{chapters/31_gauges.tex}
\input{chapters/32_feature_fields.tex}
\input{chapters/33_local_transport.tex}
\input{chapters/34_isometries.tex}

\input{chapters/40_local_gauge_cnns_intro.tex}

\input{chapters/41_pointwise_operations.tex}
\input{chapters/42_gauge_conv.tex}
\input{chapters/43_isometry_equivariance.tex}

\input{chapters/50_mobius_conv.tex}

\input{chapters/P2_intro.tex}

\input{chapters/60_bundles_intro.tex}
\input{chapters/61_general_bundles.tex}
\input{chapters/62_TM_FM_associated_GL.tex}
\input{chapters/63_associated_G_bundles.tex}
\input{chapters/64_trivializations.tex}
\input{chapters/65_bundle_transport.tex}

\input{chapters/70_global_conv_intro.tex}
\input{chapters/71_global_onexone.tex}
\input{chapters/72_GMconv.tex}

\input{chapters/80_isom_intro.tex}
\input{chapters/81_isom_action.tex}
\input{chapters/82_isom_equivariance.tex}
\input{chapters/83_quotient_kernel_fields.tex}

\input{chapters/P3_intro.tex}

\input{chapters/90_Euclidean_intro.tex}
\input{chapters/91_Euclidean_steerable_Rd.tex}

\input{chapters/92_Euclidean_affine_geom.tex}
\input{chapters/93_Euclidean_GM_conv.tex}
\input{chapters/94_Euclidean_literature.tex}

\input{chapters/100_polar_intro.tex}

\input{chapters/101_polar_Euc2_rot.tex}
\input{chapters/102_polar_Euc2_logpolar.tex}
\input{chapters/103_polar_Euc3.tex}

\input{chapters/110_spherical_intro.tex}
\input{chapters/111_spherical_geometry.tex}
\input{chapters/112_spherical_SO3.tex}
\input{chapters/113_spherical_SO2.tex}
\input{chapters/114_spherical_ico.tex}

\input{chapters/120_mesh_intro.tex}
\input{chapters/121_mesh_geometry.tex}
\input{chapters/122_mesh_SO2.tex}
\input{chapters/123_mesh_e-steer.tex}

\newpage
\input{chapters/130_conclusions.tex}

\newpage
\appendix
\mypart{Appendix}

\input{chapters/apx01_coordinate_bases.tex}
\input{chapters/apx02_kernel_figures.tex}
\input{chapters/apx03_tangent_integral.tex}
\input{chapters/apx04_homogeneous_conv.tex}
\input{chapters/apx05_lifting_iso_proof.tex}
\input{chapters/apx06_spherical_kernels.tex}

\input{chapters/apx07_smoothness_kernel_field_trafo.tex}
\input{chapters/apx08_regular_fields_as_GM_functions.tex}

\newpage
\bibliographystyle{plainnat}
\bibliography{refs}

\end{document}

%% file: chapters/00_abstract.tex

\begin{minipage}[t]{\textwidth}
\begin{abstract}

    Motivated by the vast success of deep convolutional networks, there is a great interest in generalizing convolutions to non-Euclidean manifolds.
    A major complication in comparison to flat spaces is that it is unclear in which alignment a convolution kernel should be applied on a manifold.
    The underlying reason for this ambiguity is that general manifolds do not come with a canonical choice of reference frames (gauge).
    Kernels and features therefore have to be expressed relative to \emph{arbitrary coordinates}.
    We argue that the particular choice of coordinatization should not affect a network's inference -- it should be \emph{coordinate independent}.
    A simultaneous demand for coordinate independence and weight sharing is shown to result in a requirement on the network to be \emph{equivariant under local gauge transformations} (changes of local reference frames).
    The ambiguity of reference frames depends thereby on the $G$-\emph{structure} of the manifold,
    such that the necessary level of gauge equivariance is prescribed by the corresponding \emph{structure group}~$G$.
    Coordinate independent convolutions are proven to be equivariant w.r.t. those \emph{isometries} that are symmetries of the $G$-structure.
    The resulting theory is formulated in a coordinate free fashion in terms of fiber bundles.
    To exemplify the design of coordinate independent convolutions, we implement a convolutional network on the M\"obius strip.
    The generality of our differential geometric formulation of convolutional networks is demonstrated by an extensive literature review
    which explains a large number of Euclidean CNNs, spherical CNNs and CNNs on general surfaces as specific instances of coordinate independent convolutions.

    \begin{figure}[H]
        \centering
        \vspace*{2.5ex}
        \includegraphics[width=.94\columnwidth]{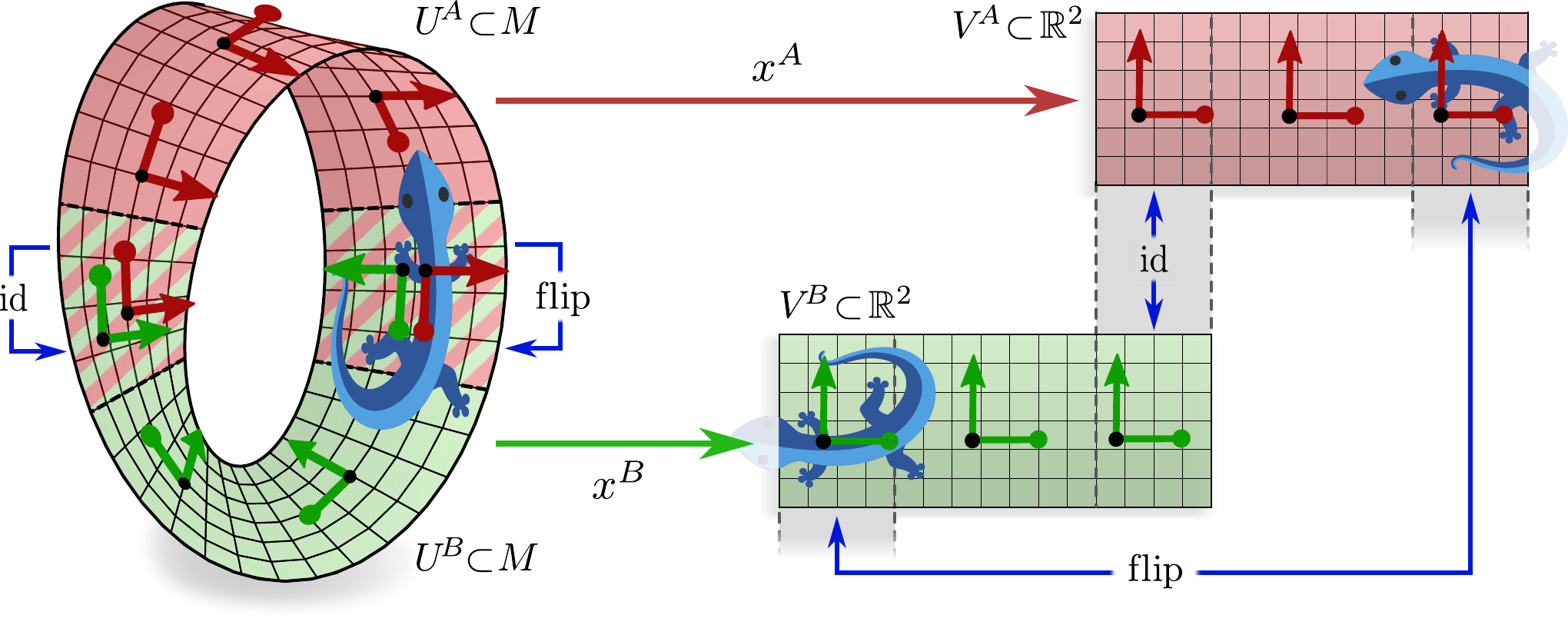}
    \end{figure}

\end{abstract}

\vspace*{-20ex}

\end{minipage}

%% file: chapters/10_intro.tex

\section{Introduction}

\begin{figure}
    \centering
    \vspace*{-2ex}
    \includegraphics[width=.94\textwidth]{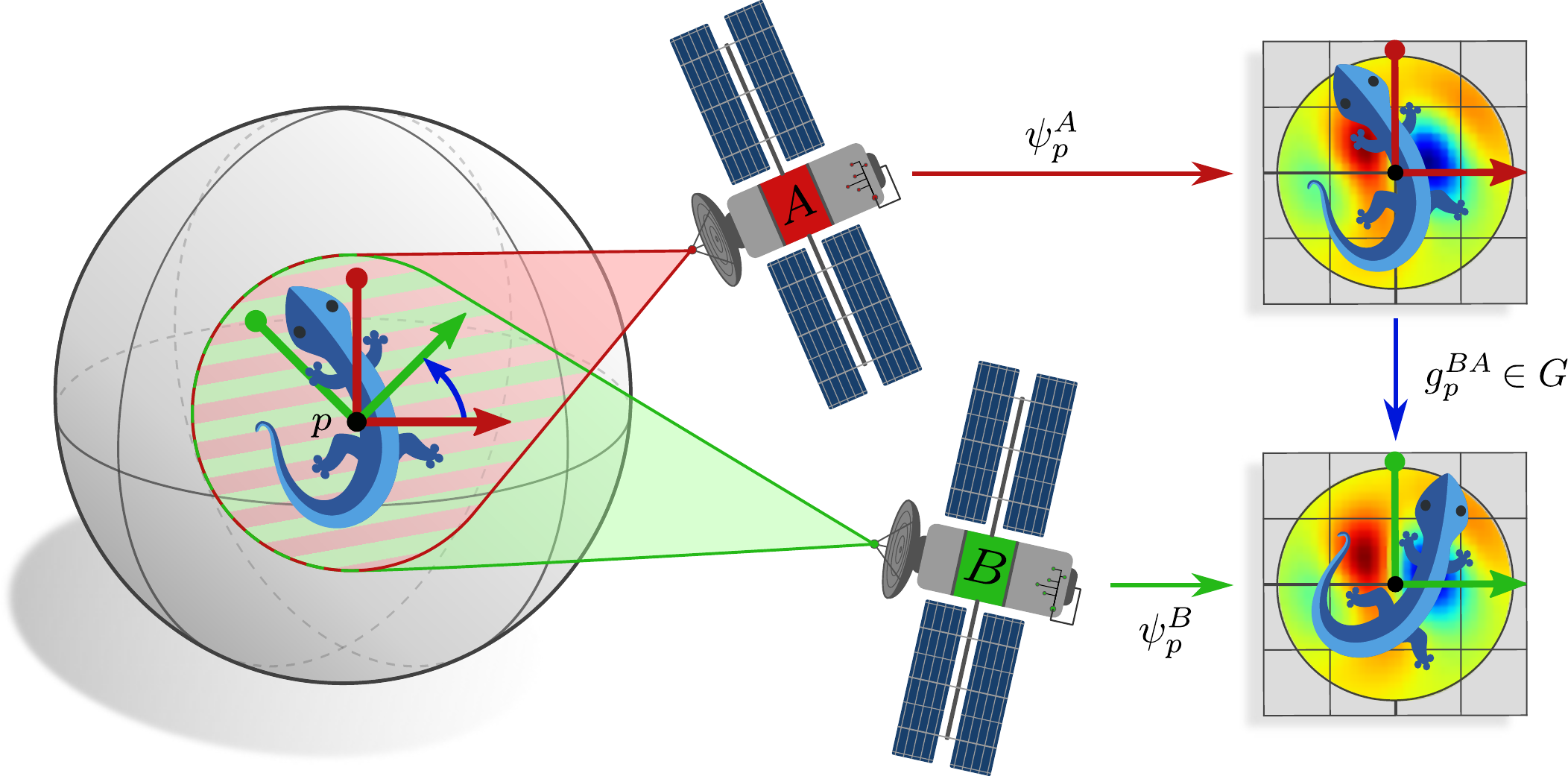}
    \caption{\small
        Different observers $A$ and $B$ may perceive a pattern of features from a different ``viewpoint''.
        The satellites in our application are convolution kernels which summarize their local field of view around~$p$ into a feature vector at~$p$.
        Their ``viewpoint'' is a choice of local reference frame (gauge) at~$p$, along which the kernel is aligned.
        Since the observations from both viewpoints represent the same pattern, the kernel responses should contain equivalent information, that is, the inference should be \emph{coordinate independent}.
        This constrains the convolution kernels to be \emph{equivariant under local gauge transformations}, i.e. changes of reference frames.
        The level of gauge equivariance is determined by the \emph{structure group}~$G$, which depends both on the manifold and the application.
        {\\
        \color{gray}
        \scriptsize
            (Lizards adapted under the Creative Commons Attribution 4.0 International
            \href{https://github.com/twitter/twemoji/blob/gh-pages/LICENSE-GRAPHICS}{\underline{license}}
            by courtesy of Twitter.)
        }
    }
    \label{fig:satellite}
\end{figure}

In recent years, deep neural networks have become the models of choice for a wide range of tasks in machine learning.
The success of deep models is often rooted in a task-specific design, reflecting the mathematical structure of the data to process.
A prominent example are convolutional neural networks (CNNs), which exploit the spatial structure of the data via a local connectivity and spatial weight sharing.
As the same kernel is applied at each point in space, convolutional networks are translation equivariant, which means that they generalize learned patterns automatically over spatial positions.
Given the considerable empirical success of Euclidean CNNs, there is a big interest in extending convolutional models to process signals on more general domains and to make them equivariant under larger symmetry groups.

This work investigates the generalization of convolutional networks to \emph{Riemannian manifolds}.
A major complication in generalizing convolutional networks from Euclidean spaces $\R^d$ to general Riemannian manifolds is that \emph{manifolds do not come with a preferred choice of reference direction}, along which a convolution kernel could be aligned to measure features.
Since no reference direction is preferred, the kernel needs to be aligned \emph{arbitrarily} on the manifold.
The central theme of this work is to regulate this arbitrariness by making the networks' inference independent from the specific alignment of convolution kernels.
It turns out that this requires kernels to be \emph{gauge equivariant}, i.e. equivariant under transformations of the kernel alignment.
The response of a gauge equivariant kernel transforms predictably when its alignment is changed
-- the extracted information content is therefore guaranteed to be the same for any (arbitrary) choice of alignment.

We formalize the alignment of a kernel at some point~$p$ of a manifold $M$ as a choice of \emph{local reference frame} -- or \emph{gauge} -- of the corresponding tangent space $\TpM$.
\emph{Gauge transformations} are therefore transformations between choices of reference frames.
Fig.~\ref{fig:intro_kernel_alignment_trivial} visualizes the concept of aligning kernels along reference frames.
Aligning the kernel relative to the canonical (uniquely preferred) \emph{frame field} of the Euclidean plane $\R^2$, as shown in the top, results in the usual \emph{kernel field} of Euclidean CNNs.
A different frame field, as shown in the bottom, implies an alternative kernel field and thus network.
As stated above, on most manifolds the choice of frames is inherently ambiguous such that no specific kernel alignment is preferred.
Fig.~\ref{fig:satellite} visualizes this issue for the sphere $S^2$, where frames are only unique up to rotations.

\begin{figure}
    \hspace*{.5ex}
    \begin{subfigure}[b]{0.105\textwidth}
        \centering
        \includegraphics[width=.66\textwidth]{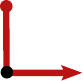}
        \\~\vspace*{-2pt}
        \caption{\small
            $G=\{e\}$
        }
        \label{fig:GpM_a}
    \end{subfigure}
    \hspace*{.5ex}
    \begin{subfigure}[b]{0.14\textwidth}
        \centering
        \includegraphics[width=1.\textwidth]{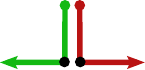}
        \\~\vspace*{-2pt}
        \caption{\small
            $G=\Flip$
        }
        \label{fig:GpM_b}
    \end{subfigure}
    \hspace*{.5ex}
    \begin{subfigure}[b]{0.14\textwidth}
        \centering
        \includegraphics[width=.8\textwidth]{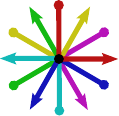}
        \caption{\small
            $G=\SO2$
        }
        \label{fig:GpM_c}
    \end{subfigure}
    \hspace*{.5ex}
    \begin{subfigure}[b]{0.14\textwidth}
        \centering
        \includegraphics[width=.66\textwidth]{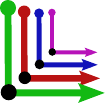}
        \\~\vspace*{-6pt}
        \caption{\small
            $G=\Scale$
        }
        \label{fig:GpM_d}
    \end{subfigure}
    \par
    \vspace*{\dimexpr-\parskip-66.pt\relax}
    \parshape 8 
        .6\textwidth .4\textwidth 
        .6\textwidth .4\textwidth
        .6\textwidth .4\textwidth
        .6\textwidth .4\textwidth
        .6\textwidth .4\textwidth
        .6\textwidth .4\textwidth
        .6\textwidth .4\textwidth
        .0\textwidth 1\textwidth 
    \makeatletter
    \setcounter{\@captype}{\value{\@captype}-1} 
    \refstepcounter{\@captype}
    \addcontentsline{\csname ext@\@captype\endcsname}{\@captype}
        {\protect\numberline{\csname the\@captype\endcsname}{ToC entry}}%
    \small 
    \csname fnum@\@captype\endcsname: 
    \makeatother
    \hyphenpenalty = 300 
        The choice of reference frames of a tangent space $\TpM$ is not always unique.
        The geometric structure ($G$-structure) of a manifold implies a preferred subset of reference frames such that gauge transformations between these frames lie in the structure group~$G\leq\GL{d}$.
        Figs.~\ref{fig:GpM_a}, \ref{fig:GpM_b}, \ref{fig:GpM_c} and~\ref{fig:GpM_d}
        show such subsets of frames for the trivial group $G=\{e\}$, reflection group $G=\Flip$, rotation group $G=\SO2$ and scaling group~$G=\Scale$, respectively.
        Features encode measurements relative to any of the distinguished frames.
        Their numerical coefficients relative to different frames are related by the action of some group representation $\rho$ of~$G$.
    \label{fig:GpM_examples}
\end{figure}

The level of ambiguity in the choice of reference frames depends on the \emph{geometric structure} of the manifold.
Such structure often allows to
\emph{disambiguate reference frames up to certain symmetry transformations} (gauge transformations); see Fig.~\ref{fig:GpM_examples}.
This statement is best explained with a few examples:
\begin{itemize}[leftmargin=1.2cm]
\item[{\rule[2.2pt]{2pt}{2pt}}]
    a naked \emph{smooth manifold} does not come with any preference in the choice of frames.
    Gauge transformations between general frames are arbitrary invertible linear maps, that is, they take values in the \emph{general linear group} ${G=\GL{d}}$.
\item[{\rule[2.2pt]{2pt}{2pt}}]
    an \emph{orientation} of the manifold allows to distinguish left-handed from right-handed frames.
    Gauge transformations between frames of either handedness are orientation preserving, i.e. they are elements of ${G=\operatorname{GL}^+(d)}$ (invertible linear maps with positive determinant).
\item[{\rule[2.2pt]{2pt}{2pt}}]
    a \emph{volume form} allows to distinguish \emph{unit volume frames}.
    Gauge transformations are then volume preserving, that is, they take values in the \emph{special linear group} $G=\operatorname{SL}(d)$.
\item[{\rule[2.2pt]{2pt}{2pt}}]
    the \emph{metric structure} of a Riemannian manifold allows to measure distances and angles in the tangent spaces and therefore allows to distinguish \emph{orthonormal frames}.
    Gauge transformations between orthonormal frames are rotations and reflections in the \emph{orthogonal group} $G=\O{d}$.
\item[{\rule[2.2pt]{2pt}{2pt}}]
    together, an \emph{orientation and metric} imply \emph{oriented orthonormal frames}.
    Gauge transformations are then only rotations in the \emph{special orthogonal group} $G=\SO{d}$.
\item[{\rule[2.2pt]{2pt}{2pt}}]
    a \emph{frame field} on the manifold consists of a \emph{unique frame} at every point of the manifold.
    Gauge transformations are in this case trivial, which is described by the \emph{trivial group} $G=\{e\}$.
\end{itemize}
All of these geometric structures have in common that they define a preferred subset (subbundle) of frames such that gauge transformations take values in some \emph{structure group}~$G\leq\GL{d}$.
To emphasize the central role of the structure group $G$, such structures are denoted as $G$-\emph{structures}~$\GM$.
Visual examples of $G$-structures for different structure groups $G$ and manifolds~$M$ are given in Fig.~\ref{fig:G_structures_intro}.

As the choice of reference frames is inherently ambiguous, any geometric quantity and network operation should be equally well representable relative to arbitrary frames of the $G$-structure~$\GM$, that is, they should be $\GM$-\emph{coordinate independent}.
Feature vectors are therefore associated with some \emph{group representation} (linear group action) $\rho$ of the structure group $G$, which determines their transformation law under gauge transformations ($G$-valued transitions between reference frames).
The particular choice of group representation determines the geometric type of a feature vector field.
Typical examples are scalar, vector or tensor fields, however, more general field types are used in practice as well.
Fig.~\ref{fig:gauge_trafos} visualizes the coordinate independence of geometric quantities at the well known example of tangent vectors.

Any network layer is required to respect the transformation laws of features, that is, it needs to guarantee that its outputs transform as expected.
Specifically for convolutions, $\GM$-coordinate independence demands that applying the shared kernel relative to different frames of the $G$-structure at some point $p\in M$ should evoke the \emph{same response up to a gauge transformation}.
We show that this requires the $G$-steerability (gauge equivariance, Eq.~\eqref{eq:G-steerable_kernel_space}) of convolution kernels.
Intuitively, one may think of $G$-steerable kernels as measuring features \emph{relative} to reference frames, which is necessary since no choice of frame, i.e. \emph{absolute} kernel alignment, is to be preferred.%
\footnote{
    Note the similarity to Einstein's \emph{principle of special relativity}, which relies on the equivalence of inertial frames instead of frames of the $G$-structure.
}
Examples of $G$-steerable kernels for the reflection group $G=\Flip$ are shown in Fig.~\ref{fig:intro_steerable_kernel}.
Fig.~\ref{fig:intro_kernel_alignment_reflect} visualizes the sharing of such kernels relative to different frames of some $\Flip$-structure.
The $\Flip$-steerability constraint enforces some symmetry on the kernels, such that the different alignments will indeed result in responses that differ exactly by gauge transformations $\rho(g)$.
We abbreviate $\GM$-coordinate independent convolutions in the following as $\GM$-\emph{convolutions}.

Besides applying \emph{gauge equivariant} kernels, $\GM$-convolutions may be \emph{isometry equivariant}, which means that they commute with the action of isometries on feature fields as illustrated in Fig.~\ref{fig:lizard_conv_egg_intro}.
Let $\phi \in \IsomM$ be some isometry (symmetry) of the manifold~$M$.
A neural network is exactly then equivariant w.r.t. the action of this isometry, if patterns at any point $p\in M$ are processed in the same way as patterns at~$\phi(p)$.
The isometry equivariance of a network is therefore in one-to-one correspondence with the \emph{isometry invariance of its neural connectivity} (kernel field); see Fig.~\ref{fig:isom_invariant_kernel_field_intro}.
Since our convolutions apply kernels relative to (arbitrary) frames of the $G$-structure $\GM$, the symmetries of the kernel field coincide with those of the $G$-structure.
Denoting the (distance preserving) \emph{symmetries of a $G$-structure} $\GM$ as $\IsomGM \leq \IsomM$, this implies that our convolutions are exactly $\IsomGM$-equivariant.
Fig.~\ref{fig:intro_invariant_kernel_fields_plane} visualizes the fact that $G$-structures and the corresponding kernel fields share the same symmetries.
The reader is encouraged to review the $G$-structures~in Fig.~\ref{fig:G_structures_intro} with regard to their symmetries and the implied equivariance properties of the corresponding $\GM$-convolutions.

The design of $\GM$-coordinate independent convolutional networks on Riemannian manifolds requires a choice of $G$-structure, which depends on multiple considerations.
Firstly, the choice of structure group~$G$ determines the \emph{local gauge equivariance} of the convolution:
a $G$-steerable kernel will automatically generalize learned patterns over all $G$-related poses of patterns; see Fig.~\ref{fig:intro_lizard}.
Secondly, the specific choice of $G$-structure determines the \emph{global isometry equivariance} of the convolution.
In medical imaging applications, patterns occur often in arbitrary rotations, reflections and positions
-- one should therefore choose an $\IsomGM = \E{d}$-invariant $\O{d}$-structure on $\R^d$, similar to the $\SO2$-structure that is shown in Fig.~\ref{fig:G_structure_intro_g}.
Images like portrait photos have a distinguished vertical axis, however, reflections over this axis leave the image statistics invariant
-- this calls for a $\Flip$-structure as in Fig.~\ref{fig:G_structure_intro_d}.
Besides such symmetry considerations, it is important to note that not any manifold (topology) admits \emph{smooth} $G$-structures for any choice of structure group~$G$.
An example is the M\"obius strip, whose twisted topology (non-orientability) prevents a smoothly varying assignment of frame orientations.
A \emph{smooth} coordinate independent convolution operation on the M\"obius strip thus \emph{necessarily} relies on reflection-steerable kernels.

This work includes an extensive \emph{literature review} on convolutional networks, which demonstrates the generality of our theory.
It covers different types of CNNs on Euclidean spaces, spherical CNNs and convolutions on general surfaces (e.g. surface meshes).
We identify the specific choices of $G$-structures that were implicitly made by the authors by analyzing the global and local equivariance properties of their models.
Table~\ref{tab:network_instantiations} gives an overview on the resulting taxonomy of $\GM$-coordinate independent convolutional networks.

To give a detailed example on how to instantiate our theory in practice, we discuss an implementation of $\GM$-convolutions on the M\"obius strip for~$G=\Flip$.
This includes a derivation of reflection-steerable kernels for different field types (group representations) and an empirical evaluation of the theoretically predicted isometry equivariance.
As expected, $\GM$-coordinate independent convolutions outperform a naive coordinate dependent implementation.
The code is available at \url{https://github.com/mauriceweiler/MobiusCNNs}.

A \emph{coordinate free} formulation of our theory is devised in the language of \emph{fiber bundles}.
$G$-structures $\GM$ are principal $G$-subbundles of the frame bundle $\FM$ over~$M$.
Feature fields are \emph{sections} of $G$-\emph{associated feature vector bundles}.
Gauges are \emph{local bundle trivialization}, while gauge transformations are transition maps between such trivializations.
The isometries w.r.t. which a $\GM$-convolution is equivariant are \emph{principal bundle automorphisms} of the $G$-structure.

Our coordinate independent CNNs are generalizations of \emph{steerable CNNs}~\cite{Cohen2017-STEER,3d_steerableCNNs,Weiler2019_E2CNN,Cohen2019-generaltheory,lang2020WignerEckart} from Euclidean (or homogeneous) spaces to Riemannian manifolds.
While steerable CNNs focus on \emph{active, global} transformations of feature fields, coordinate independent CNNs consider \emph{passive, local} transformations between reference frames.%
\footnote{
    This resembles the shift of focus from \emph{global Lorentz covariance} in \emph{special relativity} to \emph{local Lorentz covariance} in \emph{general relativity}.
}
We proposed early versions of the theory of coordinate independent CNNs (``gauge equivariant CNNs'') in previous work~\cite{gaugeIco2019,deHaan2020meshCNNs}.
In contrast to these publications, the present work develops the theory in much greater detail, formulates it in terms of fiber bundles, proves the equivariance under the action of isometries and provides a literature review.

%% file: chapters/11_outline.tex

\setcounter{tocdepth}{2}
\tableofcontents

~ 

This work is organized into an introduction, three main parts and an appendix.

Part~\ref{part:local_theory} attempts to introduce coordinate independent neural networks in an easily accessible language.
Feature fields and network layers are expressed relative to \emph{local coordinates} (bundle trivializations).
The demanded \emph{coordinate independence} requires features to be associated with some \emph{transformation law}.
Network layers are required to guarantee the correct transformation behavior of features.

Part~\ref{part:bundle_theory} formalizes the theory of coordinate independent neural networks in terms of \emph{fiber bundles}.
This allows for a \emph{global, coordinate free} formulation, which is particularly useful when investigating the networks' isometry equivariance.
The definitions from Part~\ref{part:local_theory} are recovered when expressing the coordinate free operations in local bundle trivializations (coordinates).

Part~\ref{part:literature_review} embeds our theory in \emph{related work}.
It provides detailed reviews of convolutional network architectures on various geometries and reformulates them as coordinate independent CNNs.
To facilitate the development of new network architectures, we discuss relevant characteristics of the specific geometries before reviewing the networks that operate on them.

The reader may skip Part~\ref{part:bundle_theory} at a first pass
-- the formulation from Part~\ref{part:local_theory} is fully sufficient to read the literature review in Part~\ref{part:literature_review}.

An overview of the main concepts and results of our work is provided in the following Section~\ref{sec:visual_intro}.
This overview avoids equations and builds on geometric intuition in terms of visualizations.
We hope that this Sections allows a non-technical audience to get an idea about the content of our work.

\subsubsection*{Detailed Overview}

\paragraph{Part~\ref{part:local_theory}:}

The goal of Section~\ref{sec:gauge_cnns_intro_local} is to devise coordinate independent feature spaces.
Specifically, Section~\ref{sec:21_main} introduces \emph{gauges, gauge transformations and $G$-structures}.
Gauges are a formal way to express (coordinate free) tangent vectors and functions on the tangent spaces relative to reference frames.
Gauge transformations translate between such coordinate expressions in different gauges.
Section~\ref{sec:feature_fields} introduces coordinate independent \emph{feature vector fields}.
As in the case of tangent vectors, the numerical coefficients of feature vectors transform when transitioning between reference frames.
The transformation laws of feature vectors determine in particular their \emph{parallel transport} and their pushforward when being acted on by \emph{isometries}, which are described in Sections~\ref{sec:transport_local} and~\ref{sec:isometries_local}, respectively.

Section~\ref{sec:gauge_CNNs_local} develops \emph{neural networks} that map between feature fields.
\emph{Pointwise operations}, like bias summation, \onexones\ and nonlinearities, are discussed in Section~\ref{sec:pointwise_operations}.
Section~\ref{sec:gauge_conv_main} focuses on \emph{convolutions} with spatially extended kernels.
Each of these operations is initially introduced without the weight sharing assumption, that is, allowing for instance for a different kernel at each point of the manifold.
These kernels (or biases or nonlinearities) are not constrained in any way.
However, when requiring spatial weight sharing, they become constrained to be gauge equivariant since only equivariant quantities can be shared in a coordinate independent manner.
Section~\ref{sec:gauge_conv_isom_equiv} gives a concise proof of the isometry equivariance of $\GM$-convolutions in terms of local coordinate expressions.
The key idea here is that isometries can be viewed as inducing gauge transformations (passive interpretation), which are explained away by the kernels' gauge equivariance.

Section~\ref{sec:mobius_conv} describes an implementation of \emph{orientation independent convolutions on the M\"obius strip}.
After reviewing the geometry of the M\"obius strip in Section~\ref{sec:mobius_geometry}, multiple types of feature fields are defined in Section~\ref{sec:mobius_representations}.
The following Section~\ref{sec:mobius_cnn_ops_analytical} describes orientation independent CNNs analytically.
In particular, we derive gauge equivariant convolution kernels, biases and nonlinearities for each of the field types.
Section~\ref{sec:mobius_experiment_main} closes with a numerical implementation and evaluation of the corresponding models.

\paragraph{Part~\ref{part:bundle_theory}:}

Section~\ref{sec:bundles_fields} mirrors the content of Section~\ref{sec:gauge_cnns_intro_local}, however, globally and in terms of \emph{fiber bundles}.
A~general introduction to fiber bundles is given in Section~\ref{sec:fiber_bundles_general}.
Sections~\ref{sec:GL_associated_bundles} and~\ref{sec:G_associated_bundles} introduce the tangent bundle $\TM$, the frame bundle $\FM$, $G$-structures $\GM$ and $G$-associated feature vector bundles~$\A$.
Feature fields are globally defined as sections of feature vector bundles.
\emph{Local bundle trivializations} (gauges), which are discussed in Section~\ref{sec:bundle_trivializations}, express these bundles in coordinates, thereby recovering our definitions from Section~\ref{sec:gauge_cnns_intro_local}.
We demonstrate in particular how local trivializations of the different bundles induce each other, such that their gauge transformations (transition maps) are synchronized.
Section~\ref{sec:bundle_transport} discusses \emph{parallel transporters} on $G$-bundles.

Section~\ref{sec:gauge_CNNs_global} reformulates the coordinate independent networks from Section~\ref{sec:gauge_CNNs_local} in terms of fiber bundles.
\onexonesit\ are in Section~\ref{sec:onexone} described as specific vector bundle $M$-morphisms.
Alternatively, they may be viewed as sections of a homomorphism bundle.
Section~\ref{sec:global_conv} introduces \emph{coordinate free kernel fields} and \emph{kernel field transforms}.
These operations are similar to $\GM$-convolutions but are not required to share weights, i.e. may apply a different kernel at each spatial location.
A $\GM$-\emph{convolutional kernel field} is constructed by sharing a single $G$-steerable (gauge equivariant) kernel over the whole manifold.
\emph{Coordinate free $\GM$-convolutions} are then defined as kernel field transforms with $\GM$-convolutional kernel fields.
When expressing the coordinate free formulation of $\GM$-convolutions relative to local trivializations (gauges), we recover the coordinate expressions of $\GM$-convolutions from Section~\ref{sec:gauge_conv_main}.

The \emph{isometry equivariance} of $\GM$-convolutions is investigated in Section~\ref{sec:isometry_intro}.
After introducing isometries, Section~\ref{sec:isom_background} discusses their \emph{pushforward action} on the fiber bundles.
These action may again be expressed in local trivializations, resulting in the formulation from Section~\ref{sec:isometries_local}.
Section~\ref{sec:isometry_equivariance} defines the action of isometries on kernel fields and proves that \emph{the isometry equivariance of a kernel field transform implies the isometry invariance of its kernel field and vice versa}.
$\GM$-convolutions are proven to be equivariant under the action of those isometries which are bundle automorphisms (symmetries) of the $G$-structure~$\GM$.
Section~\ref{sec:quotient_kernel_fields} investigates isometry invariant kernel fields in greater detail and proves that they are equivalent to \emph{kernel fields on quotient spaces} of the isometry action
-- intuitively speaking, isometry invariant kernel fields are required to share kernels over the isometry orbits.
This result implies in particular that isometry equivariant kernel field transforms on \emph{homogeneous spaces} are necessarily $\GM$-convolutions.

\paragraph{Part~\ref{part:literature_review}:}

The third part of this work demonstrates that a vast number of convolutional networks from the literature can be interpreted as applying $\GM$-convolutions for some choice of $G$-structure and field types.
It starts with a general discussion about the design choices of coordinate independent CNNs.
Table~\ref{tab:network_instantiations} gives an overview and classification of the models that are reviewed.
The reader is invited to have a look at the $G$-structures that are visualized in Part~\ref{part:literature_review} as these give an intuitive idea about the properties of the corresponding $\GM$-convolutions.

\emph{Euclidean CNNs} that are not only isometry equivariant but more generally equivariant under the action of \emph{affine groups} are reviewed in Section~\ref{sec:instantiations_euclidean}.
These models are essentially equivalent to \emph{steerable CNNs} on Euclidean \emph{vector} spaces $\R^d$ \cite{Cohen2017-STEER,3d_steerableCNNs,Weiler2019_E2CNN}.
Section~\ref{sec:steerable_cnns_in_coords} reviews steerable CNNs and discusses their relation to $\GM$-convolutions.
This approach is somewhat unsatisfactory since $\R^d$ comes with a canonical frame field ($\{e\}$-structure), which is implicitly ignored by equivariant models.
Section~\ref{sec:euclidean_geometry} takes a more principled approach, defining Euclidean \emph{affine} spaces $\Euc_d$ which are equipped with precisely those $G$-structures that result in $\Aff(G)$-equivariant $\GM$-convolutions.
The actual $\GM$-convolutions are defined in Section~\ref{sec:euclidean_affine_equiv}.
Section~\ref{sec:euclidean_literature} reviews affine equivariant Euclidean CNNs found in the literature.
They differ mainly in the assumed choices of structure groups and group representations.

Section~\ref{sec:instantiations_euclidean_polar} covers CNNs on \emph{punctured Euclidean spaces} $\Euc_d\backslash\{0\}$, whose origin $\{0\}$ was removed.
These models are rotation equivariant around the origin, however, they are not translation equivariant.
They are based on $G$-structures that correspond to polar coordinates, log-polar coordinates or spherical coordinates.

\emph{Spherical CNNs} are covered in Section~\ref{sec:instantiations_spherical}.
Section~\ref{sec:sphere_geometry} discusses the geometry of the (embedded) 2-sphere~$S^2$.
Interpreting the tangent spaces as two-dimensional subspaces of an embedding space~$\R^3$, we derive closed form expressions of exponential and logarithm maps, frames, gauges, transporters and isometry actions.
Section~\ref{sec:spherical_CNNs_fully_equivariant} reviews $\SO3$ and $\O3$-equivariant spherical CNNs.
We prove in particular that our theory includes the general formulation of spherical convolutions by~\citet{Cohen2019-generaltheory} as a special case.
Spherical CNNs that are merely $\SO2$ rotation equivariant around a fixed axis are described in Section~\ref{sec:spherical_CNNs_azimuthal_equivariant}.
Section~\ref{sec:spherical_CNNs_icosahedral} reviews icosahedral CNNs.
The icosahedron approximates the sphere but consists of locally flat faces which allow for an efficient implementation of convolution operations.

A survey of convolutional networks on \emph{general two-dimensional surfaces} is found in Section~\ref{sec:instantiations_mesh}.
Section~\ref{sec:surfaces_geom_main} provides a brief introduction to the classical differential geometry of embedded surfaces and their discretization in terms of triangle meshes.
The surface convolutions in the literature are categorized in two classes:
The first class, covered in Section~\ref{sec:so2_surface_conv}, is based on $G=\SO2$-steerable kernels.
These models are independent from the specific choice of right-handed, orthonormal frame.
Section~\ref{sec:e_surface_conv} reviews the second category of models, which are based on $\{e\}$-steerable, i.e. non-equivariant kernels.
These models rely explicitly on a choice of frame field.
They differ therefore mainly in the heuristics that are used to determine reference frames.
Note that such models are necessarily discontinuous on non-parallelizable manifolds like for instance topological spheres.

\paragraph{Appendix:}

The appendix covers some additional information and long proofs.

Gauges formalize an immediate assignment of reference frames to tangent spaces but refer to points on the manifold in a coordinate free manner.
A popular alternative is to choose \emph{coordinate charts}, which induce so called \emph{coordinate bases} (holonomic bases) of the tangent spaces.
Appendix~\ref{apx:coordinate_bases} gives an introduction to the chart formalism and puts it in relation to the more general gauge formalism.

Appendix~\ref{apx:coord_indep_weight_sharing} comments on the \emph{coordinate independence} of kernels and \emph{weight sharing} along reference frames.
A $\GM$-coordinate independent sharing of weights is only possible for $G$-steerable kernels.

$\GM$-convolutions are computed by expressing feature fields in geodesic normal coordinates, where they are matched with $G$-steerable convolution kernels.
This process involves an \emph{integration over the tangent spaces} which is described in Appendix~\ref{apx:tangent_integral}.

\citet{Kondor2018-GENERAL}, \citet{Cohen2019-generaltheory} and \citet{bekkers2020bspline} proposed quite general theories of \emph{convolutions on homogeneous spaces}.
As these models share weights via the action of some symmetry group, they are very similar to our isometry equivariant kernel field transforms from Sections~\ref{sec:isometry_equivariance} and~\ref{sec:quotient_kernel_fields}.
Appendix~\ref{apx:homogeneous_conv} reviews these models and explains how they relate to our $\GM$-convolutions.

Appendix~\ref{apx:lifting_iso_proof} proves that isometry invariant kernel fields on the manifold are equivalent to kernel fields on quotient spaces of the isometry action.
The special case of homogeneous spaces, on which isometry equivariant kernel field transforms are equivalent to $\GM$-convolutions, is covered in Appendix~\ref{apx:homogeneous_equivalence_proof}.

The spherical convolutions of \citet{Cohen2019-generaltheory} are in Appendix~\ref{apx:spherical_conv_main} proven to be a special case of our spherical $\GM$-convolutions
-- any spherical CNN that is covered by their theory is therefore explained by our theory as well.

Appendix~\ref{apx:smoothness_kernel_field_trafo} asserts that our kernel field transforms and $\GM$-convolutions are well defined if the kernel field is smooth and consists of compactly supported kernels.
Well-definedness means here that the integrals exist and that the resulting feature fields are smooth.

Finally, Appendix~\ref{apx:regular_field_scalar_GM} argues that feature fields that transform according to the regular representation of the structure group $G$~are equivalent to scalar fields on the $G$-structure.
This is relevant since some models, specifically group convolutions, take this viewpoint.

%% file: chapters/20_visual_intro.tex

\newpage

\begin{figure}
    \centering
    \includegraphics[width=.93\columnwidth]{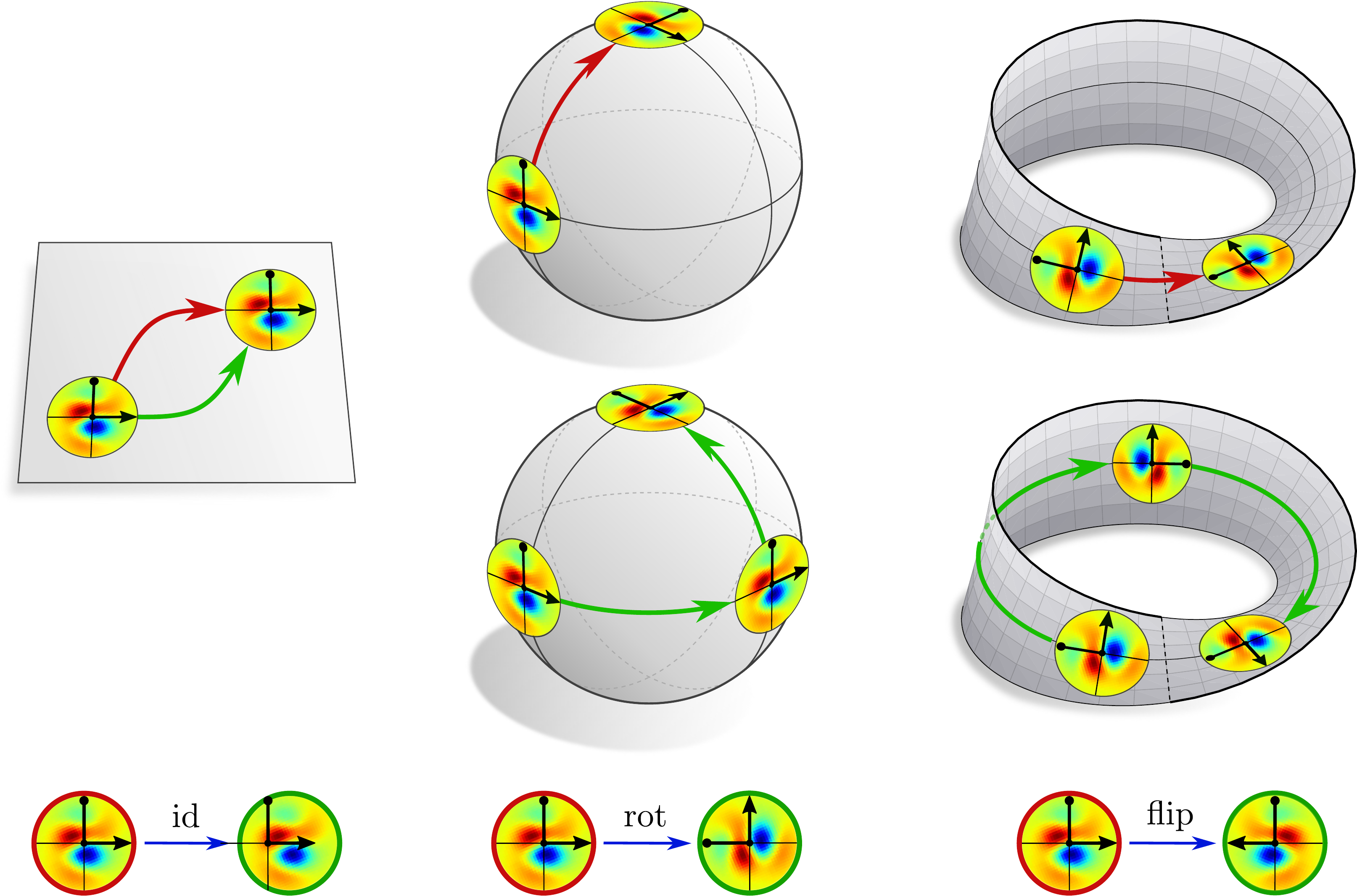}
    \caption{\small
        An intuition on the inherent ambiguity of weight sharing on manifolds.
        \ \ \emph{Left:}
        A common interpretation of weight sharing on the plane is to shift a kernel over the whole space.
        Since parallel transport is path independent on flat spaces, this is unambiguous.
        \ \ \emph{Middle:}
        On curved spaces, like the sphere, parallel transport is path dependent.
        Different paths result in kernels that are rotated relative to each other.
        \ \ \emph{Right:}
        The M\"obius strip is a non-orientable manifold.
        Different paths can therefore result in kernels that are reflected relative to each other.
        \ \ \emph{Bottom:}
        We formalize different kernel alignments by different choices of local reference frames of the corresponding tangent spaces.
        It is well known that no choice of reference frames (gauge) is preferred on general manifolds.
        Different coordinatizations are related by gauge transformations, which take values in the structure group $G$ of the manifold (the trivial group $G=\{e\}$ for the plane, rotation group $G=\SO2$ for the sphere and reflection group $G = \Flip$ for the M\"obius strip).
        Coordinate independent CNNs address the ambiguity of reference frames by applying $G$-steerable (gauge equivariant) convolution kernels.
    }
    \label{fig:weight_sharing_ambiguity}
\end{figure}

\section{Overview and visual intuition}
\label{sec:visual_intro}

The algebraic formulation of coordinate independent CNNs requires some familiarity with group theory, representation theory and differential geometry which might pose an obstacle for a non-technical audience.
However, most of our constructions and results are geometrically very intuitive and can be explained with a few visual examples.
This section attempts to give an overview and visual intuition about coordinate independent CNNs.

The following Section~\ref{sec:intro_overview_GM_conv} introduces $\GM$-coordinate independent convolutions on Riemannian manifolds.
Their equivariance under the action of isometries is discussed in Section~\ref{sec:intro_overview_isometry}.
Section~\ref{sec:intro_overview_G_structure_choice} comments on the factors that influence the choice of $G$-structure in the design of coordinate independent CNNs.

\toclesslab\subsection{\textit{G}-structures and \textit{GM}-coordinate independent convolutions}{sec:intro_overview_GM_conv}
Since convolutions are essentially characterized by their weight sharing property, a central question in this work is:
\vspace*{-1ex}
\begin{center}\it
    How should convolution kernels be shared over Riemannian manifolds?
    \footnote{
        This question applies more generally to any local template function, for instance biases or pointwise nonlinearities.
    }
\end{center}

\marginnote{} 

A common approach \marginnote{weight sharing via symmetries} is to share weights via the action of a symmetry group of the underlying space~\cite{Cohen2016-GCNN,Kondor2018-GENERAL}.
For instance, conventional CNNs share weights by translating kernels over the plane, while spherical CNNs share weights by rotating kernels over the sphere.
In order to share a kernel over the whole space, the action of the symmetry group needs to be \emph{transitive}.
As this is in general not the case for the isometries of Riemannian manifolds, this strategy is ruled out for our purpose.

The weight sharing on Euclidean spaces is often thought of as ``shifting'' a kernel over the space.
\marginnote{weight sharing via transport}
Since on flat spaces parallel transport is independent from the chosen path, this leads to an unambiguous alignment of kernels; see Fig.~\ref{fig:weight_sharing_ambiguity}~(left).
However, on curved or non-orientable spaces parallel transport becomes \emph{path dependent} and thus unsuitable for sharing weights.
Fig.~\ref{fig:weight_sharing_ambiguity} (middle and right) exemplifies this issue for the sphere and the M\"obius strip, where different paths lead to a different kernel alignment.

As the concept of ``kernel alignments'' is somewhat vague
\marginnote{weight sharing along frames}
we first need to make it mathematically precise:

\begin{minipage}{\textwidth}
\begin{center}\it
    We formalize the choice of kernel alignment at a point $p\in M$ \\
    as a choice of local reference frame (gauge)
    of the corresponding tangent space $\TpM$.
\end{center}
\end{minipage}

\begin{SCfigure}
    \centering
    \includegraphics[width=.62\textwidth]{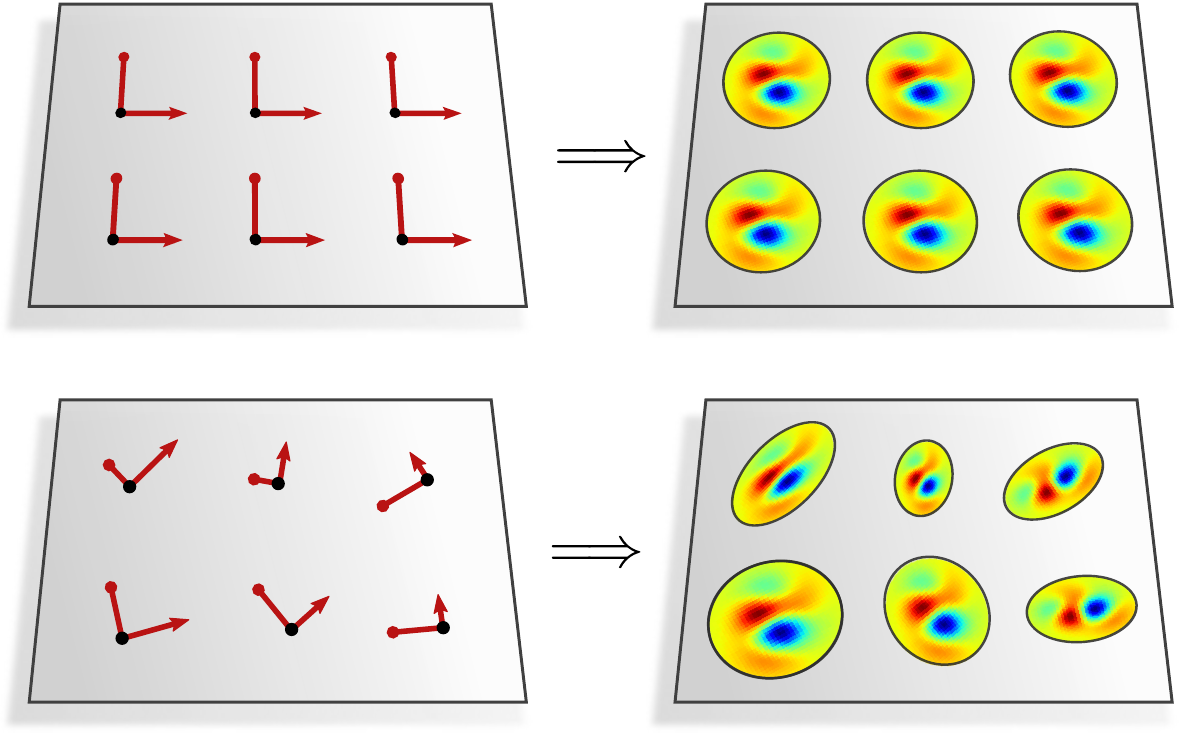}
    \captionsetup{width=.9\textwidth}
    \caption{\small
        A key property of convolutions is that they \emph{share weights} over the manifold.
        We identify the alignment of a kernel at $p\in M$ with a \emph{choice of reference frame} -- or \emph{gauge} -- of the corresponding tangent space~$\TpM$.
        Different \emph{frame fields} imply therefore different (convolutional) \emph{kernel fields}.
        \\[1ex]
        The choice of frames is often not unique.
        The ambiguity in this choice is formalized by $G$-\emph{structures}; see Fig.~\ref{fig:G_structures_intro}.
        To account for the arbitrariness of frames, the kernels are then required to be $G$-steerable (equivariant) as visualized in Figs.~\ref{fig:intro_steerable_kernel} and~\ref{fig:intro_kernel_alignment_reflect}.
        \\[0pt]
        }
    \label{fig:intro_kernel_alignment_trivial}
\end{SCfigure}

A reference frame at $p\in M$ is an ordered tuple $[e_1,\, \dots,\, e_d]$ of $d := \dim(M)$ linearly independent tangent vectors $e_i \in \TpM$, denoted as frame axes.
Since different frames at~$p$ are related by linear transformations, different choices of frames correspond to linear kernel deformations.
Fig.~\ref{fig:intro_kernel_alignment_trivial} shows two different choices of frame fields on~$M = \R^2$.
Sharing some convolution kernel along these frame fields results in the corresponding (convolutional) \emph{kernel fields}.

The identification of kernel alignments with reference frames raises the question:
\begin{center}\it
    To which extent is the choice of local reference frames on a (Riemannian) manifold ambiguous?
\end{center}
As elaborated in the following, the ambiguity of reference frames is determined by a $G$-structure with which the manifold is endowed.

\subsubsection{\textit{G}-structures}
\label{sec:visual_intro_GM_subsub}

The space of \emph{all} possible frames of $\TpM$ is denoted as~$\FpM$.
\marginnote{frame bundle $\FM$}
Taken together, the frames of all tangent spaces form the \emph{frame bundle}~$\FM$; see Fig.~\ref{fig:frame_bundle}.
No specific choice of frames in $\FM$ is on a ``naked'' smooth manifold (without further geometric structure) preferred over each other, leaving the alignment of kernels maximally ambiguous.
In order to disambiguate frames and kernel alignments, the manifold needs to be equipped with \emph{additional geometric structure}.

A Riemannian manifold is equipped with a \emph{metric structure}.
\marginnote{metric structure $\OM$}
Providing an inner product (Riemannian metric) on the tangent spaces, this structure allows to single out those specific frames whose axes are \emph{orthonormal} to each other.
\emph{Gauge transformations}, i.e. transformations between choices of reference frames (see Figs.~\ref{fig:intro_gauge_isom_induction} (left) and~\ref{fig:gauge_trafos}), take then values in the \emph{orthogonal group}~$\O{d}$.
On Riemannian manifolds the alignment of kernels is therefore always disambiguated up to rotations and reflections.

Euclidean CNNs align convolution kernels (and thus frames)
\marginnote{$G$-structures $\GM$}
parallel to each other as visualized in Fig.~\ref{fig:intro_kernel_alignment_trivial} (top).
Spherical CNNs disambiguate kernel alignments usually up to rotations, that is, they assume a preferred handedness of reference frames.
The metric structure alone is insufficient to describe these settings, which suggests that these manifolds are equipped with \emph{further geometric structure besides the metric structure}.
We propose that the suitable mathematical framework is that of $G$-\emph{structures} and postulate:

\begin{minipage}{\textwidth}
\begin{center}\it
    The ambiguity in choosing reference frames (and thus kernel alignments) \\
    on a manifold $M$ is formalized by its $G$-structure $\GM$.
\end{center}
\end{minipage}

$G$-structures $\GM$ are \emph{bundles of distinguished reference frames} on $M$ such that the \emph{gauge transformations} between frames of the same tangent space take values in the \emph{structure group}~${G\leq\GL{d}}$.
Intuitively, one may think about the set $\GpM$ of frames of $\TpM$ as ``looking like'' $G$, however, without a distinguished origin.%
\footnote{
    $\GpM$ is a \emph{principal homogeneous space} of~$G$ (or $G$-torsor).
}

The frame bundle $\FM$ is itself a $G$-structure with~${G=\GL{d}}$, while the bundle of orthonormal frames $\OM$ is a $G$-structure (metric structure) with~${G=\O{d}}$.
Conventional Euclidean CNNs rely on the canonical frame field on $\R^d$ shown in Fig.~\ref{fig:G_structure_intro_a}, which is a $G$-structure for the trivial group~${G=\{e\}}$.
Fig.~\ref{fig:G_structures_intro} visualizes $G$-structures for further manifolds and structure groups.
An overview of common structure groups is found in Table~\ref{tab:G_structures} in Section~\ref{sec:local_G-structure_G-atlas}.

We will in the following always assume the Riemannian manifolds to be equipped with an additional $G$-structure next to its metric structure.%
\footnote{
    The $G$-structure needs to be compatible with the metric structure in that the distinguished frames in $\GM$ are a subset of the orthonormal frames in $\OM$ when $G<\O{d}$.
    Specifically for $G=\O{d}$, the $G$-structure $\GM$ coincides with the metric structure $\OM$ and adds no additional geometric information.
}
The particular choice of $G$-structure determines the properties of the neural network; we will comment on this choice in Section~\ref{sec:intro_overview_G_structure_choice} below.

\afterpage{ 
\clearpage 
\begin{figure}
    \centering
    \vspace*{-4.ex}
    \input{chapters/intro_fig_G-structures.tex}
    \vspace*{1.5ex}
    \captionsetup{width=1.1\textwidth}
    \caption{\small
        Exemplary $G$-structures $\GM$ for different structure groups $G$ and on different manifolds $M$.
        The structure group $G$ signals which values gauge transformations can take, and therefore how ``big'' the subset of distinguished frames at each point~$p$ is.
        Fig.~\ref{fig:G_structure_intro_a} shows the canonical $\{e\}$-structure (frame field) of~$\R^2$, which corresponds to conventional Euclidean CNNs.
        The \mbox{$G$-structures} in
        Figs.~\ref{fig:G_structure_intro_d}, \ref{fig:G_structure_intro_g} and~\ref{fig:G_structure_intro_j}
        are constructed by adding reflected ($G=\Flip$), rotated ($G=\SO2$) and scaled ($G=\Scale$) frames, respectively.
        The corresponding $\GM$-convolutions are not only translation equivariant but equivariant under the action of affine groups~$\Aff(G)$.
        \mbox{$G$-structures} are usually not unique.
        Figs.~\ref{fig:G_structure_intro_b} and~\ref{fig:G_structure_intro_e} show alternative $G$-structures on~$\R^2$ (corresponding to an alternative metric relative to which their frames are orthonormal).
        They might not be practically relevant but demonstrate the flexibility of our framework.
        The $\{e\}$-structure in Fig.~\ref{fig:G_structure_intro_c} corresponds to polar coordinates.
        As $G$-structures are required to be continuous, we removed the origin~$0$ where polar coordinates are singular.
        One can once again define an $\Flip$-structure by adding reflected frames as shown in Fig.~\ref{fig:G_structure_intro_f}.
        These $G$-structures model convolutions on $\R^2\backslash\{0\}$ which are $\SO2$ and $\O2$-equivariant but not translation equivariant.
        Fig.~\ref{fig:G_structure_intro_h} shows the usual $\SO2$-structure on the embedded 2-sphere~$S^2$, which is underlying $\SO3$-equivariant spherical CNNs.
        Another popular choice is the $\{e\}$-structure in Fig.~\ref{fig:G_structure_intro_k}, which is induced by spherical coordinates.
        Note that this $\{e\}$-structure would be singular at the poles, which are therefore cut out.
        Continuous (i.e. non-singular) reductions of the structure group beyond~$\SO2$ are on the sphere topologically obstructed.
        $G$-steerable kernels with $G\geq\SO2$ are therefore strictly necessary for continuous convolutions on topological spheres like the mesh in Fig.~\ref{fig:G_structure_intro_i}.
        Fig.~\ref{fig:G_structure_intro_l} shows an $\Flip$-structure on the M\"obius strip.
        As the M\"obius strip is non-orientable, it does not admit a continuous reduction of the structure group beyond the reflection group~$G=\Flip$.
    }
    \label{fig:G_structures_intro}
\end{figure}
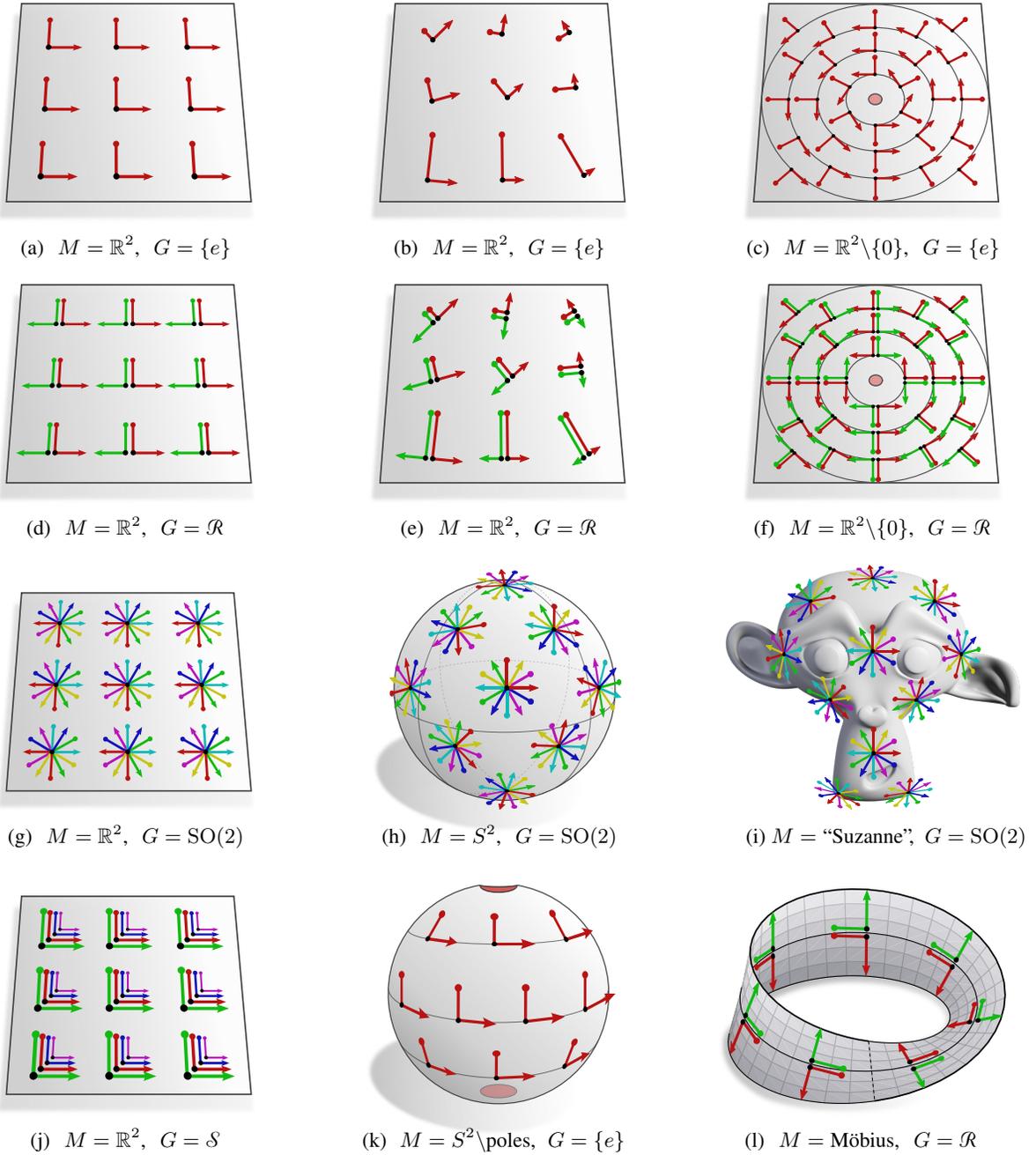
\thispagestyle{empty}
\clearpage 
}

\subsubsection{\textit{GM}-coordinate independent networks}
Our goal is
\marginnote{$\GM$-coordinate independence (covariance)}
to design neural networks on Riemannian manifolds with an additional $G$-structure.
If the structure group $G$ is non-trivial, \emph{no canonical choice of reference frames (gauge) exists} by definition.
However, in order to perform numerical computations, \emph{some} gauge, relative to which kernels and features are expressed, needs to be chosen.
Since this choice is inherently arbitrary we demand that the networks' inference should ultimately not depend on it, that is, we require:
\begin{center}\it
    Neural networks on a Riemannian manifold with $G$-structure $\GM$ \\
    should be based on ``$\GM$-coordinate independent'' operations.
\end{center}
``$\GM$-coordinate independence'' means hereby that \emph{all geometric quantities and functions between them should be equally well expressible in any gauge}, i.e. relative to any choice of reference frames of the $G$-structure.
The specific case of tangent vector coefficients and gauge transformations between them is visualized in Fig.~\ref{fig:gauge_trafos}.
Fig.~\ref{sec:gauges_TpM_functions} shows an example of a linear map and its coordinate independent representation in terms of matrices relative to different frames.

Note that the requirement for $\GM$-coordinate independence is quite flexible:
for $G=\GL{d}$ one has $\GM=\FM$ and therefore the maximal level of coordinate independence.
At the other end of the spectrum of structure groups one has $G=\{e\}$, for which $\GM$ is a fixed frame field and $\GM$-coordinate independence reduces to an explicit coordinate dependence.
The freedom of choosing arbitrary $G$-structures allows for a precise control over the networks' coordinate independence,
which varies in practice widely; see for instance Table~\ref{tab:network_instantiations} in Part~\ref{part:literature_review}.

Our networks process \emph{feature vector fields} on the manifold.
\marginnote{$G$-associated feature vector fields}
Feature vectors are \emph{coordinate free} geometric quantities like e.g. tangent vectors.
Relative to a chosen frame (gauge) they may be represented by \emph{numerical coefficient vectors}.
The demand for $\GM$-coordinate independence requires the numerical coefficients in different gauges to encode the same information content.
This is naturally achieved by associating the features with a group representation~$\rho$ of the structure group~$G$ which determines their coefficients' transformation behavior under gauge transformations:
\begin{center}\it
    Feature vector fields are associated with a $G$-representation $\rho$ which specifies \\
    the transformation law of their numerical coefficients when transforming between reference frames.
    \\[1ex]
    Technically speaking, feature fields are sections of $G$-associated feature vector bundles.
\end{center}
Typical examples are scalar fields, tangent vector fields or other tensor fields, however, any transformation law is admissible and can be chosen by the user.
Other common choices for~$\rho$ in the deep learning context are irreducible or regular representations;
more examples are listed in Table~\ref{tab:network_instantiations} in Part~\ref{part:literature_review}.

We want to emphasize that the requirement for $\GM$-coordinate independence is a mere \emph{consistency condition}, guaranteeing that different observers (frames) agree on the same coordinate free geometric observation.
It~does \emph{not} constrain the set of admissible functions in any way, but only how their expressions relative to different frames relate.
In particular, \emph{$\GM$-coordinate independent neural networks are in general not required to be gauge equivariant}.
A requirement for gauge equivariance follows when simultaneously demanding weight sharing and coordinate independence, that is, $\GM$-coordinate independent convolutions rely on gauge equivariant kernels.

\begin{SCfigure}[2.5]
    \centering
    \small
    $\begin{array}{c@{\hspace{2pt}}c}
        \rotatebox{-90}{
            \includegraphicstotab[width=.12\textwidth]{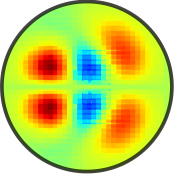}
        } &
        \reflectbox{\rotatebox{-90}{
            \includegraphicstotab[width=.12\textwidth]{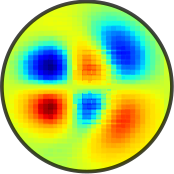}
        }} \\[60pt]
        \textup{scalar} & \textup{scalar} \\
        \updownarrow & \updownarrow \\
        \textup{scalar} & \textup{pseudoscalar}
        \\[-2pt] ~
    \end{array}$
    \hspace*{-2.5ex}
    \caption[]{\small
        Exemplary $G$-steerable kernels for the group $G=\Flip$ of reflections along the first frame axis (parity transformations).
        The kernels' equivariance constraints depend on the field types $\rhoin$ and $\rhoout$ of the convolution input and output.
        Scalar and pseudoscalar fields are described by the trivial representation and the sign-flip representation of $\Flip$, respectively.
        Steerable kernels which map between such fields are constrained to be symmetric or antisymmetric.
        \\[1ex]
        Note that steerable kernels are in general not scalar-valued but ${\cout \!\times\! \cin}$-matrix valued, where $\cout$ and $\cin$ are the dimensionalities of the output and input feature vectors, respectively.
        A derivation and more examples for other (higher-dimensional) field types of $G=\Flip$ are found in Section~\ref{sec:mobius_kernel_spaces} and Table~\ref{tab:reflection_steerable_kernels}.
        Steerable kernels for $G=\SO2$ or $G=\SO3$ are constructed from circular \cite{Weiler2019_E2CNN,Worrall2017-HNET,Weiler2018SFCNN} or spherical \cite{3d_steerableCNNs,Thomas2018-TFN} harmonics.
        If~$G$ is compact, $G$-steerable kernels are described by a generalization of the Wigner-Eckart theorem~\cite{lang2020WignerEckart}.
        \\[-20pt]
    }
    \label{fig:intro_steerable_kernel}
\end{SCfigure}

\subsubsection{\textit{GM}-coordinate independent convolutions}
\begin{figure}
    \centering
    \includegraphics[width=1.\textwidth]{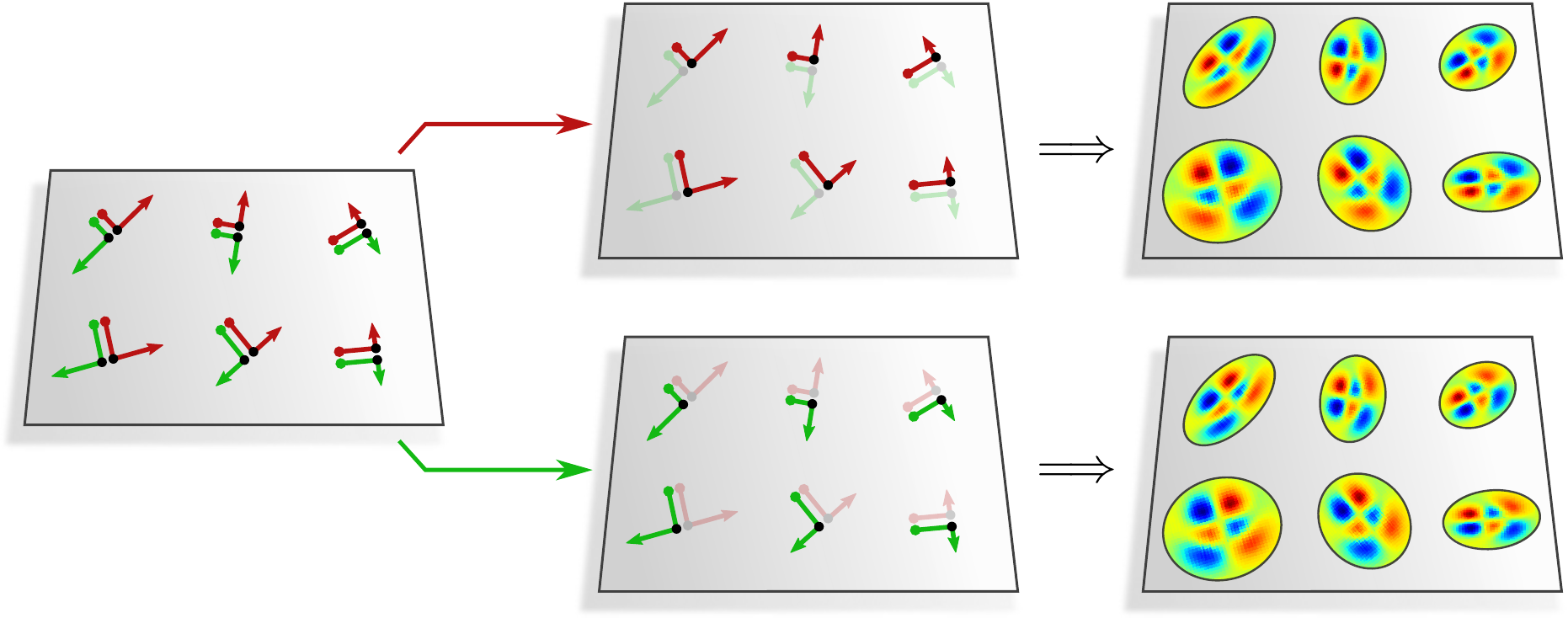}
    \caption{\small
        Sharing an $\Flip$-steerable kernel according to a given $\Flip$-structure $\RM$ over a manifold~$M = \R^2$.
        There are two continuous gauges (red and green) along which the kernel could be shared.
        Due to its $\Flip$-equivariance, the particular choice is ultimately irrelevant.
        The visualized kernel is antisymmetric and maps therefore between scalar and pseudoscalar fields.
        It is easily verified that this is indeed the case:
        the numerical coefficients of a scalar input field stay invariant under gauge transformations but the kernels are reflected.
        As they are antisymmetric, their responses will negate -- which is the transformation law of the numerical coefficients of a pseudoscalar field.
        A similar reasoning holds for mappings from pseudoscalars to scalars.
        How could we have mapped from scalars to scalars?
        In this case both the input and the output should be gauge invariant, requiring the kernel to be symmetric instead of antisymmetric.
        Symmetric kernels map furthermore between pseudoscalar fields.
    }
    \label{fig:intro_kernel_alignment_reflect}
\end{figure}

We come back to our initial question
\marginnote{$\GM$-coordinate independent weight sharing $~\mkern22mu~\Longleftrightarrow$ $G$-steerability}
about how to share a convolution kernel given that its alignment is inherently ambiguous when the structure group~$G$ is non-trivial.
As the convolution operation should be $\GM$-coordinate independent, the underlying weight sharing procedure needs to be independent from arbitrary choices of reference frames as well.
Aligning the kernel along different frames of the $G$-structure at some point~$p$ should therefore always yield equivalent results.
We demonstrate that the $\GM$-coordinate independence of the weight sharing process imposes a \emph{gauge equivariance constraint} (``$G$-steerability'') on the convolution kernels:
\begin{align*}
    \qquad
    \begin{array}{r@{}}
        \GM \mkern.0mu \rule[2.25pt]{4pt}{.5pt} \mkern1.5mu \textit{coordinate independence}\ \ \\[4pt]
        \textit{weight sharing}\ \ 
    \end{array}
    \Bigg\}
    \mkern-10mu \boldsymbol{\longrightarrow}
    \ G \mkern.75mu \rule[2.25pt]{4pt}{.5pt} \mkern1.5mu \textit{steerability}\ \ \textit{(gauge equivariance)}
\end{align*}
Fig.~\ref{fig:intro_steerable_kernel} and Table~\ref{tab:reflection_steerable_kernels} show examples of reflection steerable kernels.
The equivariance (steerability) constraint enforces some type of $G$-symmetry which ensures that the \emph{kernel responses transform predictably} under gauge transformations.
An intuitive demonstration of how $G$-steerable kernels resolve the ambiguity of reference frames is given in Fig.~\ref{fig:intro_kernel_alignment_reflect}.

\begin{samepage}
$\GM$-coordinate independent convolutions (or $\GM$-\emph{convolutions} in short)
{\setlength{\marginparwidth}{2.5cm}
\marginnote{$\GM$-convolutions}}%
are defined as convolutions with $G$-steerable kernels:
\begin{center}\it
    A $\GM$-convolution with a $G$-steerable kernel $K$ processes features by applying this \\
    kernel at each point $p\in M$ relative to an arbitrary choice of reference frame in $\GpM$. \\
\end{center}
\end{samepage}
The arbitrariness of the chosen gauge is thereby ensured by the kernel's $G$-steerability.
While being constructed in local coordinates, 
$\GM$-convolutions correspond to well defined \emph{coordinate free} convolution operation on the manifold.

Not only convolution kernels but \emph{any shared template function is required to be gauge equivariant} in order to preserve $\GM$-coordinate independence.
Section~\ref{sec:pointwise_operations} derives the equivariance constraints on shared biases, \onexone\ kernels and nonlinearities.
Note that the equivariance constraints on shared template functions can be viewed as a form of weight sharing over $G$-related reference frames
-- the ambiguity of sharing weights over the manifold~$M$ is therefore resolved by extending weight sharing over the whole $G$-structure~$\GM$.

\toclesslab\subsection{Equivariance under global symmetries of the manifold}{sec:intro_overview_isometry}

\begin{SCfigure}
    \centering
    \hspace*{2ex}
    \includegraphics[width=.5\textwidth]{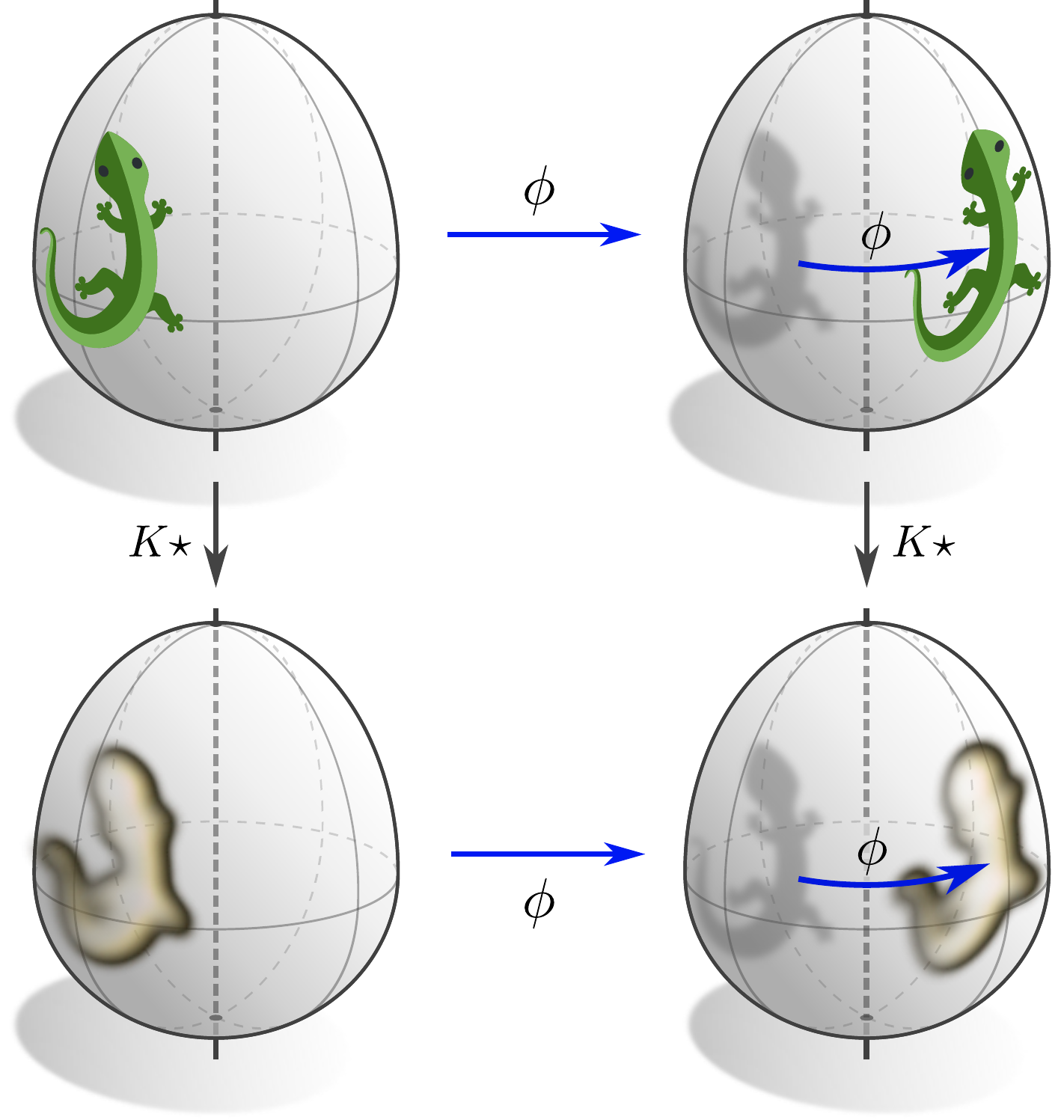}
    \captionsetup{width=.9\textwidth}
    \caption[]{\small
        Visualization of an isometry equivariant $\GM$-convolution.
        An isometry $\phi$ acts via pushforward on feature fields.
        The $\GM$-convolution $K\star$ with a $G$-steerable kernel $K$ is said to be equivariant w.r.t. this isometry when the convolution response of a transformed input ($\rightarrow,\downarrow$) agrees with the pushforward of the untransformed input's response ($\downarrow,\rightarrow$).
        \\[1ex]
        $\GM$-convolutions are equivariant w.r.t. the (sub)group $\IsomGM$ of isometries that are symmetries of the $G$-structure.
        The visualized azimuthal rotation equivariance requires therefore a $G$-structure that is invariant under rotations around the polar axis -- this condition is for instance met by the analogue of the spherical $G$-structures in Figs.~\ref{fig:G_structure_intro_h} and~\ref{fig:G_structure_intro_k} on the egg.
        The isometry group of the egg-shaped manifold contains not only rotations but also reflections.
        To achieve reflection equivariance, the $G$-structures would additionally have to contain reflected frames and the $\GM$-convolution would have to apply reflection steerable kernels.
        {\\
        \color{gray}
        \scriptsize
            (Lizards adapted under the Creative Commons Attribution 4.0 International
            \href{https://github.com/twitter/twemoji/blob/gh-pages/LICENSE-GRAPHICS}{\underline{license}}
            by courtesy of Twitter.)
        }
        \\[-16pt]
        }
    \label{fig:lizard_conv_egg_intro}
\end{SCfigure}

Convolution operations are often designed to be equivariant w.r.t. the symmetries of the underlying space \cite{Cohen2016-GCNN,Kondor2018-GENERAL}.
The symmetries of a Riemannian manifold form its \emph{isometry group} $\IsomM$, which is the group of all distance preserving maps~${\phi:M\to M}$.
Isometries act naturally on geometric quantities like feature vectors by ``moving them along'' with the isometry action (pushforward); see Fig.~\ref{fig:intro_gauge_isom_induction} (middle).
Despite only being designed to be equivariant under local gauge transformations,
$\GM$-convolutions are equivariant w.r.t. the action of specific isometry subgroups $\IsomGM \leq \IsomM$ on feature fields.
These subgroups $\IsomGM$ contain those \emph{isometries which are symmetries of the $G$-structure}.
The design of isometry equivariant $\GM$-convolutions is therefore linked to the design of invariant $G$-structures.
Fig.~\ref{fig:lizard_conv_egg_intro} visualizes the idea of an isometry equivariant $\GM$-convolution graphically.

We will in the following briefly discuss symmetry constraints which the requirement of isometry equivariance imposes on kernel fields, i.e. on the neural connectivity.
These conditions on the kernel fields correspond to symmetry constraints on the $G$-structures along which convolution kernels are shared.
We will finally comment on why $\GM$-convolutions are only isometry equivariant and are in general not equivariant w.r.t. more general diffeomorphisms.

\pagebreak

\subsubsection{Isometry invariant kernel fields}
\label{sec:visual_intro_inv_kernel_fields}

The equivariance properties of a neural network
\marginnote{kernel field transforms}
depend ultimately on symmetries in its neural connectivity~\cite{ravanbakhsh2017EquivarianceParameterSharing}.
For convolutional networks this amounts to \emph{symmetry constraints on the kernel field}.
To study these constraints in full generality, we do not require the kernel fields to be \emph{convolutional}, i.e. determined by a single shared kernel,
but assume \emph{general kernel fields} which may apply a different kernel at each point in space.
We denote the corresponding network layers as \emph{``kernel field transforms''}:%
\footnote{
    Kernel field transforms are in the computer vision literature sometimes called ``\emph{locally connected networks}''.
}
\begin{center}\it
    ``Kernel field transforms'' are similar to $\GM$-convolutions but may apply a different kernel at every point. \\
    They are therefore parameterized by general kernel fields.
\end{center}
$\GM$-convolutions are specific kernel field transforms with $\GM$-convolutional kernel fields.
The general results for isometry equivariant kernel field transforms translate therefore immediately to $\GM$-convolutions.

Theorem~\ref{thm:isometry_equivariant_kernel_field_trafos} proves
\marginnote{isometry invariant kernel~fields}
that the invariance of a kernel field under the isometry action and the equivariance of the corresponding kernel field transform imply each other:
\begin{center}\it
    isometry invariant kernel field
    $\quad \Longleftrightarrow \quad$
    isometry equivariant kernel field transform
\end{center}
Invariant kernel fields are constrained in two qualitatively different ways:
Firstly, kernels need to be shared over the \emph{isometry orbits}.
Secondly, the shared kernels are themselves constrained by the \emph{stabilizer subgroup} of the corresponding orbit.
Note that the full, symmetric kernel field can be recovered from a single representative kernel for each orbit --
Theorems~\ref{thm:tangent_quotient_repr_kernel_fields} and~\ref{thm:manifold_quotient_repr_kernel_fields} formalize this statement by proving isomorphisms between invariant kernel fields on the manifold and kernel fields on quotient spaces under the isometry action.

\begin{figure}
    \hspace{1.ex}
    \begin{subfigure}[b]{0.24\textwidth}
        \includegraphics[width=.92\textwidth]{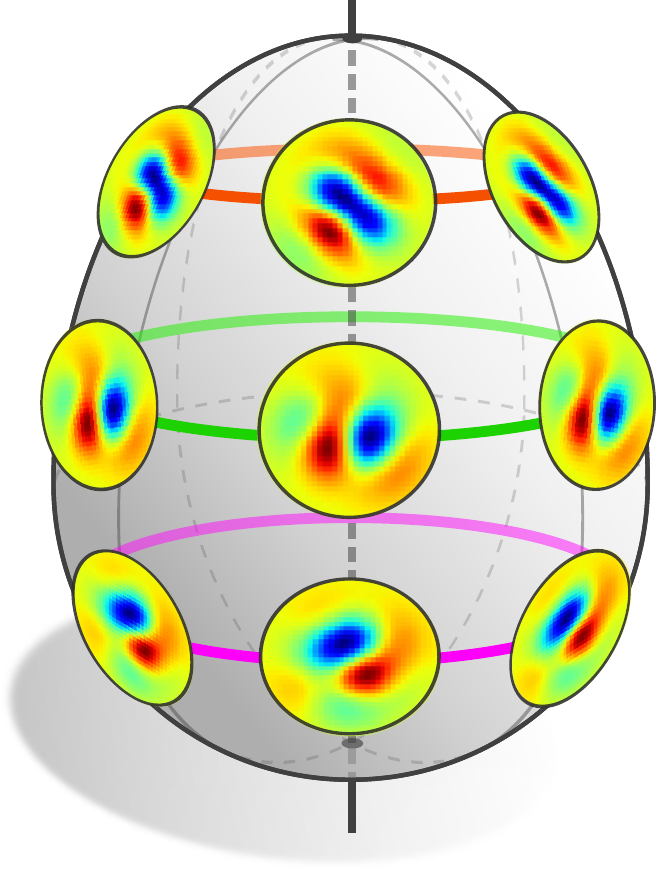}
        \vspace*{-.5ex}
        \captionsetup{format=hang, width=.7\textwidth}
        \caption{\small
            $\SO2$-invariant kernel field
        }
        \label{fig:isom_invariant_kernel_field_intro_SO2}
    \end{subfigure}
    \hspace*{1.5ex}
    \begin{subfigure}[b]{0.24\textwidth}
        \includegraphics[width=.92\textwidth]{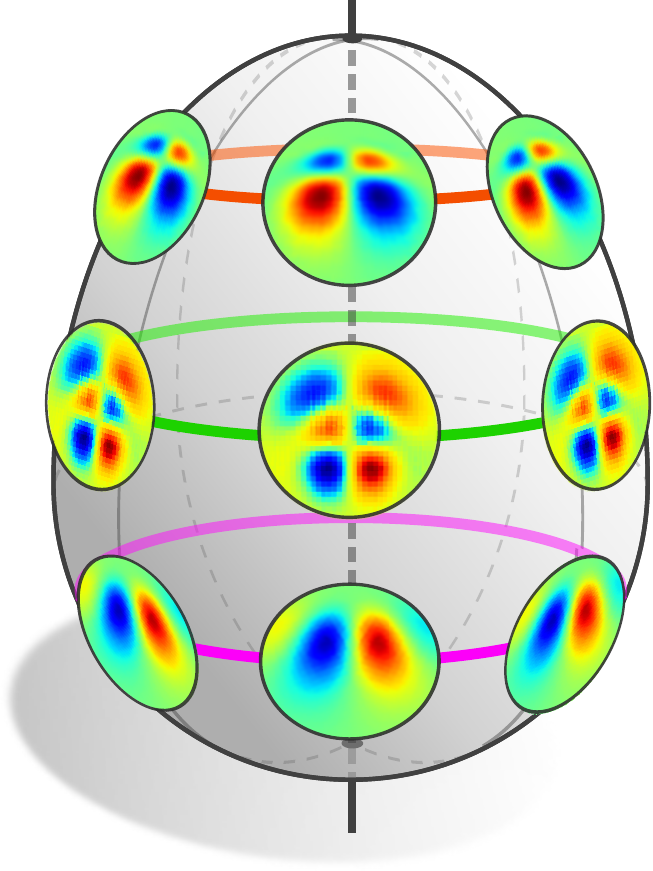}
        \vspace*{-.5ex}
        \captionsetup{format=hang, width=.7\textwidth}
        \caption{\small
            $\O2$-invariant kernel field
        }
        \label{fig:isom_invariant_kernel_field_intro_O2}
    \end{subfigure}
    \par
    \vspace*{\dimexpr-\parskip-158.pt\relax}
    \parshape 17 
        .55\textwidth .45\textwidth 
        .55\textwidth .45\textwidth
        .55\textwidth .45\textwidth
        .55\textwidth .45\textwidth
        .55\textwidth .45\textwidth
        .55\textwidth .45\textwidth
        .55\textwidth .45\textwidth
        .55\textwidth .45\textwidth
        .55\textwidth .45\textwidth
        .55\textwidth .45\textwidth
        .55\textwidth .45\textwidth
        .55\textwidth .45\textwidth
        .55\textwidth .45\textwidth
        .55\textwidth .45\textwidth
        .55\textwidth .45\textwidth
        .55\textwidth .45\textwidth
        .01\textwidth .98\textwidth 
    \makeatletter
    \setcounter{\@captype}{\value{\@captype}-1} 
    \refstepcounter{\@captype}
    \addcontentsline{\csname ext@\@captype\endcsname}{\@captype}
        {\protect\numberline{\csname the\@captype\endcsname}{ToC entry}}%
    \small 
    \csname fnum@\@captype\endcsname: 
    \makeatother
    \hyphenpenalty = 300 
        \emph{Kernel field transforms} are similar to $\GM$-convolutions but may apply a different kernel at each point.
        Theorem~\ref{thm:isometry_equivariant_kernel_field_trafos} proves that \emph{isometry equivariant kernel field transforms} are in one-to-one correspondence with \emph{isometry invariant kernel fields}.
        This implies
        1) weight sharing over the \emph{isometry orbits} (colored rings) and
        2) a constraint on the kernels to be steerable w.r.t. the \emph{stabilizer subgroup} of their respective orbit.
        Fig.~\ref{fig:isom_invariant_kernel_field_intro_SO2} shows an $\SO2$-invariant kernel field on an egg-shaped manifold.
        Different orbits are free to use different kernels without compromising $\SO2$-equivariance.
        The kernels themselves are unconstrained since the stabilizer subgroups of the $\SO2$ action on the orbits are trivial (except for at the poles).
        Fig.~\ref{fig:isom_invariant_kernel_field_intro_O2} visualizes the case of an $\O2$-invariant kernel field.
        The orbits are the same, however, the stabilizer subgroups impose a reflectional equivariance constraint.
        Note that the notion of invariance depends on the group action on the kernel, Def.~\ref{dfn:isometry_action_kernel_fields}, which depends in turn on the chosen feature field representations.
        The antisymmetric kernels are in this sense invariant under reflections.
    \label{fig:isom_invariant_kernel_field_intro}
\end{figure}

Figs.~\ref{fig:isom_invariant_kernel_field_intro_SO2} and~\ref{fig:isom_invariant_kernel_field_intro_O2}
\marginnote{examples}
exemplify these results for an egg-shaped manifold, whose isometries are rotations and reflections around the vertical axis.
A general kernel field transform could apply any kernel field.
If only $\SO2$ isometry equivariance is required (no reflections), the orbits are rings around the egg and the stabilizer subgroup of the $\SO2$-action on these orbits is trivial; see Fig.~\ref{fig:isom_invariant_kernel_field_intro_SO2}.%
\footnote{
    We are for brevity ignoring the poles, where the stabilizer subgroup is $\SO2$ in Fig.~\ref{fig:isom_invariant_kernel_field_intro_SO2} and $\O2$ in Fig.~\ref{fig:isom_invariant_kernel_field_intro_O2}.
}
The isometry invariant kernel field is therefore sharing unconstrained kernels over these orbits.
Fig.~\ref{fig:isom_invariant_kernel_field_intro_O2} shows an $\O2$-invariant kernel field.
The orbits, and therefore the spatial weight sharing pattern, are here the same as in the previous case.
However, the stabilizer subgroup of the $\O2$-action on the orbits is the reflection group.
The kernels are therefore constrained to be reflection steerable, with the exact constraint depending on the types $\rhoin$ and $\rhoout$ of the input and output feature field.

An interesting special case
\marginnote{homogeneous spaces}
is that of manifolds which are \emph{homogeneous spaces} of their isometry group.
In this case there is only one single orbit, such that an invariant kernel field is determined by a \emph{single shared (convolution) kernel}.
Theorem~\ref{thm:GM_conv_homogeneous_equivalence} proves:
\begin{center}\it
    Isometry equivariant kernel field transforms on homogeneous spaces are necessarily convolutions.
\end{center}
The kernels are again required to be steerable w.r.t. the stabilizer subgroup of the isometry action.
This recovers the results of \citet{Kondor2018-GENERAL}, \citet{Cohen2019-generaltheory} and \citet{bekkers2020bspline}, who investigated group equivariant CNNs on homogeneous spaces; see Appendix~\ref{apx:homogeneous_conv} for an in-depth comparison.

\subsubsection{Isometry equivariance of \textit{GM}-convolutions}
\label{sec:visual_intro_isom_equiv_conv}

\begin{SCfigure}
    \centering
    \includegraphics[width=.62\textwidth]{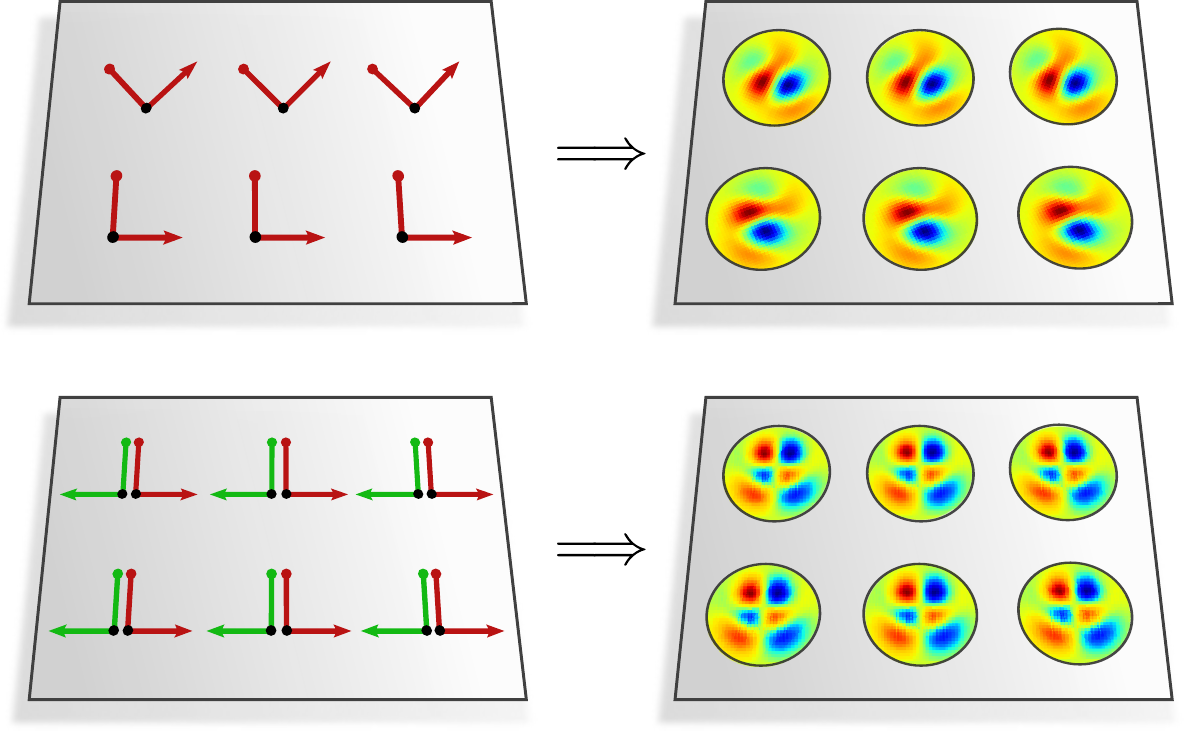}
    \captionsetup{width=.92\textwidth}
    \caption{\small
        \mbox{$\GM$-convolutional} kernel fields are constructed by sharing some $G$-steerable kernel along (arbitrary) frames of the \mbox{$G$-structure} $\GM$.
        The symmetries of the kernel fields agree therefore with those of the $G$-structure, i.e. with $\IsomGM$.
        The $\{e\}$-structure that is visualized at the top corresponds thus to a \mbox{$\GM$-convolution} which is equivariant w.r.t. translations in horizontal direction but not in vertical direction.
        As the \mbox{$\Flip$-structure} that is visualized at the bottom is invariant under arbitrary translations and horizontal reflections, the implied \mbox{$\GM$-convolution} is translation and reflection equivariant.
        \\[-5pt]
        }
    \label{fig:intro_invariant_kernel_fields_plane}
\end{SCfigure}

\paragraph{Isometry invariant \textit{G}-structures:}
$\GM$-convolutions are specific kernel field transforms
\marginnote{isometry invariant $G$-structures}
which rely on $\GM$-convolutional kernel fields.
They are therefore isometry equivariant if the $\GM$-convolutional kernel field is invariant under the isometry action.
Recall that $\GM$-convolutional kernel fields are defined by sharing some $G$-steerable kernel along (arbitrary) frames of the $G$-structure.
Their symmetries (isometries) coincide therefore with those of the $G$-structure.
\begin{center}\it
    We define $\IsomGM \leq \IsomM$ as the subgroup of those isometries which are \\
    symmetries of the $G$-structure $\GM$ (principal bundle automorphisms).
\end{center}
\begin{minipage}{\textwidth}
It follows that
\begin{center}\it
    $\GM$-convolutional kernel fields are $\IsomGM$-invariant.
\end{center}
\vspace*{1ex}\end{minipage}
\begin{minipage}{\textwidth}
As a consequence, which is proven rigorously in Theorem~\ref{thm:isom_equiv_GM_conv}, we find:
\begin{center}\it
    $\GM$-convolutions are $\IsomGM$-equivariant.
\end{center}
\vspace*{1ex}\end{minipage}
The design of isometry equivariant $\GM$-convolutions is therefore reduced to the design of isometry invariant $G$-structures.
Fig.~\ref{fig:intro_invariant_kernel_fields_plane} shows two examples of $G$-structures $\GM$ and corresponding $\GM$-convolutional kernel fields, which share the same symmetries.
For more examples we refer back to the $G$-structures in Fig.~\ref{fig:G_structures_intro}.
Note that $\IsomGM$ contains for $G=\O{d}$ (or supergroups of it) all possible isometries, implying that the corresponding $\GM$-convolutions are fully $\IsomM$-equivariant.
A similar statement holds for $G=\SO{d}$ and orientation preserving isometries.

\paragraph{Isometry induced gauge transformations:}

The argumentation in terms of invariant kernel fields
\marginnote{isometry induced gauge transformations}
and invariant $G$-structures did not rely on any choice of gauge but was formulated in a purely geometric, \emph{coordinate free} setting.
An alternative viewpoint explains the isometry equivariance of $\GM$-convolutions in coordinates, where isometries act via induced gauge transformations.
These gauge transformations are explained away by the kernel's $G$-steerability, which provides a link between the concepts of \emph{gauge equivariance} and \emph{isometry equivariance}.
We elaborate in the following concisely on this alternative viewpoint.

\begin{figure}
    \centering
    \includegraphics[width=.98\columnwidth]{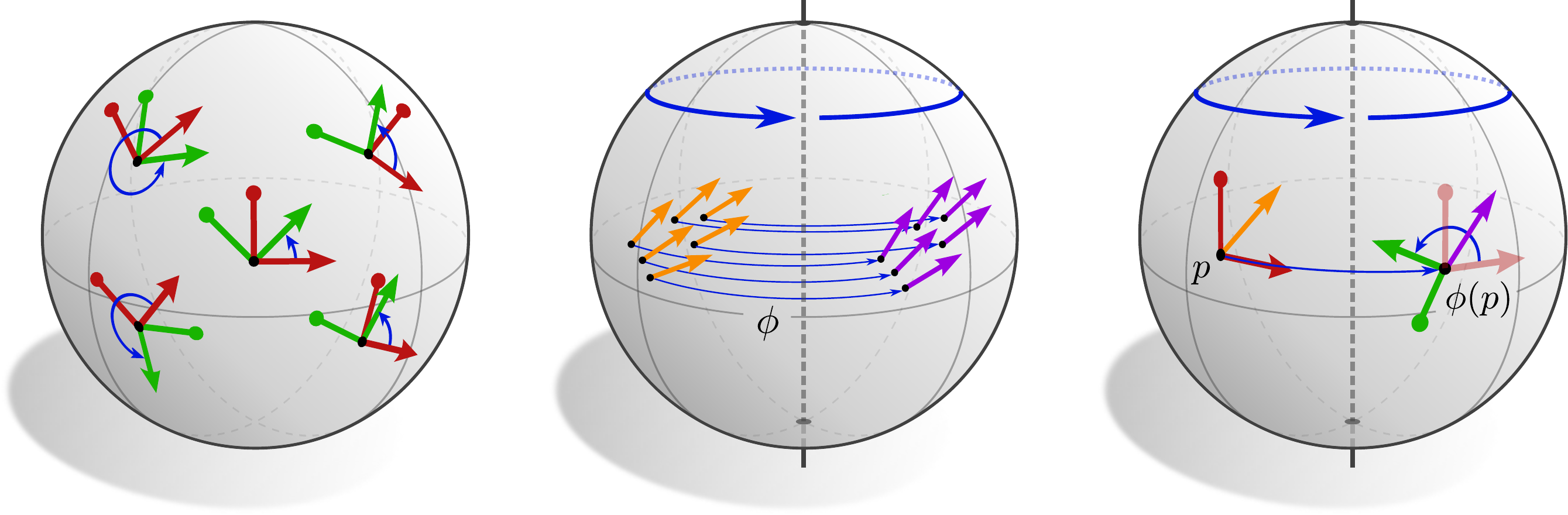}
    \vspace*{-1ex}
    \caption{\small
        Gauge transformations, isometries and their mutual relation.
        \ \ \emph{Left:}
        A gauge is a choice of local reference frames (a frame field), relative to which geometric quantities may be expressed.
        If the manifold's structure group $G$ is non-trivial, the choice of gauge is not unique but many equivalent choices exist.
        Different choices of gauges (red or green) are related by gauge transformations (blue) which are by definition taking values in the structure group~$G$.
        Visualized are orthonormal, right-handed frames, for which the gauge transformations are $G=\SO{2}$-valued.
        \ \ \emph{Middle:}
        Isometries are the symmetries of Riemannian manifolds.
        They are defined as distance preserving functions $\phi: M \to M$, mapping the manifold to itself.
        Isometries act via pushforward on tangent vectors, reference frames and feature vectors.
        While gauge transformations are passive coordinate transformations, isometries are actively moving points and geometric quantities over the manifold.
        \ \ \emph{Right:}
        When being expressed relative to local reference frames, the action of isometries can be thought of as inducing gauge transformations.
        Assume frames at $p$ (red) and $\phi(p)$ (green) to be given.
        A geometric quantity at $p$ (orange) is by the isometry pushed to $\phi(p)$ (purple).
        Since $\phi$ is an isometry, the Riemannian geometry around $p$ and $\phi(p)$ is indistinguishable, however, the pushforward of the geometric quantity is expressed relative to a new reference frame (green instead of red).
        One can therefore view isometries as inducing gauge transformations.
        If these induced gauge transformations take values in the structure group $G$, they are explained away by the $G$-steerability (gauge equivariance) of convolution kernels -- $\GM$-convolutions are then isometry equivariant.
        This condition is always met for $G\geq\O{d}$.
    }
    \label{fig:intro_gauge_isom_induction}
\end{figure}

When being expressed \emph{relative to local reference frames}, isometries can be thought of as acting via \emph{induced gauge transformations} on feature vector coefficients.
Fig.~\ref{fig:intro_gauge_isom_induction} (right) visualizes this concept:
assume that we picked reference frames at some point $p$ and at its image $\phi(p)$ under the action of an isometry~$\phi$.
The isometry pushes a feature from $p$ to $\phi(p)$.
As the Riemannian geometries around $p$ and $\phi(p)$ are indistinguishable, the only thing that changed from the viewpoint of the feature vector is with respect to which reference frame it is being expressed.
This change of frames is the isometry induced gauge transformation.

Recall that $\GM$-convolutions apply the same $G$-steerable kernel at each point of the manifold.
The kernel's $G$-equivariance accounts for any $G$-valued (induced) gauge transformation, such that:
\begin{center}\it
    If an isometry induces gauge transformations that take values in the structure group~$G$, \\
    then any $\GM$-convolution is equivariant w.r.t. this isometry.
\end{center}

The induced gauge transformations for general isometries might not take values in the chosen structure group~$G$.
For instance, the $G$-structures in Figs.~\ref{fig:G_structure_intro_a} and~\ref{fig:G_structure_intro_d} have structure groups $G=\{e\}$ and $G=\Flip$, respectively, but rotations of $M=\R^2$ induce $\SO2$- or $\O2$-valued gauge transformations.
However, if an isometry is a symmetry of the $G$-structure, i.e. an element of $\IsomGM$, it maps frames in~$\GM$ to frames in~$\GM$.
Since frames in~$\GM$ are related by $G$-valued gauge transformations it follows that:
\begin{center}\it
    Isometries in $\IsomGM$ induce gauge transformations that take values in the structure group $G$.
\end{center}
This implies that $\GM$-convolutions are $\IsomGM$-equivariant, as already found in the coordinate free setting.
While this alternative viewpoint is less geometrically intuitive, it emphasizes the role of the kernels' $G$-steerability in the networks' isometry equivariance.

\subsubsection{Diffeomorphisms, isometries and affine transformations:}
\label{sec:visual_intro_diffeo_equiv}

Most of the arguments in the previous paragraphs would not only hold for isometries, but also for diffeomorphisms.
This raises the question whether $\GM$-convolutions are equivariant w.r.t. more general diffeomorphism groups.
In some settings this is indeed the case, however, $\GM$-convolutions rely in general (additionally) on the manifolds' metric structure, which is only preserved by isometries.

Further above, we described $\GM$-convolutions as ``applying a $G$-steerable kernel relative to frames of the $G$-structure''.
More precisely, the kernel is applied in \emph{geodesic normal coordinates}.
This means that the feature field is at each point $p\in M$ via the \emph{Riemannian exponential map} $\exp_p$ pulled back to the tangent space $\TpM$, where it is matched with the kernel; see Fig.~\ref{fig:pullback_field_exp_TpM}.
While the kernel field would itself be invariant under those diffeomorphisms $\DiffGM\leq\DiffM$ that are symmetries of the $G$-structure, the exponential map depends explicitly on the metric structure.
$\GM$-convolutions with \emph{spatially extended kernels} can therefore in general only be isometry equivariant.

Any network layer that shares weights and does not rely on the manifolds' metric structure can be diffeomorphism equivariant.
Important examples of $\DiffGM$-equivariant layers are \emph{pointwise operations} like bias summation, nonlinearities and \onexones, which are introduced in Section~\ref{sec:pointwise_operations}.
It should be possible to generalize our theory to \emph{neural PDEs on smooth manifolds} which replace the non-local interactions of $\GM$-convolutions with \emph{local interactions} in terms of learnable $G$-steerable differential operators.
Note that $\GM$-convolutions are in practice often using compactly supported kernels and are therefore \emph{quasi-local}~\cite{tomboulis2015nonlocal}.
The $\DiffGM$-equivariance of such $\GM$-convolutions with small kernels should hold approximately.

A special case is that of $\GM$-convolutions on Euclidean spaces.
The exponential map is on Euclidean spaces not only preserved by isometries, but by any \emph{affine transformation}.
Theorem~\ref{thm:affine_equivariance_Euclidean_GM_conv} proves that Euclidean $\GM$-convolutions are indeed equivariant under the action of affine groups $\Aff(G)$ (Eq.~\eqref{eq:AffG_def}) if affine invariant $G$-structures as in the first column of Fig.~\ref{fig:G_structures_intro} are chosen.
This includes the isometries $\E{d}$ of Euclidean spaces, but also e.g. \emph{scale equivariant CNNs}.

\toclesslab\subsection{On the choice of \textit{G}-structures}{sec:intro_overview_G_structure_choice}

\begin{SCfigure}
    \centering
    \includegraphics[width=.46\columnwidth]{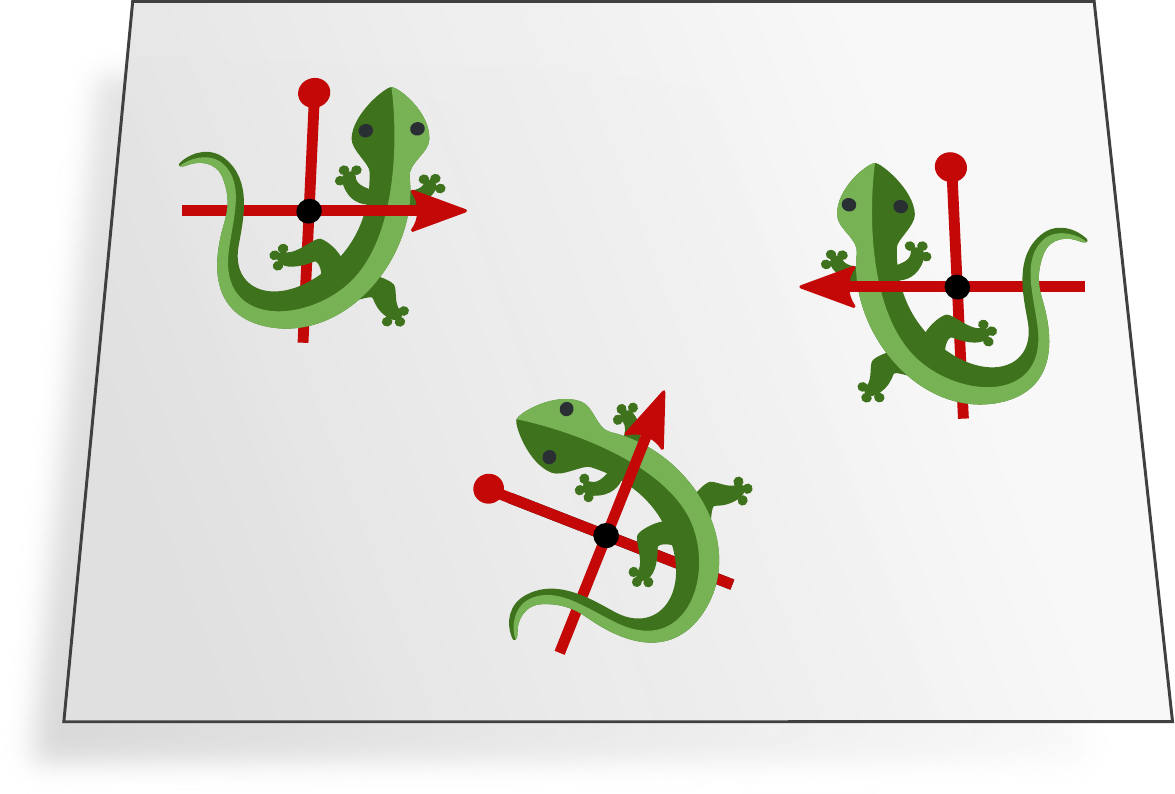}
    \captionsetup{width=1\textwidth}
    \hspace{1.5ex}
    \caption{\small
        Typical patterns in signals appear commonly in different geometric poses.
        $\GM$-convolutions generalize over all poses that are related by the action of the chosen structure group~$G$.
        While a conventional CNN on ${M=\R^2}$ would have to learn to detect all lizards individually, a $\GM$-coordinate independent CNN would for $G=\SO2$ generalize between the left and bottom lizard and for $G=\Flip$ generalize between the left and right lizard.
        For $G=\O2$, it would be guaranteed to encode all three lizards as the \emph{same feature} in \emph{different poses}.
        {
        \color{gray}
        \scriptsize
            (Lizards adapted under the Creative Commons Attribution 4.0 International
            \href{https://github.com/twitter/twemoji/blob/gh-pages/LICENSE-GRAPHICS}{\underline{license}}
            by courtesy of Twitter.)
        }
        \\\protect\rule{0ex}{2.5ex}
        }
    \label{fig:intro_lizard}
\end{SCfigure}

A point that was so far left open is the choice of $G$-structure.
In general, a Riemannian manifold comes with a metric structure, i.e. an $\O{d}$-structure.
The reduction of the structure group to subgroups $G<\O{d}$ might be obstructed by the topology of the manifold
-- \emph{smooth} (or continuous) $G$-structures \emph{do not exist} if this is the case.
Except from this constraint, the choice of $G$-structure is mainly an engineering question, depending for instance on the desired equivariance properties.

In most applications
\marginnote{topological obstructions}
it is sensible to demand that the prediction of the neural network varies \emph{smoothly} over the manifold.
As $\GM$-convolutions share kernels relative to frames of some $G$-structure, the convolution response is only then guaranteed to be smooth when the $G$-structure is smooth.
It is a well known fact that the topology of a manifold obstructs the (smooth) reduction of its structure group beyond a certain level.
This implies that \emph{a minimum level of gauge equivariance is strictly necessary} for a smooth convolution operation.
We prove in Theorem~\ref{thm:existence_kernel_field_trafo_compact_kernels} that our $\GM$-convolutions are indeed preserving the smoothness of feature fields.

An intuitive example of such a topological obstruction is the non-orientability of the M\"obius strip from Fig.~\ref{fig:G_structure_intro_l}:
a reduction of the structure group to the trivial group $\{e\}$ would correspond to some choice of smooth frame field on the strip.
However, due to the twist of the strip, this reduction would necessarily lead to a discontinuity in form of a frame reflection at some point
-- a reduction to $G=\{e\}$ is therefore topologically obstructed and $\Flip$-steerable kernels are inevitable on the M\"obius strip.
Another example is the 2-sphere~$S^2$, which does not admit a reduction beyond~$G=\SO2$.

It is in the deep learning literature not uncommon
\marginnote{discontinuous networks}
to ignore such topological obstructions and to implement discontinuous networks.
An example are spherical CNNs which rely on the $\{e\}$-structure in Fig.~\ref{fig:G_structure_intro_k}; see Section~\ref{sec:spherical_CNNs_azimuthal_equivariant} for a detailed review of such models.
Note that a removal of the poles, where the frame field is singular, turns the spherical topology into a cylindrical topology, where a smooth reduction to $G=\{e\}$ is possible.
Section~\ref{sec:e_surface_conv} discusses further discontinuous $\GM$-convolutions which apply non-steerable kernels relative to heuristically (algorithmically) fixed $\{e\}$-structures.

It is important to understand that many different $G$-structures $\GM$ exist
\marginnote{non-uniqueness of $G$-structures}
for a given structure group~$G$ and manifold~$M$.
Figs.~\ref{fig:G_structure_intro_a} and~\ref{fig:G_structure_intro_b} show different choices of $\{e\}$-structures on $M=\R^2$ while
Figs.~\ref{fig:G_structure_intro_d} and~\ref{fig:G_structure_intro_e} show different $\Flip$-structures.
Different $\O{d}$-structures on $M$ correspond to different Riemannian metrics.
While the structure group~$G$ implies a kernel's $G$-steerability, the $G$-structure determines how exactly this kernel is to be shared.

Within the topological constraints,
\marginnote{global and local equivariance}
the structure group~$G$ and $G$-structure may be freely chosen by the user.
If the manifold comes with a non-trivial isometry group $\IsomM$, the $G$-structure $\GM$ is often designed such that the corresponding $\GM$-convolution is equivariant w.r.t. some subgroup~$\IsomGM$ of isometries.
Examples are given in Fig.~\ref{fig:G_structures_intro} and throughout our literature review in Part~\ref{part:literature_review}.
Even if the manifold is asymmetric, local patterns of features appear often in multiple different geometric poses.
The choice of structure group~$G$ provides a powerful tool to exploit such symmetries in the learning task:
$\GM$-convolutions generalize learned patterns automatically over any $G$-related pose; see Fig.~\ref{fig:intro_lizard}.
The optimal choice of structure group might thereby vary with the \emph{length scale} (field of view) and thus with the depth in the network~\cite{Weiler2019_E2CNN}.
For example, even though natural images are aligned on a global scale, local patterns like edges and corners usually appear in several directions, such that local gauge equivariance still proves to be useful.

%% file: chapters/intro_fig_G-structures.tex

\begin{subfigure}[b]{0.26\textwidth}
    \centering
    \includegraphics[width=1.\textwidth]{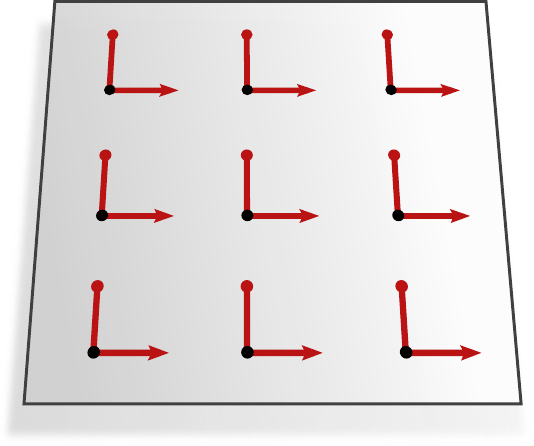}
    \captionsetup{format=hang}
    \caption{\small
        \,  $M = \R^2$,
        \,\ $G = \{e\}$
    }
    \label{fig:G_structure_intro_a}
\end{subfigure}
\hfill
\begin{subfigure}[b]{0.26\textwidth}
    \centering
    \includegraphics[width=1.\textwidth]{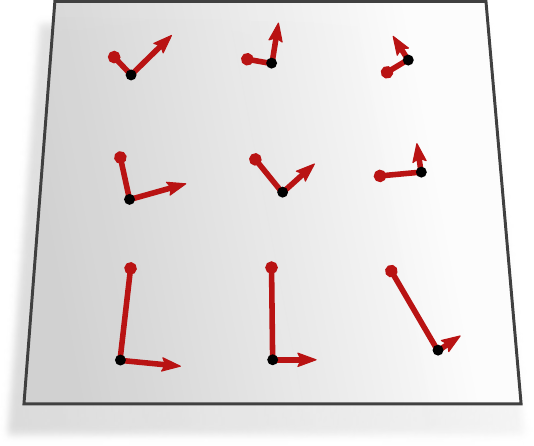}
    \captionsetup{format=hang}
    \caption{\small
        \,  $M = \R^2$,
        \,\ $G = \{e\}$
    }
    \label{fig:G_structure_intro_b}
\end{subfigure}
\hfill
\begin{subfigure}[b]{0.26\textwidth}
    \centering
    \includegraphics[width=1.\textwidth]{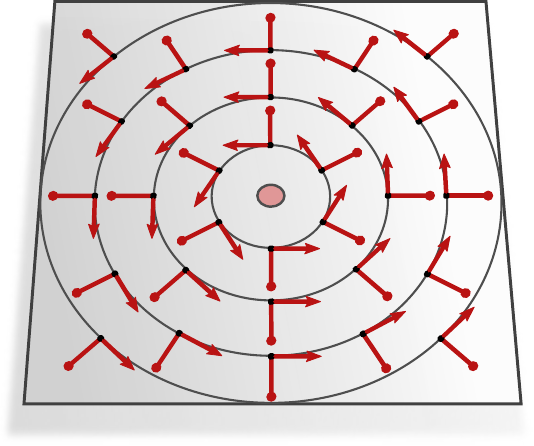}
    \captionsetup{format=hang}
    \caption{\small
        \,  $M = \R^2\backslash\{0\}$,
        \,\ $G = \{e\}$
    }
    \label{fig:G_structure_intro_c}
\end{subfigure}
\\[2ex]
\begin{subfigure}[b]{0.26\textwidth}
    \centering
    \includegraphics[width=1.\textwidth]{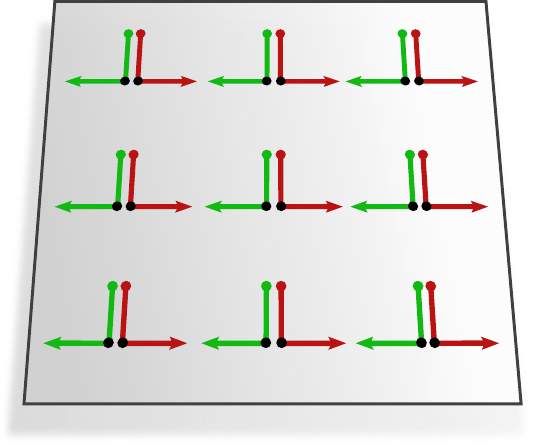}
    \captionsetup{format=hang}
    \caption{\small
        \,  $M = \R^2$,
        \,\ $G = \Flip$
    }
    \label{fig:G_structure_intro_d}
\end{subfigure}
\hfill
\begin{subfigure}[b]{0.26\textwidth}
    \centering
    \includegraphics[width=1.\textwidth]{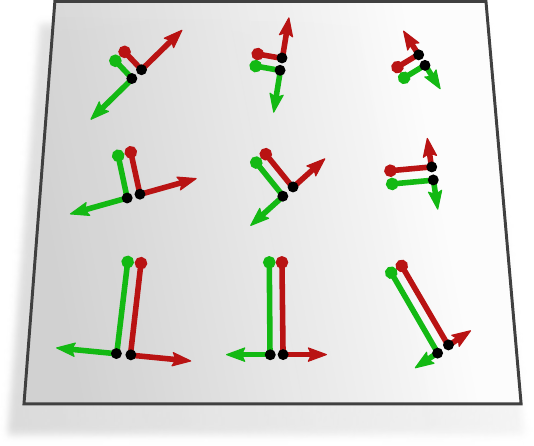}
    \captionsetup{format=hang}
    \caption{\small
        \,  $M = \R^2$,
        \,\ $G = \Flip$
    }
    \label{fig:G_structure_intro_e}
\end{subfigure}
\hfill
\begin{subfigure}[b]{0.26\textwidth}
    \centering
    \includegraphics[width=1.\textwidth]{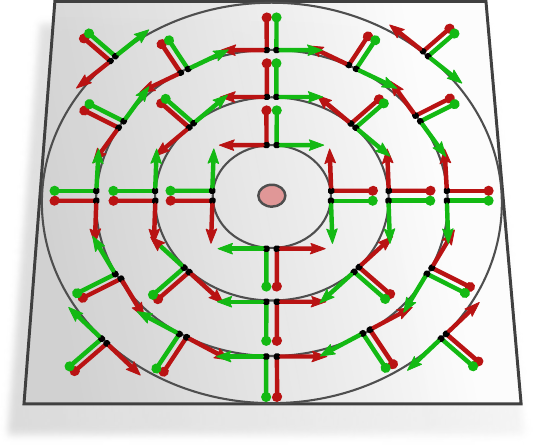}
    \captionsetup{format=hang}
    \caption{\small
        \,  $M = \R^2\backslash\{0\}$,
        \,\ $G = \Flip$
    }
    \label{fig:G_structure_intro_f}
\end{subfigure}
\\[2ex]
\begin{subfigure}[b]{0.26\textwidth}
    \centering
    \includegraphics[width=1.\textwidth]{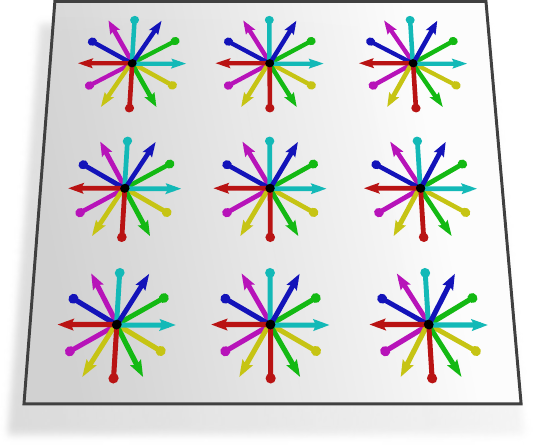}
    \captionsetup{format=hang}
    \caption{\small
        \,  $M = \R^2$,
        \,\ $G = \SO2$
    }
    \label{fig:G_structure_intro_g}
\end{subfigure}
\hfill
\begin{subfigure}[b]{0.26\textwidth}
    \centering
    \includegraphics[width=.95\textwidth]{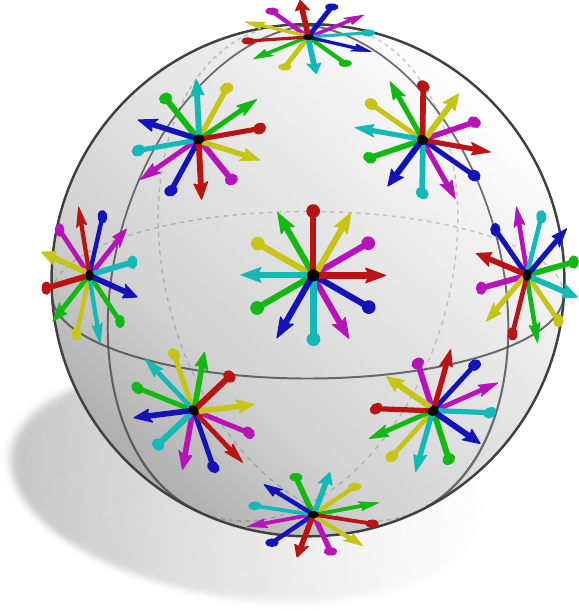}
    \vspace*{-2ex}
    \captionsetup{format=hang}
    \caption{\small
        \,  $M = S^2$,
        \,\ $G = \SO2$
    }
    \label{fig:G_structure_intro_h}
\end{subfigure}
\hfill
\begin{subfigure}[b]{0.26\textwidth}
    \centering
    \makebox[\textwidth][c]{ 
    \includegraphics[width=1.12\textwidth]{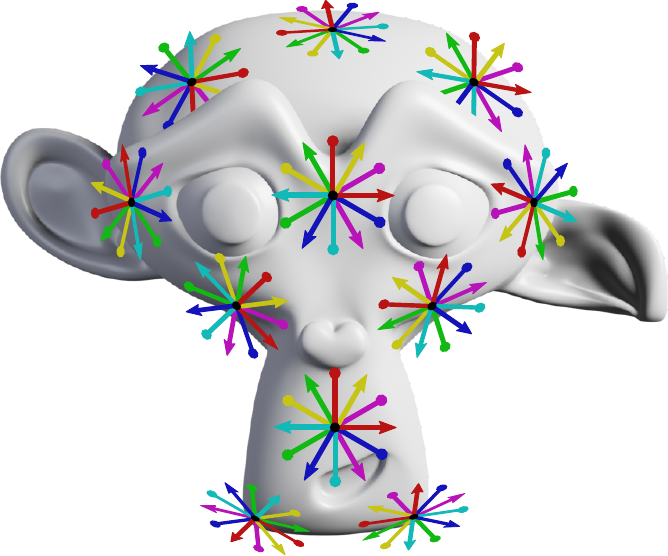}
    }
    \vspace*{-3.ex}
    \captionsetup{format=hang, width=1.1\textwidth}
    \caption{\small
        $M = \textup{``\href{https://en.wikipedia.org/wiki/Blender_(software)\#Suzanne}{Suzanne}''}$\!,
        \ $G = \SO2$
    }
    \label{fig:G_structure_intro_i}
\end{subfigure}
\\[2ex]
\begin{subfigure}[b]{0.26\textwidth}
    \centering
    \includegraphics[width=1.\textwidth]{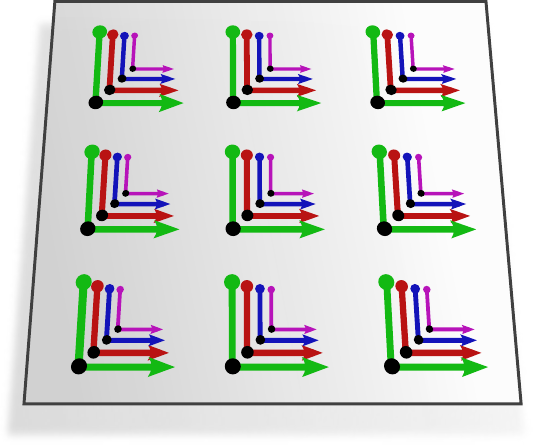}
    \captionsetup{format=hang}
    \caption{\small
        \,  $M = \R^2$,
        \,\ $G = \Scale$
    }
    \label{fig:G_structure_intro_j}
\end{subfigure}
\hfill
\begin{subfigure}[b]{0.26\textwidth}
    \centering
    \includegraphics[width=.95\textwidth]{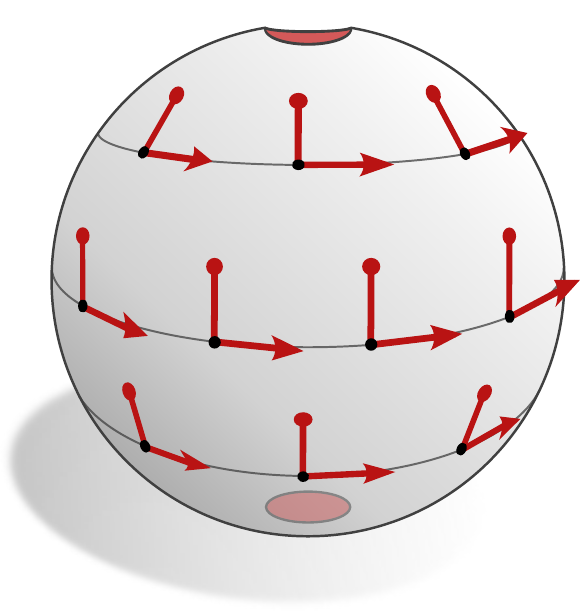}
    \vspace*{-2ex}
    \captionsetup{format=hang}
    \caption{\small
        \,  $M = S^2\backslash$poles,
        \,\ $G = \{e\}$
   }
    \label{fig:G_structure_intro_k}
\end{subfigure}
\hfill
\begin{subfigure}[b]{0.26\textwidth}
    \centering
    \makebox[\textwidth][c]{ 
    \includegraphics[width=1.1\textwidth]{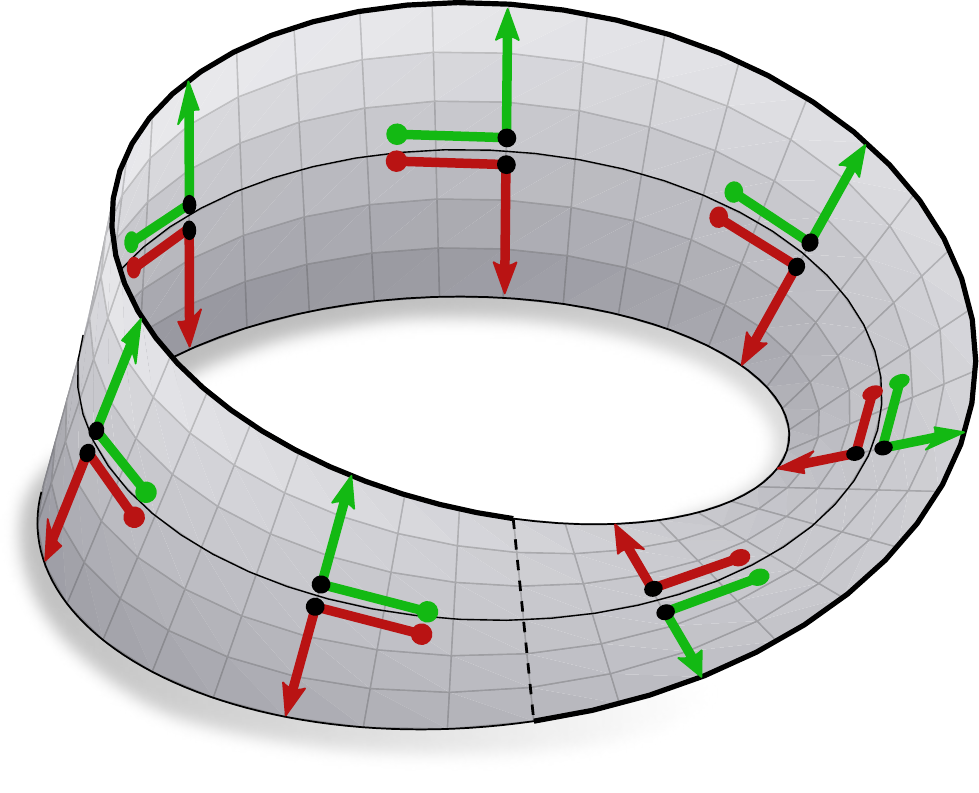}
    }
    \vspace*{-3.ex}
    \captionsetup{format=hang}
    \caption{\small
        \,  $M = \textup{M\"obius}$,
        \,\ $G = \Flip$
    }
    \label{fig:G_structure_intro_l}
\end{subfigure}

%% file: chapters/P1_intro.tex

\mypart{An introduction to coordinate independent CNNs}
\label{part:local_theory}

Convolutional networks extract a hierarchy of feature fields from an input signal on a manifold.
Features are thereby computed via kernels, optimized to detect characteristic spatial patterns in lower level features.
We demand that this inference process should be solely based on the relative arrangement of features but independent from the particular choice of coordinatization.
Features are therefore required to be coordinate independent geometric quantities, similar to scalars, vectors or tensors.
While such geometric quantities exist independently of coordinates, a (non-symbolic) computer implementation requires them to be expressed in terms of numerical coefficients in \emph{some} gauge, i.e. relative to some choice of reference frame.
The specific choice of coordinates is irrelevant -- it represents just one of multiple equivalent descriptions.
The appropriate mathematical framework to regulate such redundant degrees of freedom are \emph{gauge theories}.
A~gauge theory accounts for the equivalence of different gauges by consistently relating them to each other via \emph{gauge transformations}.
Fields of coordinate independent features are therefore associated with a certain transformation law, i.e. a group action of the structure group which describes how features transform under gauge transformations.
Any neural network layer processing such feature fields is required to respect their transformation laws in order to preserve their coordinate independence.

The aim of this first part of our work is to introduces coordinate independent CNNs in an easily accessible language.
Geometric intuition and visualizations are therefore favored over mathematical formalism.
A more formal exposition of the presented definitions and results is provided in Part~\ref{part:bundle_theory}.

\etocsettocdepth{2}
\etocsettocstyle{}{} 
\localtableofcontents

\vspace*{2.ex}

Section~\ref{sec:gauge_cnns_intro_local} introduces gauges and gauge transformations, based on which fields of coordinate independent feature vectors are defined.
Neural networks that map between such feature fields are developed in Section~\ref{sec:gauge_CNNs_local}.
Section~\ref{sec:mobius_conv} presents an exemplary instantiation of feature fields and network layers on the M\"obius strip.

%% file: chapters/30_local_intro.tex

\section{Coordinate independent feature fields}
\label{sec:gauge_cnns_intro_local}

The feature spaces of coordinate independent neural networks are spaces of feature vector fields.
The goal of this section is to define such feature fields and their geometric properties.

\etocsettocdepth{3}
\etocsettocstyle{}{} 
\localtableofcontents

Feature vectors are in practice represented relative to some reference frame and are characterized by their transformation laws when transitioning between different frames.
Section~\ref{sec:21_main} begins therefore with a discussion of the coordinatization of tangent spaces.
In particular, Section~\ref{sec:gauges_gauge_trafos} introduces gauges and gauge transformations of the tangent spaces as a formal way of describing choices of local reference frames and transformations between them.
Section~\ref{sec:gauges_TpM_functions} explains how functions on tangent spaces are represented relative to different coordinatizations
-- this introduces the idea of coordinate independent mappings, which we use later to define coordinate independent network layers.
Section~\ref{sec:local_G-structure_G-atlas} defines $G$-structures and $G$-atlases.
Coordinate independent feature fields and their gauge transformations are introduced in Section~\ref{sec:feature_fields}.
While Section~\ref{sec:individual_fields} describes the construction of individual feature fields,
Section~\ref{sec:stacked_fields} defines full feature spaces, consisting of multiple independent feature fields.
Parallel transporters of feature vectors and their representation relative to different coordinatizations are introduced in Section~\ref{sec:transport_local}.
Section~\ref{sec:isometries_local} discusses isometries and their action on geometric quantities like tangent vectors and feature vectors.

%% file: chapters/31_gauges.tex

\subsection{Gauges, gauge transformations and \textit{G}-structures}
\label{sec:21_main}

\subsubsection{Tangent spaces and reference frames}
\label{sec:gauges_gauge_trafos}

A $d$-dimensional (smooth) manifold $M$ has a tangent space $\TpM\cong\R^d$ attached to each point $p\in M.$
The tangent spaces are $d$-dimensional vector spaces, however, in contrast to $\R^d$ they do in general not come with any preferred choice of reference frame.
A tangent vector $v\in \TpM$ is a \emph{coordinate free} object and is thus not immediately represented numerically by a coordinate tuple $(v_1,\dots,v_d)\in\R^d$.
More abstractly stated, each tangent space $\TpM$ is isomorphic~to~$\R^d$ but in general no canonical isomorphism between them is given.
Both spaces are therefore structurally equivalent but are not identified with each other in any preferred way.

\begin{figure}
    \centering
    \includegraphics[width=\columnwidth]{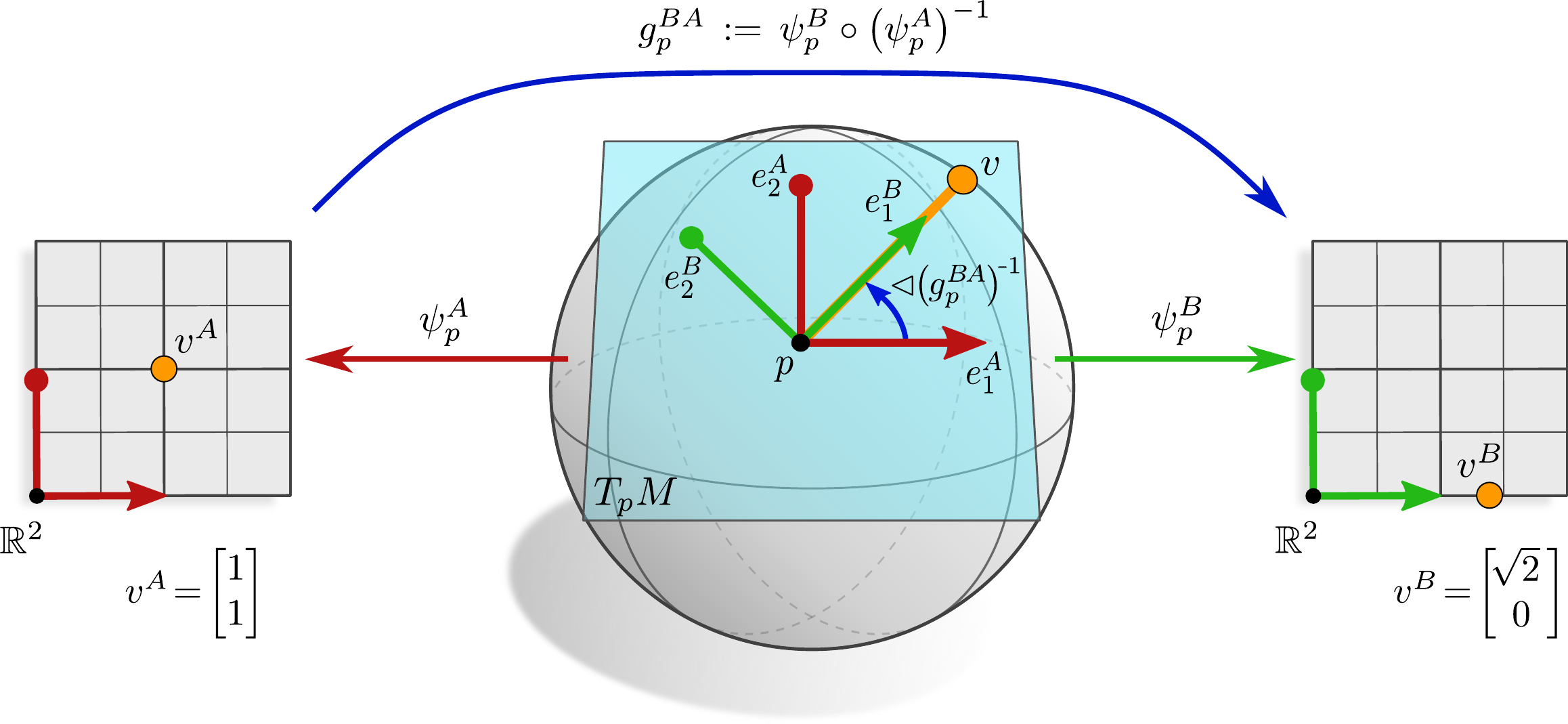}
    \caption{\small
        Identification of $\TpM\!\cong\!\R^2$ with $\R^2$ via different gauges.
        A (coordinate free) tangent vector $v\in \TpM$ (orange) can be represented numerically by a coordinate tuple $v^A=\psi_p^A(v)=\big(1,1\big)^\top$ relative to gauge $\psi_p^A$ (red) or, equivalently, by $v^B=\psi_p^B(v)=(\sqrt{2},0)^\top$ relative to gauge $\psi_p^B$ (green).
        A choice of gauge corresponds to a choice $[e_1^A,e_2^A]$ or $[e_1^B,e_2^B]$ of reference frame.
        On a general manifold no choice of gauge or coordinatization is preferred a~priori.
        Different gauges, and thus reference frames, are related by gauge transformations $g_p^{BA}:=\psi_p^B\circ(\psi_p^A)^{-1}$ (blue) which take values in the thus defined structure group $G$.
        This figure is a graphical interpretation of the commutative diagrams in Eq.~\eqref{eq:commutative_diagram_TpM} and Fig.~\ref{fig:trivialization_TM}.
        Note that gauges are immediately assigning coordinates to tangent spaces.
        Fig.~\ref{fig:affine_charts} in Section~\ref{sec:euclidean_geometry} shows a similar diagram for (affine) charts, which assign coordinates to the manifold, thereby \emph{inducing} gauges (``coordinate bases'').
        }
    \label{fig:gauge_trafos}
\end{figure}

\begin{SCfigure}
    \centering
    \includegraphics[width=.45\columnwidth]{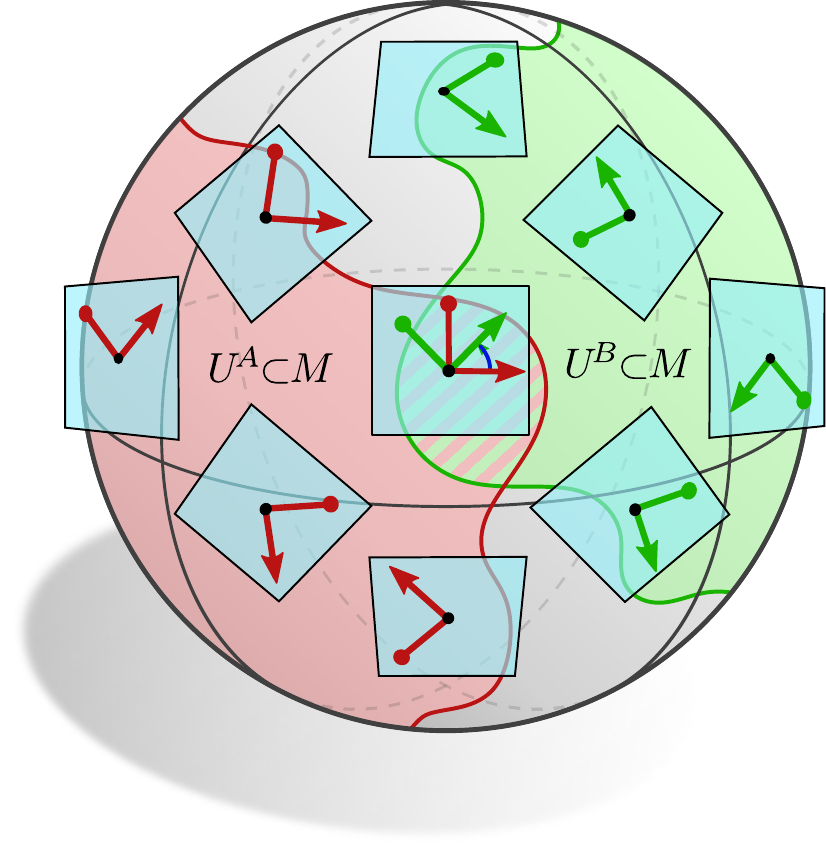}
    \captionsetup{width=.9\textwidth}
    \hspace{1ex}
    \caption{\small
        Each point $p$ of a Riemannian manifold $M$ has a tangent space $\TpM$ attached.
        A smooth gauge $\psi^A$ on a suitably chosen subset $U^A \subseteq M$ (red) coordinatizes all tangent spaces $\TpM$ for $p$ in $U^A$ as shown in Fig.~\ref{fig:gauge_trafos}.
        It is equivalent to a choice of \emph{smooth frame field} on $U^A$.
        Since it is in general not possible to extend a gauge globally over the whole manifold, it is necessary to consider a \mbox{$G$-\emph{atlas}}, consisting of gauges which cover $M$.
        Different coordinatizations $\psi^A$ on $U^A$ (red) and $\psi^B$ on $U^B$ (green) are patched together via gauge transformations (or transition maps) ${g^{BA}: U^A \cap U^B \to G}$ which are defined on the overlap ${U^A\cap U^B}$ (striped) and take values in the structure group $G \leq \GL{d}$.
        \\\protect\rule{0ex}{7.5ex}
        }
    \label{fig:gauge_trafos_manifold}
\end{SCfigure}

A \emph{gauge} (local trivialization of the tangent bundle) on $U^A\subseteq M$ is defined as a smoothly position dependent collection of invertible linear maps
\begin{align}\label{eq:gauge_definition}
    \psi_p^A:\TpM\to\R^d \,,\ \ p\in U^A \,,
\end{align}
specifying the missing vector space isomorphisms between $\TpM$ and $\R^d$.
As visualized in Fig.~\ref{fig:gauge_trafos}, they coordinatize the tangent spaces by assigning a \emph{coefficient vector}
\begin{align}
    v^A\ :=\ \psi_p^A(v) \ \in\R^d
\end{align}
to each coordinate free tangent vector $v\in \TpM$.
An inversion of this relation yields
\begin{align}
    v\ =\ \left(\psi_p^A\right)^{-1}\! (v^A)
     \ =\ \left(\psi_p^A\right)^{-1}\!\! \left(\sum\nolimits_i v_i^A \epsilon_i\right)
     \ =\ \sum\nolimits_i v_i^A\, \left(\psi_p^A\right)^{-1}\!(\epsilon_i)
     \ =:\ \sum\nolimits_i v_i^A\, e_i^A \,,
\end{align}
where we denoted by $\{\epsilon_1,\dots,\epsilon_d\}$ the standard basis of $\R^d$ and made use of the linearity of the gauge to pull out the summation.
This shows that the gauge can be thought of as endowing each tangent space $\TpM$ with a \emph{reference frame}
\begin{align}\label{eq:framefield_gauge_equivalence}
    \left[e^A_{1}, \,\dots,\, e^A_{d}\right]
    \ :=\ \Big[(\psi_p^A)^{-1}(\epsilon_1), \:\dots\,,\; (\psi_p^A)^{-1}(\epsilon_d)\Big] \,,
\end{align}
defined as that $d$-tuple of linearly independent tangent vectors which results when mapping the standard frame of $\R^d$ back through the inverse gauge map.
For brevity, we will in the following use the shorthand notation $\big[e_i^A \big]_{i=1}^d$ for frames $\big[e_1^A, \dots, e_d^A \big]$.
The coefficients $v^A$ are the coordinates of $v$ relative to this frame.
The collection of frames induced by the $\psi_p^A$ on $U^A$ is called (smooth) \emph{frame field}; see Fig~\ref{fig:gauge_trafos_manifold} for a visualization.

Gauges $\psi^X$ coordinatize tangent spaces only on local neighborhoods $U^X\subseteq M$, and can due to topological obstructions in general not be extended to the whole manifold without violating the smoothness assumption.
One therefore considers an \emph{atlas}
\begin{align}
    \mathscr{A} \,=\, \big\{\! \big(U^X, \psi^X\big) \!\big\}_{X\in \mathfrak{X}} \,,
\end{align}
consisting of smooth gauges on a set of neighborhoods $U^X$ covering the manifold, that is, satisfying $\bigcup_{X\in \mathfrak{X}} U^X = M$, where $\mathfrak{X}$ is an index set.%
\footnote{
    An atlas of gauges is very similar to usual atlases of charts of a manifold (Appendix~\ref{apx:coordinate_bases}).
    The difference is that the here considered atlases directly assign coordinates to the tangent bundle $\TM$ instead of to the manifold $M$.
}
On the overlaps $U^A\cap U^B\neq\varnothing$ of neighborhoods, different gauges $\psi_p^A$ and $\psi_p^B$ are stitched together by smooth \emph{transition functions}%
\begin{align}\label{eq:transition_fct_local_def_21}
    g^{BA}\!:\, U^A\cap U^B\to\GL{d},\quad p \mapsto g_p^{BA} := \psi_p^B \circ \left(\psi_p^A\right)^{-1}.
\end{align}
Here we assume the codomain (for now) to be given by the \emph{general linear group} $\GL{d}$, consisting of all invertible matrices in $ \R^{d\times d}$, which explain the relation between any pair of vector space isomorphisms (gauges) or reference frames.
The action of such a transition function on a given gauge defines a \emph{gauge transformation}
\begin{align}\label{eq:gauge_trafo_local_def_21}
    \psi_p^B\ =\ g_p^{BA} \!\cdot \psi_p^A.
\end{align}
In terms of a commutative diagram, the relation between different gauges is visualized as:%
\footnote{
    \mbox{
        \emph{Diagrams} give a visual overview of functions and the spaces between which they map.
        For instance, the diagram
    }
    \begin{minipage}{.2\textwidth}
    ${
    \mkern30mu
    \begin{tikzcd}[column sep=30pt, row sep=8pt, ampersand replacement=\&]
        X
            \arrow[r, pos=.55, "f" description]
            \arrow[rd, "h\!"']
        \& Y
            \arrow[d, "\;g"]
        \\
        \& Z
    \end{tikzcd}}$
    \end{minipage}
    \hfill
    \begin{minipage}{.8\textwidth}
    implies that there are functions $f:X\to Y$, \ $g:Y\to Z$ and $h:X\to Z$.
    If the compositions of functions along \emph{all} paths with the same start and endpoint agree, the diagram is said to be \emph{commutative}.
    Our example diagram is commutative if (and only if) \ $h = g\circ f$ holds.
    \end{minipage}
}
\vspace*{-1ex}
\begin{equation}\label{eq:commutative_diagram_TpM}
\begin{tikzcd}[column sep=50pt, row sep=5pt, font=\normalsize]
    \R^d
        \arrow[rr, rounded corners, to path={ 
                -- ([yshift=4.5ex]\tikztostart.north) 
                --node[above, pos=.5]{\small$g_p^{BA} \mkern2mu\cdot$} ([yshift=4.5ex]\tikztotarget.north) 
                -- (\tikztotarget.north)
                }]
    & \TpM
        \arrow[l, "\psi_p^A"']
        \arrow[r, "\psi_p^B"]
    & \R^d
        \arrow[ll, rounded corners, to path={ 
                -- ([yshift=-4.5ex]\tikztostart.south) 
                --node[below, pos=.5]{\small$g_p^{AB} \mkern2mu\cdot \,=\, \big( g_p^{BA} \big)^{-1} \mkern2mu\cdot$} ([yshift=-4.5ex]\tikztotarget.south) 
                -- (\tikztotarget.south)
                }]
\end{tikzcd}
\vspace*{-1ex}%
\end{equation}
Compare this diagram to its graphical interpretation in Fig~\ref{fig:gauge_trafos}.

A gauge transformation alters the coordinatization of the tangent spaces such that the same coordinate free tangent vector $v$ is represented by a different component vector
\begin{align}\label{eq:components_leftaction}
  v^B\ =\ g_p^{BA}v^A \,.
\end{align}
Since a gauge corresponds to a choice of frame field, a gauge transformation corresponds to a transformation between frame fields.
Specifically, a frame $\big[e_i^A\big]_{i=1}^d = \big[e_1^A,\dots,e_d^A\big]$ at $p\in M$ transforms to another frame
\begin{alignat}{3}\label{eq:frame_rightaction}
    \qquad
    \left[e_{i}^B\right]_{i=1}^d\ \notag
    :=&\ \left[ \left(\psi_p^B\right)^{-1} (\epsilon_i) \right]_{i=1}^d
        \qquad\quad && \big( \text{\small gauge induced frame, Eq.~\eqref{eq:framefield_gauge_equivalence} } \big) \notag\\
    =&\ \left[ \left(g_p^{BA} \cdot \psi_p^A\right)^{-1} \left(\epsilon_i\right) \right]_{i=1}^d
        \qquad\quad && \big( \text{\small gauge trafo, Eq.~\eqref{eq:gauge_trafo_local_def_21}} \big) \notag\\
    =&\ \left[ \left(\psi_p^A\right)^{-1} \left(\left(g_p^{BA}\right)^{-1} \epsilon_i\right) \right]_{i=1}^d
        \qquad\quad && \big( \text{\small expanded inverse} \big) \notag\\
    =&\ \left[ \left(\psi_p^A\right)^{-1} \left(\sum\nolimits_j \epsilon_j \epsilon_j^\top \left(g_p^{BA}\right)^{-1} \epsilon_i\right) \right]_{i=1}^d
        \qquad\quad && \big( \text{\small inserted identity}\ {\textstyle\mathds{1}=\sum_j\epsilon_j\epsilon_j^\top}\ \big) \notag\\
    =&\ \left[ \left(\psi_p^A\right)^{-1} \left(\sum\nolimits_j \epsilon_j \left(\!\left(g_p^{BA}\right)^{-1}\right)_{ji} \right) \right]_{i=1}^d
        \qquad\quad && \big( \text{\small identified matrix elements of $\big(g_p^{BA}\big)^{-1}$} \big) \notag\\
    =&\ \left[ \sum\nolimits_j \left(\psi_p^A\right)^{-1}\! (\epsilon_j)\ \left(\!\left(g_p^{BA}\right)^{-1}\right)_{ji} \right]_{i=1}^d
        \qquad\quad && \big( \text{\small linearity of $\psi_p^A$} \big) \notag\\
    =&\ \left[ \sum\nolimits_j e_{j}^A \left(\!\left(g_p^{BA}\right)^{-1}\right)_{ji} \right]_{i=1}^d
        \qquad\quad && \big( \text{\small gauge induced frame, Eq.~\eqref{eq:framefield_gauge_equivalence} } \big) \notag\\
    =&\mkern-6mu: \left[ e_{i}^A \right]_{i=1}^d \lhd \left(g_p^{BA}\right)^{-1}
        \qquad\qquad &&
\end{alignat}
via the thus defined \emph{right action}
\begin{align}\label{eq:right_action_mapsto}
    \lhd:\ \pig( [e_i]_{i=1}^d,\ g \pig)\ \mapsto\ [e_i]_{i=1}^d \lhd g\, :=\,  \Big[ \sum\nolimits_j e_j\, g_{ji} \Big]_{i=1}^d
\end{align}
of group elements on frames.
Note that the inverse in this action in Eq.~\eqref{eq:frame_rightaction} is due to the definition of Eq.~\eqref{eq:gauge_trafo_local_def_21} without inverse.%
\footnote{
    Other conventions might flip the choice of inverses in
    $\psi^B = g^{BA}\psi^A$ and $[e_i^B]_{i=1}^d = [e_i^A]_{i=1}^d\lhd \big(g^{BA}\big)^{-1}\!$.
    An inverse in either of the two equations is necessary to make the left action $\cdot$ on gauges and right action $\lhd$ on frames compatible.
}
One usually refers to the transformation behavior of reference frames as \emph{covariant} transformation while the transformation of gauges and vector coefficients is denoted as \emph{contravariant} transformation; see Appendix~\ref{apx:coordinate_bases}.

Since the transformation behavior of the coefficients in Eq.~\eqref{eq:components_leftaction} and the basis in Eq.~\eqref{eq:frame_rightaction} are inverse to each other they compensate, that is, they leave the tangent vector $v = \sum_i v_i^A e_{i}^A = \sum_i v_i^B e_{i}^B$ invariant:
\begin{align}\label{eq:vector_in_different_bases}
    v\ =\ \sum\nolimits_i v_i^B e_{i}^B
    \ &=\ \sum\nolimits_i v_i^B \sum\nolimits_j e_{j}^A \left(\left(g_p^{BA}\right)^{-1}\right)_{ji} \notag\\
    \ &=\ \sum\nolimits_j \left(\sum\nolimits_i \left(\left(g_p^{BA}\right)^{-1}\right)_{ji} v_i^B\right) e_{j}^A \notag\\
    \ &=\ \sum\nolimits_j v_j^A e_{j}^A \,.
\end{align}
This construction ensures that any calculation is ultimately independent of the chosen gauge, which is usually denoted as \emph{coordinate independence}.
In general, any coordinate representation of a coordinate free object or function is for consistency reasons required to be coordinate independent.

\begin{figure}
    \centering
    \includegraphics[width=.86\textwidth]{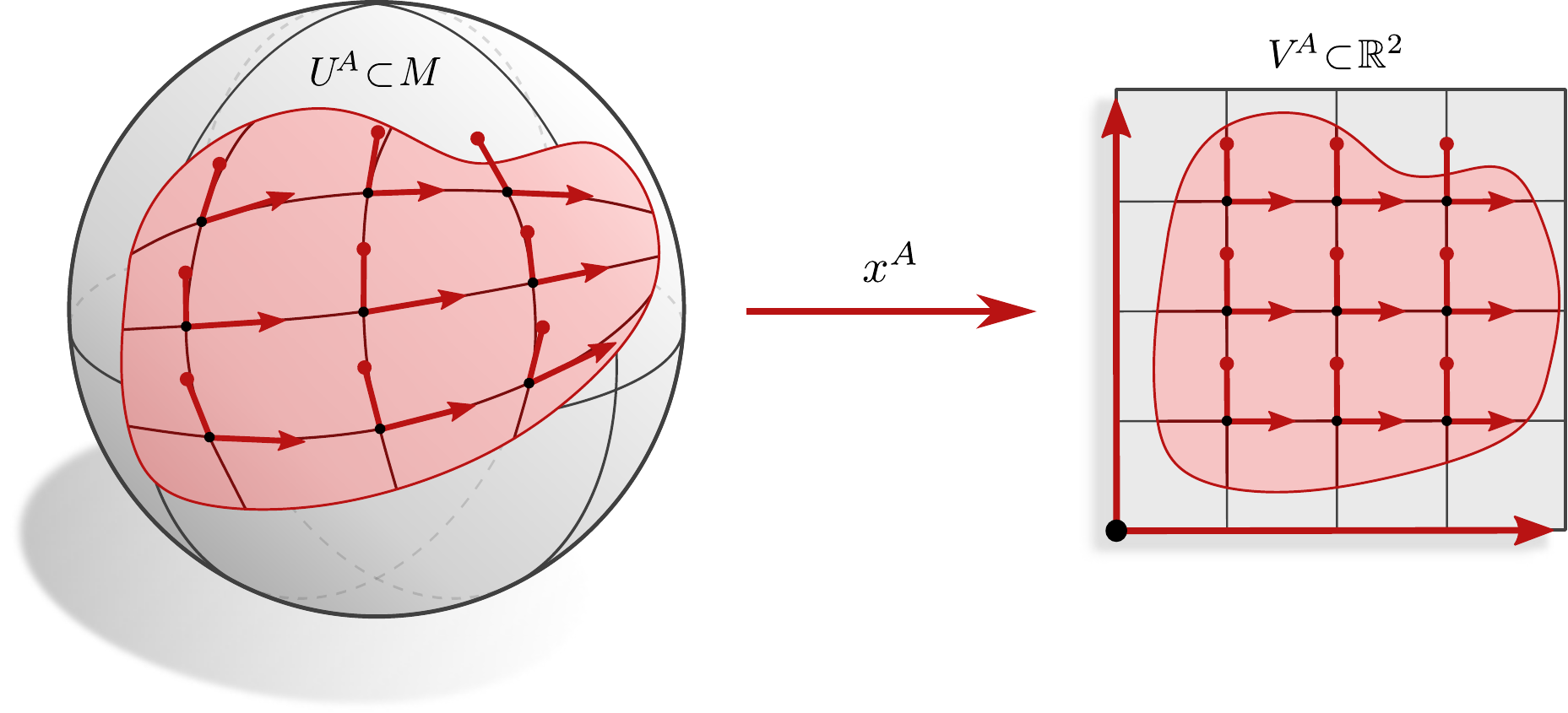}
    \hspace*{4ex}
    \caption{\small
        A \emph{chart} $x^A: U^A \to V^A$ assigns coordinates $V^A \subseteq \R^d$ to regions $U^A \subseteq M$ of the manifold.
        It induces \emph{coordinate bases}
        $\big[\frac{\partial}{\partial x^A_1} \big|_p ,\,\dots\,,\ \frac{\partial}{\partial x^A_d} \big|_p \big]$
        and corresponding gauges $\psi_p^A = \hat{d}x_p^A$ of the tangent spaces $\TpM$ over~$U^A$.
        We will mostly not work with charts but rather refer to points $p\in M$ in a \emph{coordinate free} manner.
        Gauges (frames) are then directly assigned to the tangent spaces instead of being induced.
    }
    \label{fig:sphere_chart}
\end{figure}

For completeness we want to mention that the here presented formalism defines general bases of the tangent spaces, sometimes referred to as \emph{non-coordinate bases} (non-holonomic bases), in terms of local gauges.
A~very popular but less general alternative are \emph{coordinate bases} (holonomic bases)
\begin{align}
    \bigg[\frac{\partial}{\partial x^A_1} \bigg|_p ,\,\dots\,,\ \frac{\partial}{\partial x^A_d} \bigg|_p \bigg] \,,
\end{align}
which are induced by \emph{coordinate charts}
\begin{align}\label{eq:chart_21}
    x^A: U^A \to V^A \subseteq \R^d
\end{align}
of the manifold \cite{nakahara2003geometry}.
The corresponding gauges are given by the the \emph{chart differentials}, that is,
\begin{align}
    \psi_p^A \,=\, \hat{d}x^A_p \,=\, \big( \hat{d}x^A_{p,1} \,,\dots,\, \hat{d}x^A_{p,d} \big)^\top\ :\ \TpM \to \R^d \,.
\end{align}
Gauge transformations coincide in this setting with the \emph{Jacobians}
\begin{align}
    g_p^{BA} \,=\, \frac{\partial x^B}{\partial x^A} \bigg|_{x^A(p)} \,\in\, \GL{d}
\end{align}
of chart transition maps.
An exemplary chart and its induced coordinate bases are visualized in Fig.~\ref{fig:sphere_chart}.
Appendix~\ref{apx:coordinate_bases} discusses the relationship between both formalisms in detail; an overview is given in Table~\ref{tab:coord_charts_gauge_trafos}.

In the remainder of this paper we will mainly work in the gauge formalism, which assigns reference frames immediately to the tangent spaces instead of inducing them from charts.
Exceptions are
the M\"obius convolutions in Section~\ref{sec:mobius_conv},
Euclidean CNNs in Section~\ref{sec:instantiations_euclidean},
log-polar coordinates in Section~\ref{sec:polar_Euc2_logpolar}
and icosahedral CNNs in Section~\ref{sec:spherical_CNNs_icosahedral}.
In all of these cases the manifolds are \emph{locally flat} and the charts are \emph{isometric}, such that they induce \emph{orthonormal frames}.
$\GM$-convolutions on $U^A$ can then be computed in an efficient manner by running Euclidean convolutions with $G$-steerable kernels on the charts' codomains~$V^A$.

\subsubsection{Coordinate independent functions on tangent spaces}
\label{sec:gauges_TpM_functions}

Just as the vectors $v\in \TpM$, \emph{functions} on the tangent spaces are coordinate free, that is, they are defined without referring to any reference frame.
A chosen gauge allows to represent such coordinate free mappings by functions which operate on coefficient vectors in $\R^d$.
Similar to the coefficient vectors, coordinatizations of functions need to transform in a specific way under gauge transformations in order to be consistently defined, i.e. to respect coordinate independence.
We will later apply the here presented concept of expressing coordinate free mappings in terms of local coordinates to define $\GM$-coordinate independent convolutions.

\begin{figure}
    \centering
    \includegraphics[width=\columnwidth]{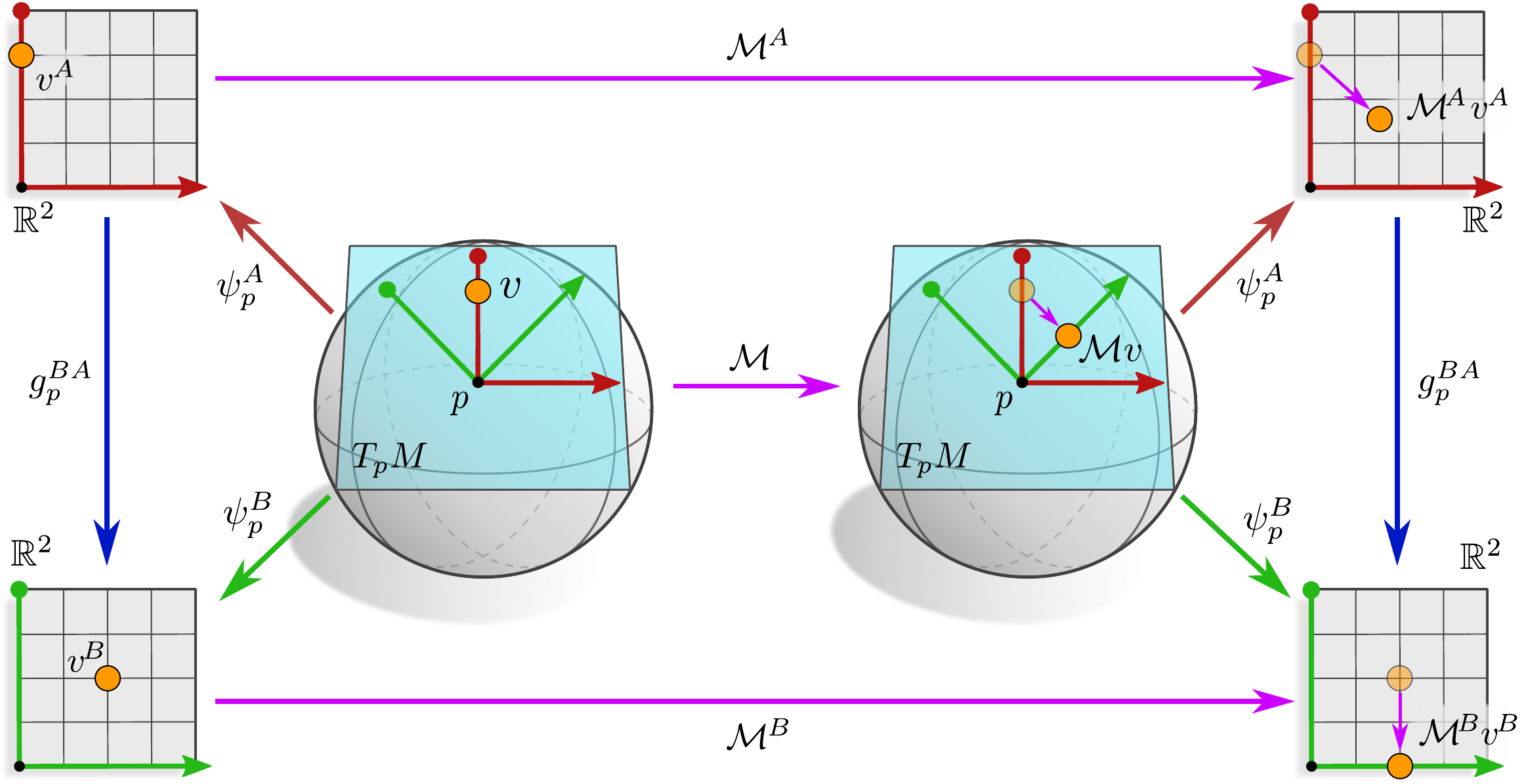}
    \vspace*{.5ex}
    \caption{\small
            Graphical interpretation of the commutative diagram in Eq.~\eqref{eq:linear_map_TpM_diagram}.
            A coordinate free map $\mathcal{M}: \TpM \to \TpM$ can be equivalently represented by functions $\MAA\!: \R^d \to \R^d$ or $\MBB\!: \R^d \to \R^d$ \mbox{relative to different gauges $\psi_p^A$ or~$\psi_p^B$,} respectively.
            These coordinatizations of $\mathcal{M}$ are defined by a pre- and postcomposition with gauges in the domain and codomain, for instance, following the arrows, $\MAA := \psi_p^A \circ \mathcal{M} \circ \big(\psi_p^A\big)^{-1}$.
            As a consequence, gauge transformations $\MBB = g_p^{BA} \MAA \big(g_p^{BA}\big)^{-1}$ between coordinatizations are given by a pre- and postcomposition with transition maps~$g_p^{BA}$ in the domain and codomain.
            \emph{All} quantities and mappings in this work will either be coordinate free (like $\mathcal{M}$) or will be expressed in a coordinate independent way in different gauges (like $\MAA$ and $\MBB$).
            We will therefore need to define (or derive) transformation laws for any quantity and function.
        }
    \label{fig:gauge_trafos_functions}
\end{figure}

As a simple example for a coordinate free operation, let us consider the case of a \emph{linear map}
\begin{align}
    \mathcal{M}:\TpM\to \TpM \,.
\end{align}
Let $v_\text{in}\in \TpM$ be a tangent vector which is by $\mathcal{M}$ being mapped to $v_\text{out} = \mathcal{M} v_\text{in} \in \TpM$.
Linear maps are in numerical implementations usually modeled by \emph{coefficient matrices} which map between \emph{coefficient vectors} relative to some choice of reference frame.
To make this precise, assume some gauge $\psi_p^A$ to be given such that the coordinate free vectors $v_\textup{in}$ and $v_\textup{out}$ in $\TpM$ are represented by coefficient vectors $v_\text{in}^A=\psi_p^A(v_\text{in})$ and $v_\text{out}^A=\psi_p^A(v_\text{out})$ in~$\R^d$.
The linear map $\mathcal{M}$ is in this gauge represented by the matrix%
\begin{align}\label{eq:matrix_trivialization}
    \MAA\ :=\ \psi_p^A \circ \mathcal{M} \circ \big(\psi_p^A\big)^{-1} \ \ \in\ \R^{d\times d}
\end{align}
whose definition is visualized by the commutative diagram below:
\begin{equation}\label{eq:bundle_morphism_onexone}
\begin{tikzcd}[row sep=4.em, column sep=4.em]
    \R^d
            \arrow[rrr, pos=.5, rounded corners, to path={ 
                -- ([yshift=-3.5ex]\tikztostart.south) 
                --node[below]{\small$
                    \MAA
                    $} ([yshift=-3.5ex]\tikztotarget.south) 
                -- (\tikztotarget.south)
                    }]
    & \TpM  \arrow[r, "\mathcal{M}"]
            \arrow[l, "\psi_p^A"']
    & \TpM  \arrow[r, "\psi_p^A"]
    & \R^d
\end{tikzcd}
\end{equation}
The matrix is consistent with the coordinate free mapping since both imply each other:
\begin{align}
    \MAA v^A_\text{in}
    \ &=\ \pig[ \psi_p^A \circ \mathcal{M} \circ \big(\psi_p^A\big)^{-1} \pig] \circ \pig[ \psi_p^A(v_\text{in}) \pig] \notag \\
    \ &=\ \psi_p^A \big( \mathcal{M}\mkern2mu v_\text{in} \big) \notag \\
    \ &=\ \psi_p^A (v_\text{out}) \notag \\
    \ &=\ v_\text{out}^A
\end{align}
Of course one could have represented $\mathcal{M}$ relative to any other choice of gauge $\psi_p^B$.
We know from Eq.~\eqref{eq:components_leftaction} that coefficient vectors in different gauges are related by $v^B = g_p^{BA} v^A$.
Similarly, $\MBB$ relates to $\MAA$ by the gauge transformation
\begin{alignat}{2}\label{eq:matrix_gaugetrafo}
    \MBB
    \ &=&\ \ \psi_p^B \circ\, &\mathcal{M} \circ \big(\psi_p^B\big)^{-1} \notag \\
    \ &=&\ \ \psi_p^B \circ \big(\psi_p^A\big)^{-1} \circ\, &\MAA \circ \psi_p^A \circ \big(\psi_p^B\big)^{-1} \notag \\
    \ &=&\ \ g_p^{BA} &\MAA \big(g_p^{BA}\big)^{-1} \,,
\end{alignat}
which acts here both on the domain and codomain.%
\footnote{
    The transformation of the matrix coefficients via the left and right multiplication with $g^{BA}$ and $\big(g^{BA}\big)^{-1}$, respectively, identify the linear map as a tensor of type $(1,1)$.
}
This transformation law is again seen to be consistent by the mutual transformations canceling out:
\begin{align}
    \MBB v^B_\text{in}
    \ &=\ \pig[ g_p^{BA} \MAA \big(g_p^{BA}\big)^{-1} \pig]\, \pig[ g_p^{BA} v_\text{in}^A \pig] \notag \\
    \ &=\ g_p^{BA} \MAA v_\text{in}^A \notag \\
    \ &=\ g_p^{BA} v_\text{out}^A \notag \\
    \ &=\ v_\text{out}^B
\end{align}
The derived gauge transformations therefore assert that all coordinatized computations are ultimately coordinate independent.
The relations between the coordinate free mapping and its coordinatizations is summarized by the following commutative diagram
\begin{equation}\label{eq:linear_map_TpM_diagram}
\begin{tikzcd}[column sep=50pt, row sep=25pt, font=\normalsize]
    \R^d
        \arrow[dd, "g_p^{BA}\cdot\ "']
        \arrow[rrr, "\MAA"]
    & &[-2ex] &
    \R^d
        \arrow[dd, "\ g_p^{BA}\cdot"]
    \\
    &
    \TpM
        \arrow[ul, "\psi_p^A"]
        \arrow[dl, "\psi_p^B"']
        \arrow[r, "\mathcal{M}"]
    &
    \TpM
        \arrow[ur, "\psi_p^A"']
        \arrow[dr, "\psi_p^B"]
    \\
    \R^d
        \arrow[rrr, "\MBB"']
    & & &
    \R^d
\end{tikzcd}
\end{equation}
which is graphically interpreted in Fig.~\ref{fig:gauge_trafos_functions}.

In practice one can not instantiate the coordinate free linear map $\mathcal{M}$ numerically without referring to a choice of coordinatization.
However, its existence is implied if (and only if) its coordinatizations relate to each other as specified by Eq.~\eqref{eq:matrix_gaugetrafo}, which ensures that the correct transformation behavior of the input and output vector coefficients in Eq.~\eqref{eq:components_leftaction} is preserved.

\subsubsection{Structure groups, \textit{G}-structures and \textit{G}-atlases}
\label{sec:local_G-structure_G-atlas}

We will later on require neural networks to operate in a coordinate independent manner, that is, we demand their inference to be independent from arbitrary choices of reference frames.
This raises the question to which extent the choice of reference frames on a manifold is arbitrary.
In the previous Sections~\ref{sec:gauges_gauge_trafos} and~\ref{sec:gauges_TpM_functions} we allowed for any possible choice of gauge or reference frame, which were thus related by general $\GL{d}$-valued gauge transformations.
In many applications the manifold does, however, come with additional structure which allows to distinguish a \emph{preferred subset of reference frames} or gauges, whose transition functions take values in a \emph{reduced structure group} $G\leq\GL{d}$.
Such geometric structures -- or rather the subsets of preferred reference frames themselves, which encode equivalent information -- are denoted as \emph{$G$-structures}.

$G$-structures are best understood by considering some specific examples.
The following list gives such examples, classified by their structure group $G \leq \GL{d}$:
\begin{itemize}[leftmargin=9.4ex]
    \item[$\O{d}$:]
        Consider the \emph{metric} structure of a Riemannian manifold, which allows to measure distances and angels, and therefore to distinguish orthonormal frames, that is, those frames that satisfy $\eta(e_i,e_j) = \delta_{ij}$ for any $i,j=1,\dots,d$.
        Correspondingly, a Riemannian metric allows to talk about isometric gauges $\psi_p^A$, which identify the metric of $\R^d$ with that of~$\TpM$, i.e. which satisfy $\eta(v,w) = \langle \psi_p^A(v),\, \psi_p^A(w) \rangle_{\R^d}$ for any $v,w \in \TpM$.
        Since orthonormal frames and isometric gauges are defined up to rotations and reflections, any gauge transformation between them will take values in the orthogonal group~$\O{d}$, which is that subgroup of $\GL{d}$ that preserves angles and distances.
    \item[$\operatorname{GL}^+(d)$:]
        Similarly, an \emph{orientation} of the manifold distinguishes left-handed from right-handed frames and orientation preserving gauges from orientation reversing gauges.
        Gauge transformations between frames of a given handedness take values in $\operatorname{GL}^+(d)$, that is, that subgroup of $\GL{d}$ which preserves orientations.
    \item[$\SO{d}$:]
        Together, a given \emph{metric and orientation} specify orthonormal frames of a certain handedness.
        Gauge transformations between such frames are guaranteed to lie in the subgroup~$\SO{d}$ of $\GL{d}$.
    \item[$\{e\}$:]
        A \emph{globally smooth frame field} defines an $\{e\}$-structure on $M$.
        In this case there is only one single distinguished frame at each position, such that gauge transformations lie in the trivial group ${\{e\} \leq \GL{d}}$.
    \item[$\GL{d}$:]
        If no additional structure is imposed, \emph{any} reference frame of the tangent spaces is equally valid.
        Gauge transformations are in this case general invertible liner maps in $\GL{d}$ and the corresponding $G$-structure is just the frame bundle $\FM$.
\end{itemize}

\begin{samepage}
The common theme in those motivating examples is that they are all defined by
\begin{enumerate}
    \item a (spatially smoothly varying) subset of distinguished reference frames,
    \item a corresponding subset of preferred gauges and
    \item a subgroup $G \leq \GL{d}$ of gauge transformations which preserve the distinguished notion of\\ frames and gauges.
\end{enumerate}
\end{samepage}
Such smoothly varying subsets of reference frames are denoted as \emph{$G$-structures}~$\GM$ on~$M$
and the group~$G$ is denoted as (reduced) \emph{structure group}
-- see Section~\ref{sec:G_associated_bundles} for a more rigorous definition.%
\footnote{
    Formally, $\GM$ is defined as a principal $G$-subbundle of frame bundle $\FM$, which is a principal $\GL{d}$-bundle.
}
The process of specifying a $G$-structure is known as a \emph{reduction of the structure group} from $\GL{d}$ to~$G$.
An atlas ${\mathscr{A}^G = \big\{\! \big(U^X, \psi^X\big) \!\big\}_{X\in \mathfrak{X}}}$ is denoted as $G$\emph{-atlas} if all of its transition functions
\begin{align}\label{eq:transition_fct_local_def_21_G_atlas}
    g^{BA}\!:\, U^A\cap U^B\to G, \quad p \mapsto g_p^{BA} := \psi_p^B \circ \left(\psi_p^A\right)^{-1}
\end{align}
lie in a reduced structure group~$G \leq \GL{d}$ (cf. Eq.~\eqref{eq:transition_fct_local_def_21}).
The relation between reference frames and gauges in Eq.~\eqref{eq:framefield_gauge_equivalence} implies that any $G$-atlas encodes a corresponding $G$-structure.

Multiple choices of $G$-structures may exist for a given structure group~$G$.
To connect to the examples above:
different Riemannian metrics specify different subsets of reference frames as being orthonormal, that is, they correspond to different $\O{d}$-structures $\OM$.
A choice of metric is therefore equivalent to a choice of $\O{d}$-structure.
Similarly, different choices of orientations of an orientable manifold specify a different set of frames as being right-handed.
The two possible choices of orientations therefore correspond to two possible choices of $\operatorname{GL}^+(d)$-structures $\operatorname{GL}^+\mkern-5muM$.
$\SO{d}$-structures $\SOM$ may differ in both the choice of orientation and metric.
A further example are $\{e\}$-structure $\eM$.
They do not allow for (non-trivial) gauge transformations and therefore correspond to choices of smooth, global frame fields on $M$.
Table~\ref{tab:G_structures} gives more examples of structure groups~$G$ and the corresponding $G$-structures.

\begin{table}
    \centering
    \renewcommand\arraystretch{1.1}
    \small
    \begin{tabular}{cll}
       \toprule
       structure group $G\leq\GL{d}$ & $G$-structure $\GM$                      & equivalent structure on $M$       \\[.25ex]
       \midrule
       $\operatorname{GL}^+(d)$      & positively oriented frames               & orientation of $M$                \\
       $\operatorname{SL}(d)$        & unit volume frames                       & volume form                       \\
       $\operatorname{CO}(d)$        & conformal frames                         & ---                                \\
       $\operatorname{Sp}(d)$        & symplectic frames                        & ---                                \\
       $\operatorname{O}(d)$         & orthonormal frames                       & Riemannian metric                 \\
       $\operatorname{O}(d-n,\,n)$   & pseudo-orthonormal frames                & pseudo-Riemannian metric          \\
       $\operatorname{SO}(d)$        & positively oriented orthonormal frames   & Riemannian metric + orientation   \\
       $\{e\}$                       & parallelization (global frame field)     & ---                                \\[.25ex]
       \bottomrule
    \end{tabular}
    \vspace*{2ex}
    \caption{
        Examples of $G$-structures $\GM$ on $M$ and their corresponding, reduced structure groups~$G\leq\GL{d}$~\cite{kobayashi1972transformation}.
        A~$G$-structure is defined as a smoothly varying subset of reference frames (a principal $G$-subbundle of the frame bundle $\FM$), where the frames of any tangent space are mutually related by $G$-valued gauge transformations.
        While this is a quite abstract definition, it allows to view many geometric structures on~$M$ in a unified way.
        For instance, a Riemannian metric on~$M$ allows to distinguish orthonormal frames.
        Conversely, a specification of orthonormality uniquely implies a metric.
        A Riemannian metric and an orthonormal structure are thus equivalent to each other.
        Similarly, there is a one-to-one correspondence between volume forms and unit volume frames.
        Note that a choice of structure group $G$ does not uniquely specify a $G$-structure.
        For example, different Riemannian metrics could be chosen as $\O{d}$-structure, different volume forms as $\operatorname{SL}(d)$-structure or different global frame fields as $\{e\}$-structure.
        Coordinate independent CNNs are designed to respect a given $G$-structure -- which particular structure this is depends on the learning task.
    }
    \label{tab:G_structures}
\end{table}

A reduction of the structure group to $G$, i.e. the existence of a $G$-structure, might be obstructed by the topology of the manifold.
This implies that there is an \emph{``irreducible'' structure group} beyond which the ambiguity of reference frames can not be resolved without violating the smoothness (or even continuity) assumption of the $G$-structure.
For example, the M\"obius strip in is non-orientable, which means that it does not admit a globally consistent, smooth definition of frame handedness and thus $\{e\}$-structure (globally smooth frame field).
As~visualized in Fig.~\ref{fig:mobius_conv_gauges}, a $G$-atlas of gauges covering the M\"obius strip will unavoidably require a reflection in one of the transition maps, implying an irreducible structure group $G=\Flip$.
Coordinate independent CNNs on the M{\"o}bius strip is therefore required to be at least reflection equivariant.
Similarly, the structure group of the sphere can not be reduced further than ${G=\SO2}$.
Smooth spherical CNNs are thus necessarily based on locally rotation equivariant kernels.

Note that \emph{any} (differentiable) manifold comes with \emph{some} $G$-structure.
For instance, a raw differentiable manifold has a $\GL{d}$-structure (containing any possible frame), a Riemannian manifold an $\O{d}$-structure and $\R^d$ is canonically equipped with an $\{e\}$-structure, visualized in Fig.~\ref{fig:frame_field_automorphism_1}.
We will therefore without loss of generality refine the term ``coordinate independence'' to $\GM$-\emph{coordinate independence}, i.e. the independence w.r.t. choices of reference frames in the $G$-structure given on~$M$.
Throughout this work we will assume that gauges are part of some $G$-atlas
\begin{align}\label{eq:G_atlas_dfn}
    \mathscr{A}^G \,=\, \big\{\! \big(U^X, \psi^X\big) \!\big\}_{X\in \mathfrak{X}}
    \quad\ \textup{such that} \quad
    g_p^{BA} \in G
    \quad \textup{for any}\ \ \ \psi_p^A, \psi_p^B \in \mathscr{A}^G,\ \ p\in U^A\cap U^B \,,
\end{align}
corresponding to the given $G$-structure.
Any quantity or function can be expressed relative to any gauge from this atlas%
\footnote{
    This is a non-trivial statement since not any quantity can be expressed relative to arbitrary $\GL{d}$-related reference frames.
    For instance, the feature fields, introduced in Section~\ref{sec:feature_fields}, will only admit $G$-valued gauge transformations and are therefore only defined relative to the preferred frames in~$\GM$.
    As an intuitive example, consider the feature vectors of a conventional (non-equivariant) CNN on $\R^d$, which are extracted relative to the canonical $\{e\}$-structure of $\R^d$ and do \emph{not} carry information about the kernel responses relative to other reference frames.
},
and the coordinatizations in different gauges relate uniquely by some $G$-valued gauge transformation.
Guaranteeing the coordinate independence of all constructions, they will always correspond to some coordinate-free counterparts, in terms of which we will formulate the global theory in Sections~\ref{sec:bundles_fields}, \ref{sec:gauge_CNNs_global} and~\ref{sec:isometry_intro}.

%% file: chapters/32_feature_fields.tex

\subsection{Coordinate independent feature vector fields}
\label{sec:feature_fields}

The feature spaces of coordinate independent CNNs are spaces of feature vector fields.
Similar to the case of tangent vectors coefficients, the numerical coefficients of feature vectors are required to transform consistently under gauge transformations.
The specific transformation law (group representation) of a feature field does hereby specify its field type -- common examples are scalar fields, tangent vector fields, general tensor fields, regular feature fields or irrep fields.
Section~\ref{sec:individual_fields} introduces such feature fields and their transformation laws.
In Section~\ref{sec:stacked_fields}, we briefly define coordinate independent feature spaces.
Similar to the definition of the feature spaces of conventional CNNs as stacks of multiple feature maps, the feature spaces of coordinate independent CNNs consist of multiple independent feature fields.

\subsubsection{Individual feature vector fields}
\label{sec:individual_fields}
Convolutional feature fields assign a feature vector, encoding information inferred from a local neighborhood of the input signal, to each point of the manifold.
The spatial accumulation of information is performed by a convolutional kernel which is \emph{measuring feature fields} in its surrounding \emph{relative to its local reference frame}.
We are thus assuming a gauge $\psi^A$ which specifies the kernel alignments on a neighborhood $U^A$.
Relative to this gauge the kernel will yield a smooth local field of responses (observations)
\begin{align}
    f^A:U^A\to\R^c \,,
\end{align}
given by a $c$-dimensional numerical feature vector $f^A(p)$ at each position $p\in U^A$.
Assume a second response field $f^B:U^B\to\R^c$, inferred relative to gauge $\psi^B$ on $U^B$, to be given.
Since the response of a kernel depends in general on its alignment, it is to be expected that $f^A$ and $f^B$ do not agree on the overlap $U^A\cap U^B.$
Without further restrictions the responses of a convolution kernel will be arbitrarily gauge dependent.

\begin{figure}
    \centering
    \includegraphics[width=.75\columnwidth]{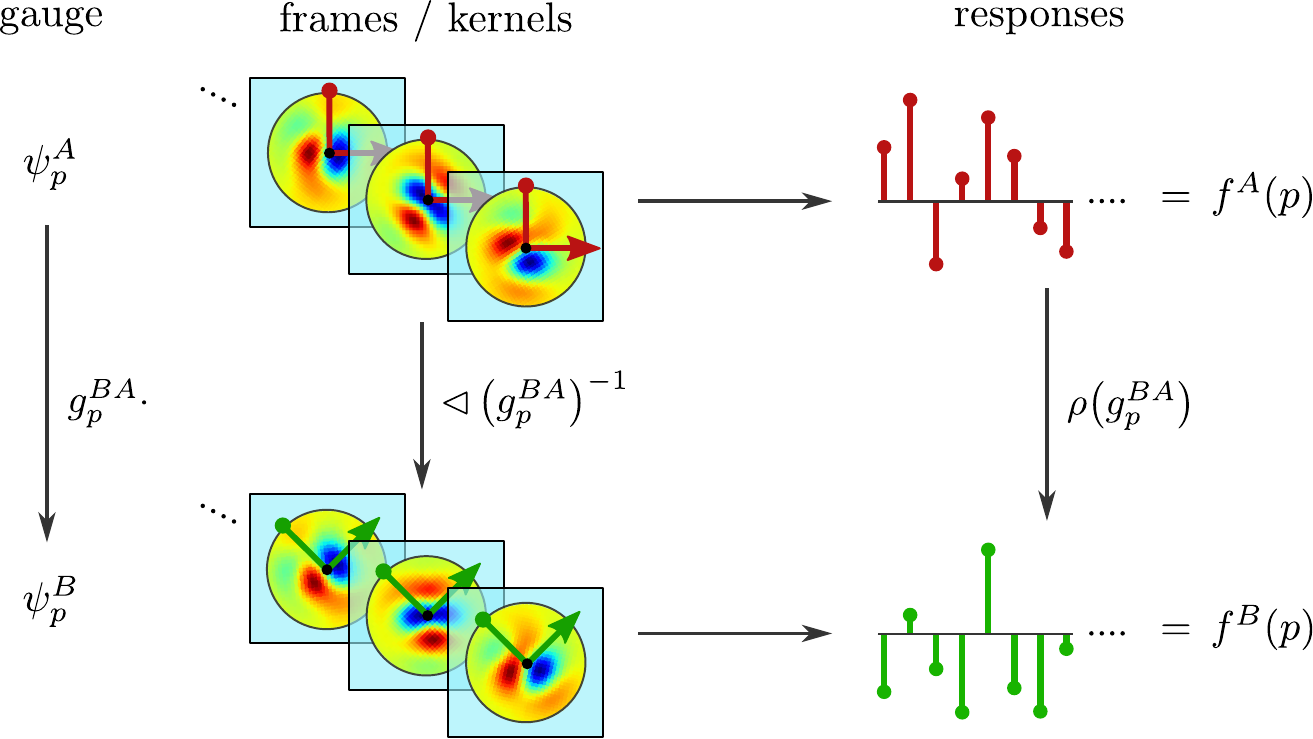}
    \vspace*{1ex}
    \caption{\small
        The numerical responses $f^A(p) \in\R^c$ and $f^B(p) \in\R^c$ of kernels that are oriented according to different frames do in general not coincide.
        In order to represent numerical coefficients of the \emph{same coordinate independent feature vector} relative to the chosen gauge, they are required to be related by gauge transformations $\rho(g_p^{BA})$ if the gauges are related by~$g_p^{BA}$.
        As derived in Section~\ref{sec:gauge_CNNs_local}, this requirement imposes a gauge equivariance constraint on the convolution kernels.
        }
    \label{fig:gauge_trafos_feature_vector}
\end{figure}

\begin{minipage}{\textwidth}
The \emph{principle of covariance}, proposed by Albert Einstein \cite{Einstein1916German,Einstein1916English}, states that:
\vspace*{1.ex}
\begin{center}
    \it
    ``Universal laws of nature are to be expressed by equations which hold good for all systems
    \\
    of coordinates, that is, are covariant with respect to any substitutions whatever.''
\end{center}
\vspace*{1.ex}
\end{minipage}
We believe that a similar principle should hold in geometric deep learning as well, that is, the inference should be independent from any arbitrariness in the choice of reference frames.
Given that this arbitrariness in coordinatizations is precisely captured by the given $G$-structure $\GM$, this requires in particular that \emph{features should be $\GM$-coordinate independent} geometric objects.%
\footnote{
    In this point we deviate from Einstein's \emph{general} covariance, which always considers $\GL{d}$-valued gauge transformations (corresponding to diffeomorphism covariance).
    His setting is in our formulation included for $G=\GL{d}$, however, we keep the assumed structure group flexible since most applications will assume a reduced structure group.
}
We thus design convolution kernels such that their responses $f^A$ and $f^B$ encode fields of \emph{feature vector coefficients} which \emph{represent a coordinate free feature vector field $f$} locally in different gauges.
A collection of such numerical coefficient fields $f^X$, expressed relative to a $G$-atlas of gauges $\psi^X$ on neighborhoods $U^X$ covering $M$, is equivalent to the global, coordinate free feature field $f$ on~$M$.

\begin{figure}
    \centering
    \includegraphics[width=.83\columnwidth]{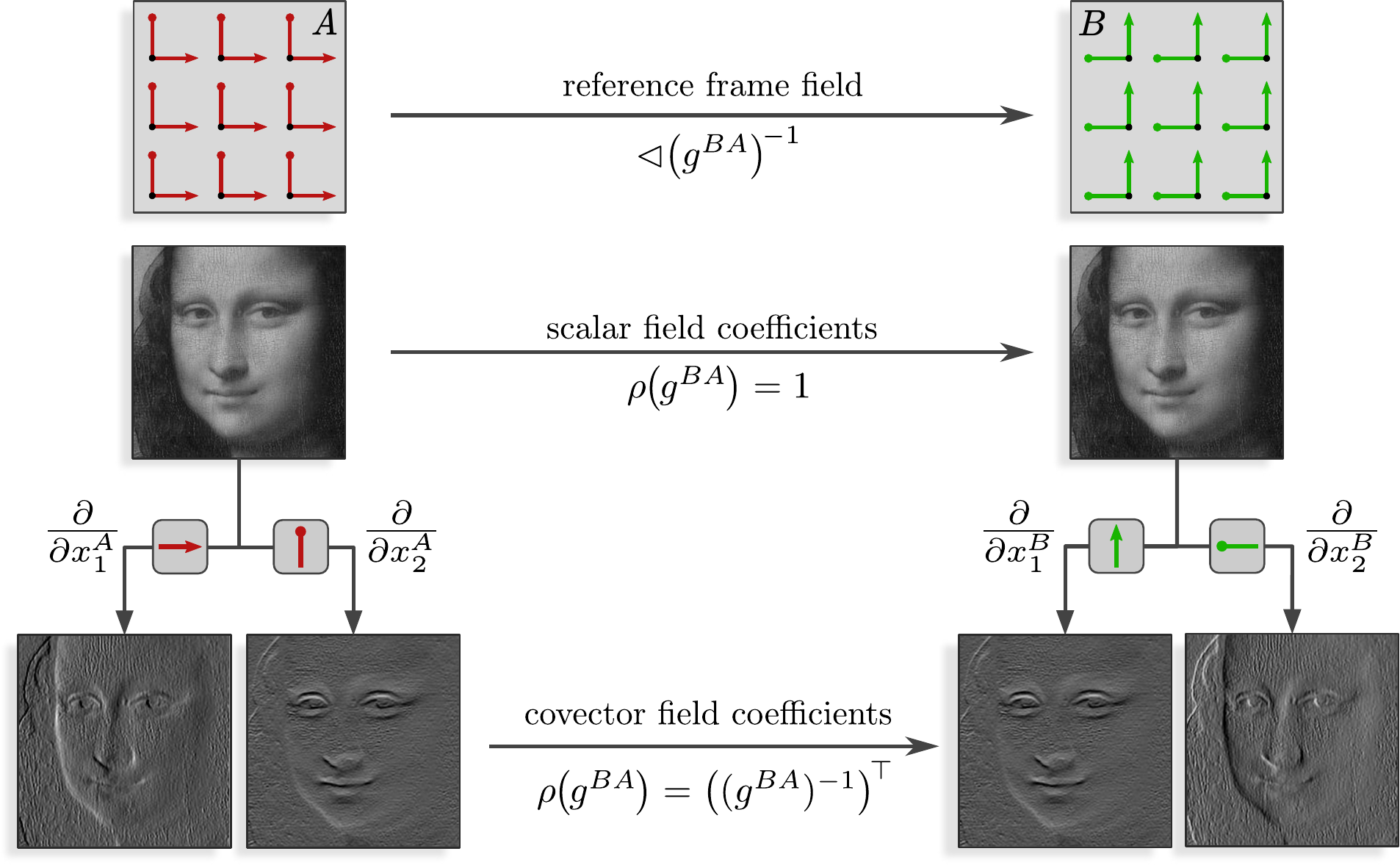}
    \caption{\small
        Examples of feature coefficient fields on $M=\R^2$ from classical image processing.
        \ \emph{Top}:
        For simplicity we assume a ``parallel'' frame field and consider the same gauge transformation, a rotation by $\pi/2$, at each point $p\in M$.
        \ \emph{Middle}:
        The intensity values of a grayscale image are independent from the choice of reference frames.
        They are therefore modeled by scalar fields, characterized by the trivial representation $\rho(g)=1\ \forall g\in G$.
        \ \emph{Bottom}:
        The two coefficient channels of a gradient image are calculated from a scalar image by taking the derivatives along the frame axes -- they are therefore gauge dependent.
        Gradient images w.r.t. different gauges are related by the group representation $\rho(g)=(g^{-1})^\top$ and are therefore identified as covector fields (tensor fields of type $(0,1)$ or 1-forms).
        For~the visualized rotation by $\pi/2$ this leads to a new first channel $(\nicefrac{\partial}{\partial x_1^B})$ equivalent to the old second channel $(\nicefrac{\partial}{\partial x_2^A})$ and a new second channel $(\nicefrac{\partial}{\partial x_2^B})$ equivalent to the negative old first channel $(\nicefrac{\partial}{\partial x_1^A})$.
        Relative to their respective reference frames, both coefficient fields encode the same (coordinate free) gradient field.
        The description is therefore automatically coordinate independent.
    }
    \label{fig:feature_field_gradient}
\end{figure}

In order for this coordinate free feature field to be well defined, i.e. $\GM$-coordinate independent, the local coefficient fields (or kernel responses) are required to be consistently stitched together via $G$-valued transition maps.
They must therefore transform in a principled manner under gauge transformations.
Since we are dealing with feature vector spaces, these transformations are typically taken to be linear, that is, they are modeled by linear \emph{group representations}
\begin{align}\label{eq:group_representation}
    \rho:G\to\GL{c}
\end{align}
of the structure group $G\leq\GL{c}$, which operates on $\R^c$ and satisfies $\rho(gh)=\rho(g)\rho(h)\ \forall\ g,h\in G$.%
\footnote{\label{footnote:repr_group_homomorphism}
    This condition ensures that representations are group homomorphisms, i.e. maps which respect the group structure of~$G$.
    The actions of the structure group on the tangent space and feature vector coefficient spaces are thus compatible.
}
Similar to the transformation of tangent vector coefficients in Eq.~\eqref{eq:components_leftaction}, the feature vector coefficients are then defined to transform under a $G$-valued gauge transformation $g_p^{BA}=\psi_p^B\!\circ\!\left(\psi_p^A\right)^{-1}$ like
\begin{align}\label{eq:gauge_trafo_features}
  f^B(p)\ :=\ \rho\big( g_p^{BA}\big) f^A(p) \,,
\end{align}
where $p\in U^A\cap U^B$; see Fig.~\ref{fig:gauge_trafos_feature_vector} for a visualization.
Being constructed to transform synchronously, the spaces of reference frames, tangent vector coefficients and feature vector coefficients are said to be $G$-\emph{associated} to each other.
Note that the construction via a $G$-representation $\rho$ does in general \emph{not} describe $\GL{d}$-valued gauge transformations, i.e. fully coordinate independent features.
The extracted feature vectors will therefore only have a well defined expression relative to the frames in the considered $G$-structure $\GM$, which is captured by the term ``$\GM$-coordinate independence''.

Different choices of representations $\rho_i$ yield different \emph{types} of feature fields as exemplified in Fig.~\ref{fig:feature_field_gradient}.
For instance, the trivial representation, $\rho(g)=1\ \forall g\in G$, describes the transformation behavior of \emph{scalar fields} $s^A(p)\ \mapsto\ s^B(p)\ :=\ 1\cdot s^A(p)$, whose numerical coefficients are invariant under gauge transformations.
Example of scalar fields are grayscale images, temperature fields, pressure fields or probability distributions on $M$.
The coefficients of \emph{tangent vector fields} transform like $v^A(p)\ \mapsto\ v^B(p)\ :=\ g_p^{BA}v^A(p)$ and correspond therefore to the group representation $\rho(g)=g$.
Examples for vector fields include optical flow or wind velocity fields.
More general tensor fields of type $(r,s)$ are described by tensor product representations $\rho(g) = \otimes^s\left(g^{-1}\right)^{\!\top} \otimes^r g$.
They model for instance diffusion tensor images, electromagnetic field tensors or stress tensors.
A common choice for discrete structure groups are \emph{regular representations} which realize the finite set of group operations by permutation matrices.
Regular representations arise as exact symmetries of crystal lattices, spin lattices or pixel grids~\cite{Cohen2016-GCNN,Hoogeboom2018-HEX,winkels3DGCNNsPulmonary2018,Worrall2018-CUBENET,gaugeIco2019}.
In addition, they are commonly used as discrete approximation of continuous structure groups, e.g. cyclic groups $C_N\leq\SO2$ to approximate continuous rotations~\cite{Weiler2018SFCNN,bekkers2018roto,Weiler2019_E2CNN,bekkers2020bspline,Marcos2017-VFN}.
They are of great practical relevance since they describe the transformation of the features of group convolutional networks~\cite{Cohen2016-GCNN}.
Feature fields which transform under \emph{irreducible representations} (irreps) were investigated in~\cite{Worrall2017-HNET,3d_steerableCNNs,Thomas2018-TFN,Kondor2018-NBN,anderson2019cormorant,Weiler2019_E2CNN,jiang2019spherical}.%
\footnote{
\label{footnote:feature_field_irrep_decomposition}
    By the Peter-Weyl Theorem, any unitary representation of a compact group can via a change of basis be decomposed into a direct sum of irreps.
    This implies that any \emph{linear} neural network operation between general representations can (after a change of basis) be understood in terms of operations between irreps~\cite{Weiler2019_E2CNN,lang2020WignerEckart}.
    In contrast, the specific choice of representation, i.e. the change of basis relative to the irreps contained in it, \emph{does} matter for any \emph{non-linear} network layer.
}
A more detailed overview and an extensive benchmark of different field types or representations in deep learning was presented in~\cite{Weiler2019_E2CNN}.

For completeness we want to mention that coordinate free feature vector fields are formally defined as smooth sections $f\in\Gamma(\A)$ of a \emph{feature vector bundle} $\A\piAarrow M$ which is associated to the $G$-structure $\GM$ and has the feature vector coefficient spaces $\R^c$ as typical fibers.
The coefficient vectors $f^A(p)$ and $f^B(p)$ in $\R^c$ are local trivializations of a coordinate free feature vector $f(p) \in \A_p \cong \R^c$, and are similarly defined as the coefficients $v^A=\psi_p^A(v)$ and $v^B=\psi_p^B(v)$ of a tangent vector $v\in \TpM$.
Note that, while being isomorphic, the feature spaces $\A_p \cong \A_q$ at different points $p\neq q$ of~$M$ are distinct from each other, such that their elements can not be summed together.
The parallel transporters, discussed in Sections~\ref{sec:transport_local} and~\ref{sec:bundle_transport}, provide isomorphisms between different feature vector spaces, which allows the summation of features (after transporting them into the same vector space).
Since these definitions are quite technical, we skip their details for now and refer the interested reader to Section~\ref{sec:G_associated_bundles}.

\subsubsection{Stacked feature fields and coordinate independent feature spaces}
\label{sec:stacked_fields}

\begin{wrapfigure}[12]{r}{0.365\textwidth}
    \vspace*{-4.25ex}
    \hfill
    \includegraphics[width=.95\linewidth]{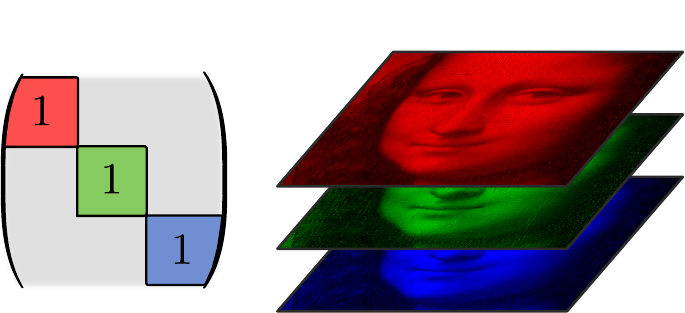}%
    \vspace*{.0ex}
    \captionsetup{width=.35\textwidth}
    \caption{\small
        The three color channels of an RGB image are identified as scalar fields, the full feature space therefore transforms according to $\rho(g) = (\mkern1.5mu 1 \mkern1.5mu) \oplus (\mkern1.5mu 1 \mkern1.5mu) \oplus (\mkern1.5mu 1 \mkern1.5mu)$.
        }
    \label{fig:feature_space_RGB}
\end{wrapfigure}
\begin{figure}
    \centering
    \includegraphics[width=.94\textwidth]{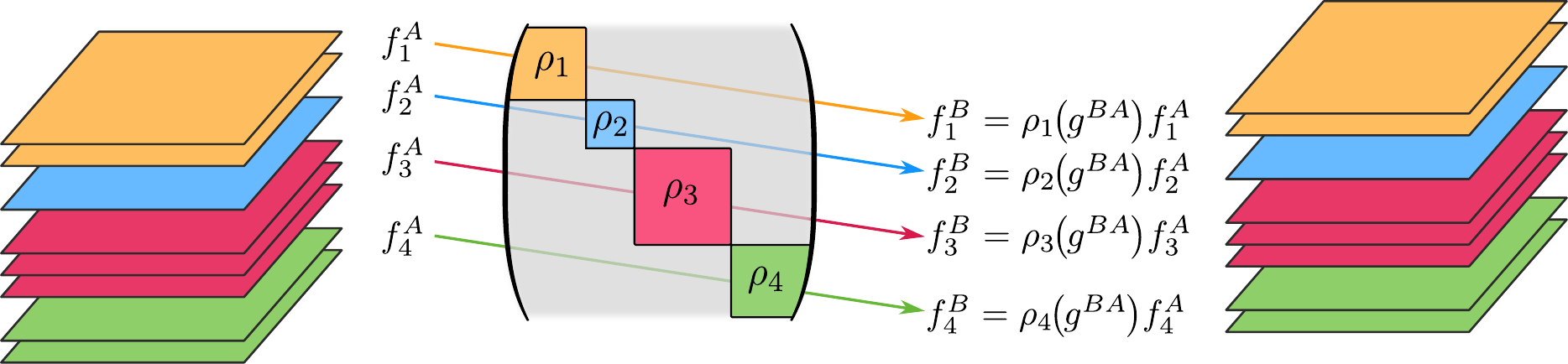}
    \vspace*{0ex}
    \caption{\small
        A full feature space comprises multiple individual feature fields $f_i$ of potentially different types $\rho_i$ and dimensionalities $c_i$.
        Via a gauge $\psi^A$, it is locally represented by coefficient fields $f_i^A: U^A \to \R^{c_i}$.
        The coefficient fields in another gauge $\psi_B$ are related via a gauge transformation $f_i^B = \rho_i\big(g^{BA}\big)f_i^A$.
        The coefficients of each individual field transform independently, the representation modeling the whole feature space is thus given by the direct sum, here $\bigoplus_i \!\rho_i = \rho_1 \oplus \rho_2 \oplus \rho_3 \oplus \rho_4$.
    }
    \label{fig:feature_spaces_oplus}
\end{figure}

%
The feature spaces of conventional CNNs consist of multiple feature maps.
In analogy, we define the feature spaces of coordinate independent CNNs to comprise multiple feature fields $f_i$ of potentially different types $\rho_i$ and dimensionalities $c_i$.
A full field of activations of a feature space of a coordinate independent CNN is therefore defined as the \emph{direct sum}%
\footnote{
    The direct sum $\oplus$ of vectors $f_i(p)$ can be thought of as ``stacking'' these into a concatenated vector.
    Consistently with this, the direct sum of representations $\rho_i$ can be thought of as building a block diagonal matrix containing the $\rho_i$ as blocks; see Figs.~\ref{fig:feature_space_RGB} and~\ref{fig:feature_spaces_oplus}.
}
\begin{align}\label{eq:feature_field_full}
    f = \scalebox{1.1}{$\bigoplus$}_i f_i
\end{align}
of the individual fields.
Every feature map of a conventional CNN encodes the position of a particular feature and transforms independently when the network's input is shifted.
The individual feature fields $f_i$ of $\GM$-coordinate independent CNNs encode both the position and the $G$-pose of a feature.
In contrast to conventional feature maps, their coefficient fields, for instance $f_i^A$, are additionally guaranteed to transform independently from each other under gauge transformations as specified by their type ${\rho_i:G\to\GL{{c_i}}}$.
A~local numerical representation $f^A = \bigoplus_i f_i^A$ of the full feature field in Eq.~\eqref{eq:feature_field_full} transforms therefore according to the direct sum of the individual representations, that is,
\begin{align}
    \rho = \scalebox{1.1}{$\bigoplus$}_i \rho_i \,.
\end{align}
The independent transformation of individual fields under $\rho$, visualized in Fig.~\ref{fig:feature_spaces_oplus}, is clear by construction:
\begin{align}
    \rho(g)f^A
    \,=\, \left(\scalebox{1.1}{$\bigoplus$}_i \rho_i(g)\right) \! \left(\scalebox{1.1}{$\bigoplus$}_i f_i^A\right)
    \,=\, \scalebox{1.1}{$\bigoplus$}_i \! \left(\rho_i(g) f_i^A\right)
\end{align}

As a practical example of a coordinate independent feature space consisting of multiple fields consider an RGB image as depicted in Fig.~\ref{fig:feature_space_RGB}.
Like the grayscale image in Fig.~\ref{fig:feature_field_gradient}, the individual color channels encode intensity values which are invariant under gauge transformations.
The full RGB image is therefore to be identified with three scalar fields, each of which ``transforms'' independently under the trivial representation.
Not all individual feature fields need to be of the same type $\rho_i$.
For instance, in a weather forecasting application the input signal might consist of scalar fields encoding features like temperature or pressure and vector fields like wind speeds.
A description as $\rho_i$ fields of corresponding types ensures the geometrically correct processing of such data.
While the field types $\rho_i$ of a network's input and output are typically given by the learning task, the field types used in hidden layers are chosen by the user as a hyperparameter similar to the choice of channels for a conventional CNN.

%% file: chapters/33_local_transport.tex

\subsection{Parallel transport of feature vectors}
\label{sec:transport_local}

The kernels of convolutional networks accumulate features from all points $q$ in a neighborhood around each point $p$ of the manifold.
Since features at different points live in different feature vector spaces and are expressed relative to different gauges, they need to be \emph{parallel transported} along some path $\gamma$ from $q$ to $p$ before they can be processed further.
We first discuss the transport of tangent vectors, which is formalized by a parallel transport map
\begin{align}
    \mathcal{P}_\gamma: \TqM \to \TpM \,.
\end{align}
This transporter is often computed from the canonical Levi-Civita connection of the manifold, however, it might in some applications correspond to an alternative ($G$-compatible) connection, as further discussed below and in our literature review in Part~\ref{part:literature_review}.
A transporter of ($G$-associated) feature vectors follows from that of the tangent vectors if the transport is $G$-compatible.

\subsubsection{Tangent vector transporters}

\begin{figure*}
    \centering
    \begin{subfigure}[b]{0.5\textwidth}
        \centering
        \includegraphics[width=.95\textwidth]{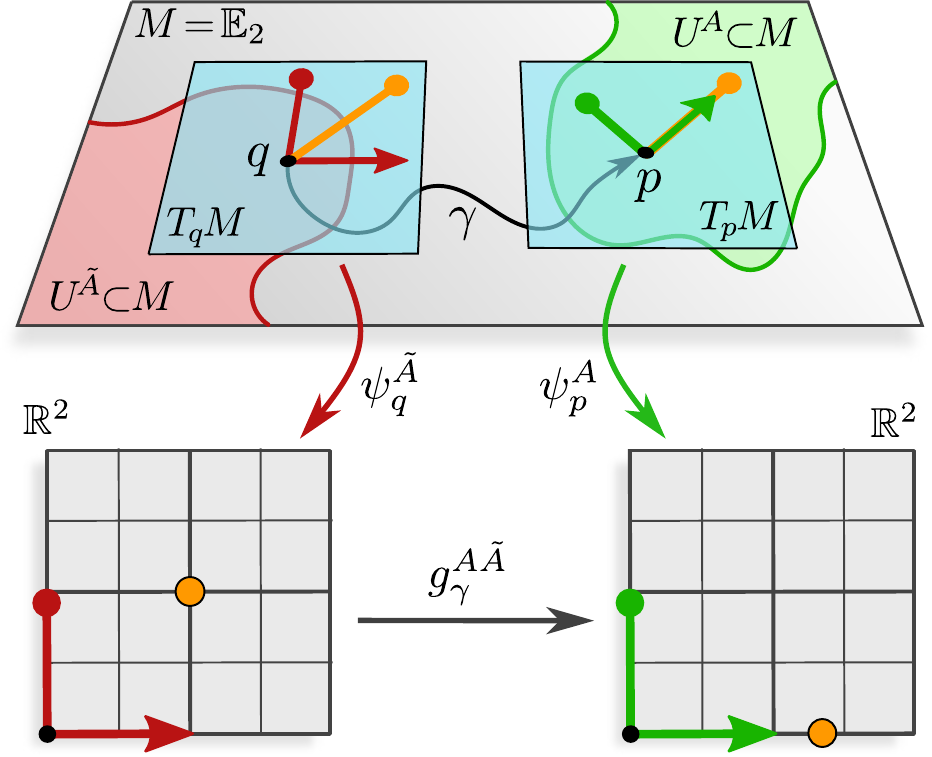}
        \vspace*{1ex}
        \caption{\small
            Parallel transport and its coordinatization on a flat space.
        }
        \label{fig:transport_flat}
    \end{subfigure}
    \hfill
    \begin{subfigure}[b]{0.4\textwidth}
        \centering
        \includegraphics[width=\textwidth]{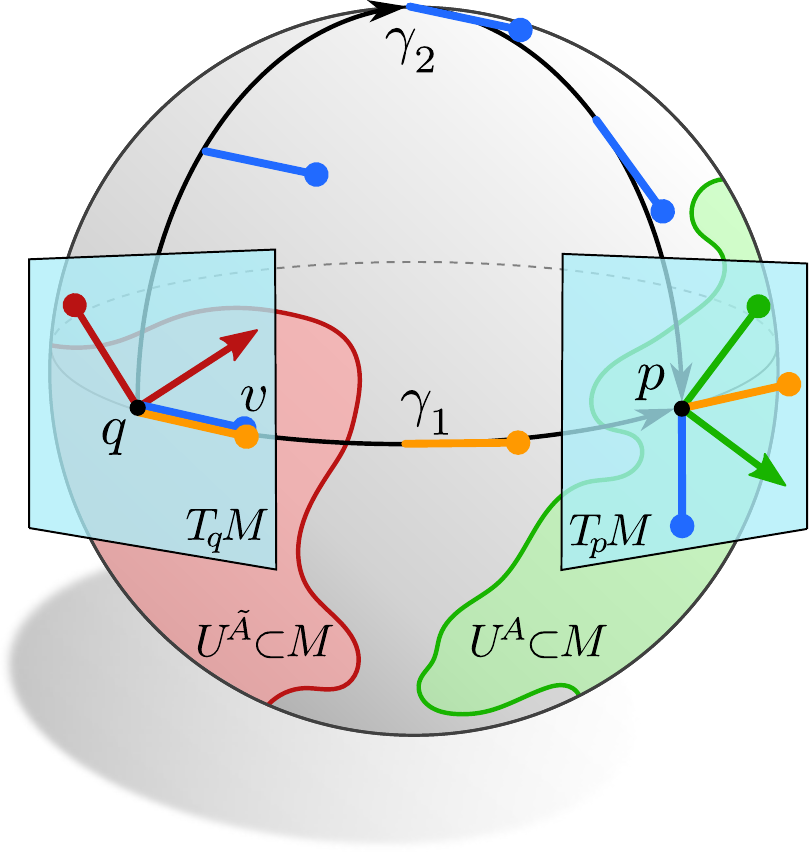}
        \vspace*{-2ex}
        \caption{\small
            Parallel transport on the 2-sphere $S^2$.
        }
        \label{fig:transport_sphere}
    \end{subfigure}
    \caption{\small
        Parallel transport of tangent vectors $v\in \TqM$ at $q$ to $\mathcal{P}_\gamma v \in \TpM$ at $p$.
        Fig.~\ref{fig:transport_flat} visualizes the special case of Levi-Civita transporters on flat Euclidean spaces $M = \Euc_2$.
        Independently from the chosen path $\gamma$, the Levi-Civita transport keeps the vector (orange) parallel in the usual sense in Euclidean spaces.
        Gauges $\psi_q^{\widetilde{A}}$ (red) and $\psi_p^A$ (green) allow to express the coordinate free transporter by a group element
        $g_\gamma^{A\widetilde{A}} = \psi_p^A \circ \mathcal{P}_\gamma \circ \big(\psi_q^{\widetilde{A}}\big)^{-1} \in \GL{d}$
        which accounts for the change of vector coefficients if the target frame does not agree with the transported source frame.
        Fig.~\ref{fig:transport_sphere} shows the Levi-Civita transport on the 2-sphere~$S^2$, Eq.~\eqref{eq:sphere_transport_embedded}.
        The transporters $\mathcal{P}_{\mkern-2mu\gamma_1}$ and $\mathcal{P}_{\mkern-2mu\gamma_2}$ along different paths $\gamma_1$ and $\gamma_2$ disagree in general.
        As before, the coordinate free transporters can be expressed by group elements that operate on coefficients relative to the coordinate frames at $q$ and $p$.
     }
    \label{fig:transport}
\end{figure*}

It is didactically reasonable to start with the specific case of Levi-Civita transporters on Euclidean spaces, depicted in Fig.~\ref{fig:transport_flat}, before proceeding to more general transporters and manifolds.
In this case the parallel transport is independent from the chosen path $\gamma$ and keeps the transported vector parallel in the usual sense on Euclidean spaces.
Note that the transporter~$\mathcal{P}_\gamma$ is mapping between the tangent spaces $\TqM$ and $\TpM$ and is therefore coordinate free.
It can, however, be expressed relative to coordinates, then operating on numerical coefficient vectors instead of tangent vectors.
An intuition is given in Fig.~\ref{fig:transport_flat}, where the frames at $q$ and $p$ are not parallel%
\footnote{
    In contrast to general manifolds, $\R^d$ comes with a canonical notion of parallelism of reference frames.
}
such that the coefficients $(1,1)^\top$ at $q$ and $\big(\sqrt{2},0\big)^\top$ at $p$ differ even though the corresponding (coordinate free) tangent vectors are parallel to each other.
To make this more precise, consider gauges $\psi_q^{\widetilde{A}}$ and $\psi_p^A$ to be given on neighborhoods $U^{\widetilde{A}}$ of $q$ (red) and $U^A$ of $p$ (green).
Let a vector $v = \big( \psi_q^{\widetilde{A}} \big)^{-1} (v^{\widetilde{A}}) \in \TqM$ be given by its coefficients $v^{\widetilde{A}} \in \R^d$.
The coefficients of the transported vector $\mathcal{P}_\gamma v$ at $p$ are then given by
$
\psi_p^A \circ \mathcal{P}_\gamma (v)
\ =\ \psi_p^A \circ \mathcal{P}_\gamma \circ \big(\psi_q^{\widetilde{A}}\big)^{-1} (v^{\widetilde{A}})
$.
It follows that the coordinate expression of a transporter is relative to gauges $\widetilde{A}$ and $A$ expressed as:%
\footnote{
    $g_\gamma^{A\widetilde{A}}$ takes values in $\GL{d}$ if we assume arbitrary ($\mathfrak{gl}(d)$-valued) connections and general structure groups $G\leq\GL{d}$.
    For the $\mathfrak{so}(d)$-valued Levi-Civita connection and orthonormal frames, i.e. $G=\O{d}$, one has $g_\gamma^{A\widetilde{A}} \in \O{d}$.
}
\begin{align}\label{eq:transporter_gauge}
    g_\gamma^{A\widetilde{A}} \ :=\ \psi_p^A \circ \mathcal{P}_\gamma \circ \big( \psi_q^{\widetilde{A}} \mkern1mu\big)^{-1} \in\, \GL{d}
\end{align}
The group element $g_\gamma^{A\widetilde{A}}$ accounts for non-parallel choices of reference frames at $q$ and $p$.
On $\R^d$, one typically assumes all frames to be parallel such that all coordinatizations of Levi-Civita transporters become trivial.%
\footnote{
    Conventional CNNs on $\R^d$ implicitly make this assumption of parallel frames (Fig.~\ref{fig:G_structure_intro_a}) and trivial transporters.
}

As the transporter in Eq.~\eqref{eq:transporter_gauge} is coordinate dependent, we are interested in its gauge transformations.
Denote by $\psi_q^{\widetilde{B}}$ and $\psi_p^B$ two alternative gauges on neighborhoods of $q$ and $p$.
From the commutative diagram
\begin{equation}\label{cd:transporter_trivialization}
\begin{tikzcd}[column sep=50pt, row sep=25pt, font=\normalsize]
    \R^d
        \arrow[dd, "g_q^{\widetilde{B}\widetilde{A}}\cdot\ "']
        \arrow[rrr, "g_\gamma^{A\widetilde{A}}\cdot"]
    & &[-1ex] &
    \R^d
        \arrow[dd, "\ g_p^{BA}\cdot"]
    \\
    &
    \TqM
        \arrow[ul, "\psi_q^{\widetilde{A}}"]
        \arrow[dl, "\psi_q^{\widetilde{B}}"']
        \arrow[r, "\mathcal{P}_\gamma"]
    &
    \TpM
        \arrow[ur, "\psi_p^A"']
        \arrow[dr, "\psi_p^B"]
    \\
    \R^d
        \arrow[rrr, "g_\gamma^{B\widetilde{B}}\cdot"']
    & & &
    \R^d
\end{tikzcd}
\end{equation}
one can then read off that the transporters in the different gauges are related by
\begin{align}\label{eq:transporter_gauge_trafo}
    g_\gamma^{B\widetilde{B}}
    \ =\ g_p^{BA}\, g_\gamma^{A\widetilde{A}}\, \big(g_q^{\widetilde{B}\widetilde{A}} \mkern1mu\big)^{-1}
\end{align}
Note the similarity of this transformation law and commutative diagram to those in Eqs.~\eqref{eq:matrix_gaugetrafo} and~\eqref{eq:linear_map_TpM_diagram}.
The difference between both is that the transporter has a different domain $\TqM$ and codomain $\TpM$, which are trivialized by different, independent gauges and transform therefore independently.

In general, the parallel transport of tangent vectors is determined by some choice of connection, for instance (but not necessarily) by the canonical Levi-Civita connection of a Riemannian manifold.
A connection can be seen as a collection of infinitesimal transporters between adjacent tangent spaces, such that the full transporter $\mathcal{P}_\gamma$ is given by integrating the connection along the path $\gamma$.
The transporters along different paths $\gamma_1$ and~$\gamma_2$ from $q$ to~$p$ need not agree, which is in Fig.~\ref{fig:transport_sphere} exemplified by Levi-Civita transporters on the 2-sphere~$S^2$, cf.~Eq.~\eqref{eq:sphere_transport_embedded}.
As for flat spaces, the coordinate free transporters can by Eq.~\eqref{eq:transporter_gauge} be expressed relative to gauges.
The gauge transformations of such coordinatized transporters are again given by Eq.~\eqref{eq:transporter_gauge_trafo}.
The transporters on a given manifold can in principle be calculated analytically from the connection~\cite{gallier2019diffgeom1,nakahara2003geometry} and can sometimes be expressed in closed form, for instance for the sphere $S^2$, Eq.~\eqref{eq:sphere_transport_embedded}.
Several numerical algorithms exist to compute parallel transporters on meshes; see Section~\ref{sec:surfaces_geom_mesh}.
We will not go into more details on how to compute tangent vector transporters $\mathcal{P}_\gamma$ but simply assume them to be given.

\subsubsection{Feature vector transporters}

Eq.~\eqref{eq:gauge_trafo_features} defines the transformation law of feature vector coefficients by their \emph{field type}~$\rho$.
Their parallel transporter, expressed relative to gauges $\psi_q^{\widetilde{A}}$ and $\psi_p^A$, is analogously given by wrapping the tangent vector coefficient transporter into this field representation, that is, by
\begin{align}
    \rho\big( g_\gamma^{A\widetilde{A}} \big) \,.
\end{align}
Note that -- since $\rho:G\to\GL{c}$ is a $G$-representation -- this construction is only then well defined when all transporters~$g_\gamma^{A\widetilde{A}}$ (for arbitrary paths $\gamma$ and frames $A$, $\widetilde{A}$) are actually taking values in the chosen structure group~$G$.
Whether this is the case depends both on the particular choice of $G$-structure (or $G$-atlas) and the transporters (or connection) considered -- they need to be \emph{compatible}~\cite{wendlLectureNotesBundles2008}.

All convolutional networks accumulate (thus transport) feature vectors in one way or the other, and assume therefore some choice of connection and $G$-structure.
If the chosen $G$-structure is incompatible with the Levi-Civita connection, this implies that these models are -- often implicitly -- assuming an alternative, $G$-compatible connection to accumulate features.
The reader should for now not worry about the specific choices of connections, which will become more clear in our literature review in Part~\ref{part:literature_review}.
In the remainder of this section, we will elaborate more on the $G$-compatibility of connections and $G$-structures.
Assuming that \emph{feature transporters will in the following always be well defined}, this part can be safely ignored at a first reading.

A more rigorous, coordinate free discussion of transporters on the associated feature vector bundles can be found in Section~\ref{sec:bundle_transport}.

\subsubsection{Compatibility of connections and \textit{G}-structures}

A connection is said to be \emph{$G$-compatible} with a $G$-structure $\GM$ if the coordinate expressions $g_\gamma^{A\widetilde{A}}$ of its transporters $\mathcal{P}_\gamma$ relative to any frames $A$, $\widetilde{A}$ of $\GM$ take values in the structure group~$G$~\cite{wendlLectureNotesBundles2008}.%
\footnote{
    Equivalently, the \emph{connection 1-form} of the connection, expressed relative to frames of $\GM$, is required to be $\mathfrak{g}$-valued, where $\mathfrak{g}$ denotes the Lie algebra of~$G$.
    More abstractly, we are interested in \emph{principal Ehresmann connections} on the principal $G$-bundle $\GM$.
}
A $G$-compatible connection gives rise to transporters of $G$-associated feature vectors.

To illuminate this somewhat abstract compatibility condition, we discuss a few specific examples.
A simple example is that of the Levi-Civita connection on $\R^2$, Fig.~\ref{fig:transport_flat}, 
Consider the two $\{e\}$-structures on $\R^2$ that are shown in Figs.~\ref{fig:frame_field_automorphism_1} and~\ref{fig:frame_field_automorphism_2}.
Here $G=\{e\}$, which means that the field type $\rho: \{e\} \to \GL{c}$ is a $\{e\}$-representation, such that the parallel transport of feature vectors can only be defined if the coordinate expressions $g_\gamma^{A\widetilde{A}}$ take values in~$\{e\}$, i.e. are trivial.
As the $\{e\}$-structure in Fig.~\ref{fig:frame_field_automorphism_1} consists of ``parallel'' frames, this is indeed the case -- the Levi-Civita connection is thus compatible with this $\{e\}$-structure.
In contrast, the frames of the $\{e\}$-structure in Fig.~\ref{fig:frame_field_automorphism_2} are ``rotated'' relative to each other, resulting in non-trivial coordinate expressions $g_\gamma^{A\widetilde{A}}$ that take values in $\SO2$ (visualized in Fig.~\ref{fig:transport_flat}).
Since the field type $\rho: \{e\} \to \GL{c}$ does not handle rotations, it is not possible to define the Levi-Civita transport of features associated to this $\{e\}$-structure -- they are incompatible.
As a second example, consider the Levi-Civita connection on $S^2$, shown in Fig.~\ref{fig:transport_sphere}.
The transport will in this case always be path dependent and lead to differently rotated vectors, implying that $g_\gamma^{A\widetilde{A}}$ will take values in $\SO2$.
Feature vectors to be transported according to Levi-Civita connection need therefore to be of some type $\rho: \SO2 \to \GL{c}$ that is an $\SO2$-representations.
This requires at least the $\SO2$-structure on $S^2$ that is shown in Fig.~\ref{fig:G_structure_S2_1}.
The $\{e\}$-structure on $S^2$ from Fig.~\ref{fig:G_structure_S2_2} is incompatible with the Levi-Civita connection.

Since the Levi-Civita connection is a metric connection, it preserves lengths of and angles between tangent vectors, and thus transports orthonormal frames to orthonormal frames.
It follows that the Levi-Civita connection is \emph{always} compatible with the $\O{d}$-structure of orthonormal frames, relative to which $g_\gamma^{A\widetilde{A}}$ takes values in~$\O{d}$.
If the manifold is orientable, the frame-handedness is preserved by Levi-Civita transporters, which means that they are guaranteed to be compatible with $\SO{d}$-structures of orthonormal, right-handed frames on~$M$.
All of the convolutional networks in our literature review in Part~\ref{part:literature_review} that are based on $\SO{d}$-structures accumulate features via Levi-Civita transporters.

If a given $G$-structure is incompatible with the Levi-Civita connection, one needs to define an alternative, $G$-compatible connection to transport the feature vectors.
The most prominent example in our literature review is that of \emph{trivial connections} on $\{e\}$-structures.
A trivial connection is characterized by the property that its transport is \emph{path independent}~\cite{craneTrivialConnectionsDiscrete2010}.
Any $\{e\}$-structure implies a unique trivial connection, which transports tangent vectors such that they keep the same angle to the reference frames of the $\{e\}$-structure.
This implies $g_\gamma^{A\widetilde{A}} = e$, i.e. they transport coefficient vectors in $\R^c$ (relative to frames of the $\{e\}$-structure) without transforming their numerical values.
Such transporters are used in convolutional networks that do not explicitly model non-trivial transporters -- which applies to \emph{all} networks with $G=\{e\}$ in Table~\ref{tab:network_instantiations}, specifically those in Sections~\ref{sec:spherical_CNNs_azimuthal_equivariant} and~\ref{sec:e_surface_conv}.
Note that the trivial connection is the only connection that is compatible with an $\{e\}$-structure.

As stated above, \emph{any} convolutional network assumes some choice of compatible $G$-structure and connection,
most often Levi-Civita connections or trivial connections.

Section~\ref{sec:bundle_transport} elaborates on the compatibility of transporters and $G$-structures from a coordinate free viewpoint.

%% file: chapters/34_isometries.tex

\subsection{Isometry actions and induced gauge transformations}
\label{sec:isometries_local}

Until now our discussion focused exclusively on the local gauge symmetries in the coordinatization of tangent spaces.
A manifold might, however, come with non-trivial symmetries itself, which are in the case of a Riemannian manifold~$M$ forming its \emph{isometry group} $\IsomM$.
This section discusses isometries and their action on manifolds, tangent vectors, reference frames and feature fields in a nutshell, summarizing results which are more rigorously derived in Section~\ref{sec:isom_background}.
We will thereby highlight the equivalence of \emph{active} isometry actions and their \emph{passive} interpretation in terms of isometry induced gauge transformations.
This equivalence will later on allow us to describe the isometry equivariance of $\GM$-convolutions.

\begin{SCfigure}[1.8]
    \centering
    \hspace{.5ex}
    \includegraphics[width=.35\textwidth]{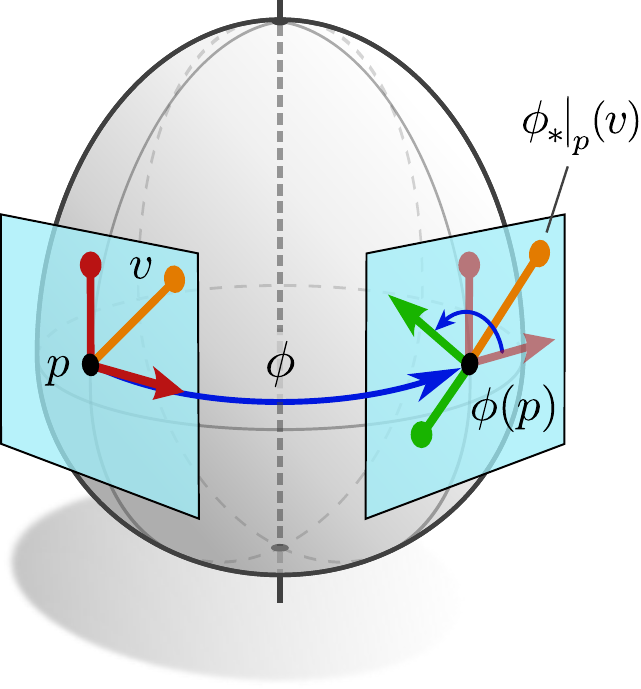}
    \caption[]{\small
        Visualization of the coordinate free pushforward of tangent vectors and its coordinate expression relative to given reference frames at the source and target location.
        The coordinate free pushforward ${\phi_*|_p: \TpM \to \TphipM}$ moves tangent vectors $\ v \!\in\! \TpM\ $ to $\ \phi_*|_p(v) \!\in\! \TphipM\,$ (orange).
        Let $\psi_p^{\widetilde{A}}$ be the gauge at $p$ that corresponds to the red reference frame and $\psi_{\phi(p)}^A$ the gauge at $\phi(p)$ that corresponds to the green reference frame.
        They explain the vectors before and after the pushforward by numerical coefficients $\psi_p^{\widetilde{A}}(v) = (1,1)^\top$ and $\psi_{\phi(p)}^A \big( \phi_*|_p(v) \big) = \big(0,-\sqrt{2}\big)^\top$.
        This transformation of vector coefficients is described by the isometry induced gauge transformation $g_\phi^{A\widetilde{A}}(p) \in \GL{d}$, that is,
        $\psi_{\phi(p)}^A \big( \phi_*|_p(v) \big) = g_\phi^{A\widetilde{A}}(p) \cdot \psi_p^{\widetilde{A}}(v)$.
        The coefficients of feature vectors transform analogously according to $\rho\big( g_\phi^{A\widetilde{A}}(p) \big)$ if $g_\phi^{A\widetilde{A}}(p) \in G$.
        \\[1ex]
        }
    \label{fig:pushforward_vector_components}
\end{SCfigure}

Isometries are defined as the symmetries of Riemannian manifolds, that is, those maps (diffeomorphisms)
\begin{align}
    \phi: M \to M,
\end{align}
that preserve the metric and thus distances on~$M$.
The set of all isometries of a Riemannian manifold~$M$ forms its isometry group, which we denote as $\IsomM$.
For instance, the Euclidean group~$\E{d}$ is the isometry group of Euclidean spaces~$\Euc_d$.
It consists of translations, rotations and reflections, all of which preserve the standard metric of~$\Euc_d$.
The isometry group of the $2$-sphere $S^2$ is given by the orthogonal group $\O3$, consisting of rotations and reflections.
Fig.~\ref{fig:pushforward_vector_components} shows an egg-shaped manifold, whose isometries are rotations and reflections in $\O2$ around the vertical axis.

\subsubsection{Pushforward of tangent vectors}

Any isometry $\phi \in \IsomM$ acts via its \emph{pushforward} (or differential)
\begin{align}
    \phi_{*}|_p:\, \TpM \to \TphipM \,,
\end{align}
naturally on tangent vectors.
As visualized in Fig.~\ref{fig:intro_gauge_isom_induction} (middle), the pushforward can intuitively be thought of as carrying tangent vectors along with the action of the isometry on the underlying manifold~$M$.
A formal definition of the pushforward on $\TM$ is given in Appendix~\ref{apx:differentials_gradients_jacobians}, however, the given intuition is sufficient for our purpose.
Since the pushforward is a coordinate free, linear map between tangent spaces, its action is in coordinates represented by some $d\times d$ matrix.
Assuming gauges $\psi_p^{\widetilde{A}}$ and $\psi_{\phi(p)}^A$ at the source and target location, respectively, this matrix is given by
\begin{align}\label{eq:isom_induced_gauge_trafo_24}
    g_\phi^{A\widetilde{A}}(p)\ :=\ \psi_{\phi(p)}^A \circ \phi_*|_p \circ \big( \psi_p^{\widetilde{A}} \mkern1mu\big)^{-1} \ \ \in\, \GL{d} \,.
\end{align}
It explains the transformation from the numerical coefficients of an original vector $v\in\TpM$ in the source gauge and its pushforward $\phi_*|_p(v) \in\TphipM$ in the target gauge, that is,
$\psi_{\phi(p)}^A \big( \phi_*|_p (v) \big) = g_\phi^{A\widetilde{A}}(p) \cdot \psi_p^{\widetilde{A}}(v)$.
The commutative diagram
\begin{equation}
\begin{tikzcd}[column sep=54pt, row sep=28pt, font=\normalsize]
    \R^d
        \arrow[dd, "g_p^{\widetilde{B}\widetilde{A}}\cdot\ "']
        \arrow[rrr, "g_\phi^{A\widetilde{A}}(p)\cdot"]
    & &[-1ex] &
    \R^d
        \arrow[dd, "\ g_{\phi(p)}^{BA}\cdot"]
    \\
    &
    \TpM
        \arrow[ul, "\psi_p^{\widetilde{A}}"]
        \arrow[dl, "\psi_p^{\widetilde{B}}"']
        \arrow[r, "\phi_*|_p"]
    &
    \TphipM
        \arrow[ur, "\psi_{\phi(p)}^A"']
        \arrow[dr, "\psi_{\phi(p)}^B"]
    \\
    \R^d
        \arrow[rrr, "g_\phi^{B\widetilde{B}}(p)\cdot"']
    & & &
    \R^d
\end{tikzcd}
\quad,
\end{equation}
which is conceptually similar to that in Eq.~\eqref{cd:transporter_trivialization}, visualizes the definition of the tangent vector pushforward's coordinate expression.
It furthermore implies that the gauge transformations between different coordinatizations are given by
\begin{align}
    g_\phi^{B\widetilde{B}}
    \ =\ g_{\phi(p)}^{BA} \; g_\phi^{A\widetilde{A}} \mkern1mu \big(g_p^{\widetilde{B}\widetilde{A}} \mkern1mu\big)^{-1} \,,
\end{align}
which is the conceptual analog to Eq.~\eqref{eq:transporter_gauge_trafo}.

\subsubsection{Pushforward of reference frames and symmetries of the \textit{G}-structure}

Since reference frames are just $d$-tuples of linearly independent frame vectors, the pushforward of tangent vectors induces a pushforward of reference frames by \emph{pushing the individual frame axes forward}.
Specifically, the pushforward of a frame $[e_i]_{i=1}^d$ at~$p$ is defined as the frame $\big[ \phi_*|_p (e_i) \big]_{i=1}^d$ at~$\phi(p)$.

This pushforward of frames is always well defined, however, it might not be compatible with the $G$-structure, that is, there is in general no guarantee that frames in $\GM$ remain in $\GM$ when being pushed forward.
Take for instance the $\{e\}$-structure in Fig.~\ref{fig:intro_invariant_kernel_fields_plane} (top left), which is preserved by horizontal translations but not by vertical translations or any other isometry of~$\R^2$.
Similarly, the $\Flip$-structure in Fig.~\ref{fig:intro_invariant_kernel_fields_plane} (bottom left) is preserved by translations and horizontal reflections, but not by rotations.
We consider therefore the subgroup
\begin{align}\label{eq:IsomGM_def_24}
    \IsomGM\ :=\ \pig\{ \phi \in \IsomM \ \pig|\ 
    \big[\phi_*(e_i)\big]_{i=1}^d \in \GM \quad \forall\ [e_i]_{i=1}^d \in \GM \pig\} \,\ \leq\ \IsomM
\end{align}
of \emph{isometries which are symmetries of the $G$-structure}, i.e. which are guaranteed to map any frame in~$\GM$ to another frame that is also contained in~$\GM$.%
\footnote{
    More formally stated, such isometries are (or induce) \emph{principal bundle automorphisms} of the $G$-structure.
}
Note that $\IsomGM$ depends in general on the specific choice of $G$-structure $\GM$, not only on the structure group~$G$.
For the special case that $G\geq\O{d}$, it is guaranteed that $\IsomGM=\IsomM$ coincide since isometries are guaranteed to map orthonormal frames to orthonormal frames.
We are interested in the subgroup $\IsomGM$ since only those isometries will induce a well defined pushforward of $\GM$-coordinate independent feature vectors, as discussed further in the following section.

Before proceeding to the isometry action on feature vectors, we discuss what we call \emph{isometry induced gauge transformations}.
For this purpose, let $\big[e_i^{\widetilde{A}} \big]_{i=1}^d$ be that frame at~$p$ that corresponds to some source gauge $\psi_p^{\widetilde{A}}$ and let $\big[e_i^A \big]_{i=1}^d$ be that frame at $\phi(p)$ that corresponds to some target gauge $\psi_{\phi(p)}^A$, as shown in Fig.~\ref{fig:pushforward_vector_components} in red (left) and green (right), respectively.
The pushforward $\big[\phi_*|_p( e_i^{\widetilde{A}}) \big]_{i=1}^d$ of the source frame from $p$ to $\phi(p)$ (translucent red, right) does in general not coincide with the target frame.
However, as proven in Section~\ref{sec:isom_coordinatization}, the two frames are related by the isometry induced gauge transformation
\begin{align}
    \pig[\phi_*|_p( e_i^{\widetilde{A}}) \pig] \raisebox{-2.5pt}{$\rule{0pt}{12pt}_{i=1}^d$}
    \ =\ \big[e_i^A \big]_{i=1}^d \lhd g_\phi^{A\widetilde{A}}(p) \,,
\end{align}
where $g_\phi^{A\widetilde{A}}(p)$ is the group element from Eq.~\eqref{eq:isom_induced_gauge_trafo_24} and $\lhd$ is the right action from Eq.~\eqref{eq:right_action_mapsto}.
The term ``isometry induced gauge transformation'' makes in so far sense that the geometries around $p$ and $\phi(p)$ are indistinguishable since $\phi$ is an isometry, i.e. a symmetry of~$M$.
Identifying the two points with each other, one can therefore reinterpret the \emph{active} action of $\phi$ on a geometric quantity as a \emph{passive} gauge transformation, i.e. an induced change from the source to the target frame.

Theorem~\ref{thm:isom_GM_in_coords} in Section~\ref{sec:isom_background} asserts that $G$-structure preserving isometries in $\IsomGM$ and $G$-valued induced gauge transformations imply each other, that is,
\begin{align}\label{eq:IsomGM_coord_in_G_24}
    \phi \in \IsomGM \quad \Longleftrightarrow \quad g_\phi^{A\widetilde{A}}(p) \in G\ \ \ \forall\ p \mkern-1mu\in\mkern-2mu M
\end{align}
holds for arbitrary gauges $\psi_p^{\widetilde{A}}$ and $\psi_{\phi(p)}^A$ of the $G$-atlas.
The reader should verify these claims at our examples in Fig.~\ref{fig:intro_invariant_kernel_fields_plane}.

\subsubsection{Pushforward of feature vectors}

If (and only if) an isometry is a symmetry of the $G$-structure, it gives rise to a \emph{pushforward of feature vectors}.
Intuitively, this pushforward moves feature vectors from points $p$ to $\phi(p)$.
When being expressed relative to the two reference frames at $p$ and $\phi(p)$, it is given by the induced gauge transformation
\begin{align}
    \rho\big( g_\phi^{A\widetilde{A}}(p) \big) \,.
\end{align}
Note that this transformation is well defined for any $\phi \in \IsomGM$, since the induced gauge transformations $g_\phi^{A\widetilde{A}}(p)$ will in this case take values in~$G$ and $\rho$ is a $G$-representation.
In contrast, if $\phi$ is not a symmetry of the $G$-structure, it is \emph{impossible} to define a corresponding feature vector pushforward.
This statement relates to the fact that the features of conventional CNNs have no specified transformation behavior under rotations or reflections in the Euclidean group~$\E{d}$.

The pushforward of individual feature vectors implies an action on the whole feature field $f$, which we denote by $\phi \rhd f$.
Relative to coordinates, this action is expressed as
\begin{align}\label{eq:feature_field_trafo_in_coords}
    \big[\phi \rhd f\big]^A(\phi(p)) \ =\ \rho\big( g_\phi^{A\widetilde{A}}(p)\big)\, f^{\widetilde{A}}(p) \,.
\end{align}
We will later prove that coordinate independent CNNs are equivariant w.r.t. the action of isometries in $\IsomGM$ on feature fields; see Fig.~\ref{fig:lizard_conv_egg}.
This property relies on the fact that the active isometry action on feature fields can by Eq.~\eqref{eq:feature_field_trafo_in_coords} be understood as a mere passive gauge transformation of feature vector coefficients.

%% file: chapters/40_local_gauge_cnns_intro.tex

\section{Coordinate independent networks and \textit{GM}-convolutions}
\label{sec:gauge_CNNs_local}

Neural networks process data by applying a series of parameterized mappings (layers) to an input signal -- in our case to a set of feature fields on a Riemannian manifold.
The \emph{principle of covariance} requires thereby that the individual network layers should be $\GM$-coordinate independent operations.
The coordinate representations of such layers will therefore have to transform such that they respect the transformation laws of their input- and output feature field.
Except form this consistency requirement, general coordinate independent layers remain \emph{unconstrained}.

A common design principle of neural networks which operate on spatial signals (feature fields) is that they are in some generalized sense convolutional.
The main characteristic which most generalizations of the convolution operation share is that their inference is \emph{position independent}.
This is achieved by \emph{sharing template functions}, for instance convolution kernels or biases, between different locations.
Whenever the structure group $G$ is non-trivial, the weight sharing process is ambiguous since template functions could be shared relative to different reference frames.
As we will argue in the following, this ambiguity is resolved by designing the shared template functions to be \emph{equivariant under $G$-valued gauge transformations}.
Gauge equivariant template functions will be indifferent to the specific reference frame in which they are applied and therefore allow for a coordinate independent weight sharing.

\etocsettocdepth{3}
\etocsettocstyle{}{} 
\localtableofcontents

In this section we will consider network layers which take fields $\fin$ of type $\rhoin$ as input and produce field $\fout$ of type $\rhoout$ as output.
Section~\ref{sec:pointwise_operations} discusses the specific case of layers which operate pointwise, that is, whose output $\fout(p)$ at any $p\in M$ depends only on the single input feature vector $\fin(p)$ at the same location.
The practically relevant examples considered here are gauge equivariant \onexones\ in Section~\ref{sec:gauge_1x1}, bias summation in Section~\ref{sec:gauge_bias_summation} and nonlinearities in Section~\ref{sec:gauge_nonlinearities}.
The more complicated case of convolutions with spatially extended kernels is treated in Section~\ref{sec:gauge_conv_main}.
As a preparation, Section~\ref{sec:observers_view} discusses feature fields as seen from the viewpoints of local observers (reference frames), relative to which the (convolution) kernels will be applied.
Such observations are formalized as a pullback of the feature field to an observer's tangent space; see Fig.~\ref{fig:pullback_field_exp_TpM}.
Section~\ref{sec:kernel_field_trafos} introduces so called kernel field transforms, which are similar to convolutions but do not assume spatial weight sharing and are therefore parameterized by a (smoothly varying) kernel field on~$M$.
The actual $\GM$-convolutions are in Section~\ref{sec:gauge_conv} defined as those kernel field transforms that are parameterized by a single, shared template kernel.
In order to ensure the coordinate independence of the weight sharing process, the convolution kernels are required to be $G$-\emph{steerable}, i.e. to satisfy a gauge equivariance constraint.
Section~\ref{sec:gauge_conv_isom_equiv} shows that $\GM$-convolutions are automatically equivariant under those isometries that are symmetries of the $G$-structure ($\IsomGM$-equivariant).
This means that $\GM$-convolutions commute with the action of isometries on feature fields as visualized in Fig.~\ref{fig:lizard_conv_egg}.

%% file: chapters/41_pointwise_operations.tex

\subsection{Pointwise gauge equivariant operations}
\label{sec:pointwise_operations}

To begin with, we consider some neural network operations for which the constraints coming from the required coordinate independence and weight sharing are particularly easy to derive.
All of these operations have in common that they act pointwise on feature vectors, that is, they compute output feature vectors $\fout(p)$ at $p\in M$ solely based on the input feature vectors $\fin(p)$ at the same location.
In order to satisfy the principle of covariance, the coordinatizations of these operations are all required to transform according to a precomposition with $\rhoin$ and a postcomposition with $\rhoout$.
When demanding that the operations are determined in terms of shared weights, these transformation laws imply a requirement for the gauge equivariance (or invariance) of the operations.

The derivations for the different pointwise operations in the following Sections~\ref{sec:gauge_1x1}, \ref{sec:gauge_bias_summation} and~\ref{sec:gauge_nonlinearities} are in the first steps mostly analogous and lead to essentially the same covariance and equivariance constraints on the template functions.
They could therefore be treated together, keeping the particular operation (or template function) abstract.
However, since the implications of the resulting constraints differ for the particular instantiations, and since we want to keep the discussion close to the application, we will omit such an abstract formulation and directly consider particular instantiations.

\subsubsection[Gauge equivariant \texorpdfstring{\onexones}{1x1-convolutions}]%
              {Gauge equivariant ${\bf1\mkern-5.mu\boldsymbol{\times}\mkern-5.mu1}$-convolutions}
\label{sec:gauge_1x1}

As a first example of pointwise operations, we consider the action of a family of \emph{linear maps} $\mathcal{C}_p$, which send the input feature vector $\fin(p)$ at each $p\in M$ to an output feature vector
\begin{align}
    \fout(p) := \mathcal{C}_p\, \fin(p) \,.
\end{align}
If we add the assumption of spatial weight sharing, the linear maps $\mathcal{C}_p$ and $\mathcal{C}_q$ at different locations $p$ and~$q$ will be coupled, and the operation can be seen as a convolution with a linear operator-valued Dirac delta kernel.
This operation is quite common in computer vision, where it is usually denoted as \onexoneit, since the spatial discretization of a linear Dirac kernel which operates on two-dimensional images is given by a (matrix-valued) kernel with a spatial extent of $1\!\times\!1$ pixels.
We will in the following derive that the demand for spatial weight sharing will result in a constraint, which forces the matrix-valued template kernels to be \emph{intertwiners}, that is, gauge equivariant matrices.

Prior to the assumption of weight sharing, the coordinate expressions of the linear maps $\mathcal{C}_p$ and the gauge transformations between them behave very similar to those of the linear maps on $\TpM$, which we discussed in Section~\ref{sec:gauges_TpM_functions}.
Since the input and output feature vectors are in coordinates represented by coefficient vectors $\fin^A(p) \in \R^{\cin}$ and $\fout^A(p) \in \R^{\cout}$, the linear map is naturally represented by that matrix $\mathcal{C}_p^A \in \R^{\cout\times\cin}$ that satisfies
\begin{align}\label{eq:linear_op_coord_A}
    \fout^A(p) = \mathcal{C}_p^A \cdot \fin^A(p) \,.
\end{align}
This relation does of course hold for arbitrary coordinatizations, such that we have $\fout^B(p) = \mathcal{C}_p^B \cdot \fin^B(p)$ for any other gauge, labeled by $B$.
The transformation law which relates $\mathcal{C}_p^B$ to $\mathcal{C}_p^A$ follows by the principle of covariance from the transformation laws of the input and output features.
Since these are given by $\fin^B(p) = \rhoin\big( g_p^{BA}\big) \fin^A(p)$ and $\fout^B(p) = \rhoout\big( g_p^{BA}\big) \fout^A(p)$, one has
\begin{alignat}{3}
    && \fout^B(p)\ &=\ \mathcal{C}_p^B \cdot \fin^B(p)
    \notag \\ \Leftrightarrow \qquad
    && \rhoout\big( g_p^{BA}\big)\, \fout^A(p)\ &=\ \mathcal{C}_p^B\, \rhoin\big( g_p^{BA}\big)\, \fin^A(p)
    \notag \\ \Leftrightarrow \qquad
    && \fout^A(p)\ &=\ \rhoout\big( g_p^{BA}\big)^{-1}\, \mathcal{C}_p^B\, \rhoin\big( g_p^{BA}\big)\, \fin^A(p) \,.
\end{alignat}
A comparison with Eq.~\eqref{eq:linear_op_coord_A} implies that the two coordinate expressions of $\mathcal{C}_p$ are necessarily related by
\begin{align}\label{eq:linear_op_trafo_law}
    \mathcal{C}_p^B\ =\ \rhoout\big( g_p^{BA}\big)\, \mathcal{C}_p^A\, \rhoin\big( g_p^{BA}\big)^{-1}
\end{align}
if they should respect the transformation laws of the feature vectors.
As usual, these considerations are concisely captured by a commutative diagram:
\begin{equation}\label{cd:linear_op_trafo_law}
\begin{tikzcd}[column sep=60pt, row sep=30pt, font=\normalsize]
    \R^{\cin}
        \arrow[d, "\rhoin\big(g_p^{BA}\big)\cdot\,"']
        \arrow[r, "\mathcal{C}_p^A \cdot"]
    &
    \R^{\cout}
        \arrow[d, "\ \rhoout\big(g_p^{BA}\big) \cdot"]
    \\
    \R^{\cin}
        \arrow[r, "\mathcal{C}_p^B \cdot"']
    &
    \R^{\cout}
\end{tikzcd}
\end{equation}
The important practical implication of this result so far is that the linear map $\mathcal{C}_p$ is not restricted in any way.
Differently formulated: as long as the coordinate expressions in different gauges are related by Eq.~\eqref{eq:linear_op_trafo_law}, one is free to parameterize $\mathcal{C}_p$ in an arbitrary, fixed gauge $A$ by an \emph{unconstrained} matrix $\mathcal{C}_p^A$.
As we will see, the situation changes when requiring the linear maps to share weights.

Consider now the case where the linear maps $\mathcal{C}_p$ and $\mathcal{C}_q$ share weights.
This means that we assume them to be parameterized by a shared set of parameters, given by a \onexone\ template kernel $K_{\!1\!\times\!1} \in \R^{\cout\times\cin}$.
The open question is how exactly the coordinate free maps should be parameterized in terms of this template kernel.
Our requirement for $\GM$-coordinate independence demands that we do not prefer any particular reference frame in the weight sharing process, that is, that we treat all coordinatizations in the same manner.
It is therefore necessary to \emph{share the template kernel with all coordinatizations at the same time}, that is, to set
\begin{align}\label{eq:weight_sharing_1x1}
    \mathcal{C}_p^X = K_{\!1\!\times\!1}
    \quad \textup{for \emph{any} gauge}\ \ \big(U^X,\psi^X) \in \mathscr{A}^G\ \ \textup{with}\ \ p\in U^X \,,
\end{align}
where $\mathscr{A}^G$ is the (maximal) $G$-atlas corresponding to the considered $G$-structure; see Eq.~\eqref{eq:G_atlas_dfn}.
As the covariance constraint in Eq.~\eqref{eq:linear_op_trafo_law} needs to hold for arbitrary $G$-related gauges, and the coordinatizations $\mathcal{C}_p^A = \mathcal{C}_p^B = K_{\!1\!\times\!1}$ of the linear maps do all coincide, the joint demand for weight sharing and $\GM$-coordinate independence is seen to imply a constraint
\begin{align}\label{eq:Konexone_constraint_intertwiner}
    K_{\!1\!\times\!1}\ =\ \rhoout(g)\, K_{\!1\!\times\!1}\, \rhoin(g)^{-1} \qquad \forall\ g\in G
\end{align}
on the template kernel.
The corresponding adaptation of the commutative diagram in Eq.~\eqref{cd:linear_op_trafo_law} with weight sharing is given by:
\begin{equation}
\begin{tikzcd}[column sep=60pt, row sep=30pt, font=\normalsize]
    \R^{\cin}
        \arrow[d, "\rhoin\big(g_p^{BA}\big)\cdot\,"']
        \arrow[r, "K_{\!1\!\times\!1} \cdot"]
    &
    \R^{\cout}
        \arrow[d, "\ \rhoout\big(g_p^{BA}\big) \cdot"]
    \\
    \R^{\cin}
        \arrow[r, "K_{\!1\!\times\!1} \cdot"']
    &
    \R^{\cout}
\end{tikzcd}
\end{equation}

The conclusion of this analysis is that the template kernels which can be \emph{unambiguously shared} are exactly those which are \emph{invariant under the gauge action}.
The vector space of such gauge invariant \onexone\ kernels is simply the space of \emph{intertwining maps} between the representations $\rhoin$ and $\rhoout$, that is,
\begin{align}\label{eq:gauge_onexone_solution_space}
    \Hom_G(\rhoin,\rhoout)\ :=\ 
    \pig\{ K_{\!1\!\times\!1} \in \R^{\cout\times\cin}\ \pig|\ 
    K_{\!1\!\times\!1} = \rhoout(g)\, K_{\!1\!\times\!1}\, \rhoin(g)^{-1}\ \ \ \forall g\in G \pig\}
    \,\ \subseteq\ \R^{\cout\times\cin} \,.
\end{align}
Note that, according to \emph{Schur's Lemma}~\cite{gallier2019harmonicRepr}, the requirement on $K_{\!1\!\times\!1}$ to be an intertwiner prevents a mapping between fields that transform under non-isomorphic irreducible representations via \onexones.
This severe restriction is unavoidable with \onexone\ kernels but will be resolved later when allowing for spatially extended kernels.

At this point we want to mention that we use the terms ``gauge equivariant template function'' and ``gauge invariant template function'' interchangeably.
This is justified by the observation that the invariance constraint in Eq.~\eqref{eq:Konexone_constraint_intertwiner} can be written as an equivariance constraint
$K_{\!1\!\times\!1}\, \rhoin(g) = \rhoout(g)\, K_{\!1\!\times\!1}\ \ \forall g\in G$.
It is in general possible to view functions which are equivariant w.r.t. some group action in their domain and codomain as the invariants of the corresponding action on the function itself.
In our application, the equivariance viewpoint highlights that a transformation of the input field will lead to a corresponding transformation of the output field, which ensures that all involved quantities transform covariantly with each other.
On the other hand, the invariance viewpoint emphasizes that the template function can be shared in an arbitrary gauge.

\subsubsection{Gauge equivariant bias summation}
\label{sec:gauge_bias_summation}

After applying a convolution operation, it is common to sum a (shared) bias vector to the individual feature vectors.
Together with the requirement of coordinate independence, the weight sharing will again lead to a linear constraint.
This constraint will only allow for biases to be summed to the invariant subspaces of the gauge action on the input feature field.

As before, we first consider the bias summation without requiring weight sharing.
We thus have biases~$\mathscr{b}_p$, depending on the position $p$ on the manifold, which are summed to an input feature vector to produce an output feature vector
\begin{align}
    \fout(p) = \fin(p) + \mathscr{b}_p \,.
\end{align}
Relative to gauges $\psi_p^A$ and $\psi_p^B$, the bias is represented by those coefficient vectors $\mathscr{b}_p^A$ and $\mathscr{b}_p^B$ in $\R^c$ that satisfy $\fout^A(p) = \fin^A(p) + \mathscr{b}_p^A$ and $\fout^B(p) = \fin^B(p) + \mathscr{b}_p^B$.
Since the summation of vectors does not allow to change their transformation laws, the group representations associated with the input and output feature necessarily agree, that is,
\begin{align}
    \rhoin = \rhoout =: \rho \,.
\end{align}
Together with the requirement for coordinate independence, this implies that the diagram
\begin{equation}\label{eq:bias_trafo_law}
\begin{tikzcd}[column sep=75pt, row sep=30pt, font=\normalsize]
    \R^c
        \arrow[d, "\rho \big(g_p^{BA}\big)\cdot\,"']
        \arrow[r, "+\mathscr{b}_p^A"]
    &
    \R^c
        \arrow[d, "\ \rho\big( g_p^{BA}\big) \cdot"]
    \\
    \R^c
        \arrow[r, "+\mathscr{b}_p^B"']
    &
    \R^c
\end{tikzcd}
\ \ ,
\end{equation}
which is the analog of that in Eq.~\eqref{cd:linear_op_trafo_law}, needs to commute.
Written out as an equation, this demands the relation
$\rho\big(g_p^{BA}\big) f^A_p + \mathscr{b}_p^B \ =\ \rho\big(g_p^{BA}\big) \big(f^A_p + \mathscr{b}_p^A\big)$
to hold.
Since the linearity of $\rho(g)$ allows to rewrite the right-hand side as
$\rho\big(g_p^{BA}\big) f^A_p + \rho\big(g_p^{BA}\big) \mathscr{b}_p^A$,
a subtraction of the input feature vector leads to
\begin{align}\label{eq:bias_trafo_non_shared}
    \mathscr{b}_p^B\ =\ \rho\big(g_p^{BA}\big) \, \mathscr{b}_p^A \,.
\end{align}
The coefficient vectors which represent a coordinate independent bias relative to different gauges therefore need to transform exactly like the feature vectors to which they are summed.
As in the case of \onexones, the coordinate independence does \emph{not} restrict the bias $\mathscr{b}_p$ in any way, but only requires different coordinatizations of the same bias to be consistent with each other.
An implementation could therefore pick an arbitrary gauge and freely parameterize the bias in that gauge by parameters in $\R^{\cin}$.

The situation changes again when asking for spatial weight sharing.
Let $b \in \R^{\cin}$ be a template bias vector to be shared over the manifold.
Since the only way to do this without arbitrarily preferring any coordinatization is to share the bias vector in all gauges simultaneously, we have to require
\begin{align}
    \mathscr{b}_p^X = b
    \quad \textup{for \emph{any} gauge}\ \ \big(U^X,\psi^X) \in \mathscr{A}^G\ \ \textup{with}\ \ p\in U^X \,.
\end{align}
in analogy to Eq.~\eqref{eq:weight_sharing_1x1}.
The combination of the covariance constraint in Eq.~\eqref{eq:bias_trafo_non_shared} with this gauge independent weight sharing then leads to the invariance constraint
\begin{align}\label{eq:bias_invariance_constraint}
    b\ =\ \rho(g)\, b \qquad \forall\ g\in G
\end{align}
on the bias vector template.
To complete the analogy to the case of \onexones, we show the adapted version of the commutative diagram in Eq.~\eqref{eq:bias_trafo_law} with shared weights:
\begin{equation}
\begin{tikzcd}[column sep=75pt, row sep=30pt, font=\normalsize]
    \R^c
        \arrow[d, "\rho\big( g_p^{BA}\big)\cdot\,"']
        \arrow[r, "+b"]
    &
    \R^c
        \arrow[d, "\ \rho\big( g_p^{BA}\big)\cdot"]
    \\
    \R^c
        \arrow[r, "+b"']
    &
    \R^c
\end{tikzcd}
\end{equation}

To get an insight in the implications of the invariance constraint in Eq.~\eqref{eq:bias_invariance_constraint}, assume it to be satisfied for a given template vector~$b$.
Due to the linearity of the constraint, any scaled vector $\alpha \!\cdot\! b$ for $\alpha\in\R$ will then satisfy it as well, that is, any solution spans a \emph{one-dimensional subspace of~$\R^c$} which is \emph{invariant under the action of~$\rho$}.
Such an invariant subspace is denoted as a subrepresentation of~$\rho$.
Since the subspaces in consideration are one-dimensional, they have themselves no proper subspace and are therefore trivial irreducible subrepresentations.
If follows that the vector space
\begin{align}\label{eq:gauge_bias_solution_space}
    \mathscr{B}^G_\rho\ :=\ \big\{ b \in\R^c \;\big|\; b = \rho(g)\mkern2mu b\ \ \ \forall g\in G \big\}
\end{align}
of gauge equivariant biases coincides with the (subspaces of) trivial subrepresentations of $\rho$.
The dimensionality of $\mathscr{B}^G_\rho$ -- and therefore the number of learnable parameters -- coincides with the multiplicity of trivial subrepresentations contained in~$\rho$.
For compact groups~$G$, Schur's orthogonality relations imply that this dimensionality is given by $\dim\!\big(\mathscr{B}^G_\rho\big) = \int_G \tr\big(\rho(g)\big) dg$.
This statement covers the practically important cases of the orthogonal groups $G=\O{d}$ and all of its subgroups.

Two simple examples of feature fields to which one might want to sum shared biases are scalar fields and tangent vector fields.
By definition, the coefficient field of a scalar field is invariant under gauge transformations, that is, it transforms according the trivial representation ${\rho(g)=1\ \ \forall g\in G}$.
One can therefore sum a (scalar) bias $b \in \R$ to them.
In contrast, the coefficient field of a tangent vector field transforms according to the non-trivial, irreducible group representation $\rho(g)=g$.
Since this representation does not contain any trivial subrepresentation, it is impossible to sum a shared bias vector to tangent vector fields while maintaining coordinate independence.
As a third example, consider regular representations of compact groups, which describe for instance the feature fields of group convolutional networks.
By the Peter-Weyl theorem, it is known that regular representations contain exactly one trivial subrepresentation~\cite{gurarie1992symmetries,gallier2019harmonicRepr}.
The bias to be summed to regular feature fields is therefore seen to be described by a single parameter.

\subsubsection{Gauge equivariant nonlinearities}
\label{sec:gauge_nonlinearities}

Except from linear (convolution) operations and bias summations, the most basic operations used in any neural network are nonlinearities.
We will here consider the usual case of nonlinearities $\sigma_p$ which act in a spatially localized way, that is, which compute output feature vectors as $\fout(p) = \sigma_p\big( \fin(p) \big)$.
A shared nonlinearity will again be required to be gauge equivariant.
As the reasoning which leads to this conclusion is similar to that in the previous cases, we will only summarize it shortly.
Due to the generality of nonlinear maps it is impossible to derive linear solution spaces as in Eqs.~\eqref{eq:gauge_onexone_solution_space} and~\eqref{eq:gauge_bias_solution_space}, however, we will discuss some specific examples.

Similar to before, any coordinate free nonlinearity $\sigma_p$ is relative to gauges $A$ and $B$ given by coordinate expressions $\sigma_p^A: \R^{\cin} \to \R^{\cout}$ and $\sigma_p^B: \R^{\cin} \to \R^{\cout}$, which are by the demand for coordinate independence required to be related by $\sigma_p^B = \rhoout\big( g_p^{BA}\big) \circ \sigma_p^A \circ \rhoin\big( g_p^{BA}\big)^{-1}$.
A nonlinear template function $\mathscr{s}: \R^{\cin} \to \R^{\cout}$ can only be shared in a coordinate independent way when sharing it with all gauges simultaneously.
This turns the covariance constraint in an invariance constraint $\mathscr{s} = \rhoout(g) \circ \mathscr{s} \circ \rhoin(g)^{-1}\ \ \forall g\in G$ on the template function, or, equivalently, in the corresponding equivariance constraint
\begin{align}\label{eq:gauge_constraint_nonlinearities}
    \rhoout(g) \circ \mathscr{s} = \mathscr{s} \circ \rhoin(g)^{-1} \qquad \forall\ g\in G \,.
\end{align}

Due to the nonlinearity of this constraint we are forced to investigate it on a case-by-case basis -- we will therefore limit our discussion to some specific examples.
The arguably most simple case is that of nonlinearities which map between scalar fields, i.e. for which $\rhoin(g) = \rhoout(g) = 1$ are for any $g \in G$ invariant.
In this case the equivariance constraint in Eq.~\eqref{eq:gauge_constraint_nonlinearities} becomes $\mathscr{s} = \mathscr{s}$, which is trivially satisfied for \emph{any} nonlinearity $\mathscr{s}: \R \to \R$.
A more interesting example is that of unitary representation $\rhoin$.
One possible nonlinearity for this case is given by the norm of feature vectors.
Since $\lVert \rhoin(g) \fin^A(p) \rVert = \lVert \fin^A(p) \rVert$ is due to the unitarity of $\rhoin$ invariant, $\rhoout$ will be the trivial representation.
Taking the norm is thus seen to be a nonlinear, gauge equivariant operation that maps any unitary field to a scalar field.
A nonlinear map which preserves the field type, i.e. which satisfies $\rhoout = \rhoin$, can very similarly be defined as $\fin^A(p) \mapsto \lVert\fin^A(p)\rVert \cdot \fin^A(p)$.
Another option, which might play a role in learning physical interactions, and which was investigated in~\cite{kondor2018ClebschGordan,Kondor2018-NBN,anderson2019cormorant,alex2020lorentz}, are tensor product nonlinearities.
Given two fields $f_\textup{\;\!\!in,1}^A(p)$ and $f_\textup{\;\!\!in,2}^A(p)$, transforming according to
$\rho_{\overset{}{\protect\scalebox{.6}{\;\!\!\textup{in},1}}}$ and $\rho_{\overset{}{\protect\scalebox{.6}{\;\!\!\textup{in},2}}}$,
respectively, such nonlinearities compute a tensor product feature $\fout^A = f_\textup{\;\!\!in,1}^A(p) \otimes f_\textup{\;\!\!in,2}^A(p)$, which transforms equivariantly according to the tensor product representation
$\rhoout = \rho_{\overset{}{\protect\scalebox{.6}{\;\!\!\textup{in},1}}} \otimes \rho_{\overset{}{\protect\scalebox{.6}{\;\!\!\textup{in},2}}}$.

All of these examples satisfy the gauge equivariance constraint in Eq.~\eqref{eq:gauge_constraint_nonlinearities}.
Which particular nonlinearity works well in practice is, however, still an open research question, which requires much more empirical investigation before it can be answered.
A first attempt in this direction has been made in~\cite{Weiler2019_E2CNN}.

%% file: chapters/42_gauge_conv.tex

\subsection{Kernel field transforms and \textit{GM}-convolutions}
\label{sec:gauge_conv_main}

The central operation of convolutional networks is the convolution operation, which linearly accumulates characteristic patterns of features from a local neighborhood around each point $p\in M$ into a new feature vector $\fout(p)$.
A spatially extended convolution kernel determines thereby the specifics of this accumulation.
The principle of covariance requires coordinate independence, and therefore a specific transformation law of kernels under gauge transformations.
As in the previous examples, an additional demand for weight sharing results in a requirement on the template kernel to be gauge equivariant ($G$-steerable).

In accordance with the previous section, we clearly distinguish between the requirements for coordinate independence and weight sharing.
Section~\ref{sec:kernel_field_trafos} therefore starts by discussing fields of kernels and their transformations laws without demanding the kernels at individual positions to be tied together.
Such unrestricted kernel fields give rise to \emph{kernel field transforms}, which are integral transforms that can be seen as precursors of convolutions.
The actual $\GM$-convolutions, which are parameterized by a shared, necessarily gauge equivariant template kernel, are defined in Section~\ref{sec:gauge_conv}.
As a preparation, we will in the following Section~\ref{sec:observers_view} describe local representations of feature fields on the tangent spaces, where they will be matched with the convolution kernels.

\subsubsection{A local observer's view on feature fields}
\label{sec:observers_view}

\begin{figure}
    \centering
    \includegraphics[width=\textwidth]{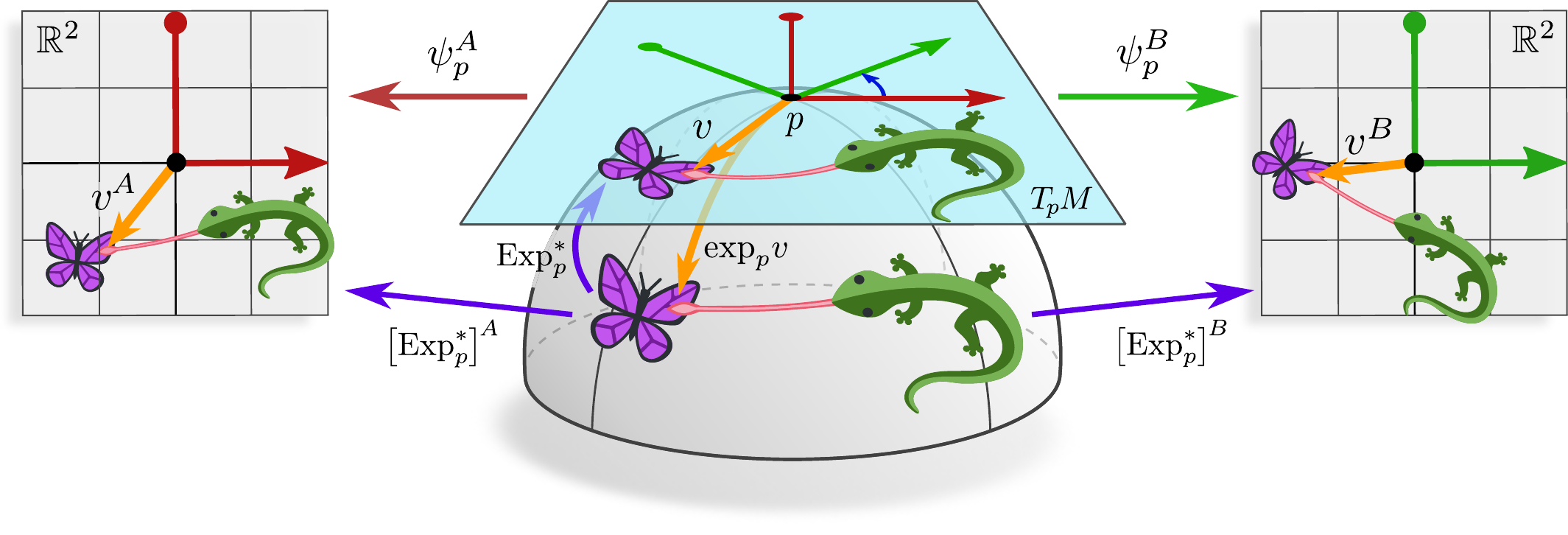}
    \caption{\small
        A feature field $f$ on $M$ and its local representation $\Expspf$ on $\TpM$ via the \emph{transporter pullback} $\Expsp$.
        Just like the usual pullback $\exp_p^*f$ of~$f$ along the exponential map $\exp_p: \TpM \to M$, the transporter pullback assigns feature vectors $f(\exp_p(v))$ to tangent vectors $v\in\TpM$.
        However, as we aim to accumulate the pulled back features by means of a convolution kernel, they need to be given in the same space and be expressed relative to the same gauge at~$p$.
        The transporter pullback therefore additionally applies the ($G$-compatible) parallel transporter along the geodesic from $\exp_p(v)$ to~$p$.
        Via a gauge $\psi_p^X$, the transporter pullback of $f$ on $\TpM$ can be expressed on $\R^d$ as $[\Expspf]^X: \R^d \to \R^c$ -- different choices of reference frames (observers) correspond hereby to different linear deformations of the feature field.
        Kernel field transforms and $\GM$-convolutions compute an output feature~$\fout(p)$ at~$p$ by matching a kernel~$\Kp$ on $\TpM$ with $\Expspf$ (i.e. integrate their product over the tangent space; see Eq.~\eqref{eq:kft_coord_expression}).
        { \\
        \color{gray}
        \scriptsize
            (Lizards and butterflies adapted under the Creative Commons Attribution 4.0 International
            \href{https://github.com/twitter/twemoji/blob/gh-pages/LICENSE-GRAPHICS}{\underline{license}}
            by courtesy of Twitter.)
        }
    }
    \label{fig:pullback_field_exp_TpM}
\end{figure}

In contrast to Euclidean spaces or more general homogeneous spaces like the sphere, the local geometry of a general Riemannian manifold varies from point to point.
It is therefore not immediately clear how a convolution kernel should be defined on~$M$ and how it could be shared between different locations.
A~common solution is to define the kernel as usual on a flat, Euclidean vector space~$\R^d$, and to share it over the tangent spaces instead of the manifold itself;
see Sections~\ref{sec:kernel_field_trafos} and \ref{sec:gauge_conv} or prior work~%
\cite{masci2015geodesic,poulenard2018multi,sun2018zernet,coors2018spherenet,gaugeIco2019,Wiersma2020,deHaan2020meshCNNs,Yang2020parallelFrameCNN}.
Subsequently, the kernel can via the Riemannian exponential map be mapped down to the manifold.
It can be thought of as being applied by a local observer, who is measuring features in its surrounding relative to its local reference frame.
We will in this section shortly elaborate on how feature fields are perceived from the perspective of different local observers.
Mathematically, this is formalized as the pullback and parallel transport of the feature field to the tangent spaces; see Fig.~\ref{fig:pullback_field_exp_TpM} for a visualization.

In order to map between the tangent spaces and the manifold, we consider the \emph{Riemannian exponential map} (corresponding to the Levi-Civita connection).%
\footnote{
    Even models which assume an \emph{alternative ($G$-compatible) connection to transport features} utilize usually the canonical \emph{Levi-Civita connection to compute geodesics} and exponential maps.
}
Assuming the manifold for simplicity to be geodesically complete%
\footnote{
    The assumption that $M$ is \emph{geodesically complete} means that the exponential maps $\exp_p$ are for each $p\in M$ defined on the whole tangent space $\TpM$.
    In cases where this assumption is violated one can resort to \emph{zero padding}, which is commonly used in convolutional networks for finitely supported images.
},
the exponential map at a specific point~$p\in M$ is a map
\begin{align}
    \exp_p: \TpM \to M \,.
\end{align}
It identifies vectors $v\in \TpM$ with those points $\exp_p(v) \in M$ that are reached when following the geodesic through~$p$ with an initial velocity of~$v$ for one unit of time.
While preserving radial distances, the exponential map does in general distort angels and fails to be injective.
For instance, if the manifold is a sphere, the exponential maps wrap their corresponding tangent space infinitely often around it.
It is, however, guaranteed that the exponential map is a local diffeomorphism if its domain is restricted to distances shorter than the distance to the cut locus (where injectivity fails).

Given the exponential maps, one can pull feature fields on the manifold back to the tangent spaces.
Specifically, let~$f$ be some feature field on~$M$, then the \emph{pullback} $\exp_p^*f := f \circ \exp_p$ is defined as that map that assigns the feature vector $f(\exp_p(v))$ from $\exp_p(v)$ to $v\in \TpM$.
Note that, due to the missing injectivity of the exponential map, each tangent vector might be assigned to multiple tangent vectors $v_1$ and $v_2$ if $\exp_p(v_1) = \exp_p(v_2)$ \:--\, this is somewhat similar to gravitational lensing effects in physics.
For the case that the exponential map is injective, or when restricting it to its injectivity radius, the pullback corresponds to an expression of the feature fields in \emph{geodesic normal coordinates}~\cite{masci2015geodesic}.

Recall that the purpose of pulling the feature vectors back to the tangent spaces is to enable that they can be accumulated by a convolution kernel.
Unfortunately, this is not immediately possible since the feature vectors at different locations live in different vector spaces and are expressed relative to different gauges.%
\footnote{
    A very similar circumstance motivates the definition of covariant derivatives, which also needs to combine geometric objects that live in different spaces.
}
It is therefore necessary to express all feature vectors $[\exp_p^*f](v)$ in the same vector space and relative to the same gauge.
A natural idea, proposed by~\citet{poulenard2018multi}, is to do this by \emph{parallel transporting} the feature vectors along the geodesics that define the exponential map from $\exp_p(v)$ to~$v$.%
\footnote{
    The parallel transport along any other path would be equally valid.
}
We denote this pullback of~$f$ with additional transport as $\Expspf$ to emphasizes its close relation to the usual pullback $\exp_p^*f$ to $\TpM$.
Fig.~\ref{fig:pullback_field_exp_TpM} gives a visual idea of this \emph{transporter pullback} of feature fields to the tangent space and its representations $\big[\Expspf\big]^A$ and $\big[\Expspf\big]^B$ on $\R^d$ relative to different coordinatizations.

We formalize $\Expspf$ by defining it in terms of its coordinate expression relative to some choice of gauge.
To this end, let $\psi_p^A$ be a gauge at $p$, relative to which the transported features will ultimately be expressed and let $\psi_{\exp_p(v)}^{\widetilde{A}}$ be an arbitrary gauge at $\exp_p(v)$, which represents the feature vector at that location by a coefficient vector $f^{\widetilde{A}}(\exp_p(v)) \in \R^c$.
Denote by 
\begin{align}
    \rho \big( g^{A\widetilde{A}}_{p\leftarrow\exp_p\!v} \big)
\end{align}
the $G$-compatible parallel transporter of feature vector coefficients along the geodesic from $\exp_p(v)$ to~$p$.
Then we define the \emph{transporter pullback} in coordinates as%
\begin{align}\label{eq:transporter_pullback_in_coords}
    \big[\mkern-2mu \Expspf \big]^A:\ \R^d \to \R^c,\quad v^A &\mapsto\ \big[\mkern-2mu \Expspf \big]^A (v^A)
        \notag \\[1.5ex]
        & \,:=\ 
        \rho\pig( g^{A\widetilde{A}}_{p \,\leftarrow\, \exp_p (\psi_p^A)^{\shortminus1}(v^A)} \pig) \cdot
        f^{\widetilde{A}} \pig( \exp_p \big(\psi_p^A\big)^{\mkern-2mu-1}(v^A) \pig) \,,
\end{align}
where $v = \big(\psi_p^A\big)^{-1} (v^A) \in \TpM$ is the coordinate free tangent vector referred to by the coefficients $v^A$ via~$\psi_p^A$.
As claimed before, the choice of gauge $\psi_{\exp_p(v)}^{\widetilde{A}}$ at $\exp_p(v)$ is by the coordinate independence of all equations irrelevant and cancels out.
Specifically, one could have used any other gauge $\psi_{\exp_p(v)}^{\widetilde{B}}$ at $\exp_p(v)$, implying gauge transformations
$
    \rho\big( g^{A\widetilde{B}}_{p \,\leftarrow\, \exp_p(v)} \big)
  = \rho\big( g^{A\widetilde{A}}_{p \,\leftarrow\, \exp_p(v)} \big)
    \rho\big( g_{\exp_p(v)}^{\widetilde{B}\widetilde{A}} \big)^{-1}
$
of the transporter by Eq.~\eqref{eq:transporter_gauge_trafo} and
$
    f^{\widetilde{B}} \big( \exp_p(v) \big)
  = \rho\big( g_{\exp_p(v)}^{\widetilde{B}\widetilde{A}} \big)
    f^{\widetilde{A}} \big( \exp_p(v) \big)
$
of the feature vector coefficients by Eq.~\eqref{eq:gauge_trafo_features}, which annihilate when composing both expressions.

The transporter pullback $[\Expspf]^A$ depends, however, still on the gauge at $p$, and therefore transforms under gauge transformations $g_p^{BA}$ at~$p$.
As for any coordinatized function, its transformation law is determined by the gauge transformations on its domain~$\R^d$ and codomain~$\R^c$.
It is therefore given by
\begin{align}\label{eq:trafo_law_transporter_pullback}
    \big[ \Expspf \big]^B\ =\ \rho\big( g_p^{BA} \big) \circ \big[ \Expspf \big]^A \circ \big(g_p^{BA} \big)^{-1} \,,
\end{align}
which is summarized by the following commutative diagram:
\begin{equation}\label{cd:}
\begin{tikzcd}[column sep=70pt, row sep=32pt, font=\normalsize]
    \R^d
        \arrow[d, "g_p^{BA} \cdot\,"']
        \arrow[r, "{\big[\mkern-1mu \Expspf \big]^A}"]
    &
    \R^c
        \arrow[d, "\ \rho\big(g_p^{BA}\big) \cdot"]
    \\
    \R^d
        \arrow[r, "{\big[\mkern-1mu \Expspf \big]^B}"']
    &
    \R^c
\end{tikzcd}
\end{equation}
As visualized in Fig.~\ref{fig:pullback_field_exp_TpM}, $\big[\Expspf\big]^A$ and $\big[\Expspf\big]^B$ should be thought of as the perspective of different local observers (reference frames) on the feature field.

In principle, one could consider alternative constructions for the pullback of feature fields from~$M$ to~$\TpM$.
Our definition of kernel field transforms and $\GM$-convolutions in Sections~\ref{sec:kernel_field_trafos} and~\ref{sec:gauge_conv} below is independent from this particular choice.

\subsubsection{Coordinate independent kernels and kernel field transforms}
\label{sec:kernel_field_trafos}

$\GM$-convolutions are coordinate independent operations which apply the same, shared kernel at each point of the manifold.
To clearly separate the assumptions being made, we first discuss more general \emph{kernel field transforms}, which are coordinate independent operations but drop the requirement of weight sharing.
They are therefore similar to $\GM$-convolutions but apply a potentially different kernel $\Kp$ to each point~$p$ of the manifold.
In order to respect the principle of covariance, the coordinate expressions of those kernels are required to transform in a principled manner, however, the kernels themselves are left unconstrained.

\begin{figure}
    \centering
    \includegraphics[width=.9\columnwidth]{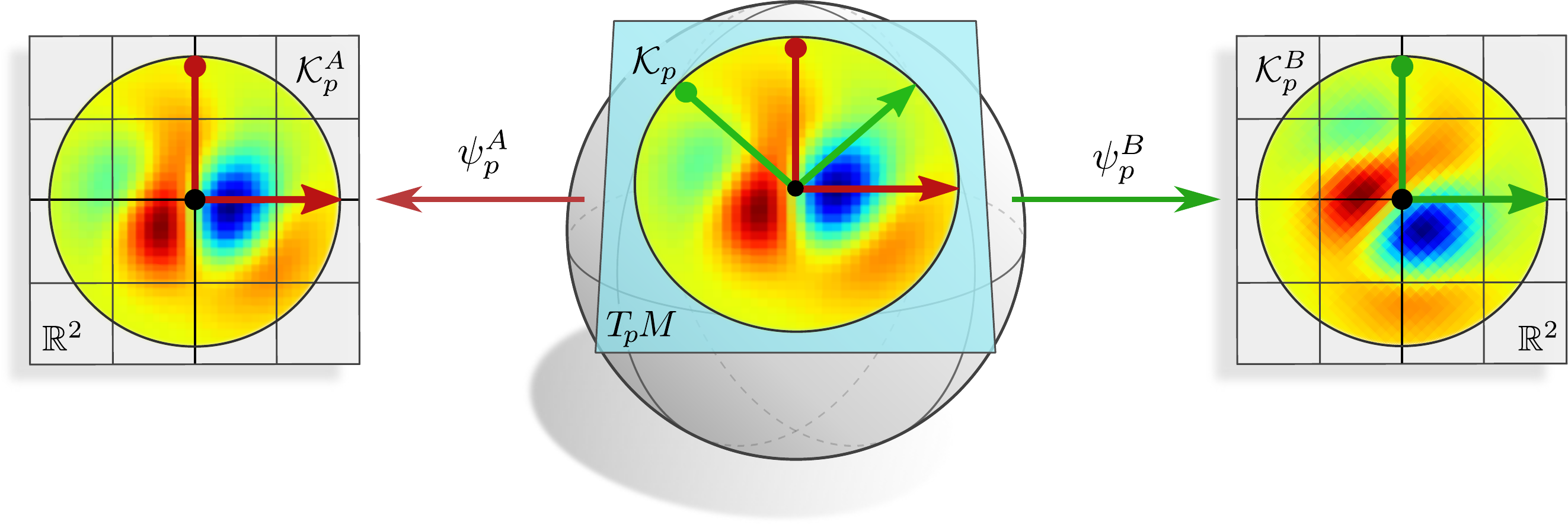}
    \caption{\small
        A coordinate free kernel $\Kp$ on $\TpM$ and its coordinate expressions ${\Kp^X \!: \R^d \to \R^{\cout\times\cin}}$ relative to gauges $\psi_p^X$ (only one of the ${\cout \times \cin}$ kernel channels is shown).
        The gauge transformations that relate different coordinatizations of a kernel follow from the transformation laws of their domain $\R^d$ and codomain $\R^{\cout\times\cin}$.
        They are therefore for any $\mathscr{v} \in \R^d$ given by
        $\Kp^B\big(g_p^{BA} \mathscr{v}\big) = \rhoout\big(g_p^{BA}\big) \:\Kp^A (\mathscr{v})\: \rhoin\big(g_p^{BA}\big)^{-1}$.
        A kernel field $\K$ on $M$ is a smooth assignment of kernels over the tangent spaces (Def.~\ref{dfn:kernel_field_general}).
        \\
        Note that we are here assuming the kernel on $\TpM$ to be given and express it subsequently relative to different gauges on $\R^d$.
        This is conceptually different from the situation depicted in
        Figs.~\ref{fig:satellite}, \ref{fig:intro_kernel_alignment_trivial}, \ref{fig:gauge_trafos_feature_vector} and~\ref{fig:kernel_apx_sharing},
        where we assume a template kernel~$K$ to be given on $\R^d$ and subsequently define $\Kp$ on $\TpM$ via convolutional weight sharing relative to some reference frame.
        In order to preserve coordinate independence during the weight sharing process, the shared kernel needs to be \emph{invariant} (or equivariant) under gauge transformations; see Section~\ref{sec:gauge_conv} and Appendix~\ref{apx:coord_indep_weight_sharing}.
    }
    \label{fig:kernel_coordinatization}
\end{figure}

\paragraph{Coordinate independent kernels:}
Since convolutions in deep learning map between fields of feature vectors of dimensionalities $\R^\cin$ and $\R^\cout$, the convolution kernels are ${\cout \mkern-3mu\times\mkern-1.5mu \cin}$ matrix-valued.
Discretized implementations of $d$-dimensional convolutions on Euclidean spaces typically represent such kernels as arrays of shape $(s_1,\dots,s_d, \; c_\text{out},c_\text{in} \mkern1mu)$.
The first $d$~axes represent hereby a spatial grid of $s_1 \times\dots\times s_d$ pixels, each of which is assigned a ${\cout \mkern-3mu\times\mkern-1.5mu \cin}$ matrix, encoded in the last two axes.%
\footnote{
    The actual memory layout depends on the particular deep learning framework in consideration.
}
In the continuous, Euclidean setting, such kernels can be described as maps
\begin{align}\label{eq:conv_kernel_unrestricted}
    K: \R^d \to \R^{\cout\times\cin} \,,
\end{align}
which assign a ${\cout \mkern-3mu\times\mkern-1.5mu \cin}$ matrix to each point of $\R^d$.
As mentioned in the previous Section~\ref{sec:observers_view}, we define $\GM$-convolutions as matching the transporter pullback $\Expspfin$
on the tangent space $\TpM$ with a kernel $\Kp$ on~$\TpM$.
Since the tangent spaces are flat, it is natural to define this matching as in the usual, fully Euclidean setting.
We do therefore define the kernels $\Kp$ via their coordinate expressions, which take the form in Eq.~\eqref{eq:conv_kernel_unrestricted}, that is,
\begin{align}
    \Kp^A: \R^d \to \R^{\cout\times\cin} \,.
\end{align}
Fig.~\ref{fig:kernel_coordinatization} shows a \emph{given} coordinate free kernel on~$\TpM$ and its representations on $\R^d$ relative to different reference frames.%
\footnote{
    We emphasize that we are here assuming a coordinate free kernel $\Kp$ which is \emph{given} on $\TpM$ and consider its coordinate expressions $\Kp^X$ on $\R^d$ relative to reference frames~$X$.
    Convolutional weight sharing will later on pose us with the question of how to \emph{define} a coordinate free kernel $\Kp$ on $\TpM$ given a template kernel $K$ on $\R^d$.
    Appendix~\ref{apx:coord_indep_weight_sharing} elaborates on these two concepts and their relation to the kernel's $G$-steerability.
}

The transformation law between the coordinate representations $\Kp^A$ and $\Kp^B$ of a kernel $\Kp$ on $\TpM$ follows as usual from the transformation laws of their domain and codomain.
On the domain $\R^d$ the transformation law is given by $g_p^{BA}$, while the transformation law of $\R^{\cout\times\cin}$ is, as in Eq.~\eqref{eq:linear_op_trafo_law}, given by a simultaneous left multiplication with $\rhoout\big(g_p^{BA}\big)$ and right multiplication with $\rhoin\big(g_p^{BA}\big)^{-1}$.
The two coordinatizations of the kernel $\Kp$ relate thus for any $\mathscr{v} \in \R^d$ by
\begin{align}\label{eq:kernel_trafo_law}
    \Kp^B\big(g_p^{BA} \mathscr{v}\big)\ =\ 
    \rhoout\big(g_p^{BA}\big) \cdot\mkern1mu 
    \Kp^A (\mathscr{v})
    \mkern1mu\cdot \rhoin\big(g_p^{BA}\big)^{-1} \,,
\end{align}
which is visualized by the following commutative diagram:
\begin{equation}\label{cd:kernel_trafo_law}
\qquad
\begin{tikzcd}[column sep=70pt, row sep=32pt, font=\normalsize]
    \R^d
        \arrow[d, "g_p^{BA} \cdot\,"']
        \arrow[r, "\Kp^A"]
    &
    \R^{\cout\times\cin}
        \arrow[d, "\ \rhoout\big(g_p^{BA}\big) \; \scalebox{1.1}{$[\,\cdot\,]$} \; \rhoin\big(g_p^{BA}\big)^{-1}"]
    \\
    \R^d
        \arrow[r, "\Kp^B"']
    &
    \R^{\cout\times\cin}
\end{tikzcd}
\end{equation}
As in the examples from Section~\ref{sec:pointwise_operations}, the principle of covariance only requires a consistent transformation behavior between different kernel coordinatizations but does not lead to a constraint on the kernel itself.
One might therefore parameterize the kernels $\Kp$ for any $p\in M$ and an arbitrary gauge at $p$ by some unrestricted, matrix-valued kernel.
We denote smooth fields of such kernels as kernel fields, which play a major role in our analysis of the isometry equivariance of $\GM$-convolutions in Section~\ref{sec:isometry_intro}.

\paragraph{Coordinate independent kernel field transforms:}
Given a smooth kernel field $\K$, we can define \emph{kernel field transforms}, which are similar to convolutions but differ in that they might apply a different kernel at each spatial position.
They compute a field of output feature vectors $\fout(p)$ by
integrating the product of the corresponding kernel $\Kp$ and transporter pullback $\Expspfin$ of $\fin$ over $\TpM$, that~is,
\begin{align}\label{eq:kft_coord_free}
    \fout(p)\ =\ 
    \int_{\TpM}
    \Kp(v) \,
    \Expspfin(v) \;
    dv \,.
\end{align}
To express this coordinate free definition in terms of coordinates, one has to replace all quantities by their coordinate expressions and to pull the integration via the chosen gauge from $\TpM$ to $\R^d$.
As described in Appendix~\ref{apx:tangent_integral}, the appropriate (gauge invariant) Riemannian volume element is for a gauge $\psi_p^A$ given by
\begin{align}
    \sqrt{|\eta_p^A|} \,\ dv^A ,
\end{align}
where the factor $\sqrt{|\eta_p^A|}$, defined in Eq.~\eqref{eq:volume_element_def}, is the (positive) volume spanned by the reference frame $[e_i^A]_{i=1}^d$ at~$p$.
The coordinate expression of the kernel field transform thus reads
\begin{align}\label{eq:kft_coord_expression}
    \fout^A(p)
    \ =&\ 
        \int_{\R^d}
        \Kp^A(v^A) \ 
        \big[\mkern-2mu \Expspfin\big]^A \mkern-1mu(v^A) \ 
        \sqrt{|\eta_p^A|}\ dv^A
    \,.
\end{align}

The coordinate independence of the kernel field transform is asserted by expressing it relative to an alternative gauge $\psi_p^B$ and showing that the resulting output field transforms as expected, which is indeed the case:
\begin{align}\label{eq:kft_coordinate_independence}
    \fout^B(p)\ 
    \overset{(1)}{=}&\ 
        \int_{\R^d}
        \Kp^B(v^B) \ 
        \big[\mkern-2mu \Expspfin\big]^B \mkern-2mu(v^B) \,\ 
        \sqrt{|\eta_p^B|}\ dv^B
    \notag \\ 
    \overset{(2)}{=}&\ 
        \int_{\R^d}
        \Big[ \rhoout\big(g_p^{BA}\big)\, \Kp^A\pig(\! \big(g_p^{BA}\big)^{-1} v^B \big)\!\pig)\, \rhoin\big(g_p^{BA}\big)^{-1} \Big] \,\ 
        \big[\mkern-1mu \Expspfin\big]^B \mkern-2mu(v^B) \,\ 
        \sqrt{|\eta_p^B|}\ dv^B
    \notag \\ 
    \overset{(3)}{=}&\ \ 
        \rhoout\big(g_p^{BA}\big)
        \int_{\R^d}
        \Kp^A(v^A) \,\ 
        \Big[ \rhoin\big(g_p^{BA}\big)^{-1} \big[\mkern-1mu \Expspfin\big]^B \mkern-2mu \big( g_p^{BA} v^A\big) \Big] \ 
        \sqrt{|\eta_p^A|}\ dv^A
    \notag \\ 
    \overset{(4)}{=}&\ \ 
        \rhoout\big(g_p^{BA}\big)
        \int_{\R^d}
        \Kp^A(v^A) \,\ 
        \big[\mkern-1mu \Expspfin\big]^A \mkern-1mu (v^A) \ 
        \sqrt{|\eta_p^A|}\ dv^A
    \notag \\ 
    \overset{(5)}{=}&\ \ 
    \rhoout\big(g_p^{BA}\big) \, \fout^A(p)
\end{align}
Here we used the definition of kernel field transforms and the transformation law of kernels (Eq.~\eqref{eq:kernel_trafo_law}) in the first two steps.
The third step follows by pulling $\rhoout$ out of the integral and substituting $v^B$ with $v^A = \big(g_p^{BA}\big)^{-1} v^B$, using that the volume element $\sqrt{|\eta_p^B|}\ dv^B = \sqrt{|\eta_p^A|}\ dv^A$ is by design gauge invariant.
The last two steps follow then by identifying the transformation law of the transporter pullback of the feature field in Eq.~\eqref{eq:trafo_law_transporter_pullback} and the definition of the kernel field transform in gauge $\psi_p^A$.
Note that the coordinate independence of the kernel field transform affirms the correctness of the kernel transformation law in Eq.~\eqref{eq:kernel_trafo_law}.

A kernel field transform is only well defined if the integrals over the tangent spaces converge, which is more rigorously discussed in Section~\ref{sec:global_conv} and Appendix~\ref{apx:smoothness_kernel_field_trafo}.
Theorem~\ref{thm:existence_kernel_field_trafo_compact_kernels} proves that a compact support of the kernels $\Kp$ is sufficient to guarantee this well-definedness.
It further proves that kernel field transforms that are based on smooth kernel fields will map smooth input feature fields to smooth output feature fields.

\subsubsection{\textit{GM}-convolutions and \textit{G}-steerable kernels}
\label{sec:gauge_conv}

The freedom of kernel field transforms to apply a different kernel at each location does not allow them to generalize learned inference over different locations and thus makes them data inefficient.
One does therefore typically consider convolutions, which can be seen as those specific kernel field transforms that are based on convolutional kernel fields, i.e. kernel fields that are parameterized by a single, shared template kernel.
As before, a coordinate independent weight sharing requires the template kernels to be gauge equivariant ($G$-steerable).
This gauge equivariance of the template kernels implies that patterns which appear in different, $G$-related geometric poses are guaranteed to evoke the same response up to a corresponding transformation of the feature vector via $\rhoout$.

\paragraph{Convolutional weight sharing:}
Let $K: \R^d \to \R^{\cout\times\cin}$ be a template kernel to be shared over all tangent spaces.
In order to not prefer any particular gauge -- which would contradict our requirement for coordinate independence -- we are forced to share the kernel with coordinatizations in all gauges simultaneously.
Naively, this seems to suggest to share the template kernel by setting $\Kp^X = K$ for any point $p\in M$ and any gauge $\psi_p^X$ at~$p$.
While such a definition of kernel sharing seems reasonable, it does not follow our principle of sharing local template functions in a strict sense:
instead of directly sharing the kernel, it is important to share the \emph{whole} local operation -- which is here the whole 
integral transform in Eq.~\eqref{eq:kft_coord_expression}.
Since this operation is parameterized in terms of the kernel field $\K$, this leads indirectly to a sharing of the template kernel, however, with a slightly different result as the naive sharing considered above.

To find the correct definition of $\GM$-convolutional kernel fields according to our principle of sharing local template functions, we first need to identify these local operations.
We do this by abstracting kernel field transforms (in coordinates) as a collection of local integral operators of the form
\begin{align}\label{eq:local_integral_operator_general}
    \mathscr{I}_{\K,p}^A:\ 
    C^\infty \big(\R^d, \R^c\big) \to \R^c, \quad
    F \,\mapsto \int_{\R^d} \Kp^A(\mathscr{v})\, F(\mathscr{v})\, \sqrt{|\eta_p^A|}\ d\mathscr{v} \,,
\end{align}
where $C^\infty \big(\R^d, \R^c\big)$ denotes the space of smooth maps from $\R^d$ to $\R^c$.
In our application, these smooth maps are just the local feature field representations $[\Expspf]^A: \R^d \to \R^c$ as seen from the tangent spaces at~$p$, which are by the kernel field transform mapped to an output feature vector $\fout^A(p) = \mathscr{I}_{\K,p}^A \big([\Expspf]^A\big)$ at~$p$.
Given our template kernel $K: \R^d \to \R^{\cout\times\cin}$, we define a corresponding integral operator template
\begin{align}\label{eq:local_integral_operator_template}
    \mathfrak{I}_K:\ 
    C^\infty \big(\R^d, \R^c\big) \to \R^c, \quad
    F \,\mapsto \int_{\R^d} K(\mathscr{v})\, F(\mathscr{v})\ d\mathscr{v} \,,
\end{align}
which multiplies the local field representation $F$ with the template kernel $K$ and then integrates their product.
Note that $\mathfrak{I}_K$ is as a template function necessarily agnostic to specific choices of gauges and does therefore not involve a frame volume factor.
A $\GM$-coordinate independent convolutional weight sharing scheme is imposed by demanding that this template functional agrees with all the individual integral operators at any point and in any gauge, that is,
\begin{align}\label{eq:weight_sharing_integral_operator}
    \mkern28mu
    \mathscr{I}_{\K,p}^X = \mathfrak{I}_K
    \quad \textup{for \emph{any} gauge}\ \ \big(U^X,\psi^X) \in \mathscr{A}^G\ \ \textup{with}\ \ p\in U^X \,,
\end{align}
where $\mathscr{A}^G$ is the (maximal) $G$-atlas corresponding to the considered $G$-structure; see Eq.~\eqref{eq:G_atlas_dfn}.
This is equivalent to directly sharing the local template kernel according to
\begin{align}\label{eq:weight_sharing_kernel}
    \Kp^X = \frac{K}{\sqrt{|\eta_p^X|}\,}
    \quad \textup{for \emph{any} gauge}\ \ \big(U^X,\psi^X) \in \mathscr{A}^G\ \ \textup{with}\ \ p\in U^X \,,
\end{align}
where the normalization factor reduces the ``kernel density'' by the reference frame volume $\sqrt{|\eta_p^X|}$.
As discussed below, this normalization is important for the equivariance under non-volume-preserving symmetry groups

We denote kernel fields which are parameterized by a shared kernel $K$ according to Eq.~\eqref{eq:weight_sharing_kernel} as $\GM$-\emph{convolutional kernel fields}.
The simultaneous requirement for weight sharing and coordinate independence leads to an equivariance constraint on the template kernels.
To derive this constraint, insert the kernel sharing in Eq.~\eqref{eq:weight_sharing_kernel} into the kernel transformation law in Eq.~\eqref{eq:kernel_trafo_law}, which results in
\begin{align}
    \frac{1}{\sqrt{|\eta_p^B|\,}}\ K\big(g_p^{BA} \mathscr{v}\big)
    \ =\ 
    \frac{1}{\sqrt{|\eta_p^A|\,}}\ 
    \rhoout\big(g_p^{BA}\big) \cdot\mkern1mu 
    K(\mathscr{v})
    \mkern1mu\cdot \rhoin\big(g_p^{BA}\big)^{-1} \,.
\end{align}
Since the volumes of different reference frames are related by
$\sqrt{|\eta_p^A|} = \big|\mkern-2mu \det(g_p^{BA})\big|\, \sqrt{|\eta_p^B|}$
and since the transformation law needs to hold for arbitrary $G$-related gauges, this implies the \emph{$G$-steerability} constraint%
\footnote{
    In contrast to prior work~\cite{3d_steerableCNNs,Weiler2019_E2CNN,gaugeIco2019,kicanaoglu2019gaugeSphere,deHaan2020meshCNNs}, this constraint contains the factor $\detg$.
    It did not appear in these works since they all considered (subgroups of) orthonormal structure groups $\O{d}$, for which the determinant factor vanishes.
}
\begin{align}\label{eq:kernel_constraint}
    K(g\mkern1.5mu \mathscr{v})\ = \ \frac{1}{\detg}\, \rhoout(g) \cdot K(\mathscr{v}) \cdot \rhoin(g)^{-1}
    \qquad \forall\ \ \mathscr{v} \in \R^d,\ \ g\in G \,.
\end{align}
on template kernels.
As proven by~\citet{lang2020WignerEckart}, this constraint requires template kernels to be \emph{representation operators}~\cite{jeevanjee2011reprOp} (generalizations of e.g. spherical tensor operators in quantum mechanics).
Diagrammatically, a $G$-steerable kernels $K$ is required to satisfy the commutativity of
\begin{equation}\label{cd:kernel_steerability}
\qquad\qquad
\begin{tikzcd}[column sep=70pt, row sep=35pt, font=\normalsize]
    \R^d
        \arrow[d, "g\mkern2mu \cdot\,"']
        \arrow[r, "K"]
    &
    \R^{\cout\times\cin}
        \arrow[d, "\ \scalebox{.96}{$\displaystyle \frac{1}{\detg}$}\, \rhoout(g) \; \scalebox{1.1}{$[\,\cdot\,]$} \; \rhoin(g)^{-1}"]
    \\
    \R^d
        \arrow[r, "K"']
    &
    \R^{\cout\times\cin}
\end{tikzcd}
\end{equation}
for any $g\in G$.
Note that the inverse determinant factor $\detg$ in the kernel's transformation law makes it transform like a \emph{matrix-valued $-1$-density}; see Table~\ref{tab:density_factors} for more details.
Intuitively, $G$-steerable kernels are exactly those kernels that can be shared relative to arbitrary $G$-related reference frames without that the particular choice of gauge would influence the result.%
\footnote{
    The $G$-steerability constraint can be rewritten as
    ${K(\mathscr{v}) = \detg^{\minus1} \rhoout(g) \mkern-1mu\cdot\mkern-1.5mu K(g^{\minus1}\mathscr{v}) \mkern-1mu\cdot\mkern-1.5mu \rhoin(g)^{\minus1}}$
    $\mkern8mu \forall\ \mathscr{v} \in \R^d,\ g\in G$,
    which emphasizes that $G$-steerable kernels are the \emph{invariants} under the gauge action on the right-hand-side.
    Being invariant under gauge transformations, a $G$-steerable kernel leads to the same coordinate free kernel $\Kp$ at $p$ when being shared relative to any reference frame in~$\GpM$.
}
The ambiguity of kernel alignments -- which motivated this work in the first place -- is thus resolved by additional weight sharing over all the equivalent reference frames (all gauges) in the considered $G$-structure~$\GM$.

\begin{table}
    \centering
    \setlength\aboverulesep{0pt}
    \setlength\belowrulesep{0pt}
    \renewcommand{\arraystretch}{1.8}
    \setlength{\tabcolsep}{2.4ex}
    \scalebox{.95}{
    \begin{tabular}{l|ccccccc}
        object     & $\Kp^X$ & $K$  & $\big[\mkern-2mu\Expspf\big]^{\mkern-2muX}\mkern-2mu$ and $F$ & $\sqrt{|\eta^X|}$ & $dv^X\!$ and $d\mathscr{v}$ \\
       \midrule
       density $s$ & $0$     & $-1$ & $0$                                                           & $-1$              & $1$
    \end{tabular}
    }
    \vspace*{2ex}
    \caption{
        An overview of the density exponents $s$ of different objects involved in general kernel field transforms and $\GM$-convolutions.
        The coordinate expression of an $s$-density transforms with a factor of $\detg^s$ when the coordinates are transformed via $g\in G$.
        A general matrix-valued kernel $\Kp^X$ is according to Eq.~\eqref{eq:kernel_trafo_law} a $0$-density.
        The same holds for feature fields and their pullbacks, whose transformation laws are given in Eqs.~\eqref{eq:gauge_trafo_features} and~\eqref{eq:trafo_law_transporter_pullback}.
        The whole integrand $\Kp^X(v^X) [\Expspf]^{\mkern-1muX}(v^X) \sqrt{|\eta^X|}\, dv^X$ of a general kernel field transforms in Eq.~\eqref{eq:kft_coord_expression} is seen to be a $0$-density as well -- note that this is necessary for its coordinate independence as demonstrated in Eq.~\eqref{eq:kft_coordinate_independence}.
        As the integral operator template $\mathfrak{I}_K$ in Eq.~\eqref{eq:local_integral_operator_template} is agnostic of any choice of gauge, it does not involve the frame volume factor $\sqrt{|\eta^X|}$.
        Since it should nonetheless behave like the integral operators $\mathscr{I}_{\K,p}^X$ underlying kernel field transforms, the whole integrand $K(\mathscr{v}) F(\mathscr{v})\, d\mathscr{v}$ of $\mathfrak{I}_K(F)$ is required to be a $0$-density.
        This necessitates the shared template kernels $K$ themselves to transform like $-1$-densities, which is reflected in the $G$-steerability constraint in Eq.~\eqref{eq:kernel_constraint}.
        Note that this transformation law of template kernels is strictly necessary for the local $G$-equivariance of $\GM$-convolutions if the output features should transform like densities of weight~$0$; see Eq.~\eqref{eq:active_local_gauge_trafo}.
        For an alternative perspective, we point the interested reader to Corollary~1 in~\cite{bekkers2020bspline}, where the determinant factor is derived from Haar measures on Lie groups.
    }
    \label{tab:density_factors}
\end{table}

Before coming to $\GM$-convolutions, we comment on the space of $G$-steerable kernels.
Note that the set
\begin{align}\label{eq:unconstrained_kernel_space}
    \mathscr{K}\ :=\ \pig\{ K\!: \R^d \to \R^{\cout\times\cin} \pig\} \,.
\end{align}
of general, i.e. not necessarily equivariant kernels forms a vector space when being equipped with the standard summation and scalar multiplication on $\R^{\cout\times\cin}$.
Since the $G$-steerability constraint in Eq.~\eqref{eq:kernel_constraint} is linear, it restricts the kernel space to a \emph{linear subspace}
\begin{align}\label{eq:G-steerable_kernel_space}
    \KG\, :=\ \Big\{ K\!: \R^d \to \R^{\cout\times\cin} \,\Big|\,
    K(g\mkern1mu \mathscr{v}) = \frac{1}{\detg}\, \rhoout(g) \mkern-1.5mu\cdot\mkern-1.5mu K(\mathscr{v}) \mkern-1.5mu\cdot\mkern-1.5mu \rhoin(g)^{-1} \ \ \ \forall\,\ \mathscr{v}\in \R^d,\,\ g\in G \Big\} \,.
\end{align}
It is therefore possible to solve for a basis of $G$-steerable kernels, in terms of which the $\GM$-convolution can be parameterized.
While this space is in theory usually infinite-dimensional, it is in practice often being discretized, such that one ends up with a finite basis $\{K_1,\dots,K_N\}$ of $G$-steerable kernels.
A~set $\{w_1,\dots,w_N\}$ of real-valued weights $w_i\in\R$, which are optimized during the training process, then parameterize the convolution with $K = \sum_{i=1}^N w_i K_i$.
Note that the reduced dimensionality of the (sub)space of $G$-steerable kernels implies an improved parameter efficiency in comparison to conventional convolutions.

Section~\ref{sec:mobius_kernel_spaces} discusses exemplary analytical solutions of reflection equivariant kernels spaces for different group representations of the reflection group~$\Flip$.
The resulting kernels, which are characterized by different types of reflectional symmetries, are visualized in Table~\ref{tab:reflection_steerable_kernels}.

Further examples can be found in the literature on steerable CNNs:
an analytical solution of the kernel space constraint for the special orthogonal structure group $\SO3$ in three dimensions and its irreducible representations was presented by~\citet{3d_steerableCNNs}.
\citet{Weiler2019_E2CNN} generalized this approach to cover arbitrary group representations and solved the kernel space constraint for any representation of $\O2$ and all of its subgroups $G\leq\O2$
-- an implementation is available at \url{https://quva-lab.github.io/e2cnn/api/e2cnn.kernels.html}.
For finite structure groups the constraint might alternatively be solved numerically as explained by~\citet{Cohen2017-STEER}.
A more general solution strategy, applying to arbitrary compact structure groups~$G$ (and thus all above mentioned cases), was proposed by~\citet{lang2020WignerEckart}.
This solution generalizes the classical \emph{Wigner-Eckart} theorem~\cite{agrawalla1980WignerEckart,jeevanjee2011reprOp,wigner1931gruppentheorie,wigner1993matrices} to a Wigner-Eckart theorem for $G$-steerable kernels, which expresses the kernels in terms of harmonic basis functions, Clebsch-Gordan coefficients and endomorphisms of the representations (generalized reduced matrix elements).
We refer to~\cite{lang2020WignerEckart} for a more detailed overview on prior and related work on steerable kernels.

\paragraph{\textit{GM}-coordinate independent convolutions:}

Given a $G$-steerable template kernel $K\in\KG$, the $\GM$-convolution~$K\star$ with this kernel is defined as the kernel field transform with the corresponding $\GM$-convolutional kernel field, satisfying $\Kp^X = K / \sqrt{|\eta^X|}$ for any point $p\in M$ and any gauge $\psi_p^X$.
By inserting the $\GM$-convolutional kernel field into Eq.~\eqref{eq:kft_coord_expression}, i.e. the kernel field transform, the coordinate expression of the $\GM$-convolution boils down to
\begin{align}\label{eq:gauge_conv_coord_expression}
    \fout^A(p)
    \ =\ 
    \big[ K \star \fin \big]^A(p)
    \ :=&\,\ 
        \int_{\R^d} \!
        K(\mathscr{v}) \,
        \big[\mkern-2mu \Expspfin\big]^{\mkern-2mu A} \mkern-1mu(\mathscr{v})
        \; d\mathscr{v}
    \ \ =\,\ \mathfrak{I}_K \big( [\Expspfin]^A \big)
    \,.
\end{align}
It is thus simply given by matching the transporter pullback $[\Expspfin]^A$ of the feature field in an \emph{arbitrarily chosen gauge} $\psi_p^A$ with the \emph{gauge independent} convolution kernel~$K$.
$\GM$-coordinate independent convolutions are therefore easily implemented by
1) choosing arbitrary reference frames,
2) pulling (and transporting) the feature fields back to the tangent space coordinatizations and
3) contracting them there with a (trainable) $G$-steerable kernel.

\begin{figure}
    \centering
    \includegraphics[width=.65\columnwidth]{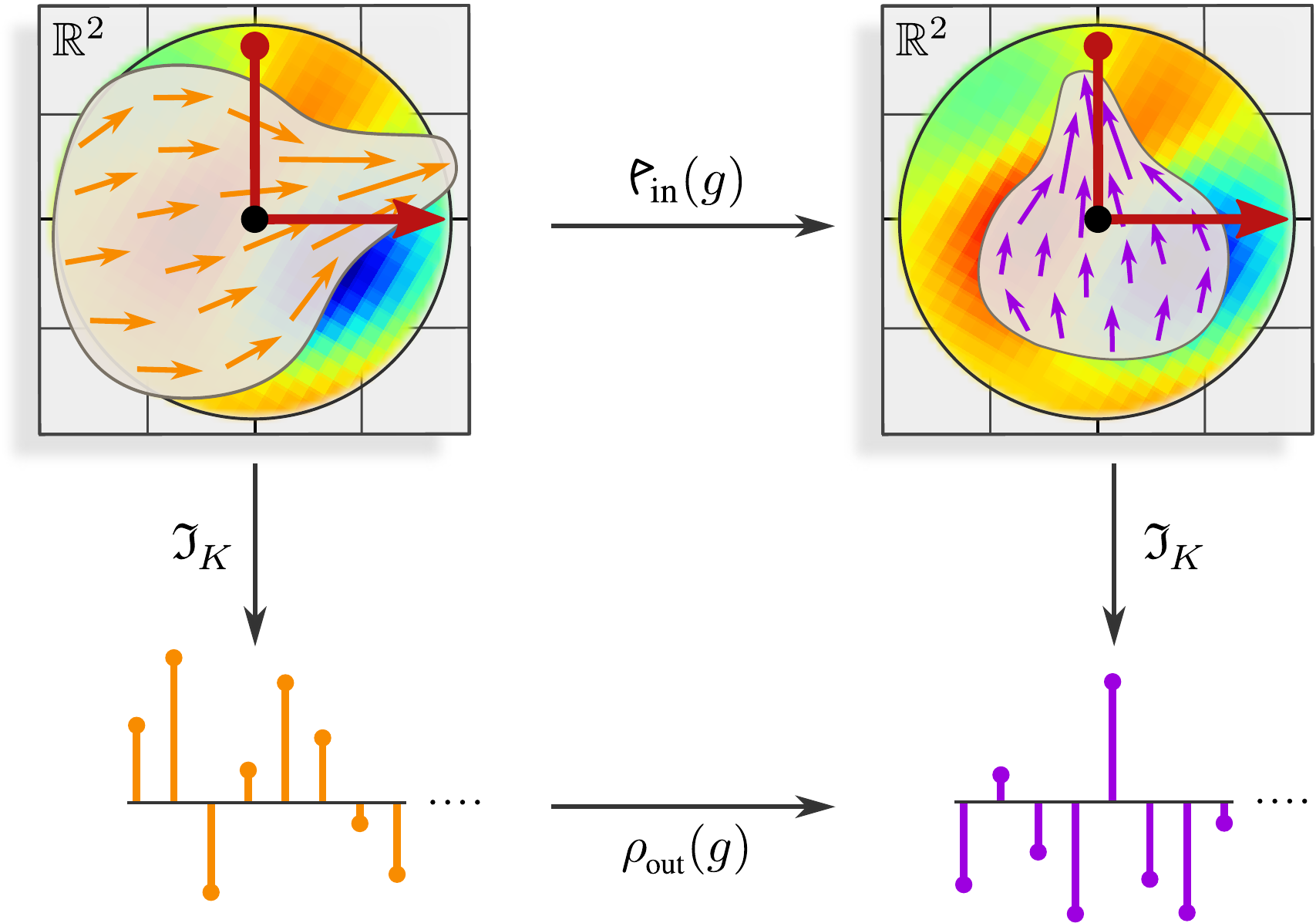}
    \vspace*{1ex}
    \caption{\small
        Local $G$-equivariance of the shared integral operator template $\mathfrak{I}_K$ underlying a $\GM$-convolution~$K\star$.
        An active $G$-transformation $\textphnc{b}_{\mkern-5mu\textup{in}}(g)$ of a local field representation on $\R^d$ moves feature vectors from $g^{-1}\mathscr{v}$ to~$\mathscr{v}$ and transforms them additionally via $\rhoin(g)$.
        While the former moves features spatially, the latter transforms their numerical coefficients (visualized as rotation and scaling of the individual (tangent) vectors in the figure).
        The application of $\mathfrak{I}_K$ to both inputs results in different output feature vectors.
        However, by the $G$-equivariance of $\mathfrak{I}_K$, the responses are guaranteed to be related by $\rhoout(g)$; see Eq.~\eqref{eq:active_local_gauge_trafo}.
        An active $G$-transformation of an input field therefore results in a corresponding active $G$-transformation of the output feature vector.
        Note that the $G$-equivariance of $\mathfrak{I}_K$ is a direct consequence of the $G$-steerability of $K$.
        \\\protect\rule{0ex}{0.5ex}
        }
    \label{fig:active_TpM_equivariance}
\end{figure}

$\GM$-convolutions exhibit multiple related symmetry properties:
\begin{itemize}[leftmargin=13em]
    \item[\it coordinate independence:]
        As specific instances of kernel field transforms, $\GM$-convolutions are (passively) coordinate independent, i.e. Eq.~\eqref{eq:kft_coordinate_independence} applies to them.
    \item[\it global isometry equivariance:]
        They are equivariant under the \emph{active}, \emph{global} action of $G$-structure preserving isometries in $\IsomGM$ on feature fields.
        Sections~\ref{sec:gauge_conv_isom_equiv} and specifically~\ref{sec:isometry_intro} discuss this property in detail.
    \item[\it local $G$-equivariance:]
        The integral operator template $\mathfrak{I}_K$ is by the $G$-steerability of~$K$ itself $G$-equivariant.
        Any $G$-transformation of a local feature field representation on $\R^d$ will therefore result in a corresponding transformation of the resulting feature vector; see Fig.~\ref{fig:active_TpM_equivariance}.
        Independent $G$-transformations of patterns that are centered at different points $p_i\in M$ will therefore lead to independent output feature transformations at these points (this holds \emph{only} at these points and requires compactly supported kernels whose entire \emph{field of view} transforms according to the $G$-transformation).
\end{itemize}

To make the last point precise, we define active $G$-transformations of local feature field representations in $C^\infty(\R^d,\R^c)$ as%
\footnote{
    ${\textphnc{b} = \Res_G^{\R^d\mkern-1mu\rtimes G}\Ind_G^{\R^d\mkern-1mu\rtimes G}\!\rho}$ is formally given by the induction of the $G$-representation $\rho$ to a ${(\R^d\mkern-4mu\rtimes\mkern-2muG)}$-representation ${\Ind_G^{\R^d\mkern-1mu\rtimes G}\!\rho}$, followed by a restriction back to $G$.
    An intuitive explanation of induced and restricted representations can be found in Appendix~B of~\cite{Weiler2019_E2CNN} while \cite{gallier2019harmonicRepr} treats the topic more formally.
}
\begin{align}\label{eq:local_field_G_trafo}
    \textphnc{b}_{\mkern-5mu\textup{in}}:\ 
    G \times C^\infty(\R^d,\R^c) \to C^\infty(\R^d,\R^c) \,, \quad
    F \,\mapsto\, \textphnc{b}_{\mkern-5mu\textup{in}}(g)F \,:=\, \rhoin(g) \circ F \circ g^{-1} \,,
\end{align}
where we assume $F$ to be of type $\rhoin$.
Intuitively, $\textphnc{b}_{\mkern-5mu\textup{in}}$ acts on fields $F$ by actively moving feature vectors $F(g^{-1}\mathscr{v}) \in \R^c$ from $g^{-1}\mathscr{v}$ to $\mathscr{v}$, thereby transforming them with $\rhoin(g)$
-- this is the ``usual'' definition of active transformations of feature fields~$F$ on Euclidean spaces~$\R^d$.
The claimed $G$-equivariance of $\mathfrak{I}_K$ is easily seen by applying it to a transformed input, followed by a substitution and making use of the $G$-steerability of~$K$:
\begin{alignat}{3}\label{eq:active_local_gauge_trafo}
\qquad
    \mathfrak{I}_K \big( \textphnc{b}_{\mkern-5mu\textup{in}}(g) F \big)
    \ =&\ \mathfrak{I}_K \big( \rhoin(g) \circ F \circ g^{-1} \big)
        \qquad\quad && \big( \text{\small def. of $\textphnc{b}_{\mkern-5mu\textup{in}}$, Eq.~\eqref{eq:local_field_G_trafo} } \big) \notag\\
    \ =&\ \int_{\R^d} K(\mathscr{v})\; \rhoin(g)\, F\big(g^{-1}\mathscr{v})\ d\mathscr{v}
        \qquad\quad && \big( \text{\small def. of $\mathfrak{I}_K$, Eq.~\eqref{eq:local_integral_operator_template} } \big) \notag\\
    \ =&\ \int_{\R^d} K(g\mkern1mu\tilde{\mathscr{v}})\; \rhoin(g)\, F\big(\tilde{\mathscr{v}})\; \detg \ d\tilde{\mathscr{v}}
        \qquad\quad && \big( \text{\small substitution of $\tilde{\mathscr{v}} = g^{-1} \mathscr{v}$ } \big) \notag\\
    \ =&\ \int_{\R^d} \rhoout(g)\, K(\tilde{\mathscr{v}})\; F\big(\tilde{\mathscr{v}})\ d\tilde{\mathscr{v}}
        \qquad\quad && \big( \text{\small $G$-steerability of $K$, Eq.~\eqref{eq:kernel_constraint} } \big) \notag\\
    \ =&\ \rhoout(g)\; \mathfrak{I}_K(F)
        \qquad\quad && \big( \text{\small def. of $\mathfrak{I}_K$, Eq.~\eqref{eq:local_integral_operator_template} } \big)
\end{alignat}
An active transformation of a local feature field representation $F$ on some tangent space coordinatization by~$\textphnc{b}_{\mkern-5mu\textup{in}}(g)$ is therefore guaranteed to lead to a transformation of the resulting output feature vector by~$\rhoout(g)$.
In other words, features which appear in different $G$-related geometric poses will evoke the same response up to a transformation via~$\rhoout$.
In terms of a commutative diagram, this is concisely summarized as:
\begin{equation}\label{cd:}
\begin{tikzcd}[column sep=56pt, row sep=34pt, font=\normalsize]
    C^\infty(\R^d,\R^{\cin})
        \arrow[d, "\mathfrak{I}_K\,"']
        \arrow[r, "\textphnc{b}_{\mkern-5mu\textup{in}}(g)"]
    &
    C^\infty(\R^d,\R^{\cin})
        \arrow[d, "\ \mathfrak{I}_K"]
    \\
    \R^{\cout}
        \arrow[r, "\rhoout(g)"']
    &
    \R^{\cout}
\end{tikzcd}
\end{equation}
Fig.~\ref{fig:active_TpM_equivariance} gives a visual interpretation of this equivariance property of $\mathfrak{I}_K$.

Note that the equivariance under local $G$-transformations in Eq.~\eqref{eq:active_local_gauge_trafo} requires the $G$-steerability constraint exactly as it is in Eq.~\eqref{eq:kernel_constraint}, that is, in particular, \emph{with} the determinant factor $\detg^{-1}$ which makes the kernel transform like a $-1$-density.
This factor is traced back to our definition of convolutional weight sharing in Eq.~\eqref{eq:weight_sharing_kernel} \emph{with} the normalization by the reference frame volumes $\sqrt{|\eta_p^X|}$.
The naive weight sharing mentioned in the beginning of this section would therefore not have lead to the desired transformation behavior.
In other words: both the naive and the normalized version of the kernel sharing are coordinate independent and behave therefore both consistently under passive gauge transformations -- in particular such which change the frame volume.
However, in the case of the naive kernel sharing, this is taken care of by the invariance of the Riemannian volume element $\sqrt{|\eta_p^A|}\ dv^A = \sqrt{|\eta_p^B|}\ dv^B$.
By \emph{canceling} this factor in the normalized weight sharing, the consistency of the transformation behavior is not guaranteed by the integration measure itself anymore -- which requires the $G$-steerable kernels themselves to explain volume changes via the determinant factor.
Only the latter generalizes to active transformations, where only the feature field is transformed, while the integration measure stays invariant.

As our definition of $\GM$-convolutions allows for arbitrary Riemannian manifolds, G-structures and field types, it is quite general and covers a wide range of related work.
We substantiate this claim in Part~\ref{part:literature_review}, where we explain many CNNs on Euclidean affine spaces $\Euc_d$, the sphere $S^2$ and general manifolds or meshes as specific instantiations of Eq.~\eqref{eq:gauge_conv_coord_expression}.
For an overview and a classification of these models, we refer to Table~\ref{tab:network_instantiations}.

%% file: chapters/43_isometry_equivariance.tex

\subsection{Isometry equivariance}
\label{sec:gauge_conv_isom_equiv}

\begin{SCfigure}
    \centering
    \includegraphics[width=.56\textwidth]{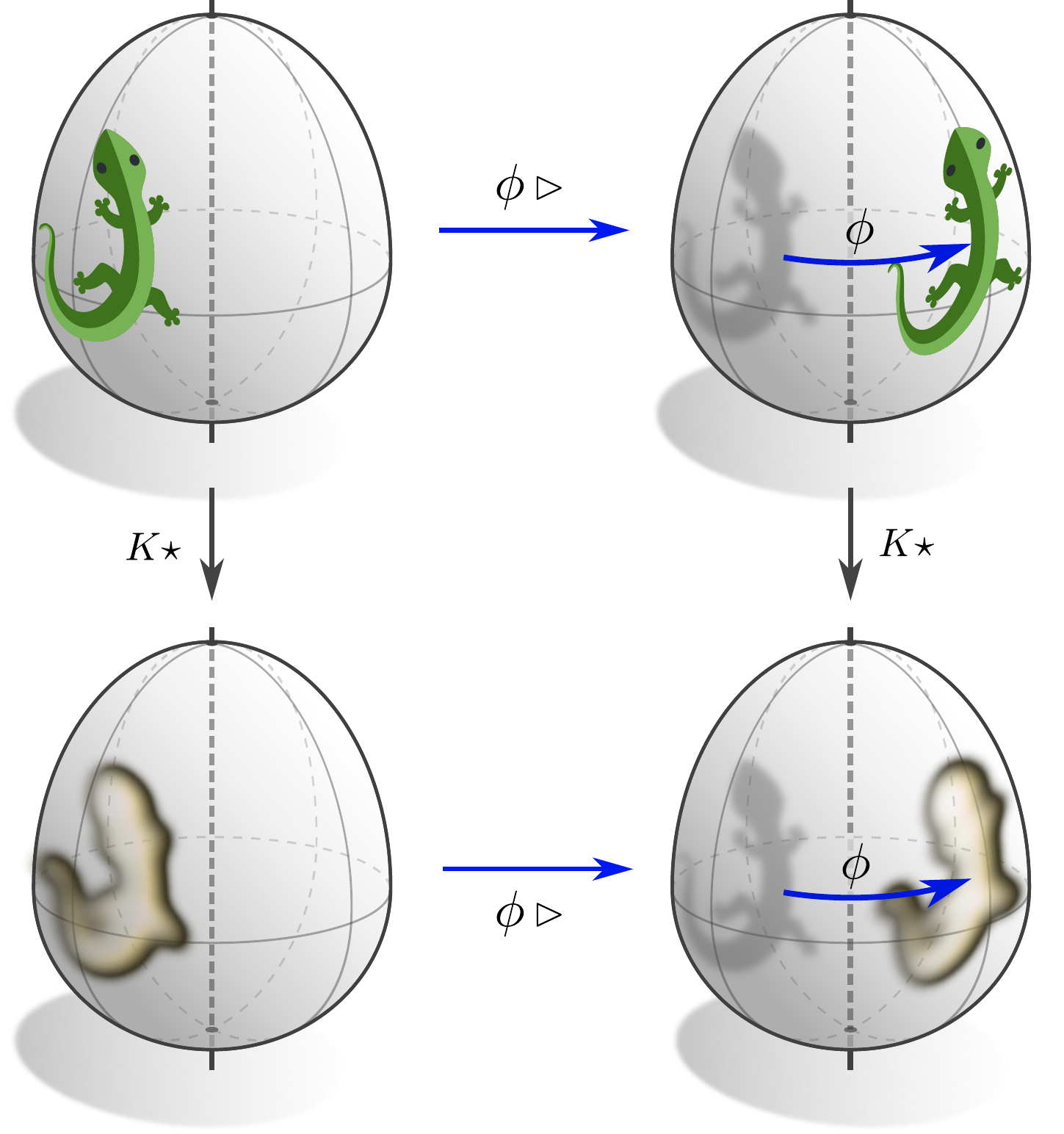}
    \captionsetup{width=.89\textwidth}
    \caption[]{\small
        A network layer is said to be equivariant under isometries when it commutes with their action on feature fields.
        $\GM$-convolutions are by design equivariant w.r.t. those isometries $\phi\in \IsomGM$ that are symmetries of the $G$-structure.
        In equations, the convolution $K\star$ is equivariant under the action of an isometry $\phi$ when it satisfies the relation
        ${K \star \big( \phi\rhd\! \fin \big) = \phi\rhd\! \big(K \star \fin \big)}$
        for any choice of input field~$\fin$.
        This relation is visualized by the commutative diagram in Eq.~\eqref{cd:isometry_equivariance_conv_mapsto}, a graphical interpretation of which is shown in this figure.
        \\[1ex]
        That $\GM$-convolutions are $\IsomGM$-equivariant relies on the facts that
        1) kernels are shared over the whole manifold,
        2) isometries preserve the transporter pullback of feature fields and
        3) that $\IsomGM$ induces $G$-valued gauge transformations, which are accounted for by the kernel's $G$-steerability.
        {\\
        \color{gray}
        \scriptsize
            (Lizards adapted under the Creative Commons Attribution 4.0 International
            \href{https://github.com/twitter/twemoji/blob/gh-pages/LICENSE-GRAPHICS}{\underline{license}}
            by courtesy of Twitter.)
        }
        \\[0ex]
        }
    \label{fig:lizard_conv_egg}
\end{SCfigure}

Given that a manifold exhibits symmetries it is usually desirable that neural networks respect these symmetries, i.e. are equivariant under their action on feature fields.
$\GM$-convolutions are by design guaranteed to be $\IsomGM$-equivariant, which means that they commute with the action of isometries in $\IsomGM$ on feature fields, as visualized in Fig.~\ref{fig:lizard_conv_egg}.%
\footnote{
    Recall that an action on $\GM$-coordinate independent feature fields can only be defined for the $G$-structure preserving isometries in $\IsomGM$.
    It is therefore not even possible to define a notion of isometry equivariance for isometries that are not symmetries the $G$-structure.
    Note that this is without loss of generality since one can always choose a structure group $G=\O{d}$, for which $\IsomGM=\IsomM$ coincides with the full isometry group.
}.
Expressed in equations, the $\GM$-convolution $K\star$ is equivariant when it satisfies the relation
\begin{align}\label{eq:isometry_equivariance_conv_def_33}
    K \star \big( \phi\rhd\! \fin \big) \ =\ \phi\rhd\! \big( K \star \fin \big)
    \qquad \forall\ \ \phi\in \IsomGM
\end{align}
for any possible input field $\fin$, that is, when the following diagram commutes:
\begin{equation}\label{cd:isometry_equivariance_conv_mapsto}
\begin{tikzcd}[column sep=60pt, row sep=35, font=\normalsize]
    \fin
        \arrow[r, "\phi\,\rhd", mapsto]
        \arrow[d, "K\star"', mapsto]
    &
    \phi\,\rhd \fin
        \arrow[d, "K\star", mapsto]
    \\
    \fout
        \arrow[r, "\phi\,\rhd"', mapsto]
    &
    \phi\,\rhd \fout
\end{tikzcd}
\end{equation}

As a first step towards proving the isometry equivariance of $\GM$-convolutions, recall that they are pointwise defined as the contraction of a kernel $K$ with the transporter pullback $[\Expspfin]^A$ of the input field~$\fin$.
Since isometries preserve the Riemannian geometry of~$M$ by definition, they preserve in particular the Riemannian exponential map and Levi-Civita transporters; see Section~\ref{sec:isom_expmap_transport} and Fig.~\ref{fig:isom_exp_transport}.%
\footnote{
    More generally, whenever an alternative $G$-compatible connection is chosen to transport feature vectors, we assume this connection to be invariant under the action of $\IsomGM$; see Section~\ref{sec:isom_expmap_transport}.
    This assumption is satisfied for all models that are covered in the literature review in Part~\ref{part:literature_review}.
}
This implies that the transporter pullback of the pushforward field $\phi\rhd\fin$ at $\phi(p)$ will only differ from the transporter pullback of the original field $\fin$ at $p$ by the isometry induced gauge transformation, that is,
\begin{align}\label{eq:transporter_pullback_pushforward_field}
    \big[\mkern-2mu \Expsphip (\phi\rhd\fin) \big]^A\ =\ 
    \rhoin\big( g_\phi^{A\widetilde{A}}(p)\big) \circ \big[\mkern-2mu \Expspfin\big]^{\widetilde{A}} \circ g_\phi^{A\widetilde{A}}(p)^{-1} \,;
\end{align}
cf. Eq.~\eqref{eq:trafo_law_transporter_pullback} and, for the coordinate free formulation, Theorem~\ref{thm:transporter_pullback_isometry_action}.

Given this identity, the isometry equivariance of $\GM$-convolutions
is proven by the following simple calculation, which crucially leverages the $G$-steerability of the template kernel~$K$ to explain away the isometry induced gauge action:
\begin{align}
    \big[ K \star (\phi \rhd \fin) \big]^A (\phi(p))
    \ \overset{(1)}{=}&\ \ 
        \int_{\R^d} K(\mathscr{v})\  \big[\mkern-2mu \Expsphip (\phi\rhd\fin) \big]^A (\mathscr{v})\ d\mathscr{v} \notag \\
    \ \overset{(2)}{=}&\ \ 
        \int_{\R^d} K(\mathscr{v})\ 
        \Big[ \rhoin\pig( g_\phi^{A\widetilde{A}}(p)\pig) \big[\mkern-2mu \Expspfin\big]^{\widetilde{A}} \pig(g_\phi^{A\widetilde{A}}(p)^{-1} \mathscr{v} \pig) \Big]\ d\mathscr{v} \notag \\
    \ \overset{(3)}{=}&\ \ 
        \int_{\R^d} \Big[ K\pig( g_\phi^{A\widetilde{A}}(p)\, \tilde{\mathscr{v}}\pig)\ \rhoin\pig( g_\phi^{A\widetilde{A}}(p)\pig) \Big]\ 
        \big[\mkern-2mu \Expspfin\big]^{\widetilde{A}} (\tilde{\mathscr{v}})\ 
        \pig|\mkern-3mu\det g_\phi^{A\widetilde{A}}(p)\pig|\; d\tilde{\mathscr{v}} \notag \\
    \ \overset{(4)}{=}&\ \ 
        \int_{\R^d} \Big[\rhoout\pig( g_\phi^{A\widetilde{A}}(p)\pig)  K\big(\tilde{\mathscr{v}}\big) \Big]\ 
        \big[\mkern-2mu \Expspfin\big]^{\widetilde{A}} (\tilde{\mathscr{v}})\ d\tilde{\mathscr{v}} \notag \\
    \ \overset{(5)}{=}&\ \ 
        \rhoout\big( g_\phi^{A\widetilde{A}}(p) \big) \cdot \fout^{\widetilde{A}}(p) \notag \\
    \ \overset{(6)}{=}&\ \ 
        \big[ \phi \rhd \fout \big]^A (\phi(p)) \notag \\
    \ \overset{(7)}{=}&\ \ 
        \big[ \phi \rhd \big(K \star \fin\big) \big]^A (\phi(p))
\end{align}
The first step follows hereby from the definition of $\GM$-convolutions in Eq.~\eqref{eq:gauge_conv_coord_expression} while the second step inserted the induced gauge transformation according to Eq.~\eqref{eq:transporter_pullback_pushforward_field}.
A substitution from $\mathscr{v}$ to $\tilde{\mathscr{v}} = g_\phi^{A\widetilde{A}}(p)^{-1} \mkern1mu \mathscr{v}$ justifies step three.
In the fourth step the $G$-steerability of the template kernel, i.e. Eq.~\eqref{eq:kernel_constraint}, is applied (recall Eq.~\eqref{eq:IsomGM_coord_in_G_24}, which states that the $\IsomGM$-induced gauge transformations are $G$-valued).
What follows is that the resulting output feature vector is transformed by the induced gauge transformation.
After identifying this as the coordinate expression of the pushforward of the output field in Eq.~\eqref{eq:feature_field_trafo_in_coords}, the statement follows.
As all steps are valid for arbitrary isometries in $\IsomGM$, we see that \emph{$\GM$-convolutions are automatically equivariant w.r.t. any $G$-structure preserving isometry}.
They are \emph{not} necessarily equivariant w.r.t. general isometries in $\IsomM$, which might disrespect the $G$-structure, however, full isometry equivariance is guaranteed for orthonormal structure groups $G=\O{d}$ (or supergroups of~it).

\paragraph{Invariant kernel fields:}
A more in depth analysis of the isometry equivariance of general kernel field transforms can be found in Sections~\ref{sec:isometry_equivariance} and~\ref{sec:quotient_kernel_fields}.
The central result of this investigation is Theorem~\ref{thm:isometry_equivariant_kernel_field_trafos}, which states that \emph{the isometry equivariance of a kernel field transform implies the isometry invariance of its kernel field} and vice versa.
Fig~\ref{fig:isom_invariant_kernel_field_multiple_orbits} visualizes such an invariant kernel field, which is required to share weights over the orbits of the isometry action.
The required invariance of the kernel field is intuitively plausible since isometry equivariance certainly requires the inference of the network to be the constant on each orbit.
This abstract results implies the isometry equivariance of $\GM$-convolutions by observing that $\GM$-convolutional kernel fields -- which are determined by a single, shared template kernel -- are invariant under isometries in $\IsomGM$; see Theorem~\ref{thm:isom_equiv_GM_conv} and Fig.~\ref{fig:intro_invariant_kernel_fields_plane}.
The template kernel's $G$-steerability accounts thereby for the invariance of kernels under the action of stabilizer subgroups of the isometry group.

\paragraph{Homogeneous spaces:}
While the demand for isometry equivariance requires kernels to be shared over orbits of the isometry group, it does in general not require convolutional weight sharing over the whole manifold.
An important exception is the case of manifolds that are \emph{homogeneous spaces} of their isometry group, like for instance $\R^d$ or the sphere $S^2$.
By definition, the isometry action is on such spaces transitive, that is, there exists only one single orbit.
Consequently, there will only be one independent kernel, which is via the action of the isometry group being shared over the whole space.
Theorem~\ref{thm:GM_conv_homogeneous_equivalence} in Section~\ref{sec:quotient_kernel_fields} proves that \emph{isometry equivariant kernel field transforms on homogeneous spaces are necessarily coordinate independent convolutions}.
This observation establishes a formal link between our theory and prior work on convolutional networks on homogeneous spaces
by~\citet{Kondor2018-GENERAL}, \citet{Cohen2019-generaltheory} and~\citet{bekkers2020bspline},
who are defining convolutions via their equivariance w.r.t. global symmetries of the underlying space.

\paragraph{Diffeomorphism equivariance:}
The reader might wonder whether it is possible to make our coordinate independent CNNs fully diffeomorphism equivariant.
As one can easily see, the pointwise operations from Section~\ref{sec:pointwise_operations}, i.e. \onexones, biases and nonlinearities, are already diffeomorphism equivariant.
Specifically, let
\begin{align}\label{eq:DiffGM_def_part1}
    \DiffGM\ :=\ \pig\{ \phi \in \Diff(M) \ \pig|\ 
    \big[\phi_*(e_i)\big]_{i=1}^d \in \GM \quad \forall\ [e_i]_{i=1}^d \in \GM \pig\} \,\ \leq\ \Diff(M)
\end{align}
be the subgroup of $G$-structure preserving diffeomorphisms, i.e. the analog to Eq.~\eqref{eq:IsomGM_def_24} without the requirement on~$\phi$ to be an isometry.
Similarly to Eq.~\eqref{eq:IsomGM_coord_in_G_24} and Theorem~\ref{thm:isom_GM_in_coords}, the coordinate expressions (induced gauge transformations) of $G$-structure preserving diffeomorphisms are guaranteed to take values in $G$, that is,
\begin{align}
    \phi \in \DiffGM \quad \Longleftrightarrow \quad g_\phi^{A\widetilde{A}}(p) \in G\ \ \ \forall\ p \mkern-1mu\in\mkern-2mu M \,.
\end{align}
The $G$-equivariance of the shared pointwise template functions will guarantee that they commute with these $\DiffGM$-induced gauge transformations -- and therefore with the active diffeomorphism action itself.

$\GM$-convolutions with spatially extended kernels, on the other hand, are in general \emph{not} equivariant w.r.t. diffeomorphisms.
The reason for this is that the transporter pullback $\Expspf$ relies on exponential maps, which are inherently Riemannian constructions that do not commute with diffeomorphisms.
However, as the kernels are $G$-steerable, $\DiffGM$-equivariance should nonetheless hold in the limit of the kernel support going to zero.
Given that convolution kernels are in typical deep learning applications quite small, diffeomorphism equivariance should in practice hold approximately.

\paragraph{Affine equivariance:}
Euclidean spaces constitute a special case since they allow for $\GM$-convolutions that are equivariant under the action of \emph{affine groups}~$\Aff(G)$.
That this is the case relies on the fact that the exponential map commutes on Euclidean spaces not only with the action of isometries but more generally with affine transformations.
The affine group equivariance of Euclidean $\GM$-convolutions is proven in Section~\ref{sec:euclidean_affine_equiv}.

%% file: chapters/50_mobius_conv.tex

\section{Toy model: reflection equivariant M{\"o}bius convolutions}
\label{sec:mobius_conv}

To make the theoretical considerations in the previous sections more tangible, we turn now to an example application.
While not being of immediate practical importance, $\GM$-convolutions on the M\"obius strip are a suitable toy model since its geometry and the involved representation theory are particularly simple.
Due to its non-orientability, reference frames can only be (smoothly) preferred up to reflections.
As expected, coordinate independent CNNs, applying reflection equivariant template functions, outperform a naive, coordinate dependent implementation.
They are furthermore shown to be equivariant under the action of the M\"obius strip's isometry group.

\etocsettocdepth{3}
\etocsettocstyle{}{} 
\localtableofcontents

The following Section~\ref{sec:mobius_geometry} discusses the geometry of the flat M\"obius strip.
Due to its twist, its structure group can not be reduced further than to the reflection group~$G=\Flip$, such that one needs to consider a $\Flip$-atlas of gauges as visualized in Fig.~\ref{fig:mobius_conv_gauges}.
The isometry group is given by rotations along the strip and induces $\Flip$-valued gauge transformations.
$\RM$-coordinate independent feature fields, some of which are introduced in Section~\ref{sec:mobius_representations}, necessarily have to transform according to some representation of the reflection group.
Section~\ref{sec:mobius_cnn_ops_analytical} discusses orientation independent convolutional network operations.
This clarifies in particular the concept of $G$-steerable kernels but also covers reflection equivariant biases and nonlinearities.
A numerical implementation of the proposed model family is discussed and evaluated in Section~\ref{sec:mobius_experiment_main}.
The code is publicly available at \url{https://github.com/mauriceweiler/MobiusCNNs}.

\subsection{Geometry of the M\"obius strip}
\label{sec:mobius_geometry}

The manifold $M$ under consideration is the flat M\"obius strip with boundary as shown in Fig.~\ref{fig:weight_sharing_ambiguity} (right).
It~can be thought of as being constructed by taking a rectangular subset $[0,X] \times [0,Y]$ of $\R^2$ and gluing two opposing ends together in a twisted way.
Such defined, the M\"obius strip inherits the canonical metric of $\R^2$, which endows it with a Riemannian structure.
The metric specifies in particular a Levi-Civita connection and therefore exponential maps and parallel transporters, which are further discussed below.

\begin{figure}
    \centering
    \includegraphics[width=\columnwidth]{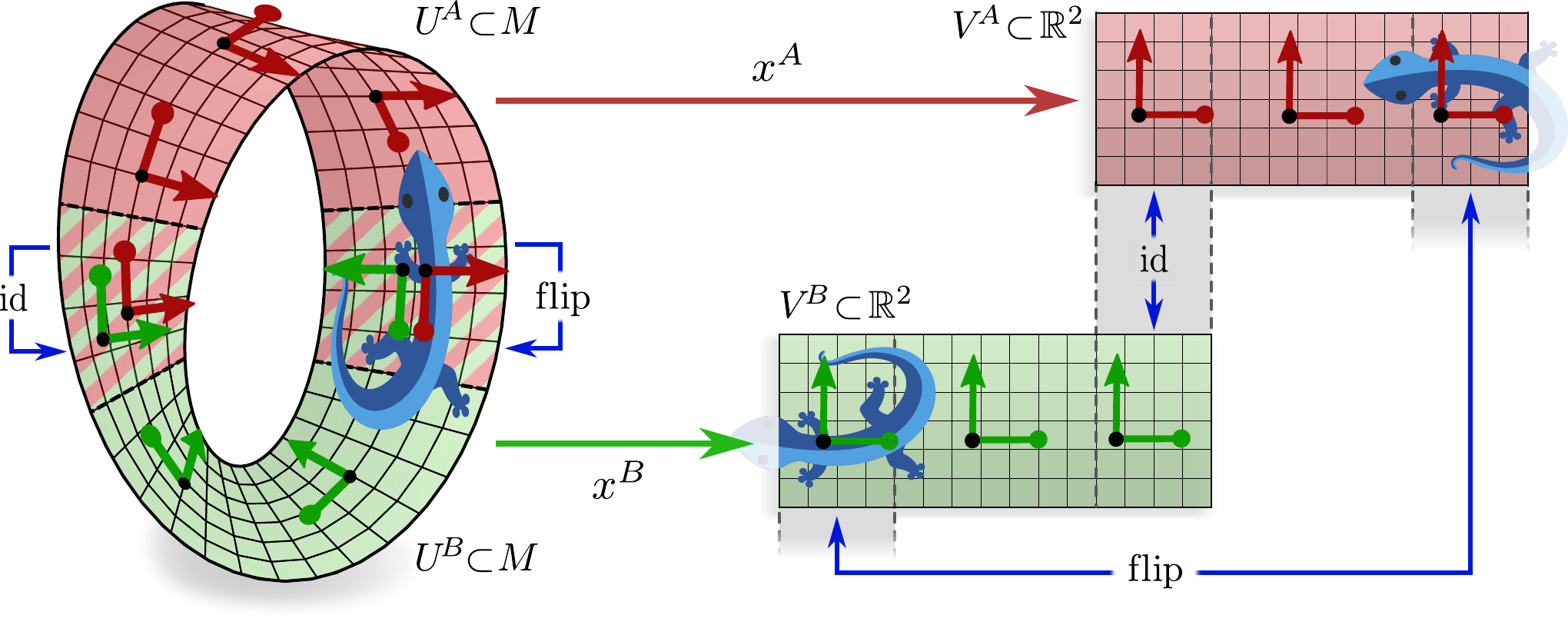}
    \vspace*{.5ex}
    \caption{\small
        The flat geometry of the M\"obius strip allows for local subsets which can be isometrically identified with corresponding subsets of~$\R^2$.
        We fix an isometric atlas, consisting of two charts $x^A$ and $x^B$ on $U^A$ (red) and $U^B$ (green), which cover the whole strip.
        Gauges $\psi_p^X = \hat{d}x_p^X: \TpM \to \R^d$ for $p\in U^A$ are induced as chart differentials.
        Due to the twist of the M\"obius strip, the transition functions $g_p^{BA}$ will at one of the overlapping regions be trivial, while the other region will necessarily transition between gauges via flips~$s$.
        The chosen atlas of charts therefore induces an $\Flip$-atlas of gauges and implies a corresponding $\Flip$-structure $\RM$, consisting of two reflected frames at each point of~$M$.
        Each of the charts $x^X$ induces a smooth local frame field, given by the coordinate bases
        $\Big[\frac{\partial}{\partial x_i^X} \mkern-1mu\big|_p \Big] \raisebox{-2pt}{$\rule{0pt}{11pt}_{i=1}^d$}$.
        The flip in the transition functions at one overlap shows in a reflection of frames.
        {
        \\ \color{gray} \scriptsize
            (Lizards adapted under the Creative Commons Attribution 4.0 International
            \href{https://github.com/twitter/twemoji/blob/gh-pages/LICENSE-GRAPHICS}{\underline{license}}
            by courtesy of Twitter.)
        }
    }
    \label{fig:mobius_conv_gauges}
\end{figure}

A first question to answer when constructing a coordinate independent CNN is to which extent the choice of reference frames is ambiguous.
Given the Riemannian metric on the strip, we can restrict our attention to orthonormal frames.
One can furthermore single out one of the two directions \emph{along} the strip to (smoothly) disambiguate the rotation of the reference frames by aligning their first axes with this direction.
This leaves us with an ambiguity of frame handedness, with the two orientations corresponding to the two possible directions of the second frame axis perpendicular to the strip.
Being a non-orientable manifold, the M\"obius strip does not admit a globally smooth (or even continuous) choice of frame orientations.
To get an intuition about this statement, consider the attempt of constructing a smooth frame field by picking an arbitrary frame at a random position and to smoothly extend this choice over the whole strip.
After one revolution around the strip the constructed frames will unavoidably be reflected w.r.t. the initial frames, and therefore contradict the desired smoothness.
It is thus topologically \emph{impossible} to define an $\{e\}$-structure, i.e. a globally smooth field of frames, on the M\"obius strip.
We are thus left with an irreducible structure group
\begin{align}
    G \,=\, \Flip \,\cong\, \Z/2\Z \,,
\end{align}
which models the reflection of frames.
The reflection group contains only two elements, the identity $e$ and the reflection (Spiegelung) $s$, which are composed according to the following simple multiplication table:
\begin{align}\label{eq:reflection_multiplication_table}
\begin{tabular}{c|c@{\hspace{8pt}}c}
        & $e$ & $s$ \\ \hline
    $e$ & $e$ & $s$ \\
    $s$ & $s$ & $e$
\end{tabular}
\end{align}
The only nontrivial statement in this table is that two reflections annihilate, that is, $s^2=e$, or, equivalently, $s^{-1}=s$.
Given the irreducibility of the structure group $\Flip$, we will in the following need to consider the corresponding $\Flip$-structure $\RM$ which consists of two frames of opposing handedness at each point on the M\"obius strip.

To encode smooth $\RM$-coordinate independent feature fields on $M$, one needs to specify an $\Flip$-atlas, consisting of $\Flip$-related gauges that cover the whole strip.
We choose to do this by fixing an atlas of \emph{charts}
\begin{align}
    x^X: U^X \to V^X \subset \R^2
\end{align}
which cover the strip, and subsequently induce the gauges from it.
Fig.~\ref{fig:mobius_conv_gauges} visualizes such an atlas, consisting of two charts $x^A$ and $x^B$ on $U^A$ (red) and $U^B$ (green) which map two overlapping halves of the strip isometrically to corresponding rectangular regions of $\R^2$.
As described in Appendix~\ref{apx:chart_induced_bases_main}, the charts induce gauges, which are given by the chart differentials, that is,
\begin{align}
    \psi_p^X \,:=\, \hat{d}x^X_p :\ \TpM \to \R^2 \ \ \ \textup{for any}\,\ p\in U^X\,\ \text{and}\,\ X=A,B \,.
\end{align}
The transition functions coincide then with the Jacobians
$g^{BA} = \frac{\partial x^B}{\partial x^A}$.
Due to the twist, the transition maps are at one of the two overlapping regions all trivial, that is, $g_p^{BA} = e$, and on the other end necessarily reflected, i.e. $g_p^{BA} = s$.
The induced atlas of gauges is therefore indeed identified as an $\Flip$-atlas.
Being derived from coordinate charts, the smooth local frame fields corresponding to the gauges are just the usual coordinate bases, that is, the frames $\big[e_i^X \big]_{i=1}^d$ at $p\in U^X$ are given by
$\pig[\frac{\partial}{\partial x_i^X} \mkern-1mu\big|_p \pig] \raisebox{-2pt}{$\rule{0pt}{11pt}_{i=1}^d$}$.
Since the charts are isometric, the induced frame field is automatically orthonormal.
However, the two rectangular regions $V^A$ and $V^B$ in $\R^2$ must not be rotated relative to each other in order to induce an $\Flip$-atlas and a corresponding $\Flip$-structure~$\RM$.

We need to emphasize that the approach of inducing gauges via coordinate charts is not strictly necessary
-- it is just a convenient option since the \emph{flat} M\"obius strip is locally identified with regions of $\R^2$ in an \emph{isometric} way.
This will later allow us to transfer regular sampling grids from $\R^2$, like for instance the pixel grid $\Z^2$, to regular sampling grids on the strip.
As this is not possible for manifolds that are not locally flat, for instance meshes in computer graphics, most implementations on general manifolds (or meshes) assign coordinates immediately to the tangent spaces; see Section~\ref{sec:instantiations_mesh}.

The canonical Levi-Civita connection on the M\"obius strip defines a notion of parallel transport of tangent vectors.
Since the strip is locally isometric to the plane $\R^2$, this transport can on local patches be understood as flattening these patches out into a plane and moving the vectors as usual on~$\R^2$.
If no single patch can cover a path~$\gamma$, there will be an open covering such that the full transport is explained by a sequence of transporters over the local patches.
It is easy to see that the transport will relative to frames of the chosen $\Flip$-structure take values $g_\gamma^{A\widetilde{A}}$ in the reflection group $\Flip$.
This means that the Levi-Civita connection is $\Flip$-compatible with~$\RM$.
It does therefore imply well defined transporters $\rho\big( g_\gamma^{A\widetilde{A}} \big)$ of $\Flip$-associated feature vectors.

\begin{figure}
    \centering
    \includegraphics[width=\columnwidth]{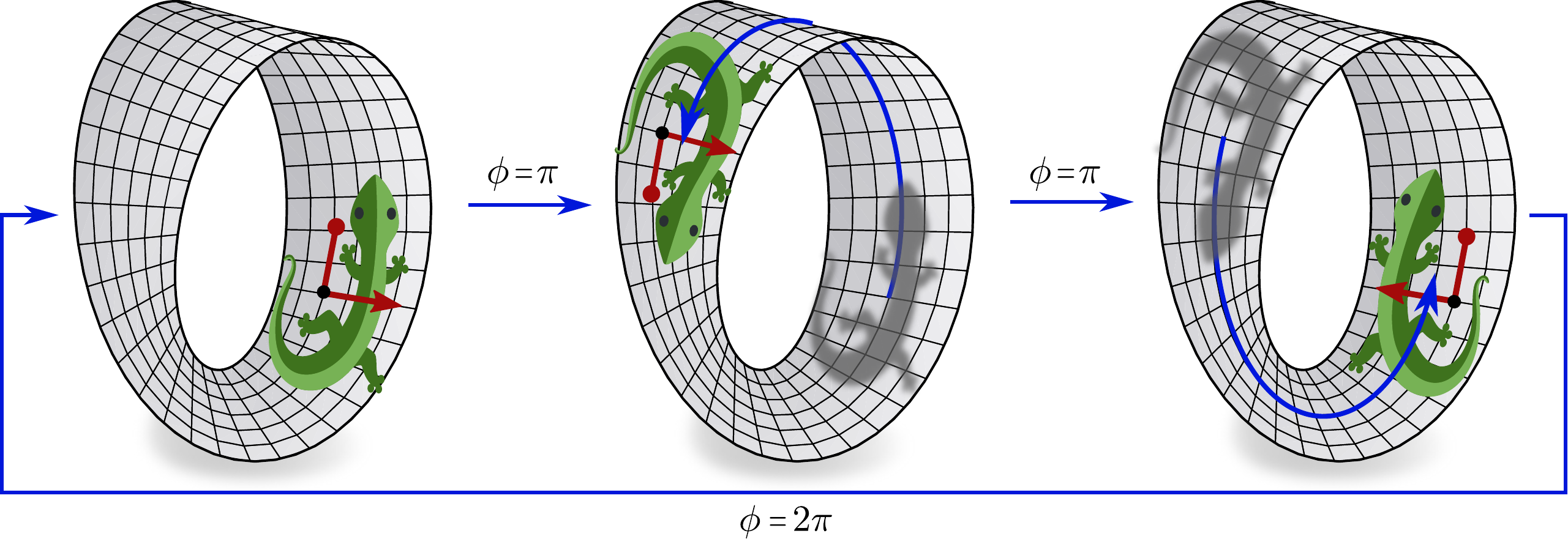}
    \caption{\small
        Visualization of the group of $\Flip$-structure preserving isometries $\IsomRM$ of the M\"obius strip, which is isomorphic to~$\SO2$.
        It consists of all rotations along the strip.
        Due to the twist, a rotation by $2\pi$, i.e. once around the strip, does not yet map it back to itself but results in a reflection.
        After a second revolution, that is, a total rotation of $4\pi$, the strip is mapped back to itself.
        Induced gauge transformations take values in $\Flip$.
        {
        \\ \color{gray} \scriptsize
            (Lizards adapted under the Creative Commons Attribution 4.0 International
            \href{https://github.com/twitter/twemoji/blob/gh-pages/LICENSE-GRAPHICS}{\underline{license}}
            by courtesy of Twitter.)
        }
    }
    \label{fig:mobius_conv_isometries}
\end{figure}

The group $\IsomRM$ of isometries that preserve the $\Flip$-structure contains all rotations which shift the strip along itself.
Note that a rotation once around the strip, which we denote by an angle of $2\pi$, does \emph{not} correspond to the identity but rather maps the strip in a reflected way on itself.
Only a rotation by $4\pi$, i.e. two full revolutions, map the strip back to itself.%
\footnote{
    The M\"obius strip is therefore seen to have the cylinder as double cover.
}
The action of the isometry group on the manifold and on reference frames is visualized in Fig.~\ref{fig:mobius_conv_isometries}.
Relative to coordinates, the isometry action will induce $\Flip$-valued gauge transformations.

\subsection{Orientation independent feature fields}
\label{sec:mobius_representations}

The principle of covariance requires the feature fields on the M\"obius strip to be $\RM$-coordinate independent, that is, they need to be equivalently expressible relative to frames of either handedness.
They are therefore characterized by a choice of group representation $\rho: \Flip \to \GL{c}$ of the reflection group, which specifies the transformation of numerical feature vectors when switching between the two orientations.
We will in the following discuss a few possible choices of such field types.
The reader might want to check that the proposed representations are indeed group homomorphisms, satisfying $\rho(gh) = \rho(g)\rho(h)\,\ \forall\, g,h\in \Flip$, as demanded in Section~\ref{sec:feature_fields} and footnote~\ref{footnote:repr_group_homomorphism}.

The most basic example, which exists for any structure group, is the \emph{trivial representation}
\begin{align}
    \rhotriv: \Flip \to \GL{1}\ , \qquad 
    \begin{aligned}
        e &\mapsto \big[\mkern2mu 1 \mkern2mu\big] \\[2pt]
        s &\mapsto \big[\mkern2mu 1 \mkern2mu\big] \\
    \end{aligned}
    \quad ,
\end{align}
which assigns the ${1\!\times\!1}$ identity matrix to both group elements.
It models scalar fields $f_{\textup{triv}}$, which consist of one-dimensional feature vectors whose coordinatizations $f^A_{\textup{triv}}(p) \in \R^1$ stay \emph{invariant under frame reflections}.
A second one-dimensional representation is the \emph{sign-flip representation}
\begin{align}
    \rhosign: \Flip \to \GL{1}\ , \qquad 
    \begin{aligned}
        e &\mapsto \big[\mkern2mu 1 \mkern2mu\big] \\[2pt]
        s &\mapsto \big[\! \shortminus\!1 \big]
    \end{aligned}
    \quad .
\end{align}
It assigns the negative ${1\!\times\!1}$ identity matrix to reflections, and therefore describes pseudoscalar fields, i.e. one-dimensional feature fields $f_{\textup{sign}}$, whose numerical coefficients $f^A_{\textup{sign}}(p) \in \R^1$ \emph{change their sign under reflections}, \mbox{that is,} ${\rhosign(s)\cdot f^A_{\textup{sign}}(p) = -f^A_{\textup{sign}}(p)}$.
Since the trivial representation and the sign-flip representation are one-dimensional, they are both irreducible representations (irreps) of the reflection group.
In fact, they are the only two irreps of the reflection group.%
\footnote{
    The reflection group is isomorphic to the cyclic group $\Z/2\Z$ of order two.
    It is well known that the irreps of cyclic groups of order $N$ correspond to the $N$-th roots of unity, which are for $N=2$ just $+1$ (trivial) and $-1$ (sign-flip).
}

Since $\Flip$ is a finite group, it has a finite-dimensional (two-dimensional) \emph{regular representation}
\begin{align}
    \rhoreg: \Flip \to \GL{2}\ , \qquad 
    \begin{aligned}
        e &\mapsto
            \begin{bmatrix} \hspace{1.5pt}
                1 &\mkern-4mu 0 \hspace*{1.5pt} \\ \hspace{1.5pt} 0 &\mkern-4mu 1 \hspace*{1.5pt}
            \end{bmatrix} \\[2pt]
        s &\mapsto 
            \begin{bmatrix} \hspace{1.5pt}
                0 &\mkern-4mu 1 \hspace*{1.5pt} \\ \hspace{1.5pt} 1 &\mkern-4mu 0 \hspace*{1.5pt}
            \end{bmatrix}
    \end{aligned}
    \quad ,
\end{align}
which represents the group elements by permutation matrices.
By definition, the regular representation models the permutation of the group elements in $\Flip$ when acting on themselves.
Compare this to the columns of the multiplication table in Eq.~\eqref{eq:reflection_multiplication_table}:
the middle column can be thought of as originating from the action of $\rhoreg(e)$ on the leftmost column, while the swapped group elements in the right column correspond to the permutation described by the action of $\rhoreg(s)$ on the left column.
The regular feature fields $f_{\textup{reg}}$ of $\Flip$ are numerically represented by two-dimensional feature vectors $f^A_{\textup{reg}}(p) \in \R^2$ whose two \emph{channels are swapped under reflections}, that is,
$
    \rhoreg(s) \mkern-1mu\cdot\mkern-2mu f^A_{\textup{reg}}(p)
    \,=\,
    \begin{bmatrix} \hspace{1.5pt} 0 &\mkern-4mu 1 \hspace*{1.5pt} \\ \hspace{1.5pt} 1 &\mkern-4mu 0 \hspace*{1.5pt} \end{bmatrix}
    \mkern-4mu \cdot\mkern-4mu 
    \begin{bmatrix} f^A_{\textup{reg},1} \\ f^A_{\textup{reg},2} \end{bmatrix} \!(p)
    \,=\,
    \begin{bmatrix} f^A_{\textup{reg},2} \\ f^A_{\textup{reg},1} \end{bmatrix} \!(p)
$.

The regular representation is reducible, that is, it contains two proper invariant subspaces, which correspond in this case to the trivial and the sign-flip representation.
It can therefore equivalently be thought of as being constructed of the direct sum $\rhotriv \oplus \rhosign$ of those two irreps and a change of basis $Q$\,:
\begin{align}\label{eq:rho_reg_decomposition}
    \rhoreg(g)\ =\ Q\, \big( \rhotriv \!\oplus\mkern-1mu \rhosign \big)\mkern-2mu(g)\; Q^\top
    \quad\ \textup{where } \quad
    Q = \frac{1}{\sqrt{2}}
    \begin{bmatrix} \hspace{1.5pt}
        1 &\mkern-7mu \shortminus1 \hspace*{1.5pt} \\ \hspace{1.5pt} 1 &\mkern-7mu \phantom{\shortminus}1 \hspace*{1.5pt}
    \end{bmatrix}
\end{align}
The validity of this statement is easily asserted by inserting the r.h.s. for both group elements:
\begin{align}
    Q\, \big( \rhotriv \!\oplus\mkern-1mu \rhosign \big)\mkern-2mu(e)\; Q^\top
    \ =\ \frac{1}{2}
    \begin{bmatrix} \hspace{1.5pt}
        1 &\mkern-7mu \shortminus1 \hspace*{1.5pt} \\ \hspace{1.5pt} 1 &\mkern-7mu \phantom{\shortminus}1 \hspace*{1.5pt}
    \end{bmatrix} \mkern-6mu \cdot\mkern-6mu
    \begin{bmatrix} \hspace{1.5pt}
        1 &\mkern-7mu 0 \hspace*{1.5pt} \\ \hspace{1.5pt} 0 &\mkern-7mu 1 \hspace*{1.5pt}
    \end{bmatrix} \mkern-6mu \cdot\mkern-6mu
    \begin{bmatrix} \hspace{1.5pt}
        \phantom{\shortminus}1 &\mkern-7mu 1 \hspace*{1.5pt} \\ \hspace{1.5pt} \shortminus1 &\mkern-7mu 1 \hspace*{1.5pt}
    \end{bmatrix}
    \ =\ 
    \begin{bmatrix} \hspace{1.5pt}
        1 &\mkern-7mu 0 \hspace*{1.5pt} \\ \hspace{1.5pt} 0 &\mkern-7mu 1 \hspace*{1.5pt}
    \end{bmatrix}
    \ =\ \rhoreg(e)
\end{align}
\begin{align}
    Q\, \big( \rhotriv \!\oplus\mkern-1mu \rhosign \big)\mkern-2mu(s)\; Q^\top
    \,=\, \frac{1}{2}
    \begin{bmatrix} \hspace{1.5pt}
        1 &\mkern-7mu \shortminus1 \hspace*{1.5pt} \\ \hspace{1.5pt} 1 &\mkern-7mu \phantom{\shortminus}1 \hspace*{1.5pt}
    \end{bmatrix} \mkern-6mu \cdot\mkern-6mu
    \begin{bmatrix} \hspace{1.5pt}
        1 &\mkern-7mu \phantom{\shortminus}0 \hspace*{1.5pt} \\ \hspace{1.5pt} 0 &\mkern-7mu \shortminus1 \hspace*{1.5pt}
    \end{bmatrix} \mkern-6mu \cdot\mkern-6mu
    \begin{bmatrix} \hspace{1.5pt}
        \phantom{\shortminus}1 &\mkern-7mu 1 \hspace*{1.5pt} \\ \hspace{1.5pt} \shortminus1 &\mkern-7mu 1 \hspace*{1.5pt}
    \end{bmatrix}
    \,=\, \frac{1}{2}
    \begin{bmatrix} \hspace{1.5pt}
        1 &\mkern-7mu \phantom{\shortminus}1 \hspace*{1.5pt} \\ \hspace{1.5pt} 1 &\mkern-7mu \shortminus1 \hspace*{1.5pt}
    \end{bmatrix} \mkern-6mu \cdot\mkern-6mu
    \begin{bmatrix} \hspace{1.5pt}
        \phantom{\shortminus}1 &\mkern-7mu 1 \hspace*{1.5pt} \\ \hspace{1.5pt} \shortminus1 &\mkern-7mu 1 \hspace*{1.5pt}
    \end{bmatrix}
    \,=\, 
    \begin{bmatrix} \hspace{1.5pt}
        0 &\mkern-7mu 1 \hspace*{1.5pt} \\ \hspace{1.5pt} 1 &\mkern-7mu 0 \hspace*{1.5pt}
    \end{bmatrix}
    \,=\, \rhoreg(s)
\end{align}
More generally, any finite-dimensional representations of compact (including finite) groups is \emph{completely reducible} into a direct sum of irreps~\cite{gallier2019harmonicRepr,din2017reprLectureNotes,serre1977linear}.
This suggests that any covariant feature vector, transforming under a compact structure group, can up to a change of basis be constructed from irrep features.
As argued in~\cite{Weiler2019_E2CNN}, it is in this case indeed possible to reduce any \emph{linear} network operation to equivalent operations between irrep fields, which simplifies the construction of the space of $G$-steerable kernels in Eq.~\eqref{eq:G-steerable_kernel_space} and of invariant biases in Eq.~\eqref{eq:gauge_bias_solution_space}.
However, as we will see below, the specific choice of basis of equivalent field types has a quite significant impact on the model performance.
The reason for this is that \emph{nonlinear} network operations are sensitive to the chosen basis, i.e. to a specific choice from equivalent field types.

\subsection{Orientation independent convolutional networks}
\label{sec:mobius_cnn_ops_analytical}

In order to construct orientation independent CNNs on the M\"obius strip we need to instantiate the gauge equivariant layers from Section~\ref{sec:gauge_CNNs_local} for the reflection group~$\Flip$.
More specifically, each of the shared equivariant template functions defining the orientation independent layers needs to be instantiated for any choice of the considered field types $\rhotriv$, $\rhosign$ and $\rhoreg$.
The following Section~\ref{sec:mobius_bias} starts by solving for the spaces $\mathscr{B}^R_\rho$ of gauge invariant bias templates from Eq.~\eqref{eq:gauge_bias_solution_space}
Some admissible choices of gauge equivariant nonlinearities for the different field types are proposed in Section~\ref{sec:mobius_nonlin}.
Section~\ref{sec:mobius_kernel_spaces} will then derive the spaces $\KR$ of $\Flip$-steerable kernels (Eq.~\eqref{eq:G-steerable_kernel_space}) for each possible pair of input and output irreps.
While this section will mainly consist of theoretical derivations, the following Section~\ref{sec:mobius_experiment_main} will cover more practical implementation details.

\subsubsection{Orientation independent bias summation}
\label{sec:mobius_bias}

The space of biases templates that can be summed to a field of type $\rho$ without interfering with the coordinate independence assumption was in Section~\ref{sec:gauge_bias_summation} shown to be given by
\begin{align}
    \mathscr{B}^{\Flip}_\rho\ :=\ \big\{ b \in\R^c \;\big|\; b = \rho(g)\mkern2mu b\ \ \ \forall g\in \Flip \big\} \,.
\end{align}
For the case of the reflection group, there are only two group elements and thus two constraints.
The constraint for the identity element $g=e$ is trivially satisfied since $\rho(e) = \id_{\R^c}$ is by definition always the identity on~$\R^c$.
In the following it is therefore sufficient to restrict attention to the constraint $b = \rho(s)\mkern2mu b$ coming from the reflection $g=s$.

We start with the case of scalar fields, i.e. the trivial representation.
The reflection constraint then reads $b = \rhotriv(s)\mkern2mu b = b$, which is always satisfied.
It follows that the space of bias templates
\begin{align}
    \mathscr{B}^{\Flip}_{\scalebox{.8}{$\rho_{{\textup{triv}}}$}} = \R
\end{align}
remains unconstrained such that arbitrary real-valued biases can be summed to scalar fields.
For the sign-flip representation the reflection constraint becomes $b = \rhotriv(s)\, b = -b$ and is therefore only satisfied for biases which are zero:
\begin{align}
    \mathscr{B}^{\Flip}_{\scalebox{.8}{$\rho_{{\textup{sign}}}$}} = \{0\}
\end{align}
It is thus impossible to sum biases to sign-flip fields while maintaining coordinate independence.
Our third exemplary field type is the two-dimensional regular representation.
The corresponding reflectional constraint on $b\in\R^2$ reads
\begin{align}
    \begin{bmatrix} b_1 \\ b_2 \end{bmatrix}
    \ =\ 
    b
    \ =\ 
    \rhoreg(s)\mkern2mu b
    \ =\ 
    \begin{bmatrix} \hspace{1.5pt}
        0 &\mkern-7mu 1 \hspace*{1.5pt} \\ \hspace{1.5pt} 1 &\mkern-7mu 0 \hspace*{1.5pt}
    \end{bmatrix}
    \!\cdot\! \begin{bmatrix} b_1 \\ b_2 \end{bmatrix}
    \ =\ 
    \begin{bmatrix} b_2 \\ b_1 \end{bmatrix}
\end{align}
and leads to the one-dimensional solution space
\begin{align}\label{eq:bias_solution_space_regular}
    \mathscr{B}^{\Flip}_{\scalebox{.8}{$\rho_{{\textup{reg}}}$}} = 
    \big\{ b\in\R^2 \,\big|\, b_1=b_2 \big\} =
    \bigg\{ \begin{bmatrix} \beta \\ \beta \end{bmatrix} \,\bigg|\, \beta\in\R \bigg\} \ .
\end{align}
The coordinate independence of this constraint is intuitively clear:
since the regular representation swaps the two channels which make up the field, the bias summation is only then coordinate independent when the values summed to both channels are equal, such that their order does not matter.

As already claimed in Section~\ref{sec:gauge_bias_summation}, the solution space $\mathscr{B}^{\Flip}_\rho$ for a representation $\rho$ coincides exactly with its trivial subrepresentations.
This is certainly true for the trivial representation, to which one can sum any bias, and the sign-flip representation, which has itself no trivial subrepresentation and therefore does not admit biases at all.
A more interesting example is the regular representation, which was in Eq.~\ref{eq:rho_reg_decomposition} shown to decompose into a direct sum of the trivial and the sign-flip representation.
The one-dimensional solution space in Eq.~\ref{eq:bias_solution_space_regular} corresponds exactly to the single trivial subrepresentation contained in~$\rhoreg$.
To check the validity of this statement, note that the admissible biases for the direct sum representation $\rhotriv \oplus \rhosign$ are of the form $(\beta,\mkern1mu 0\mkern1mu)^\top$, where $\beta\in\R$.
This results can via the change of basis $Q$ be translated back to the regular representation, which indeed recovers our solution in Eq.~\eqref{eq:bias_solution_space_regular}:
\begin{align}
    Q \cdot\mkern-2mu \begin{bmatrix} \beta \\ 0 \end{bmatrix}
    \ \propto\ 
    \begin{bmatrix} \hspace{1.5pt}
        1 &\mkern-7mu \shortminus1 \hspace*{1.5pt} \\ \hspace{1.5pt} 1 &\mkern-7mu \phantom{\shortminus}1 \hspace*{1.5pt}
    \end{bmatrix}
    \!\cdot\! \begin{bmatrix} \beta \\ 0 \end{bmatrix}
    \ =\ 
    \begin{bmatrix} \beta \\ \beta \end{bmatrix}
\end{align}

\subsubsection{Orientation independent nonlinearities}
\label{sec:mobius_nonlin}

To construct a deep network, we need to come up with equivariant nonlinearities for each of the field types.
As already discussed in Section~\ref{sec:gauge_nonlinearities}, scalar fields can due to their invariance under gauge transformations be acted on by any nonlinearity $\mathscr{s}_{\textup{triv}}: \R\to\R$.
Usual choices are the pointwise ReLU or ELU nonlinearities.
For the sign-flip fields one might take the absolute value $\lVert f^A_{\textup{sign}}(p) \rVert$ of feature vectors, which maps the sign-flip field to a scalar field.
In our implementation below we instead use nonlinearities of the form
\begin{align}\label{eq:signflip_nonlin}
    \mathscr{s}_{\textup{sign}}: \mathscr{f} \ \mapsto\ 
    \operatorname{ReLU} \!\big( \lVert \mathscr{f} \rVert - b \big) \mkern-2mu\cdot\!
    \frac{\mathscr{f}}{\lVert \mathscr{f} \rVert},
\end{align}
where $b\in\R^+$ is a learnable bias parameter.
This choice is easily seen to map sign-flip fields to sign-flip fields since the first multiplicand is acting on the gauge invariant norm of feature vectors while the second multiplicand is preserving the feature vector's sign.
As a permutation representation, the regular representation allows for any pointwise nonlinearity, for instance ReLUs, to act on the individual field channels without changing the field type:
\begin{align}
    \rhoreg(s) \circ \mathscr{s}_{\textup{reg}} \begin{bmatrix} \mathscr{f}_1 \\ \mathscr{f}_2 \end{bmatrix}
    \ =\ 
    \begin{bmatrix} \hspace{1.5pt}
        0 &\mkern-7mu 1 \hspace*{1.5pt} \\ \hspace{1.5pt} 1 &\mkern-7mu 0 \hspace*{1.5pt}
    \end{bmatrix}
    \begin{bmatrix} \operatorname{ReLU}(\mathscr{f}_1) \\ \operatorname{ReLU}(\mathscr{f}_2) \end{bmatrix}
    \ =\ 
    \begin{bmatrix} \operatorname{ReLU}(\mathscr{f}_2) \\ \operatorname{ReLU}(\mathscr{f}_1) \end{bmatrix}
    \ =\ 
    \mathscr{s}_{\textup{reg}} \begin{bmatrix} \mathscr{f}_2 \\ \mathscr{f}_1 \end{bmatrix}
    \ =\ 
    \mathscr{s}_{\textup{reg}} \circ \rhoreg(s) \begin{bmatrix} \mathscr{f}_1 \\ \mathscr{f}_2 \end{bmatrix}
\end{align}
While the regular representation is linearly equivalent to $\rhotriv\oplus\rhosign$, we can not apply independent pointwise nonlinearities on the two channels in the irrep basis.
This substantiates the claim that nonlinearities make the networks sensitive to the particular choice of basis of the representation.

\subsubsection{Orientation independent convolutions}
\label{sec:mobius_kernel_spaces}

The last operations that we instantiate here are reflection equivariant convolutions.
This requires us on the one hand to explain the exponential map and parallel transport on the strip, and on the other hand to solve for the $\Flip$-steerable kernel spaces.
Due to the flat geometry of $M$ and the use of isometric charts, the former is almost trivial:
features are transported similarly as on $\R^2$, with the only difference that the feature vectors might experience a reflection by~$\rho(s)$.
Our implementation of this part is visualized in Fig.~\ref{fig:mobius_conv_numerical}, which is further explained in the following Section~\ref{sec:mobius_implementation}.
The current section focuses exclusively on the analytical solution of the kernel spaces.

For the M\"obius strip with $\Flip$-structure, the space of $\Flip$-steerable kernels from Eq.~\eqref{eq:G-steerable_kernel_space} is given by
\begin{align}
    \KR :=\ \Big\{ K\!: \R^2 \to \R^{\cout\times\cin} \,\Big|\,
    K(g\mkern1mu\mathscr{v}) = \rhoout(g) \mkern-2mu\cdot\mkern-2mu K(\mathscr{v}) \mkern-2mu\cdot\mkern-2mu \rhoin(g) \quad \forall\ \mathscr{v} \in \R^2,\,\ g\in \Flip \Big\} \,,
\end{align}
where we dropped the determinant factor $\detg=1$ and removed all inverses since ${g^{-1} \!=\! g\ \ \ \forall\mkern4mu g \in \Flip}$.
As~in the case of equivariant biases, the constraint coming from the identity element is trivially satisfied, such that only the reflectional constraint remains.
Reflection equivariant kernels are therefore characterized by the single constraint
\begin{align}
    K(s.\mathscr{v})\ =\ \rhoout(s) \mkern-1mu\cdot\mkern-1mu K(\mathscr{v}) \mkern-1mu\cdot\mkern-1mu \rhoin(s) \qquad \forall\: \mathscr{v} \in \R^2 \,,
\end{align}
which requires that the reflected kernel $K(s.\mathscr{v})$ equals the non-reflected kernel $K(\mathscr{v})$ after being acted on by the input and output field representation.
We will in the following solve this constraint for all nine pairs of field types.
The resulting kernels, all of which are in one or another sense symmetric under reflections, are visualized in Table~\ref{tab:reflection_steerable_kernels}.

\begin{SCtable}
    \hspace*{-5.ex}
    \def\figfolder{figures/kernels/}
    \def\figheight{0.08\columnwidth}
    \newlength{\vertSkip}
    \newlength{\eqToKernelSkip}
    \setlength\vertSkip{4.5ex}
    \setlength\eqToKernelSkip{1.5ex}
    \newcommand\cellTstrut{\rule{0pt}{3.5ex}}
    \footnotesize
    \setlength{\tabcolsep}{4pt}
    \begin{tabular}{r|c|c|c}
        \diagbox[height=18pt]{\raisebox{3pt}{$\rhoout$}}{\raisebox{-0pt}{$\rhoin$}}
        & trivial & sign-flip & regular \\
        \hline
        \cellTstrut
        \multirow{2}{*}{
            \hspace*{-1.5ex}
            trivial
            \hspace*{-1ex}
            \rule{0pt}{5.5ex}
            }
        & $\Koo(s.\mathscr{v}) = \Koo(\mathscr{v})$
        & $\Koo(s.\mathscr{v}) = \minus\Koo(\mathscr{v})$
        & $\Koo(s.\mathscr{v}) = \Kot(\mathscr{v})$
        \\[\eqToKernelSkip]
        & $\begin{pmatrix} \includegraphicstotab[height=\figheight]{\figfolder K_symmetric.png} \end{pmatrix}$
        & $\begin{pmatrix} \includegraphicstotab[height=\figheight]{\figfolder K_antisymmetric.png} \end{pmatrix}$
        & $\begin{pmatrix}
            \includegraphicstotab[height=\figheight]{\figfolder K_crisp_1.png} \mkern4mu, & \mkern-16mu
            \includegraphicstotab[height=\figheight]{\figfolder K_crisp_1f.png}
          \end{pmatrix}$ 
        \hspace{-6pt}
        \\[\vertSkip]
        \hline
        \cellTstrut
        \multirow{2}{*}{
            \hspace*{-1.5ex}
            sign-flip
            \hspace*{-1ex}
            \rule{0pt}{5.5ex}
            }
        & $\Koo(s.\mathscr{v}) = \minus\Koo(\mathscr{v})$
        & $\Koo(s.\mathscr{v}) = \Koo(\mathscr{v})$
        & $\Koo(s.\mathscr{v}) = \minus\Kot(\mathscr{v})$
        \\[\eqToKernelSkip]
        & $\begin{pmatrix} \includegraphicstotab[height=\figheight]{\figfolder K_antisymmetric.png} \end{pmatrix} $
        & $\begin{pmatrix} \includegraphicstotab[height=\figheight]{\figfolder K_symmetric.png} \end{pmatrix} $
        & $\begin{pmatrix}
            \includegraphicstotab[height=\figheight]{\figfolder K_crisp_1.png} \mkern4mu, & \mkern-16mu
            \includegraphicstotab[height=\figheight]{\figfolder K_crisp_1f_inverted.png}
          \end{pmatrix} $ 
        \hspace{-6pt}
        \\[\vertSkip]
        \hline
        \rule{0pt}{5.ex}
        \multirow{2}{*}{
            \hspace*{-1.5ex}
            regular
            \hspace*{-1ex}
            \rule{0pt}{10.5ex}
            }
        & $\Koo(s.\mathscr{v}) = \Kto(\mathscr{v})$
        & $\Koo(s.\mathscr{v}) = \minus\Kto(\mathscr{v})$
        & \makecell{
            $\Koo(s.\mathscr{v}) = \Ktt(\mathscr{v})$ \\[.6ex]
            $\Kot(s.\mathscr{v}) = \Kto(\mathscr{v})$
          }
        \hspace{-6pt}
        \\[3.ex]
        & $\begin{pmatrix}
            \includegraphicstotab[height=\figheight]{\figfolder K_crisp_1.png} \\
            \includegraphicstotab[height=\figheight]{\figfolder K_crisp_1f.png}
          \end{pmatrix}$
        & $\begin{pmatrix}
            \includegraphicstotab[height=\figheight]{\figfolder K_crisp_1.png} \\
            \includegraphicstotab[height=\figheight]{\figfolder K_crisp_1f_inverted.png}
          \end{pmatrix}$
        & $\begin{pmatrix}
            \includegraphicstotab[height=\figheight]{\figfolder K_crisp_1.png} \mkern4mu, & \mkern-16mu
            \includegraphicstotab[height=\figheight]{\figfolder K_crisp_3.png} \\
            \includegraphicstotab[height=\figheight]{\figfolder K_crisp_3f.png} \mkern4mu, & \mkern-16mu
            \includegraphicstotab[height=\figheight]{\figfolder K_crisp_1f.png}
          \end{pmatrix}$
    \end{tabular}
    \captionsetup{width=1.\textwidth}
    \caption[]{\small
        \protect\rule{0ex}{8ex} 
        Visualization of reflection-steerable kernels in $\KR$ for all considered pairs of input and output field types $\rhoin$ and $\rhoout$.
        In general, these kernels need to satisfy the $\Flip\mkern-1.2mu$-steerability kernel constraint $K(s.\mathscr{v})\ =\ \rhoout(s) \mkern-1mu\cdot\mkern-1mu K(\mathscr{v}) \mkern-1mu\cdot\mkern-1mu \rhoin(s)$ where $K: \R^2 \to \R^{\cout\times\cin}$.
        Each entry of the table states the specific constraint for the corresponding input and output representations and visualizes one exemplary steerable kernel.
        Note that the constraint binds the reflected kernel $K(s.\mathscr{v})$ to a linear transformation of the unreflected kernel $K(\mathscr{v})$ by the input and output representation.
        It results therefore in reflectional symmetries of the kernels.
    }
    \label{tab:reflection_steerable_kernels}
\end{SCtable}

\begin{itemize}
\setlength{\itemindent}{-2em}
\setlength{\itemsep}{.5em}
    \item[{\rule[2.0pt]{2pt}{2pt}}]
    \textbf{scalar $\bm\to$ scalar:}
        Kernels
        $K = \pig[\Koo\pig]: \R^2 \to \R^{1\times1}$
        which map between scalar fields are required to satisfy the constraint
        \begin{align}
                \pig[\Koo\pig] \mkern-1mu (s.\mathscr{v})
            \ =\ 
                \big[\mkern2mu 1 \mkern2mu\big]
                \!\cdot\!
                \pig[\Koo\pig] \mkern-1mu (\mathscr{v})
                \cdot\!
                \big[\mkern2mu 1 \mkern2mu\big]
            \ =\ 
                \pig[\Koo\pig] \mkern-1mu (\mathscr{v})
            \qquad \forall\ \mathscr{v} \in \R^2 \,.
        \end{align}
        They are necessarily \emph{symmetric} (invariant) under reflections; see the upper left entry in Table~\ref{tab:reflection_steerable_kernels}.
    \item[{\rule[2.0pt]{2pt}{2pt}}]
    \textbf{scalar $\bm\to$ sign-flip:}
        The kernels $K = \pig[\Koo\pig]: \R^2 \to \R^{1\times1}$ which map a scalar field to a sign-flip field need to satisfy
        \begin{align}
                \pig[\Koo\pig] \mkern-1mu (s.\mathscr{v})
            \ =\ 
                \big[\! \shortminus\!1 \big]
                \!\cdot\!
                \pig[\Koo\pig] \mkern-1mu (\mathscr{v})
                \cdot\!
                \big[\mkern2mu 1 \mkern2mu\big]
            \ =\
                -\pig[\Koo\pig] \mkern-1mu (\mathscr{v})
            \qquad \forall\ \mathscr{v} \in \R^2 \,.
        \end{align}
        This implies \emph{antisymmetric} kernels as visualized in the middle row in the first column of Table~\ref{tab:reflection_steerable_kernels}.
    \item[{\rule[2.0pt]{2pt}{2pt}}]
    \textbf{scalar $\bm\to$ regular:}
        In order to map from a scalar field to a regular feature field one needs to apply kernels of the form
        $K = \begin{bmatrix} \Koo \\ \Kto \end{bmatrix}: \R^2 \to \R^{2\times1}$,
        which map from one input channel to two output channels.
        The demanded permutation of the output channels is guaranteed if the kernel satisfies
        \begin{align}\label{eq:constraint_s2r}
                \begin{bmatrix} \Koo \\ \Kto \end{bmatrix} \mkern-4mu (s.\mathscr{v})
            \ =\ 
                \begin{bmatrix} \hspace{1.5pt} 0 &\mkern-4mu 1 \hspace*{1.5pt} \\ \hspace{1.5pt} 1 &\mkern-4mu 0 \hspace*{1.5pt} \end{bmatrix}
                \!\cdot\!
                \begin{bmatrix} \Koo \\ \Kto \end{bmatrix} \mkern-4mu (\mathscr{v})
                \cdot\!
                \big[\mkern2mu 1 \mkern2mu\big]
            \ =\ 
                \begin{bmatrix} \Kto \\ \Koo \end{bmatrix} \mkern-4mu (\mathscr{v})
            \qquad \forall\ \mathscr{v} \in \R^2 \,.
        \end{align}
        This constraint requires that the two channels contain kernels which are \emph{reflected copies} of each other, that is,
        $\Koo(s.\mathscr{v}) = \Kto(\mathscr{v})$ for all $\mathscr{v} \in \R^2$
        (this already covers the second line of the constraint in Eq.~\eqref{eq:constraint_s2r}).
        This case is visualized in the bottom left entry of Table~\ref{tab:reflection_steerable_kernels}.
    \item[{\rule[2.0pt]{2pt}{2pt}}]
    \textbf{sign-flip $\bm\to$ scalar:}
        Kernels
        $K = \pig[\Koo\pig]: \R^2 \to \R^{1\times1}$
        that map from sign-flip to scalar fields are again \emph{antisymmetric} since they need to satisfy the same constraint
        \begin{align}
                \pig[\Koo\pig] \mkern-1mu (s.\mathscr{v})
            \ =\ 
                \big[\mkern2mu 1 \mkern2mu\big]
                \!\cdot\!
                \pig[\Koo\pig] \mkern-1mu (\mathscr{v})
                \cdot\!
                \big[\! \shortminus\!1 \big]
            \ =\ 
                -\pig[\Koo\pig] \mkern-1mu (\mathscr{v})
            \qquad \forall\ \mathscr{v} \in \R^2
        \end{align}
        like kernels which map in the opposite direction.
    \item[{\rule[2.0pt]{2pt}{2pt}}]
    \textbf{sign-flip $\bm\to$ sign-flip:}
        The kernels
        $K = \pig[\Koo\pig]: \R^2 \to \R^{1\times1}$
        which preserve the transformation behavior of sign-flip fields are \emph{symmetric} since the two sign inversions in the constraint
        \begin{align}
                \pig[\Koo\pig] \mkern-1mu (s.\mathscr{v})
            \ =\ 
                \big[\! \shortminus\!1 \big]
                \!\cdot\!
                \pig[\Koo\pig] \mkern-1mu (\mathscr{v})
                \cdot\!
                \big[\! \shortminus\!1 \big]
            \ =\ 
                \pig[\Koo\pig] \mkern-1mu (\mathscr{v})
            \qquad \forall\ \mathscr{v} \in \R^2
        \end{align}
        cancel out.
    \item[{\rule[2.0pt]{2pt}{2pt}}]
    \textbf{sign-flip $\bm\to$ regular:}
        In the case of kernels
        $K = \begin{bmatrix} \Koo \\ \Kto \end{bmatrix}: \R^2 \to \R^{2\times1}$
        which map from sign-flip to regular feature fields, we get the constraint
        \begin{align}
                \begin{bmatrix} \Koo \\ \Kto \end{bmatrix} \mkern-4mu (s.\mathscr{v})
            \ =\ 
                \begin{bmatrix} \hspace{1.5pt} 0 &\mkern-4mu 1 \hspace*{1.5pt} \\ \hspace{1.5pt} 1 &\mkern-4mu 0 \hspace*{1.5pt} \end{bmatrix}
                \!\cdot\!
                \begin{bmatrix} \Koo \\ \Kto \end{bmatrix} \mkern-4mu (\mathscr{v})
                \cdot\!
                \big[\! \shortminus\!1 \big]
            \ =\ 
                - \begin{bmatrix} \Kto \\ \Koo \end{bmatrix} \mkern-4mu (\mathscr{v})
            \qquad \forall\ \mathscr{v} \in \R^2 \,.
        \end{align}
        The two lines imply each other, such that they can be summarized by the single kernel constraint
        ${\Koo(s.\mathscr{v}) = -\Kto(\mathscr{v})}\ \ \forall \mathscr{v}\in \R^2$.
        This constraint requires that the two channels of the kernel contain \emph{reflected, negated copies} of each other; see the visualization in the middle of the bottom row of Table~\ref{tab:reflection_steerable_kernels}.
    \item[{\rule[2.0pt]{2pt}{2pt}}]
    \textbf{regular $\bm\to$ scalar:}
        The kernels which map regular feature fields to scalar fields have two input channels and one output channel and are therefor of the form
        $K = \pig[\Koo \,,\, \Kot\pig]: \R^2 \to \R^{1\times2}$.
        The constraint
        \begin{align}
                \pig[\Koo \,,\, \Kot\pig] \mkern-1mu (s.\mathscr{v})
            \ =\ 
                \big[\mkern2mu 1 \mkern2mu\big]
                \!\cdot\!
                \pig[\Koo \,,\, \Kot\pig] \mkern-1mu (\mathscr{v})
                \cdot\!
                \begin{bmatrix} \hspace{1.5pt} 0 &\mkern-4mu 1 \hspace*{1.5pt} \\ \hspace{1.5pt} 1 &\mkern-4mu 0 \hspace*{1.5pt} \end{bmatrix}
            \ =\ 
                \pig[\Kot \,,\, \Koo\pig] \mkern-1mu (\mathscr{v})
            \qquad \forall\ \mathscr{v} \in \R^2 \,,
        \end{align}
        which can be reduced to the requirement
        $\Koo(s.\mathscr{v}) = \Kot(\mathscr{v})\ \ \forall \mathscr{v} \in \R^2$,
        again demands that the two entries of the kernel contain \emph{reflected copies} of each other.
    \item[{\rule[2.0pt]{2pt}{2pt}}]
    \textbf{regular $\bm\to$ sign-flip:}
        Mappings from regular feature fields to sign-flip fields utilize kernels
        $K = \pig[\Koo,\, \Kot\pig]: \R^2 \to \R^{1\times2}$
        that satisfy
        \begin{align}
                \pig[\Koo \,,\, \Kot\pig] \mkern-1mu (s.\mathscr{v})
            \ =\ 
                \big[\! \shortminus\!1 \big]
                \!\cdot\!
                \pig[\Koo \,,\, \Kot\pig] \mkern-1mu (\mathscr{v})
                \cdot\!
                \begin{bmatrix} \hspace{1.5pt} 0 &\mkern-4mu 1 \hspace*{1.5pt} \\ \hspace{1.5pt} 1 &\mkern-4mu 0 \hspace*{1.5pt} \end{bmatrix}
            \ =\ 
                - \pig[\Kot \,,\, \Koo\pig] \mkern-1mu (\mathscr{v})
            \qquad \forall\ \mathscr{v} \in \R^2 \,,
        \end{align}
        or, equivalently, $\Koo(s.\mathscr{v}) = -\Kot(\mathscr{v})\ \ \forall \mathscr{v} \in \R^2$.
        As probably already expected, they are made up from kernels whose two channels contain \emph{reflected, negated copies} of another.
    \item[{\rule[2.0pt]{2pt}{2pt}}]
    \textbf{regular $\bm\to$ regular:}
        Lastly, we need to consider kernels
        $K = \begin{bmatrix} \Koo &\mkern-12mu \Kot \\ \Kto &\mkern-12mu \Ktt \end{bmatrix}: \R^2 \to \R^{2\times2}$
        which map regular fields to regular fields and therefore have $2\times2$ matrices as codomain.
        Their constraint, coming from a left and right multiplication with the regular representation, becomes
        \begin{align}
                \begin{bmatrix} \Koo &\mkern-12mu \Kot \\ \Kto &\mkern-12mu \Ktt \end{bmatrix} \mkern-4mu (s.\mathscr{v})
            \ =\ 
                \begin{bmatrix} \hspace{1.5pt} 0 &\mkern-4mu 1 \hspace*{1.5pt} \\ \hspace{1.5pt} 1 &\mkern-4mu 0 \hspace*{1.5pt} \end{bmatrix}
                \!\cdot\!
                \begin{bmatrix} \Koo &\mkern-12mu \Kot \\ \Kto &\mkern-12mu \Ktt \end{bmatrix} \mkern-4mu (\mathscr{v})
                \cdot\!
                \begin{bmatrix} \hspace{1.5pt} 0 &\mkern-4mu 1 \hspace*{1.5pt} \\ \hspace{1.5pt} 1 &\mkern-4mu 0 \hspace*{1.5pt} \end{bmatrix}
            \ =\ 
                \begin{bmatrix} \Ktt &\mkern-12mu \Kto \\ \Kot &\mkern-12mu \Koo \end{bmatrix} \mkern-4mu (\mathscr{v})
            \qquad \forall\ \mathscr{v} \in \R^2 \,.
        \end{align}
        This is equivalent to the two independent constraints
        \begin{align}
            \Koo(s.\mathscr{v}) = \Ktt(\mathscr{v}) \quad \forall \mathscr{v} \in \R^2
        \end{align}
        and
        \begin{align}
            \Kot(s.\mathscr{v}) = \Kto(\mathscr{v}) \quad \forall \mathscr{v} \in \R^2 \,,
        \end{align}
        which couple the four kernel entries such that there are \emph{two pairs of mutually reflected kernels}.
        This case is visualized in the bottom right entry of Table~\ref{tab:reflection_steerable_kernels}.
\end{itemize}

While the derived results tell us how to map between individual feature fields, neural networks typically assume feature spaces that consist of multiple, potentially differing feature fields.
The kernels which map between these stacks of feature fields can be thought of as being built from blocks which map between the individual fields.
To give an example, consider the case where both the input and output feature spaces contain one of the discussed representations each, that is, $\rhoin = \rhoout = \rhotriv \oplus \rhosign \oplus \rhoreg$.
The number of input and output channels is then $\cin=\cout=1+1+2=4$, such that the full kernel is of the form $K: \R^2 \to \R^{4\times4}$.
Since the input and output representations are defined as direct sums, they are block diagonal.
The full constraint decouples thus into nine independent constraints between all pairs of individual input and output fields, which correspond in this case exactly to the nine solutions presented above.
The $4\times4$ entries of the full kernel will therefore be required to have the same symmetries as the $4\times4$ kernels which are visualized in Table~\ref{tab:reflection_steerable_kernels} as a whole.

For completeness, we briefly elaborate on general kernel field transforms and isometry equivariant kernel field transforms on the M\"obius strip.
In the general case the smooth kernel field remains entirely unrestricted, that is, no weights need to be shared and the individual kernels are not required to have any reflectional symmetries whatsoever.
In order for the kernel field transform to be equivariant w.r.t. isometries, the applied kernel field is required to be invariant under their action.
This requires weights to be shared over the isometry orbits, which come in two different types.
The first type corresponds to that single orbit which lies exactly in the middle of the strip.
Points on this orbit return back to themselves after one revolution around the strip, while the strip itself is reflected over this central orbit.
An $\IsomRM$-invariant kernel field will have some kernel shared over this central orbit, however, this kernel is required to have reflectional symmetries like the $\Flip$-steerable kernels in Table~\ref{tab:reflection_steerable_kernels}.
This is the case since the kernels are after one revolution mapped in a reflected way on themselves while the kernel field is required to be isometry invariant.%
\footnote{
    Section~\ref{sec:quotient_kernel_fields} describes such situations in a more general setting as \emph{stabilizer subgroup} constraints of the isometry group.
    In the current case, the subgroup of rotations once around the strip stabilizes the points on the central orbit.
    It is isomorphic to the reflection group and therefore leads to reflectional symmetries in the kernels.
}
Any other orbit is of the second orbit type.
Consider some point at a given distance from the central orbit.
The isometry action will move this point at this distance from the center along the strip.
However, it will not return to the initial point after one revolution but to that point which lies at the same distance on the opposite of the central orbit.
Only after a second revolution around the strip the orbit will close.
The demanded isometry invariance of the kernel field will thus require kernels to be shared over all points with the same distance in either direction of the strip's center (but allows for different kernels at different distances).
In contrast to the central orbit's kernel, these kernels are not required to be reflection equivariant themselves.
This analysis shows that any isometry equivariant kernel field transform requires $\Flip$-steerable kernels, although strictly only on the central orbit.
Conversely, the convolutional kernel field, corresponding to the application of the same $\Flip$-steerable kernels on the whole manifold, is certainly invariant under isometries.
The orientation independent convolution on the M\"obius strip is therefore $\IsomRM$-equivariant, which is empirically confirmed below.

\subsection{Numerical implementation and evaluation of M\"obius convolutions}
\label{sec:mobius_experiment_main}

Being prepared with the analytical derivations in the previous sections we are ready to discuss a numerical implementation of orientation independent CNNs on the M\"obius strip.
After doing so in Section~\ref{sec:mobius_implementation}, we evaluate the models for different choices of field types and compare them with a naive, orientation dependent implementation in Section~\ref{sec:mobius_evaluation}.

The implementation is publicly available at \url{https://github.com/mauriceweiler/MobiusCNNs}.

\subsubsection{Numerical implementation}
\label{sec:mobius_implementation}

\paragraph{Feature spaces:}
The first question to answer when implementing convolutions on the M\"obius strip is how the feature fields should be represented numerically.
Since the M\"obius strip is a flat manifold, we can conveniently inherit (subsets of) the regular sampling grid $\Z^2$ from $\R^2$ over to the strip.
This intuition is formalized by the pullbacks
$f^X\circ \big(x^X\big)^{-1}: V^X \to \R^c$
of the local feature field coordinatizations
$f^X: U^X \to \R^c$
via the (inverse) charts
$\big(x^X\big)^{-1}: V^X \to U^X$
to a new domain $V^X \subset \R^2$,
where $X=A,B$.
The numerical discretization is then defined as a restriction
$f^X\circ \big(x^X\big)^{-1} \big|_{V^X\cap\mkern2mu\Z^2}$
of the pullback to the sampling grid, which is in Fig.~\ref{fig:mobius_conv_gauges} shown as an overlay.

\begin{figure}
    \begin{subfigure}[t!]{.6\linewidth}
        \includegraphics[width=\linewidth]{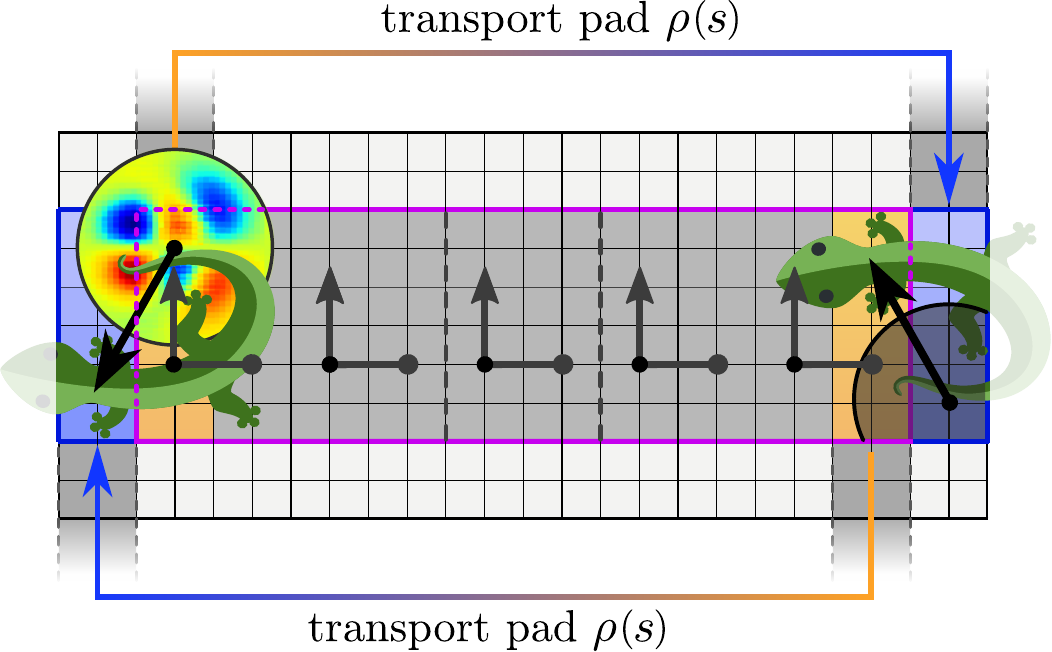}
    \end{subfigure}
    \par
    \vspace*{\dimexpr-\parskip-160.pt\relax}
    \parshape 18 
        .62\textwidth .37\textwidth 
        .62\textwidth .37\textwidth
        .62\textwidth .37\textwidth
        .62\textwidth .37\textwidth
        .62\textwidth .37\textwidth
        .62\textwidth .37\textwidth
        .62\textwidth .37\textwidth
        .62\textwidth .37\textwidth
        .62\textwidth .37\textwidth
        .62\textwidth .37\textwidth
        .62\textwidth .37\textwidth
        .62\textwidth .37\textwidth
        .62\textwidth .37\textwidth
        .62\textwidth .37\textwidth
        .62\textwidth .37\textwidth
        .62\textwidth .37\textwidth
        .62\textwidth .37\textwidth
        .01\textwidth .98\textwidth 
    \makeatletter
    \setcounter{\@captype}{\value{\@captype}-1} 
    \refstepcounter{\@captype}
    \addcontentsline{\csname ext@\@captype\endcsname}{\@captype}
        {\protect\numberline{\csname the\@captype\endcsname}{ToC entry}}%
    \small 
    \csname fnum@\@captype\endcsname: 
    \makeatother
        Numerical representation of feature fields on the M\"obius strip and Levi-Civita transport of feature vectors during the convolution.
        The flat geometry of the strip allows to cut it open and flatten it out isometrically to the magenta rectangle.
        When assigning the canonical reference frames of~$\R^2$ this corresponds to a gluing of the two charts $V^A$ and $V^B$ from Fig.~\ref{fig:mobius_conv_gauges} at their overlap with trivial gauge transformations (``$\id$'').
        In order to avoid redundancies, we assign half of the width of the second chart overlap with reflective gauge transformations (``$\operatorname{flip}$'') to either end of the flattened magenta strip (orange pixels).
        Feature fields are stored as an array with spatial dimensions corresponding to the magenta box and $c$~channels.
        During the convolution operation, the kernel collects features from all pixels that it covers.
        Choosing a kernel size of $5\times5$ pixels, we need to specify all values in a radius of $2$ pixels around its center, which overall requires to pad a border region of $2$ pixels around the magenta rectangle.
        The border at the top and bottom correspond to the boundary of the strip.
        Since no valid feature values can be assigned there, we zero-pad the array as commonly done in computer vision.
        The left and right border of the flattened strip are glued together with a twist.
        We implement the parallel transport of those features by cutting an area of two pixels width from either end of the strip (orange) and padding them in a reflected way to the opposite ends (blue).
        As the twist implies a gauge transformation, the feature fields need to be acted on by $\rho(s)$ when being reflected.
        After padding, the convolution is run with ``valid'' boundary conditions, such that its output again has the size of the magenta box.
        Operations which act pointwise do not require the padding but can be applied to the magenta array right away.
        {
        \\ \color{gray} \scriptsize
            (Lizards adapted under the Creative Commons Attribution 4.0 International
            \href{https://github.com/twitter/twemoji/blob/gh-pages/LICENSE-GRAPHICS}{\underline{license}}
            by courtesy of Twitter.)
        }
    \label{fig:mobius_conv_numerical}
\end{figure}

Note that this representation is due to the overlap $U^A \cap U^B \neq \varnothing$ of the charts redundant.
To remove this redundancy one needs to identify those regions that are represented twice and store only one, shared copy of the corresponding feature vectors.
One possible scheme to do so, which we use in our numerical implementation, is to store the feature fields in the multidimensional array corresponding to the magenta rectangle in Fig.~\ref{fig:mobius_conv_numerical}.
It can be thought of as being defined by ``gluing'' those regions in $V^A$ and $V^B$ which are identified by \emph{trivial} gauge transformations $g^{BA}$ together (``$\id$'' in Fig.~\ref{fig:mobius_conv_gauges} and central four pixels in Fig.~\ref{fig:mobius_conv_numerical}).
What remains is a redundancy of feature vectors at the second overlapping region with reflecting gauge transformations (``$\operatorname{flip}$'' in Fig.~\ref{fig:mobius_conv_gauges}).
It is resolved by assigning those feature vectors in equal parts to either end of the ``glued'' local field representation (orange pixels in Fig.~\ref{fig:mobius_conv_numerical}).
Together, the pixels in the magenta box represent the feature space in a non-redundant way by assigning a $c$-dimensional feature vector to each of them.
The ring of two pixels around the magenta rectangle is \emph{not} part of the feature space but visualizes a padding region which will only be used during the forward pass of the convolution operation as discussed below.

The actual dimensions (shape) of the array that encodes a feature space depend on the chosen field multiplicities.
Let $m_{\textup{triv}}$, $m_{\textup{sign}}$ and $m_{\textup{reg}}$ be those integer multiplicities of feature fields which make up a feature space.
Since the scalar and sign-flip fields are one-dimensional and the regular feature fields are two-dimensional, the overall number of channels (or dimensionality of stacked feature vectors) is given by $c = m_{\textup{triv}} + m_{\textup{sign}} + 2 m_{\textup{reg}}$.
Assume further that the spatial resolution of the magenta rectangle is $X\times Y$ pixels and assume a batch size of $N$ samples.
The array that encodes a feature space is then of shape $(N,c,X,Y)$, as usual in image processing.
Note that this numerical representation of the feature space is both agnostic to the twisted geometry of the strip and the actual type of the contained feature fields (except for their dimensionality).
The actual geometric information is therefore solely carried by the network layers which process the fields.

\paragraph{Bias summation:}
To implement the orientation independent bias summation, recall the results from Section~\ref{sec:mobius_bias} that the vector spaces of reflection equivariant template biases are for scalar fields and regular fields one-dimensional and for sign-flip fields zero-dimensional.
At initialization of the bias module we therefore allocate an $m_{\textup{triv}}$-dimensional parameter vector $\beta_{\textup{triv}}$ and an $m_{\textup{reg}}$-dimensional parameter vector $\beta_{\textup{reg}}$.
During the forward pass we expand these parameters into a $c$-dimensional bias vector $b_{\textup{full}}$, which is to be summed to the full stack of feature fields.
This is done by allocating a $c$-dimensional array of zeros and filling the first $m_{\textup{triv}}$ elements with the scalar field bias parameters and the last $2m_{\textup{reg}}$ elements with the $m_{\textup{reg}}$ regular field bias parameters, each repeated twice to satisfy the structure of the solution space in Eq.~\eqref{eq:bias_solution_space_regular}.
After this expansion the full bias vector
\begin{align}
    b_{\textup{full}}\ =\ \big[
        \underbrace{
            \beta_{\textup{triv},1}, \,\dots,\, \beta_{\textup{triv}, m_{\textup{triv}}}
        }_{m_{\textup{triv}}},\ 
        \underbrace{
            0, \,\dots,\, 0,\ 
        }_{m_{\textup{sign}}},\ 
        \underbrace{
            \beta_{\textup{reg},1},\beta_{\textup{reg},1}, \,\dots,\, \beta_{\textup{reg}, m_{\textup{reg}}}, \beta_{\textup{reg}, m_{\textup{reg}}}
        }_{2m_{\textup{reg}}}
    \big]^\top \ \in\ \R^c
\end{align}
is summed to the feature field array as usual.
Its orientation independence (gauge invariance) justifies the summation to the array in Fig.~\ref{fig:mobius_conv_numerical}, despite it being glued from feature vectors in two different gauges.

\paragraph{Nonlinearities:}
The nonlinearities can be implemented straightforwardly as defined in Section~\ref{sec:mobius_nonlin}.
We do this by splitting the full stack of feature fields into three stacks of fields of the same type, applying the respective reflection equivariant nonlinearities to them, and finally concatenating the results.
Due to the definition of the nonlinearity for sign-flip fields in Eq.~\eqref{eq:signflip_nonlin} with a learnable bias, the nonlinearity module has $m_{\textup{sign}}$ trainable parameters.

\paragraph{\textit{GM}-convolutions:}
Since convolution operations do not operate pointwise but accumulate all features covered by the kernel, their implementation is less trivial.
The forward pass is split in three steps, namely
1) the expansion of reflection symmetric kernels from parameter arrays,
2) a Levi-Civita transport of feature vectors and
3) the actual $\GM$-convolution.

Recall that the space $\KR$ of $\Flip$-steerable kernels is a linear subspace of the space of unconstrained kernels $\mathscr{K}$ in Eq.~\eqref{eq:unconstrained_kernel_space}.
To parameterize $\Flip$-steerable kernels it is necessary to choose a basis of $\KR$, in terms of which the kernels are expanded.
The trainable parameters of the convolution operation are the expansion coefficients in this basis.
Our implementation parameterizes all kernels that correspond to the same pair of input and output field types jointly
since they share the same symmetries and thus basis.
Considering the nine pairs of field types shown in Table~\ref{tab:reflection_steerable_kernels}, this means that the convolution module holds nine corresponding parameter arrays.
The actual kernels are then expanded from these parameters during each forward pass.
To give an example, consider the subset of kernel channels that map from $m_{\textup{triv}}$ scalar fields to $m_{\textup{sign}}$ sign-flip fields and assume a kernel size of $s\times s$ pixels.
The corresponding parameter array is then of shape $(m_{\textup{sign}},\, m_{\textup{triv}},\, \frac{s}{2},\, s)$ and represents the $m_{\textup{triv}}\times m_{\textup{sign}}$ individual kernel channels with a basis of  $\frac{s}{2}\times s$ \emph{antisymmetric} kernels each.
The expansion is implemented as filling the upper $\frac{s}{2}\times s$ pixels with the unaltered parameters while the bottom $\frac{s}{2}\times s$ pixels are filled with the negated and reflected parameters.
As a second example, consider the kernel channels that map from $m_{\textup{reg}}$ regular fields to $m_{\textup{triv}}$ scalar fields.
The parameter array for this case is of shape $(m_{\textup{triv}},\, m_{\textup{reg}},\, s,\, s)$ and stores one of the two kernel channels per input and output field.
The second, symmetric channels are during the forward pass expanded by reflecting the first kernel channels as shown in Table~\ref{tab:reflection_steerable_kernels}.
After expanding the full kernel in this fashion from all of its sub-blocks corresponding to the different combinations of field types, it has the usual shape of kernels in deep learning but is guaranteed to respect the symmetries derived in Section~\ref{sec:mobius_kernel_spaces}.
Note that the kernel symmetries make $\GM$-convolutions more parameter efficient than a corresponding non-equivariant CNN with the same number of channels~$c$.
Specifically for $\Flip$-steerable kernel the number of parameters reduced by a factor of two%
\footnote{
    The improved parameter efficiency of $\Flip$-steerable kernels by a factor of $2$ is exact for continuous kernels or for even kernel sizes $s$.
    If $s$ is odd, the number of parameters scales for symmetric kernels like $s(s+1)/2$ and for antisymmetric kernels like $s(s-1)/2$ since the former are freely parameterizing the central row of pixels while the latter need to set them to zero.
}.

After expanding the kernels, they are convolved with the feature fields.
This requires an implementation of the exponential map and the $\Flip$-compatible Levi-Civita transporters on the M\"obius strip -- or rather on its numerical representation by the magenta array from Fig.~\ref{fig:mobius_conv_numerical}.
The flat geometry of the M\"obius strip makes the implementation almost trivial, however, its boundaries and circular connectivity require some special care.
We therefore need to distinguish between three qualitatively different cases, which correspond to
1) exponential maps that lie completely within the magenta array,
2) exponential maps that would cross a boundary and are therefore not well defined and
3) exponential maps whose geodesics run out at one end of the array and enter it (twisted) at the other end.
The first case is trivial and corresponds to the exponential map on $\R^2$ itself.
Since the strip is flat and the reference frames within the array are all parallel, the transport along these geodesics is trivial.
Within the interior region of the array, where the (finitely supported) kernels do not protrude out of it, one can therefore implement the convolution as usual on $\R^2$.
The second case concerns the top and bottom rows of the array where the exponential maps might cross the boundary of the strip (or array).
This is analogous to the boundary problems for usual flat, rectangular images, where the issue is most commonly solved via zero-padding.
Adopting this solution, we pad the array with rows of zeros, shown as the two light gray strips above and below the magenta rectangle in Fig.~\ref{fig:mobius_conv_numerical}.
Given a kernel size of $s\times s$ pixels with $s$ being odd, one needs to pad $(s-1)/2$ rows of zeros at both sides.
The third case occurs at the left and right end of the array, where the strip was cut open to flatten it out.
Fig.~\ref{fig:mobius_conv_numerical} visualizes an exemplary geodesic which crosses the cut line and therefore enters the array in a reflected direction at the opposite side.
Due to the reflection, the parallel transporter across the cut is given by $\rho(s)$.
In order to be able to run a conventional convolution routine, we implement the transport across the cut by copying a region of $(s-1)/2$ pixels from both ends of the array (orange), reflecting them upside down to model the twist, acting on them with $\rho(s)$ to account for the reflected gauges and finally appending them to the opposite side of the array (blue).
Having padded the array in this way, all relevant geodesics and transporters are reduced to their trivial counterparts on $\R^2$.

Overall, our implementation of the convolution operation applies the three steps mentioned above.
It first expands the $\Flip$-steerable kernels and pads the magenta feature field array with zeros and the field values which are transported over the cut.
The expanded kernel is then convolved with the padded feature fields, calling a conventional convolution routine for flat images.
We use ``valid'' boundary settings for the convolution, which means that the operation does not implicitly pad further zeros and only computes feature vectors for those points where the kernel does not protrude beyond the boundaries of our manually padded array.
The resulting feature field will therefore again have the same spatial dimensions as the original magenta rectangle.

\paragraph{Pooling:}
Conventional CNNs usually apply spatial pooling operations which summarize feature vectors from a given pooling window, for instance a region of $2\times2$ pixels, into a new feature vector.
Such operations reduce the spatial resolution, which lowers the computational cost and increases the effective field of view of the convolution kernels.
A common way of pooling is the so called ``max-pooling'', which takes the maximum value of each individual feature channel in the pooling region.
This operation can be applied to scalar fields right away since they are gauge invariant.
It is further admissible for regular feature fields since taking the maximum commutes with the permutation of channels.
However, as sign-flip fields change their sign under gauge transformations, max-pooling is not equivariant w.r.t. their transformation law.
An equivariant alternative is average pooling, which takes the average of features in the pooling region and therefore commutes with a change of sign.
Another option, that we use in our experiments below since it performs slightly better, is to pool sign-flip fields based on their absolute value, which is again invariant under sign inversions.
We then multiply the sign of the pooled field values with maximum norm back in to preserve the original transformation law.

While such defined pooling operations are equivariant w.r.t. gauge transformations, their design principle interferes fundamentally with the desired isometry equivariance.
This is the case since they reduce the spatial resolution of the numerical discretization, such that the output is only exactly equivariant w.r.t. the subgroup of symmetries of the lower resolution grid.
This effect is well known for conventional CNNs~\cite{azulay2018shift}.
Even though some attempts to rectify the situation were made~\cite{zhang2019CNNsShiftInvariant}, the partial loss of translation (or isometry) equivariance to a subgroup is usually accepted as it is.

\paragraph{Unit tests:}
All of the proposed coordinate independent operations are unit tested in order to guarantee their gauge equivariance and isometry equivariance.
The gauge equivariance tests pass for all of the proposed operations as well as for the whole networks described in the following section.
For the convolution, bias summation and nonlinearities, our unit tests confirm isometry equivariance to hold exactly.
As expected, the spatial pooling operations are not exactly equivariant w.r.t. the symmetries of the high resolution grid.%
\footnote{
    Note that this issue is inherent for pooling operations and applies to conventional CNNs as well~\cite{azulay2018shift,zhang2019CNNsShiftInvariant}.
}
However, we confirm their isometry equivariance for that subgroup of isometries which are simultaneously a symmetry of the lower resolution grid.
Our empirical results, which we discuss next, suggest that the inexact isometry equivariance affects the isometry invariance of a full network's classification predictions in most cases only marginally.

\subsubsection{Empirical evaluation}
\label{sec:mobius_evaluation}

We evaluate the coordinate independent operations and their claimed equivariance properties on a simple classification task of MNIST images which are projected on the M\"obius strip.
Different combinations of field types are compared by instantiating similar model architectures for them.
As a baseline we train a non coordinate independent CNN on the M\"obius strip, which is significantly outperformed by the equivariant models.

The M\"obius MNIST dataset is constructed by taking the standard MNIST digits of $28\times28$ pixels and projecting them on the strip by identifying the left and right border with an additional twist.
In compliance with the rotated MNIST dataset, which is a standard benchmark for rotation equivariant Euclidean CNNs, we reduce the training set size to $12000$~samples~\cite{Weiler2018SFCNN,Weiler2019_E2CNN}.
Since MNIST contains single channel grayscale digits, which are invariant under gauge transformations, its samples are identified as scalar fields.
Each sample is therefore represented by an array of shape $(1,28,28)$, corresponding to the magenta rectangle in Fig.~\ref{fig:mobius_conv_numerical}.
Note that the identification of the left and right border does not lead to any discontinuities for the specific case of MNIST digits since their background color is constant black (i.e. zero).
In order to demonstrate the induced isometry equivariance of the coordinate independent CNNs, we construct two versions of this dataset.
The first one contains digits which are all \emph{centered}, that is, which occur at the same position on the strip.
The second dataset puts the digits at random positions around the strip, i.e. \emph{shifts} them by randomly sampled isometries as visualized in Fig.~\ref{fig:mobius_conv_isometries}.
Any isometry equivariant model is expected to generalize their inference from the dataset of centered digits to the isometry shifted dataset, which is confirmed by our experiments.
While M\"obius MNIST clearly is a toy dataset, it exhibits all the theoretical properties which we are interested in and serves as a convenient test case to demonstrate the difference between conventional CNNs and coordinate independent CNNs.

All network architectures are as usual constructed as a series of convolutional layers, followed by a global pooling operation and an invariant, fully connected classifier; see Table~\ref{tab:mobius_model_architectures} for a comparison.
The convolutional parts are built from six convolutional blocks with spatial pooling operations after the second and fourth convolution block.
The convolution blocks are pretty basic and consist of only one convolutional layer followed by a bias summation and a nonlinearity layer.
All intermediate pooling operations utilize pooling windows of $2\times2$ pixels and therefore halve the spatial resolution.
In the case of reflection equivariant models, the last convolutional layer maps to $64$ scalar fields.
Their invariance under gauge transformations guarantees that the subsequent global max-pooling operation produces \emph{both position and gauge invariant} features.
An MLP with a final softmax activation takes those features to produce invariant predictions.
It consists for all models of the same two MLP blocks, which apply a batch-normalization, ELU nonlinearity, dropout with $30\%$ dropping probability and a linear (or affine) layer, whose number of output neurons is listed in Table~\ref{tab:mobius_model_architectures}.
The differences between the different models are therefore restricted to the convolutional part.

\begin{table}
    \centering
    \footnotesize
    \setlength{\tabcolsep}{8pt}
    \begin{tabular}{lccccccc}
        \toprule
        layer              & \multicolumn{6}{c}{output field multiplicities $(m_{\textup{triv}},m_{\textup{sign}},m_{\textup{reg}})$ / channels / neurons}   \\
                           & scalar        & sign-flip     & regular       & irreps          & mixed          & CNN \\[.25ex]
        \midrule
        network input      &  $( 1, 0, 0)$ &  $(1,  0, 0)$ & $(1, 0,  0)$  &  $(1,  0, 0)$   &    $(1, 0, 0)$ & $1$ \\
        conv block         &  $(16, 0, 0)$ &  $(0, 16, 0)$ & $(0, 0,  8)$  &  $(8,  8, 0)$   &    $(4, 4, 2)$ & $\lfloor 16/\sqrt{\alpha}\rfloor$ \\
        conv block         &  $(32, 0, 0)$ &  $(0, 32, 0)$ & $(0, 0, 16)$  & $(16, 16, 0)$   &    $(8, 8, 4)$ & $\lfloor 32/\sqrt{\alpha}\rfloor$ \\
        pooling            & \ditto        & \ditto        & \ditto        & \ditto          & \ditto         & \ditto                            \\
        conv block         &  $(64, 0, 0)$ &  $(0, 64, 0)$ & $(0, 0, 32)$  & $(32, 32, 0)$   & $(16, 16,  8)$ & $\lfloor 64/\sqrt{\alpha}\rfloor$ \\
        conv block         & $(128, 0, 0)$ & $(0, 128, 0)$ & $(0, 0, 64)$  & $(64, 64, 0)$   & $(32, 32, 16)$ & $\lfloor128/\sqrt{\alpha}\rfloor$ \\
        pooling            & \ditto        & \ditto        & \ditto        & \ditto          & \ditto         & \ditto                            \\
        conv block         & $(256, 0, 0)$ & $(0, 256, 0)$ & $(0, 0, 128)$ & $(128, 128, 0)$ & $(64, 64, 32)$ & $\lfloor256/\sqrt{\alpha}\rfloor$ \\
        conv block         &  $(64, 0, 0)$ &  $(64, 0, 0)$ &  $(64, 0, 0)$ & $(64, 0, 0)$    & $(64, 0, 0)$   & $64$                              \\
        \midrule
        global max-pooling & $64$          & $64$          & $64$          & $64$            & $64$           & $64$           \\
        MLP block          & $32$          & $32$          & $32$          & $32$            & $32$           & $32$           \\
        MLP block + softmax \hspace*{-3ex}
                           & $10$          & $10$          & $10$          & $10$            & $10$           & $10$           \\
        \bottomrule
    \end{tabular}
    \vspace*{2.ex}
    \caption[]{\small
        Overview of the compared model architectures.
        All models consist of a convolutional part on the M\"obius strip, followed by a global max-pooling operation and an MLP classifier.
        The five orientation independent CNNs differ in their multiplicities $(m_{\textup{triv}},m_{\textup{sign}},m_{\textup{reg}})$ of field types but agree exactly in their number of channels and approximately in their number of parameters.
        Their inputs, i.e. the MNIST digits, are assumed to be scalar fields.
        All orientation independent models map in their last convolution to $64$ gauge invariant scalar fields.
        A subsequent global pooling operation therefore produces position and coordinate independent features.
        The baseline CNN model comes in two flavors, which differ by their factor of $\sqrt{\alpha}$ in the number of channels.
        A first version assumes $\alpha=1$, and therefore utilizes the same number of channels like the coordinate independent models.
        Due to the inferior parameter efficiency of non-equivariant CNNs, this model uses approximately twice as many parameters.
        For a fair comparison we add a second version with $\alpha=2$ and therefore approximately the same number of parameters like the equivariant models.
    }
    \label{tab:mobius_model_architectures}
\end{table}

The five coordinate independent models that we instantiate differ in the utilized field types:
there are three pure models, denoted by ``scalar'', ``sign-flip'' and ``regular'', which assume only the suggested field type.
Due to their higher dimensionality, the field multiplicities of the regular feature fields are halved in comparison to those of scalar and sign-flip fields.
A fourth model, denoted by ``irrep'', uses a mixture of scalar and sign-flip fields in equal proportions.
Note that the feature fields of this model are linearly equivalent to those of the ``regular'' models since the change of basis from Eq.~\eqref{eq:rho_reg_decomposition} translates between both.
A fifth, ``mixed'' model applies all three field types.
The nonlinearities in use for the different field types are those described in Section~\ref{sec:mobius_nonlin}.
As stated before, all models assume scalar inputs and outputs.

All coordinate independent layers are unit testes and found to be exactly gauge equivariant, implying that the models are overall exactly gauge invariant.
Since they apply two pooling steps, which reduce the spatial resolution by a factor of~$2$ each before the global pooling, the isometry equivariance (invariance) holds only for the subgroup of shifts by multiples of~$4$ pixels.
The theoretically claimed properties therefore hold as expected.

As a baseline, we compare the reflection equivariant models to conventional coordinate dependent CNNs on the M\"obius strip.
In order to respect the topology of the strip, we apply a naive version of the transport padding operation.
Since CNNs are agnostic to field types, this is done by taking the orange strips of two pixels from Fig.~\ref{fig:mobius_conv_numerical} and padding them to the opposite side of the array after applying a reflection but without acting with the unspecified group representation -- formally, this corresponds to transporting the features according to a trivial connection.
Since the non-equivariant operations are less parameter efficient, we consider two different versions:
the first version uses the same number of \emph{channels} like the coordinate independent CNNs, and therefore requires approximately twice as many parameters.
The number of channels of the second version is scaled down by a factor $\sqrt{2}$ such that the number of \emph{parameters} is approximately equivalent to that of the orientation independent models.

All models are trained for $20$ epochs with a batch size of $128$ samples, a weight decay of $10^{-6}$ and using the Adam optimizer~\cite{Kingma2015-yq}.
The initial learning rate of $5\cdot10^{-3}$ is chosen as high as possible without leading to a divergence of the training process.
A fixed learning rate decay schedule reduces the step size every $4$ epochs by a factor of $2$.

\begin{table}
    \centering
    \renewcommand\arraystretch{1.1}
    \setlength{\tabcolsep}{12pt} 
    \small
    \begin{tabular}{lc@{\hspace{6pt}}c@{\hspace{6pt}}crc@{\hspace{8pt}}c}
       \toprule
       model                   & \multicolumn{3}{c}{field types $\rho_i$}    & params        & \multicolumn{2}{c}{test error (\%)}                                              \\
                               & trivial & sign-flip & regular               &               & shifted train digits               & centered train digits                       \\[.25ex]
       \midrule
       CNN (channels)          &         & ---       &                       & $1501$\,k    & $1.97 \scriptstyle\,\pm\, 0.11$    & $           42.99 \scriptstyle\,\pm\, 2.65$ \\
       CNN (params)            &         & ---       &                       &  $832$\,k    & $2.08 \scriptstyle\,\pm\, 0.10$    & $           43.68 \scriptstyle\,\pm\, 2.85$ \\
       gauge CNN (scalar)      & \checkmark  & $\bm\times$ & $\bm\times$     &  $902$\,k    & $1.60 \scriptstyle\,\pm\, 0.10$    & $\phantom{4} 1.60 \scriptstyle\,\pm\, 0.09$ \\
       gauge CNN (sign-flip)   & $\bm\times$ & \checkmark  & $\bm\times$     &  $820$\,k    & $4.27 \scriptstyle\,\pm\, 0.24$    & $\phantom{4} 4.89 \scriptstyle\,\pm\, 0.36$ \\
       gauge CNN (regular)     & $\bm\times$ & $\bm\times$ & \checkmark      &  $752$\,k    & $1.24 \scriptstyle\,\pm\, 0.08$    & $\phantom{4} 1.23 \scriptstyle\,\pm\, 0.07$ \\
       gauge CNN (irreps)      & \checkmark  & \checkmark  & $\bm\times$     &  $752$\,k    & $1.65 \scriptstyle\,\pm\, 0.09$    & $\phantom{4} 1.64 \scriptstyle\,\pm\, 0.12$ \\
       gauge CNN (mixed)       & \checkmark  & \checkmark  & \checkmark      &  $752$\,k    & $1.43 \scriptstyle\,\pm\, 0.09$    & $\phantom{4} 1.42 \scriptstyle\,\pm\, 0.10$ \\[.25ex]
       \bottomrule
    \end{tabular}
    \vspace*{2ex}
    \caption{
        Test errors of the different network architectures, each averaged over $32$~runs.
        The column ``shifted train digits'' reports the performance for a setting where both the training and test samples are placed at random locations on the strip.
        While not being $\RM$-coordinate independent, the conventional CNNs are able to learn to detect the digits as seen from their discontinuous frame field.
        Almost all coordinate independent CNNs achieve significantly better results.
        The inferior performance of the sign-flip model shows that coordinate independent CNNs might not work very well when bad choices of field types or nonlinearities are made.
        The training digits in the column ``centered train digits'' are all placed at the same position on the strip while the test digits remain randomly shifted.
        The coordinate independent CNNs are able to generalize their inference between both situations which affirms their isometry equivariance.
        In contrast, the performance of the conventional CNNs deteriorates, which reflects their missing equivariance under isometries.
    }
    \label{tab:mobius_mnist_results}
\end{table}

Table~\ref{tab:mobius_mnist_results} shows the resulting test errors of all models, each averaged over $32$~runs.
The first setting, reported in the column ``shifted train digits'', uses randomly located digits both in the training and test dataset.
Both versions of the non-equivariant CNN achieve approximately the same test error.
In contrast, most coordinate independent CNNs achieve a significantly better result.
Only the model which is purely based on sign-flip fields performs worse -- this suggests that the utilized combination of sign-flip fields and nonlinearities is not a good choice, despite being coordinate independent.
Bad choices of feature fields and nonlinearities are therefore seen to harm the model performance.
The model achieving the best results is based on regular feature fields.
This observation is in alignment with previous findings, for instance the systematic comparison of field types in~\cite{Weiler2019_E2CNN}.
Our interpretation of this result is that the kernel constraints involving regular feature fields allow for essentially unconstrained kernel channels, with the additional requirement of applying two reflected copies of them -- view this in contrast to the $\Flip$-steerable kernels between irrep fields, which are required to be symmetric \emph{within} one kernel channel.
The model based on scalar fields achieves an intermediate performance between the conventional CNNs and the regular field model.
Both models which use mixed field types have performances lying between those of the field types of the mix.
We want to emphasize again that the regular model and the irrep model contain exactly the same irrep field types but are expressed in a different basis.
Since this change of basis could be interpreted as part of the applied nonlinearities, this result implies that the used nonlinearities have a major impact on the model performance.
Despite being investigated in~\cite{Weiler2019_E2CNN}, the landscape of equivariant nonlinearities is still largely unexplored territory.

The second training setting, reported in the column ``centered train digits'', investigates the capability of the models to generalize over all poses that are related by isometries.
All models are trained on digits which occur at the same location on the strip but test on randomly shifted digits.
As expected, the conventional CNNs' performances degrade significantly in this setting -- this implies that they are indeed not equivariant under the isometries of the M\"obius strip.
In contrast, the performance of most coordinate independent CNNs stays within the standard deviation unchanged.
Despite only being exactly equivariant (invariant) to the subgroup of isometries which shifts by multiples of $4$ pixels, the full isometry invariance of the models therefore seems to hold very well.
While the sign-flip model becomes significantly worse in comparison to the first training setting, it is still approximately isometry equivariant and therefore performs much better than the conventional~CNNs.

In conclusion, the conducted experiments confirm the claimed properties of coordinate independent CNNs and show their superiority over coordinate dependent models.

%% file: chapters/P2_intro.tex

\mypart{Theory of coordinate independent CNNs}
\label{part:bundle_theory}

Part~\ref{part:local_theory} introduced feature fields and network layers in terms of their coordinate expressions relative to some choice of gauge on \emph{local} neighborhoods $U\subseteq M$.
As the existence of \emph{global} gauges is in general topologically obstructed, global coordinate representations of feature fields do in general not exist.
Part~\ref{part:local_theory} addressed this issue by assembling the global content of feature fields from their local coordinate expressions relative to an atlas of gauges that cover~$M$.
A more elegant alternative is to define global feature fields in an abstract, \emph{coordinate free} formalism in terms of fiber bundles.
Bundle trivializations allow to recover the local coordinate expressions of feature fields and network layers.

\etocsettocdepth{2}
\etocsettocstyle{}{} 
\localtableofcontents

~

The following sections develop a global, coordinate free description of the neural networks and feature spaces from Part~\ref{part:local_theory}.
Section~\ref{sec:bundles_fields} introduces fiber bundles, in particular the tangent bundle, $G$-structures and $G$-associated feature vector bundles.
Neural network operations like kernel field transforms and $\GM$-convolutions are defined in Section~\ref{sec:gauge_CNNs_global}.
Section~\ref{sec:isometry_intro} investigates the isometry equivariance of these operations.

%% file: chapters/60_bundles_intro.tex

\section{Associated bundles and coordinate free feature fields}
\label{sec:bundles_fields}

Fields of geometric quantities on manifolds are formalized as sections of fiber bundles (Eq.~\eqref{cd:section_proj_idM}).
Any smooth manifold is naturally endowed with its tangent bundle and frame bundle.
A choice of $G$-structure, which is a $G$-bundle of reference frames, allows to define $G$-associated feature vector bundles.
The feature spaces of our coordinate independent neural networks are spaces of feature fields, i.e. sections of these feature vector bundles.

Fiber bundles in general are reviewed in Section~\ref{sec:fiber_bundles_general}.
Section~\ref{sec:GL_associated_bundles} discusses the tangent bundle $\TM$ and the frame bundle $\FM$.
$G$-structures $\GM$, which are subsets of reference frames which are distinguished by the given geometric structure on the manifold, are introduced in Section~\ref{sec:G_associated_bundles}.
Associated $G$-bundles, including the feature vector bundles $\A$, are constructed from the $G$-structure.
Section~\ref{sec:bundle_trivializations} gives details on the local trivializations (gauges) of $\TM$, $\FM$, $\GM$ and $\A$, which reintroduces coordinates and recovers the formulation in Section~\ref{sec:gauge_cnns_intro_local}.
The mutual transformation of the trivialized feature fields with trivialized tangent vector coefficients and reference frames follows thereby from the coordinate free formulation via associated $G$-bundles.
Section~\ref{sec:bundle_transport} discusses parallel transporters on the associated bundles, in particular how they induce each other.

All concepts presented here are well established in differential geometry and can easily be found in the literature~\cite{schullerGeometricalAnatomy2016,nakahara2003geometry,husemollerFibreBundles1994a,steenrodTopologyFibreBundles,shoshichikobayashiFoundationsDifferentialGeometry1963,marshGaugeTheoriesFiber2016,wendlLectureNotesBundles2008,sternberg1999lectures,piccione2006theory,crainic2013GStructuresExamples}.
Our contribution is to give a comprehensive exposition which bridges between the mathematical theory and its application in geometric deep learning.

%% file: chapters/61_general_bundles.tex

\subsection{A brief introduction to fiber bundles}
\label{sec:fiber_bundles_general}

Intuitively, a \emph{fiber bundle} can be thought of as a space which is constructed by taking a so called \emph{base space}, in our case the manifold $M$, and attaching another space $F$, denoted as \emph{fiber}, to each of its points.
A trivial example would be the direct product \mbox{$M\times F$}.
However, the fibers can in general be connected in a twisted way such that the resulting bundle is topologically different from a product.
\begin{wrapfigure}[13]{r}{0.54\textwidth}
    \vspace*{-1.4ex}
    \hfill
    \includegraphics[width=.98\linewidth]{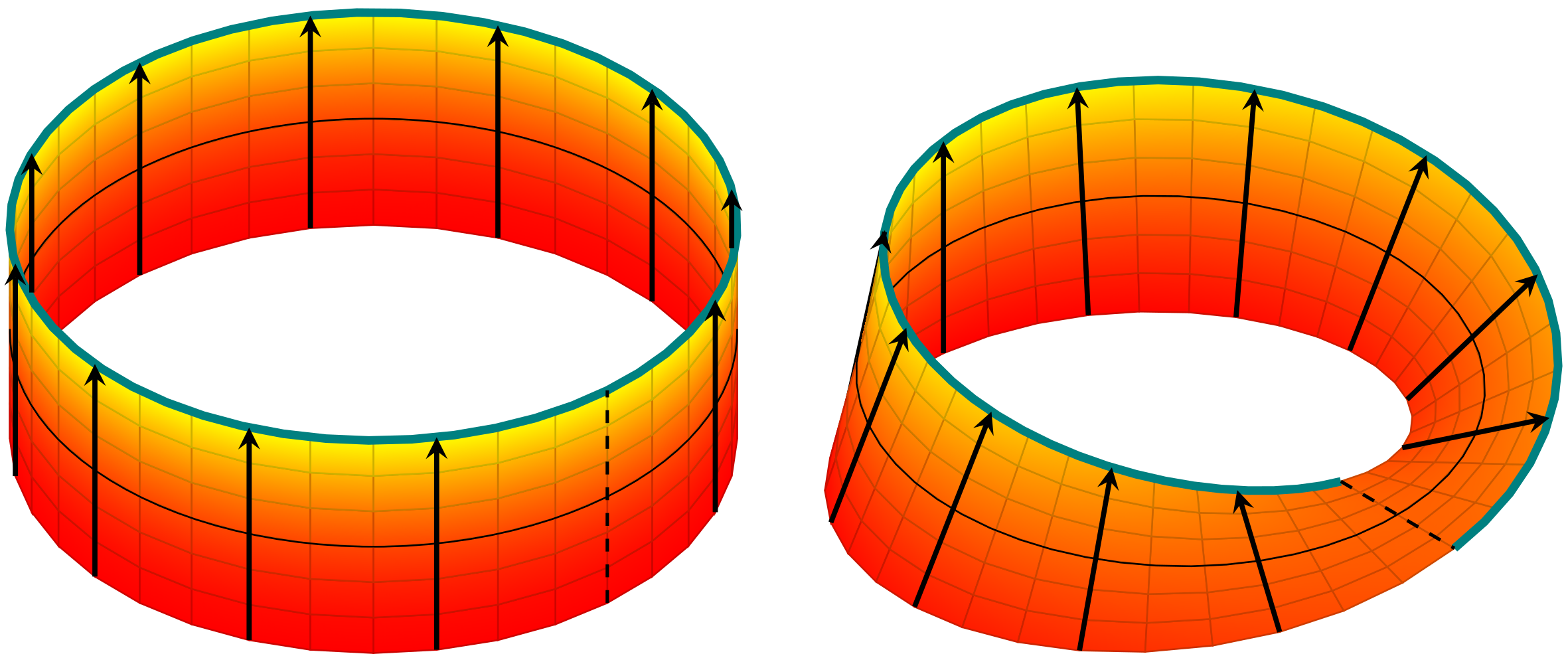}%
    \vspace*{.2ex}
    \caption{\small
        A cylinder and a M{\"o}bius strip.
        Both bundles share the circle $S^1$ as base space and line segments $[-1,1]$ as fibers, however, their topological structure differs by a twist in the fibers.
        {\\
            \color{gray}
            \scriptsize
            (Figure based on Jake's code from
            \href{https://tex.stackexchange.com/questions/118563/moebius-strip-using-tikz}{\underline{tex.stackexchange.com}}.)
        }
        }
    \label{fig:moebius}
\end{wrapfigure}%
For instance, let the base space be the circle $M=S^1$ and let the fiber be the line segment $F=[-1,1]$.
Their direct product $S^1\times[-1,1]$ then forms a cylinder; see Fig.~\ref{fig:moebius} (left).
In contrast, if the fibers are attached such that they are twisted "upside down" after one revolution around the circle, one obtains the M{\"o}bius strip, a non-trivial fiber bundle which is topologically different from the cylinder; see Fig.~\ref{fig:moebius}~(right).%
\footnote{
    To prevent confusion, we emphasize that this example considers the M{\"o}bius strip as a fiber bundle with base space (manifold) $M=S^1$.
    In contrast, all previous figures that contained the M{\"o}bius strip considered it as the base space (manifold) $M$ to convolve over.
}%
\footnote{
    We furthermore need to mention that the arrows shown in the figure are just meant to emphasize the twist in the M{\"o}bius strip.
    They do \emph{not} imply a gluing direction as in \emph{gluing diagrams}.
}
Note that the M{\"o}bius strip \emph{locally} looks like a direct product $U\times F$ of a line element $U\subsetneq S^1$ with the fiber~$F$.
As discussed below, fiber bundles can by definition always be locally trivialized to direct products.

We are interested in fiber bundles since they allow for a global description of fields on manifolds.
For instance, a wind field on the globe $M=S^2$ is a tangent vector field which assigns a tangent vector in $\TpM$ to each point $p$ of~$M$.
The corresponding fiber bundle is the tangent bundle $\TM$ which connects all the tangent spaces together and is therefore identified as a fiber bundle with base space $M=S^2$ and fiber $\R^d\cong \TpM$.
Similar to the fibers of a M{\"o}bius strip, the tangent spaces of a curved manifold are in general not connected in a canonical way but are inherently twisted relative to each other.
The tangent bundle is therefore in general topologically distinct from a product, that is, $\TM \ncong M\times\R^d$.
In order to define $c$-dimensional feature vector fields, we will later consider bundles with base space $M$ and feature vector spaces $\R^c$ as fibers.

\paragraph{Fiber bundles in general:}
Formally, a fiber bundle is a structure $(E,M,\pi,F)$ consisting of topological spaces $E$ (total space), $M$ (base space) and $F$ (typical fiber) and a continuous surjective projection map $\pi:E\to M$.
A fiber bundle is \emph{locally trivializable}, which means that for each point $p\in M$ there exists a local neighborhood $U\subseteq M$ of~$p$, restricted to which the bundle looks like a direct product $U\times F$.
The local triviality is formalized by homeomorphisms%
\footnote{
    A \emph{homeomorphism} is a topological isomorphism, i.e. a continuous, invertible map between topological spaces with continuous inverse.
}
$\Psi:\pi^{-1}(U)\to U\times F$ satisfying the commutative diagram below
\\[-2ex]
\begin{equation}\label{cd:trivialization_general_intro}
\begin{tikzcd}[row sep=3em, column sep=4em]
    \:E\: \supseteq
    &[-4.2em]
    \pi^{-1}(U) \arrow[d, swap, "\pi"] \arrow[r, "\Psi"]
    & U\times F \arrow[ld, "\proj_1"] \\
    M\, \supseteq
    & U
\end{tikzcd}
\quad ,
\end{equation}
that is,
\begin{align}\label{eq:local_triviality_proj}
   \pi\ =\  \proj_1 \circ \mkern2mu \Psi \,,
\end{align}
where $\proj_1: U\times F\to U$ denotes the natural projection on the first factor.
A bundle which is globally homeomorphic to the product $M\times F$ is called \emph{trivial}.
Bundles are often shortly written $E\!\xrightarrow{\pi}\!M$ or just $E$ with the typical fiber and base space left implicit.
Since we are considering smooth frame fields, we assume $E$, $M$ and $F$ to be smooth manifolds and $\pi$ and $\Psi$ to be smooth maps (diffeomorphisms).

The local triviality of $E\!\xrightarrow{\pi}\!M$ implies that the preimage $E_p:=\pi^{-1}(p)$ of any point $p\in M$, called the \emph{fiber over}~$p$, is diffeomorphic to the typical fiber $F$.
As in Section~\ref{sec:gauge_cnns_intro_local}, we denote the diffeomorphisms which identify the fibers over different points with the typical fiber by $\psi_p:E_p\to F$.
The local trivializations are then in terms of these diffeomorphisms given by
\begin{align}\label{eq:Psi_via_psi}
    \Psi:\pi^{-1}(U)\to U\times F,\ \ \ e\mapsto \big(\pi(e),\: \psi_{\pi(e)}(e)\big) \,.
\end{align}
If the typical fiber $F$ and the fibers $E_p$ over $p$ carry additional structure, the diffeomorphisms $\psi_p: E_p\to F$ are required to respect this structure, i.e. to be isomorphisms.%
\footnote{
    Alternatively, assume that $F$ carries structure which is respected by the transition functions $\psi_p^B \circ (\psi_p^A)^{-1} = g^{BA}(p) \in \Aut(F)$ (see the next paragraph).
    Then the trivializations $\psi_p^X: E_p \to F$ consistently \emph{induce} the structure of~$F$ on~$E_p$ and are automatically isomorphisms.
}
For instance, if $F$ and $E_p$ carry a vector space structure, then $\psi_p$ is required to be linear.

In general the specific choice of local trivializations (or diffeomorphisms) over $U$ is not canonically specified by the bundle.
One therefore has to consider different choices (gauges) and \emph{transition functions} (gauge transformations) between them.
To make this precise, consider two overlapping trivializing neighborhoods $U^A$ and $U^B$ with local trivializations $\Psi^A$ and $\Psi^B$.
From Eq.~\eqref{eq:Psi_via_psi} it follows that the transition between both local trivializations is on $U^{AB}:=U^A\cap U^B\neq\varnothing$ given by
\begin{align}\label{eq:transition_function_general_bdl}
    \Psi^B\!\circ\!\big(\Psi^A\big)^{-1}\!:\ U^{AB}\times F \to U^{AB}\times F,
    \quad (p,\mathscr{f}) \mapsto \left(p,\, \pig[\psi_p^B \!\circ\! \big(\psi_p^A\big)^{-1}\, \pig]\mkern-1.5mu(\mathscr{f}) \right)
    =: \left(p,\: g_p^{BA} \,\btr\mkern1.5mu \mathscr{f}\right)
\end{align}
where we implicitly defined the smooth \emph{transition functions}%
\footnote{
    The automorphism group $\operatorname{Aut}(F)$ of a space~$F$ consists of all invertible, structure preserving maps (isomorphisms) from~$F$ to itself.
    For instance, if $F=\R^n$ is a vector space, the automorphism group is the general linear group $\GL{n}$, which consists of all invertible $n\!\times\!n$ matrices.
}
\begin{align}\label{eq:transition_functions_psi}
    g^{BA}: U^{AB} \to \Aut(F),\ \ \ p \mapsto g_p^{BA} := \psi_p^B \circ \big(\psi_p^A\big)^{-1}
\end{align}
and their left action
\begin{align}\label{eq:gauge_trafo_leftaction}
    \blacktriangleright\,:\, \Aut(F) \times F\to F, \quad
    \left(g_p^{BA}, \mathscr{f}\right) \;\mapsto\; g_p^{BA}\,\btr\mkern1.5mu \mathscr{f} \;:=\; \left[\psi_p^B\circ\big(\psi_p^A\big)^{-1}\right]\!(\mathscr{f}).
\end{align}
on the typical fiber $F$; cf. Eqs.~\eqref{eq:gauge_trafo_local_def_21} and~\eqref{eq:transition_fct_local_def_21}.
To see that the first factor in Eq.~\eqref{eq:transition_function_general_bdl} is indeed given by the identity, note that, for any $p\in U^{AB}$ and any $\mathscr{f}\in F$, the repeated application of Eq.~\eqref{eq:local_triviality_proj} implies
$
    \pig[ \proj_1 \circ \Psi^B \circ \big(\Psi^A\big)^{-1} \mkern1mu\pig] (p,\mathscr{f})
    \ =\ \pig[ \pi \circ \big(\Psi^A\big)^{-1}  \mkern1mu\pig] (p,\mathscr{f})
    \ =\ \proj_1 (p,\mathscr{f})
    \ =\ p \,.
$
The transition between different trivializations is visualized by the following (commuting) extension of the commutative diagram in Eq.~\eqref{cd:trivialization_general_intro}:
\begin{equation}\label{cd:trivialization_general_trafo}
\begin{tikzcd}[row sep=3.5em, column sep=5.em]
    U^{AB}\!\times\!F
        \arrow[rd, "\proj_1"']
        \arrow[rr, rounded corners, to path={ 
                    -- ([yshift=4ex]\tikztostart.north) 
                    --node[above, pos=.5]{\small$\id \times g^{BA}\btr$} ([yshift=4ex]\tikztotarget.north) 
                    -- (\tikztotarget.north)
                    }]
    & \pi^{-1}(U^{AB})
        \arrow[d, swap, "\pi"]
        \arrow[l, "\Psi^A"']
        \arrow[r, "\Psi^B"]
    & U^{AB}\!\times\!F
        \arrow[ld, "\proj_1"]
    \\
    & U^{AB}
\end{tikzcd}
\end{equation}
Restricted to a single point $p\in U^{AB}$, and for the specific case of the tangent bundle (defined below), this diagram corresponds to that in Eq.~\eqref{eq:commutative_diagram_TpM} and its graphical version in Fig~\ref{fig:gauge_trafos}.

\begin{figure*}
    \centering
    \begin{subfigure}[b]{0.46\textwidth}
        \includegraphics[width=\textwidth]{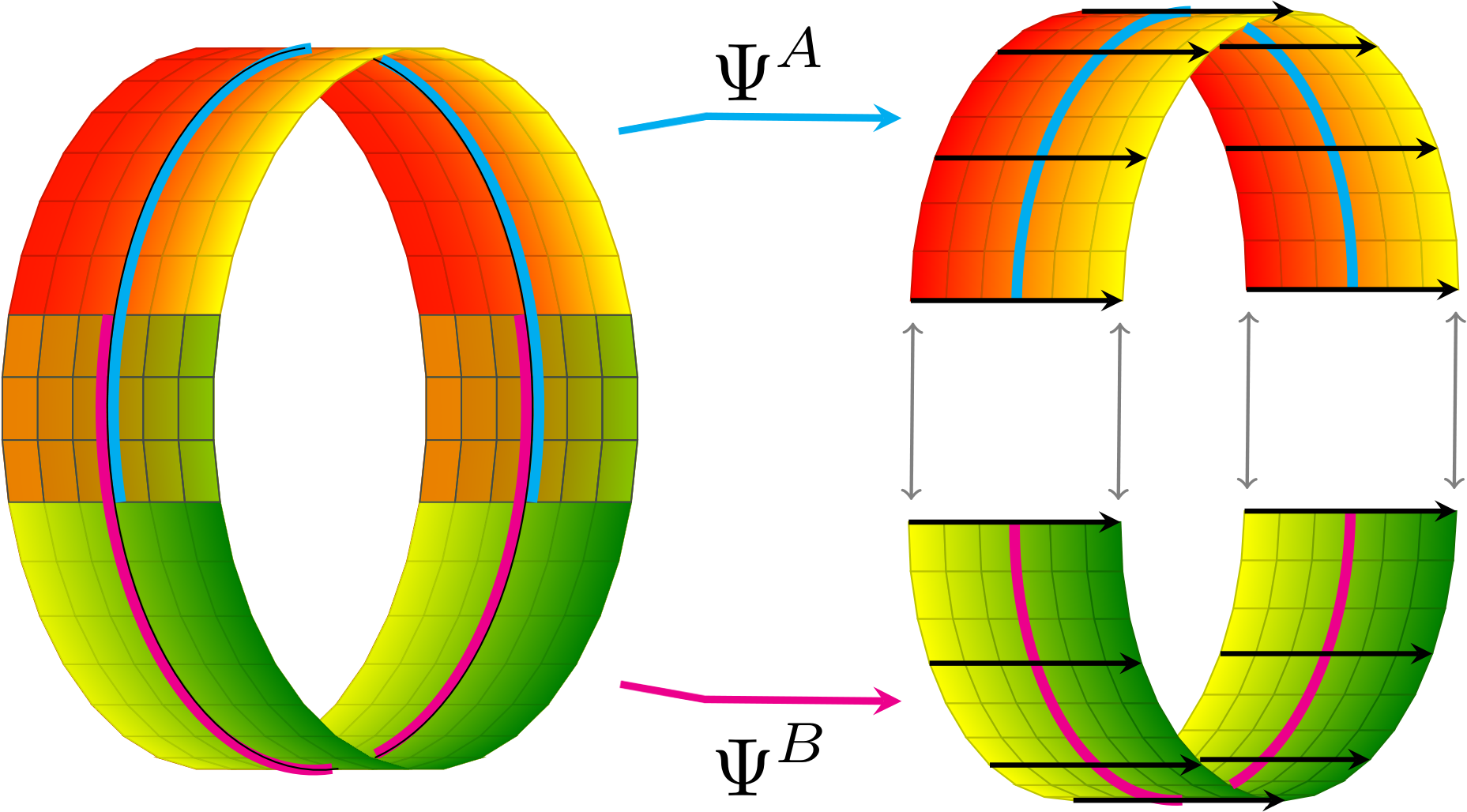}
    \end{subfigure}
    \hfill
    \begin{subfigure}[b]{0.46\textwidth}
        \centering
        \includegraphics[width=\textwidth]{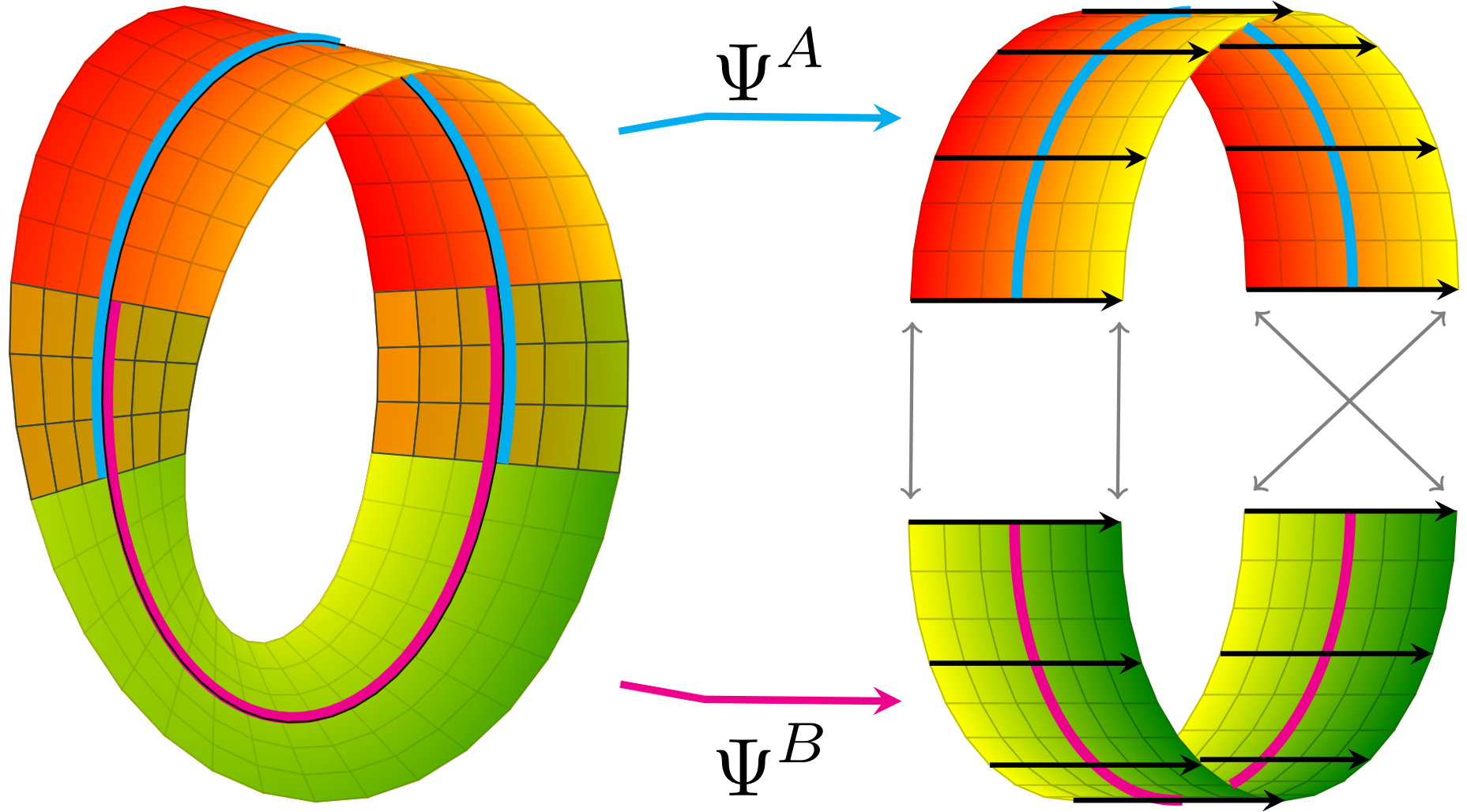}
    \end{subfigure}
    \vspace{2.ex}
    \caption{\small
        Description of the cylinder and the M{\"o}bius strip in terms of $G$-atlases consisting of two local trivializations each.
        \textit{Left:}~Since the cylinder is a trivial bundle, all transition functions can be chosen to be identity maps such that the structure group is reduced to the trivial group $G=\{e\}$.
        Differing from the visualized situation, it is possible to choose a single, global trivialization.
        \textit{Right:}~The topology of the M{\"o}bius strip forces the transition functions at one of the overlaps to glue the fibers together in an inverted way.
        The structure group can therefore not be reduced further than the group $G=\Flip$ which models the reflection of fibers.
        Global trivializations of the M{\"o}bius strip do therefore not exists.
        Note that the arrows on the M{\"o}bius strip should not be confused with the arrows in gluing diagrams, that is, the twist glues the vectors at one of the cuts in opposite direction.
    }
    \label{fig:trivializations_moebius}
\end{figure*}

By definition, the transition functions in Eq.~\eqref{eq:transition_functions_psi} satisfy the following three conditions:%
\footnote{%
    Conditions $i)$ and $ii)$ follow from the cocycle condition $iii)$ but are often stated explicitly.
}
\begin{alignat}{3}
       i) \qquad&& g_p^{AA}         \;&=\ e                     \quad&&\forall p\in U^A \label{eq:transition_condition_1}\\
      ii) \qquad&& g_p^{BA}         \;&=\big(g_p^{AB}\big)^{-1} \quad&&\forall p\in U^A\cap U^B \label{eq:transition_condition_2}\\
     iii) \qquad&& g_p^{CB} g_p^{BA}\;&=\ g_p^{CA}              \quad&&\forall p\in U^A\cap U^B\cap U^C \qquad \text{(cocycle condition)} \label{eq:transition_condition_3}
\end{alignat}
By the \emph{fiber bundle construction theorem}, any fiber bundle can be fully specified globally in terms of an \emph{atlas}
$\mathscr{A}\ =\ \big\{\big( U^X, \Psi^X \big) \,\big|\, X\in\mathfrak{X} \big\}$
of local trivializations $\big(U^X,\Psi^X\big)$ which cover $M$ and whose transition functions satisfy Eqs.~\eqref{eq:transition_condition_1}, \eqref{eq:transition_condition_2} and~\eqref{eq:transition_condition_3} (here $\mathfrak{X}$ denotes some index set).
The individual trivializations can be thought of as being ``glued together'' by the transition maps, which is visualized in Fig.~\ref{fig:trivializations_moebius}.
Note that this is similar to the global description of a manifold in terms of an atlas of local charts.

\paragraph{Vector bundles:}
Several more specific notions of fiber bundles, carrying additional mathematical structure, exist.
An important example are \emph{vector bundles}, which, as the name suggests, are bundles consisting of vector spaces attached to a manifold.
Formally, a (real) vector bundle of rank~$k$ is a bundle $(E,M,\pi,\R^k)$ with typical fiber~$\R^k$ and fibers $E_p \cong \R^k$ over $p$ such that the local trivializations are fiber wise vector space isomorphisms (linear maps).
The transition functions $\psi^B_p \circ \big(\psi^A_p\big)^{-1} \in \Aut(\R^k) = \GL{k}$ then take values in the general linear group.

Alternatively, given the fiber $\R^k$ and an atlas of local trivializations whose transition functions take values in $\Aut(\R^k) = \GL{k}$, a vector space structure of $E_p$ is induced by setting
\begin{align}
    \alpha v + \beta w \ :=\ \big(\psi_p^A \big)^{-1} \big( \alpha \psi_p^A(v) + \beta \psi_p^A(w) \big) \qquad \forall\ v,w\in E_p,\ \ \alpha,\beta\in\R
\end{align}
for an arbitrary gauge $\psi_p^A: E_p \to \R^k$ from the $\GL{k}$-atlas.
That the vector space structure is consistently defined is clear as
\begin{align}
       &\big( \psi_p^B \big)^{-1} \pig( \alpha \psi_p^B(v) + \beta \psi_p^B(w) \pig) \notag \\
    =\ &\big( \psi_p^A \big)^{-1} \pig( \big(g^{BA}_p)^{-1} \big( \alpha g^{BA}_p \psi_p^A(v) + \beta g^{BA}_p \psi_p^A(w) \big) \pig) \notag \\
    =\ &\big( \psi_p^A \big)^{-1} \pig( \alpha \psi_p^A(v) + \beta \psi_p^A(w) \pig)
\end{align}
yields the same result.
Note that the last step required the linearity of $g_p^{BA} \in \GL{d}$.
The gauges $\psi_p^A$ or $\psi_p^B$ are then automatically vector space isomorphisms.

The most relevant examples for us are the tangent bundle and feature vector bundles, which are introduced in the following sections.

\paragraph{\textit{G}-bundles}
Depending on the topology of the bundle, it might be possible to define an atlas of local trivializations
$\mathscr{A}^G\ =\ \big\{\big( U^X, \Psi^X \big) \,\big|\, X\in\mathfrak{X} \big\}$
whose \emph{transition functions are restricted to a subgroup} ${G \leq \Aut(F)}$, that is, they satisfy
\begin{align}
    g_p^{BA} \in G\quad\ \ \textup{for all}\ \ A,B\in\mathfrak{X}\ \ \textup{and all}\ \ p\in U^A\cap U^B \,.
\end{align}
Any such atlas is called $G$-\emph{atlas} and $G$ is denoted as \emph{structure group} of the bundle.
Two different $G$-atlases are equivalent (or compatible), if their union is again a $G$-atlas.
A bundle equipped with an equivalence class of $G$-atlases is known as a $G$-bundle.%
\footnote{
    The equivalence class ensures thereby that no single of the equivalent $G$-atlases is preferred.
    Equivalently, one could take the \emph{maximal} $G$-atlas, defined as the unique $G$-atlas in which any other compatible $G$-atlas is contained.
    Note that an equivalence class of $G$-atlases is uniquely implied by a single given $G$-atlas.
}

The topology of a bundle determines how far its structure group can be reduced.
For instance, the cylinder in Fig.~\ref{fig:trivializations_moebius} (or any other trivial bundle) can be described by an $\{e\}$-atlas, consisting of local trivializations with identity transition functions only.
This corresponds to a reduction to a trivial structure group $G=\{e\}$.
In contrast, the twisted topology of the M{\"o}bius strip requires any $G$-atlas to contain transition functions which glue the fibers together in an inverted orientation; see Fig~\ref{fig:trivializations_moebius} (right).
The structure group of the M{\"o}bius strip can therefore not be restricted further than the group $G=\Flip$ which models the reflection of fibers.
On Riemannian manifolds the structure group of the tangent bundle $\TM$, and thus the associated feature vector bundles, can in general not be reduced further than to an orthogonal structure group $\O{d}$ which motivated this work on coordinate independent CNNs in the first place.

\paragraph{Associated \textit{G}-bundles:}
Two $G$-bundles are said to be \emph{associated} to each other if they share the same base space, structure group and, most importantly, \emph{same transition functions}.
Associated bundles $(E,M,\pi,F)$ and $(\widetilde{E},M,\widetilde{\pi},\widetilde{F})$ with structure group $G$ might differ in their typical fibers~$F$ and $\widetilde{F}$ and therefore also in their left actions $\blacktriangleright: G \times F\to F$ and $\widetilde{\blacktriangleright}: G \times \widetilde{F}\to \widetilde{F}$ of~$G$ on the respective fiber.
Given two $G$-atlases
$\big\{\big( U^X,            \Psi^X  \big) \,\big|\, X\in\mathfrak{X} \big\}$ and
$\big\{\big( U^X, \widetilde{\Psi}^X \big) \,\big|\, X\in\mathfrak{X} \big\}$
of the bundles over the same open cover of $M$, the requirement for the equivalence of the transition functions (up to the different left actions) means:
\begin{align}
    \Psi^B \circ \big(\Psi^A \big)^{-1}\ =\ \big(\id \times g^{BA}\blacktriangleright \big)
    \qquad \Longleftrightarrow \qquad
    \widetilde{\Psi}^B \circ \big(\widetilde{\Psi}^A \big)^{-1}\ =\ \big(\id \times g^{BA}\, \widetilde{\blacktriangleright} \big)
\end{align}
Intuitively, the typical fibers~$F$ and~$\widetilde{F}$ of~$E$ and~$\widetilde{E}$ are ``glued together'' in the same way over~$M$.

An important example of bundles which are $\GL{d}$-associated to each other are the tangent bundle $\TM$, the cotangent bundle $\TsM$, any other tensor bundle $\TrsM$ and the tangent frame bundle~$\FM$, (the first and the latter are introduced in Section~\ref{sec:GL_associated_bundles}).
The associatedness of these bundles is reflected in that their components relative to chosen bases transform according to the same gauge transformation (e.g. Jacobian $g^{BA}_{\mu\nu} = \frac{\partial x^B_\mu}{\partial x^A_\nu}$, see Appendix~\ref{apx:coordinate_bases}).
The different actions of a gauge transformation on the respective fibers is in this example denoted as being a contravariant transformation ($\TM$), covariant transformation ($\TsM$), $r$-times contra- and $s$-times covariant transformation ($\TrsM$) and, again, covariant transformation ($\FM$), respectively.
We will later on introduce the $G$-structure $\GM$, the tangent bundle $\TM$ and the feature vector bundles $\A$ as associated $G$-bundles.
The associatedness does in this case come from the fact that changes of reference frames in $\GM$ lead to a simultaneous transformations of the tangent vector coefficients and feature vector coefficients.

We want to mention that any associated bundles are additionally associated to a uniquely specified principal $G$-bundle (defined in the next paragraph).
In turn, any associated bundle can be constructed from the respective associated principle bundle -- we will make heavy use of this construction to define feature vector bundles in Section~\ref{sec:G_associated_bundles}.

\paragraph{Principal \textit{G}-bundles:}
A fiber bundle $(P,M,\pi,G)$ is called a (smooth) \emph{principal G-bundle} $(P,M,\pi,G,\lhd)$ if 1) its typical fiber coincides with its structure group $G$ and 2) it is endowed with a smooth \emph{right $G$-action}
\begin{align}
    \lhd: P \times G \to G,\ \ (\mathscr{p},g) \mapsto \mathscr{p}\lhd g
\end{align}
which preserves the fibers, that is,
\begin{align}
    \pi(\mathscr{p}\lhd g)\ =\ \pi(\mathscr{p})\quad \forall\ \mathscr{p}\in P,\ g\in G
\end{align}
and acts \emph{transitively} and \emph{freely} on them.%
\footnote{%
    A (right) group action $\phi:X\times G\to X,\ (x,g)\mapsto x.g$ is called \emph{transitive} if any point of $X$ can be mapped to any other point i.e. if for each $x,y\in X$ there exists a $g\in G$ such that $y=x.g$.
    It is called (fixed point) \emph{free} if for any $x\in X$ the equation $x=x.g$ implies that $g=e$, that is, if only the action of the identity element leaves $p$ invariant.
    Note that the same statements can be made for left actions.
}
The last two conditions (transitivity and freedom) together require that the fibers of a principal $G$-bundle are $G$-torsors (or principal homogeneous $G$-spaces), which intuitively means that they ``look like~$G$'' but come without any specified origin or identity element.%
\footnote{
    Formally, a (right) $G$-torsor $P$ satisfies $P\times G \cong P\times P$ where the isomorphism is given by $(p,g) \mapsto (p,p.g)$.
    This condition implies that there is a \emph{unique} group element connecting \emph{any} two points in the torsor.
}
The local trivializations $\Psi: \pi^{-1}(U) \to U \times G$ are required to respect the right $G$-action, that is, to be right $G$-equivariant
\begin{align}\label{eq:right_G_equiv_principal_bdl_general}
    \Psi(\mathscr{p} \lhd g)\ =\ \Psi(\mathscr{p}) (\id \times \cdot\mkern1mu g)
    \quad \textup{or, equivalently,}\quad
    \psi_{\pi(\mathscr{p})} (\mathscr{p}\lhd g)\ =\ \psi_{\pi(\mathscr{p})} (\mathscr{p}) \cdot\mkern0mu g
    \qquad \forall\ \mathscr{p}\in P,\ g\in G \,,
\end{align}
where $\cdot g$ denotes the canonical right multiplication with group elements on the typical fiber~$G$.
This extends the diagram in Eq.~\eqref{cd:trivialization_general_intro} to the diagram
\begin{equation}\label{cd:trivialization_principal_intro}
\begin{tikzcd}[row sep=3.5em, column sep=4.5em]
      \pi^{-1}(U)
            \arrow[r, "\Psi"]
    & U\times G
    \\
      \pi^{-1}(U)
            \arrow[d, "\pi\,"']
            \arrow[r, "\Psi"]
            \arrow[u, "\lhd g\ "]
    & U\times G
            \arrow[ld, "\proj_1"]
            \arrow[u, "\ (\id \times \cdot g)"']
    \\
    U
\end{tikzcd}
\quad,
\end{equation}
which is required to commute for any $g\in G$.

Principal $G$-bundles are of great relevance for the study of general $G$-bundles.
In particular, any $G$-bundle $(E,M,\pi_E,F)$ is associated to some (unique) principal $G$-bundle $(P,M,\pi_P,G,\lhd)$ over~$M$ and any associated $G$-bundle can be constructed from $P$.
In the following sections we will present the frame bundle~$\FM$ and $G$-structures~$\GM$ as specific instances of principal bundles, which will make the claims made here less abstract and uncover some consequences of them.

\paragraph{Sections and fields:}
Smooth $F$-valued fields over $M$ are formalized as smooth \emph{sections} $\sigma$ of a bundle $E\!\xrightarrow{\pi}\!M$ with fiber $F$.
A smooth section is thereby defined as a smooth map $\sigma:M\to E$ that assigns to each point $p$ of the base space an element in the fiber $E_p$ over $p$, that is, it satisfies $\pi\circ\sigma=\id_M$, which the following commutative diagram visualizes:
\begin{equation}\label{cd:section_proj_idM}
\begin{tikzcd}[row sep=3em, column sep=4.5em]
      M \arrow[r, "\sigma"]
        \arrow[rr, rounded corners, to path={ 
                  -- ([yshift=-3.ex]\tikztostart.south) 
                  --node[below, pos=.5]{\small$\id_M$} ([yshift=-3.ex]\tikztotarget.south) 
                  -- (\tikztotarget.south)
                  }]
    & E \arrow[r, "\pi"]
    & M
\end{tikzcd}
\end{equation}
An important example are tangent vector fields, which are modeled as sections $v: M\to \TM$ that assign a tangent vector $v(p)\in \TpM$ to each point $p$ in $M$.
Note that the projection map is, by its nature, non-invertible, such that $\sigma\circ\pi\neq\id_E$.
The following diagram does therefore in general \emph{not} commute:
\begin{equation}\label{cd:section_proj_noncommutative}
\begin{tikzcd}[row sep=3em, column sep=4.5em,
               execute at end picture={
                    \node [] at (-.04, -.46) {$\noncommutative$};
                    }]
      E \arrow[r, "\pi"]
        \arrow[rr, rounded corners, to path={ 
                  -- ([yshift=-3.5ex]\tikztostart.south) 
                  --node[below, pos=.5]{\small$\id_E$} ([yshift=-3.5ex]\tikztotarget.south) 
                  -- (\tikztotarget.south)
                  }]
    & M \arrow[r, "\sigma"]
    & E
\end{tikzcd}
\end{equation}
In cases below where a diagram does not commute, which is mostly the case for sections, we emphasize this visually by adding the symbol $\raisebox{-2pt}{\noncommutative}$.
Smooth sections do not necessarily exist globally but can always be defined on trivializing neighborhoods $U\subseteq M$.
Via a local trivialization, a local section can be identified with a function $s:U\to F$ by setting $s(p) = \psi_p(\sigma(p))$ for $p\in U$.
We denote the space of global sections by $\Gamma(E)$ while the space of local sections is written $\Gamma(U,E)$.

\paragraph{Bundle morphisms:}

The morphisms (maps) in the category of fiber bundles are called \emph{bundle morphisms} or bundle maps.
They differ from mere diffeomorphisms between the total spaces in that they are additionally required to respect the bundle structure, i.e. to \emph{map fibers to fibers}.
In general, a smooth bundle map between two smooth fiber bundles $(E,M,\pi,F)$ and $(\widetilde{E},\widetilde{M},\widetilde{\pi},\widetilde{F})$ is a smooth map $\phi: E \to \widetilde{E}$ between the total spaces such that there exists a second smooth map $\widehat{\phi}: M \to \widetilde{M}$ between the base spaces which satisfies $\widetilde{\pi} \circ \phi = \widehat{\phi} \circ \pi$, that is, the following diagram is required to commute:
\begin{equation}\label{cd:general_bundle_morphism}
\begin{tikzcd}[row sep=3.em, column sep=4.5em]
    E
            \arrow[r, "\phi"]
            \arrow[d, "\pi\,"']
    & \widetilde{E}
            \arrow[d, "\,\widetilde{\pi}"]
    \\
    M
            \arrow[r, "\widehat{\phi}"]
    & \widetilde{M}
\end{tikzcd}
\end{equation}
The map on the base space ensures that the bundle morphism maps fibers at $p\in M$ to fibers at $\widehat{\phi}(p) \in \widetilde{M}$ instead of ``shearing them apart''.
Obvious generalizations to bundle \emph{isomorphisms} and bundle \emph{automorphism} exist.
For instance, bundle isomorphisms require $\phi$ and $\widetilde{\phi}$ to be invertible, i.e. diffeomorphisms (and to respect further structure if defined).

The specific kind of bundle map under consideration can be narrowed down further by demanding additional requirements.
A \emph{bundle $M$-morphism} between two bundles $(E,M,\pi,F)$ and $(\widetilde{E},\widetilde{M},\widetilde{\pi},\widetilde{F})$ \emph{over the same base space $M$} is required to map fibers $E_p$ over any $p \in M$ to fibers $\widetilde{E}_p$ over the same point $p$, that is, $\widehat{\phi} = \id_M$.
In terms of a commutative diagram this reads:
\begin{equation}\label{cd:bundle_M_morphism}
\begin{tikzcd}[row sep=3.em, column sep=2em]
    E
            \arrow[rr, "\phi"]
            \arrow[dr, "\pi\,"']
    & & \widetilde{E}
            \arrow[dl, "\,\widetilde{\pi}"]
    \\
    & M
\end{tikzcd}
\end{equation}
From this perspective, we identify the bundle trivialization in Diagram~\eqref{cd:trivialization_general_intro} as a bundle $U$-morphism~$\Psi$ between the trivial bundles $\pi^{-1}(U)$ and $U\times F$ over~$U$.

If the fibers carry additional structure, this structure is typically required to be preserved by the bundle map.
For instance, \emph{vector bundle morphisms} $\phi$ between $(E,M,\pi,\R^k)$ and $(\widetilde{E},M,\widetilde{\pi},\R^{\widetilde{k}})$ are demanded to respect the vector space structure on the fibers, and therefore to restrict to \emph{fiber wise linear maps $\phi|_p: E_p \to \widetilde{E}_{\phi(p)}$}.
Similarly, \emph{principal bundle morphisms} are required to respect the property of the fibers to be right $G$-torsors, i.e. to be right $G$-equivariant.
Given two principal bundles $(P,M,\pi,G,\lhd)$ and $(\widetilde{P},\widetilde{M},\widetilde{\pi},\widetilde{G},\widetilde{\lhd})$ and some group homomorphism $\theta:G \to \widetilde{G}$, a principal bundle morphism is required to make the following diagram commute for any $g\in G$:
\begin{equation}\label{cd:principal_bundle_morphism}
\begin{tikzcd}[row sep=3.em, column sep=4.5em]
    P
            \arrow[r, "\phi"]
    & \widetilde{P}
    \\
    P
            \arrow[d, "\pi\,"']
            \arrow[r, "\phi"]
            \arrow[u, "\lhd g\ "]
    & \widetilde{P}
            \arrow[d, "\,\widetilde{\pi}"]
            \arrow[u, "\ \widetilde{\lhd}\, \theta(g)"']
    \\
    M
            \arrow[r, "\widehat{\phi}"]
    & \widetilde{M}
\end{tikzcd}
\end{equation}
The local trivialization of principal bundles in Diagram~\eqref{cd:trivialization_principal_intro} is thus seen as a principal bundle $U$-morphism $\Psi$ between $\pi^{-1}(U)$ and $U \times G$ where the group homomorphism $\theta: G\to G,\ g\mapsto g$ is given by the identity on $G$.

Bundle morphisms are of particular importance in Section~\ref{sec:isometry_intro}, where they describe the transformation of bundles and feature fields under the action of isometries.
Coordinate independent CNNs are proven to be equivariant w.r.t. these actions on bundles and their sections.

For more background on fiber bundles in general we refer to~\cite{schullerGeometricalAnatomy2016,nakahara2003geometry,husemollerFibreBundles1994a,steenrodTopologyFibreBundles,shoshichikobayashiFoundationsDifferentialGeometry1963,marshGaugeTheoriesFiber2016,wendlLectureNotesBundles2008}.

%% file: chapters/62_TM_FM_associated_GL.tex

\subsection{The tangent bundle \textit{TM} and frame bundle \textit{FM}}
\label{sec:GL_associated_bundles}

Any differentiable (and thus any Riemannian) manifold $M$ is canonically equipped with its tangent bundle $\TM$ and the (general) frame bundle $\FM$, consisting of all local reference frames of the tangent spaces.
The two bundles are naturally associated to each other, with their structure group a-priori given by $\Aut(\R^d) = \GL{d}$.
This fact will be emphasized by ``reconstructing'' $\TM$ from $\FM$ via an associated bundle construction which will later allow us to define associated feature vector bundles.
To clearly separate the concepts introduced and assumptions made, we will describe $\TM$ and $\FM$ here as $\GL{d}$-bundles.
The following Section~\ref{sec:G_associated_bundles} will additionally assume a $G$-structure imposed on $\TM$ and $\FM$, which will establish them as $G$-bundles.
While bundles are locally trivializable by definition, we will take the specific trivializations for now as granted and postpone their exact definition to Section~\ref{sec:bundle_trivializations}.

\paragraph{Tangent bundle \textit{TM}:}

Any smooth manifold $M$ comes with a set of tangent spaces $\TpM\cong\R^d$.
Their disjoint union%
\footnote{%
    The disjoint union $\coprod_{p\in M} \TpM = \bigcup_{p\in M} \left\{(p,v) \,|\, v\in \TpM\right\}$ of tangent spaces can be thought of as ``remembering'' from which particular tangent space $\TpM$ a certain vector $v\in \TM$ originates, which is necessary for the definition of the projection map $\piTM$.
}
\begin{align}
    \TM\ :=\, \coprod_{p\in M} \TpM \,,
\end{align}
together with a canonically given smooth structure and projection map, defines a smooth fiber bundle known as the \emph{tangent bundle}.
The projection map ${\piTM: \TM\to M}$ is thereby given by the obvious choice ${\piTM(v)=p}$ for $v\in \TpM$.
As derived in Appendix~\ref{apx:coordinate_bases}, local trivializations $\PsiTM: \piTM^{-1}(U) \to U\times \R^d$ of the tangent bundle are canonically induced by charts $x:U\to V\subseteq \R^d$ of the manifold.
We can therefore take the trivializability of $\TM$ as granted and postpone their discussion to Section~\ref{sec:bundle_trivializations}.
A smooth structure on $\TM$ is induced from the smooth structure of $M$ via the above mentioned trivializations from charts.
We skip the technicalities on this construction and refer the interested reader to~\cite{schullerGeometricalAnatomy2016,nakahara2003geometry}.

The thus defined tangent bundle is a \emph{vector bundle} since its typical fiber $\R^d$ is a vector space.
Tangent vector fields, describing for instance a flow on $M$, are formalized as sections $\sigma: M\to \TM$ of the tangent bundle.
Smooth global sections of vector bundles always exist; a standard example is the zero section which assigns the zero vector of $\TpM$ to each $p\in M$.
We want to emphasize that the tangent spaces -- and therefore the tangent bundle -- are defined without reference to coordinate frames, such that sections describe vector fields in a coordinate free way.

After introducing the tangent frame bundle $\FM$ below, we will come back to the tangent bundle and its explicit construction as associated $\GL{d}$-bundle which emphasizes its coordinate free nature.
In Section~\ref{sec:G_associated_bundles} we will analogously construct $\TM$ as a associated $G$-bundle to a $G$-structure $\GM$.

\begin{figure}
    \centering
    \includegraphics[width=1.\columnwidth]{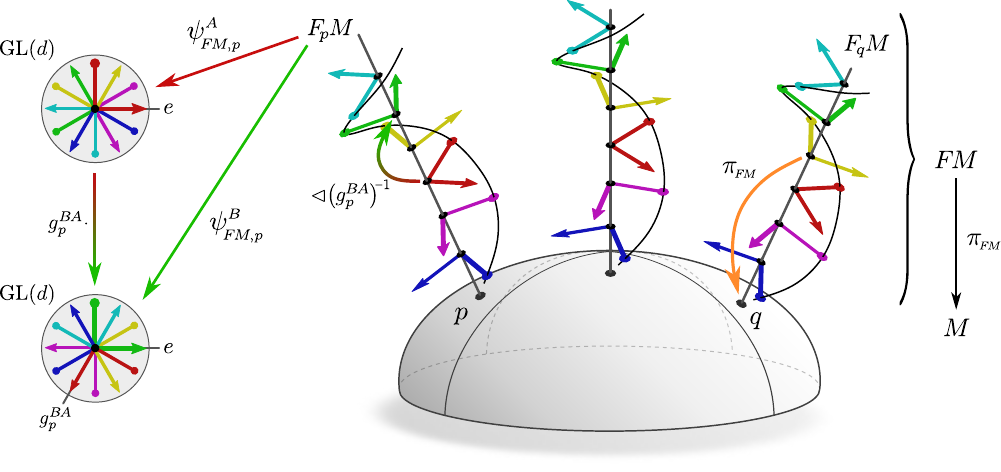}
    \vspace*{-1ex}
    \caption{\small
        A graphical interpretation of the frame bundle $\FM$ over~$M$ and its trivializations.
        The fiber~$\FpM$ over~$p$ is defined as the space of all possible reference frames of~$\TpM$.
        All frames in $\FpM$ are by the projection map $\piFM$ being mapped to that point $p$ in $M$ to which the fiber is attached.
        The fibers $\FpM$ are isomorphic to $\GL{d}$, but come without an origin which would distinguish a preferred choice of reference frame.
        Gauges $\psiFMp^A: \FpM \to \GL{d}$ or $\psiFMp^B: \FpM \to \GL{d}$, introduced in Section~\ref{sec:bundle_trivializations} below, identify the fibers with $\GL{d}$, thereby specifying a preferred frame.
        Different gauges are related by gauge transformations $g_p^{BA} \in \GL{d}$.
        We need to warn the reader about two potential misconceptions:
        Firstly, the frames in different fibers are a-priori not identified with each other in a canonical way, which the redundant colors might suggest.
        Secondly, to minimize clutter, the visualization shows only right-handed, orthonormal frames instead of all possible reference frames.
        As we will discuss in the following Section~\ref{sec:G_associated_bundles}, the shown orthonormal, right-handed frames would correspond to a $G$-structure $\GM$ (a principal $G$-subbundle of $\FM$) for the structure group $G=\SO2$.
        }
    \label{fig:frame_bundle}
\end{figure}

\paragraph{Frame bundle \textit{FM}:}
The space of local reference frames of all tangent spaces $\TpM$ forms the (tangent) \emph{frame bundle}.
Consider the spaces of reference frames (ordered bases) of the individual tangent spaces $\TpM$:
\begin{align}
    \FpM\ :=\ \pig\{ \big[e_{1},\dots,e_{d}\big]\, \pig|\ \{e_{1},\dots,e_{d}\}\ \text{is a basis of } \TpM \pig\}
\end{align}
The frame bundle is defined as their disjoint union $\FM := \coprod_{p\in M} \FpM$ together with the projection map $\piFM: \FM\to M$ which sends frames in $\FpM$ to $p$ and a smooth structure induced from $\TM$.
The typical fiber of the frame bundle is
the general linear group $\GL{d}\cong \FpM$, i.e. the group of invertible $d\!\times\!d$ matrices whose linearly independent columns can be thought of as defining a frame of $\R^d$.
As the frame bundle is constructed from the tangent bundle, its local trivializations $\PsiFM: \piFM^{-1}(U) \to U \times \GL{d}$ are immediately induced from those of $\TM$; see Section~\ref{sec:bundle_trivializations}.
Fig.~\ref{fig:frame_bundle} shows a graphical interpretation of the frame bundle.

Smooth local sections $\sigma:U\to\piFM^{-1}(U)\subseteq \FM$ of the frame bundle map points $p\in U$ to frames in $\FpM$.
They define smooth local frame fields, that is, smoothly varying choices of reference frames for $\TpM,\ p\in U$; visualized in Fig.~\ref{fig:gauge_trafos_manifold}.
As argued in Eq.~\eqref{eq:framefield_gauge_equivalence}, a choice of frame field on $U$ is \emph{equivalent} to a choice of gauge or local trivialization on $U$.
This implies that global frame fields exist only if $\FM$ -- and thus $\TM$ -- are trivial.
We will discuss this equivalence in more depth in Section~\ref{sec:bundle_trivializations}.

A~transitive and free right action on the individual fibers $\FpM\cong\GL{d}$ of the frame bundle is naturally given by the change of frames defined in Eq.~\eqref{eq:frame_rightaction}~\cite{schullerGeometricalAnatomy2016}.
The corresponding action
\begin{align}\label{eq:rightaction_FM}
    \lhd:\ \FM\times \GL{d} \to \FM, \quad
    \big( [e_i]_{i=1}^d,\ g \big)
    \ \mapsto\ 
    [e_i]_{i=1}^d\! \lhd g \ :=\ 
    \left[ \sum\nolimits_j e_j\, g_{ji} \right]_{i=1}^d
\end{align}
on $\FM$ as a whole makes the frame bundle to a principal $\GL{d}$-bundle.
The lack of origin or preferred identity element of the fibers $\FpM$ as $\GL{d}$-torsors reflects the inherent ambiguity of reference frames.

\paragraph{\textit{TM} as GL(\textit{d})-associated vector bundle (\textit{FM}$\boldsymbol{\times \mathds{R}^d)/}$GL(\textit{d})\,:}
In Section~\ref{sec:gauges_gauge_trafos} we expressed~tangent vectors in $\TpM$ in terms of their coefficients in $\R^d$ relative to some reference frame.
The particular choice of frames was thereby irrelevant since the transformation of the coefficients in Eq.~\eqref{eq:components_leftaction} cancels with the transformation of reference frames in Eq.~\eqref{eq:frame_rightaction} such that $v = \sum_i v^A_i e^A_{i} = \sum_i v^B_i e^B_{i}$ are equivalent coordinate representations of the same coordinate free vector $v\in \TpM$.
Following this idea, one can construct the tangent bundle from the frame bundle by pairing reference frames with coefficient vectors and taking a quotient to collapse the resulting redundant descriptions of tangent vectors relative to different frames to one unique element.

In order to construct the tangent bundle in this way, consider the product $\FM\times\R^d$ which can be seen as a fiber bundle with base space $M$ and a typical fiber $\GL{d}\times\R^d$.
This bundle consists of pairs of (mutually unrelated) reference frames and coefficients.
Motivated by the equivalent expression of tangent vectors in different reference frames we define the \emph{equivalence relation}%
\footnote{\label{footnote:equiv_rel}%
    An \emph{equivalence relation} on a set $X$ is a binary relation $\sim$ which is
    \emph{reflexive} ($x\!\sim\! x$),
    \emph{symmetric} ($x\!\sim\! y \Leftrightarrow y\!\sim\! x$) and
    \emph{transitive} ($x\!\sim\! y \wedge y\!\sim\! z \Rightarrow x\!\sim\! z$).
    It defines a partitioning of $X$ into \emph{equivalence classes} $[x] := \{y\in X | x\sim y\}$ of elements $x\in X$.
    The space of equivalence classes $X/\!\!\sim\ := \{[x] \,|\, x\in X\}$ is called the \emph{quotient set} of $X$ by $\sim$.
}
\begin{align}\label{eq:equiv_relation_TM}
    \big([e_i]_{i=1}^d,\,\mathscr{v}\big)\ \sim_{\GL{d}}\, \big([e_i]_{i=1}^d\!\lhd g^{-1},\, g\cdot \mathscr{v}\big) \qquad\forall\, g\in \GL{d}
\end{align}
on $\FM\times\R^d$.
As an equivalence relation, it partitions $\FM\times\R^d$ into \emph{equivalence classes} $\big[[e_i]_{i=1}^d,\,\mathscr{v}\big]$.
The space of these equivalence classes is the quotient space $(\FM\times\R^d)/\GL{d}$.
The projection map
\begin{align}\label{eq:associated_TM_proj}
    \pi_{\scalebox{.75}{$\sim_{\GL{d}}$}}:\ 
    (\FM\times\R^d)/\GL{d} \to M,\ \ 
    \big[[e_i]_{i=1}^d,\,\mathscr{v}\big] \mapsto \piFM\!\left([e_i]_{i=1}^d\right) \,,
\end{align}
which is induced from the frame bundle, makes $(\FM\times\R^d)/\GL{d}$ to a fiber bundle with base space $M$ and typical fiber~$\R^d$.
Note that the projection map in Eq.~\eqref{eq:associated_TM_proj} is well defined as it is independent of the representative of the equivalence class, i.e.
$\pi_{\scalebox{.75}{$\sim_{\GL{d}}$}}\left(\big[[e_i]_{i=1}^d\lhd g^{-1},\, g\cdot \mathscr{v}\big]\right) := \piFM\left([e_i]_{i=1}^d\!\lhd g^{-1}\right) = \piFM\left([e_i]_{i=1}^d\right)$,
where we used that the right action $\lhd$ preserves the fibers of $\FM$.
The vector space structure of $\R^d$ makes $(\FM\times\R^d)/\GL{d}$ to a vector bundle with linear combinations within the same fiber being defined by
\begin{align}\label{eq:associated_bdl_linear_combination}
    \alpha \big[[e_i]_{i=1}^d,\,\mathscr{v}\big] + \beta \big[[e_i]_{i=1}^d,\,\mathscr{w}\big]
    \ :=\ \big[[e_i]_{i=1}^d,\,\alpha\mathscr{v} + \beta\mathscr{w} \big] \,,
\end{align}
for arbitrary $\alpha,\beta\in\R$ and $\mathscr{v},\mathscr{w}\in\R^d$.
This definition is easily checked to be independent of the choice of representative in both summands.

The thus defined bundle is isomorphic to the tangent bundle,
\begin{align}
    \TM\, \cong\, (\FM\times\R^d)/\GL{d} \,,
\end{align}
with the vector bundle $M$-isomorphism given by the fiber wise linear map
\begin{align}\label{eq:A_TM_isomorphism}
    \chi:(\FM\times\R^d)/\GL{d}\ \to \TM,\ \ \ \left[[e_i]_{i=1}^d,\, \mathscr{v}\right] \mapsto \sum_{i=1}^d \mathscr{v}_i e_i
\end{align}
which takes some representative tuple of frame and coefficient vector from the equivalence class and maps them to the corresponding tangent vector.
By the definition of the equivalence relation $\sim_{\GL{d}}$, this function is independent of the choice of representative, that is,
$
    \forall g\in \GL{d}: \ 
    \chi\left( \big([e_i]_{i=1}^d\lhd g^{-1},\, g\cdot \mathscr{v}\big) \right) = 
    {\sum_i (g\!\cdot\!\mathscr{v})_i \left([e_j]_{j=1}^d\!\lhd g^{-1}\right)_i} =
    \sum_i \mathscr{v}_i e_i \, ;\,
$
cf.~Eq.~\eqref{eq:vector_in_different_bases}.
As discussed in~\cite{schullerGeometricalAnatomy2016}, the inverse is given by taking a tangent vector, projecting it on an arbitrary frame and taking the equivalence class.

The bundle $(\FM \times \R^d) / \GL{d}$ is by construction associated to $\FM$ as $\GL{d}$-bundle, that is, it has the same transition functions in $\GL{d}$ as $\FM$, as we will derive in Section~\ref{sec:bundle_trivializations}.
The construction of $\TM$ as quotient $(\FM\times\R^d)/\GL{d}$ emphasizes the \emph{coordinate free} nature of the tangent bundle in a very intuitive way:
it considers all possible choices of coordinatizations of the tangent spaces and treats them as being equivalent by taking a quotient.

%% file: chapters/63_associated_G_bundles.tex

\subsection%
    [\textit{G}-structures \textit{GM} and associated feature vector bundles \texorpdfstring{$\A$}{A}]%
    {\textit{G}-structures \textit{GM} and associated feature vector bundles $\boldsymbol{\A}$}
\label{sec:G_associated_bundles}

We will now introduce $G$-structures $\GM$ as distinguished subsets of frames in $\FM$, which encode additional geometric structure on $M$ that is to be respected by coordinate independent CNNs.
The tangent bundle is via a similar associated bundle construction to that in the last section reintroduced as an associated $G$-bundle.
This approach can be generalized to construct any other associated $G$-bundle, which we use to define the feature vector bundles $\A$.
All such constructed bundles are associated to each other, that is, they differ only in their fiber $F$ but share the same base space $M$, structure group $G$ and transition functions $g^{BA}$ between trivializing neighborhoods.
The local trivializations of the bundles and their mutual gauge transformations are discussed in detail in the next Section~\ref{sec:bundle_trivializations}.

\paragraph{\textit{G}\hspace{.5pt}-\hspace{.5pt}structures \textit{GM}:}
As discussed in Section~\ref{sec:21_main} and Table~\ref{tab:G_structures}, it is often possible to work with a \emph{distinguished subset of reference frames} which are related by the action of a \emph{reduced structure group} $G \leq \GL{d}$.
This is best understood by discussing a few examples before coming to a technical definition below.
For instance, a restriction to orthonormal frames 
\begin{align}
    \OpM\ :=\ \pig\{ \big[e_{1},\dots,e_{d}\big]\, \pig|\ 
    \{e_{1},\dots,e_{d}\}\ \text{is an \emph{orthonormal} basis of } \TpM\ \text{w.r.t.}\ \eta \pig\}\ \cong\ \O{d}
\end{align}
gives rise to a principal subbundle $\OM$ of $\FM$ with structure group $\O{d}$.
Note that the orthonormality of reference frames is judged by the metric $\eta$ on $M$ -- different choices of metrics on a manifold therefore correspond to different subsets of preferred reference frames for the same structure group $\O{d}$.
As a second example, consider a choice of orientation on an orientable manifold, which allows to specify a preferred notion of frames%
\footnote{
    Conversely, non-orientable manifolds do not allow for a reduction of structure group to $\operatorname{GL}^{\!+}\!(d)$.
}
\begin{align}
    \operatorname{GL}^{\!+}_p\!M\ :=\ \pig\{ \big[e_{1},\dots,e_{d}\big]\, \pig|\ \{e_{1},\dots,e_{d}\}\ \text{is a \emph{positively oriented} basis of } \TpM \pig\}\ \cong\ \operatorname{GL}^{\!+}\!(d)
\end{align}
and a corresponding principal subbundle $\operatorname{GL}^{\!+}\!(d)M$ of $\FM$ with structure group $\operatorname{GL}^{\!+}\!(d)$.
Again, the two different choices of orientations correspond to two different choices of subbundles of accordingly oriented frames.
Combining both requirements for the orthonormality and right-handedness of frames results in an $\SO{d}$-structure with fibers
\begin{align}
    \SOpM\ :=\ \pig\{ \big[e_{1},\dots,e_{d}\big]\, \pig|\ 
    \{e_{1},\dots,e_{d}\}\ \text{is a \emph{positively oriented}, \emph{orthonormal} basis of } \TpM \pig\}\ \cong\ \SO{d} \,,
\end{align}
Fig.~\ref{fig:frame_bundle} can be thought of as showing an $\SO{2}$-structure since only right-handed, orthonormal frames are shown (the typical fiber $\GL{d}$ should then be labeled $\SO2$).
Different choices of $\SO{d}$-structures correspond either to an opposite handedness of frames, sticking to the same notion of orthonormality, or to a different choice of metric (or both).
The exact same pattern repeats for volume forms $\omega$ (on orientable manifolds $M$):
they allow to specify a preferred notion of frames
\begin{align}
    \operatorname{SL}_p\!M\ :=\ \pig\{ \big[e_{1},\dots,e_{d}\big]\, \pig|\ \{e_{1},\dots,e_{d}\}\ \text{is a basis of }\, \TpM\ \text{with \emph{unit volume} w.r.t.}\ \omega \pig\}\ \cong\ \operatorname{SL}(d)
\end{align}
and thus principal subbundles $\operatorname{SL}\!M$ of $\FM$ with structure group $\operatorname{SL}(d)$.
The specific set of frames which are preferred depends here on the specific choice of volume form.
As a last example, consider $\{e\}$-structures, corresponding to a trivial structure group $G=\{e\}$ and therefore consisting of one single frame at each point~$p$.
By definition, $\{e\}$-structures are equivalent to global (smooth) frame fields $\sigma \in \Gamma(\FM)$:
\begin{align}
    \epM\ :=\ \pig\{ \big[e_{1},\dots,e_{d}\big] = \sigma(p) \pig\}\ \cong\ \{e\}
\end{align}
They do therefore only exist on trivial manifolds.
Figs.~\ref{fig:frame_field_automorphism_1} and~\ref{fig:frame_field_automorphism_2} visualize two different choices of $\{e\}$-structures~$\eM$ on~$M=\R^2$.

All of these examples represent specific choices of $G$\emph{-structures} $\GM$ on $M$.
In general, a $G$-structure on~$M$ is a principal $G$-\emph{subbundle} of $\FM$, that is, a ``smoothly varying'' choice of \emph{subsets} $\GpM \subseteq \FpM$ which are right $G$-torsors w.r.t. $\lhd$ for any $p\in M$ \cite{sternberg1999lectures,piccione2006theory,crainic2013GStructuresExamples}.%
\footnote{\label{footnote:GpM_G_orbit_in_FpM}
    As $\FpM$ is a right $\GL{d}$-torsor, any $G$-orbit $\GpM$ in $\FpM$ is automatically guaranteed to be a right $G$-torsor.
}
The smoothness can hereby be formalized by requiring that around each frame $[e_i]_{i=1}^d \in \GpM$ there exists a neighborhood $U$ of $p$ on which a smooth section $\sigma: U \mapsto \piGM^{-1}(U) \subseteq GM$ with $\sigma(p) = [e_i]_{i=1}^d$ exists.
The projection
\begin{align}
    \piGM :=\, \piFM\mkern-2mu \big|_{\scalebox{.6}{$\GM$}}:\ \ \GM \to M
\end{align}
of $\GM$ is hereby simply given by the restriction of the projection map of $\FM$ to $\GM$.
Together with the restriction
\begin{align}\label{eq:rightaction_GM}
    \lhd:\ \GM\times G \to \GM, \quad
    \big( [e_i]_{i=1}^d,\ g \big)
    \ \mapsto\ 
    [e_i]_{i=1}^d\! \lhd g \ :=\ 
    \left[ \sum\nolimits_j e_j\, g_{ji} \right]_{i=1}^d
\end{align}
of the right action of $\GL{d}$ on $\FM$ in Eq.~\eqref{eq:rightaction_FM} to an action of $G \leq \GL{d}$ on $\GM \subseteq \FM$, this makes the $G$-structure to a \emph{principal $G$-bundle} $\GM\!\xrightarrow{\piGM}\!M$.
However, it is important to note that there are \emph{multiple choices} of such subbundles, corresponding to different $G$-structures for the same structure group~$G$; compare this claim with the examples above.
As discussed earlier, the topology of a bundle might obstruct the reduction to a structure group~$G$, and thus the existence of a corresponding $G$-structure $\GM$.

While the above definition of $G$-structures would be sufficient, it is instructive to briefly review some alternative, equivalent definitions.
The claim that $\GM$ is a principal $G$-\emph{subbundle} of $\FM$ is made precise by defining it as a tuple $(P, \mathscr{E})$ consisting of a choice of an (also non-unique) principal $G$-bundle $P$ over $M$ together with a smooth, right $G$-equivariant embedding $\mathscr{E}: P \to \FM$ (over $M$).%
\footnote{
    The embedding is a principal $G$-bundle $M$-morphism as introduced in Section~\ref{sec:fiber_bundles_general} with the group homomorphism $\theta:G\to\GL{d}$ being the canonical inclusion of the subgroup $G\leq\GL{d}$ into $\GL{d}$.
}
This is visualized by the following diagram, which is required to commute for any $g\in G$:
\begin{equation}\label{cd:GM_def_embedding}
\begin{tikzcd}[row sep=3.em, column sep=2.5em]
    P
        \arrow[rr, "\mathscr{E}", hook]
    && \FM
    \\
    P
        \arrow[rr, "\mathscr{E}", hook]
        \arrow[u, "\lhd_{\overset{}{\mkern-2muP}}\mkern2mu g\:"]
        \arrow[dr, "\pi_P"']
    && \FM
        \arrow[u, "\:\lhd\mkern2mu g"']
        \arrow[dl, "\piFM"]
    \\
    & M
\end{tikzcd}
\end{equation}
Different subsets of preferred frames correspond in this viewpoint to different choices of embeddings ${\GM = \mathscr{E}(P)}$ of $P$ in $\FM$.
$G$-structures are furthermore equivalent to sections of the form $s: M \mapsto \FM/G$ with $\GM = s(M)$, which emphasizes that $\GpM = s(p) \in \FpM/G$ is indeed a choice of $G$-orbit in $\FpM$ as stated in footnote~\ref{footnote:GpM_G_orbit_in_FpM}.
Yet another definition of $G$-structures is in terms of (equivalence classes of) $G$-atlases~\cite{wendlLectureNotesBundles2008}.
As this is the viewpoint which might be taken in an implementation of $\GM$-convolutions, we discuss it in more detail in the following Section~\ref{sec:bundle_trivializations}.
For the interested reader we want to mention that $G$-structures are a specific case of the more general concept of a \emph{reduction (or lift) of structure groups}~\cite{sternberg1999lectures,piccione2006theory,crainic2013GStructuresExamples}.

$G$-structures are of pivotal importance for the theory of $\GM$-convolutions.
The particular choice of $G$-structure determines the specific set of reference frames over which the $G$-steerable template kernel is shared.
By the gauge equivariance of the kernels, $\GM$-convolutions are guaranteed to respect the $G$-structure, i.e. to be $\GM$-\emph{coordinate independent}.
As derived in Section~\ref{sec:isometry_intro}, the isometries with respect to which a $\GM$-convolution is equivariant are exactly those which preserve the $G$-structure (i.e. those which induce automorphisms of~$\GM$).

\paragraph{\textit{TM} as \textit{G}-associated vector bundle (\textit{GM}$\boldsymbol{\times \mathds{R}^d)/}$\textit{G}\,:}

Given a $G$-structure $\GM$, one can adapt the associated $\GL{d}$-bundle construction of $\TM$ from $\FM$ in Section~\ref{sec:GL_associated_bundles} to a similar associated $G$-bundle construction of $\TM$ based on $\GM$.
Instead of expressing tangent vectors relative to general frames in $\FM$, they will thereby be expressed relative to the distinguished frames in~$\GM$ and the quotient is taken w.r.t. the reduced structure group~$G$ instead of~$\GL{d}$.
The resulting bundle is by design associated to $\GM$ (or to $\FM$ with a $G$-atlas, which is equivalent as explained in the next section) and therefore has transition functions which take values in~$G$.
The restriction of $\chi$ in Eq.~\eqref{eq:A_TM_isomorphism} to $(\GM\times\R^d)/G$ yields a vector bundle isomorphism
\begin{align}
    \TM\, \cong\, (\GM\times\R^d)/G \,.
\end{align}
While all three bundles $\TM$, $(\FM\times\R^d)/\GL{d}$ and $(\GM\times\R^d)/G$ are thus isomorphic \emph{as vector bundles}, they are only isomorphic as associated $G$-bundles if $\TM$ and $(\FM\times\R^d)/\GL{d}$ are endowed with a $G$-structure (or $G$-atlas), which is a-priori not the case.
In contrast, the bundle $(\GM\times\R^d)/G$ comes with a $G$-structure by design.
For a precise definition of associated $G$-bundle isomorphisms we refer to \cite{schullerGeometricalAnatomy2016}.

\paragraph{Associated feature vector bundles $\boldsymbol{\A}$:}
The associated $G$-bundle construction $(\GM\times\R^d)/G$ can be generalized to attach other fibers with other group actions to the $G$-structure $\GM$.
Indeed, \emph{any} bundle associated to $\GM$ can be constructed in this way.
Important examples in differential geometry are the cotangent bundle $\TsM$ with its typical fiber being the dual $\R^{d*}$ of $\R^d$, acted on by the dual action, or the $(r,s)$ tensor bundles $T^r_s\!M$ with fibers $\big(\R^d\big)^{\otimes r}\! \otimes\! \big(\R^{d*}\big)^{\otimes s}$ being acted on by the corresponding tensor product representation of~$G$.

In the following we consider \emph{associated feature vector bundles} with feature vector coefficients $\R^c$ as typical fibers.
Under gauge transformations, these fibers are acted on from the left by a multiplication with a group representation $\rho:G\to\GL{c}$, that is, Eq.~\eqref{eq:gauge_trafo_leftaction} is instantiated by $\blacktriangleright_\rho:\ {G\times\R^c \to \R^c,}\ \ {(g,\mathscr{f}) \mapsto \rho(g)\mathscr{f}}$.
Similar to before, feature vector bundles are then constructed as a quotient
\begin{align}\label{eq:associated_bundle_def}
    \A\ :=\ (\GM\times\R^c)/\!\sim_{\!\rho}
\end{align}
with the equivalence relation $\sim_{\!\rho}$ here given by
\begin{align}\label{eq:equiv_relation_A}
    \big([e_i]_{i=1}^d,\, \mathscr{f} \mkern1mu\big)\ \sim_{\!\rho}\ 
    \big([e_i]_{i=1}^d\!\lhd g^{-1},\ \rho(g) \mathscr{f} \mkern1mu\big) \qquad\forall g\in G.
\end{align}
The elements of $\A$ are the equivalence classes $\big[[e_i]_{i=1}^d,\ \mathscr{f}\big]$ of feature vector coefficients relative to reference frames and are therefore \emph{coordinate free}.
A (well defined) projection map is again induced from the projection of the $G$-structure:
\begin{align}\label{eq:associated_A_proj}
    \piA: \A \to M,\ \ 
    \big[[e_i]_{i=1}^d,\,\mathscr{f}\big] \mapsto \piGM \big( [e_i]_{i=1}^d \big)
\end{align}
Linear combinations on the fibers are defined in analogy to Eq.~\eqref{eq:associated_bdl_linear_combination}.
Since such defined feature vector bundles are associated to $\GM$, their structure group is $G\leq\GL{d}$, as we will explicitly derive in the next Section~\ref{sec:bundle_trivializations}.%
\footnote{
    Strictly speaking, the transition functions will take values in $\rho(G) \leq \GL{c}$ instead of $G \leq \GL{d}$, however, since the resulting transitions are still $G$\emph{-compatible}, the term ``$G$-valued'' is usually adapted to include such cases~\cite{wendlLectureNotesBundles2008}.
}
Note that this definition includes tangent vector fields and scalar fields, which can of course be processed as feature fields, for $\rho(g)=g$ and $\rho(g)=1$, respectively.

The construction of $\A$ as an associated $G$-bundle models $\GM$-coordinate independent feature vectors on~$M$:
features $f_p \in \A$ are equivalently expressed relative to arbitrary frames in $\GM$, with feature coefficients in different coordinatizations being related via Eq.~\eqref{eq:equiv_relation_A}, but do not have a well defined coordinate expression relative to other frames.
From an engineering viewpoint, this is reflected in the $G$-steerability of convolution kernels which ensures that measurements of features are performed \emph{relative} to frames in $\GM$ but allows to discriminate between patterns whose poses are not related by a $G$-valued gauge transformation in an \emph{absolute} sense.

\paragraph{Associated feature vector field and feature spaces:}
Smooth, coordinate free feature fields are defined as smooth global sections $f\in\Gamma(\A)$ of the feature vector bundles, that is, as smooth maps $f:M\to\A$ satisfying $\piA\circ f=\id_M$.
As discussed before, such feature fields are guaranteed to exist since vector bundles always admit smooth global sections.
In the following Section~\ref{sec:bundle_trivializations} we show how a local bundle trivialization over $U^A$ allows to represent $f$ by a field $f^A:U^A\to\R^c$ of feature vector coefficients.
A different trivialization over $U^B$ will lead to a different coefficient field $f^B:U^B\to\R^c$ representing $f$ locally.
From the transition maps between bundle trivializations it will follow that both coefficient fields are on the overlap $U^{AB}=U^A\cap U^B$ of their domains related by $f^B(p)=\rho\big(g_p^{BA}\big) f^A(p)$.
The commutative diagram in Fig.~\ref{fig:trivialization_A_sections} visualizes the relations between feature vector fields and their local trivializations.

The feature spaces of coordinate independent CNNs usually consist of multiple independent feature fields over the same base space.
The bundle describing a feature space as a whole is the \emph{Whitney sum} $\bigoplus_i\A_i$ of the feature vector bundles $\A_i\xrightarrow{\scalebox{.85}{$\pi_{\scalebox{.6}{$\!\A_i$}}$}}M$ underlying its individual fields.
As such it has the same base space $M$, a typical fiber $\bigoplus_i\R^{c_i} \cong \R^{\sum_i\!c_i}$ defined as direct sum of the individual fields' fibers and is equipped with the obvious projection map.
It is associated to $\TM$, $\FM$, $\GM$ and the $\A_i$ as $G$-bundles and can therefore equivalently be defined as
\begin{align}
    \scalebox{1.1}{$\bigoplus$}_i\,\A_i\ \ \cong\,\ \left(\GM\times\R^{\sum_i\!c_i}\right)\!\!\Big/\!\!\sim_{\oplus_i\rho_i}
\end{align}
Note that the direct sum $\bigoplus_i\rho_i$ of representations $\rho_i$ defining $\A_i$ guarantees that the transition maps of $\bigoplus_i\A_i$ transform each individual field independently.
The feature spaces are then defined as the spaces $\Gamma\!\left(\bigoplus_i\A_i\right)$ of global sections of the Whitney sum bundle.

%% file: chapters/64_trivializations.tex

\subsection%
    [Local bundle trivializations of \texorpdfstring{\textit{TM}, \textit{FM}, \textit{GM} and $            \A $}{TM, GM and A}]%
    {Local bundle trivializations of \texorpdfstring{\textit{TM}, \textit{FM}, \textit{GM} and $\boldsymbol{\A}$}{TM, GM and A}}
\label{sec:bundle_trivializations}

While the global theory of coordinate independent CNNs is elegantly formalized in terms of coordinate free fiber bundles, a numerical implementation requires coordinate free feature vectors $f(p)\in\A_p$ to be expressed by coefficient vectors ${f^A(p):=\psiAp^A\big(f(p)\big)\in\R^c}$ relative to some choice of reference frame $\big[e_{i}^A\big]_{i=1}^d\in \GpM$ as described in Section~\ref{sec:feature_fields}.
In the language of fiber bundles, this corresponds to a choice of local trivializations or gauges $\PsiGM^A$, $\PsiTM^A$, $\PsiFM^A$ and $\PsiA^A$, all of which transform simultaneously if $\GM$, $\TM$, $\FM$ and $\A$ are taken to be $G$-associated to each other.
Recall that a local description and thus implementation via a $G$-atlas, consisting of local trivializations which cover $M$ and satisfy the three conditions~\eqref{eq:transition_condition_1}, \eqref{eq:transition_condition_2} and~\eqref{eq:transition_condition_3}, is fully equivalent to the global, coordinate free theory.

In this section we work out the associated trivializations of $\TM$, $\FM$, $\GM$ and $\A$ and their synchronous gauge transformations.
To stay consistent with Section~\ref{sec:21_main}, we start out by assuming trivializations of $\TM$ to be given and discuss how they induce trivializations of $\FM$ and corresponding local frame fields.
If a $G$-atlas is chosen for $\TM$ and thus $\FM$, it gives rise to a $G$-structure $\GM$ whose $G$-atlas agrees with that of $\FM$.
The local trivializations of any associated $G$-bundle, in particular those of the feature vector bundles $\A$, follow from those of $\GM$.
These trivializations recover the transformation law of feature fields from Section~\ref{sec:feature_fields}.

\paragraph{Trivializations of \textit{TM}:}

As the tangent bundle has $\R^d$ as typical fiber, its local trivializations are given by maps of the form
\begin{align}\label{eq:trivialization_TM}
    {\PsiTM: \piTM^{-1}\!\left(U\right) \to U\times\R^d} .
\end{align}
These trivializations correspond to the (spatially smoothly varying) pointwise gauges
\begin{align}\label{eq:def_psiTMp}
    \psiTMp: \TpM \to \R^d
\end{align}
from Eq.~\eqref{eq:gauge_definition} by identifying $\PsiTM(v) = \left(\piTM(v),\, \psiTMp(v) \right)$ for $p=\piTM(v)$.
In order to respect the vector space structures of the fiber $\R^d$ and the tangent spaces $\TpM$, the trivializations $\PsiTM$ are defined as \emph{vector bundle isomorphisms} between $\piTM^{-1}(U)$ and $U\times\R^d$, that is, the maps $\psiTMp$ are required to be linear and invertible (i.e. vector space isomorphisms).
The transition maps between different trivializations of $\TM$ will in general take values in the general linear group $\GL{d}$, the (linear) automorphism group of $\R^d$.

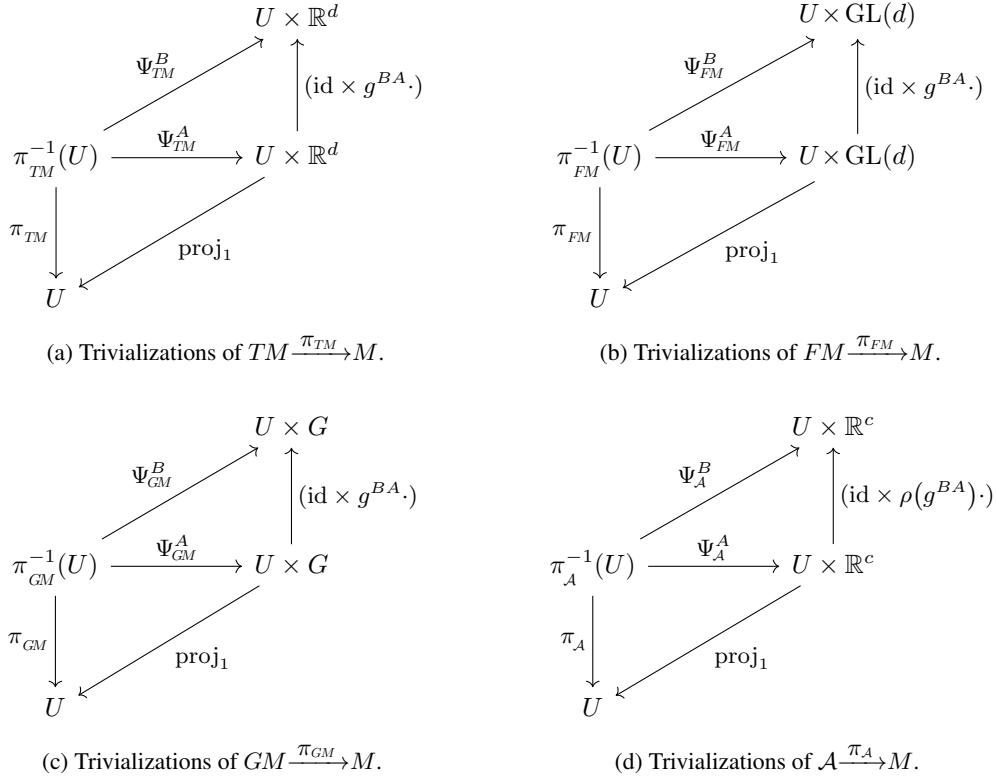
\begin{figure}
    \centering
    \begin{subfigure}[b]{.4\textwidth}
        \centering
        \begin{tikzcd}[row sep=3.5em, column sep=5.em]
            & U\times \R^d
            \\
              \piTM^{-1}(U) \arrow[d, swap, "\piTM"]
                            \arrow[r, "\PsiTM^A"]
                            \arrow[ru, "\PsiTM^B"]
            & U\times \R^d  \arrow[u, swap, "(\id\times g^{BA}\cdot)"]
                            \arrow[ld, "\proj_1"]
            \\
              U
        \end{tikzcd}
        \caption{\small
            Trivializations of
            $\TM {\xrightarrow{\scalebox{1}{$\,\pi_{\scalebox{.55}{$\TM$}}\,$}}} M$.
        }
        \label{fig:trivialization_TM}
    \end{subfigure}
    \hspace*{7ex}
    \begin{subfigure}[b]{.4\textwidth}
        \centering
        \begin{tikzcd}[row sep=3.5em, column sep=5.em]
            & U \!\times\mkern-2mu \GL{d} 
            \\
              \piFM^{-1}(U) \arrow[d, swap, "\piFM"]
                            \arrow[r, "\PsiFM^A"]
                            \arrow[ru, "\PsiFM^B"]
            & U \!\times\mkern-2mu \GL{d}
                            \arrow[u, swap, "(\id\times g^{BA}\cdot)\ "]
                            \arrow[ld, "\proj_1"]
            \\
              U
        \end{tikzcd}
        \caption{\small
            Trivializations of
            $\FM {\xrightarrow{\scalebox{1}{$\,\pi_{\scalebox{.55}{$\FM$}}\,$}}} M$.
        }
        \label{fig:trivialization_FM_simplified}
    \end{subfigure}
    \\[3ex]
    \begin{subfigure}[b]{.4\textwidth}
        \centering
        \begin{tikzcd}[row sep=3.5em, column sep=5.em]
            & U\times G 
            \\
              \piGM^{-1}(U) \arrow[d, swap, "\piGM"]
                            \arrow[r, "\PsiGM^A"]
                            \arrow[ru, "\PsiGM^B"]
            & U\times G     \arrow[u, swap, "(\id\times g^{BA}\cdot)"]
                            \arrow[ld, "\proj_1"]
            \\
              U
        \end{tikzcd}
        \caption{\small
            Trivializations of
            $\GM {\xrightarrow{\scalebox{1}{$\,\pi_{\scalebox{.55}{$\GM$}}\,$}}} M$.
        }
        \label{fig:trivialization_FM_simplified}
    \end{subfigure}
    \hspace*{7ex}
    \begin{subfigure}[b]{.4\textwidth}
        \centering
        \begin{tikzcd}[row sep=3.5em, column sep=5.em]
            & U\times \R^c
            \\
              \piA^{-1}(U)  \arrow[d, swap, "\piA"]
                            \arrow[r, "\PsiA^A"]
                            \arrow[ru, "\PsiA^B"]
            & U\times \R^c  \arrow[u, swap, "(\id\times \rho\big(g^{BA}\big)\cdot)"]
                            \arrow[ld, "\proj_1"]
            \\
            U
        \end{tikzcd}
        \caption{\small
            Trivializations of
            $\A {\xrightarrow{\scalebox{1}{$\,\pi_{\scalebox{.55}{$\A$}}\,$}}} M$.
        }
        \label{fig:trivialization_A}
    \end{subfigure}
    \vspace*{1ex}
    \caption{\small
        Visualization of the local trivializations of the associated $G$-bundles $\TM$, $\FM$, $\GM$ and $\A$ in terms of commutative diagrams where we abbreviate $U=U^A\cap U^B$.
        A~$G$-atlas $\big\{ U^X, \PsiTM^X \big\}$ of the tangent bundle with transition maps $g^{BA}: U \to G$ implies a $G$-structure $\GM$ and induces $G$-atlases for $\FM$, $\GM$ and $\A$ with compatible transition functions.
        More detailed commutative diagrams which show sections $\sigma:U\to\pi_{\scriptstyle\!F\!M}^{-1}(U)$ and the right action $\lhd$ on the frame bundle are given in Figs.~\ref{fig:trivialization_FM_non-collapsed} and~\ref{fig:trivialization_FM_section}.
        Feature fields, modeled as sections $f:M\to\A$ of the associated feature vector bundle $\A$, and their local trivializations $f^A:U^A\to\R^c$ are shown in Fig.~\ref{fig:trivialization_A_sections}.
        A graphical interpretation of the commutative diagram for $\TM$, restricted to one single tangent space $\TpM$, is given in Fig.~\ref{fig:gauge_trafos}.
    }
    \label{fig:trivializations_TM_FM_A}
\end{figure}

If further structure is specified on the tangent bundle, the trivializations are required to respect this structure.
For instance, if a metric is defined on $M$ and thus $\TM$, the maps $\psiTMp$ are required to be isometric, i.e. to map vectors in $\TpM$ in such a way to vectors in $\R^d$ that norms and angles are preserved.
As the trivializations are then only allowed to differ in their direction and orientation, different trivializations are guaranteed to be related by a reduced structure group $\O{d}$, corresponding to the metric as $\O{d}$-structure.
More generally, a $G$-structure on $\TM$ requires -- or is implied by -- a choice of $G$-atlas $\big\{ \big(U^X, \PsiTM^X \big) \big\}_{\!X\in\mathfrak{X}}\,$.
Two different trivializations $\PsiTM^A$ and $\PsiTM^B$ of such a $G$-atlas are on $U^A\cap U^B$ related by ${\PsiTM^B\circ\big(\PsiTM^A\big)^{-1}}$ as defined in Eq.~\eqref{eq:transition_function_general_bdl} with $G$-valued transition functions
\begin{align}\label{eq:transition_fct_TM_gAB}
    g^{BA}: U^A\mkern-1mu \cap\mkern-1mu U^B \to G,\ \ \ p \mapsto \psiTMp^B\circ \big(\psiTMp^A\big)^{-1} \,,
\end{align}
which define the left action $\btr: G\times \R^d \to \R^d,\ \ (g,\mathscr{v}) \mapsto g\cdot \mathscr{v}$ on the typical fiber.
For a graphical intuition on the pointwise action of the transition functions on individual fibers we refer back to Fig.~\ref{fig:gauge_trafos}.
A diagrammatic visualization of local trivializations of $\TM$ and their transitions is given in Fig.~\ref{fig:trivialization_TM}.

\paragraph{Induced trivializations of \textit{FM} and frame fields:}

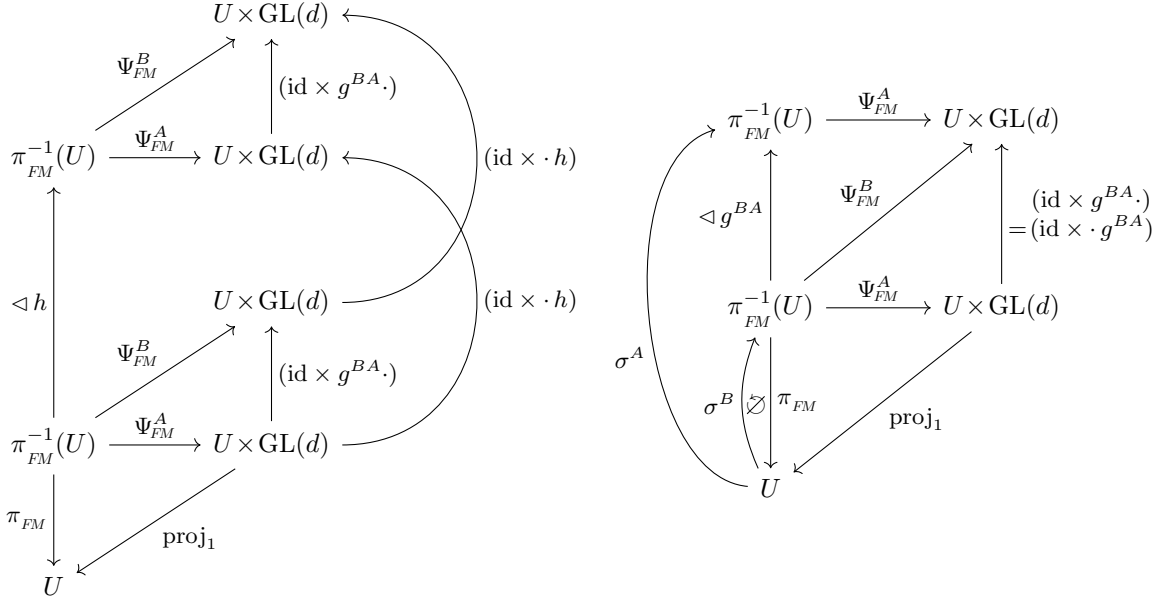
\begin{figure}
    \centering
    \begin{subfigure}[b]{0.47\textwidth}
        \begin{tikzcd}[row sep=3.5em, column sep=3.5em]
            & U\mkern-3mu\times\mkern-2.5mu \GL{d}
            \\
              \piFM^{-1}(U) \arrow[r, "\PsiFM^A"]
                            \arrow[ru, "\PsiFM^B"]
            & U\mkern-3mu\times\mkern-2.5mu \GL{d}
                            \arrow[u, swap, "(\id\times g^{BA}\cdot)"]
            \\
            & U\mkern-3mu\times\mkern-2.5mu \GL{d}
                            \arrow[uu, swap, "(\id\times \cdot\,h)", bend right=90, looseness=1.6]
            \\
            \piFM^{-1}(U)   \arrow[d, swap, "\piFM"]
                            \arrow[r, "\PsiFM^A"]
                            \arrow[ru, "\PsiFM^B"]
                            \arrow[uu, "\lhd\,h"]
            & U\mkern-3mu\times\mkern-2.5mu \GL{d}
                            \arrow[u, swap, "(\id\times g^{BA}\cdot)"]
                            \arrow[ld, "\proj_1"]
                            \arrow[uu, swap, "(\id\times \cdot\,h)", bend right=90, looseness=1.6]
            \\
            U
        \end{tikzcd}
        \hfill
        \caption{\small
            The trivializations of the frame bundle are right equivariant, i.e. they satisfy
            $\PsiFM\circ\lhd\, h\, =\, (\id\times\cdot h)\circ\PsiFM$ for any $h\in \GL{d}$.
        }
        \label{fig:trivialization_FM_non-collapsed}
    \end{subfigure}
    \hfill
    \begin{subfigure}[b]{0.47\textwidth}
        \hfill
        \begin{tikzcd}[row sep=5em, column sep=4em,
                       execute at end picture={
                            \node [] at (-1.83, -1.4) {$\noncommutative$};
                            }]
              \piFM^{-1}(U)
                            \arrow[r, "\PsiFM^A"]
            & U\mkern-3mu\times\mkern-2.5mu \GL{d}
            \\
              \piFM^{-1}(U) \arrow[d, "\piFM", bend left=0]
                            \arrow[r, "\PsiFM^A"]
                            \arrow[ru, "\PsiFM^B"]
                            \arrow[u, "\lhd\,g^{BA}"]
            & U\mkern-3mu\times\mkern-2.5mu \GL{d}
                            \arrow[u, swap, "$\phantom{=}\,(\id\times g^{BA}\cdot)$\\$=\!(\id\times \cdot\,g^{BA})$" align=left]
                            \arrow[ld, "\proj_1"]
            \\
              U             \arrow[u,  "\sigma^B", pos=0.5, shift left=.5, bend left=22.5]
                            \arrow[uu, "\sigma^A", pos=0.45, bend left=80, looseness=.8]
        \end{tikzcd}
        \vspace*{8ex}
        \caption{\small
            If identity sections $\sigma^A$ and $\sigma^B$ are added to the diagram, the left and right actions agree with each other
            since $\psiFMp^A\circ\sigma^A(p)=e$ and $g\cdot e=e\cdot g\ \ \forall g\in \GL{d}$.
        }
        \label{fig:trivialization_FM_section}
    \end{subfigure}
    \caption{\small
        Extended diagrams of the frame bundle trivializations which capture the interplay of the transition functions $g^{BA}\cdot$, the right actions $\lhd\,h$ and $\,\cdot\,h$ and the identity sections $\sigma^A$ and $\sigma^B$.
        As before, we abbreviate $U=U^{AB}=U^A\cap U^B$.
        Except for $\sigma^A\circ\piFM \neq \id_{\FM}$ and $\sigma^B\circ\piFM \neq \id_{\FM}$, the diagrams commute.
        If the trivializations are part of some $G$-atlas, similar diagrams, with $\FM$ and $\GL{d}$ being replaced by $\GM$ and $G$, apply to the corresponding $G$-structure.
    }
    \label{fig:trivializations_FM_complete}
\end{figure}

Any atlas
$\big\{ \big(U^X, \PsiTM^X \big) \big\}_{\!X\in\mathfrak{X}}\,$
of the tangent bundle is in one-to-one correspondence to an atlas
$\big\{ \big(U^X, \PsiFM^X \big) \big\}_{\!X\in\mathfrak{X}}\,$
of the frame bundle.
Specifically, given a local trivialization $\PsiTM^A$ of $\TM$, a corresponding local trivialization
\begin{align}\label{eq:trivialization_FM}
    \PsiFM^A: \piFM^{-1}\big(U^A\big)\to U^A\times \GL{d}, \quad
    [e_{i}]_{i=1}^d \mapsto \pig(p,\ \psiFMp^A\big([e_{i}]_{i=1}^d\big) \pig) \,,
\end{align}
of $\FM$, where we abbreviated $p=\piFM\left( [e_{i}]_{i=1}^d\right)$, is induced by defining
\begin{align}\label{eq:trivialization_FM_p}
    \psiFMp^A: \FpM\to \GL{d}, \quad
    [e_{i}]_{i=1}^d \mapsto\, \psiFMp^A \big([e_{i}]_{i=1}^d\big) := \big(\psiTMp^A(e_{i})\big)_{i=1}^d
\end{align}
as a map from tangent frames to invertible $d\!\times\!d$ matrices whose $i$-\emph{th} column is given by $\psiTMp^A(e_{i})\in\R^d$.
As~required for associated bundles, the trivializations of $\TM$ and $\FM$ share the \emph{same transition functions},
\begin{align}\label{eq:transition_functions_FM}
    \psiFMp^B\big([e_{i}]_{i=1}^d\big)
    \ &=\ \big(\psiTMp^B(e_{i}) \big)_{i=1}^d \notag \\
    \ &=\ \big(g_p^{BA} \psiTMp^A(e_{i}) \big)_{i=1}^d \notag \\
    \ &=\ g_p^{BA} \big(\psiTMp^A(e_{i}) \big)_{i=1}^d \notag \\
    \ &=\ g_p^{BA} \psiFMp^A\big([e_{i}]_{i=1}^d \big) \ ,
\end{align}
since the action of $g^{BA}$ on the individual trivialized frame axes in the second line agrees with its action on the trivialized frame matrix in the third line.
Furthermore, as claimed for principal bundles in Eq.~\eqref{eq:right_G_equiv_principal_bdl_general}, the trivializations of the frame bundle are \emph{right $\GL{d}$-equivariant}, that is, for any $h\in \GL{d}$ one has:
\begin{align}\label{eq:right_equivariance_FM}
    \psiFMp^A\big([e_{i}]_{i=1}^d \lhd h\big)
    \ &=\ \psiFMp^A\left(\left(\sum\nolimits_j e_{j}\, h_{ji} \right)_{i=1}^d\right) \notag \\
    \ &=\ \left(\psiTMp^A\left(\sum\nolimits_j e_{j}\, h_{ji} \right)\right)_{i=1}^d \notag \\
    \ &=\ \left(\sum\nolimits_j \psiTMp^A\left(e_{j}\right) h_{ji} \right)_{i=1}^d \notag \\
    \ &=\ \left( \psiTMp^A\left(e_{i}\right) \right)_{i=1}^d  \cdot h \notag \\
    \ &=\ \psiFMp^A\big( [e_{i}]_{i=1}^d \big) \cdot h
\end{align}
Here we used the linearity of $\psiTMp^A$ in the third step and identified the index expression as a right matrix multiplication in the fourth step.
Fig.~\ref{fig:trivialization_FM_non-collapsed} summarizes the left action on the trivialization via transition functions $\PsiFM^B \circ \left(\PsiFM^A\right)^{-1} = (\id\times g^{BA}\cdot)$ as derived in Eq.~\eqref{eq:transition_functions_FM} and the right equivariance $\PsiFM^A\circ(\lhd\, h) = (\id\times\cdot h)\circ\PsiFM^A$ of the trivializations as derived in Eq.~\eqref{eq:right_equivariance_FM}.

As indicated in Eq.~\eqref{eq:framefield_gauge_equivalence} and visualized in Figs.~\ref{fig:gauge_trafos} and~\ref{fig:gauge_trafos_manifold}, a smooth local trivialization $\PsiTM^A$ on $U^A$ of the tangent bundle induces a \emph{frame field} on $U^A$.
It is formalized as a smooth \emph{local section}
\begin{align}\label{eq:section_FM}
    \sigma^A:U^A\to \piFM^{-1}\!\left(U^A\right),\ \ p\mapsto \left[\big(\psiTMp^A\big)^{-1}(\epsilon_i)\right]_{i=1}^d
\end{align}
of the frame bundle, defined by mapping the standard frame vectors $\epsilon_i$ of $\R^d$ back to the tangent spaces in $\piTM^{-1}\big(U^A\big)\subseteq \TM$.
Following Eq.~\ref{eq:frame_rightaction}, a gauge transformation from $\PsiTM^A$ to $\PsiTM^B = (\id\times g^{BA}\cdot)\PsiTM^A$ corresponds to a transformation
\begin{align}\label{eq:section_FM_rightaction}
    \sigma^B(p)\ =\ \sigma^A(p) \lhd \left(g^{BA}_p\right)^{-1}
\end{align}
of sections on $U^{AB}$.
Being defined in terms of $\PsiTM^A,$ the trivializations $\PsiFM^A$ of $\FM$ have the nice property that they map the corresponding sections $\sigma^A$ to the identity frame $e \in \GL{d} \subset \R^{d\times d}$ of $\R^d$, which can be seen by inserting both definitions:
\begin{align}\label{eq:identity_section_prop}
    \psiFMp^A\circ\sigma^A(p)
    \ =\ \psiFMp^A \Big(\Big[ \big(\psiTMp^A\big)^{-1} (\epsilon_i) \Big]_{i=1}^d \Big)
    \ =\ \Big(\psiTMp^A \circ \big(\psiTMp^A\big)^{-1} (\epsilon_i) \Big)_{i=1}^d
    \ =\ (\epsilon_i)_{i=1}^d
    \ =\ e
\end{align}
This property is often used to define sections of $\FM$ given trivializations $\PsiFM^A$ as
\begin{align}\label{eq:identity_section_def}
    \sigma^A\!:U^A\to\piFM^{-1}\big(U^A\big),\ \ \ p \mapsto \big(\PsiFM^A\big)^{-1}(p,e) = \big(\psiFMp^A\big)^{-1}(e) \,,
\end{align}
which ultimately coincides with our definition in Eq.~\eqref{eq:section_FM}.
Since $\sigma^A$ and $\PsiFM^A$ constructed this way imply each other they are sometimes called \emph{identity sections} and \emph{canonical local trivializations}.
Extending the diagram in Fig.~\ref{fig:trivialization_FM_non-collapsed} with identity sections $\sigma^A$ and $\sigma^B$, related by Eq.~\ref{eq:section_FM_rightaction}, fixes $h=g^{BA}$ and thus leads to the commutative diagram in Fig.~\ref{fig:trivialization_FM_section}.
The left and right multiplications with $g^{BA}$ on the typical fiber $\GL{d}$ coincide hereby only since $\psiFMp^A\circ\sigma^A = \psiFMp^B\circ\sigma^B = e$ for which $g^{BA}\cdot e = g^{BA} = e\cdot g^{BA}$.
Compare Fig.~\ref{fig:trivialization_FM_section} to Fig.~\ref{fig:frame_bundle}, which shows the left gauge action $g_p^{BA}\cdot$ on $\GL{d}$ and the right action $\lhd\big( g_p^{BA} \big)^{-1}$ of the inverse group element which transforms between the corresponding identity section frames.

\paragraph{\textit{G}-atlas induced \textit{G}-structure \textit{GM}:}

The agreement of the transition functions of the tangent bundle and the frame bundle in Eq.~\eqref{eq:transition_functions_FM} implies that a $G$-atlas of $\TM$ induces a $G$-atlas for $\FM$.
As we will derive in the following, such $G$-atlases fix a corresponding $G$-structure $\GM$, i.e. a principal $G$-subbundle of $\FM$, consisting of preferred frames.

To motivate the definition of $\GM$ in terms of a given $G$-atlas $\big\{ \big(U^X, \PsiFM^X \big) \big\}_{\!X\in\mathfrak{X}}\,$ of $\FM$, consider two of its local trivializations $\PsiFM^A$ and $\PsiFM^B$ with overlapping domains and let $p \in U^A\cap U^B$.
The trivializations define reference frames $\sigma^A(p)$ and $\sigma^B(p)$ in $\FpM$, which are according to Eq.~\eqref{eq:section_FM_rightaction} related by the right action of some element $g_p^{BA}$ of the reduced structure group $G \leq \GL{d}$.
Any such defined frame is therefore seen to be an element of a $G$-orbit $\GpM \cong G$ in $\FpM \cong \GL{d}$.
Specifically, expressing the identity sections via Eq.~\eqref{eq:identity_section_def} as $\sigma^A(p) = \big( \psiFMp^A \big)^{-1} (e)$ and
$
\sigma^B(p)
= \big( \psiFMp^B \big)^{-1} (e)
= \big( g_p^{BA} \psiFMp^A \big)^{-1} (e)
= \big( \psiFMp^A \big)^{-1} \pig( \big( g_p^{BA} \big)^{-1} \pig)
$
suggests the pointwise definition of the $G$-structure in terms of inverse images of~$G$ by (arbitrary) gauge maps:
\begin{align}\label{eq:G_atlas_induced_G_structure_GM_def_ptwise}
    \GpM\ :=\ \pig\{ \big(\psiFMp^A \big)^{-1}(g) \;\pig|\; g\in G\, \pig\} \ =\ \big( \psiFMp^A \big)^{-1} (G)
\end{align}
The independence from the chosen gauge of the $G$-atlas is clear as any other choice
$
  \big( \psiFMp^B \big)^{-1} (G)
= \big( \psiFMp^A \big)^{-1} \pig( \big(g_p^{BA}\big)^{-1} G \pig)
= \big( \psiFMp^A \big)^{-1} (G)
$
would yield the same result.
As one can easily check, $\GpM$ is indeed a right $G$-torsor since~$G$ is a right $G$-torsor and $\psiFMp^A$ is by Eq.~\eqref{eq:right_equivariance_FM} a right $\GL{d}$-equivariant -- and thus in particular right $G$-equivariant -- isomorphism.
The required smoothness of $\GM = \coprod_{p\in M} \GpM$ follows from the smoothness of the trivializations $\PsiFM^A$.

A $G$-atlas of local trivializations of $\GM$ is given by restricting the trivializations in the $G$-atlas of $\FM$ to frames in $\GM$, that is,
\begin{align}
    \PsiGM^A := \PsiFM^A \big|_{\piGM^{-1}(U^A)} :\ \ \piGM^{-1} \big(U^A\big) \to U^A \times G \,,
\end{align}
or, locally,
\begin{align}
    \psiGMp^A := \psiFMp^A \big|_{\GpM} :\ \ \GpM \to G \,.
\end{align}
It follows immediately that the $G$-valued transition functions agree with those of $\TM$ and $\FM$, that is,
\begin{align}\label{eq:transition_functions_GM}
    \psiGMp^B\big([e_{i}]_{i=1}^d\big)
    \ =\ g_p^{BA} \psiGMp^A\big([e_{i}]_{i=1}^d \big) \,,
\end{align}
and that the trivializations are right $G$-equivariant:
\begin{align}\label{eq:right_equivariance_GM}
    \psiGMp^A\big([e_{i}]_{i=1}^d \lhd h\big)
    \ =\ \psiGMp^A\big( [e_{i}]_{i=1}^d \big) \cdot h \qquad \forall h \in G
\end{align}
The frame fields are also given by an equivalent expression
\begin{align}\label{eq:GM_section_psi_inverse_def}
    \sigma^A(p)\ =\ \big( \psiGMp^A \big)^{-1}(e)
\end{align}
to that in Eq.~\eqref{eq:identity_section_def}.
The commutative diagrams in Figs.~\ref{fig:trivialization_FM_non-collapsed} and~\ref{fig:trivialization_FM_section} hold as well when replacing $\FM$ with $\GM$ and $\GL{d}$ with $G$.

\paragraph{Induced trivializations of associated bundles $\A$:}

A $G$-atlas
$\big\{ \big(U^X, \PsiA^X \big) \big\}_{\!X\in\mathfrak{X}}\,$,
consisting of local trivializations
$\PsiA^X:\piA^{-1}\big(U^X\big)\to U^X\times\R^c$
of the associated feature vector bundles
$\A=(\GM\times\R^c)/\!\sim_{\!\rho}$
is induced from the corresponding trivializations $\PsiGM^X$ of the $G$-structure.
In order to construct these trivializations, recall that $\A$ is defined in terms of equivalence classes
$\big[ [e_{i}]_{i=1}^d,\ \mathscr{f}\,\big]$
consisting of pairs of reference frames and feature coefficient vectors which are related by the equivalence relation $\sim_{\!\rho}$ defined in Eq.~\eqref{eq:equiv_relation_A}.
A natural idea is thus to trivialize
$\big[ [e_{i}]_{i=1}^d,\ \mathscr{f}\,\big]\in\A_p$
by picking one representative of its equivalent coefficient vectors in~$\R^c$.
A preferred choice of representative is hereby given by that coefficient vector belonging to the identity section frame $\sigma^A(p)$ corresponding to $\PsiGM^A$.

Let $[e_{i}]_{i=1}^d := \sigma^A(p)\lhd h\ \in \GpM$ be some frame that is defined by an offset $h\in G$ relative to section~$\sigma^A$.
This offset can be recovered by the trivialization of the $G$-structure:
\begin{align}
    \psiGMp^A \!\left([e_{i}]_{i=1}^d\right)
    \ =\ \psiGMp^A \!\left( \sigma^A(p)\lhd h \right)
    \ =\ \psiGMp^A \!\left( \sigma^A(p) \right) \cdot h
    \ =\ h
\end{align}
Here we used the right $G$-equivariance of $\psiGMp^A$ and that $\sigma^A$ is defined as identity section; see Eqs.~\eqref{eq:right_equivariance_GM} and~\eqref{eq:identity_section_prop}, the latter adapted to $\psiGMp^A$.
We can therefore rewrite any frame via its offset as:
\begin{align}
    [e_{i}]_{i=1}^d
    \ =\ \sigma^A(p) \lhd \psiGMp^A \!\left([e_{i}]_{i=1}^d\right)
\end{align}
Similarly, we can rewrite any feature vector $\big[ [e_{i}]_{i=1}^d,\ \mathscr{f}\,\big]\in\A_p$ by different representatives of the equivalence class:
\begin{align}
    \left[ [e_{i}]_{i=1}^d,\ \mathscr{f}\,\right]
    \ =\ \left[ \sigma^A(p) \lhd \psiGMp^A\big([e_{i}]_{i=1}^d\big),\,\ \mathscr{f}\,\right]
    \ =\ \left[ \sigma^A(p),\ \rho\left(\psiGMp^A\big([e_{i}]_{i=1}^d\big)\right) \mathscr{f}\,\right]
\end{align}

Based on these insights we define induced trivializations of $\A$ by setting
\begin{align}\label{eq:trivialization_A}
    \PsiA^A: \piA^{-1}\big(U^A\big)\to U^A\times\R^c,\quad
    \big[[e_{i}]_{i=1}^d,\ \mathscr{f}\,\big]\ \mapsto\ 
    \Big(\piGM \big([e_{i}]_{i=1}^d \big),\ \psiAp^A \pig(\big[ [e_{i}]_{i=1}^d,\ \mathscr{f}\,\big]\pig) \Big) \ ,
\end{align}
with
\begin{align}\label{eq:trivialization_A_p}
    \psiAp^A:\A_p\to\R^c,\quad
    \big[[e_{i}]_{i=1}^d,\ \mathscr{f}\,\big]
    \, =\, \Big[\sigma^A(p),\ \rho\pig( \psiGMp^A \big([e_{i}]_{i=1}^d \big)\pig) \mathscr{f}\,\Big]
    \ \mapsto\ \rho\pig(\psiGMp^A \big([e_{i}]_{i=1}^d \big)\pig)\, \mathscr{f} \,,
\end{align}
which picks that particular representative coefficient vector
$f^A = \rho\left(\psiGMp^A \left([e_{i}]_{i=1}^d\right)\right) \mathscr{f}\in\R^c$
that is distinguished by the reference frame $\sigma^A(p)$ corresponding to the chosen gauge.
For later convenience we note that this implies in particular that the inverse of Eq.~\eqref{eq:trivialization_A_p} is given by
\begin{align}\label{eq:trivialization_A_p_inv}
    \big(\psiAp^A\big)^{-1}\!: \R^c \to \A_p:\ \ 
    \mathscr{f} \,\mapsto \big[\sigma^A(p),\; \mathscr{f}\,\big] \ .
\end{align}
The such defined trivialization is independent of the chosen representative since for any $k\in G$ we have:
\begin{align}
    \psiAp^A \pig(\big[ [e_{i}]_{i=1}^d \!\lhd k^{-1},\,\ \rho(k)\mathscr{f} \,\big]\pig)
    \ &=\ \rho\big(\psiGMp^A \big([e_{i}]_{i=1}^d \!\lhd k^{-1} \big)\big) \rho(k)\mathscr{f} \notag\\
    \ &=\ \rho\big(\psiGMp^A \big([e_{i}]_{i=1}^d \big) \cdot k^{-1}\big) \rho(k)\mathscr{f} \notag\\
    \ &=\ \rho\big(\psiGMp^A \big([e_{i}]_{i=1}^d \big)\big) \mathscr{f} \notag\\
    \ &=\ \psiAp^A \pig(\big[ [e_{i}]_{i=1}^d,\,\ \mathscr{f}\,\big]\pig)
\end{align}
By construction, the transition functions are given by $\rho\big(g_p^{BA}\big)$:
\begin{align}\label{eq:transition_fct_A}
    \psiAp^B \pig(\big[ [e_{i}]_{i=1}^d,\,\ \mathscr{f}\, \big]\pig)
    \ &=\ \rho\big(\psiGMp^B \big([e_{i}]_{i=1}^d \big)\big) \mathscr{f} \notag\\
    \ &=\ \rho\big(g_p^{BA}\psiGMp^A \big([e_{i}]_{i=1}^d \big)\big) \mathscr{f} \notag\\
    \ &=\ \rho\big(g_p^{BA}\big) \rho\big(\psiGMp^A \big([e_{i}]_{i=1}^d \big)\big) \mathscr{f} \notag\\
    \ &=\ \rho\big(g_p^{BA}\big) \psiAp^A \big(\big[ [e_{i}]_{i=1}^d,\,\ \mathscr{f}\,\big]\big)
\end{align}
If the tangent bundle is taken as a $G$-associated vector bundle $\TM\cong(\GM\times\R^d)/G$, its trivializations are recovered from Eq.~\eqref{eq:trivialization_A} for the specific choice $\rho(g)=g$.

\begin{figure}
    \centering
    \begin{tikzcd}[row sep=4.em, column sep=7.em, crossing over clearance=.6ex,
                   execute at end picture={
                        \node [] at (-4.27, -1.1) {$\noncommutative$};
                        }]
        & U\times \R^c  \arrow[r, "\proj_2"]
        &[7ex] \R^c
        \\
          \piA^{-1}(U)  \arrow[d, "\piA"] \arrow[r, "\PsiA^A"] \arrow[ru, "\PsiA^B"]
        & U\times \R^c  \arrow[u, swap, "\big(\id\times \rho\big(g^{BA}\big)\cdot\big)"]
                        \arrow[ld, "\proj_1"']
                        \arrow[r, "\proj_2"']
        & \R^c          \arrow[u, "\rho\big(g^{BA}\big)\cdot"']
        \\
            U           \arrow[u, bend left=25, shift left=.6, "f"]
                        \arrow[rru, bend right=13, "f^A"']
                        \arrow[rruu, bend right=12, "f^B"', crossing over]
    \end{tikzcd}
    \caption{\small
        Coordinate free feature fields are defined as global sections $f\in\Gamma(\A)$.
        On local neighborhoods $U^A$ and $U^B$ they trivialize to fields of feature coefficient vectors $f^A:U^A\mapsto\R^c$ and $f^B:U^B\mapsto\R^c$ which are on $U=U^A\cap U^B$ related by $f^B(p)=\rho\big(g_p^{BA}\big)f^A(p)$.
        Except for $f\circ\piA \neq \id_{\A}$, the diagram commutes.
    }
    \label{fig:trivialization_A_sections}
\end{figure}
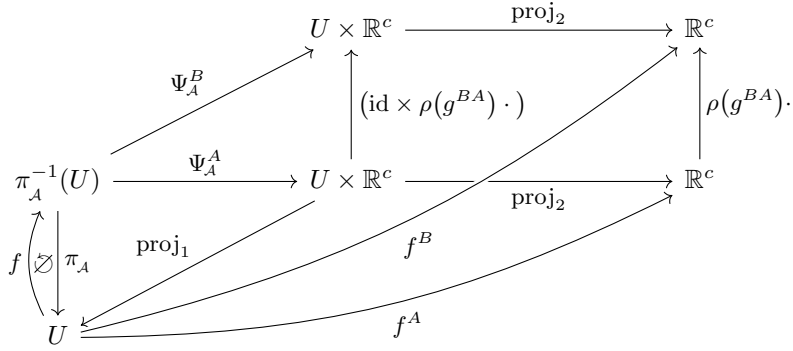

Assume a coordinate free feature field $f\in\Gamma(\A)$ to be given.
Relative to gauge $\PsiA^A$, it can be locally represented as a coefficient vector field $f^A:U^A\to\R^c$ by defining
\begin{align}
    f^A\ :=\ \proj_2 \circ\mkern2mu \PsiA^A \circ f
\end{align}
which is equivalent to the pointwise definition
\begin{align}
    f^A(p)\ =\ \psiAp^A \circ f(p) \,.
\end{align}
As apparent from the commutative diagram in Fig.~\ref{fig:trivialization_A_sections}, the transition functions in Eq.~\eqref{eq:transition_fct_A} carry over to the local coefficient fields such that we get
\begin{align}
    f^B(p)\ =\ \rho\left(g_p^{BA}\right) f^A(p)
\end{align}
for $p\in U^A\cap U^B$.
This agrees with and justifies our definition of the gauge transformations of feature coefficient vectors in Eq.~\eqref{eq:gauge_trafo_features}.

\paragraph{Summarizing remarks:}

The here defined local trivializations and transition functions formalize and justify the definitions of gauges and gauge transformations from Section~\ref{sec:feature_fields}.
Local trivializations of $\TM$ and $\FM$ were shown to induce each other.
If a $G$-atlas is chosen for either of both, it defines a $G$-structure $\GM$, whose $G$-atlas essentially coincides with that of $\FM$.
It furthermore induces a $G$-atlas for any other associated bundle, including $\A$.
As visualized in Fig.~\ref{fig:trivializations_TM_FM_A}, the transition functions of all $G$-atlases for $\TM$, $\FM$, $\GM$ and $\A$ agree, making the bundles $G$-associated to each other.
Specifically, when switching from gauge $A$ to gauge $B$, the trivializations of $\TM$, $\FM$ and $\GM$ transform according to a left multiplication with $g^{BA}$ while the feature vector bundle trivializations transform according to a left multiplication with $\rho\big(g^{BA}\big)$; see Eqs.~\eqref{eq:transition_fct_TM_gAB}, \eqref{eq:transition_functions_FM}, \eqref{eq:transition_functions_GM} and~(Eq.~\eqref{eq:transition_fct_A}).
At the same time, frame fields transform according to the right action $\lhd \big( g^{BA} \big)^{-1}$ (Eq.~\eqref{eq:section_FM_rightaction}).

%% file: chapters/65_bundle_transport.tex

\subsection{Parallel transporters on associated bundles}
\label{sec:bundle_transport}

Section~\ref{sec:transport_local} gave an intuitive introduction to the parallel transport of tangent vectors and feature vectors along a path~$\gamma$ from $q\in M$ to $p\in M$.
Here we briefly discuss how coordinate free parallel transporters on the fiber bundles induce each other and derive coordinate expressions relative to given trivializations for them.
We start by assuming coordinate free transporters
\begin{alignat}{2}
    \PTMgamma:&\ &\TqM &\to \TpM \\
\intertext{
    on the tangent bundle $\TM$ to be \emph{given} and explain how they \emph{induce} transporters
}
    \PFMgamma:&\ &\FqM &\to \FpM \\
\intertext{
    on the frame bundle $\FM$.
    If these transporters are $G$-\emph{compatible} with the chosen $G$-structure, as discussed below, they further induce transporters
}
    \PGMgamma:&\ &\GqM &\to \GpM \\
    \PAgamma:&\  &\A_q &\to \A_p
\end{alignat}
on the associated $G$-bundles $\GM$ and $\A$.
In practice, that is, in our literature review in Part~\ref{part:literature_review}, most convolutional networks assume either transporters that are based on the Levi-Civita connection or some trivial connection.

A more formal definition of bundle transporters might take a different route, starting by introducing a so called principal Ehresmann connection on the principal $G$-bundle $\GM$ (which would by definition be $G$-compatible).
Such an Ehresmann connection can either be defined by a choice of horizontal subbundle $HGM$ of the tangent bundle $TGM$ of $\GM$ or, equivalently, by a Lie algebra-valued connection 1-form $\omega:TGM\to\mathfrak{g}$ on $\GM$.
The transport on $\GM$ would subsequently be defined via the horizontal lift $\gamma^\uparrow:[0,1]\to \GM$ of curves $\gamma:[0,1]\to M$ on the base space such that the tangent vectors of the lift in $\GM$ are horizontal, i.e.~$\dot{\gamma}^\uparrow \in HGM$.
All transporters on $\TM$, $\FM$ and $\A$ as associated $G$-bundles would then be induced from the transporters on the $G$-structure.
Instead of following this formal approach, which would be rather technical and can be found in the literature~\cite{schullerGeometricalAnatomy2016,wendlLectureNotesBundles2008,husemollerFibreBundles1994a,nakahara2003geometry,marshGaugeTheoriesFiber2016,shoshichikobayashiFoundationsDifferentialGeometry1963},
we focus on how the different transporters interrelate by inducing each other.

\paragraph{Transport on \textit{TM}:}
To this end, we take a shortcut by assuming the coordinate free transporters $\PTMgamma$ on $\TM$ to be given.
Recall that, given gauges $\PsiTM^{\widetilde{A}}$ on a neighborhood $U^{\widetilde{A}}$ of $q$ and $\PsiTM^A$ on a neighborhood $U^A$ of $p$, the tangent vector transporter is coordinatized according to Eq.~\eqref{eq:transporter_gauge}, that is,
\begin{align}\label{eq:transporter_gauge_copy}
    g_\gamma^{A\widetilde{A}} \ :=\ \psiTMp^A \circ \PTMgamma \circ \Big( \psiTMq^{\widetilde{A}} \Big)^{-1} \ \in\, \GL{d} \,,
\end{align}
and that its coordinatizations transform under gauge transformations at $q$ and $p$ according to Eq.~\eqref{eq:transporter_gauge_trafo}:
\begin{align}\label{eq:transporter_gauge_trafo_copy}
    g_\gamma^{B\widetilde{B}} \ =\ g_p^{BA}\, g_\gamma^{A\widetilde{A}} \Big(g_q^{\widetilde{B}\widetilde{A}}\Big)^{-1}
\end{align}
We refer back to Eq.~\eqref{cd:transporter_trivialization} for a visualization of these definitions in terms of a commutative diagram.

\paragraph{Transport on \textit{FM}:}
Given the transporter on the tangent bundle, the transporter on the frame bundle follows immediately from the transport of individual frame axes.
In equations, let $[e_i]_{i=1}^d \in \FqM$ be a frame at $q$, then the individual axes $e_i$ for $i=1,\dots,d$ are tangent vectors in $\TqM$ which can be transported via $\PTMgamma$.
We thus define the transporter on the frame bundle as:%
\footnote{
    The transport of a frame along $\gamma$ describes a curve $\gamma^\uparrow$ (horizontal lift) in $\FM$.
    The space spanned by all tangent vectors $\dot{\gamma}^\uparrow$ in $TFM$ along such curves is the horizontal subbundle $HFM$ of $TFM$, mentioned above.
}
\begin{align}\label{eq:transporter_FM_def}
  \PFMgamma\!:\ \FqM \to \FpM, \quad
  [e_i]_{i=1}^d \mapsto \PFMgamma\big([e_i]_{i=1}^d\big) := \big[\PTMgamma(e_i)\big]_{i=1}^d
\end{align}
In order to derive the explicit form of its coordinatization
$\psiFMp^A \circ \PFMgamma \circ \big(\psiFMq^{\widetilde{A}}\big)^{-1}\! \in \GL{d}$,
consider its action on a group element $h\in \GL{d}$, representing a trivialized frame of $\R^d$ which is spanned by the matrix columns $h_{:,i}\in\R^d,\ i=1,\dots,d$\,:
\begin{alignat}{3}\label{eq:transporter_gauge_FM}
    \Big[ \psiFMp^A \circ \PFMgamma \circ \big(\psiFMq^{\widetilde{A}}\big)^{-1} \Big](h)
    \ &=\ \Big[ \psiFMp^A \circ \PFMgamma \Big] \Big(\! \big[\big(\psiTMq^{\widetilde{A}}\big)^{-1}(h_{:,i})\big]_{i=1}^d \Big)
        \qquad && \big( \text{\small def. of $\psiFMp^{\widetilde{A}}$, Eq.~\eqref{eq:trivialization_FM_p}} \big) \notag \\
    \ &=\ \psiFMp^A \Big( \big[\PTMgamma \circ \big(\psiTMq^{\widetilde{A}}\big)^{-1}(h_{:,i})\big]_{i=1}^d \Big)
        \qquad && \big( \text{\small def. of $\PFMgamma$, Eq.~\eqref{eq:transporter_FM_def}} \big) \notag \\
    \ &=\ \Big( \psiTMp^A \circ \PTMgamma \circ \big(\psiTMq^{\widetilde{A}}\big)^{-1}(h_{:,i}) \Big)_{i=1}^d
        \qquad && \big( \text{\small def. of $\psiFMp^{\widetilde{A}}$, Eq.~\eqref{eq:trivialization_FM_p}} \big) \notag \\
    \ &=\ \Big( g_\gamma^{A\widetilde{A}} (h_{:,i}) \Big)_{i=1}^d
        \qquad && \big( \text{\small triv. of $\PTMgamma$, Eq.~\eqref{eq:transporter_gauge_copy}} \big) \notag \\
    \ &=\ g_\gamma^{A\widetilde{A}} \, h
\end{alignat}
The coordinatizations of the frame transporters are therefore equivalent to those of the tangent vector transporters in Eq.~\eqref{eq:transporter_gauge_copy} but act on trivialized frames in $\GL{d}$ instead of acting on coefficient vectors in $\R^{d}$.
Their gauge transformations are from the commutative diagram
\begin{equation}\label{cd:FM_transport_trivialization}
\begin{tikzcd}[column sep=53pt, row sep=30, font=\normalsize]
    \GL{d}
        \arrow[dd, "g_q^{\widetilde{B}\widetilde{A}}\cdot\ "']
        \arrow[rrr, "g_\gamma^{A\widetilde{A}}\cdot"]
    & &[-3ex] &
    \GL{d}
        \arrow[dd, "\ g_p^{BA}\cdot"]
    \\
    &
    \FqM
        \arrow[ul, "\psiFMq^{\widetilde{A}}", pos=.45]
        \arrow[dl, "\psiFMq^{\widetilde{B}}"', pos=.45]
        \arrow[r, "\PFMgamma"]
    &
    \FpM
        \arrow[ur, "\psiFMp^A"', pos=.45]
        \arrow[dr, "\psiFMp^B", pos=.45]
    \\
    \GL{d}
        \arrow[rrr, "g_\gamma^{B\widetilde{B}}\cdot"']
    & & &
    \GL{d}
\end{tikzcd}
\quad
\end{equation}
seen to coincide with those of the coordinatized transporters on $\TM$ in Eq.~\eqref{eq:transporter_gauge_trafo_copy}.

\paragraph{Compatibility of connections and \textit{G}-structures:}

Not any choice of connection or definition of transporters on the $\GL{d}$-bundles $\TM$ and $\FM$ is compatible with any $G$-structure.
Specifically, a $G$-structure might not be closed under the transport of frames, that is,
while a frame in $\GqM \subseteq \FqM$ will by $\PFMgamma$ be transported to some frame in $\FpM$, this frame is \emph{not} necessarily contained in $\GpM$.%
\footnote{
    In terms of a principal Ehresmann connection on $\FM$, this is the case if the horizontal subbundle $HFM \subseteq TFM$ is not contained in $TGM \subseteq TFM$.
    An immediate definition of parallel transport in terms of a choice of horizontal subbundle $HGM$ on the $G$-structure will always (by definition) lead to a well defined transport on $\GM$.
}
Relative to trivializations of $\GM$, such an incompatibility would reflect in coordinatized transporters ${g_\gamma^{A\widetilde{A}} \notin G}$, whose left multiplication is well defined on the fibers $\R^d$ and $\GL{d}$ of the $\GL{d}$-bundles $\TM$ and $\FM$, but not on the fiber $G$ of~$\GM$.
If the subbundle $\GM$ is not closed under the parallel transport on $\FM$, this means that no well defined corresponding transport on $\GM$ -- and thus on any associated $G$-bundles $\A$ -- exists.

As an example, consider the Levi-Civita connection on Euclidean spaces, whose transporters keep tangent vectors and frames parallel in the usual sense on~$\Euc_d$.
The $\{e\}$-structure (frame field) in Fig.~\ref{fig:frame_field_automorphism_1} is closed under this transport, and therefore compatible.
The $\{e\}$-structure in Fig.~\ref{fig:frame_field_automorphism_2}, on the other hand, is not closed under the transport, and thus incompatible with the Levi-Civita connection.
Similarly, the $\SO2$-structure on $S^2$ in Fig.~\ref{fig:G_structure_S2_1} is compatible with the Levi-Civita connection on the sphere, while the $\{e\}$-structure in Fig.~\ref{fig:G_structure_S2_2} is not.

The reader might wonder which general statements about the compatibility of connections (or transporters) and $G$-structures can be made.
In general, the Levi-Civita connection, or any other metric connection, are compatible with the $\O{d}$-structure $\OM$ that corresponds to the metric.%
\footnote{
    This statement holds by definition since metric connections preserve angles and lengths between vectors and thus the orthonormality of frames.
    One can furthermore \emph{define} metric connections as principal Ehresmann connections on~$\OM$.
}
If the manifold is orientable, the Levi-Civita connection is furthermore compatible with any $\SO{d}$-structure that corresponds to the metric.
An example is the $\SO2$-structure on $S^2$ in Fig.~\ref{fig:G_structure_S2_1}.
A necessary (but not sufficient) condition for a $G$-structure to be compatible with a given connection is that the holonomy group of the connection is a subgroup of the structure group~$G$.

An important special case is that of $\{e\}$-structures, since they imply a \emph{unique trivial connection}.%
\footnote{
    A connection is \emph{trivial} if its holonomy group, i.e. its parallel transport around any closed loop, is trivial~\cite{craneTrivialConnectionsDiscrete2010}.
}%
\footnote{
    Only one principal Ehresmann connection $H\eM = T\eM$ can be chosen on $\eM$ since the vertical subbundle $V\eM$ is the zero-section of $T\eM$.
}
The corresponding transporters move frames in such a way that the stay parallel with the frames of the $\{e\}$-structure.
Trivial connections might not seem to be of particular importance for the theory of $\GM$-convolutions, however, they are actually utilized by many convolutional networks.
Specifically, any network that relies on an $\{e\}$-structure is implicitly assuming a trivial connection.
This includes all of the models in Table~\ref{tab:network_instantiations} with $G=\{e\}$, specifically those which are reviewed in Sections~\ref{sec:spherical_CNNs_azimuthal_equivariant} and~\ref{sec:e_surface_conv}.%
\footnote{
    These models are \emph{implicitly} assuming a trivial connection by not modeling non-trivial transporters of feature vectors:
    they accumulate feature vector coefficients without transforming them.
}
Note that these models assume the trivial connection only for their feature vector transport but compute geodesics for the transporter pullback, Eq.~\eqref{eq:transporter_pullback_in_coords}, based on the original Levi-Civita connection.

\paragraph{Transport on \textit{GM}:}
Assuming that $\GM$ is compatible with (i.e. closed under) the transport on $\FM$, a well defined transporter is given by restricting the frame bundle transporter to the $G$-structure:
\begin{align}\label{eq:transporter_GM_def}
  \PGMgamma := \PFMgamma \big|_{\scalebox{.62}{$\GM$}}:\ \ \GqM \to \GpM
\end{align}
The transition functions between different coordinatizations of $\PGMgamma$ do then agree with those of $\PFMgamma$ and thus also $\PTMgamma$.
We obtain the following commutative diagram, which visualizes the restriction of the diagram in Eq.~\eqref{cd:FM_transport_trivialization} from $\FqM$, $\FpM$ and $\GL{d}$ to $\GqM$, $\GpM$ and $G$:
\begin{equation}\label{cd:transport_GM_triv}
\begin{tikzcd}[column sep=60pt, row sep=30, font=\normalsize]
    G
        \arrow[dd, "g_q^{\widetilde{B}\widetilde{A}}\cdot\ "']
        \arrow[rrr, "g_\gamma^{A\widetilde{A}}\cdot"]
    & &[-3ex] &
    G
        \arrow[dd, "\ g_p^{BA}\cdot"]
    \\
    &
    \GqM
        \arrow[ul, "\psiGMq^{\widetilde{A}}", pos=.45]
        \arrow[dl, "\psiGMq^{\widetilde{B}}"', pos=.45]
        \arrow[r, "\PGMgamma"]
    &
    \GpM
        \arrow[ur, "\psiGMp^A"', pos=.45]
        \arrow[dr, "\psiGMp^B", pos=.45]
    \\
    G
        \arrow[rrr, "g_\gamma^{B\widetilde{B}}\cdot"']
    & & &
    G
\end{tikzcd}
\quad
\end{equation}
We will in the remainder of this work assume that the transport on $\GM$ is well defined.

\paragraph{Transport on $\boldsymbol{\A}$:}
If the transporters of a connection are well defined on $\GM$, they induce transporters on any associated $G$-bundle, including the feature vector bundles $\A=(\GM\times\R^c)/\!\sim_\rho$.
Let $f_q := \big[[e_i]_{i=1}^d,\,\mathscr{f}\:\!\big]$ be a coordinate free feature vector in $\A_q$.
Its parallel transport is given by that equivalence class defined by keeping some representative coefficients $\mathscr{f}\in \R^c$ fixed and transporting the corresponding frame $[e_i]_{i=1}^d$:
\begin{align}\label{eq:transporter_A_def}
    \PAgamma: \A_q &\to \A_p, \quad
    f_q \,\mapsto\, \PAgamma(f_q) \,:=\, \pig[\PGMgamma\big([e_i]_{i=1}^d),\: \mathscr{f} \,\pig]
\end{align}
In Section~\ref{sec:transport_local} we claimed that the transporter of numerical feature vector coefficients is given by $\rho\big(g_\gamma^{A\widetilde{A}}\big)$ provided that $g_\gamma^{A\widetilde{A}}\in G$, which is the case if the transport on $\GM$ is well defined.
This coordinate expression of $\PAgamma$ can be derived by evaluating the action of
$\psiAp^A \circ \PAgamma \circ \big(\psiAq^{\widetilde{A}}\big)^{-1} \in\, \rho(G)\, \leq\, \GL{c}$
on a feature coefficient vector $\mathscr{f}\in\R^c$ step by step:

\begin{alignat}{3}\label{eq:transporter_gauge_A}
    \Big[\psiAp^A \circ \PAgamma \circ \big(\psiAq^{\widetilde{A}}\big)^{-1}\Big] (\mathscr{f})
    \ &=\ \Big[\psiAp^A \circ \PAgamma\Big] \big(\big[\sigma^{\widetilde{A}}(q),\, \mathscr{f}\,\big]\big)
        \qquad && \big( \text{\small def. of $\big(\psiAp^{\widetilde{A}}\big)^{-1}$, Eq.~\eqref{eq:trivialization_A_p_inv}} \big) \\
    \ &=\ \psiAp^A  \Big(\big[\PGMgamma\big(\sigma^{\widetilde{A}}(q)\big),\, \mathscr{f}\,\big]\Big)
        \qquad && \big( \text{\small def. of $\PAgamma$, Eq.~\eqref{eq:transporter_A_def}} \big) \notag \\
    \ &=\ \rho\Big(\psiGMp^A \circ \PGMgamma \circ \sigma^{\widetilde{A}}(q)\Big) \cdot \mathscr{f}
        \qquad && \big( \text{\small def. of $\psiAp^A$, Eq.~\eqref{eq:trivialization_A_p}} \big) \notag \\
    \ &=\ \rho\Big(\psiGMp^A \circ \PGMgamma \circ \big(\psiGMq^{\widetilde{A}}\big)^{-1}(e)\Big) \cdot \mathscr{f}
        \qquad && \big( \text{\small def. of identity section $\sigma^{\widetilde{A}}$, Eq.~\eqref{eq:identity_section_def}} \big) \notag \\
    \ &=\ \rho\big(g_\gamma^{A\widetilde{A}}\big) \!\cdot\! \mathscr{f}
        \qquad && \big( \text{\small $\PGMgamma$ in coordinates Eq.~\eqref{cd:transport_GM_triv}} \big) \notag
\end{alignat}
The commutative diagram
\begin{equation}
\begin{tikzcd}[column sep=65pt, row sep=30, font=\normalsize]
    \R^c
        \arrow[dd, "\rho\big(g_q^{\widetilde{B}\widetilde{A}}\big)\ "']
        \arrow[rrr, "\rho\big(g_\gamma^{A\widetilde{A}}\big)"]
    & &[-3ex] &
    \R^c
        \arrow[dd, "\ \rho\big(g_p^{BA}\big)"]
    \\
    &
    \A_q
        \arrow[ul, "\psiAq^{\widetilde{A}}"]
        \arrow[dl, "\psiAq^{\widetilde{B}}"']
        \arrow[r, "\PAgamma"]
    &
    \A_p
        \arrow[ur, "\psiAp^A"']
        \arrow[dr, "\psiAp^B"]
    \\
    \R^c
        \arrow[rrr, "\rho\big(g_\gamma^{B\widetilde{B}}\big)"']
    & & &
    \R^c
\end{tikzcd}
\end{equation}
implies that the gauge transformations of the coordinatized feature vector transporters read:
\begin{align}
    \rho\big(g_\gamma^{B\widetilde{B}}\big)
    \ =\
    \rho\big(g_p^{BA}\big)
    \rho\big(g_\gamma^{A\widetilde{A}}\big)
    \rho\big(g_q^{\widetilde{B}\widetilde{A}}\big)^{-1}
\end{align}
Note that this transformation law is in agreement with that in Eq.~\eqref{eq:transporter_gauge_trafo_copy}.

%% file: chapters/70_global_conv_intro.tex

\section{Coordinate free kernel field transforms and \textit{GM}-convolutions}
\label{sec:gauge_CNNs_global}

The associated $G$-bundles introduced in Section~\ref{sec:bundles_fields} allow to describe feature fields -- and therefore convolutional networks -- on a global level.
Given a sequence
${\A_0\! \xrightarrow{\scalebox{.85}{$\pi_{{\scalebox{.65}{$\!\!\A_0$}}}$}}\! M ,}\ \dots,\ 
 {\A_N\! \xrightarrow{\scalebox{.85}{$\pi_{{\scalebox{.65}{$\!\!\A_N$}}}$}}\! M}$,
of $G$-associated feature vector bundles over $M$, we describe coordinate free convolutional networks as sequences
\begin{align}
    \Gamma(\A_0)\, \xrightarrow{\ \ L_1\ \ }\, \Gamma(\A_1)\, \xrightarrow{\ \ L_2\ \ }\ \ \dots\ \ \xrightarrow{\ \ L_N\ \ }\, \Gamma(\A_N)
\end{align}
of parameterized layers $L_1,\dots, L_N$ which map between the feature spaces $\Gamma(\A_0), \dots, \Gamma(\A_N)$, i.e. between spaces of feature fields modeled by the corresponding bundles.
While the field types (or transformation laws) ${\rho_i:G\to\GL{{c_i}}}$ of the intermediate bundles
$\A_i := (\GM\times\R^{c_i})/\!\sim_{\!\rho_i}$ for $i=1,\dots,N-1$
have to be specified by the user as a hyperparameter, the field types $\rho_0:G\to\GL{{c_0}}$ and $\rho_N:G\to\GL{{c_N}}$ of the network input and output are typically determined by the learning task.
The modular construction of neural networks allows to restrict attention to individual layers, mapping between feature spaces $\Gamma(\Ain)$ and $\Gamma(\Aout)$ of dimensionality $\cin$ and $\cout$ and type $\rhoin$ and $\rhoout$.

\etocsettocdepth{3}
\etocsettocstyle{}{} 
\localtableofcontents

The main goal of this section is to introduce coordinate free $\GM$-convolutions, which are the central building blocks of $\GM$-coordinate independent networks on Riemannian manifolds.
To get started, and to introduce concepts which are required later on, we will in Section~\ref{sec:onexone} first focus on the simpler case of \onexoneGMsit, which apply point-like kernels.
Section~\ref{sec:global_conv} shifts the focus to $\GM$-convolutions and kernel field transforms with spatially extended kernels.
They are parameterized in terms of smooth, global \emph{kernel fields}, which are introduced in Section~\ref{sec:kernel_fields}.
$\GM$-\emph{convolutional kernel fields} are required to share weights between different spatial positions.
In order for this weight sharing to be $\GM$-coordinate independent, the template kernels that parameterize $\GM$-convolutional kernel fields are required to be $G$-steerable (Eq.~\eqref{eq:kernel_constraint_rhohom}).
The actual kernel field transforms and $\GM$-convolutions are introduced in Section~\ref{sec:KFTs_GM-conv_global}.
Their global definition is guided by replacing the local coordinate expressions from Section~\ref{sec:gauge_conv_main} with their global, coordinate free counterparts.
As shown in Section~\ref{sec:KFTs_GM-conv_local}, these coordinate free definitions reduce in local trivializations to the coordinate expressions form Section~\ref{sec:gauge_conv_main}.

%% file: chapters/71_global_onexone.tex

\subsection[\texorpdfstring{${1\kern-2.7pt\times\kern-2.9pt1}$}{1x1} \textit{GM}-convolutions]%
           {\texorpdfstring{$\bm{1\kern-2.7pt\times\kern-2.9pt1}$}{1x1} \textit{GM}-convolutions}
\label{sec:onexone}

\onexoneGMs\ map input feature fields $\fin\in\Gamma(\Ain)$ to output feature fields $\fout\in\Gamma(\Aout)$ by linearly mapping each individual input feature vector $\fin(p)\in\Ainp\cong\R^{\cin}$ to an output feature vector $\fout(p)\in\Aoutp\cong\R^{\cout}$ at the same location $p\in M$.
The convolutional character is implemented by \emph{sharing} the linear map from $\Ainp$ to $\Aoutp$ between different spatial locations.
However, while the feature spaces $\Ainp$ and $\Ainq$ as well as $\Aoutp$ and $\Aoutq$ are for different $p,q\in M$ isomorphic to each other, there is no canonical isomorphism between them given if the considered structure group~$G$ is non-trivial.
It is therefore not obvious how the linear map could be shared between different locations.
As already suggested in the introduction of this section, this issue is resolved by considering $G$-equivariant kernels which are indifferent to the specific choice of isomorphism or gauge.
The arbitrariness of the trivialization which is chosen from the $G$-atlas reflects the $\GM$-coordinate independence of \onexoneGMs.

Mathematically, \onexoneGMs\ can be formulated either as specific vector bundle $M$-morphisms or via the corresponding sections of (associated) homomorphism bundles $\Hom(\Ain,\Aout)$.
Since we require both concepts later on, we will introduce both viewpoints in the following Sections~\ref{sec:onexone_M_morphism} and~\ref{sec:onexone_hom_section}.

\subsubsection[\texorpdfstring{${1\kern-2.7pt\times\kern-2.9pt1}$}{1x1} \textit{GM}-convolutions as vector bundle \textit{M}-morphisms]%
              {\texorpdfstring{$\bm{1\kern-2.7pt\times\kern-2.9pt1}$}{1x1} \textit{GM}-convolutions as vector bundle \textit{M}-morphisms}
\label{sec:onexone_M_morphism}

\onexoneGMs\ can be formalized in terms of specific smooth \emph{vector bundle $M$-morphisms} which share weights over spatial positions.
Ignoring the requirement for shared weights for now, such a vector bundle $M$-morphism $\mathcal{C}$ is a smooth bundle map satisfying the following commutative diagram:
\begin{equation}\label{eq:bundle_morphism_onexone}
    \begin{tikzcd}[row sep=3.5em, column sep=2.5em]
        \Ain
            \arrow[rd, "\piAin"']
            \arrow[rr, "\mathcal{C}"]
        & &
        \Aout
            \arrow[ld, "\piAout"]
        \\
        & M
    \end{tikzcd}
\end{equation}
The commutativity $\piAin = \piAout \!\circ\, \mathcal{C}$ ensures that each fiber $\Ainp$ is mapped to the fiber $\Aoutp$ over the same point $p\in M$ (which gives rise to the ``$M$'' in the term $M$-morphism).
As a vector bundle morphism, the restriction $\mathcal{C}\!\!\;|_p: \Ainp\to\Aoutp$ to a single fiber is further defined to be linear.
Relative to a local trivialization $\PsiAin^A$ of $\Ain$ and $\PsiAout^A$ of $\Aout$, the bundle map is therefore at each point $p\in U^A$ represented by a matrix
\begin{align}\label{eq:bundle_morphism_onexone_triv_local}
    \mathcal{C}^A\!\!\;|_p\ :=\  \psiAoutp^A \circ \mathcal{C}\!\;|_p \circ \big(\psiAinp^A\big)^{-1}\ \in\, \R^{\cout\times\cin} \,.
\end{align}
Its relationship to a second coordinatization $\mathcal{C}^B$ is at $p\in U^A\cap U^B$ given by
\begin{align}\label{eq:onexone_gaugetrafo}
    \mathcal{C}^B\!\!\;|_p \ =\ \rhoout\big(g_p^{BA}\big)\; \mathcal{C}^A\!\!\;|_p\; \rhoin\big(g_p^{BA}\big)^{-1} \,,
\end{align}
which is evident from the commutative diagram below:
\begin{equation}
\begin{tikzcd}[column sep=65pt, row sep=30, font=\normalsize]
    \R^{\cin}
        \arrow[dd, "\rhoin\big(g_p^{BA}\big)\ "']
        \arrow[rrr, "\mathcal{C}^A\!\!\;|_p"]
    & &[-3ex] &
    \R^{\cout}
        \arrow[dd, "\ \rhoout\big(g_p^{BA}\big)"]
    \\
    &
    \Ainp
        \arrow[r, "\mathcal{C}\!\!\;|_p"]
        \arrow[ul, "\psiAinp^A"]
        \arrow[dl, "\psiAinp^B"']
    &
    \Aoutp
        \arrow[ur, "\psiAoutp^A"']
        \arrow[dr, "\psiAoutp^B"]
    \\
    \R^{\cin}
        \arrow[rrr, "\mathcal{C}^B\!\!\;|_p"']
    & & &
    \R^{\cout}
\end{tikzcd}
\end{equation}

The bundle map $\mathcal{C}$ acts on input feature fields $\fin \in \Gamma(\Ain)$ to produce output feature fields
\begin{align}
    \fout = \mathcal{C} \circ \fin
    \quad \in\ \ \Gamma(\Aout) \,.
\end{align}
In terms of a commutative diagram, this mapping is visualized as:
\begin{equation}\label{eq:bundle_morphism_onexone_section}
    \begin{tikzcd}[row sep=3.5em, column sep=2.5em]
        \Ain
            \arrow[rr, "\mathcal{C}"]
        & &
        \Aout
        \\
        & M \arrow[ul, "\fin"]
            \arrow[ur, "\fout"']
    \end{tikzcd} ,
\end{equation}

In order for a vector bundle $M$-morphism $\mathcal{C}_{K_{\!1\!\times\!1}}$ to represent a \onexoneGM, it needs to be parameterized in terms of a \onexoneGM\ kernel template $K_{\!1\!\times\!1} \in \R^{\cout\times\cin}$ which is shared with coordinatizations at all spatial positions.
As argued before, no particular gauge must thereby be preferred in order to ensure \emph{$\GM$-coordinate independence}.
It is therefore necessary to \emph{share the weights with all trivializations $X \in \mathfrak{X}$ of the $G$-atlas $\mathscr{A}^G$ simultaneously}, that is, to require:
\begin{align}\label{eq:weight_sharing_onexone}
    \mathcal{C}_{K_{\!1\!\times\!1}}^X\!\big|_p\ =\ K_{\!1\!\times\!1}
    \qquad \textup{for \emph{any} gauge}\,\ X \in \mathfrak{X}\,\ \textup{with}\,\ p\in U^X \,.
\end{align}
From the transformation behavior between different coordinatizations in Eq.~\eqref{eq:onexone_gaugetrafo} it follows that the kernel template has to satisfy the linear constraint
\begin{align}\label{eq:onexone_kernel_constraint}
    \rhoout(g)\, K_{\!1\!\times\!1}\, \rhoin(g)^{-1}  =\, K_{\!1\!\times\!1} \qquad\forall g\in G,
\end{align}
that is, it has to be an intertwiner (an equivariant linear map).
The vector space
\begin{align}
    \Hom_G(\rhoin,\rhoout)\ :=\ 
    \pig\{ K_{\!1\!\times\!1} \in \R^{\cout\times\cin}\ \pig|\ 
    K_{\!1\!\times\!1} \rhoin(g) = \rhoout(g) K_{\!1\!\times\!1}\ \ \forall g\in G \pig\}
\end{align}
of intertwining maps characterizes the space of $\GM$-coordinate independent \onexone\ kernels fully.
As already mentioned in Section~\ref{sec:gauge_1x1}, \emph{Schur's Lemma}~\cite{gallier2019harmonicRepr} implies that the requirement on $K_{\!1\!\times\!1}$ to be an intertwiner prevents a mapping between fields which transform under non-isomorphic irreducible representations via \onexoneGMs.
The more general $\GM$-convolutions with spatially extended kernels, defined in Section~\ref{sec:global_conv}, will resolve this issue.

With these preparations we are ready to give a concise definition of \onexoneGMs:
\begin{dfn}[\onexoneGM]
\label{dfn:onexone}
    A \onexoneGM\ is a map
    \begin{align}
        K_{\!1\!\times\!1} \ostar :\ \Gamma(\Ain) \to \Gamma(\Aout),\ \ \ 
        \fin \,\mapsto\, K_{\!1\!\times\!1} \,\ostar\, \fin \,:=\, \mathcal{C}_{K_{\!1\!\times\!1}}\! \circ \fin
    \end{align}
    which is parameterized by an \emph{intertwining} \onexoneGM\ kernel $K_{\!1\!\times\!1} \in \Hom_G(\rhoin,\rhoout)$.
    Here $\mathcal{C}_{K_{\!1\!\times\!1}}$ is the unique smooth vector bundle $M$-morphism between $\Ain$ and $\Aout$ which is in \emph{arbitrary} gauges $\psiAinp$ and $\psiAoutp$ from the considered $G$-atlas pointwise defined by
    \begin{align}
        \mathcal{C}_{K_{\!1\!\times\!1}}|_p\ :=\ \psiAoutp^{-1} \circ K_{\!1\!\times\!1} \circ \psiAinp \,.
    \end{align}
    The independence of the chosen gauges ($\GM$-coordinate independence) is guaranteed by $K_{\!1\!\times\!1}$ being an intertwiner.
\end{dfn}
To show the independence of the chosen gauge explicitly, consider any $G$-related trivializations $\rhoin(g)\,\psiAinp$ and $\rhoout(g)\,\psiAoutp$ for an arbitrary structure group element $g\in G$, which leave the construction of
\begin{align}
    \mathcal{C}_{K_{\!1\!\times\!1}} \mkern-2mu\big|_p\ 
    =\ \ &\big(\rhoout(g)\, \psiAoutp \big)^{-1} \circ K_{\!1\!\times\!1} \circ \big(\rhoin(g)\, \psiAinp\big) \notag \\
    =\ \ &\psiAoutp^{-1} \circ \big( \rhoout(g)^{-1} K_{\!1\!\times\!1}\, \rhoin(g) \big) \circ \psiAinp \notag \\
    =\ \ &\psiAoutp^{-1} \circ K_{\!1\!\times\!1} \circ \psiAinp
\end{align}
invariant.
That such defined \onexoneGMs\ are indeed mapping to sections in $\Gamma(\Aout)$ follows from $\mathcal{C}_{K_{\!1\!\times\!1}}$ being a bundle map.
An overview of local coordinatizations of \onexoneGMs\ is given in Fig.~\ref{fig:triv_bundle_morphism_onexone}.

\begin{figure}
    \centering
    \begin{tikzcd}[row sep=4.5em, column sep=4.35em, crossing over clearance=.6ex,
                   execute at end picture={
                        \node [] at (-1.16, -1.05) {$\noncommutative$};
                        \node [] at ( 1.09, -1.05) {$\noncommutative$};
                        }]
          U\times \R^{\cin}
                            \arrow[rrrr, pos=.5, rounded corners, to path={ 
                                    -- ([yshift=2.5ex]\tikztostart.north) 
                                    --node[above]{\small$
                                        \mathcal{C}_{K_{\!1\!\times\!1}}^B
                                        := (\id\times K_{\!1\!\times\!1})
                                    $} ([yshift=2.5ex]\tikztotarget.north) 
                                    -- (\tikztotarget.north)
                                    }]
        &[-3.0ex] & &
        &[-3.0ex] U\times \R^{\cout}
        \\
          U\times \R^{\cin}
                            \arrow[drr, pos=.5, "\proj_1"']
                            \arrow[u, "\big(\id\!\times\! \rhoin\big(g^{BA}\big)\!\cdot\big)"]
                            \arrow[rrrr, pos=.5, rounded corners, to path={ 
                                    -- ([yshift=-17.5ex]\tikztostart.south) 
                                    --node[below]{\small$
                                        \mathcal{C}_{K_{\!1\!\times\!1}}^A
                                        := (\id\times K_{\!1\!\times\!1})
                                    $} ([yshift=-17.5ex]\tikztotarget.south) 
                                    -- (\tikztotarget.south)
                                    }]
        & \piAin^{-1}(U)    \arrow[dr, shorten <=-3pt, shift right=.25, pos=.2, "\piAin\mkern-12mu"']
                            \arrow[l,  "\PsiAin^A"']
                            \arrow[lu, "\PsiAin^B"']
                            \arrow[rr,  "\mathcal{C}_{K_{\!1\!\times\!1}}"]
        &
        & \piAout^{-1}(U)   \arrow[dl, shorten <=-2pt, shift left=.25, pos=.2, "\mkern-6mu\piAout"]
                            \arrow[r,  "\PsiAout^A"]
                            \arrow[ru, "\PsiAout^B"]
        & U\times \R^{\cout}
                            \arrow[u, swap, "\big(\id\!\times\! \rhoout\big(g^{BA}\big)\!\cdot\big)"]
                            \arrow[lld, pos=.5, "\proj_1"]
        \\[1.5ex]
        &&
          U \arrow[ul, shorten >=-5pt, bend right=22, looseness=.5, pos=.6, "\!\fin"']
            \arrow[ur, shorten >=-5pt, bend left =22, looseness=.5, pos=.6, "K_{\!1\!\times\!1} \mkern-1mu\ostar\mkern-1mu \fin\!\!"]
    \end{tikzcd}
    \caption{\small
        Coordinatization of an \onexoneGM\ $K_{\!1\!\times\!1} \protect\ostar: \Gamma(\Ain) \to \Gamma(\Aout)$ and its corresponding vector bundle $M$-morphism $\mathcal{C}_{K_{\!1\!\times\!1}}$.
        The convolutional character is encoded into the morphism by sharing a kernel matrix $K_{\!1\!\times\!1}\in\R^{\cout\times\cin}$ over different spatial positions $p\in M$.
        Since no gauge is to be preferred, the kernel is furthermore shared over different trivializations
        $\mathcal{C}_{K_{\!1\!\times\!1}}^A$ and $\mathcal{C}_{K_{\!1\!\times\!1}}^B$.
        The commutativity of the diagram for any choices
        $\Psi_{\!\!\A_\text{in} }^A$,
        $\Psi_{\!\!\A_\text{out}}^A$ and
        $\Psi_{\!\!\A_\text{in} }^B$,
        $\Psi_{\!\!\A_\text{out}}^B$
        therefore enforces the constraint
        $\rho_\text{out}(g) K_{\!1\!\times\!1} \rho_\text{in}(g)^{-1} = K_{\!1\!\times\!1}\,\ \forall g\!\in G$
        which restricts the kernel matrix to be an intertwiner (an equivariant linear map), that is,
        $K_{\!1\!\times\!1} \in \Hom_G(\rhoin,\rhoout) \subseteq \R^{\cout\times\cin}$.
        Except for $\fin \circ \piAin \neq \id_{\Ain}$ and $\big[K_{\!1\!\times\!1} \protect\ostar \fin\big] \circ \piAout \neq \id_{\Aout}$, the diagram commutes.
    }
    \label{fig:triv_bundle_morphism_onexone}
\end{figure}

\subsubsection[\texorpdfstring{${1\kern-2.7pt\times\kern-2.9pt1}$}{1x1} \textit{GM}-convolutions as homomorphism bundle sections]%
              {\texorpdfstring{$\bm{1\kern-2.7pt\times\kern-2.9pt1}$}{1x1} \textit{GM}-convolutions as homomorphism bundle sections}
\label{sec:onexone_hom_section}

While the vector bundle $M$-morphism with gauge independent coordinatizations from Def.~\ref{dfn:onexone} and Fig.~\ref{fig:triv_bundle_morphism_onexone} fully specifies a \onexoneGM, we will now adopt an alternative viewpoint which describes \onexoneGMs\ in terms of the \emph{homomorphism bundle} $\Hom(\Ain,\Aout) \xrightarrow{\,\piHom\,} M$.
To this end, recall that the vector bundle morphism $\mathcal{C}$ in Eq.~\eqref{eq:bundle_morphism_onexone} restricts to linear maps ${\mathcal{C}\!\:|_p}: \Ainp\to\Aoutp$ over each $p\in M$.
The set of such linear maps (or vector space homomorphisms) between $\Ainp$ and $\Aoutp$ is denoted as $\Hom(\Ainp,\Aoutp)$.
Since it is closed under linear combinations, it forms a vector space itself.
It can be shown that the disjoint union
\begin{align}
    \Hom(\Ain,\Aout)\ :=\ \coprod_{p\in M} \Hom(\Ainp,\Aoutp)
\end{align}
of these homomorphism spaces forms a vector bundle, the homomorphism bundle between $\Ain$ and $\Aout$, when being equipped with the projection map $\piHom: \Hom(\Ain,\Aout) \to M$ which sends elements in $\Hom(\Ainp,\Aoutp)$ to $p$ and a smooth structure induced from that of $\Ain$ and $\Aout$~\cite{dundas2018differentialTopology}.
The fibers over~$p$ satisfy $\Hom(\Ainp,\Aoutp) \cong \Hom(\R^\cin,\R^\cout) \cong \R^{\cout\times\cin}$ such that we can take the typical fiber to be the vector space of real-valued ${\cout\!\times\!\cin}$ matrices.
The trivializations
\begin{align}
    \PsiHom:\ \piHom^{-1}(U) \to U\!\times\R^{\cout\times\cin},\ \ H\mapsto \big(p,\ \psiHomp(H)\big) ,
\end{align}
where we abbreviated $p=\piHom(H)$, are \emph{induced} from the trivializations of $\Ain$ and $\Aout$ by defining
\begin{align}\label{eq:Hom_bdl_triv_ptwise}
    \psiHomp\!:\ \Hom(\Ainp,\Aoutp) \to \R^{\cout\times\cin} ,\ \ \ H\mapsto \psiAoutp \circ H \circ \big(\psiAinp\big)^{-1}
\end{align}
in analogy to Eqs.~\eqref{eq:bundle_morphism_onexone_triv_local} and \eqref{eq:matrix_trivialization}.
This implies transition maps
\begin{alignat}{3}
    H^B
    \ &=&\ \psiAoutp^B \circ\,&H \circ \big(\psiAinp^B\big)^{-1} \notag \\
    \ &=&\ \psiAoutp^B \circ \big(\psiAoutp^A\big)^{-1}\,&H^A\ \psiAinp^A \circ \big(\psiAinp^B\big)^{-1} \notag \\
    \ &=&\ \rhoout\big(g^{BA}\big)\, &H^A\, \rhoin\big(g^{BA}\big)^{-1} \notag \\
    \ &=:&\ \rhoHom\big(g^{BA}\big)\, &H^A
\end{alignat}
between gauges $\PsiHom^A$ and $\PsiHom^B$ on $U^A\cap U^B$, where we introduced the homomorphism group representation $\rhoHom:G\to\GL{\R^{\cout\times\cin}}$ as left and right multiplication with $\rhoout$ and $\rhoin$ for notational convenience.%
\footnote{
    In general, a homomorphism bundle between two \emph{non-associated} vector bundles with structure groups $G_1$ and $G_2$ would have a structure group $G_1\times G_2$.
    Since $\Ain$ and $\Aout$ are associated, they transform synchronously under the same structure group $G_1=G_2=G$ such that their transition maps take values in the diagonal subgroup $G$ of $G\times G$.
}
The homomorphism bundle $\Hom(\Ain,\Aout)$ is by construction associated to $\TM$, $\GM$, $\Ain$ and $\Aout$, that is, its trivializations transform synchronously with those of the other bundles.
As a $G$-associated vector bundle, it can be identified with $(\GM\times\R^{\cout\times\cin})/\!\sim_{\!\rhoHom}$.
Fig.~\ref{fig:trivialization_hom} gives an overview of the local trivializations of $\Hom(\Ain,\Aout)$.
Note the similarity to the trivializations of the other associated $G$-bundles in Fig.~\ref{fig:trivializations_TM_FM_A}.

\begin{figure}
    \centering
    \begin{subfigure}[b]{0.48\textwidth}
        \begin{tikzcd}[row sep=4.em, column sep=5.5em]
            & U\times \R^{\cout\times\cin} \\
              \piHom^{-1}(U)
                    \arrow[d, swap, "\piHom"]
                    \arrow[r, "\PsiHom^A"]
                    \arrow[ru, "\PsiHom^B"]
            & U\times \R^{\cout\times\cin}
                    \arrow[u, swap, "\big( \id\times \rhoHom\big(g^{BA}\big)\!\cdot \big)"]
                    \arrow[ld, "\proj_1"] \\
            U
        \end{tikzcd}
        \centering
        \caption{\small
            Trivialization of $\Hom(\Ain,\Aout)$.
            Being associated to $\TM$, $\GM$, $\Ain$ and $\Aout$, the transition maps of the homomorphism bundle are determined by the same group element $g^{BA}$ of the shared structure group $G$ (compare this to Fig.~\ref{fig:trivializations_TM_FM_A}).
            Unconstrained vector bundle $M$-morphisms as shown in Eq.~\eqref{eq:bundle_morphism_onexone} correspond to unconstrained smooth sections of $\Hom(\Ain,\Aout)$.
        }
        \label{fig:trivialization_hom}
    \end{subfigure}
    \hfill
    \begin{subfigure}[b]{0.49\textwidth}
        \centering
        \begin{tikzcd}[row sep=5.5em, column sep=7.em, crossing over clearance=.6ex,
                       execute at end picture={
                            \node [] at (-2.98, -.11) {$\noncommutative$};
                            }]
              \piHom^{-1}(U)    \arrow[d, "\piHom", shift left=.2]
                                \arrow[r, "\PsiHom^A"', bend right=3, shift right=.5, looseness=.5, pos=.45]
                                \arrow[r, "\PsiHom^B",  bend left=3,  shift left=.5,  looseness=.5, pos=.45]
            & U\!\times\! \underbrace{\Hom_G(\rhoin,\rhoout)}_{\subseteq\ \R^{\cout\times\cin}}
                                \arrow[loop, distance=3.5em, in=125, out=55, "\big(\id\times \rhoHom\big(g^{BA}\big)\!\cdot\!\big)"']
                                \arrow[ld, "\proj_1", shorten <= -15pt]
            \\
              U                 \arrow[u, bend left=20, shift left=.5, "\sigma_{K_{1\!\times\!1}}"]
        \end{tikzcd}
        \caption{\small
            The sections $\sigma_{K_{1\!\times\!1}}: M\to \Hom(\Ain,\Aout)$ of the homomorphism bundle which correspond to \onexoneGMs\ are exactly those which trivialize to the \emph{same} (intertwining) matrix
            $K_{\!1\!\times\!1} \in \Hom_G(\rhoin,\rhoout) \subseteq \R^{\cout\times\cin}$
            in all gauges.
            Such sections correspond to bundle maps which trivialize as specified in Fig.~\ref{fig:triv_bundle_morphism_onexone}.
            \\~
        }
        \label{fig:trivialization_hom_onexone_section}
    \end{subfigure}
    \caption{\small
        Local trivializations of the homomorphism bundle $\Hom(\Ain,\Aout)$, which is the vector bundle of linear maps between the spaces $\Ainp$ and $\Aoutp$ for any $p\in M$.
        As usual we abbreviate $U=U^A\cap U^B$.
        Except for $\sigma_{K_{1\!\times\!1}} \circ \piHom \neq \id_{\Hom(\Ain,\Aout)}$ the diagrams commute.
    }
    \label{fig:trivializations_hom_bundle}
\end{figure}

From the viewpoint of homomorphism bundles, unconstrained bundle maps as in Eq.~\eqref{eq:bundle_morphism_onexone} correspond to the action of unconstrained smooth homomorphism bundle sections
\begin{align}\label{eq:hom_bdl_section_unconstrained}
    \sigma_{\Hom}:M \mapsto \Hom(\Ain,\Aout)
    \quad \textup{such that} \quad
    \piHom \circ \sigma_{\Hom} = \id_M
\end{align}
which can be interpreted as $1\!\times\!1$ \emph{kernel fields} that do not share weights.
Their global existence is guaranteed by $\Hom(\Ain,\Aout)$ being a vector bundle.
Sections corresponding to \onexoneGMsit\ require in addition that the linear transformations $\sigma_{\Hom}(p)\in\Hom(\Ainp,\Aoutp)$ are determined by a template kernel $K_{\!1\!\times\!1} \in\R^{\cout\times\cin}$ which is shared over different positions $p\in M$ and any choice of gauge.
They can therefore for any $p \in \!M$ be defined as
\begin{align}
    \sigma_{K_{1\!\times\!1}}(p)\ :=\ \psiHomp^{-1}\big(K_{\!1\!\times\!1}\big), \qquad K_{\!1\!\times\!1} \in \Hom_G(\rhoin,\rhoout) \,,
\end{align}
where the chosen trivialization $\PsiHom$ is arbitrary if (and only if) $K_{\!1\!\times\!1}$ satisfies the intertwiner constraint
\begin{align}\label{eq:onexone_intertwiner_constraint_rhoHom}
    \rhoHom(g) K_{\!1\!\times\!1} = K_{\!1\!\times\!1} \qquad\forall g\in G \,,
\end{align}
which is equivalent to Eq.~\eqref{eq:onexone_kernel_constraint}.%
\footnote{
    The required smoothness of the section follows from the smoothness of the local trivializations.
}
The gauge irrelevance of such sections is visualized in the commutative diagram in Fig.~\ref{fig:trivialization_hom_onexone_section} (compare this to the equivalent bundle map trivialization in Fig.~\ref{fig:triv_bundle_morphism_onexone}).

\paragraph{Summarizing remarks:}
A smooth \onexoneGM\ layer $K_{\!1\!\times\!1} \ostar: \Gamma(\Ain)\to \Gamma(\Aout),\ \fin\mapsto \fout$ can equivalently be defined via a smooth bundle map as $\fout(p) := \mathcal{C}_{K_{\!1\!\times\!1}} \!\circ \fin(p)$ or via a smooth homomorphism bundle section as $\fout(p) := \sigma_{K_{\!1\!\times\!1}}(p) \circ \fin(p)$.
By definition, both trivialize in an arbitrarily chosen gauge $\PsiHom^A$ to $\fout^A(p) = K_{\!1\!\times\!1} \fin^A(p)$.
The $\GM$-coordinate independence of this definition is guaranteed by the intertwining property of the kernel in Eq.~\eqref{eq:onexone_kernel_constraint} or, equivalently, Eq.~\eqref{eq:onexone_intertwiner_constraint_rhoHom}.
This can be seen by considering a different trivialization via~$\PsiHom^B$:
\begin{align}
    K_{\!1\!\times\!1} \fin^B(p)
    \ &=\ K_{\!1\!\times\!1} \left( \rhoin\big(g^{BA}_p\big) \fin^A(p) \right) \notag \\
    \ &=\ \rhoout\big(g^{BA}_p\big) K_{\!1\!\times\!1} \fin^A(p) \notag \\
    \ &=\ \rhoout\big(g^{BA}_p\big) \fout^A(p) \notag \\
    \ &=\ \fout^B(p)
\end{align}

%% file: chapters/72_GMconv.tex

\subsection{Kernel field transforms and \textit{GM}-convolutions}
\label{sec:global_conv}

We now turn to kernel field transforms and $\GM$-convolutions with spatially extended (convolution) kernels.
Section~\ref{sec:kernel_fields} introduces general, unconstrained kernel fields and more specific $\GM$-convolutional kernel fields, which are defined in terms of a shared, $G$-steerable template kernel.
General kernel field transforms and $\GM$-convolutions are introduced in Section~\ref{sec:KFTs_GM-conv_global}.
As both are defined \emph{globally}, their formulation is necessarily \emph{coordinate free}.
Section~\ref{sec:KFTs_GM-conv_local} expresses both operations relative to local trivializations, recovering our local definitions from Section~\ref{sec:gauge_conv_main}.

\subsubsection[Kernel fields]{Coordinate free kernels fields and \textit{G}-steerable kernels}
\label{sec:kernel_fields}

To detect spatial pattens in feature fields, convolutional networks apply spatially extended kernels which linearly accumulate features from a local neighborhood around each point.
In Eq.~\eqref{eq:conv_kernel_unrestricted} we defined (unconstrained) template kernels for a $d$-dimensional manifold and $\cin$- and $\cout$-dimensional input and output feature fields as maps
$K: \R^d \to \R^{\cout\times\cin}$
which assign a $\cout\times\cin$ matrix to each point of their domain.
The definition of convolution kernels as maps with domain $\R^d\cong \TpM$ and codomain $\R^{\cout\times\cin}\cong \Hom(\Ainp,\Aoutp)$ suggests a coordinate free definition of kernels
as maps between the tangent spaces and the corresponding homomorphism spaces:

\begin{dfn}[Kernel field]
\label{dfn:kernel_field_general}
    We define (unconstrained) \emph{kernel fields} of type $\rhoin,\rhoout$ on a manifold $M$ as smooth bundle $M$-morphisms between the tangent bundle $\TM$ and the feature vector homomorphism bundle $\Hom(\Ain,\Aout)$.
    By its definition as an $M$-morphism, a kernel field $\K$ lets the following diagram commute:
    \begin{equation}\label{eq:kernel_bundle_map}
        \begin{tikzcd}[row sep=3.5em, column sep=2.5em]
            \TM  \arrow[rd, start anchor={[xshift=-1ex]}, "\piTM"']
                \arrow[rr, "\K"]
            & &
            \mkern-3mu
            \Hom(\Ain,\Aout)
                \arrow[ld, start anchor={[xshift=-3ex]}, shorten >=.1ex, shorten <=-.75ex, "\piHom"] \\
            & M &
        \end{tikzcd}
    \end{equation}
    Despite smoothly mapping between two vector bundles, $\K$ is \emph{not} assumed to be a \emph{vector} bundle morphism, that is, the restrictions $\Kp: \TpM \to \Hom(\Ainp, \Aoutp)$ are not assumed to be linear.%
    \footnote{
        This reflects that convolution kernels are in general not linear as maps $K:\R^d \to \R^{\cout\times\cin}$.
        Note that this does not interfere with the linearity of $K(\mathscr{v}) \in \R^{\cout\times\cin}$ (as map $\R^{\cin} \to \R^{\cout}$) for any $\mathscr{v}\in\R$ or, here, the linearity of $\Kp(v) \in \Hom(\Ainp,\Aoutp)$ (as map $\Ainp \to \Aoutp$) for any $v\in\TpM$.
    }
\end{dfn}
The name \emph{kernel field} is motivated by the fact that such defined bundle maps $\K$ assign a (potentially different) coordinate free kernel $\Kp: \TpM \to \Hom(\Ainp,\Aoutp)$ to each point $p$ of the manifold.%
\footnote{
    We expect that it is possible to work out a well defined notion of \emph{kernel bundles} whose sections are in one to one correspondence to our definition of kernel fields as bundle maps (this reformulation would mirror the transition from Eq.~\eqref{eq:bundle_morphism_onexone} to Eq.~\eqref{eq:hom_bdl_section_unconstrained}).
}
In practice, kernels $\Kp$ are often designed to detect local patterns around $p$ and are therefore assumed to be compactly supported around the origin of $\TpM$.

A coordinate free kernel $\Kp$ at $p$ is relative to gauges $\psiTMp^A$ and $\psiHomp^A$ of the $G$-atlases given by the map
\begin{align}\label{eq:kernel_field_general_coord_expression}
    \Kp^A: \R^d \to \R^{\cout\times\cin}, \qquad
    \Kp^A\: :=\: \psiHomp^A \circ \Kp \circ \big(\psiTMp^A\big)^{-1} \,.
\end{align}
Fig.~\ref{fig:kernel_coordinatization} visualizes a coordinate free kernel on $\TpM$ and its coordinatizations on $\R^d$ relative to different gauges.
From the commutative diagram
\begin{equation}
    \begin{tikzcd}[column sep=45pt, row sep=30, font=\normalsize]
        \R^d    \arrow[rrr, pos=.4, "\Kp^A"]
                \arrow[dd, "g_p^{BA}\cdot"']
        & & &[-5pt]
        \R^{\cout\times\cin}
                \arrow[dd, "\,\rhoHom\big(g_p^{BA}\big)"]
        \\
        &
        \TpM    \arrow[r, "\Kp"]
                \arrow[lu, pos=.4, "\psiTMp^A"']
                \arrow[ld, pos=.4, "\psiTMp^B"]
        &
        \Hom\!\big(\Ain|_p,\Aout|_p\big)
                \arrow[ru, start anchor={[xshift=2.em]}, pos=.4, "\psiHomp^A"]
                \arrow[rd, start anchor={[xshift=2.em]}, pos=.4, "\psiHomp^B"']
        &
        \\
        \R^d    \arrow[rrr, pos=.4, "\Kp^B"']
        & & &
        \R^{\cout\times\cin}
    \end{tikzcd}
\end{equation}
it follows that different kernel coordinatizations are related by
\begin{align}\label{eq:kernel_field_unconstrained_gaugetrafo}
    \Kp^B \ =\ \rhoHom\big(g_p^{BA}\big) \circ \Kp^A \circ \big(g_p^{BA}\big)^{-1} \,.
\end{align}
Note that this relation only implies $\GM$-coordinate independence but does not constrain the coordinate free kernel in any way.
As before, the situation changes when sharing weights over spatial positions.

In order for a kernel field $\KK$ to correspond to a convolution, it needs to be fully specified by a single template kernel $K: \R^d \to \R^{\cout\times\cin}$ which is shared over all spatial positions.
We are again forced to share weights with all gauges $X \in \mathfrak{X}$ simultaneously in order to preserve their equivalence and thus $\GM$-coordinate independence.
As argued in Section~\ref{sec:gauge_conv}, the appropriate way of sharing $K$ with kernel coordinatizations $\KKp^X$ involves a normalization by the reference frame volume $\sqrt{|\eta_p^X|}$ and is defined by
\begin{align}\label{eq:weight_sharing_kernel_62}
    \KKp^X\ =\ \frac{K}{\sqrt{|\eta_p^X|}}
    \qquad \textup{for \emph{any} gauge}\,\ X \in \mathfrak{X}\,\ \textup{with}\,\ p\in U^X \,.
\end{align}
The reason for the frame normalization factor is that convolutions will later be defined in terms of integrals over the tangent spaces.
We are therefore actually required to share the integral operator itself in different coordinatizations,
which is equivalent to identifying the matrix-valued integration measures $\KKp^X(\mathscr{v})\, \sqrt{|\eta_p^X|}\; d\mathscr{v}$ for any gauge $X \in \mathfrak{X}$ at $p\in M$ with a template measure $K(\mathscr{v})\, d\mathscr{v}$.
The form of the kernel sharing in Eq.~\eqref{eq:weight_sharing_kernel_62} follows by equating both expressions.

Together with the relation $\sqrt{|\eta_p^A|} = \big|\!\det(g_p^{BA})\big| \mkern1.5mu \sqrt{|\eta_p^B|}$ between different frame volumes, the kernel transformation law in Eq.~\eqref{eq:kernel_field_unconstrained_gaugetrafo} and the weight sharing in Eq.~\eqref{eq:weight_sharing_kernel_62} imply the \emph{$G$-steerability kernel constraint}
\begin{align}\label{eq:kernel_constraint_rhohom}
    \frac{1}{\detg} \, \rhoHom(g) \circ K \circ g^{-1}\ =\ K \qquad\forall\: g\in G \,.
\end{align}
Valid template kernels are thus given by the invariants under the simultaneous gauge action of $\detg^{-1}$, $\rhoHom(g)$ and $g^{-1}$.
Writing out the representation $\rhoHom$, acting on $\R^{\cout\times\cin}$ via multiplication with $\rhoout$ and $\rhoin^{-1}$ from the left and right, respectively, the constraint in Eq.~\eqref{eq:kernel_constraint_rhohom} is seen to be equivalent to that in Eq.~\eqref{eq:kernel_constraint}, i.e.
$K(g\mkern1mu\mathscr{v}) = \detg^{-1} \rho_\text{out}(g)\, K(\mathscr{v})\, \rho_\text{in}(g)^{-1} \ \ \forall\, g\in G,\ \mathscr{v}\in\R^d$.

We cast these insights into definitions:
\begin{dfn}[$G$-steerable kernel]
\label{dfn:G-steerable_kernel_def_43}
    $G$-steerable kernels are characterized by their \emph{invariance under the gauge action}.
    The vector space of smooth $G$-steerable kernels that map between field types $\rhoin$ and $\rhoout$ is defined by
    \begin{align}
        \KG \!:=&\,
        \Big\{ K\!: \R^d \to \R^{\cout\times\cin}\ \text{smooth} \,\Big|\,
        \frac{1}{\detg}\, \rhoHom(g) \circ K \circ g^{-1} =\, K \ \ \ \forall g\in G \Big\} \,,
        \label{eq:G_steerable_space_in_dfn_Hom} \\[1ex]
        =&\ 
        \Big\{ K\!: \R^d \to \R^{\cout\times\cin}\ \text{smooth} \,\Big|\,
        \frac{1}{\detg}\mkern1mu \rhoout(g) K(g^{-1}\mathscr{v}) \rhoin(g)^{-1} \mkern-2mu= K(\mathscr{v}) \ \ \ \forall\, g\in G,\ \mathscr{v} \in \R^d \Big\} ,
        \label{eq:G_steerable_space_in_dfn_classical}
    \end{align}
    where $\rhoHom(g)H := \rhoout(g)H \rhoin(g)^{-1}$ for any $H\in \R^{\cout\times\cin}$ and $G\leq\GL{d}$.
    The gauge invariance of $G$-steerable kernels allows for $\GM$-coordinate independent weight sharing.
\end{dfn}
$G$-steerable kernels were in~\cite{Cohen2017-STEER} introduced to equivariant deep learning, where finite groups were assumed.
The current formulation in Def.~\ref{dfn:G-steerable_kernel_def_43} was proposed in~\cite{3d_steerableCNNs}.
A complete solution for the $G$-steerable kernel spaces for arbitrary representations $\rhoin$ and $\rhoout$ of structure groups $G\leq\O2$ has been derived in~\cite{Weiler2019_E2CNN}, an implementation is publicly available at \url{https://quva-lab.github.io/e2cnn/api/e2cnn.kernels.html}.
Mathematically, steerable kernel are equivalent to \emph{representation operators} like for instance the spherical tensor operators from quantum mechanics.
A generalization of the \emph{Wigner-Eckart theorem} describes $G$-steerable kernels as being composed from harmonic basis functions, Clebsch-Gordan coefficients and endomorphisms of irreducible representations~\cite{lang2020WignerEckart}.

\begin{dfn}[$\GM$-convolutional kernel field]
\label{dfn:conv_kernel_field}
    A \emph{$\GM$-convolutional kernel field} $\KK$ of type $\rhoin,\rhoout$ is a kernel field which is determined by a shared, \emph{$G$-steerable} template kernel $K \in \KG$.
    It is in \emph{arbitrary gauges} $\psiTMp^X$ and $\psiHomp^X\,$ from the considered $G$-atlas pointwise defined by:
    \begin{align}\label{eq:conv_kernel_field_def_ptwise}
        \KKp \,:=\ \big(\psiHomp^X\big)^{\mkern-2mu-1} \circ \frac{K}{\sqrt{|\eta_p^X|}} \circ \psiTMp^X
    \end{align}
    The smoothness of $\KK$ follows from the smoothness of the gauges, the metric and the template kernel.
\end{dfn}
As in the case of \onexoneGMs, the arbitrariness of the particular choice of gauge in Eq.~\eqref{eq:conv_kernel_field_def_ptwise} -- and therefore the $\GM$-coordinate independence of the definition -- is guaranteed by the $G$-steerability of $K\in\KG\!$.
To show this explicitly, one may define the kernel field relative to some gauge $B$ and then apply a transformation to any other gauge $A$, which cancels out and therefore leads to an equivalent expression:
\begin{align}\label{eq:arbitrariness_gauge_GM_kernel_field_def}
    \KKp
    \,\ =&\,\ \big(\psiHomp^B \big)^{-1} \,\circ \,\frac{K}{\sqrt{|\eta^B|}} \circ \psiTMp^B \notag \\
    \,\ =&\,\ \big( \rhoHom\big(g_p^{BA}\big)\, \psiHomp^A \big)^{-1} \,\circ \,\frac{K}{\sqrt{|\eta^A|} \,/\, |\! \det(g_p^{BA})| } \circ \big( g_p^{BA}\cdot \psiTMp^A \big) \notag \\
    \,\ =&\,\ \big(\psiHomp^A \big)^{-1} \,\circ \frac{\,\big|\! \det(g_p^{BA})\big|\, \rhoHom\big(g_p^{BA}\big)^{-1} \circ K \circ g_p^{BA}}{\sqrt{|\eta^A|}} \circ \psiTMp^A \notag \\
    \,\ =&\,\ \big(\psiHomp^A \big)^{-1} \,\circ \,\frac{K}{\sqrt{|\eta^A|}} \circ \psiTMp^A
\end{align}
Fig.~\ref{fig:triv_kernel_bundle_morphism} gives an overview of the local trivializations of $\GM$-convolutional kernel fields in terms of a commutative diagram.

\begin{figure}
    \centering
    \begin{tikzcd}[row sep=4.5em, column sep=5.2em]
          U\times\R^d
                        \arrow[rrrr, pos=.48, rounded corners, to path={ 
                                -- ([yshift=2.5ex]\tikztostart.north) 
                                --node[above]{\small$
                                    \KK^B = \big(\id\times K/\mkern-2mu \sqrt{|\eta^B|} \,\big)
                                $} ([yshift=2.5ex]\tikztotarget.north) 
                                -- (\tikztotarget.north)
                                }]
        & &[-3.25em] &[-3.25em] &
        U\times\R^{c_\text{out}\times c_\text{in}}
        \\
        U\times\R^d
                        \arrow[u, "\big(\id\times g^{BA}\!\!\cdot\big)"]
                        \arrow[rrd, "\proj_1"']
                        \arrow[rrrr, pos=.5, rounded corners, to path={ 
                                -- ([yshift=-16.ex]\tikztostart.south) 
                                --node[below]{\small$
                                    \KK^A = \big(\id\times K/\mkern-2mu \sqrt{|\eta^A|} \,\big)
                                $} ([yshift=-16.ex]\tikztotarget.south) 
                                -- (\tikztotarget.south)
                                }]
        &
        \piTM^{-1}(U)   
                        \arrow[rd, "\piTM\!\!"', pos=0.3]
                        \arrow[rr, "\KK"]
                        \arrow[lu, "\PsiTM^B"']
                        \arrow[l,  "\PsiTM^A"']
        & &
        \piHom^{-1}(U)
                        \arrow[ld, "\!\piHom", pos=0.3]
                        \arrow[r,  "\PsiHom^A"]
                        \arrow[ru, "\PsiHom^B"]
        &
        U\times\R^{c_\text{out}\times c_\text{in}}
                        \arrow[u, "\big(\id\times \rhoHom\big(g^{BA}\big)\!\cdot\big)"']
        \arrow[lld, "\proj_1"] \\
        & &
        U
        & &
    \end{tikzcd}
    \caption{\small
        Commutative diagram showing local coordinatizations of a $\GM$-\emph{convolutional kernel field} $\KK$ as defined in Def.~\ref{dfn:conv_kernel_field}.
        Convolutional weight sharing requires the coordinate expression of the kernel field $\KK$ at any point $p\in M$ and any gauge~$X$ at~$p$ to be determined by the shared template kernel ${K: \R^d \to \R^{\cout\times\cin}}$ as
         $\KKp^X = K/\sqrt{|\eta_p^X|}$.
        The commutativity of the diagram then implies the $G$-steerability constraint ${\detg^{-1} \rhoHom(g) \circ K \circ g^{-1} = K} \ \ \forall g\in G$ on the space~$\KG$ of template kernels.
        We want to emphasize that, despite looking similar to the diagram in Fig.~\ref{fig:triv_bundle_morphism_onexone}, the diagram in the current figure should be seen as analog to that in Fig.~\ref{fig:trivialization_hom_onexone_section}.
        The difference between the current diagram and that in Fig.~\ref{fig:trivialization_hom_onexone_section} is that the linear maps in the homomorphism bundle are via $\KK: \TM\to \Hom(\Ain,\Aout)$ determined by an element of the tangent bundle $\TM$ instead of the section $\sigma_{K_{1\!\times\!1}}: M\to \Hom(\Ain,\Aout)$.
    }
    \label{fig:triv_kernel_bundle_morphism}
\end{figure}

Note that the $G$-steerability constraint in Eq.~\eqref{eq:G_steerable_space_in_dfn_classical} or~\eqref{eq:G_steerable_space_in_dfn_Hom} reduces to the constraint on \onexoneGM\ kernels in Eq.~\eqref{eq:onexone_kernel_constraint} or~\eqref{eq:onexone_intertwiner_constraint_rhoHom} when being evaluated at the origin $\mathscr{v}=0$ of $\R^d$, which is invariant under the action of any~$g\in G$.
The results on \onexoneGMs, derived in the previous section, are therefore seen to be a special case for the choice of point-like kernels.%
\footnote{
    To make this statement precise, one would have to generalize Def.~\ref{dfn:G-steerable_kernel_def_43} to operator-valued distributions and define \onexoneGM\ kernels as operator-valued Dirac deltas.
    We omit this generalization here for brevity.
}
We further want to mention that the constraint on spatially extended kernels does in general not require their codomain to be restricted to $\Hom_G(\rhoin,\rhoout)$, i.e. the space of intertwiners.
In contrast to \onexoneGMs, this allows $\GM$-convolutions with spatially extended kernels to map between fields that transform according to non-isomorphic irreducible representations.

\subsubsection{Kernel field transforms and \textit{GM}-convolutions}
\label{sec:KFTs_GM-conv_global}

Having defined both feature fields and kernel fields, we are ready to introduce kernel field transforms and $\GM$-convolutions.
They are pointwise defined in terms of integral operators which compute output feature vectors $\fout(p)$ at points~$p\in M$ by matching the kernel $\Kp$ at~$p$ with the feature field~$\fin$ ``as seen from~$p$''.

The local representation of an input field ``as seen from~$p$'' is formally given by its \emph{transporter pullback}, which is visualized in Fig.~\ref{fig:pullback_field_exp_TpM}.
It~is defined as the usual pullback from $M$ to $\TM$ via the Riemannian exponential map%
\footnote{
    We define the exponential map on the full tangent bundle as
    $\exp: \TM \!\to M,\ \ v \mapsto \exp_{\scalebox{.85}{$\pi_{\overset{}{\protect\scalebox{.6}{$T\mkern-2muM$}}}\mkern-1mu(v)$}}(v)$.
    Recall that we assumed the manifold to be geodesically complete, such that the exponential map is well defined on the whole tangent bundle (and resort to zero-padding if this assumption fails to hold).
}
with the additional application of a parallel transporter (Eq.~\eqref{eq:transporter_A_def}), which is necessary in order to express the pulled back features in $\mathcal{A}_{\textup{in},\exp(v)}$ as features in~$\Ainp$.
Denoting this parallel transporter along the geodesic path $\gamma_v(t) := \exp((1-t) \,v)$ between $\gamma(0) = \exp(v)$ and $\gamma(1) = \pi(v) =: p$ by
\begin{align}
    \mathcal{P}_{\mkern-4mu\overset{}{\protect\scalebox{.75}{$\!\A$},\protect\scalebox{.75}{$\, p\!\leftarrow\!\exp(v)$}}}
    : \A_{\exp(v)} \to \A_p \,,
\end{align}
we thus define the pulled back feature field representations on the tangent spaces as follows:
\begin{dfn}[Transporter pullback of feature field to \textit{TM}]
\label{dfn:Expf_pullback_field}
    Given a feature field $f \in \Gamma(\A)$, we define its (redundant) representation on the tangent bundle as
    \begin{align}
        \Expsf :\ \TM \to \A, \quad\ 
        v \,\mapsto\, \PAexpv \!\circ f \circ \exp(v) \,.
    \end{align}
    The Riemannian exponential map $\exp$ corresponds hereby to the Levi-Civita connection, while the transporter $\PAexpv$ relies on some $G$-compatible connection; see Sections~\ref{sec:transport_local} and~\ref{sec:bundle_transport}.

    From the construction it is clear that $\Expsf(v) \in \A_p\,$ for any $v \in \TpM$, that is, $\Expsf$ is a bundle $M$-morphism, satisfying the following commutative diagram:
    \begin{equation}\label{eq:pullback_field_bundle_map}
        \begin{tikzcd}[row sep=3.5em, column sep=2.5em]
            \TM  \arrow[rd, start anchor={[xshift=.6ex]}, "\piTM"']
                \arrow[rr, "\Expsf"]
            & &
            \mkern-3mu
            \A
                \arrow[ld, "\piA"] \\
            & M &
        \end{tikzcd}
    \end{equation}
    Despite smoothly mapping between two vector bundles, $\Expsf$ is \emph{not} assumed to be a \emph{vector} bundle morphism, that is, the restrictions $\Expspf: \TpM \to \A_p$ are usually not linear.
\end{dfn}
The restriction $\Expspf := \Expsf\big|_{\TpM}$ of the transporter pullback's domain to $\TpM$ captures the feature field from the perspective of an observer at~$p$ as shown in Fig.~\ref{fig:pullback_field_exp_TpM}.
Note that this definition resembles a local representation of the feature field in terms of \emph{geodesic normal coordinates}, with the difference that it is not restricted to the injectivity radius of the exponential map.%
\footnote{
    Any feature vector $f(q)$ might therefore be represented multiple times on the same tangent space $\TpM$, once for each $v\in\TpM$ with $\exp(v)=q$.
    If this is not desired, one may restrict the kernel support to the injectivity radius of the exponential map, such that only the geodesically nearest occurrence will be measured.
}
We furthermore want to mention that the transporter may be replaced with any other isomorphism between $\A_{\exp(v)}$ and $\A_p$, as done for instance in~\cite{sommer2019horizontal}.

As stated before, kernel field transforms and $\GM$-convolutions are defined as matching the local feature field representations on the tangent spaces with kernels.
Working towards these definitions, note that the bundle $M$-morphisms of kernels $\K: \TM \to \Hom(\Ain,\Aout)$ and local field representations $\Expsfin: \TM \to \Ain$, can be combined to yet another (nonlinear) $M$-morphism from $\TM$ to $\Aout$,
\begin{equation}\label{eq:integrand_bundle_map}
    \quad
    \begin{tikzcd}[row sep=3.5em, column sep=6.5em]
        \TM  \arrow[rd, start anchor={[xshift=-1ex]}, "\piTM"']
            \arrow[r, "\K \mkern-1mu\times\mkern-1mu \Expsfin"]
        &
        \mkern-3mu
        \Hom(\Ain,\Aout) \!\times\! \Ain
            \arrow[r, "\ev"]
        &
        \Aout
            \arrow[ld, "\piAout"] \\
        & M &
    \end{tikzcd}
    \quad,
\end{equation}
where $\ev: \big(\K(v),\, \Expsfin(v)\big) \mapsto \K(v) \Expsfin(v)$ is the evaluation map on $\Hom(\Ain,\Aout) \!\times\! \Ain$.
Kernel field transforms compute output feature vectors at~$p$ by integrating this product of kernels and input fields over the respective tangent space $\TpM$:
\begin{dfn}[Kernel field transform]
\label{dfn:kernel_field_trafo}
    Let $\K$ be any smooth kernel field.
    The corresponding \emph{kernel field transform} is a smooth integral transform
    \begin{align}\label{eq:kernel_field_trafo_def_signature}
        \TK: \Gamma(\Ain)\to \Gamma(\Aout)
    \end{align}
    which is pointwise defined by%
    \footnote{
        The integration over $\TpM$ via the Riemannian volume density $dv$ is discussed in Appendix~\ref{apx:tangent_integral}.
    }
    \begin{align}\label{eq:kernel_field_trafo_def_ptwise}
        \big[ \TK (\fin)\big] (p)
        \,\ :=\, \int\limits_{\TpM}\!\!
            \K(v) \,
            \Expsfin (v)
            \ dv
        \,\ =\, \int\limits_{\TpM}\!\!
            \K(v) \ 
            \PAinexppv \; \fin(\exp_p\!v)
            \ dv \,.
    \end{align}
    In order to be well defined, the integral needs to exist and the resulting output field $\TK(f)$ needs to be smooth.
    This requires $\K$ to be chosen suitably, e.g. by assuming it to decay rapidly or to be compactly supported.
\end{dfn}
Note that general kernel field transforms do not necessarily model convolutions as they do not assume weights (kernels) to be shared between spatial positions.
Such general kernel field transforms will become handy in Section~\ref{sec:isometry_intro}, where we derive a requirement for spatial weight sharing from the requirement for isometry equivariance.

Appendix~\ref{apx:smoothness_kernel_field_trafo} discusses the existence and smoothness of kernel field transforms.
A sufficient condition for kernel field transforms to be well defined is the restriction of kernel supports to balls of a fixed radius $R>0$:
\begin{thm}[Kernel field transform existence for compactly supported kernels]
\label{thm:existence_kernel_field_trafo_compact_kernels}
    Let $\K$ be a kernel field whose individual kernels $\Kp$ at any $p\in M$ are (at most) supported on a closed ball of radius $R>0$ around the origin of $\TpM$, that is,
    \begin{align}
        \supp\!\big(\Kp\big)\ \subseteq\ \big\{ v\in\TpM \,\big|\, \lVert v\rVert \leq R \big\} \quad \forall p\in M \,.
    \end{align}
    The corresponding kernel field transform $\TK$ is then guaranteed to be well defined, i.e. the integral in Eq.~\eqref{eq:kernel_field_trafo_def_ptwise} exists and the output field $\TK(f) \in \Gamma(\Aout)$ is smooth for any smooth input field $f\in\Gamma(\Ain)$.
\end{thm}
\begin{proof}
    See Appendices~\ref{apx:smoothness_kernel_field_trafo} and~\ref{apx:proof_sufficiency_ball_kernel_support}.
\end{proof}
The requirement to restrict the kernel support to a closed ball of certain radius is common practice in deep learning.
Note, however, that a compactly supported kernel is at odds with scale equivariant convolutions, which, by the corresponding $G$-steerability kernel constraints, require infinitely far extending kernels.
Current implementations of scale equivariant convolutions usually approximate scale equivariant kernel spaces by restricting their support~\cite{marcos2018scale,Worrall2019DeepScaleSpaces,ghosh2019scale,zhu2019scale,bekkers2020bspline,Sosnovik2020scale,naderi2020scalesteerable} and are therefore covered by Theorem~\ref{thm:existence_kernel_field_trafo_compact_kernels}.

Based on general kernel field transforms, we define \emph{coordinate free $\GM$-convolutions} by adding the assumption of spatial weight sharing, i.e. by assuming $\GM$-\emph{convolutional kernel fields}:
\begin{dfn}[$\GM$-convolution]
\label{dfn:coord_free_conv}
    Let $\Ain$ and $\Aout$ be $G$-associated feature vector bundles with types $\rhoin$ and $\rhoout \,$, respectively.
    We define the \emph{$\GM$-convolution} with a $G$-steerable kernel $K\in\KG$ as the kernel field transform with the corresponding $\GM$-\emph{convolutional kernel field}~$\KK$:
    \begin{align}\label{eq:coord_free_conv_def_signature}
        K\,\star\,:\, \Gamma(\Ain)\to \Gamma(\Aout), \quad
        \fin \mapsto K\star \fin \,:=\, \TKK(\fin)
        \,=\! \int\limits_{\TpM}\! \KK(v)\ \Expsfin(v)\ dv
    \end{align}
\end{dfn}
As $\GM$-convolutions do not prefer any reference frame in the $G$-structure, they are guaranteed to generalize their inference over all ``poses'' of patterns which are related by the action of the structure group~$G$; see Eq.~\eqref{eq:active_local_gauge_trafo} and Fig.~\ref{fig:active_TpM_equivariance}.

\subsubsection{Kernel field transforms and \textit{GM}-convolutions in local coordinates}
\label{sec:KFTs_GM-conv_local}

What is left to show is that the coordinate free definitions of transporter pullbacks, kernel field transforms and $\GM$-convolutions introduced in this section reduce to the coordinate expressions from Section~\ref{sec:gauge_conv_main} when being expressed relative to some local trivialization.

The local coordinate expression of the transporter pullback $\Expsf$ of a feature field $f$ is, as usual, defined by pre- and post-composing it with local trivializations of the corresponding bundles, that is:
\begin{align}\label{eq:kft_coordinate_expression_62}
    \big[\Expsf \big]^A: U\times \R^d \to U\times \R^c,\ \ \ 
    (p,\mathscr{v}) \mapsto&\phantom{=} \PsiA^A \circ \Expsf \circ \big(\PsiTM^A \big)^{-1} (p, \mathscr{v}) \notag \\
                           &= \pig(p,\,\ \psiAp^A \circ \Expspf \circ \big(\psiTMp^A \big)^{-1} (\mathscr{v}) \pig)
\end{align}
Local gauge transformations at $p\in M$ are from this definition seen to be given by
\begin{align}
    \big[\Expspf \big]^B \ =\ \rho\big( g_p^{BA}\big) \circ \big[\Expspf \big]^A \circ \big(g_p^{BA}\big)^{-1} \,.
\end{align}
We visualize these coordinate expressions in terms of a commutative diagram, which is very similar to that for the local trivializations of kernel fields in Fig.~\ref{fig:triv_kernel_bundle_morphism}:
\begin{equation}
\begin{tikzcd}[row sep=4.5em, column sep=5.2em]
      U\times\R^d
                    \arrow[rrrr, pos=.48, rounded corners, to path={ 
                            -- ([yshift=2.5ex]\tikztostart.north) 
                            --node[above]{\small$
                                \big[\Expsf \big]^B
                            $} ([yshift=2.5ex]\tikztotarget.north) 
                            -- (\tikztotarget.north)
                            }]
    & &[-3.25em] &[-3.25em] &
    U\times\R^c
    \\
    U\times\R^d
                    \arrow[u, "\big(\id\times g^{BA}\!\!\cdot\big)"]
                    \arrow[rrd, "\proj_1"']
                    \arrow[rrrr, pos=.5, rounded corners, to path={ 
                            -- ([yshift=-16.ex]\tikztostart.south) 
                            --node[below]{\small$
                                \big[\Expsf \big]^A
                            $} ([yshift=-16.ex]\tikztotarget.south) 
                            -- (\tikztotarget.south)
                            }]
    &
    \piTM^{-1}(U)   
                    \arrow[rd, "\piTM\!\!"', pos=0.3]
                    \arrow[rr, "\Expsf"]
                    \arrow[lu, "\PsiTM^B"']
                    \arrow[l,  "\PsiTM^A"']
    & &
    \piA^{-1}(U)
                    \arrow[ld, "\!\piA", pos=0.3]
                    \arrow[r,  "\PsiA^A"]
                    \arrow[ru, "\PsiA^B"]
    &
    U\times\R^c
                    \arrow[u, "\big(\id\times \rho\big(g^{BA}\big)\!\cdot\big)"']
    \arrow[lld, "\proj_1"] \\
    & &
    U
    & &
\end{tikzcd}
\end{equation}

For an implementation it is useful to further resolve the coordinate expression of the transporter pullback into those of its individual components, i.e. of the transporter
$\mathcal{P}_{\mkern-4mu\overset{}{\protect\scalebox{.75}{$\!\A$},\protect\scalebox{.75}{$\, p\!\leftarrow\!\exp(v)$}}}$,
the feature field $f$ and the exponential map $\exp$.
This is achieved by expanding it with an identity of the form
$\id_{\A_{\exp(v)}} = \big(\psiAexpnop^{\widetilde{A}} \big)^{-1} \circ \psiAexpnop^{\widetilde{A}}$,
where the choice of gauge $\widetilde{A}$ at $\exp(v)$ is irrelevant as it ultimately drops out:
\begin{align}\label{eq:coordinate_expression_transporter_pullback_62}
    \big[\Expspf \big]^A (\mathscr{v})
    \ &=\ \pig[ \psiAp^A \circ \Expspf \circ \big(\psiTMp^A \big)^{-1} \pig] (\mathscr{v}) \notag \\
    \ &=\ \psiAp^A \circ 
        \mathcal{P}_{\mkern-4mu\overset{}{\protect\scalebox{.75}{$\!\A$},\protect\scalebox{.75}{$\, p\!\leftarrow\!\exp(v)$}}}
     \circ f \big( \exp \circ \big(\psiTMp^A\big)^{-1} (\mathscr{v})\big) \notag \\
    \ &=\ \psiAp^A \circ 
        \mathcal{P}_{\mkern-4mu\overset{}{\protect\scalebox{.75}{$\!\A$},\protect\scalebox{.75}{$\, p\!\leftarrow\!\exp(v)$}}}
     \circ \big(\psiAexpnop^{\widetilde{A}_v} \big)^{-1} \circ \psiAexpnop^{\widetilde{A}_v}
     \circ f \big( \exp \circ \big(\psiTMp^A\big)^{-1} (\mathscr{v})\big) \notag \\
    \ &=\ \rho\pig( g^{A\widetilde{A}}_{p\leftarrow\exp\circ (\psiTMp^A)^{-1}(\mathscr{v})} \pig) \mkern1mu
     \cdot f^{\widetilde{A}} \big( \exp \circ \big(\psiTMp^A\big)^{-1} (\mathscr{v})\big) \notag \\
\end{align}
As expected, we recover our definition from Eq.~\eqref{eq:transporter_pullback_in_coords} in Section~\ref{sec:observers_view}, which approves that Def.~\ref{dfn:Expf_pullback_field} is indeed its coordinate free counterpart.

The coordinate expression of a kernel field transform, which coincides with Eq.~\eqref{eq:kft_coord_expression} in Section~\ref{sec:observers_view}, is given by the following theorem:
\begin{thm}[Kernel field transform in coordinates]
\label{thm:kernel_field_trafo_in_coords}
    Relative to some gauge $A$ at $p\in U^A$, the kernel field transform is given by the coordinate expression
    \begin{align}\label{eq:kernel_field_trafo_in_coords}
        \!\big[ \TK(\fin) \big]^A (p)
        \ &=
        \int\limits_{\R^d}
        \Kp^A (v^A) \ 
        \big[\Expspfin\big]^A (v^A)
        \ \sqrt{|\eta_p^A|}\,\ dv^A
        \notag \\
        \ &=
        \int\limits_{\R^d}
        \Kp^A (v^A) \ 
        \rho\pig( g^{A\widetilde{A}}_{p\leftarrow\exp\circ (\psiTMp^A)^{-1}(v^A)} \pig)
         \cdot \fin^{\widetilde{A}} \big( \exp \circ \big(\psiTMp^A\big)^{-1} (v^A) \big)
        \ \sqrt{|\eta_p^A|}\ dv^A ,\!
    \end{align}
    where the gauges $\widetilde{A}$ at $\exp(v)$ are chosen arbitrarily as they cancel out.%
    \footnote{
        Note that the gauges at $\exp(v)$ might differ for different $v\in\TpM$ and should more correctly be labeled by $\widetilde{A}_v$.
        We suppress this dependency for brevity.
    }
\end{thm}
\begin{proof}
    The first expression is derived by a simple calculation which translates all involved quantities into their corresponding coordinate expressions:
    \begin{align}
        & \ 
            \big[ \TK(\fin) \big]^A (p) \notag \\[.5ex]
        \overset{(1)}{=} & \ \ 
            \psiAoutp^A \big[ \TK(\fin) \big] (p) \notag \\[.5ex]
        \overset{(2)}{=} & \ \ 
            \psiAoutp^A
            \int\limits_{\TpM}\mkern-4mu
            \Kp(v) \,
            \big[\Expspfin\big] (v)
            \,\ dv
        \notag \\[.5ex]
        \overset{(3)}{=} & \ \ 
            \psiAoutp^A
            \int\limits_{\R^d}
            \Kp \pig(\! \big(\psiTMp^A\big)^{-1}(v^A) \pig) \;
            \big[\Expspfin\big] \pig(\!\big( \psiTMp^A \big)^{-1} (v^A) \pig)
            \ \sqrt{|\eta_p^A|}\,\ dv^A
        \notag \\[.5ex]
        \overset{(4)}{=} & \ \ 
            \int\limits_{\R^d}
            \Big[ \psiAoutp^A \circ
            \Kp \pig(\! \big(\psiTMp^A\big)^{-1}(v^A) \pig) \circ
            \big( \psiAinp^A \big)^{-1} \Big]
            \Big[ \psiAinp^A \circ
            \big[\Expspfin\big] \circ \big( \psiTMp^A \big)^{-1} \Big] (v^A)
            \ \sqrt{|\eta_p^A|}\,\ dv^A
        \notag \\[.5ex]
        \overset{(5)}{=} & \ \ 
            \int\limits_{\R^d}
            \Big[ \psiHomp^A \circ \Kp \circ \big(\psiTMp^A\big)^{-1} \Big] (v^A) \,\ 
            \Big[ \psiAinp^A \circ \big[\Expspfin\big] \circ \big( \psiTMp^A \big)^{-1} \Big] (v^A)
            \,\ \sqrt{|\eta_p^A|}\,\ dv^A
        \notag \\[.5ex]
        \overset{(6)}{=} & \ \ 
            \int\limits_{\R^d}
            \Kp^A (v^A) \ 
            \big[\Expspfin\big]^A (v^A)
            \ \sqrt{|\eta_p^A|}\,\ dv^A
    \end{align}
    Step~(1) expresses the output feature vector at $p$ explicitly in terms of gauge $\psiAoutp^A$, acting on the coordinate free kernel field transform.
    This coordinate free expression is in step~(2) expanded as defined in Def.~\ref{dfn:kernel_field_trafo}.
    Step~(3) pulls the integral over $\TpM$ via the chosen gauge back to $\R^d$, which is in more detail described in Appendix~\ref{apx:tangent_integral}.
    Step~(4) inserts an identity map of the form $\id = \big(\psiAinp^A \big)^{-1} \circ \psiAinp^A$ and pulls $\psiAoutp^A$ into the integral
    while step~(5) identifies the definition of $\psiHomp^A$ from Eq.~\eqref{eq:Hom_bdl_triv_ptwise}.
    Lastly, we identify the coordinate expressions of $\Kp$ and $\Expspfin$ from Eqs.~\eqref{eq:kernel_field_general_coord_expression} and~\eqref{eq:kft_coordinate_expression_62}.

    The second expression follows from the first one by expanding the coordinate expression of the transporter pullback according to Eq.~\eqref{eq:coordinate_expression_transporter_pullback_62}.
\end{proof}

The coordinate expression for the coordinate free $\GM$-convolutions follows immediately:
\begin{thm}[$\GM$-convolutions in coordinates]
\label{thm:gauge_equiv_conv_from_coordinate_free}
    A coordinate free $\GM$-convolution
    $K\star:\, \Gamma(\Ain) \to \Gamma(\Aout)$
    with a $G$-steerable kernel $K\in\KG$ is relative to some gauge $A$ at $p\in U^A$ given by
    \begin{align}
        &\mkern-40mu
        \big[ K\star f \big]^A (p)
        \ =\ \big[ \TKK(f) \big]^A(p)
        \ =\ \int\limits_{\R^d}
            K \mkern-1mu (v^A) \,\ 
            \big[ \Expspf \big]^A (v^A)
            \,\ dv^A \,,
    \end{align}
    that is, by the coordinate expression that was introduced in Eq.~\eqref{eq:gauge_conv_coord_expression}.
    This expression may be written out further as done for general kernel field transforms in Eq.~\eqref{eq:kernel_field_trafo_in_coords}.
\end{thm}
\begin{proof}
    The result follows from Theorem~\ref{thm:kernel_field_trafo_in_coords} by observing that the coordinate free $\GM$-convolution $K\star$ is just a kernel field transform with the corresponding $\GM$-convolutional kernel field $\KK$; see Def.~\ref{dfn:coord_free_conv}.
    Specifically, the coordinate expression of a $\GM$-convolutional kernel field $\KK$ is according to Def.~\ref{dfn:conv_kernel_field} given by the frame volume normalized $G$-steerable kernel $K$, that is, $\KKp^A = K/ \sqrt{|\eta_p^A|}$.
    Inserting this identity in Eq.~\eqref{eq:kernel_field_trafo_in_coords} leads to the claimed coordinate expression for $\GM$-convolutions.
\end{proof}

This result assures that a global, coordinate free $\GM$-convolution can be implemented in terms of its local coordinate expressions relative to some $G$-atlas of local trivializations that cover $M$.

%% file: chapters/80_isom_intro.tex

\section{Isometry equivariance}
\label{sec:isometry_intro}

A main characteristic of the convolution operation and its various generalizations is their equivariance w.r.t. symmetries of the underlying manifold.
For instance, the conventional convolution on Euclidean spaces is translation equivariant while spherical convolutions are rotation equivariant.
More generally, any locally compact group and their homogeneous spaces admit group convolutions%
~\cite{gurarie1992symmetries,kowalski2010introduction,chirikjian2001engineering,gallier2019harmonicRepr},
which were recently picked up by the deep learning community to generalize convolutional networks to such spaces~\cite{Cohen2016-GCNN,Kondor2018-GENERAL,Cohen2019-generaltheory,bekkers2020bspline}.
However, as these approaches rely fundamentally on the global, \emph{transitive} symmetries of the homogeneous space, they do not immediately apply to general Riemannian manifolds.

$\GM$-convolutions on the other hand shift the focus from \emph{global symmetries of the space itself} to \emph{local symmetries in the coordinatization of the space}.
As it turns out, the local gauge equivariance of $\GM$-convolutions, together with convolutional weight sharing, induces their equivariance under \mbox{the action of global symmetries.}
Stated more precisely, $\GM$-convolutions are equivariant under the action of \emph{$G$-structure preserving isometries} (Def.~\ref{dfn:IsomGM}), which form a subgroup $\IsomGM \leq \IsomM$ of the full isometry group.
The requirement on the symmetry to be an isometry (i.e. to preserve the metric) comes hereby from the use of exponential maps, which rely on the Levi-Civita connection and thus Riemannian metric.
The additional requirement on these isometries to preserve the $G$-structure is a consequence of the definition of feature vector bundles as associated $G$-bundles, whose elements have a well defined meaning only relative to those reference frames that are contained in~$\GM$.
Note that the latter is not really a restriction, as one may always choose structure groups $G\geq\O{d}$, for which \emph{any} isometry respects the corresponding $G$-structure.
On the contrary, this design allows for a precise control of the level of isometry equivariance.
For instance, the conventional convolution on Euclidean vector spaces relies on the canonical $\{e\}$-structure of $\R^d$, visualized in Fig.~\ref{fig:frame_field_automorphism_1}, and is therefore solely translation equivariant.
An $\SO{d}$-structure on $\R^d$, visualized in Fig.~\ref{fig:SO2_structure_SE2}, is additionally preserved by rotations, and thus corresponds to $\SE{d}$-equivariant convolutions.
Equivariance under the full isometry group $\E{d}$ of $\R^d$ is implied when choosing an $\O{d}$-structure on $\R^d$.

\etocsettocdepth{3}
\etocsettocstyle{}{} 
\localtableofcontents

The goal of this section is to derive theorems which formally characterize the isometry equivariance of $\GM$-convolutions and kernel field transforms.
Section~\ref{sec:isom_background} lays the foundations of this investigation by introducing isometry groups of Riemannian manifolds and discussing a range of well known relation and constructions which they induce.
Specifically, Section~\ref{sec:isometry_groups} introduces isometries and isometry groups while Section~\ref{sec:isom_action_bundles} defines their induced action (``pushforwards'') on the associated bundles in a coordinate free setting.
In Section~\ref{sec:isom_coordinatization} we express these actions on bundles relative to local trivializations and discuss their passive interpretation as isometry induced gauge transformations, visualized in Figs.~\ref{fig:intro_gauge_isom_induction} (right) and~\ref{fig:pushforward_vector_components}.
Section~\ref{sec:isom_expmap_transport} briefly states how the quantities involved in kernel field transforms behave under the action of isometries.

Based on these properties, we study the isometry equivariance of kernel field transforms and $\GM$-convolutions in Section~\ref{sec:isometry_equivariance}.
After defining the term ``isometry equivariance'' formally, Section~\ref{sec:isometry_constraint} proves a central result, which asserts that the demand for \emph{isometry equivariance requires the invariance of the kernel field under isometries}; see Fig.~\ref{fig:isom_invariant_kernel_field_multiple_orbits}.
Section~\ref{sec:isom_equiv_GM_conv} considers the more specific $\GM$-convolutions and proves that they are by design equivariant under any isometry that preserves the $G$-structure.
This result implies in particular, that $\OM$-convolutions are equivariant w.r.t. any isometry.

The invariance constraint on kernel fields enforces that they share weights over the orbits of the isometry group.
This suggests that invariant kernel fields can equivalently be described by representative kernels on orbit representatives, which we formalize in Section~\ref{sec:quotient_kernel_fields}.
Section~\ref{sec:isom_quotients} discussed isometry induced quotient spaces and their representatives.
In Section~\ref{sec:quotient_kernels_stabilizers} we use these mathematical definitions to prove that the space of isometry invariant kernel fields is indeed isomorphic to kernel fields on quotient representatives.
This implies in particular that isometry equivariant kernel field transforms on homogeneous spaces are necessarily convolutions, which closes the loop to prior work.

%% file: chapters/81_isom_action.tex

\subsection{Isometries and their action on manifolds, bundles and fields}
\label{sec:isom_background}

\begin{figure}
    \centering
    \subcaptionbox{\small Action of different subgroups of the isometry group on fields.
        \label{fig:isom_egg_actions}}%
        [.6\linewidth][l]{
            \includegraphics[width=.575\textwidth]{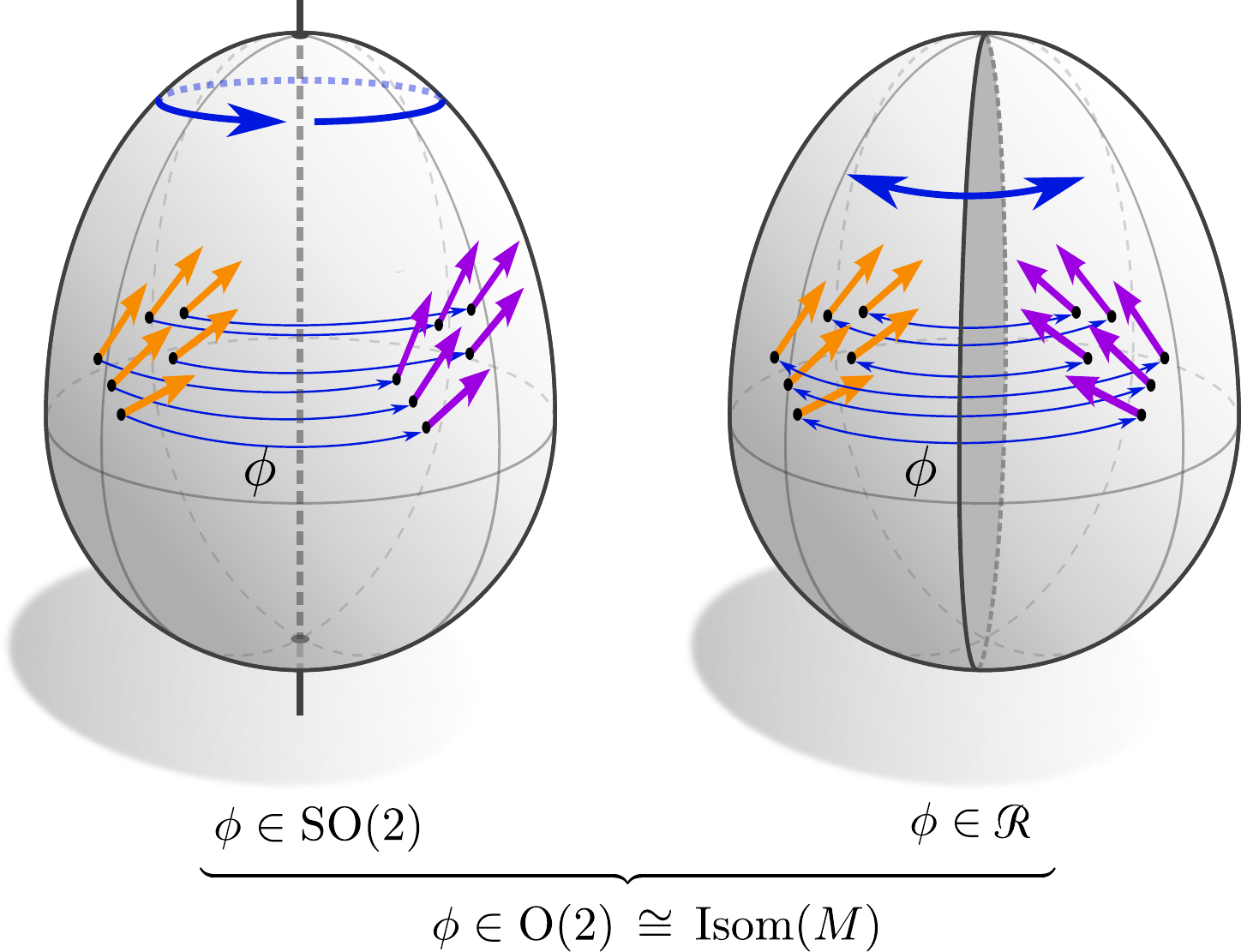}
        }
    \hfill
    \subcaptionbox{\small Orbits of the isometry group.
        \label{fig:isom_egg_orbits}}%
        [.3\linewidth][r]{
            \includegraphics[width=.26\textwidth]{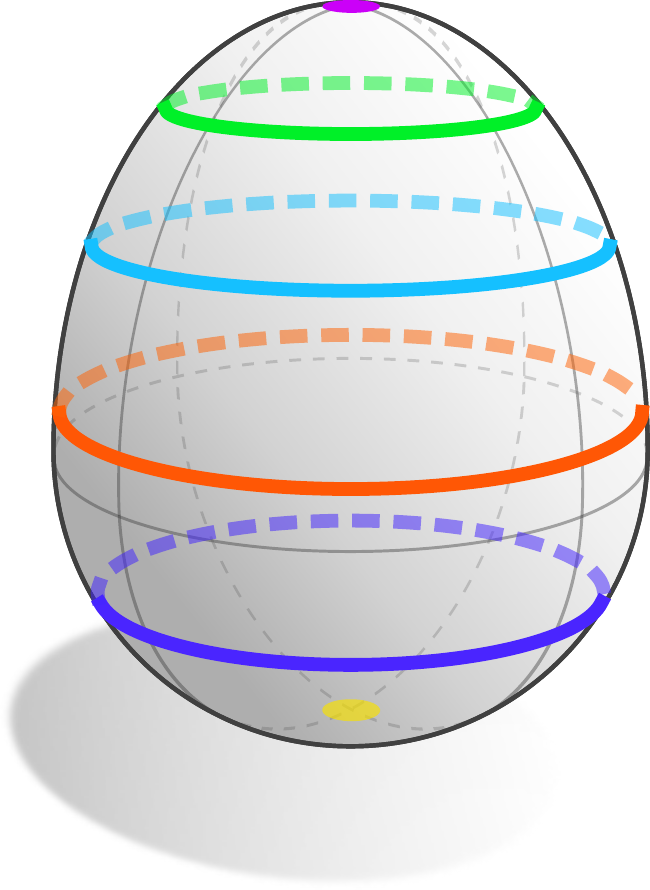}
            \hspace*{2ex}
            \vspace*{7.0ex}
        }
    \caption{\small
        Visualizations of the isometry group $\IsomM \cong \O2$ of an egg $M$, which we will use throughout this section to exemplify different concepts and constructions relating to isometries.
        Fig.~\ref{fig:isom_egg_actions} shows the action of the isometry group on (tangent or feature) vector fields.
        It can be thought of as consisting of the subgroups of rotations in $\SO2$ and reflections in $\Flip$.
        The action of the isometry group partitions the egg into orbits $\IsomM.p = \big\{\phi(p) \,\big|\, \phi\in\IsomM\big\}$ of points $p\in M$, shown in Fig.~\ref{fig:isom_egg_orbits} in different colors.
        Note that not all orbits are homeomorphic to each other -- the orbits at the poles are single points while any other orbit traces out a circle around the egg.
        The isometry group of the egg acts non-transitively on it, that is, not every point can be reached from any other point.
        A~kernel field transform is isometry equivariant if it commutes with the isometry action on feature fields.
        We show that isometry equivariance is guaranteed if and only if the kernel field is invariant under the action of isometries.
        This implies in particular that isometry equivariance require weight sharing along the isometry orbits; see Fig.~\ref{fig:isom_invariant_kernel_field_multiple_orbits}.
    }
    \label{fig:isom_egg_main}
\end{figure}

In this section we introduce most of the mathematical concepts required for our study of the isometry equivariance of kernel field transforms and $\GM$-convolutions.
After defining isometries in Section~\ref{sec:isometry_groups}, we discuss in Section~\ref{sec:isom_action_bundles} how they induce natural actions on tangent vectors and reference frames.
For structure groups $G<\O{d}$, not any isometry is compatible with any $G$-structure.
We define the subgroup $\IsomGM \leq \IsomM$ of those isometries which do act on (induce automorphisms of) a $G$-structure $\GM$ and their $G$-associated feature bundles.
While these constructions are kept coordinate free, Section~\ref{sec:isom_coordinatization} expresses the action of isometries on fiber bundles relative to local bundle trivializations.
In preparation for our investigation of isometry equivariant kernel field transforms later on, Section~\ref{sec:isom_expmap_transport} discusses how isometries commute with the exponential map and with parallel transporters, which allows to derive how isometries act on the transporter pullback $\Expspf$ of feature fields~$f$.
While staying mostly mathematical, we try to draw connections to the application wherever possible.

\subsubsection{Isometry groups}
\label{sec:isometry_groups}

A (global) \emph{isometry} $\phi: M \to \hat{M}$ is a diffeomorphism between Riemannian manifolds $(M,\eta)$ and $(\hat{M},\hat{\eta})$, which \emph{preserves the metric}.
In terms of the pushforward (differential) $\dphiTM: \TM \to T\mkern-1.5mu\hat{M}$ of tangent vectors, which we introduce in Appendix~\ref{apx:differentials_gradients_jacobians} and in Section~\ref{sec:isom_action_bundles} below, this statement is made precise by requiring that isometries satisfy
\begin{align}\label{eq:isometry_def}
    \eta_p(v,\, w)\ =\ \hat{\eta}_{\phi(p)}( \dphiTM v,\, \dphiTM w) \qquad \forall\ \ p\in M,\,\ v,w\in \TpM \,,
\end{align}
i.e. that they preserve distances and angles between tangent vectors.
Intuitively, an isometry is thought of as a distance preserving map between manifolds.
Note that the inverse of an isometry is necessarily an isometry as well.
Since isometries (and their inverses) respect the metric, they constitute the \emph{isomorphisms in the category of Riemannian manifolds}.

The set of all isometries $\phi: M \to M$ from a Riemannian manifold to itself, equipped with the usual function composition $\circ: (\phi_1, \phi_2) \mapsto \phi_1 \circ \phi_2$, defines a group, known as \emph{isometry group} $\IsomM$ of~$M$.
This group is the automorphism group of a Riemannian manifold, which contains all of its (metric) ``symmetries''.
It is a subgroup of the diffeomorphism group $\Diff(M)$ of~$M$.
The full isometry group might have non-trivial subgroups, which we will in the following denote by $\I \leq \IsomM$.
An example is given in Fig.~\ref{fig:isom_egg_actions}, which visualizes the isometry group $\IsomM \cong \O2$ of an egg.
The full isometry group splits (for instance) into the subgroups of rotations in $\I_1 \cong \SO2$ and reflections in $\I_2 \cong \Flip$.

\begin{SCfigure}
    \centering
    \hspace{1.ex}
    \includegraphics[width=.42\textwidth]{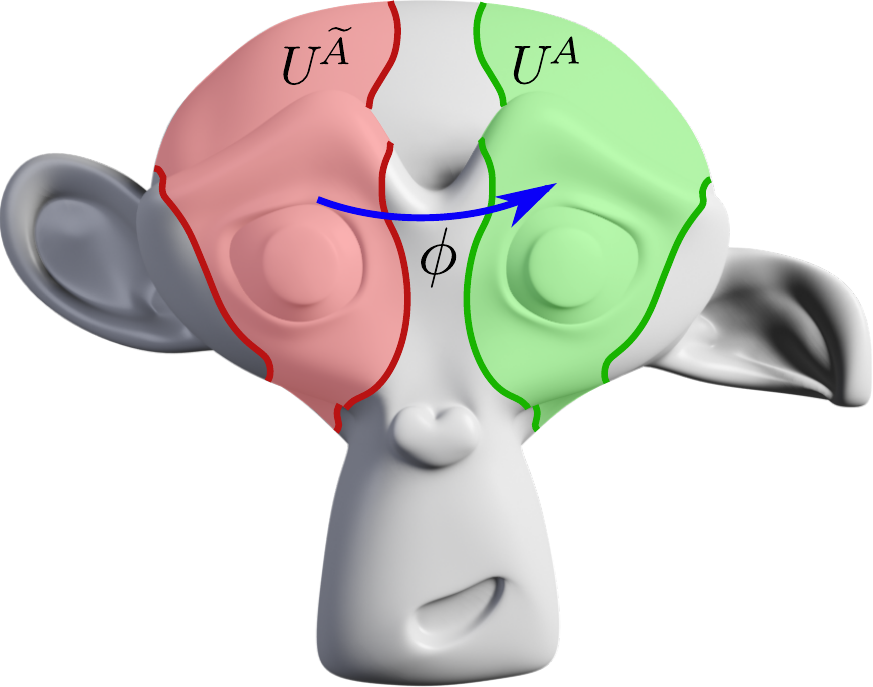}
    \hspace{2.ex}
    \captionsetup{width=.89\textwidth}
    \caption[]{\small
        An asymmetric manifold, whose \emph{global} isometry group is trivial.
        Since the asymmetry is limited to the ears and the mouth of
        ``\href{https://en.wikipedia.org/wiki/Blender_(software)\#Suzanne}{Suzanne}'',
        the monkey, there are non-trivial localized symmetries left.
        For instance, the smooth map ${\phi: U^{\widetilde{A}} \to U^A}$ between the red and green highlighted subsets preserves the metric locally.
        All concepts developed in Section~\ref{sec:isom_background} as well as the isometry equivariance of point-wise operations like \onexones\ generalize immediately to such isometries between local subsets. 
        The isometry equivariance of kernel field transforms with spatially extended kernels generalizes up to boundary effects.
        }
    \label{fig:suzanne_local_isometry}
\end{SCfigure}

In general, the isometry group of a manifold is non-transitive, that is, not every point of $M$ can be reached from any other point by its action.
The manifold is then partitioned into disjoint \emph{orbits}, visualized for the example of $M$ being an (Easter) egg in Fig.~\ref{fig:isom_egg_orbits}.
The isometry group of a manifold $M$ might be trivial, given that $M$ is sufficiently asymmetric.
In this case there might still exist non-trivial isometries between open subsets $U^{\widetilde{A}}$ and $U^A$ of $M$, restricted to which Eq.~\eqref{eq:isometry_def} holds.
Fig.~\ref{fig:suzanne_local_isometry} shows an example of a manifold which is globally asymmetric but has non-trivial isometries between local subsets of itself.
We will in the following only consider global isometries of~$M$, however, all concepts of the current Section~\ref{sec:isom_background} generalize in an obvious way to isometries between local subsets.
Without proof, we claim that the same holds for the isometry equivariance of any neural network operation which acts pointwise, for instance \onexones, nonlinearities or bias summation.
The equivariance of kernel field transforms with spatially extended kernels holds up to boundary effects.

\subsubsection{Isometry action on fiber bundles}
\label{sec:isom_action_bundles}

Isometries act naturally on tangent vectors in~$\TM$ and reference frames in~$\FM$ by ``carrying them along'' with the group action as visualized in Fig.~\ref{fig:isom_egg_actions}.
If an isometry is in addition compatible with the $G$-structure, that is, if it gives rise to an automorphism of~$\GM$, it furthermore acts on any associated $G$-bundle, in particular the feature vector bundles~$\A$.
We discuss these actions of isometries on the associated bundles and on feature fields in the following.

\paragraph{Isometry action on the tangent bundle \textit{TM}:}

Any isometry $\phi \in \IsomM$ gives rise to a \emph{pushforward}
\begin{align}
    \dphiTM: \TM \to \TM \,, \qquad \phi\in \IsomM
\end{align}
on the tangent bundle, which is just the differential of $\phi$ as introduced in Appendix~\ref{apx:differentials_gradients_jacobians}.
It can at each point $p \in M$ be thought of as a \emph{linear} approximation of $\phi$, which maps vectors $v \in \TpM$ to $\dphiTM(v) \in T_{\mkern-1mu\phi(p)}\mkern-2mu M$, that is, it satisfies
\begin{align}\label{eq:pushfwd_bundle_automorphism}
    \piTM \circ \dphiTM \,=\, \phi \circ \piTM.
\end{align}
As argued in Appendix~\ref{apx:differentials_gradients_jacobians}, the pushforward is invertible with $(\dphiTM)^{-1} = (\phi^{-1})_{\mkern-2mu*\mkern-1mu\scalebox{.55}{$,\mkern-2muT\mkern-3muM$}}$, for which we will unambiguously write $\dphiTMinv$.%
\footnote{
    The invertibility does not hold for pushforwards in general but only for those of diffeomorphisms and thus isometries.
}
The pushforward of an element $\phi$ of the isometry group is therefore seen to be an (isometric) vector bundle automorphism of $\TM$ over $\phi$, satisfying the following commutative diagram:
\begin{equation}\label{cd:pushforward_TM}
\begin{tikzcd}[column sep=70pt, row sep=40, font=\normalsize]
    \TM
        \arrow[r, shift left=2.5pt, "\dphiTM"]
        \arrow[d, "\piTM"']
    &
    \TM
        \arrow[d, "\piTM"]
        \arrow[l, shift left=2.5pt, "\dphiTMinv"]
    \\
    M
        \arrow[r, shift left=2.5pt, "\phi"]
    &
    M
        \arrow[l, shift left=2.5pt, "\phiinv"]
\end{tikzcd}
\end{equation}
By the \emph{definition of isometries}, their pushforward preserves distances and angles, that is, 
\begin{align}\label{eq:metric_pushfwd_isometry}
    \eta_{\phi(p)} \big(\dphiTM v,\, \dphiTM w\big) \,=\, \eta_p(v,w)
    \qquad\ \forall\ \ p\in M,\ \  v,w\in \TpM,\ \ \phi \in \IsomM .
\end{align}
More details about pushforwards between tangent bundles are easily found in the literature, for instance in~\cite{schullerGeometricalAnatomy2016}.

\paragraph{Isometry action on the frame bundle \textit{FM}:}
The pushforward on $\TM$ immediately induces a corresponding principal bundle automorphism $\dphiFM$ on $\FM$ by pushing forward the individual frame vectors:
\begin{align}\label{eq:pushforward_FM_def}
    \dphiFM\!: \FM \to \FM,\ \ \ 
    [e_i]_{i=1}^d \mapsto \dphiFM\big([e_i]_{i=1}^d\big) := \big[\dphiTM(e_i)\big]_{i=1}^d \,,
    \qquad \phi\in \IsomM
\end{align}
It maps frames in $\FpM$ for arbitrary $p\in M$ to frames at $F_{\mkern-1mu\phi(p)}\mkern-2mu M$, that is, $\piFM \circ \dphiFM = \phi \circ \piFM$.
To see this, let $[e_p]_{i=1}^d \in \FpM$, then $\phi\circ \piFM\big([e_i]_{i=1}^d\big) = \phi(p)$ and
$ \piFM \circ \dphiFM \big([e_i]_{i=1}^d\big)
= \piFM \pig(\big[ \dphiTM(e_i) \big]_{i=1}^d \pig)
= \piTM \circ \dphiTM (e_j)
= \phi \circ \piTM(e_j)
= \phi(p)$
for any $j=1,\dots,d$.
It can further be checked to be invertible with
$(\dphiFM)^{-1} = (\phi^{-1})_{\mkern-2mu*\mkern-1mu\scalebox{.55}{$,\mkern-2mu F\mkern-3muM$}}$,
again abbreviated by $\dphiFMinv$.
The \emph{left action} of the $\dphiFM$ on the frame bundle commutes with the \emph{right action} $\lhd$ on its fibers, that is, for arbitrary $g\in \GL{d}$ and $\phi \in \IsomM$ we have that:
\begin{alignat}{3}
\label{eq:dpsiFM_right_GL_equiv}
    \qquad
    \Big(\dphiFM \pig( [e_i]_{i=1}^d \pig)\Big) \lhd g
    \ &=\ \big[\dphiTM (e_i)\big]_{i=1}^d \lhd g
        \qquad\quad && \big( \text{\small def. of $\dphiFM$, Eq.~\eqref{eq:pushforward_FM_def}} \big) \notag \\
    \ &=\ \Big[\sum\nolimits_j \dphiTM (e_j)\, g_{ji} \Big]_{i=1}^d
        \qquad\quad && \big( \text{\small def. of $\lhd$, Eq.~\eqref{eq:rightaction_FM} } \big) \notag \\
    \ &=\ \Big[\dphiTM \Big(\sum\nolimits_je_j g_{ji}\Big) \Big]_{i=1}^d
        \qquad\quad && \big( \text{\small linearity of $\dphiTM$ } \big) \notag \\
    \ &=\ \dphiFM \Big(\Big[\sum\nolimits_je_j g_{ji}\Big] \Big)_{i=1}^d
        \qquad\quad && \big( \text{\small def. of $\dphiFM$, Eq.~\eqref{eq:pushforward_FM_def}} \big) \notag \\
    \ &=\ \dphiFM \Big( [e_i]_{i=1}^d \lhd g \Big)
        \qquad\quad && \big( \text{\small def. of $\lhd$, Eq.~\eqref{eq:rightaction_FM} } \big)
\end{alignat}
A gauge transformation of a frame at $p\in M$ by $g\in \GL{d}$, followed by a pushforward to $\phi(p)$, is therefore equal to a pushforward of the untransformed frame, followed by a gauge transformation by the same group element $g$ but at $\phi(p)$.
Different frames in the fiber $\FpM$ are hence mapped in such a way to frames at $F_{\mkern-1mu\phi(p)}\mkern-2mu M$ that their relative offset is preserved.
The derived properties of $\dphiFM$ are summarized by the statement that the diagram
\begin{equation}\label{eq:FM_isom_induced_automorphism}
\begin{tikzcd}[column sep=70pt, row sep=35, font=\normalsize]
    \FM
        \arrow[r, "\dphiFM"]
    &
    \FM
    \\
    \FM
        \arrow[r, "\dphiFM"]
        \arrow[d, "\piFM"']
        \arrow[u, "\lhd g"]
    &
    \FM
        \arrow[d, "\piFM"]
        \arrow[u, "\lhd g"']
    \\
    M
        \arrow[r, "\phi"']
    &
    M
\end{tikzcd}
\end{equation}
commutes for any $\phi\in\IsomM$ and any $g\in \GL{d}$.
Satisfying the commutativity of this diagram, the pushforward $\dphiFM$ on the frame bundle is identified as a \emph{principal bundle automorphism}%
\footnote{
    I.e. a principal bundle isomorphism from the frame bundle to itself; cf. Eq.~\ref{cd:principal_bundle_morphism}.
}
over $\phi$.

Note that the inverses, which are shown explicitly in the diagram~\eqref{cd:pushforward_TM}, are omitted to reduce clutter.

\paragraph{Isometry action on \textit{G}-structures \textit{GM}:}

\begin{figure}
    \centering
    \subcaptionbox{$\SE2$-invariant $\SO2$-structure $\SOM$ over $M = \R^2$.
        \label{fig:SO2_structure_SE2}}%
        [.5\linewidth][l]{
            \includegraphics[width=.5\textwidth]{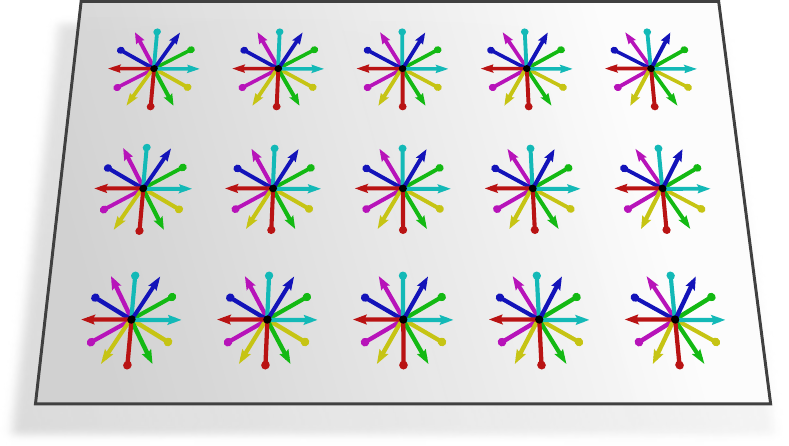}
            \rule{0pt}{20pt}
        }
    \subcaptionbox{$\SO3$-invariant $\SO2$-structure $\SOM$ over $M = S^2$\!\!.
        \label{fig:SO2_structure_SO3}}%
        [.48\linewidth][l]{
            \includegraphics[width=.4\textwidth]{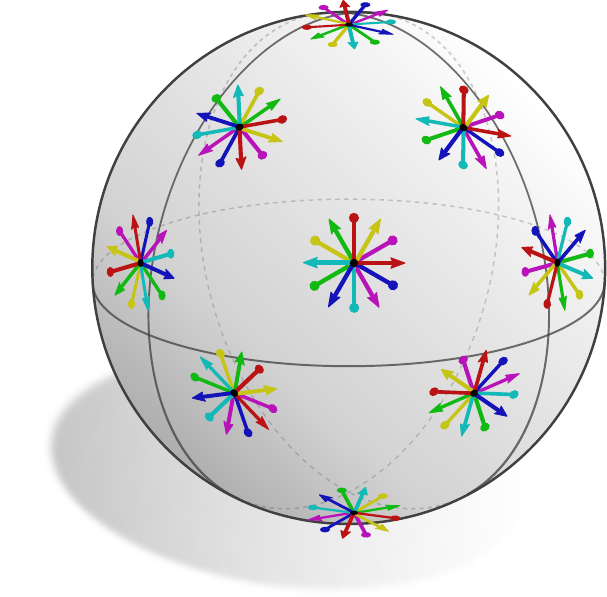}
        }
    \caption[]{\small
        Two examples of $\SO2$-structures $\SOM$ over the plane $M = \R^2$ and the sphere $M = S^2$.
        For ${M = \R^2}$, shown in Fig.~\ref{fig:SO2_structure_SE2}, the $\SO2$-structure is invariant under translations and rotations.
        As it consists only of right-handed frames (mind the arrow tips on the first and the circle tips on the second frame axes) it is not invariant under reflections.
        The isometries which preserve $\SOM$ therefore form the group $\IsomSOM = \SE2$, which is a subgroup of the full isometry group $\IsomM = \E2$.
        In the case of $M = S^2$, shown in Fig.~\ref{fig:SO2_structure_SO3}, the $\SO2$-structure is invariant under rotations but not under reflections.
        The $\SO2$-structure automorphisms are here $\IsomSOM = \SO3$ while the full isometry group is $\IsomM = \O3$.
        }
    \label{fig:SO2_structures_SE2_SO3}
\end{figure}

As $G$-structures are principal subbundles of the frame bundle, one can consider the restriction of the domain of the pushforward on~$\FM$ to~$\GM$, that is,
\begin{align}
    \dphiFM \mkern-2mu \big|_{\scalebox{.58}{$\GM$}} :\ \GM \to \FM \,, \qquad \phi\in \IsomM \,.
\end{align}
It is hereby necessary to keep the full frame bundle~$\FM$ as codomain since there is in general no guarantee that frames in~$\GpM$ are mapped to frames of $\GphipM$ but only to $\FphipM$.
\begin{figure}
    \centering
    \subcaptionbox{Canonical $\{e\}$-structure of $\R^2$
        \label{fig:frame_field_automorphism_1}}%
        [.49\linewidth][l]{
            \includegraphics[width=.46\textwidth]{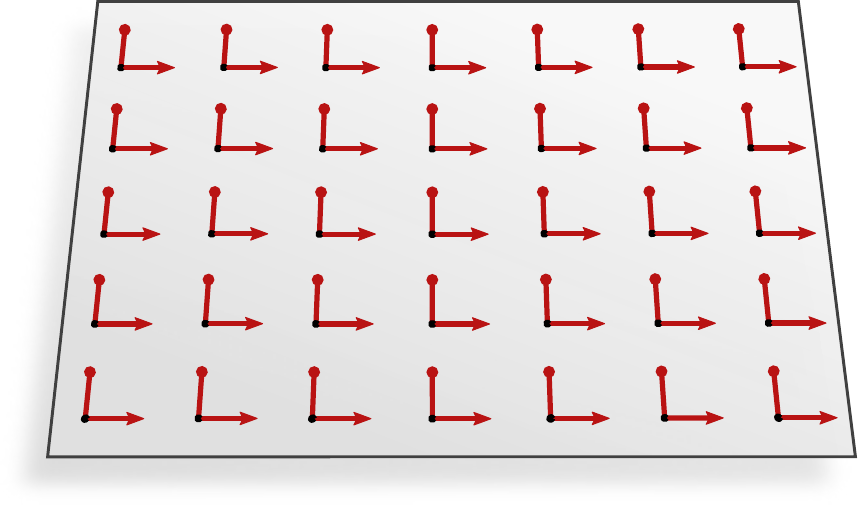}
        }
    \subcaptionbox{An alternative $\{e\}$-structure on $\R^2$
        \label{fig:frame_field_automorphism_2}}%
        [.49\linewidth][l]{
            \includegraphics[width=.46\textwidth]{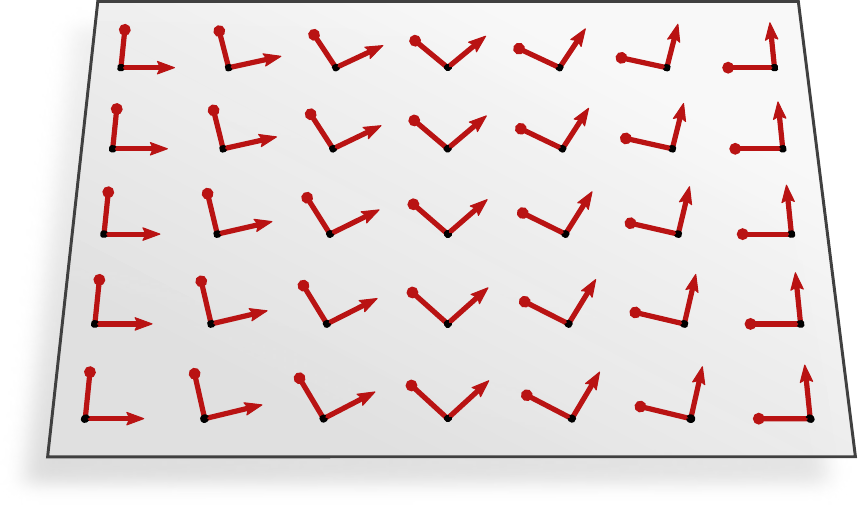}
        }
    \caption[]{\small
        Two specific choices of $\{e\}$-structures (global frame fields) $\eM$ on $M = \R^2$, which we use to visualize the concept of \emph{$G$-structure preserving isometries}.
        The full isometry group of~$M$ is the Euclidean group~$\IsomM=\E2$, consisting of translations, rotations and reflections.
        Fig.~\ref{fig:frame_field_automorphism_1} shows the canonical $\{e\}$-structure of $\R^2$, which is invariant under translations but not under rotations or reflections.
        More abstractly stated, translations make up the subgroup $\IsomeM = \Trans_2 := (\R^2,+)$ of isometries that induce automorphisms of $\eM$.
        In contrast, rotations or reflections map frames in $\epM$ to frames in $\FphipM$ but fail to send them to~$\ephipM$.
        They do therefore not induce automorphisms of the $\{e\}$-structure and are not part of $\IsomeM$.
        Group actions of such isometries on~$\eM$ or any of its $\{e\}$-associated bundles \emph{are not defined}.
        Fig.~\ref{fig:frame_field_automorphism_2} shows an alternative choice of $\{e\}$-structure on $M = \R^2$ (or $M=\Euc_2$), which is only invariant under translations in the ``up-down'' direction, i.e. $\IsomeM \cong \Trans_1 = (\R,+)$.
        The examples in Figs.~\ref{fig:frame_field_automorphism_1} and~\ref{fig:frame_field_automorphism_2} exemplify that the $G$-structure automorphisms do not only depend on the structure group $G$ but on the particular choice of $G$-structure~$\GM$.
        The general case for $G$ being non-trivial is harder to visualize since $\GpM$ will then not be a single frame but a set of frames.
        }
    \label{fig:frame_field_automorphism}
\end{figure}
Since $G$-structures are in general not closed under the action of isometries on~$\FM$, it might be \emph{impossible} to define a group action of the full isometry group on~$\GM$ or any other associated $G$-bundle.
To remedy this shortcoming, we will in the following consider the subgroup of those isometries that respect the $G$-structure, i.e. which map preferred frames in~$\GM$ to frames in~$\GM$.
\begin{dfn}[$G$-structure preserving isometries]
\label{dfn:IsomGM}
    Given a $G$-structure $\GM$, we define the corresponding subgroup of $G$-structure preserving isometries $\IsomGM$ as:
    \begin{align}\label{eq:isomGM_def}
        \IsomGM\ :=\ \big\{ \phi \in \IsomM \,\big|\, \dphiFM(\GpM) = \GphipM\ \ \ \forall p \in M \big\} \ \leq\ \IsomM
    \end{align}
\end{dfn}
For such isometries, we define the induced action on~$\GM$ as
\begin{align}
    \dphiGM\, :=\, \dphiFM \mkern-2mu \big|_{\scalebox{.58}{$\GM$}} \,:\ \GM \to \GM \,, \qquad \phi\in \IsomGM \,.
\end{align}
Such defined actions for $\phi \in \IsomGM$ are \emph{$G$-structure automorphisms}, that is, they make the following diagram commute for any $g\in G$ (which follows by restricting Eq~\eqref{eq:FM_isom_induced_automorphism} from $\FM$ to $\GM$ and $\GL{d}$ to $G$):
\begin{equation}\label{cd:isom_induced_GM_automorphism}
\begin{tikzcd}[column sep=70pt, row sep=35, font=\normalsize]
    \GM
        \arrow[r, "\dphiGM"]
    &
    \GM
    \\
    \GM
        \arrow[r, "\dphiGM"]
        \arrow[d, "\piGM"']
        \arrow[u, "\lhd g"]
    &
    \GM
        \arrow[d, "\piGM"]
        \arrow[u, "\lhd g"']
    \\
    M
        \arrow[r, "\phi"']
    &
    M
\end{tikzcd}
\end{equation}
Fig.~\ref{fig:frame_field_automorphism} shows two examples of $\{e\}$-structures on $M=\R^2$, i.e. global frame fields.
From these examples it is apparent that the subgroups $\IsomGM$ do really depend on the particular choice of $G$-structure $\GM$, not only on the structure group~$G$.
In Fig.~\ref{fig:SO2_structure_SE2} we visualize an $\SO2$-structure on $M=\R^2$.
Its isometry group $\IsomSOM = \SE2$ is larger than those of the $\{e\}$-structures in Fig.~\ref{fig:frame_field_automorphism}.
An $\SO2$-structure on the sphere $S^2$, which is preserved by all rotations $\IsomSOM = \SO3$, is shown in Fig.~\ref{fig:SO2_structure_SE2}.

For specific choices of structure groups $G$ it is possible to make more general statements about which isometries are contained in the subgroup $\IsomGM$.
Most importantly, for orthonormal structure groups $G=\O{d}$ (which are compatible with $\eta$) \emph{any} isometry will induce an automorphism of~$\OM$, that is, one always has $\IsomOM = \IsomM \,.$
To prove this claim, let $[e_i]_{i=1}^d \in \OpM \subset \FpM$ be an orthonormal frame, which is by an arbitrary isometry $\phi \in \IsomM$ being sent to $\dphiFM \big|_{\scalebox{.58}{$\OM$}} [e_i]_{i=1}^d = \big[\dphiTM e_i\big]_{i=1}^d$; see Eq.~\eqref{eq:pushforward_FM_def}.
Applying Eq.~\eqref{eq:metric_pushfwd_isometry} to the individual axes of the pushforward frame yields
\begin{align}\label{eq:isom_orthonormal_to_orthonormal_frames}
    \eta\big( \dphiTM e_i, \dphiTM e_j \big) \,=\, \eta(e_i, e_j) \,=\, \delta_{ij}\ \quad\ \forall\ \ i,j \in 1,\dots,d \,,
\end{align}
which implies the orthonormality of of the pushforward frame $\dphiFM \big|_{\scalebox{.58}{$\OM$}} [e_i]_{i=1}^d \in \OphipM$ and therefore allows to define $\dphiOM := \dphiFM \big|_{\scalebox{.58}{$\OM$}}$ for any $\phi \in \IsomM$.
More generally, this result implies:
\begin{align}\label{eq:isomM_isomOM}
    \IsomGM = \IsomM \quad \ \forall\ G \geq \O{d}
\end{align}
It is similarly possible to show
\begin{align}\label{eq:isomplusM_isomSOM}
    \IsomSOM = \IsomplusM \,,
\end{align}
that is, that any orientation preserving isometry in $\IsomplusM$ induces an automorphism of an $\SO{d}$-structure $\SOM$.
Note that these statements all depend only on the structure group~$G$ but are independent from the specific choice of $G$-structure.
This is ultimately a result of only considering isometries, which are adapted to $\O{d}$-structures by definition, instead of considering more general diffeomorphisms.
As mentioned before, the subgroup $\IsomGM$ \emph{does} in general depend on the specific choice of $G$-structure $\GM$, not only the structure group~$G$.

\paragraph{Isometry action on associated vector bundles $\A$:}
From the pushforward of isometries in $\IsomGM$ on $\GM$ one can construct a pushforward $\dphiA$ on any $G$-associated vector bundle $\A = (\GM\times\R^c)/\!\sim_{\!\rho}$ by defining
\begin{align}\label{eq:pushforward_A_def}
    \dphiA\!: \A \to \A,\ \ \
    \pig[[e_i]_{i=1}^d,\, \mathscr{f}\pig]
    \mapsto \dphiA\!\Big(\! \pig[[e_i]_{i=1}^d,\, \mathscr{f}\pig] \!\Big)
    := \Big[\dphiGM \big([e_i]_{i=1}^d \big),\, \mathscr{f}\Big] \,,
    \qquad \phi \in \IsomGM \,.
\end{align}
This action is well defined since the construction is by the right $G$-equivariance of $\dphiGM$ in Eq.~\eqref{cd:isom_induced_GM_automorphism} independent from the chosen representative of the equivalence class.
Similar to before, one has $\piA\mkern2mu\circ\mkern2mu \dphiA = \phi\mkern2mu \circ\mkern2mu \piA$, that is, $\dphiA$ maps feature vectors at $\A_p$ to feature vectors at $\A_{\phi(p)}$, which can be checked by acting on a feature vector and using the corresponding property of $\dphiGM$.
Since $\dphiA$ is defined by the action of $\dphiGM$ on the first factor in $(\GM\times\R^c)/\!\sim_{\!\rho}$\,, it does not interfere with linear combinations which act on the second factor as defined in Eq.~\eqref{eq:associated_bdl_linear_combination}.
This implies that the pushforward on associated bundles maps linearly between their fibers.
The invertibility of $\dphiA$ follows from the invertibility of $\dphiGM$ such that one again gets $(\dphiA)^{-1} = (\phi^{-1})_{\mkern-2mu*\mkern-1mu\scalebox{.55}{$,\mkern-2mu \A$}}$, which we write as $\dphiAinv$.
These properties, together with the fact that $\dphiGM \in \Aut(\GM)$ is in particular a principal bundle automorphism, identify $\dphiA$ as an \emph{associated vector bundle automorphism}, satisfying the following commutative diagram:
\begin{equation}\label{cd:associated_bdl_automorphism}
\begin{tikzcd}[column sep=70pt, row sep=35, font=\normalsize]
    \A
        \arrow[r, "\dphiA"]
        \arrow[d, "\piA"']
    &
    \A
        \arrow[d, "\piA"]
    \\
    M
        \arrow[r, "\phi"']
    &
    M
\end{tikzcd}
\end{equation}

The associated bundle resulting from the specific choices of $\rho(g)=g$ as group representation and $\R^d$ as typical fiber is via the bundle morphism $\chi:(\GM\times\R^d)/\!\!\sim\; \to \TM$ from Eq.~\eqref{eq:A_TM_isomorphism} isomorphic to the tangent bundle $\TM$ (as $G$-bundle).
Our definition of pushforwards on associated $G$-bundles is consistent with this identification since $\chi\circ\dphiA = \dphiTM\circ\chi$.
To see this, let $\big[[e_i]_{i=1}^d,\, \mathscr{v}\big] \in (\GM\times\R^d)/\!\!\sim\;$ be an element of the isomorphic associated bundle that is mapped to $\chi\big(\big[ [e_i]_{i=1}^d,\, \mathscr{v} \big]\big) = \sum_i e_i \mathscr{v}_i$.
Then we have
${\chi \circ \dphiA \big( \big[[e_i]_{i=1}^d,\, \mathscr{v}\big] \big)}
 = \chi \big( \big[[\dphiTM(e_i)]_{i=1}^d,\, \mathscr{v}\big] \big)
 = \sum_i \dphiTM(e_i) \mathscr{v}_i
 = \dphiTM \big(\sum_i e_i \mathscr{v}_i \big)
 = \dphiTM \circ \chi\big(\big[ [e_i]_{i=1}^d,\, \mathscr{v} \big]\big) \,,
$
which shows the consistency of the definitions.

As an associated bundle, the pushforward $\dphiHom$ on the homomorphism bundle $\Hom(\Ain,\Aout) \cong {(\GM\times\R^{\cout\times\cin})/\!\sim_{\!\rhoHom}}$ is specified by Eq.~\eqref{eq:pushforward_A_def} as well.
However, we will later on require an expression of $\dphiHom$ in terms of the pushforwards $\dphiAin$ and $\dphiAout$ of $\Ain$ and $\Aout$, respectively, which we will shortly derive here.
For that purpose, let $H\in \Hom(\Ain|_p,\Aout|_p)$ be a homomorphism at $p$ and $f_p\in \Ainp$ be a feature vector at~$p$.
Then $H(f_p)$ is by definition a feature vector in $\Aoutp$.
In order to be consistently defined, the pushforward of the input feature vector $f_p$, being acted on by the pushforward of the homomorphism $H$, needs to agree with the pushforward of the output feature vector $H(f_p)$.
This implies
\begin{align}
    \dphiAout \big[ H(f_p) \big]
    \ =\  \big[ \dphiAout H\, \dphiAin^{-1} \big] \big( \dphiAin f_p \big)
    \ =:\ \big[ \dphiHom H \big] \big( \dphiAin f_p \big) \,,
\end{align}
where we defined the pushforward on the homomorphism bundle as:
\begin{align}\label{eq:pushforward_Hom_def}
    \dphiHom\!: \Hom(\Ain,\Aout) \to \Hom(\Ain,\Aout), \quad
    H \mapsto
    \dphiAout H\, \dphiAininv \,,
    \qquad\ \phi \in \IsomGM
\end{align}
Note that the composition of an element $H\in \Hom(\Ain,\Aout)$ with $\dphiAout$ on the left and with $\dphiAininv$ on the right mirrors the style of Eq.~\eqref{eq:Hom_bdl_triv_ptwise}.

\paragraph{Isometry action on feature fields:}
The actions of isometries in $\IsomGM$ on the associated bundles give rise to actions on their sections, in particular on feature fields.
This pushforward of sections is defined as follows:
\begin{dfn}[Isometry pushforward of feature field:]
\label{dfn:isometry_pushforward}
    Let $f\in \Gamma(\A)$ be a feature field and let $\phi\in \IsomGM$ be a $G$-structure preserving isometry.
    The isometry acts on the feature field via the \emph{pushforward}%
    \footnote{
        Note the similarity of this definition to that of the \emph{induced representation}, which is the group action w.r.t. which steerable CNNs are designed to be equivariant~\cite{Cohen2017-STEER,3d_steerableCNNs,Weiler2019_E2CNN}.
    }
    \begin{align}\label{eq:pushforward_section_A}
        \rhd\!:\ \IsomGM \mkern-2mu\times\mkern2mu \Gamma(\A) \to \Gamma(\A), \quad
        (\phi,f) \mapsto \phi\rhd\!f \,:=\, \dphiA\! \circ f \circ \phiinv \,.
    \end{align}
    In terms of a commutative diagram, this definition is visualized as:
    \begin{equation}\label{cd:pushforward_section_A}
    \qquad\qquad
    \begin{tikzcd}[column sep=70pt, row sep=35, font=\normalsize]
        \A
            \arrow[r, "\dphiA"]
        &
        \A
        \\
        M
            \arrow[r, "\phi"']
            \arrow[u, "f\ "]
        &
        M
            \arrow[u, "\ \phi\rhd\!f\ :=\: \dphiA \!\circ\mkern-2mu f \circ \phiinv"']
    \end{tikzcd}
    \end{equation}
\end{dfn}
Intuitively, this definition states that the pushforward section $\phi\rhd\!f$, evaluated at $p\in M$, returns the feature vector of $f$ from $\phi^{-1}(p)$, pushed forward to $p$ via $\dphiA$.
Note that such pushforwards do indeed yield well defined sections, which satisfy
\begin{align}
    \piA \circ (\phi\rhd \!f)
    \ &=\ \piA \circ \dphiA\! \circ f \circ \phiinv \notag \\
    \ &=\ \phi \circ \piA \circ f \circ \phiinv \notag \\
    \ &=\ \phi \circ \id_M \circ \phiinv \notag \\
    \ &=\ \id_M
\end{align}
as required by Eq.~\eqref{cd:section_proj_idM}.
Fig.~\ref{fig:isom_egg_actions} visualizes the action of isometries on fields.
The action of isometries on the transporter pullback $\Expspf$ of a fields~$f$ is derived in Section~\ref{sec:isom_expmap_transport} below.

\subsubsection{Isometry action in local coordinates}
\label{sec:isom_coordinatization}

Most of the derivations on the isometry equivariance of kernel field transforms in Sections~\ref{sec:isometry_equivariance} and~\ref{sec:quotient_kernel_fields} will be kept in a coordinate free setting.
However, since $\GM$-convolutions are defined relative to a choice of $G$-atlases of the associated bundles, the investigation of their isometry equivariance will require us to study coordinate expressions of the isometry pushforwards $\dphiTM$, $\dphiFM$, $\dphiGM$ and $\dphiA$ relative to local bundle trivializations.
Coordinate expressions of the isometry action are furthermore useful in numerical implementations, which are necessarily encoding feature fields relative to fields of reference frames.

In the following, we assume gauges $\PsiTM^{\widetilde{A}}$ and $\PsiTM^A$ on neighborhoods $U^{\widetilde{A}}$ of $p$ and $U^A$ of $\phi(p)$ to be given.
For convenience, let $U^A = \phi\big(U^{\widetilde{A}}\big)$ coincide with the image of $U^{\widetilde{A}}$ under the isometry, which is always possible without losing generality.

\paragraph{Pushforward on \textit{TM} in coordinates:}
Recall that the pushforward on the tangent bundle is a linear map from vectors $v\in \TpM$ to vectors $\dphiTM v \in \TphipM$.
Relative to the given gauges, the pushforward is therefore coordinatized by a field of matrices%
\footnote{
    Given charts
    $x^{\widetilde{A}}: U^{\widetilde{A}} \to x^{\widetilde{A}}\big(U^{\widetilde{A}}\big) \subseteq \R^d$
    and
    $x^A: U^A \to x^A\big(U^A\big) \subseteq \R^d$
    of $M$, an isometry $\phi$ can be locally represented by a map
    $x^A \circ \phi \circ \big(x^{\widetilde{A}}\big)^{\mkern-1mu-1}\!: x^{\widetilde{A}}\big(U^{\widetilde{A}}\big) \to x^A\big(U^A\big)$
    between coordinates.
    For the special case that the gauges at $p$ and $\phi(p)$ correspond to the coordinate bases of those charts, $g_\phi^{A\widetilde{A}}$ is simply given by the \emph{Jacobian} of $x^A \circ \phi \circ \big(x^{\widetilde{A}}\big)^{\mkern-1mu-1}$.
}
\begin{align}\label{eq:pushforward_TM_coord}
    g_\phi^{A\widetilde{A}}\!: U^{\widetilde{A}} \to \GL{d},\ \ \ 
    p \mapsto g_\phi^{A\widetilde{A}}(p) := \psiTMphip^A \mkern-2mu\circ \dphiTM \mkern-2mu\circ \big(\psiTMp^{\widetilde{A}}\big)^{-1}
    , \quad\ \ \phi \in \IsomM \,,
\end{align}
which transforms between the corresponding numerical coefficients $\psiTMp^{\widetilde{A}}(v)$ of $v$ at $p$ and $\psiTMphip^A(\dphiTM v) = g_\phi^{A\widetilde{A}}(p) \psiTMp^{\widetilde{A}}(v)$ of $\dphiTM v$ at $\phi(p)$.
More precisely, $g_\phi^{A\widetilde{A}}$ takes values in the subgroup $\langle\, G\cup\O{d} \,\rangle$ of $\GL{d}$, which is generated by the elements of $\O{d}$ (due to $\dphiTM$ preserving the metric) and $G$ (since the transition functions might form a supergroup of $\O{d}$).
The definition of the pushforward in local coordinates is visualized by the following commutative diagram:
\begin{equation}\label{cd:pushforward_TM_coord}
    \begin{tikzcd}[row sep=4.em, column sep=5em]
        \R^d
            \arrow[rrr, pos=.5, rounded corners, to path={ 
                    -- ([yshift=-3.5ex]\tikztostart.south) 
                    --node[below]{\small$
                        g_\phi^{A\widetilde{A}}(p) \mkern2mu\cdot
                        $} ([yshift=-3.5ex]\tikztotarget.south) 
                    -- (\tikztotarget.south)
                    }]
        &
        \TpM
            \arrow[l, "\psiTMp^{\widetilde{A}}"']
            \arrow[r, "\dphiTM"]
        &
        \TphipM
            \arrow[r, "\psiTMphip^A"]
        &
        \R^d
    \end{tikzcd}
\end{equation}
Fig.~\ref{fig:pushforward_vector_components} gives a graphical interpretation of the pushforward in coordinates.

\paragraph{Pushforward on \textit{FM} in coordinates:}
The coordinatization of the pushforward on the frame bundle is defined in analogy to Eq.~\eqref{eq:pushforward_TM_coord}.
It turns out to be given by the left action of the same group element $g_\phi^{A\widetilde{A}}$ on trivialized frames as shown in the commutative diagram below:
\begin{equation}\label{eq:pushforward_FM_coord}
    \begin{tikzcd}[row sep=4.em, column sep=5em]
        \GL{d}
            \arrow[rrr, pos=.5, rounded corners, to path={ 
                    -- ([yshift=-3.5ex]\tikztostart.south) 
                    --node[below]{\small$
                        g_\phi^{A\widetilde{A}}(p) \mkern2mu\cdot
                        $} ([yshift=-3.5ex]\tikztotarget.south) 
                    -- (\tikztotarget.south)
                    }]
        &
        \FpM
            \arrow[l, "\psiFMp^{\widetilde{A}}"']
            \arrow[r, "\dphiFM"]
        &
        \FphipM
            \arrow[r, "\psiFMphip^A"]
        &
        \GL{d}
    \end{tikzcd}
\end{equation}
To prove this claim, we compute the action on a trivialized frame, given by a matrix $h \in \GL{d}$ whose $i$-th column $h_{:,i}$ represents the $i$-th frame vector:
\begin{alignat}{3}
    \qquad\qquad
       & \Big[ \psiFMphip^A \circ \dphiFM \circ \big(\psiFMp^{\widetilde{A}}\big)^{-1} \Big](h) \notag \\
    =\ & \Big[ \psiFMphip^A \circ \dphiFM \Big] \!\Big( \!\Big( \big(\psiTMp^{\widetilde{A}}\big)^{-1} (h_{:,i})\Big)_{i=1}^d \Big)
        \qquad\quad && \big( \text{\small def. of $\psiFMp^{\widetilde{A}}$, Eq.~\eqref{eq:trivialization_FM_p} } \big) \notag \\
    =\ & \psiFMphip^A \!\Big( \!\Big( \dphiTM \circ \big(\psiTMp^{\widetilde{A}}\big)^{-1} (h_{:,i})\Big)_{i=1}^d \Big)
        \qquad\quad && \big( \text{\small def. of $\dphiFM$, Eq.~\eqref{eq:pushforward_FM_def} } \big) \notag \\
    =\ & \Big( \!\Big( \psiTMphip^A \circ \dphiTM \circ \big(\psiTMp^{\widetilde{A}}\big)^{-1} (h_{:,i})\Big)_{i=1}^d \Big)
        \qquad\quad && \big( \text{\small def. of $\psiFMphip^A$, Eq.~\eqref{eq:trivialization_FM_p} } \big) \notag \\
    =\ & \Big( g_\phi^{A\widetilde{A}}(p) \cdot h_{:,i} \Big)_{i=1}^d
        \qquad\quad && \big( \text{\small def. of $g_\phi^{A\widetilde{A}}$, Eq.~\eqref{eq:pushforward_TM_coord} } \big) \notag \\
    =\ & g_\phi^{A\widetilde{A}}(p) \cdot h
\end{alignat}

The action of the pushforward on local trivializations can be thought of as \emph{inducing a gauge transformation}.
A graphical intuition for this statement was given in Fig.~\ref{fig:pushforward_vector_components} where the initial gauges at $p$ and $\phi(p)$ are visualized by choices of reference frames.
A pushforward of the frame at $p$ to $\phi(p)$ (red) does in general not agree with the original frame at $\phi(p)$ (green).
The transition between these two frames is the induced gauge transformation by at $\phi(p)$.
We will in the following construct this transformation; first in terms of local trivializations, then in terms of the corresponding frame fields.

From the commutative diagram in Eq.~\eqref{eq:pushforward_FM_coord} it is clear that the gauge $\psiFMp^{\widetilde{A}}: \FpM \to \GL{d}$ at $p$ can via $\dphiFM^{-1}$ be pulled back to a gauge at $\phi(p)$, which is given by
\begin{align}\label{eq:isom_pullback_section_FM}
    \psiFMp^{\widetilde{A}} \circ \dphiFM^{-1}
    \ =\ \big( g_\phi^{A\widetilde{A}}(p) \big)^{-1} \psiFMphip^A
    \,:\,\ \FphipM \to \GL{d} \,.
\end{align}
The corresponding extension of the commutative diagram in Eq.~\eqref{eq:pushforward_FM_coord} visualizes the equivalence of both expressions and makes an algebraic proof superfluous:
\begin{equation}\label{cd:pushforward_FM_coord_extended}
    \begin{tikzcd}[row sep=4.em, column sep=5em]
        \GL{d}
            \arrow[rrr, pos=.5, rounded corners, to path={ 
                    -- ([yshift=-3.5ex]\tikztostart.south) 
                    --node[below]{\small$
                        g_\phi^{A\widetilde{A}}(p) \mkern2mu\cdot
                        $} ([yshift=-3.5ex]\tikztotarget.south) 
                    -- (\tikztotarget.south)
                    }]
        &
        \FpM
            \arrow[l, "\psiFMp^{\widetilde{A}}"']
            \arrow[r, "\dphiFM"]
        &
        \FphipM
            \arrow[r, "\psiFMphip^A"]
            \arrow[ll, pos=.5, rounded corners, to path={ 
                    -- ([yshift=5.ex]\tikztostart.north) 
                    --node[above]{\small$
                        \psiFMp^{\widetilde{A}}\, \dphiFM^{-1}\ =\ 
                        \big( g_\phi^{A\widetilde{A}}(p) \big)^{-1} \mkern-2mu\cdot \psiFMphip^A
                        $} ([yshift=5.ex]\tikztotarget.north) 
                    -- (\tikztotarget.north)
                    }]
        &
        \GL{d}
    \end{tikzcd}
\end{equation}
The transition map (gauge transformation) between the isometry induced gauge $\psiFMp^{\widetilde{A}} \, \dphiFM^{-1}$ and the original gauge $\psiFMphip^A$ at $\phi(p)$ is read off to be given by the inverse%
\footnote{
    The inverse is a matter of convention.
    It arises here since we defined $g_\phi^{A\widetilde{A}}$ as coordinate expression of the covariant pushforward of frames while gauges transform contravariantly.
}
group element
\begin{align}\label{eq:gauge_trafo_pushforward_gauge}
    \big( \psiFMp^{\widetilde{A}} \, \dphiFM^{-1} \big) \circ \big( \psiFMphip^A \big)^{-1}
    \ =\ \big( g_\phi^{A\widetilde{A}}(p) \big)^{-1}
    \ \ \ \in\,\ \langle\, G\cup\O{d} \,\rangle\ \leq\ \GL{d} \,.
\end{align}
Note that this group element does for $G\leq\O{d}$ not necessarily lie in the structure group, that is, the isometry induced gauge might not be $G$-compatible (it can not be added to an existing $G$-atlas of $\FM$).
In the next paragraph on $G$-structures we will show that this happens exactly when $\phi \notin \IsomGM$, i.e. for isometries which do not respect the $G$-structure.

To derive the isometry action on frame fields, consider the identity sections $\sigma^{\widetilde{A}}: U^{\widetilde{A}} \to \piFM^{-1}\big(U^{\widetilde{A}}\big)$ over $U^{\widetilde{A}}$ and $\sigma^A: U^A \to \piFM^{-1}\big(U^A\big)$ over $U^A$.
These sections model the original frame fields from Fig.~\ref{fig:pushforward_vector_components}.
The new frame field is then given by the pushforward section
\begin{align}\label{eq:pushforward_section_FM}
    \phi \mkern2mu\rhd \sigma^{\widetilde{A}} \,:=\, \dphiFM \circ \sigma^{\widetilde{A}} \circ \phiinv
    \ :\ U^A \to \piFM^{-1}\big( U^A\big) \,,
\end{align}
which is equivalently defined to that in Eq.~\ref{eq:pushforward_section_A}.
An alternative expression for the pushforward frame field in terms of the right action of $g_\phi^{A\widetilde{A}}$ is found by applying $\psiFMphip^A$:
\begin{alignat}{3}
    \qquad\qquad\qquad
     &\ \ \psiFMphip^A \pig(\big[ \phi \rhd \sigma^{\widetilde{A}} \,\big] (\phi(p)) \pig) \notag \\
    =&\ \ \psiFMphip^A \, \dphiFM \, \sigma^{\widetilde{A}}(p)
        \qquad\quad && \big( \text{\small def. of $\phi \rhd \sigma^{\widetilde{A}}$, Eq.~\eqref{eq:pushforward_section_FM} } \big) \notag \\
    =&\ \ g_\phi^{A\widetilde{A}}(p) \, \psiFMp^{\widetilde{A}} \, \sigma^{\widetilde{A}}(p) 
        \qquad\quad && \big( \text{\small equivalent expressions in Eq.~\eqref{eq:isom_pullback_section_FM} } \big) \notag \\
    =&\ \ g_\phi^{A\widetilde{A}}(p)
        \qquad\quad && \big( \text{\small identity section $\sigma^{\widetilde{A}}$, Eq.~\eqref{eq:identity_section_prop} } \big) \notag \\
    =&\ \ \psiFMphip^A \pig( \sigma^A \big(\phi(p)\big) \pig) \, g_\phi^{A\widetilde{A}}(p)
        \qquad\quad && \big( \text{\small identity section $\sigma^A$, Eq.~\eqref{eq:identity_section_prop} } \big) \notag \\
    =&\ \ \psiFMphip^A \pig( \sigma^A \big(\phi(p)\big) \lhd g_\phi^{A\widetilde{A}}(p) \pig)
        \qquad\quad && \big( \text{\small right $\GL{d}$ equivariance, Eq.~\eqref{eq:right_equivariance_FM} } \big)
\end{alignat}
Since $\psiFMphip^A$ is an isomorphism, it follows that
\begin{align}\label{eq:pushfwd_section_right_action}
    \pig(\phi \rhd \sigma^{\widetilde{A}} \pig) \big( \phi(p) \big)
    \ =\ \dphiFM \sigma^{\widetilde{A}} (p)
    \ =\ \sigma^A \big(\phi(p)\big) \lhd g_\phi^{A\widetilde{A}}(p) \,,
\end{align}
that is, $g_\phi^{A\widetilde{A}}(p)$ does as expected describe the transformation between identity sections.
This isometry induced transformation between reference frames is in Fig.~\ref{fig:pushforward_vector_components} visualized by the blue arrow between the (translucent) red and green frame.

The isometry transformed gauge $\psiFMp^{\widetilde{A}}\, \dphiFM^{-1}$ and the pushforward section $\phi \rhd \sigma^{\widetilde{A}}$ correspond to each other in so far that the latter is the identity section of the former:
\begin{align}
    \psiFMp^{\widetilde{A}}\, \dphiFM^{-1} \pig[ \phi \rhd \sigma^{\widetilde{A}} \pig] \big(\phi(p)\big)
    \ =\ \psiFMp^{\widetilde{A}}\, \dphiFM^{-1} \dphiFM \sigma^{\widetilde{A}}(p)
    \ =\ \psiFMp^{\widetilde{A}} \sigma^{\widetilde{A}}(p)
    \ =\ e
\end{align}

\paragraph{Pushforward on \textit{GM} in coordinates:}
As argued in the previous Section~\ref{sec:isom_action_bundles}, the pushforward on $\GM$ is only well defined for isometries $\phi$ in a subgroup $\IsomGM$.
Not surprisingly, the corresponding isometry induced gauge transformations take values in the structure group~$G$:
\begin{thm}[$\IsomGM$ in local trivializations]
\label{thm:isom_GM_in_coords}
    Let $\phi \in \IsomM$ be any isometry of $M$.
    Then the following three statements are equivalent:
    \begin{enumerate}
        \item $\phi$ is $G$-structure preserving, that is, $\phi \in \IsomGM$.
        \item The isometry pullback $\psiFMp^{\widetilde{A}}\, \dphiFM^{-1}$ of any gauge $\psiFMp^{\widetilde{A}}$ of the $G$-atlas of $\FM$ that defines~$\GM$ is $G$-compatible with that $G$-atlas.
        \item
        The coordinate expression of $\dphiFM$ relative to any gauges $\psiFMp^{\widetilde{A}}$ and $\psiFMphip^A$ from the $G$-atlas of $\FM$ takes values in the structure group, that is, $g_\phi^{A\widetilde{A}}(p) \in G\ \ \ \forall\ p \in M$.
    \end{enumerate}
\end{thm}
\begin{proof}
    The defining property of a $G$-structure preserving isometry $\phi \in \IsomGM$ is that it satisfies $\dphiFM(\GpM) = \GphipM$ for any $p \in M$; see Eq.~\eqref{eq:isomGM_def}.
    In terms of a given $G$-atlas of $\FM$, Eq.~\eqref{eq:G_atlas_induced_G_structure_GM_def_ptwise} defined the $G$-structure at $p\in M$ as $\GpM := \big(\psiFMp^{\widetilde{A}}\big)^{-1} (G)$ where $\psiFMp^{\widetilde{A}}$ is an \emph{arbitrary} gauge of the $G$-atlas.
    With this expression we expand the left-hand side of the defining property of $\IsomGM$:
    \begin{align}
        \dphiFM(\GpM)
        \ &=\ \dphiFM\, \big(\psiFMp^{\widetilde{A}}\big)^{-1} (G) \notag \\
        \ &=\ \Big(\psiFMp^{\widetilde{A}} \dphiFM^{-1} \Big)^{-1} (G)
    \intertext{
    Relative to any gauge $\psiFMphip^A$ of the $G$-atlas at $\phi(p)$, this can be further manipulated to
    }
        \dphiFM(\GpM)
        \ &=\ \Big(\big( g_\phi^{A\widetilde{A}}(p) \big)^{-1} \psiFMphip^A  \Big)^{-1} (G) \notag \\
        \ &=\ \big( \psiFMphip^A  \big)^{-1} \big(g_\phi^{A\widetilde{A}}(p)\, G\big) \,.
    \end{align}
    The right-hand side of the defining property of $\IsomGM$ is in terms of $\psiFMphip^A$ given by
    \begin{align}
        \GphipM = \big( \psiFMphip^A \big)^{-1} (G).
    \end{align}
    Setting both sides equal and using that $\psiFMphip^A$ is an isomorphism implies $g_\phi^{A\widetilde{A}}(p)\, G = G$ which leads to the claimed equivalence
    \begin{align}
        \dphiFM(\GpM) = \GphipM
        \quad \Longleftrightarrow \quad
        g_\phi^{A\widetilde{A}}(p) \in G
    \end{align}
    of statements \textit{1}. and \textit{3}.
    To prove the equivalence to statement~\textit{2.}, recall that $g_\phi^{A\widetilde{A}}(p)$ is by Eq.~\eqref{eq:gauge_trafo_pushforward_gauge} equal to the gauge transformation from $\psiFMp^{\widetilde{A}} \, \dphiFM^{-1}$ to $\psiFMphip^A$.
    As $G$-atlases have by definition transition functions in the structure group $G$, the implications (\textit{2.$\leftrightarrow$3.}) follow, such that all three statements are seen to be equivalent.
\end{proof}
These results are of central importance for our later study of the isometry equivariance of $\GM$-convolutions.
We will be able to show that such convolutions are equivariant under the action of $\phi \in \IsomGM$ on feature fields, which relies on the fact that the $G$-steerability of the convolution kernels accounts for the isometry induced gauge transformations $g_\phi^{A\widetilde{A}}(p) \in G$.

For $G$-structure automorphism inducing isometries $\phi \in \IsomGM$, we can adapt the commutative diagram for~$\FM$ in Eq.~\eqref{cd:pushforward_FM_coord_extended} to its cousin for~$\GM$:
\begin{equation}\label{cd:pushforward_GM_coord_extended}
    \begin{tikzcd}[row sep=4.em, column sep=5em]
        G
            \arrow[rrr, pos=.5, rounded corners, to path={ 
                    -- ([yshift=-3.5ex]\tikztostart.south) 
                    --node[below]{\small$
                        g_\phi^{A\widetilde{A}}(p) \mkern2mu\cdot
                        $} ([yshift=-3.5ex]\tikztotarget.south) 
                    -- (\tikztotarget.south)
                    }]
        &
        \GpM
            \arrow[l, "\psiGMp^{\widetilde{A}}"']
            \arrow[r, "\dphiGM"]
        &
        \GphipM
            \arrow[r, "\psiGMphip^A"]
            \arrow[ll, pos=.5, rounded corners, to path={ 
                    -- ([yshift=5.ex]\tikztostart.north) 
                    --node[above]{\small$
                        \psiGMp^{\widetilde{A}}\, \dphiGM^{-1}\ =\ 
                        \big( g_\phi^{A\widetilde{A}}(p) \big)^{-1} \mkern-2mu\cdot \psiGMphip^A
                        $} ([yshift=5.ex]\tikztotarget.north) 
                    -- (\tikztotarget.north)
                    }]
        &
        G
    \end{tikzcd}
\end{equation}

\paragraph{Pushforward on $\A$ in coordinates:}

The pushforward of $\phi \in \IsomGM$ on associated $G$-bundles is similarly coordinatized as those of the other bundles.
In terms of a commutative diagram we get, not surprisingly,
\begin{equation}\label{cd:pushforward_A_coord}
    \quad
    \begin{tikzcd}[row sep=4.em, column sep=5em]
        \R^c
            \arrow[rrr, pos=.5, rounded corners, to path={ 
                    -- ([yshift=-3.5ex]\tikztostart.south) 
                    --node[below]{\small$
                        \rho\big( g_\phi^{A\widetilde{A}}(p) \big)
                        $} ([yshift=-3.5ex]\tikztotarget.south) 
                    -- (\tikztotarget.south)
                    }]
        &
        \A_p
            \arrow[l, "\psiAp^{\widetilde{A}}"']
            \arrow[r, "\dphiA"]
        &
        \A_{\phi(p)}
            \arrow[r, "\psiAphip^A"]
        &
        \R^c
    \end{tikzcd}
    \quad,
\end{equation}
which follows when acting on feature vector coefficients $\mathscr{f}\in\R^c$:
\begin{alignat}{3}
    \qquad\qquad\qquad\qquad\quad
       & \Big[ \psiAphip^A \circ \dphiA \circ \big(\psiAp^{\widetilde{A}}\big)^{-1} \Big] (\mathscr{f}) \notag \\
    =\ & \Big[ \psiAphip^A \circ \dphiA \Big] \!\Big( \Big[ \sigma^{\widetilde{A}}(p), \,\mathscr{f}\, \Big] \Big)
        \qquad\quad && \big( \text{\small def. of $\big(\psiAp^A\big)^{-1}$, Eq.~\eqref{eq:trivialization_A_p_inv} } \big) \notag \\
    =\ & \psiAphip^A \!\Big( \Big[ \dphiFM\big( \sigma^{\widetilde{A}}(p) \big), \,\mathscr{f}\, \Big] \Big)
        \qquad\quad && \big( \text{\small def. of $\dphiA$, Eq.~\eqref{eq:pushforward_A_def} } \big) \notag \\
    =\ & \psiAphip^A \!\Big( \Big[ \sigma^A\big(\phi(p)\big) \lhd g_\phi^{A\widetilde{A}}(p), \,\mathscr{f}\, \Big] \Big)
        \qquad\quad && \big( \text{\small induced gauge transformation, Eq.~\eqref{eq:pushfwd_section_right_action} } \big) \notag \\
    =\ & \psiAphip^A \!\Big( \Big[ \sigma^A\big(\phi(p)\big), \, \rho\big( g_\phi^{A\widetilde{A}}(p) \big) \mathscr{f}\, \Big] \Big)
        \qquad\quad && \big( \text{\small def. of $\sim_{\!\rho}$, Eq.~\eqref{eq:equiv_relation_A} } \big) \notag \\
    =\ & \rho\Big( g_\phi^{A\widetilde{A}}(p) \Big) \cdot \mathscr{f}
        \qquad\quad && \big( \text{\small def. of $\psiAp$, Eq.~\eqref{eq:trivialization_A_p} } \big)
\end{alignat}
Note that the expression $\rho\big( g_\phi^{A\widetilde{A}}(p) \big)$ requires $g_\phi^{A\widetilde{A}}(p)$ to be a structure group element since $\rho$ is a $G$-representation.
This shows once again from another perspective that pushforwards on $\A$ can only be defined for isometries in $\IsomGM$.

For completeness, we give the following local trivialization of the commutative diagram from Eq.~\eqref{cd:pushforward_section_A}, which might be useful when implementing coordinate independent CNNs and testing their $\IsomGM$-equivariance:
\begin{equation}\label{cd:pushforward_A_coord}
\begin{tikzcd}[column sep=60pt, row sep=45, font=\normalsize,
                   execute at end picture={
                        \node [] at (-1.87, -.07) {$\noncommutative$};
                        \node [] at ( 1.87, -.07) {$\noncommutative$};
                        }]
    U^{\widetilde{A}} \times \R^c
        \arrow[rd, "\proj_1"']
        \arrow[rrr, rounded corners, to path={ 
                -- ([yshift=2em]\tikztostart.north) 
                --node[above]{\small$\phi \times \rho\big(g^{A\widetilde{A}}_\phi\big) \cdot$} ([yshift=2em]\tikztotarget.north) 
                -- (\tikztotarget.north)
                }]
    &
    \piA^{-1}\big(U^{\widetilde{A}}\big)
        \arrow[r, "\dphiA"]
        \arrow[d, "\piA"']
        \arrow[l, "\PsiA^{\widetilde{A}}"']
    &[2ex]
    \piA^{-1}\big(U^A\big)
        \arrow[d, "\piA"]
        \arrow[r, "\PsiA^A"]
    &
    U^A \times \R^c
        \arrow[ld, "\proj_1"]
    \\
    &
    U^{\widetilde{A}}
        \arrow[r, "\phi"']
        \arrow[u, "f"', bend right=28]
    &
    U^A
        \arrow[u, "\ \phi\rhd\!f", bend left=28]
\end{tikzcd}
\end{equation}

\subsubsection{Commutativity of isometry actions with the exponential map and transporters}
\label{sec:isom_expmap_transport}

In the following section we will need an expression for the action of isometries on transporter pullbacks $\Expspf$ of feature fields $f$, which we derive here.
For this purpose, we discuss the behavior of the exponential map and parallel transporters under the action of isometries.

\begin{figure}
    \centering
    \includegraphics[width=.9\columnwidth]{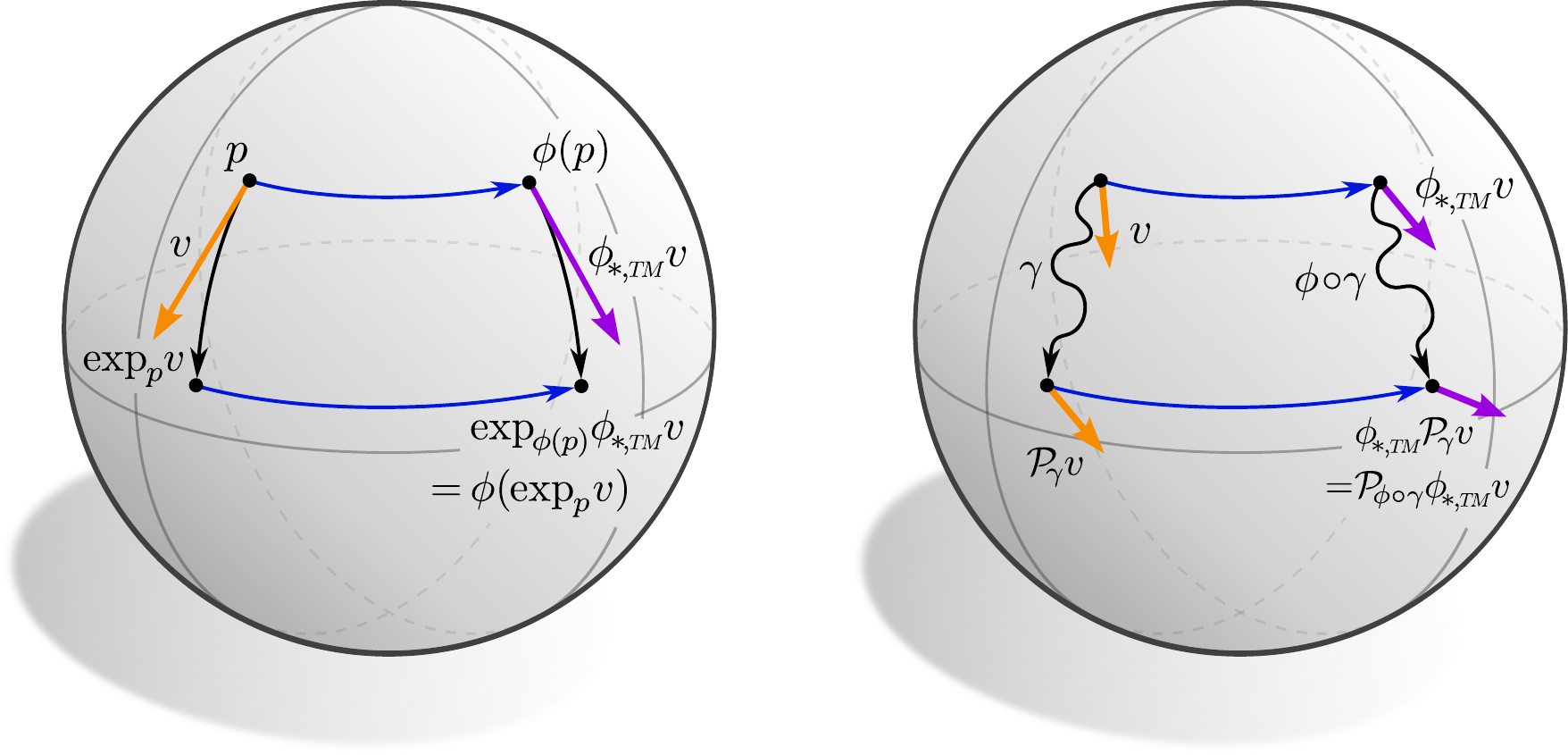}
    \vspace*{1ex}
    \caption{\small
        \ \emph{Left:}
        Isometries commute with the exponential map, that is,
        $\exp_{\phi(p)} \circ\, \protect\dphiTM (v) \,=\, \phi \circ \exp_p(v)$
        for any vector $v\in \TpM$ and isometry $\phi\in \IsomM$.
        \ \emph{Right:}
        Isometries also commute with the Levi-Civita transport of tangent vectors and feature vectors, that is,
        $\protect\dphiA\! \circ \protect\PAgamma \ =\ 
        \mathcal{P}_{\mkern-4mu\overset{}{\protect\scalebox{.6}{$\!\A$}, \mkern1mu\phi\circ\gamma}}\!
        \circ \protect\dphiA$
        for arbitrary paths $\gamma:[0,1]\to M$ and isometries $\phi\in \IsomM$.
        If an alternative, $G$-compatible connection is used, we demand that the same commutativity property holds for them.
        The isometry-invariance of exponential maps and transporters allows $\GM$-convolutions to be equivariant under the action of isometries.
    }
    \label{fig:isom_exp_transport}
\end{figure}

\paragraph{Isometries and the exponential map:}
As proven in \cite{gallier2019diffgeom1}, isometries map geodesics to geodesics and do therefore in particular commute with the exponential map.%
\footnote{\label{footnote:LeviCivita_isometry_invariance}
    The proof relies on the fact that the Levi-Civita connection $\nabla: \Gamma(\TM) \times \Gamma(\TM) \to \Gamma(\TM),\ (X,Y)\mapsto \nabla_{\!X}Y$, on which the Riemannian exponential map is based, commutes with isometries:
    $\phi\rhd\! \big( \nabla_{\!X}Y \big) \ =\ \nabla_{\mkern-2mu\phi\rhd\! X} \big(\phi\rhd\! Y\big)$;
    see~\cite{gallier2019diffgeom1}
}
More specifically, the identity
\begin{align}\label{eq:exp_isom_commutation}
    \exp_{\phi(p)} \circ\, \dphiTM (v) \,=\, \phi \circ \exp_p(v) \quad\ \forall\ v\in \TpM,\ \ \phi\in \IsomM \,,
\end{align}
holds for any isometry and any tangent vector at $p$ (still assuming a geodesically complete manifold).
It states that the result of the exponential map at $p$, evaluated with some vector $v$ and then being mapped through the isometry, equals the exponential map at $\phi(p)$ when being evaluated with the pushforward of $v$ as visualized in Fig.~\ref{fig:isom_exp_transport} (left).
This statement is diagrammatically expressed by the commutativity of (the upper square of) the following diagram:
\begin{equation}
\begin{tikzcd}[column sep=70pt, row sep=35, font=\normalsize]
    M
        \arrow[r, "\phi"]
    &
    M
    \\
    \TM
        \arrow[r, "\dphiTM"]
        \arrow[d, "\piTM"']
        \arrow[u, "\exp"]
    &
    \TM
        \arrow[d, "\piTM"]
        \arrow[u, "\exp"']
    \\
    M
        \arrow[r, "\phi"']
    &
    M
\end{tikzcd}
\end{equation}

\paragraph{Isometries and parallel transporters:}
The pushforward on the tangent bundle was in~\cite{gallier2019diffgeom1} further argued to commute with the corresponding Levi-Civita transporters, as visualized in Fig.~\ref{fig:isom_exp_transport} (right).
If an alternative, $G$-compatible connection is chosen to transport feature vectors, we \emph{demand} that it commutes with the action of isometries as well.
Since the transporters and the pushforwards on $\FM$, $\GM$ and $\A$ are induced from those on $\TM$, one can easily show that this property translates to them.
Specifically for the associated feature vector bundles this means that for arbitrary isometries $\phi \in \IsomGM$ and paths $\gamma$ we assume the relation
\begin{align}\label{eq:transport_isom_commutation}
    \dphiA\! \circ \PAgamma
    \ =\ 
    \mathcal{P}_{\mkern-4mu\overset{}{\protect\scalebox{.6}{$\!\A$}, \mkern1.5mu\phi \mkern1.5mu\circ\mkern1.5mu \gamma}}\!
    \circ \dphiA
\end{align}
to hold, such that the following diagram commutes:
\begin{equation}
\quad
\begin{tikzcd}[column sep=70pt, row sep=40, font=\normalsize]
    \A_{\gamma(0)}
        \arrow[r, "\dphiA"]
        \arrow[d, "\PAgamma"']
    &
    \A_{\phi \mkern1mu\circ\mkern1mu \gamma(0)}
        \arrow[d, "\mathcal{P}_{\mkern-4mu\overset{}{\protect\scalebox{.6}{$\!\A$}  \mkern-0mu,\phi \mkern1mu\circ\mkern1mu \gamma}}"]
    \\
    \A_{\gamma(1)}
        \arrow[r, "\dphiA"']
    &
    \A_{\phi \mkern1mu\circ\mkern1mu \gamma(1)}
\end{tikzcd}
\end{equation}

\paragraph{Isometries and transporter pullbacks of feature fields:}

Knowing the transformation laws of exponential maps and transporters under the action of isometries, we have everything at hand that is required to derive the transformation law of transporter pullbacks $\Expspf$ of feature fields~$f$:
\begin{thm}[Isometry action on transporter pullbacks of feature fields]
\label{thm:transporter_pullback_isometry_action}
    Let $f\in \Gamma(\A)$ be any feature field and let $\phi \in \IsomGM$ be any $G$-structure preserving isometry.
    Assume the feature vector transporters to commute with the action of $\IsomGM$, that is, that Eq.~\eqref{eq:transport_isom_commutation} holds
    (which is automatically guaranteed for the Levi-Civita connection).
    The transporter pullback (Def.~\ref{dfn:Expf_pullback_field}) of the pushforward field $\phi\rhd \!f$ (Def.~\ref{dfn:isometry_pushforward}) is then given by:
    \begin{align}\label{eq:transporter_pullback_isometry_commutativity}
        \Expsp (\phi \rhd f)
        \ =\ 
        \dphiAout \!\circ \big[\mkern-2mu \Expsphiinvpf \mkern1mu\big] \circ \dphiTM^{-1}
    \end{align}
\end{thm}
\begin{proof}
    We start by letting the right-hand side act on an arbitrary vector $v\in\TpM$ and work progressively to the left-hand side by using the properties derived in this section:
    \begin{alignat}{3}
         &\,\ \dphiAout \big[\mkern-2mu \Expsphiinvpf \mkern1mu\big] \, \dphiTM^{-1} (v) \\
        =&\,\ \dphiAout\, \mathcal{P}_{\mkern-5mu\overset{}{\protect\scalebox{.8}{$\!\Ain$},\protect\scalebox{.85}{$\mkern2mu\phiinv(p) \mkern-3mu\leftarrow\mkern-1mu \exp_{\phiinv(p)} \!\circ\mkern2mu \dphiTM^{-1}(v)$}}}
            \circ f \circ \exp_{\phiinv(p)} \mkern-2mu\circ\, \dphiTM^{-1} (v)
            \quad && \big( \text{\small transporter pullback, Def.~\ref{dfn:Expf_pullback_field} } \big) \notag\\
        =&\,\ \dphiAout\, \mathcal{P}_{\mkern-5mu\overset{}{\protect\scalebox{.8}{$\!\Ain$},\protect\scalebox{.85}{$\mkern2mu\phiinv(p) \mkern-3mu\leftarrow\mkern-1mu \phiinv\mkern-2mu \circ \exp_p (v)$}}}
            \circ f \circ \phiinv \circ \exp_p (v)
            \quad && \big( \text{\small isometry action on $\exp$, Eq.~\eqref{eq:exp_isom_commutation}} \big) \notag\\
        =&\,\ \mathcal{P}_{\mkern-5mu\overset{}{\protect\scalebox{.8}{$\!\Ain$},\protect\scalebox{.85}{$\mkern2mu p \mkern-3mu\leftarrow\mkern-1mu \exp_p (v)$}}}
            \circ \dphiAin \circ f \circ \phiinv \circ \exp_p (v)
            \quad && \big( \text{\small isometry action on $\mathcal{P}_{\mkern-5mu\overset{}{\protect\scalebox{.75}{$\!\Ain$}}}$, Eq.~\eqref{eq:transport_isom_commutation}} \big) \notag\\
        =&\,\ \mathcal{P}_{\mkern-5mu\overset{}{\protect\scalebox{.8}{$\!\Ain$},\protect\scalebox{.85}{$\mkern2mu p \mkern-3mu\leftarrow\mkern-1mu \exp_p (v)$}}}
            \circ (\phi \rhd f) \circ \exp_p (v)
            \quad && \big( \text{\small pushforward of fields, Eq.~\eqref{eq:pushforward_section_A}} \big) \notag\\
        =&\,\ \big[\mkern-2mu \Expsp (\phi \rhd f) \mkern1mu\big] (v)
            \quad && \big( \text{\small transporter pullback, Def.~\ref{dfn:Expf_pullback_field} } \big) \notag
    \end{alignat}
\end{proof}
Intuitively, this result just states that the transporter pullback of a pushforward field equals the pushforward of the original field's transporter pullback.
Relative to local trivializations, this pushforward can be interpreted as an isometry induced gauge transformation, which was stated in Eq.~\eqref{eq:transporter_pullback_pushforward_field}.
We will in the following assume that the $G$-compatible connection which is chosen to transport feature vectors will always be $\IsomGM$-invariant, and thus that Eq.~\eqref{eq:transporter_pullback_isometry_commutativity} holds.

That the transporter pullback and the isometry pushforward commute is a consequence of the commutativity of the exponential map and parallel transporter, in terms of which the transporter pullback is defined.
Note that general diffeomorphisms do not preserve the metric and thus the exponential map and the transporter pullback of feature fields.
Being based on these constructions, kernel field transforms and $\GM$-convolutions can only be isometry equivariant but not fully diffeomorphism equivariant.

%% file: chapters/82_isom_equivariance.tex

\subsection{Isometry equivariance of kernel field transforms and \textit{GM}-convolutions}
\label{sec:isometry_equivariance}

We now turn to investigate under which conditions kernel field transforms and $\GM$-convolutions are equivariant w.r.t. the action of isometries on feature fields.
As the action on the $G$-associated feature vector bundles is only defined for $G$-structure preserving isometries, we will formulate all statements for the subgroup $\IsomGM \leq \IsomM$ or subgroups $\I \leq \IsomGM$ thereof.
One can of course always consider structure groups $G\geq\O{d}$, for which $\IsomGM = \IsomM$.

The equivariance of a kernel field transform, and thus $\GM$-convolutions, is defined as follows:
\begin{dfn}[Isometry equivariant kernel field transform]
\label{dfn:isometry_equivariance}
    Let ${\TK: \Gamma(\Ain) \to \Gamma(\Aout)}$ be a kernel field transform.
    Then $\TK$ is said to be \emph{equivariant w.r.t the action of isometries} in a subgroup $\I \leq \IsomGM$ if it commutes with this action, that is, if the following property holds:
    \begin{align}\label{eq:isom_equivariance_kernel_field_trafo_def}
        \TK \big( \phi\rhd\!f \big) \ =\ \phi\rhd\! \big( \TK(f) \big)
        \qquad \forall\ \ f\in\Gamma(\Ain), \ \ \phi\in \I
    \end{align}
    In terms of a diagram, $\TK$ is equivariant w.r.t isometries in $\I$ if
    \begin{equation}
    \begin{tikzcd}[column sep=70pt, row sep=35, font=\normalsize]
        \Gamma(\Ain)
            \arrow[r, "\TK"]
            \arrow[d, "\phi\,\rhd"']
        &
        \Gamma(\Aout)
            \arrow[d, "\phi\,\rhd"]
        \\
        \Gamma(\Ain)
            \arrow[r, "\TK"']
        &
        \Gamma(\Aout)
    \end{tikzcd}
    \end{equation}
    commutes for all $\phi \in \I$.
\end{dfn}
A visualization of this definition is given in Fig.~\ref{fig:lizard_conv_egg}.
In the following Section~\ref{sec:isometry_constraint} we will derive a constraint on kernel fields in order for the corresponding kernel field transform to be isometry equivariant.
The geometrically intuitive result which we obtain is that the kernel field itself is required to be invariant under the action of isometries, which implies a form of weight sharing over the isometry orbits; see Fig.~\ref{fig:isom_invariant_kernel_field_multiple_orbits}.
Section~\ref{sec:isom_equiv_GM_conv} applies these insights to the more specific case of $\GM$-convolutions and $\GM$-convolutional kernel fields.
It turns out that $\GM$-convolutions are by the $G$-steerability of their template kernel automatically equivariant with respect to \emph{any} isometry in $\IsomGM$.

\subsubsection{Isometry equivariance of general kernel field transforms}
\label{sec:isometry_constraint}

The main result of this section, Theorem~\ref{thm:isometry_equivariant_kernel_field_trafos}, states that \emph{a kernel field transform $\TK$ is isometry equivariant if and only if its underlying kernel field $\K$ is invariant under isometries}.
To make sense of this statement we start by defining the transformation behavior of kernel fields when being acted on by isometries.
\begin{dfn}[Isometry action on kernel fields]
\label{dfn:isometry_action_kernel_fields}
    Let $\K: \TM \to \Hom(\Ain,\Aout)$ be a kernel field.
    An isometry $\phi \in \IsomGM$ acts on $\K$ via the kernel field pushforward
    \begin{align}\label{eq:kernel_constraint_isom_full_1}
        \dphiK \K\ :=\ \dphiHom \!\circ \K \circ \dphiTMinv \,.
    \end{align}
    Intuitively, this pushforward of kernel fields can be thought of as moving the individual kernels $\Kp$ at points $p\in M$ along the orbits of the isometry group to $\phi(p)$.
\end{dfn}
Since kernel fields are defined to be bundle $M$-morphisms, that is, to satisfy $\piHom \K = \piTM$, their pushforward is only well defined if it preserves this property.
This is guaranteed since the pushforward on the tangent bundle and homomorphism bundle are bundle maps, satisfying $\piTM \circ \dphiTM = \phi \circ \piTM$ (Eq.~\eqref{cd:pushforward_TM}) and ${\piHom \circ \dphiHom = \phi \circ \piHom}$ (Eq.~\eqref{cd:associated_bdl_automorphism}), respectively:
\begin{align}
    \piHom\, \dphiK \K
    \ &=\ \piHom\, \dphiHom \K\; \dphiTMinv \notag \\
    \ &=\ \phi\, \piHom\, \K\; \dphiTMinv \notag \\
    \ &=\ \phi\, \piTM\, \dphiTMinv \notag \\
    \ &=\ \phi\, \phiinv\, \piTM \notag \\
    \ &=\ \piTM
\end{align}
We visualize the definition of the isometry action on kernel fields by a commutative diagram:
\begin{equation}
    \begin{tikzcd}[row sep=3.5em, column sep=2.8em]
        &[3.5ex] M & \\
        \TM  \arrow[ru, "\piTM"]
            \arrow[rr, pos=.56, "\dphiK \K"']
        & &
        \mkern-3mu
        \Hom(\Ain,\Aout)
            \arrow[lu, "\piHom"']
        \\
        \TM  \arrow[rd, "\piTM"']
            \arrow[rr, pos=.57, "\K"]
            \arrow[u, "\dphiTM"]
        & &
        \mkern-3mu
        \Hom(\Ain,\Aout)
            \arrow[ld, "\piHom"]
            \arrow[u, "\dphiHom"']
        \\
        & M 
            \arrow[uuu, rounded corners, to path={ 
                    -- ([xshift=-25ex]\tikztostart.west) 
                    --node[left, pos=.515]{\small$\phi$} ([xshift=-25ex]\tikztotarget.west) 
                    -- (\tikztotarget.west)
                    }]
        &
    \end{tikzcd}
    \quad
\end{equation}
The bottom part of this diagram shows the coordinate free kernel field $\K$ from the diagram in Eq.~\eqref{eq:kernel_bundle_map} while the upper part shows its pushforward $\dphiK \K = \dphiHom \!\circ \K \circ \dphiTMinv$ by $\phi \in \IsomGM$.
The commutativity of the leftmost arrow, which asserts that $\dphiK$ moves kernels from $p$ to $\phi(p)$, follows from $\dphiTM$ and $\dphiHom$ both being bundle maps over $\phi$.

\begin{SCfigure}
    \centering
    \includegraphics[width=.28\textwidth]{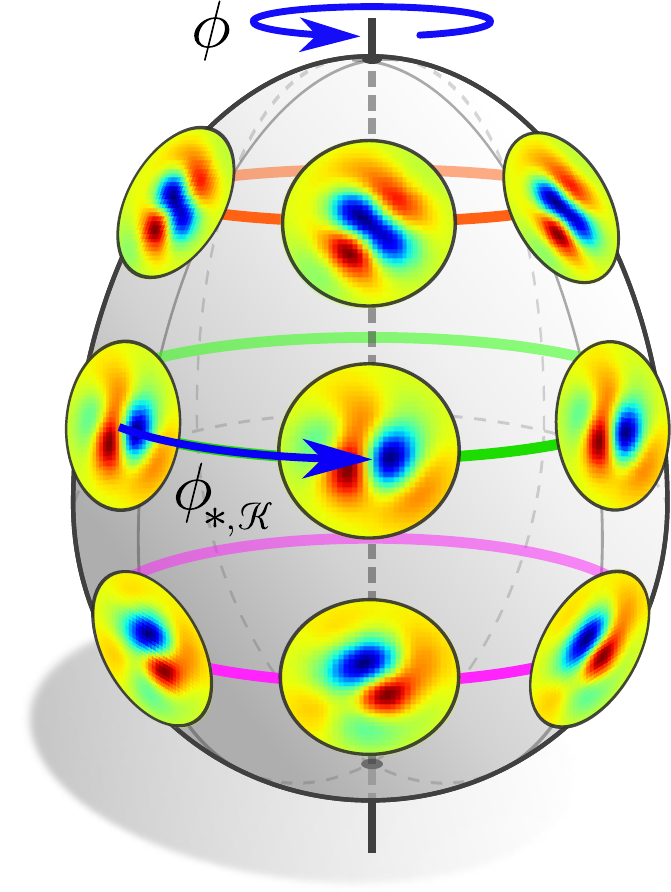}
    \hspace{10ex}
    \captionsetup{width=1.2\textwidth}
    \caption[]{\small
        Visualization of an invariant kernel field $\K$ for the case of an isometry (sub)group $\I = \SO2$.
        The invariance constraint requires $\dphiK \K := \dphiHom \K \dphiTMinv = \K$ for any $\phi$ in $\I$.
        It enforces kernels to be shared over the orbits $\I.p := \{\phi(p) \,|\, \phi \in \I\}$ of the action but allows for different kernels on different orbits.
        Theorem~\ref{thm:isometry_equivariant_kernel_field_trafos} proves that invariant kernel fields and equivariant kernel field transforms imply each other.
        This is intuitively clear since a specific pattern in the feature field at~$p\in M$ will evoke the same response when being transported to $\phi(p)$ if and only if the kernels at both points coincide.
        For the choice of $\I = \O2$ as isometry group, the kernels would additionally have to satisfy a reflectional constraint; see Fig.~\ref{fig:isom_invariant_kernel_field_quotient}.
        \\\protect\rule{0ex}{2.ex}
        }
    \label{fig:isom_invariant_kernel_field_multiple_orbits}
\end{SCfigure}

We proceed by defining isometry invariant kernels fields -- a visualization is found in Fig.~\ref{fig:isom_invariant_kernel_field_multiple_orbits}.
\begin{dfn}[Isometry invariant kernel fields]
\label{dfn:isometry_invariant_kernel_fields}
    A kernel field $\K$ is said to be \emph{invariant%
    \footnote{
        Instead of saying that $\K$ is \emph{invariant}, one could call it \emph{equivariant} since it satisfies
        $\dphiHom \!\circ \K \,=\ \K \circ \dphiTM \ \forall \phi\in\I$.
        In general, any function which is equivariant w.r.t. certain group actions on its domain and codomain is \emph{itself} invariant under pre- and postcomposition with the actions on its domain and codomain as in Eq.~\eqref{eq:kernel_constraint_isom_full_1}.
    }
    under isometries} in $\I \leq \IsomGM$ if it satisfies the constraint $\dphiK \K = \K$ for all $\phi \in \I$.
    We denote the space of isometry invariant kernel fields by
    \begin{align}\label{eq:KIfull_def}
        \KIfull :=
            \pig\{\mkern2mu \K\mkern-2mu:\TM\to \Hom(\Ain,\Aout) \ \ \textup{smooth}\ \,\pig|\; 
            \piHom\!\circ\K = \piTM \,,\ \ \ 
            \dphiK \K  = \K \quad \forall \phi\in\I \,\pig\} \,.
    \end{align}
\end{dfn}
By writing out $\dphiK$, the invariance constraint reads
\begin{align}\label{eq:kernel_constraint_isom_full_1}
    \dphiHom \!\circ \K \circ \dphiTMinv \,=\ \K \qquad \forall \phi\in\I \,,
\end{align}
which, after further expanding $\dphiHom$ as defined in Eq.~\eqref{eq:pushforward_Hom_def}, becomes:
\begin{align}\label{eq:kernel_constraint_isom_full_2}
    \dphiAout
    \K \big(\dphiTMinv v\big) \ 
    \dphiAininv
    =\,
    \K(v) \qquad \forall\ v\in \TM,\ \ \forall \phi \in \I
\end{align}
Note the similarity of these kernel field constraints in
Eqs.~\eqref{eq:kernel_constraint_isom_full_1} and~\eqref{eq:kernel_constraint_isom_full_2} with the $G$-steerability constraint on template kernels in Eqs.~\eqref{eq:G_steerable_space_in_dfn_Hom} and~\eqref{eq:G_steerable_space_in_dfn_classical}, respectively.
Indeed, both constraints are closely related and imply each other to a certain extent as we will show in the following Section~\ref{sec:isom_equiv_GM_conv} on isometry equivariant $\GM$-convolutions.

The following theorem proves that kernel fields which are invariant under isometries do indeed correspond to isometry equivariant kernel field transforms:
\begin{thm}[Equivariant kernel field transform $\Leftrightarrow$ invariant kernel field]
\label{thm:isometry_equivariant_kernel_field_trafos}
    \ A kernel field transform $\TK: \Gamma(\Ain)\to \Gamma(\Aout)$ is equivariant w.r.t. isometries in $\I \leq \IsomGM$ according to Def.~\ref{dfn:isometry_equivariance} if and only if
    the underlying kernel field~$\K$ is invariant under isometries according to Def.~\ref{dfn:isometry_invariant_kernel_fields}, that is,
    \begin{align}
        \TK(\phi \rhd f) \,=\, \phi \rhd \TK(f)
        \quad\forall\; \phi \in \I\,,\ f\in\Gamma(\Ain)
        \quad \Longleftrightarrow \quad
        \dphiK \K = \K
        \quad\forall\, \phi \in \I
    \end{align}
\end{thm}
\begin{proof}
    To prove this theorem, we write out the kernel field transforms and feature field pushforwards on both sides of the isometry equivariance condition in Eq.~\eqref{eq:isom_equivariance_kernel_field_trafo_def}.
    The statement follows from a comparison of both sides after a few algebraic manipulations.

    We start with the right-hand side of Eq.~\eqref{eq:isom_equivariance_kernel_field_trafo_def}:
    \begin{align}
        \big[ \phi \rhd \TK(f) \big](p)
        \ &\overset{(1)}{=}\ \ 
            \dphiAout \big[ \TK(f) \big] \big( \phi^{-1}(p) \big) \notag \\[1ex]
        \ &\overset{(2)}{=}\ \ 
            \dphiAout \mkern-8mu
            \int\limits_{T_{\!\phiinv\mkern-1mu(p)}\mkern-2muM}\!\!
            \mkern-6mu \K(v) \ 
            \big[\mkern-2mu \Expsphiinvpf \mkern1mu\big] (v)
            \,\ dv \notag \\
        \ &\overset{(3)}{=}\ \ 
            \dphiAout \mkern-8mu
            \int\limits_{T_{\!\phiinv\mkern-1mu(p)}\mkern-2muM}\!\!
            \mkern-6mu \K(v) \ 
            \dphiAin^{-1} \big[\mkern-2mu \Expsp\! \big(\phi \rhd f\big) \mkern1mu\big] \big( \dphiTM (v) \big)
            \,\ dv \notag \\
        \ &\overset{(4)}{=}\ \ 
            \int\limits_{\TpM}
            \dphiAout
            \K\big( \dphiTM^{-1} \tilde{v} \big) \ 
            \dphiAin^{-1} \big[\mkern-2mu \Expsp\! \big(\phi \rhd f\big) \mkern1mu\big] (\tilde{v})
            \,\ d\tilde{v} \notag \\
        \ &\overset{(5)}{=}\ \ 
            \int\limits_{\TpM}
            \big[ \dphiHom \K\, \dphiTM^{-1} \big] (\tilde{v}) \ 
            \big[\mkern-2mu \Expsp\! \big(\phi \rhd f\big) \mkern1mu\big] (\tilde{v})
            \,\ d\tilde{v} \notag \\
        \ &\overset{(6)}{=}\ \ 
            \int\limits_{\TpM}
            \big[ \dphiK \K \big] (\tilde{v}) \ 
            \big[\mkern-2mu \Expsp\! \big(\phi \rhd f\big) \mkern1mu\big] (\tilde{v})
            \,\ d\tilde{v}
    \end{align}
    Steps~$(1)$ and~$(2)$ expand the isometry action $\rhd$ on feature fields (Def.\ref{dfn:isometry_pushforward}) and the kernel field transform (Def.~\ref{dfn:kernel_field_trafo}).
    The transformation law of the field's transporter pullback in Theorem~\ref{thm:transporter_pullback_isometry_action}, which relies on the $\IsomGM$-invariance of the $G$-compatible connection, justifies step~$(3)$.
    Step~$(4)$ substitutes $v$ with $\tilde{v} = \dphiTM v$.
    Since $\phi$ is an isometry, the change of volume equates to~$1$.
    Steps~$(5)$ and~$(6)$ identify the action of the kernel pushforward~$\dphiK$, Def.~\ref{dfn:isometry_action_kernel_fields}.
    The resulting statement is quite intuitive:
    A transformation of the kernel field transform's output corresponds to a simultaneous transformation of its input \emph{and} kernel field.

    Writing out the left-hand side yields
    \begin{align}
        \big[ \TK \big(\phi \rhd f\big) \big](p)
        \ =\ \int\limits_{\TpM}
                \K(v) \ 
                \big[\mkern-2mu \Expsp\!\big( \phi\rhd f\big) \mkern1mu\big] (v)
                \,\ dv \,,
    \end{align}
    which is equivalent to the right-hand side \emph{up to} the transformation of the kernel field.

    Isometry equivariance requires both expressions to agree for arbitrary fields $f\in\Gamma(\Ain)$, points $p\in M$ and isometries $\phi\in\I$.
    This is the case if and only if $\dphiK \K = \K$ holds for any $\phi \in \I$, i.e. if the kernel field is invariant under the action of isometries.
\end{proof}
Note that this proof would have been very cumbersome to work out in (a $G$-atlas of) local trivializations.
The global, coordinate free description of kernel field transforms allows for a simple proof without having to worry that the isometries move features between different local trivializations.

At this point we could proceed with a further investigation of isometry invariant kernel fields:
since the invariance constraint implies kernels to be shared over orbits of the isometry group, a description of the entire kernel field on the full manifold is redundant.
It is therefore possible to reduce the description of such kernel fields to kernel fields on quotient spaces.
As this analysis is not required to prove the isometry equivariance of $\GM$-convolutions and requires some technical definitions, we postpone it to Section~\ref{sec:quotient_kernel_fields}

\subsubsection{Isometry equivariance of \textit{GM}-convolutions}
\label{sec:isom_equiv_GM_conv}

Recall that $\GM$-convolutions (Def.~\ref{dfn:coord_free_conv}) were defined as specific kernel field transforms with $\GM$-convolutional kernel fields (Def.~\ref{dfn:conv_kernel_field}).
The results on the isometry equivariance of kernel field transforms therefore immediately apply to $\GM$-convolutions as well.
However, in addition to the isometry invariance constraint in Eq.~\eqref{eq:kernel_constraint_isom_full_1}, $\GM$-convolutional kernel fields need to satisfy the $G$-steerability constraint on the template kernel from Eq.~\eqref{eq:G_steerable_space_in_dfn_Hom} and share weights over the $G$-structure according to Eq.~\eqref{eq:conv_kernel_field_def_ptwise}.
In order for the $\GM$-convolution to be isometry equivariant, all of these constraints have to be satisfied simultaneously.
Intuitively, this implies that the convolutional weight sharing needs to agree with the isometry induced weight sharing over orbits.
Luckily it turns out that this is automatically the case for the isometries under consideration:
$\GM$-convolutions share weights over the $G$-structure and the isometries in $\IsomGM$ preserve the $G$-structure such that \emph{$\GM$-convolutional kernel fields are guaranteed to be $\IsomGM$ invariant}.
In coordinates, this reflects in the $\IsomGM$-induced gauge transformations $g_\phi^{A\widetilde{A}}(p)$ taking values in the structure group $G$, such that they are explained away by the $G$-steerability of the template kernels.

To make these arguments more rigorous, consider a $\GM$-convolution $K\star: \Gamma(\Ain) \to \Gamma(\Aout)$ with some $G$-steerable kernel $K \in \KG$, which is by Def.~\ref{dfn:coord_free_conv} just the kernel field transform $\TKK$ with the $\GM$-convolutional kernel field $\KK$.
By Theorem~\ref{thm:isometry_equivariant_kernel_field_trafos}, the $\GM$-convolution is therefore exactly then $\IsomGM$-equivariant if~$\KK$ is $\IsomGM$-invariant, i.e. when it satisfies $\dphiK \KK = \dphiHom \circ \KK \circ \dphiTMinv = \KK$ for any $\phi \in \IsomGM$.
This constraint on the full kernel field is equivalently expressed by a set of constraints on the individual convolution kernels that make up the field:
\begin{align}\label{eq:KK_constraint_ptwise_abstract}
    \dphiHom \circ \KKp \circ \dphiTMinv \ &=\ \KKphip \qquad \forall\ p\in M,\ \ \phi \in \IsomGM
\end{align}
Considering a specific point $p\in M$, we choose arbitrary gauges $\widetilde{A}$ at $p$ and $A$ at $\phi(p)$ from the $G$-atlas.
The $\GM$-convolutional kernel field is by Def.~\ref{dfn:conv_kernel_field} at $p$ and $\phi(p)$ given by
\begin{align}
    \KKp    := \big(\psiHomp^{\widetilde{A}} \big)^{-1} \circ \frac{K}{\sqrt{|\eta_p^{\widetilde{A}}|}} \circ \psiTMp^{\widetilde{A}}
    \qquad \textup{and} \qquad
    \KKphip := \big(\psiHomphip^A            \big)^{-1} \circ \frac{K}{\sqrt{|\eta_{\phi(p)}^A|}}       \circ \psiTMphip^A \,.
\end{align}
Plugging these expressions into the constraint in Eq.~\eqref{eq:KK_constraint_ptwise_abstract} for the fixed $p$ and identifying $\dphiHom \big(\psiHomp^{\widetilde{A}} \big)^{-1}$ with $\big( \psiHomp^{\widetilde{A}}\, \dphiHom^{-1}  \big)^{-1}$ yields:
\begin{align}\label{eq:GM-conv_kernel_field_isom_invariance_1}
    \big( \psiHomp^{\widetilde{A}}\, \dphiHom^{-1}  \big)^{-1} \circ \frac{K}{\sqrt{|\eta_p^{\widetilde{A}}|}} \circ \psiTMp^{\widetilde{A}}\, \dphiTMinv
    \,\ &=\,\ \big(\psiHomphip^A \big)^{-1} \circ \frac{K}{\sqrt{|\eta_{\phi(p)}^A|}} \circ \psiTMphip^A
    \qquad \forall\ \phi \in \IsomGM
\end{align}
The isometry equivariance will therefore hold if the weight sharing via the isometry induced gauges $\psidotp^{\widetilde{A}} \dphidot$ agrees with the weight sharing via the original gauges $\psidotphip^A$ from $\phi(p)$.
Recall that the isometry induced gauges are by Theorem~\ref{thm:isom_GM_in_coords} for isometries in $\IsomGM$ guaranteed to be compatible with the $G$-atlases (of the corresponding bundle).
As shown in Eq.~\eqref{eq:arbitrariness_gauge_GM_kernel_field_def}, the particular choice of gauge, relative to which the $G$-steerable template kernel is oriented, is irrelevant, as long as the gauges are $G$-compatible.
Since all derivations were independent from the chosen point $p$ and the particular choice of gauges, this implies that $\GM$-convolutions are by design guaranteed to be $\IsomGM$-equivariant.

To gain a better intuition for this result it is worth to make the induced, $G$-valued gauge transformations $g_\phi^{A\widetilde{A}}(p)$ explicit.
To this end, note that the commutativity of the diagrams in Eqs.~\eqref{cd:pushforward_A_coord} and~\eqref{cd:pushforward_TM_coord} implies
$\psiHomp^{\widetilde{A}}\, \dphiHominv = \rhoHom\big( g_\phi^{A\widetilde{A}}(p) \big)^{-1} \psiHomphip^A$ and
$\psiTMp^{\widetilde{A}}\,  \dphiTMinv  =        \big( g_\phi^{A\widetilde{A}}(p) \big)^{-1} \psiTMphip^A$.
Inserting these coordinate expressions into the constraint in Eq.~\eqref{eq:GM-conv_kernel_field_isom_invariance_1} leads to the requirement that
\begin{align}\label{eq:GM-conv_kernel_field_isom_invariance_2}
    \Big(\mkern-1.5mu \rhoHom\big( g_\phi^{A\widetilde{A}}(p) \big)^{-1} \psiHomphip^A \mkern-1.5mu\Big)^{-1}
    \!\circ \frac{K}{\sqrt{|\eta_p^{\widetilde{A}}|}} \circ \big( g_\phi^{A\widetilde{A}}(p) \big)^{-1} \psiTMphip^A
    \ &=\ \big(\psiHomphip^A \big)^{-1} \!\circ \frac{K}{\sqrt{|\eta_{\phi(p)}^A|}} \circ \psiTMphip^A
\end{align}
needs to hold for any isometry $\phi$ in~$\IsomGM$.
By expanding the inverse on the left-hand side, using that
${\sqrt{|\eta_p^{\widetilde{A}}|} \,= \sqrt{|\eta_{\phi(p)}^A|} \cdot \big|\mkern-2mu \det g^{A\widetilde{A}}_\phi(p) \big|}$
and dropping the gauges, which is possible since they are isomorphisms, we end up with the constraint
\begin{align}\label{eq:GM_conv_isom_constraint_coords}
    \frac{1}{\big|\mkern-2mu \det g^{A\widetilde{A}}_\phi(p) \big|} \,
    \rhoHom\Big(g^{A\widetilde{A}}_\phi(p)\Big) \circ K \circ \Big(g^{A\widetilde{A}}_\phi(p)\Big)^{-1} \ =\ K
    \qquad \forall\ \phi \in \IsomGM \,,
\end{align}
which looks \emph{exactly} like the $G$-steerability kernel constraint on~$K$ from Def.~\ref{dfn:G-steerable_kernel_def_43}.
Recall that the isometry induced gauge transformations $g_\phi^{A\widetilde{A}}(p)$ are by Theorem~\ref{thm:isom_GM_in_coords} guaranteed to be $G$-valued if~$\phi$ is an element of $\IsomGM$.
The constraint in Eq.~\eqref{eq:GM_conv_isom_constraint_coords} is therefore always satisfied by the $G$-steerability of~$K$.

The derived results on the $\IsomGM$-invariance of $\GM$-convolutional kernel fields $\KK$ are concisely summarized by the statement that the following diagram is guaranteed to be commutative if $K$ is $G$-steerable and if $\phi \in \IsomGM$ is $G$-structure preserving:
\begin{equation}
    \begin{tikzcd}[row sep=4.em, column sep=5.5em] 
        & &[-2.7em] U^A &[-2.7em] & \\[-1em]
        U^A \mkern-2mu\times\mkern-1mu \R^d
            \arrow[rrrr, rounded corners, to path={ 
                    -- ([yshift=15ex]\tikztostart.north) 
                    --node[above]{\small
                            $\id\times \pig( \rhoHom\big(g'^{A\widetilde{A}}_\phi\big) \circ K \big/\mkern-2mu \sqrt{|\eta^A|} \circ \big(g'^{A\widetilde{A}}_\phi\big)^{-1} \pig)$
                            } ([yshift=15ex]\tikztotarget.north) 
                    -- (\tikztotarget.north)
                    }]
        &
        \,\piTM^{-1}\big(U^A\big)
            \arrow[ru, "\piTM"]
            \arrow[rr, "\dphiK \KK"']
            \arrow[l, "\PsiTM^A"']
        & &
        \piHom^{-1}\big(U^A\big)
            \arrow[lu, "\piHom"']
            \arrow[r, "\PsiHom^A"]
        &
        U^A \mkern-2mu\times\mkern-1mu \R^{\cout\times\cin}
        \\
        U^{\widetilde{A}} \mkern-2mu\times\mkern-1mu \R^d
            \arrow[u, "\phi \times g_\phi^{A\widetilde{A}}\cdot"']
            \arrow[rrrr, rounded corners, to path={ 
                    -- ([yshift=-15ex]\tikztostart.south) 
                    --node[below]{\small$\id\times K \big/\mkern-2mu \sqrt{|\eta^{\widetilde{A}}|}$} ([yshift=-15ex]\tikztotarget.south) 
                    -- (\tikztotarget.south)
                    }]
        &
        \piTM^{-1}\big(U^{\widetilde{A}}\big)
            \arrow[rd, "\piTM"']
            \arrow[rr, "\KK"]
            \arrow[u, "\dphiTM"]
            \arrow[l, "\PsiTM^{\widetilde{A}}"]
        & &
        \piHom^{-1}\big(U^{\widetilde{A}}\big)
            \arrow[ld, "\piHom"]
            \arrow[u, "\dphiHom"']
            \arrow[r, "\PsiHom^{\widetilde{A}}"']
        &
        U^{\widetilde{A}} \mkern-2mu\times\mkern-1mu \R^{\cout\times\cin}
            \arrow[u, "\phi \times \rhoHom\big(g_\phi^{A\widetilde{A}}\big)"]
        \\[-1em]
        & & U^{\widetilde{A}}
        & &
    \end{tikzcd}
\end{equation}
Here we defined the pullback $g_\phi'^{A\widetilde{A}} := g_\phi^{A\widetilde{A}} \circ \phi^{-1} : U^A \to G$ of the isometry pushforward coordinatization from $U^{\widetilde{A}}$ to $U^A$ for notational convenience.

Together with Theorem~\ref{thm:isometry_equivariant_kernel_field_trafos}, the $\IsomGM$-invariance of $\GM$-convolutional kernel fields implies the $\IsomGM$-equivariance of $\GM$-convolutions:
\begin{thm}[Isometry equivariance of $\GM$-convolutions]
\label{thm:isom_equiv_GM_conv}
    A $\GM$-convolution $K\star: \Gamma(\Ain)\to \Gamma(\Aout)$ with a $G$-steerable kernel $K\in\KG$ is equivariant with respect to all $G$-structure preserving isometries $\phi \in \IsomGM$, that is,
    \begin{align}
        K \star \big( \phi \rhd f \big) \ =\ 
        \phi \rhd \big( K \star  f \big)
        \qquad \forall\ \ f\in\Gamma(\Ain), \ \ \phi\in \IsomGM \,.
    \end{align}
    The following diagram commutes therefore for every $\phi \in \IsomGM$:
    \begin{equation}
    \begin{tikzcd}[column sep=60pt, row sep=35, font=\normalsize]
        \Gamma(\Ain)
            \arrow[r, "K\star"]
            \arrow[d, "\phi\,\rhd\,"']
        &
        \Gamma(\Aout)
            \arrow[d, "\,\phi\,\rhd"]
        \\
        \Gamma(\Ain)
            \arrow[r, "K\star"']
        &
        \Gamma(\Aout)
    \end{tikzcd}
    \end{equation}
\end{thm}
\begin{proof}
    The proof was given in the discussion prior to the theorem.
\end{proof}

Having this general result on $\GM$-convolutions derived, we will now discuss some special cases for specific choices of structure groups~$G$.
Firstly, for orthogonal structure groups $G=\O{d}$ (or supergroups of it), the convolution will commute with \emph{any} isometry:
\begin{thm}[Full isometry equivariance of $\OM$-convolutions]
\label{thm:Od_equiv_OM_conv}
    $\OM$-convolutions are equivariant w.r.t. the action of \emph{any} isometry $\phi \in \IsomM$ on feature fields.
    More generally, any $\GM$-convolution for $G$-structures with structure groups $G\geq\O{d}$ is fully isometry equivariant.
\end{thm}
\begin{proof}
    The statement follows from Theorem~\eqref{thm:isom_equiv_GM_conv} by observing that $\IsomGM = \IsomM$ is guaranteed for structure groups $G \geq \O{d}$.
    The latter was discussed in Eq.~\eqref{eq:isomM_isomOM}.
\end{proof}
This result relies essentially on the fact that isometries are defined as that subgroup of diffeomorphisms on~$M$ which induce $\O{d}$-structure automorphisms.
Less abstractly stated, $\IsomM$ is by definition that subgroup of diffeomorphisms which respect the Riemannian metric~$\eta$ of~$M$ and the corresponding $\O{d}$-structure $\OM$ is equivalent information to the metric.

On orientable Riemannian manifolds one can furthermore pick an orientation (frame handedness), which together with the metric defines an $\SO{d}$-structure $\SOM$.
The corresponding isometries which lift to $\SO{d}$ structure automorphisms are the orientation preserving isometries in $\IsomplusM$.
\begin{thm}[$\IsomplusM$ equivariance of $\SOM$-convolutions]
\label{thm:SOd_equiv_SOM_conv}
    $\SOM$-convolutions are equivariant w.r.t. the action of orientation preserving isometries $\phi \in \IsomplusM$ on feature fields.
\end{thm}
\begin{proof}
    This result follows from Theorem~\eqref{thm:isom_equiv_GM_conv} by observing that $\IsomSOM = \IsomplusM$.
\end{proof}
For instance, an $\SOM$-convolution for $M=\R^2$, corresponding to Fig~\ref{fig:SO2_structure_SE2}, is equivariant w.r.t. the action of the special Euclidean group $\Isom_+(\R^2) = \SE2$.
Similarly, an $\SOM$-convolution for $M=S^2$, corresponding to Fig~\ref{fig:SO2_structure_SO3}, is rotation equivariant with $\Isom_+(S^2) = \SO3$.

Note that the results of Theorems~\ref{thm:Od_equiv_OM_conv} and~\ref{thm:SOd_equiv_SOM_conv} depend only on the structure group $G$ but not on the particular choice of $G$-structure.
For subgroups $G$ of $\O{d}$ (or $\SO{d}$) things become more complicated.
In these cases the subgroups $\IsomGM$ of $\IsomM$ depend on the specific embedding of the $G$-structure~$\GM$ into~$\FM$.
This was for $G=\{e\}$ visualized in Fig.~\ref{fig:frame_field_automorphism}.
Specifically, Fig.~\ref{fig:frame_field_automorphism_1} shows the canonical $\{e\}$-structure of $\R^2$, which is fully translation equivariant, that is, $\IsomeM = \Trans_2 := (\R^2,+)$.
In contrast, Fig.~\ref{fig:frame_field_automorphism_2} shows an $\{e\}$-structure of $\R^2$ which is only translation equivariant along one axis such that $\IsomeM \cong \Trans_1 := (\R,+)$.
From the viewpoint of convolutional networks this result is very intuitive:
The $\{e\}$-steerable kernels in these examples are unconstrained, i.e. conventional convolution kernels.
They do therefore in general not carry any information about their responses when being applied relative to gauge transformed reference frames.
Since the frames, and therefore kernels, in Fig.~\ref{fig:frame_field_automorphism_2} are differently rotated along the ``left-right'' direction, the kernel responses change unpredictably when translating a signal in that direction.
If the template kernels would, however be $\SO2$-steerable, they could account for the rotation of frames.
This case corresponds to the situation in Fig~\ref{fig:SO2_structure_SE2}, i.e. an $\SOM$-convolution.

%% file: chapters/83_quotient_kernel_fields.tex

\subsection{Quotient kernel fields}
\label{sec:quotient_kernel_fields}

Theorem~\ref{thm:isometry_equivariant_kernel_field_trafos} showed that the isometry equivariance of a kernel field transform requires the invariance of the corresponding kernel field.
Since the invariance constraint implies kernels to be shared over orbits as visualized in Fig.~\ref{fig:isom_invariant_kernel_field_multiple_orbits}, the mathematical description of such invariant kernel fields is redundant:
a single kernel at one orbit representative is sufficient to reconstruct the kernel field on the whole orbit.
In Section~\ref{sec:quotient_kernels_stabilizers} we derive equivalent, reduced descriptions of invariant kernel fields in terms of kernels on orbit representatives.
These representative kernels are themselves constrained by the action of the stabilizer subgroup of the orbit representative.
We propose a (unique) lifting from representative kernels to invariant kernel fields, which establishes an isomorphism between both descriptions.
This lifting isomorphism suggest a way of parameterizing and constructing isometry equivariant kernel field transforms in an implementation.
Before deriving these results in Section~\ref{sec:quotient_kernels_stabilizers}, the following Section~\ref{sec:isom_quotients} sets up the mathematical framework.

The derivations and results of this section are close in spirit to the theory of \emph{steerable CNNs on homogeneous spaces}~\cite{Cohen2018-intertwiners,Cohen2019-generaltheory},
however, we generalize their results from homogeneous spaces to general manifolds.
When sticking to homogeneous spaces $M$, we prove that isometry equivariant kernel field transforms are equivalent to $\GM$-convolutions.

\subsubsection{Isometry induced quotient spaces}
\label{sec:isom_quotients}

The action of a symmetry group on a space partitions it into orbits, defined as the sets of all points which are connected by the group action.
The space of such orbits is the \emph{quotient space} w.r.t. this group action.
In the following we will discuss the quotient spaces arising from the actions of some isometry group $\I \leq \IsomGM$ both on the manifold and on the fiber bundles.
These definitions will later allow us to share weights over orbits by acting with isometries on kernels.

\paragraph{Manifold quotients:} 

\begin{figure}
    \centering%
    \subcaptionbox{\small Quotient map and orbit representatives for $M$.%
        \label{fig:isom_egg_quotient_M}}%
        [.44\linewidth][l]{%
            \includegraphics[width=.43\textwidth]{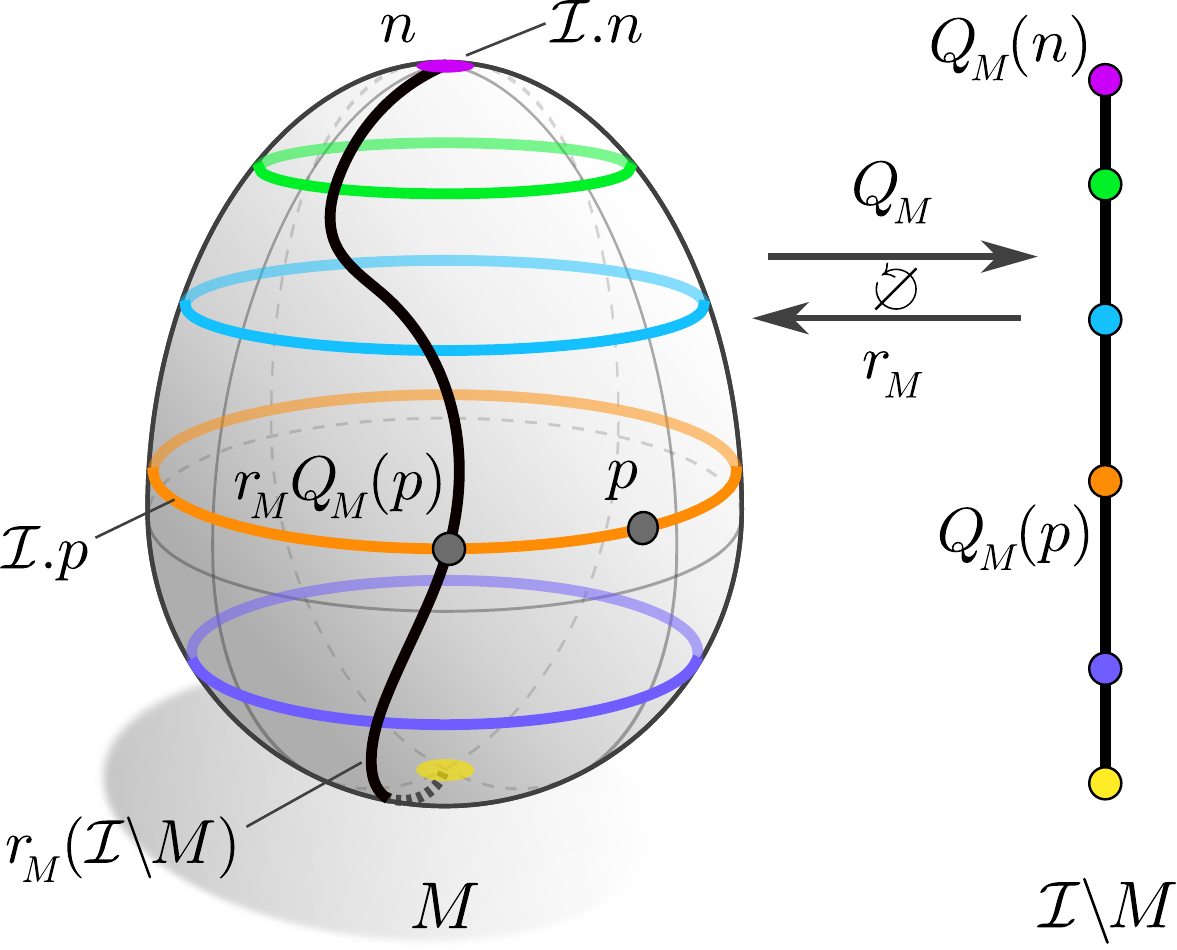}%
            \vspace*{1ex}%
        }%
    \hfill%
    \subcaptionbox{\small Quotient map and orbit representatives for $\TM$.%
        \label{fig:isom_egg_quotient_TM}}%
        [.5\linewidth][r]{%
            \includegraphics[width=.5\textwidth]{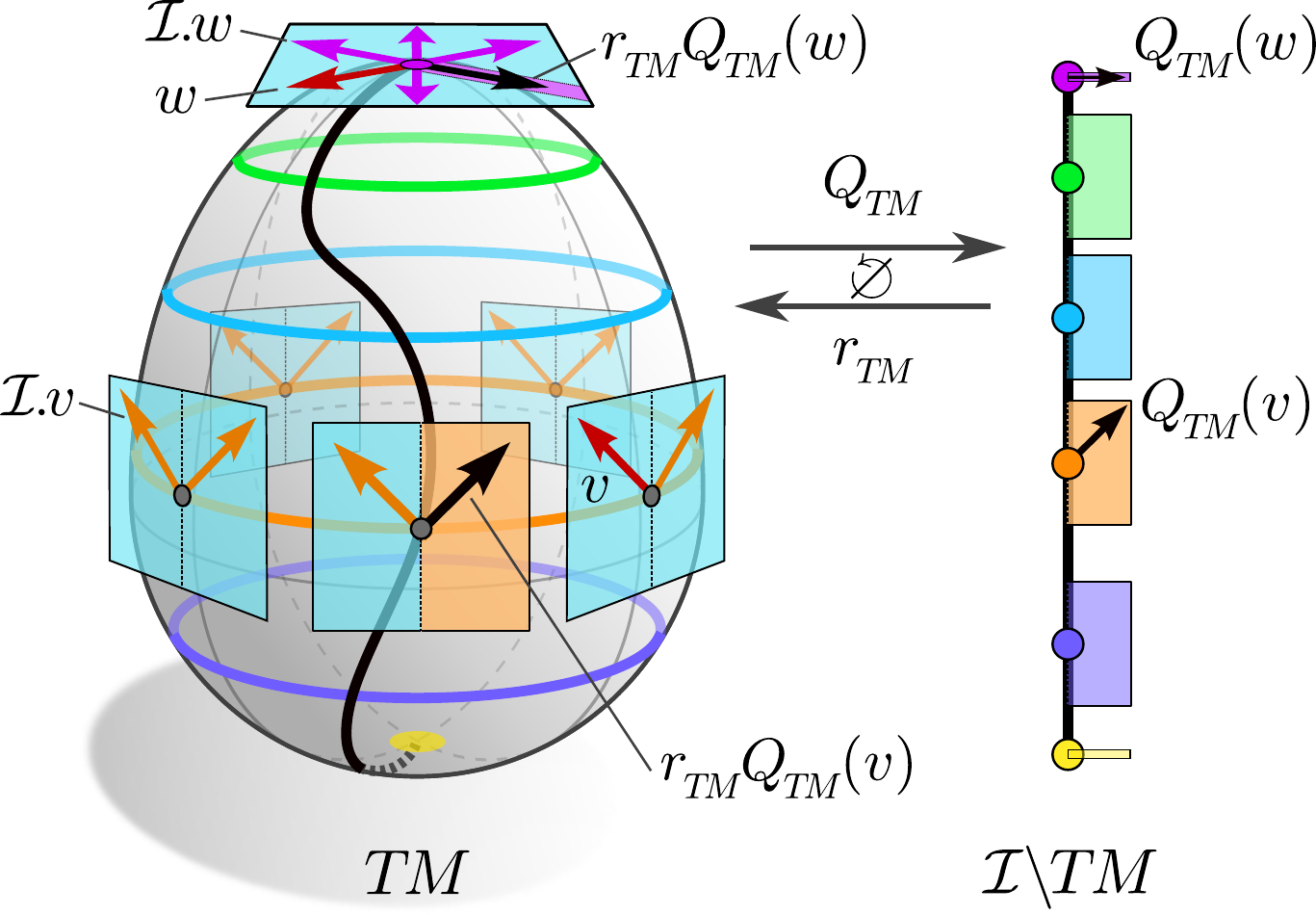}%
            \vspace*{1ex}%
        }%
    \caption{\small%
        Quotient maps $\QM$ and $\QTM$ and orbit representatives (sections) $\rM$ and $\rTM$ for the actions of the isometry group $\I=\O2$ on the manifold~$M$ in Fig.~\ref{fig:isom_egg_quotient_M} and on the tangent bundle $\TM$ in Fig.~\ref{fig:isom_egg_quotient_TM}.
        A detailed description of both visualizations is given in the main text.
    }%
    \label{fig:isom_egg_quotient_main}%
\end{figure}

Any point $p\in M$ traces out an \emph{orbit}
\begin{align}\label{eq:orbit_Ip}
    \I.p \,:=\, \big\{ \phi(p) \;\big|\; \phi \in\I \,\big\}\ \subseteq\, M \,,
\end{align}
which is defined as the set of all points reached by acting on $p$ with any isometry in $\I \leq \IsomM$.
One can easily check that the relation ``$p$ and $q$ are elements of the same orbit'' is an equivalence relation (see footnote \ref{footnote:equiv_rel}) and thus \emph{partitions} the manifold as visualized in Fig.~\ref{fig:isom_egg_quotient_M}.
The quotient space
\begin{align}\label{eq:quotientspace_IM}
    \IM \, :=\, \big\{ \I.p \,\big|\, p\in M \big\}
\end{align}
with respect to this equivalence relation is the space of all orbits, that is, each element of $\IM$ corresponds to a full orbit in~$M$.%
\footnote{%
    We write $\IM$ as a left quotient since $\I$ acts on $M$ from the left.
}
The corresponding \emph{quotient map}
\begin{align}\label{eq:quotientmap_QM}
    \QM: M\to\IM,\ \ p\mapsto \I.p
\end{align}
identifies a point $p \in M$ with its orbit $\I.p \in \IM$.
For each orbit one can select an arbitrary \emph{orbit representative}, formally determined by a \emph{section}
\begin{align}\label{eq:section_rM}
    \rM: \IM\to M \quad \textup{such that}\quad \QM\circ\rM = \id_{\IM} \,,
\end{align}
where the last condition ensures that the representative $\rM(\I.p)$ is indeed an element of the orbit $\I.p$.
One is often interested in continuous (or smooth) sections, however, these do in general not exist.
We will therefore in the following \emph{not} demand the orbit representatives to be chosen continuously and make up for this shortcoming post-hoc if necessary.
As usual for sections, they are in general only right inverses of the quotient map but not left inverses, that is, $\rM\circ\QM \neq \id_M$.
This is visualized by a commutative diagram
\begin{equation}
\begin{tikzcd}[row sep=3em, column sep=4.5em]
      \IM
            \arrow[r, "\rM"]
            \arrow[rr, rounded corners, to path={ 
                  -- ([yshift=-3.ex]\tikztostart.south) 
                  --node[below, pos=.5]{\small$\id_{\IM}$} ([yshift=-3.ex]\tikztotarget.south) 
                  -- (\tikztotarget.south)
                  }]
    & M
            \arrow[r, "\QM"]
    & \IM
\end{tikzcd}
\end{equation}
similar to that in Eq.~\eqref{cd:section_proj_idM} and a non-commutative diagram
\begin{equation}
\begin{tikzcd}[row sep=3em, column sep=4.5em,
               execute at end picture={
                    \node [] at (-.04, -.46) {$\noncommutative$};
                    }]
      M
            \arrow[r, "\QM"]
            \arrow[rr, rounded corners, to path={ 
                  -- ([yshift=-3.5ex]\tikztostart.south) 
                  --node[below, pos=.5]{\small$\id_M$} ([yshift=-3.5ex]\tikztotarget.south) 
                  -- (\tikztotarget.south)
                  }]
    & \IM
            \arrow[r, "\rM"]
    & M
\end{tikzcd}
\end{equation}
similar to that in Eq.~\eqref{cd:section_proj_noncommutative}.
The individual fibers $\preim{\QM}\mkern-4mu(\I.p) = \I.p \subseteq M$ of the quotient map $\QM$ are given by the orbits themselves.
Note that $M\xrightarrow{\QM}\IM$ is in general \emph{not} a fiber bundle since the orbits are not necessarily homeomorphic to each other and can therefore not be locally trivialized with a shared typical fiber $F$, as required by the commutative diagram in Eq.~\eqref{cd:trivialization_general_intro}.
Each orbit therefore has an own \emph{type} which is in close relation to the stabilizer subgroups of the points on that particular orbit.
The \emph{stabilizer subgroup}
\begin{align}
    \Stab{p} \,:=\, \big\{ \xi \in \I \,\big|\, \xi(p)=p \big\} \ \leq\ \I
\end{align}
of a point $p\in M$ is thereby defined as that subgroup of the isometry group which leaves $p$ fixed.
In terms of the stabilizer subgroup, it holds that the orbit of a point is identified with
\begin{align}
    \I.p \ \cong\ \I/\Stab{p} \,.
\end{align}
To see this claim, let $f_p: \I \to \I.p,\ \phi \mapsto \phi(p)$ for some $p\in M$ and observe that $f_p(\phi\circ\xi) = \phi\circ\xi(p) = \phi(p) = f_p(\phi)$ for any $\xi\in\Stab{p}$.
It can easily be shown that indeed $\preim{f_p}\!\! \big(\phi(p)\big) = \phi.\Stab{p}$ is a coset of the stabilizer subgroup of $p$ and thus that $f_p$ establishes the claimed isomorphism $\I.p \,\cong\, \I/\Stab{p}$.

To make these constructions more intuitive, consider the example in Fig.~\ref{fig:isom_egg_quotient_M} with $\I \cong \O2$.
The orbits $\I.n = \{n\}$ and $\I.s = \{s\}$ of the north and south pole are just points, which are fixed by $\I$.
This agrees with, for instance, $\I.n \cong \I/\Stab{n} = \I/\I \cong \{n\}$ since $\Stab{n}=\I$ coincides with the full isometry group.
For any other point $p\in M$, the orbits $\I.p$ are circles.
We have reflections $\Stab{p} \cong \Flip$ (flipping over $p$) as stabilizer subgroup and thus indeed get the circle $\I/\Stab{p} \cong \O2/\Flip \cong S^1$ as orbit type.
The quotient map $\QM:M\to\IM$ sends points $q\in M$ to their orbits $\QM(q) = \I.q$ in the quotient space $\IM$, shown on the right.
Since the orbits can be traversed from the north to the south pole, the quotient space $\IM$ has the topology of a line segment.
The section $\rM:\IM\to M$ picks one representative point $\rM(o) \in M$ for any orbit $o \in \IM$.
In general, this orbit representative does not recover a projected point.
For instance, we have that $\rM\QM(p) \neq p$.
One can interpret the section as embedding the quotient space $\IM$ into the manifold, shown as the black line $\rM(\IM)$ from the north to the south pole.

\paragraph{Bundle quotients:}

Since the isometry group acts not only on the manifold itself but via pushforwards also on the associated bundles, these bundles are in a similar manner partitioned into orbits.
To keep the discussion general, we are in the following considering a generic associated bundle $E\xrightarrow{\pi_E} M$, which could stand for $\TM$, $\FM$, $\GM$, $\A$ or $\Hom(\Ain,\Aout)$.
We denote elements of the total space as $e\in E$ and let $\dphiE$ be the pushforward of $\phi$ on $E$ as introduced in Section~\ref{sec:isom_action_bundles}.
The orbit of an element of the bundle is then in analogy to Eq.~\eqref{eq:orbit_Ip} given by
\begin{align}\label{eq:bundle_orbit_def}
    \I.e \:=\ \big\{ \dphiE(e) \,\big|\, \phi\in\I \big\}
\end{align}
while the quotient space, consisting of bundle orbits, is analogously to Eq.~\eqref{eq:quotientspace_IM} defined as
\begin{align}
    \IE \:=\ \big\{ \I.e \,\big|\, e\in E \big\} \,.
\end{align}
Similar to before, the (canonical) quotient map sends bundle elements to their orbit:
\begin{align}
    \QE: E\mapsto\IE,\ \ e\mapsto\I.e
\end{align}
We define a (uniquely determined) projection map 
\begin{align}\label{eq:quotient_projection_piIE}
    \piIE\!:\, \IE \to \IM, \quad \QE(e) \mapsto \QM \circ \piE(e)
\end{align}
between the bundle quotients and manifold quotient as visualized in the following commutative diagram:
\begin{equation}\label{cd:QE_IE}
\begin{tikzcd}[column sep=50pt, row sep=30, font=\normalsize]
    \IE     \arrow[d, "\piIE"']
    &
    E       \arrow[l, "\QE"']
            \arrow[d, "\piE"]
    \\
    \IM
    &
    M       \arrow[l, "\QM"]
\end{tikzcd}
\end{equation}
Note that the definition in Eq.~\eqref{eq:quotient_projection_piIE} does not depend on the particular choice of orbit representative since for any other $\dphiE(e) \in \QE(e)$ we obtain the same result:
$    \QM \circ \piE \circ \dphiE(e)
\,=\,\QM \circ \phi \circ \piE(e)
\,=\,\QM \circ \piE(e) .
$
Orbit representatives are formally determined by a choice of section
\begin{align}\label{eq:bundle_quotient_section_def}
    \rE: \IE\to E \quad \textup{such that}\quad \QE\circ\rE = \id_{\IE} \,,
\end{align}
which we again do not demand to be continuous.
However, for convenience we demand the representatives of bundle orbits to lie above the representatives $\rM(\IM)$ in the base space, that is, to satisfy
\begin{align}\label{eq:bundle_quotient_section_isom}
    \piE \circ \rE \ =\ \rM \circ \piIE
\end{align}
as shown in the commutative diagram below:
\begin{equation}
\begin{tikzcd}[column sep=50pt, row sep=30, font=\normalsize]
    \IE     \arrow[d, "\piIE"']
            \arrow[r, "\rE"]
    &
    E       \arrow[d, "\piE"]
    \\
    \IM     \arrow[r, "\rM"']
    &
    M
\end{tikzcd}
\end{equation}

The stabilizer subgroup of a bundle element $e\in E$ is defined as
\begin{align}
    \Stab{e} \,:=\, \big\{ \xi \in \I \;\big|\; \dxiE e=e \big\} \ \leq\ \Stab{\piE(e)}\ \leq\ \I \,.
\end{align}
It is necessarily a subgroup of the stabilizer subgroup $\Stab{\piE(e)}$ of the point $\piE(e)$ in the base space, which is easily seen by
$ \ \xi \in \Stab{e}                        \ \Leftrightarrow\ 
    \dxiE e = e                             \ \Rightarrow\ 
    \piE(\dxiE e) = \xi\, \piE(e) = \piE(e) \ \Leftrightarrow\ 
    \xi \in \Stab{\piE(e)} .
$
As before, the relation $\I.e \cong \I/\Stab{e}$ holds.

We extend our example from Fig.~\ref{fig:isom_egg_quotient_M} by considering the action of $\I\cong\O2$ on the tangent bundle $\TM$ of the egg $M$ in Fig.~\ref{fig:isom_egg_quotient_TM}.
The orbit (violet) of a non-zero vector $0\neq w\in T_nM$ (red) at the north pole~$n$ describes a circle in $T_nM$.
This is consistent with $\I.w \cong \I/\Stab{w} \cong \O2/\Flip \cong S^1$ since such a vector is stabilized by reflections $\Stab{w} \cong \Flip$ along its axis.
The orbit of $0\in T_nM$ is a single point in $\TM$, which is stabilized by any isometry.
Any other vector $v\in\TpM$ (red), living in a tangent space at a point $p\in M$ different from the poles, is by the action of the isometry group rotated and reflected to other tangent spaces $\TphipM$ on the orbit $\I.p$ of $p$.
The orbit $\I.v$ (orange) of any such vector, if not pointing exactly to the north or south, is given by an eastward and a westward pointing copy of the vector in each of the tangent spaces over $\I.p$.
We have $\Stab{v}=\{e\}$ for such vectors and indeed the orbit $\I.v \cong \I/\Stab{v} \cong \O2/\{e\}$ is homeomorphic to $\O2$ (or two circles).
Vectors $v'\in\TpM$ which do point exactly north- or southwards are stabilized by reflections over the axis which they define, that is, $\Stab{v'} \cong \Flip$.
Their orbit is homeomorphic to a circle $\I.v' \cong \I/\Stab{v'} \cong \O2/\Flip \cong S^1$.

The quotient map $\QTM: \TM \to \ITM$ projects the tangent bundle to the bundle quotient $\ITM$, shown in the right half of Fig.~\ref{fig:isom_egg_quotient_TM}.
To understand its structure, we consider all qualitatively different cases:
Firstly, note that the orbits of vectors at the poles correspond to circles of a certain radius, such that the set of such orbits forms a line $\piIE^{-1}(\I.n) \cong \R^+$ (pink ray under the black arrow).
Similarly, the orbits of vectors at any other point $p\in M$ intersect all tangent spaces $\TphipM$ over $\I.p$ in two reflections and therefore form a half plane $\piIE^{-1}(\I.p) \cong \R\times\R^+$ (orange).
The section $\rTM: \ITM \to \TM$ sends each bundle quotient element to some representative in $\TM$.
By the requirement in Eq.~\eqref{eq:bundle_quotient_section_isom}, these representatives are required to lie in the same fiber over the representatives $\rM(\IM)$ of the manifold quotient $\IM$, shown as the black line.
For instance, $v\in\TpM$ (red) is by the quotient map sent to $\QTM(v)\in \ITM$ (black).
The section represents $\QTM(v)$ by $\rTM\QTM(v)$ (also black), which is an element of $T_{\rM\QM(p)}M$ and does in general differ from $v$.

\subsubsection{Quotient representative kernel fields and stabilizer constraints}
\label{sec:quotient_kernels_stabilizers}

\begin{SCfigure}
    \centering
    \includegraphics[width=.51\columnwidth]{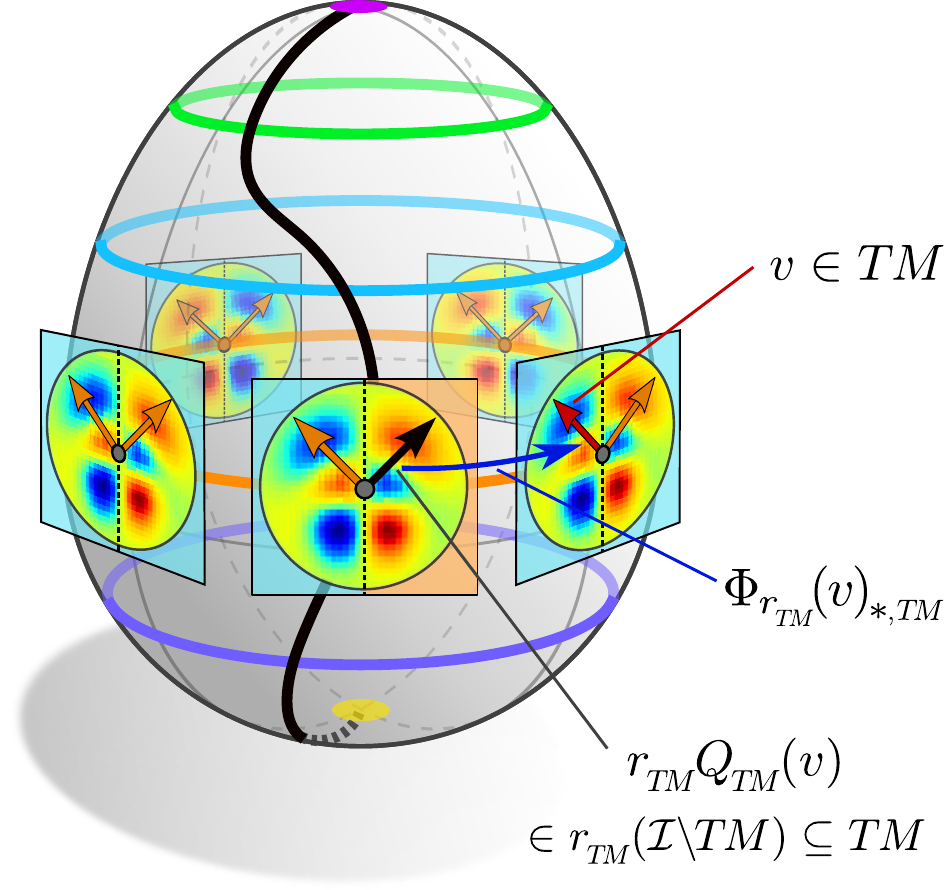}
    \captionsetup{width=1.\textwidth}
    \hfill
    \caption{\small
        Visualization of an isometry invariant kernel field, Def.~\ref{dfn:isometry_invariant_kernel_fields}, and its full reconstruction from kernels on quotient representatives only.
        In contrast to Fig~\ref{fig:isom_invariant_kernel_field_multiple_orbits}, we assume here an isometry group $\I = \O2$ instead of $\SO2$.
        The visualized kernels therefore have a reflectional symmetry, which is enforced by the stabilizer subgroups $\Stab{p} \!\cong\! \Flip$ of points on the orbit $\I.\piTM(v)$.
        Due to its symmetry, the full kernel field $\K: \TM\to \Hom(\Ain,\Aout)$ can be reconstructed from its restriction to the bundle quotient representative $\rTM(\ITM) \subseteq \TM$; see Theorem~\ref{thm:tangent_quotient_repr_kernel_fields}.
        For instance, the shown kernels are fully determined by the partial kernel on the orange half space.
        The reconstruction at $v\in \TM$ is done by evaluating the quotient representative kernel at $\rTM\QTM(v) \in \rTM(\ITM)$ and pushing the kernel via the reconstruction isometry $\PhirNoArg(v) \in \I$, defined in Eq.~\eqref{eq:reconstruction_isometry}, back to $v$.
        We want to mention that the visualized \emph{anti}symmetric kernels would result when mapping between feature fields of even and odd parity, while kernels between feature fields of the same parity would be symmetric.
        }
    \label{fig:isom_invariant_kernel_field_quotient}
\end{SCfigure}

To motivate the construction of quotient representative kernel fields and stabilizer constraints, consider the more explicit formulation
\begin{align}\label{eq:isom_invariant_kernel_constraint_explicit}
    \dphiHom \!\circ \Kp \circ \dphiTMinv \,=\ \Kphip \qquad \forall\, p\in M,\ \  \phi\in\I \,.
\end{align}
of the isometry invariance constraint from Def.~\ref{dfn:isometry_invariant_kernel_fields}, which follows by writing out Eq.~\eqref{eq:kernel_constraint_isom_full_1} for any point~${p\in M}$ individually.
This formulation emphasizes that the constraint leads to \emph{shared weights along the manifold orbits} $\I.p \in \IM$ as visualized in Figs.~\ref{fig:isom_invariant_kernel_field_multiple_orbits} and~\ref{fig:isom_invariant_kernel_field_quotient}.
It implies that the kernel $\Kr$ at an \emph{arbitrary representative point} $r = \rM(o)$ of any orbit $o = \I.r$ fully specifies the kernel field on the rest of the orbit, i.e. at all points $\phi(r)$ where $\phi\in\I$.
The kernel $\Kr$ at the representative point $r$ is itself constrained by the stabilizer subgroup of~$r$:
\begin{align}\label{eq:stab_constraint_teaser}
    \dxiHom \!\circ \Kr \circ \dxiTMinv \,=\ \Kr \qquad \forall\, \xi\in\Stab{r} \,.
\end{align}
This implies that any isometry invariant kernel field can be parameterized in terms of a field of kernels on manifold orbit representatives $r \in \rM(\IM)$ which satisfy Eq.~\eqref{eq:stab_constraint_teaser}.

In case that the stabilizer subgroup at $r$ happens to be non-trivial, the stabilizer constraint in Eq.~\eqref{eq:stab_constraint_teaser} implies further symmetries of the kernel $\Kr$ at~$r$ itself.
For instance, in the example in Fig.~\ref{fig:isom_invariant_kernel_field_quotient} one has the stabilizer subgroup ${\Stab{r} \cong \Flip}$ on the highlighted orbit, enforcing a reflectional symmetry of the kernels.
Such stabilizer symmetries allows to compress the description of isometry invariant kernel fields further:
it turns out to be sufficient to know the values $\K(w)$ of the kernel field on the tangent bundle quotient representatives $w \in \rTM(\ITM) \subseteq \TM$ only.
In Fig.~\ref{fig:isom_invariant_kernel_field_quotient} this corresponds to knowing the kernel values on the orange highlighted half space, from which the full field on the orbit can be reconstructed by the reflectional and rotational symmetries in $\I\cong\O2$.

Theorem~\ref{thm:tangent_quotient_repr_kernel_fields} below makes the latter claim precise by proving that the space $\KIfull$ of isometry invariant kernel fields is isomorphic to a space $\KIquot$ of kernel fields on tangent bundle orbit representatives $\rTM(\ITM)$.
$\KIquot$~is characterized by maximally reduced constraints and thus encodes the kernel fields in $\KIfull$ in a non-redundant way.
It can therefore be viewed as the distilled degrees of freedom contained in $\KIfull$.
In~Theorem~\ref{thm:manifold_quotient_repr_kernel_fields} we formulate a third isomorphic space $\KIquothat$, which equivalently describes isometry invariant kernel fields in terms of the stabilizer subgroup constrained kernels $\Kr$ from Eq.~\eqref{eq:stab_constraint_teaser}.
While the formulation of isometry invariant kernel fields in terms of $\KIquothat$ involves stronger constraints than that in terms of $\KIquot$, it might be more convenient for implementations, since it describes kernels on full tangent spaces instead of kernels on quotients of tangent spaces.

\paragraph{Reconstruction isometries:}
In order to reconstruct full invariant kernel fields in $\KIfull$ from single kernels on orbit representatives, the representative kernels need to be redistributed over the full manifold by applying the kernel pushforward in Eq.~\eqref{eq:isom_invariant_kernel_constraint_explicit} with $p=r$ fixed to the chosen representative points.
For the kernel reconstruction at some point $q\in M$, this requires some isometry $\phi$ which maps the orbit representative $\rM\QM(q) \in \rM(\IM) \subseteq M$ back to $q\in M$, that is, which satisfies $\phi\big(\rM\QM(q)\big) = q$.
To make this more precise, recall that kernel fields $\K: \TM\to \Hom(\Ain,\Aout)$ are defined as maps with domain $\TM$, encoding the \emph{kernel alignments} in addition to their position.
We therefore need to consider more specific isometries which push tangent bundle orbit representatives $\rTM\QTM(v) \in \rTM(\ITM) \subseteq \TM$ back to vectors $v \in \TM$.
These \emph{reconstruction isometries} are defined by:%
\footnote{
    Since the sections $\rTM$ are in general not continuous, $\PhirNoArg$ can in general not be demanded to be continuous either.
}
\begin{align}\label{eq:reconstruction_isometry}
    \PhirNoArg: \TM \to \I \quad\textup{such that}\quad \dPhirTM{v}\, \rTM\mkern1mu \QTM(v) = v \quad \forall\, v\in \TM
\end{align}
We recommend to consult Fig.~\ref{fig:isom_invariant_kernel_field_quotient} to get an intuition for the reconstruction isometries:
graphically, $\Phir{v}$ is defined as \emph{any} isometry which pushes the black vector $\rTM\QTM(v)$ on the orange orbit $\I.v$ back to the red vector $v$ on the same orbit.
Note that $\PhirNoArg$ is only unique up to the stabilizer subgroups of the orbit representatives since for any $\xi\in \Stab{\rTM\QTM(v)}$ it follows that $\Phir{v} \xi$ satisfies the defining constraint in Eq.~\eqref{eq:reconstruction_isometry} as well:%
\footnote{\label{footnote:ambiguity_reconstruction_isometry}:%
    Furthermore, the defining constraint on $\PhirNoArg$ is fulfilled when \emph{left} multiplying $\Phir{v}$ with any $\zeta\in\Stab{v}$.
    This does, however, not add any new degrees of freedom since $\Stab{v} \cong \Stab{\rTM\QTM(v)}$ and
    $\zeta\,\Phir{v} = \Phir{v} \big[\Phir{v}^{-1} \zeta \Phir{v}\big] =: \Phir{v} \widetilde{\zeta}$
    with $\widetilde{\zeta} \in \Stab{\rTM\QTM(v)}$.
}
$\big[ \Phir{v}\, \xi\, \big]_{\!*,\scalebox{.58}{$T\mkern-1.5muM$}}\mkern2mu \rTM\QTM(v) = \dPhirTM{v}\, \rTM\QTM(v) = v$.
All of the following constructions are shown to be independent from this ambiguity.
The action of reconstruction isometries on the base space~$M$ follows by applying the tangent bundle projection to both sides of the defining constraint in Eq.~\eqref{eq:reconstruction_isometry}:
\begin{alignat}{3}\label{eq:reconstruction_isometry_basespace}
    \qquad\qquad\qquad
    \piTM(v)
    \ &=\ \piTM \dPhirTM{v}\, \rTM\, \QTM(v)
        \qquad\quad && \big( \textup{\small Def. of $\PhirNoArg\,$, Eq.~\eqref{eq:reconstruction_isometry} } \big) \notag \\
    \ &=\ \Phir{v}\, \piTM\, \rTM\, \QTM(v)
        \qquad\quad && \big( \textup{\small Pushforward is a bundle map, Eq.~\eqref{eq:pushfwd_bundle_automorphism} } \big) \notag \\
    \ &=\ \Phir{v}\, \rM\, \piITM\, \QTM(v)
        \qquad\quad && \big( \textup{\small Def. of bundle sections, Eq.~\eqref{eq:bundle_quotient_section_isom} } \big) \notag \\
    \ &=\ \Phir{v}\, \rM\, \QM\,  \piTM(v)
        \qquad\quad && \big( \textup{\small Def. of $\piIE\,$, Eq.~\eqref{cd:QE_IE} } \big)
\end{alignat}
A visual summary of the properties of $\PhirNoArg$, that is, its actions on $\TM$ and $M$, is given in the following commutative diagram
\begin{equation}
    \begin{tikzcd}[row sep=3.5em, column sep=6em]
        \TM
            \arrow[d, "\piTM"']
            \arrow[r, "\PhirNoArg \,\times\ \rTM \mkern-2mu\circ\mkern-2mu \QTM"]
            \arrow[rr, rounded corners, to path={ 
                    -- ([yshift=+5.5ex]\tikztostart.north) 
                    --node[above, pos=.5]{\small$\id_{\TM}$} ([yshift=+5.5ex]\tikztotarget.north) 
                    -- (\tikztotarget.north)
                    }]
        &[10ex]
        \I \times \rTM(\ITM)
            \arrow[r, "\ev"]
            \arrow[d, "\id_{\I} \times \piTM"']
        & \TM
            \arrow[d, "\piTM"]
        \\
          M
            \arrow[rr, rounded corners, to path={ 
                    -- ([yshift=-5.5ex]\tikztostart.south) 
                    --node[below, pos=.5]{\small$\id_M$} ([yshift=-5.5ex]\tikztotarget.south) 
                    -- (\tikztotarget.south)
                    }]
        &[10ex] \I \times \rM(\IM)
            \arrow[r, pos=.52, "\ev"']
        & M
    \end{tikzcd}
    \quad
\end{equation}
where the \emph{evaluation maps} $\ev$ are, overloading the notation, given by $\ev: \I\times  M \to  M,\ (\phi,p) \mapsto \phi(p)$ and $\ev: \I\times \TM \to \TM,\ (\phi,v) \mapsto \dphiTM(v)$, respectively.

\paragraph{Quotient representative kernel fields:}
As argued above, the symmetries which are present in an \mbox{isometry} \mbox{invariant} feature field $\K \in \KIfull$ should allow for its full reconstruction from its restriction ${\Krestr: \rTM(\ITM) \to \rHom(\IHom)}$ to tangent bundle orbit representatives $\rTM(\ITM) \subseteq \TM$.%
\footnote{
    In the following we might abbreviate $\Hom(\Ain,\Aout)$ and $\I\backslash \Hom(\Ain,\Aout)$ with $\Hom$ and $\IHom$, respectively.
}
To~construct a (unique) \emph{lift}~$\Lambda$ which recovers $\K = \Lambda(\Krestr)$ from $\Krestr$, we expand tangent vectors~$v$ in the domain of $\K$ via the reconstruction isometry $\PhirNoArg$ from Eq.~\eqref{eq:reconstruction_isometry} and make use of the invariance (equivariance) of the kernel field in Eq.~\eqref{eq:kernel_constraint_isom_full_1}.
This leads to:
\begin{align}\label{eq:lambda_construction}
    \K(v)
    \ &=\,\ \K\; \dPhirTM{v}\; \rTM\, \QTM (v) \notag \\
    \ &=\,\ \dPhirHom{v}\, \K\; \rTM\, \QTM (v) \notag \\
    \ &=\,\ \dPhirHom{v}\, \Krestr\, \rTM\, \QTM (v) \notag \\
    \ &=:\  \pig[ \Lambda\big(\Krestr\big) \pig] (v)
\end{align}
Note that this construction is well defined despite the ambiguity of $\PhirNoArg$ w.r.t. the right multiplication with elements in $\Stab{\rTM\QTM(v)}$.
This is easily seen by observing that for any $w \in \TM$, any $\xi \in \Stab{w}$ and any $\K \in \KIfull$ one has $\dxiHom \K(w) = \K(\dxiTM w) = \K(w)$, which implies that $\Stab{\K(w)} \geq \Stab{w}$, and thus that the final result does not depend on the particular choice of the ambiguous $\PhirNoArg$.

Since the lift $\Lambda$ recovers invariant kernel fields from their restriction to tangent bundle orbit representatives, it can be viewed as the \emph{inverse} map of the restriction (of invariant kernel fields).
This viewpoint implies that the lift establishes an \emph{isomorphism} $\Lambda: \KIquot \to \KIfull$ between the image of the restriction $\KIquot$, which we still need to characterize, and $\KIfull$:
\begin{equation}
    \begin{tikzcd}[row sep=3.5em, column sep=12.em]
        \KIquot
            \arrow[r, bend left=8, shift left=2pt, "\Lambda"]
        &
        \KIfull
            \arrow[l, bend left=8, shift left=2pt, "\Lambda^{-1} = (\,\cdot\,)|_{\rTM(\ITM)}"]
    \end{tikzcd}
\end{equation}

In order to characterize the space $\KIquot$ which makes $\Lambda$ to an isomorphism, it is sufficient to list the properties of restricted fields $\Q := \Krestr \in \KIquot$ for $\K \in \KIfull$:
\begin{itemize}[leftmargin=0.6cm]

\item[{\rule[2.2pt]{2pt}{2pt}}]
First of all, since $\Lambda^{-1}$ is given by the restriction of the domain to $\rTM(\ITM)$, it is clear that any $\Q \in \KIquot$ is required to be of the form $\Q: \rTM(\ITM) \to \rHom(\IHom)$.

\item[{\rule[2.2pt]{2pt}{2pt}}]
Secondly, the property of kernel fields to be bundle $M$-morphisms translates under the restriction $\Lambda^{-1}$ to the requirement on $\Q$ to satisfy $\piHom\circ\Q(w) = \piTM(w)$ for any $w\in \rTM(\ITM)$

\item[{\rule[2.2pt]{2pt}{2pt}}]
Thirdly, $\Q$ is required to satisfy the \emph{(vector) stabilizer constraint} $\dxiHom \Q(w) = \Q(w)$ for any representative vector $w \mkern-2mu\in\mkern-1mu \rTM(\ITM)$ and any $\xi \mkern-1mu\in\mkern-1mu \Stab{w}$.
This requirement is a residual from the invariance constraint in Eq.~\eqref{eq:kernel_constraint_isom_full_1}, surviving the restriction.
\\
It can be deduced by considering the full constraint $\dphiHom \Q\, \dphiTMinv(w) = \Q(w)$ for any $w \in \rTM(\ITM)$ and any isometry $\phi\in\I$ which additionally satisfies that $\dphiTM(w) \in \rTM(\ITM)$, i.e. that the pushforward $\dphiTM(w)$ stays within the restricted domain of~$\Q$.
Note that $\dphiTM(w) \in \I.w$ and that $\rTM(\ITM)$ intersects each orbit exactly once.
This implies that $\I.w \cap \rTM(\ITM) = \{w\}$ such that $\phi\in\I$ is required to satisfy $\dphiTM(w)=w$, that is, $\phi\in \Stab{w}$.
The claimed (vector) stabilizer constraint follows from these considerations.
\\
For an intuition we refer back to Fig.~\ref{fig:isom_invariant_kernel_field_quotient} where the black representative vector $w = \rTM\QTM(v)$ is stabilized only by the trivial isometry $\xi=\{e\}$, implying that the corresponding value of $\Q$ is unconstrained.
Vectors $w' \in \rTM(\ITM)$ which point exactly north- or southwards, i.e. which lie on the dashed reflection axis, are stabilized by reflections in $\Stab{w'}\cong\Flip$, implying a constraint on the corresponding kernel values.%
\footnote{
    The exact constraint depends on the action $\dxiHom$ on $\Hom(\Ain,\Aout)$, which depends on $\rhoHom$ and thus on $\rhoin$ and $\rhoout$.
    The visualized kernel in Fig.~\eqref{fig:isom_invariant_kernel_field_quotient} would correspond to $\rhoHom$ being the sign-flip (odd parity) representation of the reflection group, which enforces antisymmetric kernels.
    The antisymmetry requires $\Q$ to be constrained to $\Q(w') = -\Q(w') = 0$ for $w'$ on the reflection axis; cf. Table~\ref{tab:reflection_steerable_kernels}
}

\item[{\rule[2.4pt]{2pt}{2pt}}]
As a last requirement, $\Q$ needs to lift to a \emph{smooth} kernel field, that is, $\Lambda(\Q)$ is required to be smooth.
Unfortunately, the smoothness (or even continuity) of $\Lambda(\Q)$ does not automatically follow from the smoothness (continuity) of $\Q$ since $\Lambda$ is defined in terms of $\rTM$ and $\PhirNoArg$, which can in general not be demanded to be smooth (continuous).

\end{itemize}

Before summarizing and proving these claims rigorously in Theorem~\ref{thm:tangent_quotient_repr_kernel_fields} below, we give a visual overview of the relation between $\Q = \Krestr \in \KIquot$ and its lift $\K = \Lambda(\Q) \in \KIfull$ in terms of commutative diagrams
\begin{equation}
    \begin{tikzcd}[row sep=3.5em, column sep=6em]
        \TM
            \arrow[r, "\PhirNoArg \,\times\ \Q \mkern-2mu\circ\mkern-2mu \rTM \mkern-2mu\circ\mkern-2mu \QTM"']
            \arrow[rr, rounded corners, to path={ 
                    -- ([yshift=+3.5ex]\tikztostart.north) 
                    --node[above, pos=.5]{\small$\K = \Lambda(\Q)$} ([yshift=+3.5ex]\tikztotarget.north) 
                    -- (\tikztotarget.north)
                    }]
        &[10ex]
        \I \times \rHom(\IHom)
            \arrow[r, "\ev"']
        & \Hom
    \end{tikzcd}
    \qquad
\end{equation}
and
\begin{equation}
    \begin{tikzcd}[row sep=5em, column sep=7.5em]
        \TM
            \arrow[rd, pos=.6, "\rTM\circ\QTM"]
            \arrow[rrddd, "\piTM"']
            \arrow[rrrr, "\K = \Lambda(\Q)"]
        & &[-7em] &[-7em] &
        \Hom
            \arrow[ld, pos=.6, "\rHom\circ\QHom"']
            \arrow[llddd, "\piHom"]
        \\[-1em]
        & \rTM(\ITM)
            \arrow[rr, "\Q"]
            \arrow[rd, start anchor={[xshift=-1ex]}, "\piTM"']
        & &
        \mkern-3mu
        \rHom(\IHom)
            \arrow[ld, "\piHom"]
        \\
        & &
        \rM(\IM)
        \\
        & &
        M
            \arrow[u, pos=.6, "\rM \!\circ\mkern-1mu \QM"' description]
    \end{tikzcd}
\end{equation}
In the last diagram, the commutativity of the top square follows by inserting the definition of the lift, which yields
$\rHom\QHom \Lambda(\Q) = \rHom\QHom \dPhirHom{v} \Q\, \rTM\QTM = \Q\, \rTM\QTM$.
The commutative of the bottom left and right squares follows from Eqs.~\ref{eq:bundle_quotient_section_isom} and~\ref{eq:quotient_projection_piIE}.

\begin{thm}[Tangent quotient representative kernel fields]
\label{thm:tangent_quotient_repr_kernel_fields}
    The space of isometry invariant kernel fields $\KIfull$ from Def.~\ref{dfn:isometry_invariant_kernel_fields} is isomorphic to the space $\KIquot$ of (vector) stabilizer subgroup constrained kernel fields on tangent bundle quotient representatives, defined as:%
    \footnote{
        This definition of $\KIquot$ is in cyclic dependency with that of $\Lambda$ in Eq.~\eqref{eq:lifting_isomorphism_lambda}.
        This could be avoided on the expense of 1) having to define spaces $\widetilde{\KIquot}$ and $\widetilde{\KIfull}$ without smoothness requirements, in terms of which 2) $\widetilde{\Lambda}: \widetilde{\KIquot} \to \widetilde{\KIfull}$ could be defined, which would 3) allow to demand the smoothness requirements in $\KIquot$ in terms of $\widetilde{\Lambda}$.
    }
    \begin{align}\label{eq:KIquot_def}
        \KIquot :=\,
            \Big\{ \Q \mkern-2mu: \rTM(\ITM) \to \rHom(\IHom) \,\Big|\:& 
            \piHom \mkern-5mu\circ\mkern-1mu \Q = \piTM,\quad
            \Lambda(\Q)\ \textup{smooth}, \\ &
            \dxiHom \Q(w) = \Q(w) \ \ \ \forall\; w \mkern-2mu\in\mkern-1mu \rTM(\ITM),\ \xi \mkern-1mu\in\mkern-1mu \Stab{w}
            \!\Big\} \notag
    \end{align}
    The (unique) \emph{lifting isomorphism} $\Lambda: \KIquot \to \KIfull$ between both spaces is hereby given by
    \begin{align}\label{eq:lifting_isomorphism_lambda}
        \Lambda(\Q): \TM \to \Hom(\Ain,\Aout),\ \ \ 
        v \mapsto \big[ \Lambda(\Q) \big](v) \,:=\, \dPhirHom{v} \Q\; \rTM \QTM(v) \,.
    \end{align}
    Its inverse $\Lambda^{-1}: \KIfull \to \KIquot$ is given by the restriction of invariant kernel fields to the bundle quotient representatives $\rTM(\ITM) \subseteq \TM$:
    \begin{align}\label{eq:lifting_isomorphism_lambda_inv}
        \Lambda^{-1}(\K): \rTM(\ITM) \to \rHom(\IHom),\ \ \ 
        w \mapsto \big[ \Lambda^{-1}(\K) \big](w) \,:=\, \Krestr(w)
    \end{align}
\end{thm}
\begin{proof}
    In order to prove that $\Lambda: \KIquot \to \KIfull$ is an isomorphism, we need to show that
    \textit{1)} $\Lambda^{-1}$ is indeed an inverse of $\Lambda$, that
    \textit{2)} the defining properties of $\KIfull$ and $\KIquot$ are satisfied after lifting and restricting and that
    \textit{3)} the constructions do not depend on arbitrary choices.
    In order to not overload this section, we outsource the full proof to Appendix~\ref{apx:lifting_iso_proof}.
    The individual steps of the proof are listed below:
    \begin{itemize}[leftmargin=1.25cm]
        \item[\it 1\hspace{1pt}a)] $\Lambda \circ \Lambda^{-1} = \id_{\KIfull}$,
            that is, $\Lambda^{-1}$ is a right inverse of $\Lambda$
        \item[\it 1\hspace{1pt}b)] $\Lambda^{-1} \circ \Lambda = \id_{\KIquot}$,
            that is, $\Lambda^{-1}$ is a left inverse of $\Lambda$
        \item[\it 2\hspace{1pt}a)] $\piHom \mkern-5mu\circ\mkern-2mu \Lambda(\Q) = \piTM$ for any $\Q \in \KIquot$,
            that is, the lift $\Lambda(\Q)$ is a bundle $M$-morphism
        \item[\it 2\hspace{1pt}b)] ${\piHom \mkern-5mu\circ\mkern-2mu \Lambda^{-1}(\K) = \piTM}$ for any $\K \in \KIfull$
        \item[\it 2\hspace{1pt}c)] $\dphiHom \Lambda(\Q)\, \dphiTMinv = \Lambda(\Q)\ \ \forall \phi \in \I$,
            that is, $\Lambda(\Q)$ satisfies the full isometry invariance (equivariance) constraint
        \item[\it 2\hspace{1pt}d)] $\dxiHom \big[\Lambda^{-1}(\K)\big](w) = \big[\Lambda^{-1}(\K)\big](w) \ \ \
               \forall\; w \mkern-2mu\in\mkern-1mu \rTM(\ITM),\ \xi \mkern-1mu\in\mkern-1mu \Stab{w}$,\ 
            that is, $\Lambda^{-1}(\K)$ satisfies the stabilizer constraint
        \item[\it 3)] All constructions and proofs are independent from the particular choice of $\PhirNoArg$
    \end{itemize}
    The smoothness of lifted quotient representative kernel fields holds by definition.
\end{proof}
The arbitrariness in the choice of section $\rTM$ allows for different, isomorphic quotient kernel fields, expressed on different bundle quotient representatives.

Instead of maximally restricting the kernel field to bundle orbit representatives in $\rTM(\IM)$, one could choose to restrict the description to $\piTM^{-1}\big( \rM(\IM) \big)$ only, i.e. to \emph{complete tangent spaces} $T_rM$ for any $r\in \rM(\IM)$.
In Fig.~\eqref{fig:isom_invariant_kernel_field_quotient}, this would correspond to modeling the (reflection symmetric) kernel on the full tangent space shown in the front instead of only one half.
The requirements on such restricted kernels can be derived by following the same rationale as before and results in the constraint in Eq.~\eqref{eq:stab_constraint_teaser}.
We obtain a similar theorem to Theorem~\eqref{thm:tangent_quotient_repr_kernel_fields}:

\begin{thm}[Manifold quotient representative kernel fields]
\label{thm:manifold_quotient_repr_kernel_fields}
    The space of isometry invariant kernel fields $\KIfull$ from Def.~\ref{dfn:isometry_invariant_kernel_fields} is isomorphic to the space $\KIquothat$ of (manifold) stabilizer subgroup constrained kernel fields on the tangent spaces over manifold quotient representatives $\rM(\IM)$, defined as:
    \begin{align}\label{eq:KIquothat_def}
        \KIquothat :=\,
            \Big\{ \Qhat \mkern-2mu: \piTM^{-1}\big( \rM(\IM) \big) \to \piHom^{-1}\big( \rM(\IM) \big) \,\Big|\:& 
            \piHom \mkern-5mu\circ\mkern-1mu \Qhat = \piTM,\quad
            \widehat{\Lambda}(\Qhat)\ \textup{smooth}, \\ &
            \mkern-34mu
            \dxiHom \Qhat|_r\, \dxiTM^{-1} = \Qhat|_r \quad \forall\,\ r \mkern-2mu\in\mkern-1mu \rM(\IM),\,\ \xi \mkern-1mu\in\mkern-1mu \Stab{r}
            \!\Big\} \notag
    \end{align}
    The \emph{lifting isomorphism} $\widehat{\Lambda}: \KIquothat \to \KIfull$ is in terms of $\Lambda$ and a restriction defined as
    \begin{align}
        \widehat{\Lambda}\ :=\ \Lambda \circ (\,\cdot\,)\big|_{\rTM(\ITM)}
    \end{align}
    and therefore essentially agrees with $\Lambda$\textup{:}
    \begin{align}\label{eq:lifting_isomorphism_hat}
        \widehat{\Lambda}(\Qhat): \TM \to \Hom(\Ain,\Aout),\ \ \ 
        v \mapsto \big[ \widehat{\Lambda}(\Qhat) \big](v) \,:=\, \dPhirHom{v} \Qhat\; \rTM \QTM(v)
    \end{align}
    Its inverse $\widehat{\Lambda}^{-1}: \KIfull \to \KIquothat$ is given by the restriction of invariant kernel fields to 
    the tangent spaces over manifold quotient representatives:
    $\piTM^{-1}\big(\rM(\IM)\big) \subseteq \TM$:
    \begin{align}
        \widehat{\Lambda}^{-1}(\K): \piTM^{-1}\big(\rM(\IM)\big) \to \piHom^{-1}\big( \rM(\IM) \big),\ \ \ 
        \widehat{w} \mapsto \pig[ \widehat{\Lambda}^{-1}(\K) \pig](\widehat{w}) \,:=\, \K\big|_{\piTM^{-1}(\rM(\IM))}(\widehat{w})
    \end{align}
\end{thm}
\begin{proof}
    The proof is essentially analogous to that of Theorem~\ref{thm:tangent_quotient_repr_kernel_fields} with the slight difference that the stronger constraint
    $\dxiHom \Qhat|_r\, \dxiTM^{-1} = \Qhat|_r \quad \forall\,\ r \mkern-2mu\in\mkern-1mu \rM(\IM),\,\ \xi \mkern-1mu\in\mkern-1mu \Stab{r}$
    is required.
    Since it would not add much in addition to what is presented in Appendix~\ref{apx:lifting_iso_proof}, we omit the proof.
\end{proof}

The following commutative diagram shows the isomorphisms between the three equivalent descriptions of invariant kernel fields:
\begin{equation}
    \begin{tikzcd}[row sep=3.5em, column sep=13.em]
        \KIquot
            \arrow[r, bend left=7, shift left=2pt, "\Omega"]
            \arrow[rr, rounded corners, to path={ 
                    -- ([yshift=6.ex]\tikztostart.north) 
                    --node[above, pos=.49]{\small$\Lambda$} ([yshift=6.ex]\tikztotarget.north) 
                    -- (\tikztotarget.north)
                }]
        &
        \KIquothat
            \arrow[r, bend left=7, shift left=2pt, "\widehat{\Lambda}"]
            \arrow[l, bend left=7, shift left=2pt, "\Omega^{-1} = (\,\cdot\,)\big|_{\rTM(\ITM)}"]
        &
        \KIfull
            \arrow[l, bend left=7, shift left=2pt, "\widehat{\Lambda}^{-1} = (\,\cdot\,)\big|_{\piTM^{-1}(\rM(\IM))}"]
            \arrow[ll, rounded corners, to path={ 
                    -- ([yshift=-6.5ex]\tikztostart.south) 
                    --node[below, pos=.5]{\small$\Lambda^{-1} = (\,\cdot\,)\big|_{\rTM(\ITM)}$} ([yshift=-6.5ex]\tikztotarget.south) 
                    -- (\tikztotarget.south)
                }]
    \end{tikzcd}
\end{equation}

\paragraph{Relation to \emph{GM}-convolutions:}

The difference between $\IsomGM$-equivariant $\GM$-convolutions and general $\IsomGM$-equivariant kernel field transforms via $\IsomGM$-invariant kernel fields is that the former share $G$-steerable kernels over the whole manifold while the latter are only required to share $\Stab{p}$-steerable kernels over orbits $\IsomGM\!.p \in \IsomGMM$.
The requirement to share weights over the whole manifold is not strictly necessary but is -- supported by Occam's razor -- likely to be a good inductive bias in practice.
It can be viewed as an analog to the assumption that the same physical laws apply throughout the whole universe.

Assume now that $M$ is a \emph{homogeneous space} with respect to the action of some isometry group $\I \leq \IsomGM$, that is, for any two points $p,q\in M$ there is at least one isometry $\phi \in \I$ that connects both points, i.e. $q = \phi(p)$.
In this case there is only one single orbit $\I.p$, which is just $M$ itself, and the stabilizers $\Stab{p}$ of all points $p\in M$ coincide up to isomorphism.
The quotient space $\IM$ is a singleton which is represented by a single representative point $r = \rM(\IM)$ in~$M$.
By Theorem~\ref{thm:manifold_quotient_repr_kernel_fields}, the space of $\I$-invariant kernel fields is equivalently expressed by a kernel field on orbit representatives.
Since we have only a single representative point~$r$ for homogeneous spaces, the full isometry invariant kernel field is in this case equivalent to a single kernel on~$\TrM$.
This representative kernel is required to satisfy the stabilizer subgroup constraint in Eq.~\eqref{eq:KIquothat_def}.
Via the lifting isomorphism $\widehat{\Lambda}$ in Eq.~\eqref{eq:lifting_isomorphism_hat}, the representative kernel is shared over the whole manifold.

This sounds very similar to the definition of $\GM$-convolutions, which share a single, $G$-steerability constrained kernel over the whole manifold.
Theorem~\ref{thm:GM_conv_homogeneous_equivalence} below asserts that there is indeed an equivalence between convolutions and equivariant kernel field transforms on homogeneous spaces.
The coordinate free $\Stab{r}$-steerability from the stabilizer constraint thereby translates (non-canonically) to the $H$-steerability of template kernels, where $H \cong \Stab{r}$ with $H\leq G$ is an isomorphic representation of $\Stab{r}$ relative to some coordinatization.
One can view the isometry (sub)group $\I \leq \IsomGM$ as a principal $\Stab{r}$-bundle over $M$, whose (non-canonical) embedding into $\GM$ gives rise to a $H$-(sub)structure $\HM$ of~$\GM$.
The sharing of a $\Stab{r}$-steerable kernel via the lifting isomorphism, which operates per action of $\I$,
then corresponds exactly to the sharing of an $H$-steerable kernel over~$\HM$.
This implies that $\I$-equivariant kernel fields transforms on homogeneous spaces do indeed correspond to some $\HM$-convolution.
\begin{thm}[Equivariance on homogeneous \textit{M} implies convolution]
\label{thm:GM_conv_homogeneous_equivalence}
    Let $M$ be a manifold equipped with a $G$-structure~$\GM$.
    Assume that there is an isometry group $\I \leq \IsomGM$ which acts transitively on $M$, making it a \emph{homogeneous space}.
    Let $r\in M$ be an arbitrary representative point of $M$ and $\Stab{r} \leq \I$ its stabilizer.
    Then
    \begin{itemize}
        \item[\textit{1)}] There exists a $H$-(sub)structure $\HM \subseteq \GM$ on $M$ with:
            \begin{itemize}\setlength\itemsep{1ex}
                \item $H \cong \Stab{r} \leq \I$ is a subgroup of~$G \cap \O{d}$
                \item $\HM$ is an embedding of $\I$ (as principal $\Stab{r}$-bundle $\I \to \I/\Stab{r}$) into $\GM$, which is preserved by $\I$, that is, $\IsomHM = \I$
            \end{itemize}
        \item[\textit{2)}] Any $\I$-equivariant kernel field transform shares a single $H$-steerable kernel over the whole space $M$ and is equivalent to a $\HM$-convolution with that kernel.
    \end{itemize}
    The specific choice of $H$-structure depends on the chosen isomorphism $H \cong \Stab{r}$ but is irrelevant since $\I$-equivariant kernel field transforms can be equivalently expressed in any such choice.
\end{thm}
\begin{proof}
    The proof is found in Appendix~\ref{apx:homogeneous_equivalence_proof}.
\end{proof}
Our definition of isometry equivariant kernel field transforms is on homogeneous spaces essentially equivalent to the \emph{steerable convolutions on homogeneous spaces} as proposed by~\citet{Cohen2018-intertwiners}\cite{Cohen2019-generaltheory}.
The proven equivalence between isometry equivariant kernel field transforms and $\HM$-convolutions on homogeneous spaces therefore asserts that $\HM$-convolutions and steerable convolutions are essentially similar in this case.
However, while steerable convolutions are only defined on homogeneous spaces, $\HM$-convolutions generalize to general Riemannian manifolds.
More details on convolutions on homogeneous spaces are discussed in the related work Appendix~\ref{apx:homogeneous_conv}.

%% file: chapters/P3_intro.tex

\mypart{A literature review on coordinate independent CNNs}
\label{part:literature_review}

The formulation of coordinate independent CNNs in terms of associated $G$-bundles over Riemannian manifolds is quite general and covers a wide range of possible model instantiations.
To substantiate this claim, we review a large body of convolutional models from the literature and explain them from the unifying viewpoint of coordinate independent CNNs.
Most of the papers in the literature do not explicitly formulate their models in terms of $G$-structures and associated $G$-bundles.
The \emph{implicitly assumed} $G$-structures and group representations are therefore \emph{deduced} from the models' weight sharing patterns, kernel symmetries and equivariance properties; see for instance Fig.~\ref{fig:G_structure_R3_no_origin}.
Table~\ref{tab:network_instantiations} summarizes the resulting taxonomy of coordinate independent CNNs.
The following sections discuss the covered models and their properties in detail.

\etocsettocdepth{2}
\etocsettocstyle{}{} 
\localtableofcontents

Section~\ref{sec:instantiations_euclidean} covers $\Aff(G)$ (affine group) equivariant convolutions on Euclidean spaces $\Euc_d$.
They rely on $\Aff(G)$-invariant $G$-structures as shown in Fig.~\ref{fig:G_structures_R2_main}.
The models in Section~\ref{sec:instantiations_euclidean_polar} operate on punctured Euclidean spaces $\Euc_d\backslash\{0\}$; see Figs.~\ref{fig:G_structures_R2_no_origin}, \ref{fig:G_structure_R2_no_origin_logpolar} or~\ref{fig:G_structure_R3_no_origin}.
They are equivariant w.r.t. rotations around the chosen origin $\{0\}$ but are not translation equivariant.
Spherical and icosahedral CNNs are discussed in Section~\ref{sec:instantiations_spherical}.
Most of these models assume the $G$-structures that are visualized in Figs.~\ref{fig:G_structure_S2_1} and~\ref{fig:G_structure_S2_2} and are therefore $\SO3$ or $\SO2$-equivariant, respectively.
Section~\ref{sec:instantiations_mesh} reviews $\GM$-convolutions on general surfaces, which are mostly discretized as meshes.

The next few pages discuss various design choices of coordinate independent CNNs and give a first overview of the models in Table~\ref{tab:network_instantiations}.

\afterpage{ 
\clearpage 
    \begin{table}
        \begin{center}
        \addtolength{\leftskip} {-2cm} 
        \addtolength{\rightskip}{-2cm}
            \vspace*{-6.5ex}
            \input{chapters/tab_instantiations.tex}
            \vspace*{10pt}
            \captionsetup{width=1.06\columnwidth}
            \caption{
                Classification of convolutional networks in the literature from the viewpoint of coordinate independent CNNs.
                Bold lines separate different geometries.
                The affine group equivariant convolutions on Euclidean spaces $\Euc_d$
            (rows 1-26)
                are reviewed in Section~\ref{sec:instantiations_euclidean}.
                Section~\ref{sec:instantiations_euclidean_polar} discusses $\GM$-convolutions on punctured Euclidean spaces
                ${\Euc_d \backslash \{0\}} \cong {S^{d-1} \mkern-4mu\times\! \R^+}$
            (rows 27-30).
                Details on spherical CNNs
            (rows 31-36)
                are found in Section~\ref{sec:instantiations_spherical}.
                The models in
            rows (37-41)
                operate on general surfaces, mostly represented by triangle meshes; see Section~\ref{sec:instantiations_mesh}.
                The last two lines list our M\"obius convolutions from Section~\ref{sec:mobius_conv}.
                $\Trans_d$,~$\Flip$ and $\Scale$ denote the translation, reflection and scaling group, respectively, while $\CN$ and $\DN$ are cyclic and dihedral groups.
                Infinite-dimensional representations are in implementations discretized or sampled.
                For instance, the regular representations of $\SO2$ or $\O2$ are typically approximated by the regular representations of cyclic or dihedral groups $\CN$ or $\DN$.
            }
            \label{tab:network_instantiations}
        \end{center}
    \end{table}
\thispagestyle{empty}
\clearpage 
}

\newpage 

\subsection*{Design choices and overview}
\label{sec:instantiations_taxonomy}

A coordinate independent CNN is \emph{in theory} fully specified by
\begin{itemize}
    \item[1)] a choice of \emph{Riemannian manifold} $(M,\eta)$
    \item[2)] its \emph{$G$-structure} $\GM$,
    \item[3)] a \emph{$G$-compatible connection} which specifies feature transporters $\PAgamma$,
    \item[4)] the \emph{field types} or $G$-representations $\rho$ of each feature space, and
    \item[5)] a choice of $G$-equivariant \emph{nonlinearities}.
\end{itemize}
The \emph{geodesics, exponential} and \emph{logarithmic maps} follow from the canonical Levi-Civita connection on~$M$.%
\footnote{
    It might seem strange to compute geodesics and feature transporters based on potentially different connections.
    When the transporter connection differs from Levi-Civita, this is usually due to the Levi-Civita connection not being $G$-compatible with the chosen $G$-structure when~$G<\O{d}$.
    Some examples are given in the paragraph on $G$-compatible connections below.
}
The \emph{isometry group} $\IsomGM$ w.r.t. which the network is equivariant follows from the metric and the $G$-structure.
All \emph{kernel spaces} $\KG$ are determined by the group representations of the feature spaces between which they map.
\emph{Weight sharing} is performed by placing a $G$-steerable template kernel relative to an arbitrary $G$-frame in $\GpM$ for each point $p\in M$.

In practice, the user is faced with additional design questions, for instance concerning the discretization of the geometry, the encoding of feature fields, numerical algorithms for computing geodesics and transporters,~etc.
This section gives a high level overview of all relevant design choices.
More specific details are found in the following Sections~\ref{sec:instantiations_euclidean}, \ref{sec:instantiations_euclidean_polar}, \ref{sec:instantiations_spherical} and~\ref{sec:instantiations_mesh}.

\paragraph{Discretizations of manifolds and feature fields:}
The implementations differ in their representation of the manifolds and sampling of the feature fields.

Euclidean spaces $\Euc_d$ admit regular pixels grids, for instance $\Z^d$ or the hexagonal grid~\cite{Hoogeboom2018-HEX}.
More generally, locally regular grids are suitable for locally flat manifolds like the M\"obius strip and the icosahedron; see Figs.~\ref{fig:mobius_conv_numerical} and~\ref{fig:ico_cutting}.
Feature fields on Euclidean spaces may furthermore be sampled on a non-regular point cloud.
This is for instance useful when processing atomic environments, where the atom positions serve as sampling locations~\cite{Thomas2018-TFN}.

An important difference between the two approaches is that regular pixel grids are not equivariant w.r.t. continuous translations in $\Trans_d = (\R^d,+)$, but only w.r.t. the subgroup of discrete translations which preserves the grid, for instance~$(\Z^d,+)$.
CNNs on regular grids are furthermore usually applying spatial pooling operations which reduce the models' equivariance even further.
Specifically, given that the pooling operation has a stride of $n$~pixels, it is equivariant w.r.t. translations in $(n\Z^d,+)$.
After $L$ pooling layers in a convolutional network, this implies that the model as a whole is only equivariant w.r.t. translations in $(n^L\Z^d,+)$
-- this issue was empirically investigated in~\cite{azulay2018shift}.
\citet{zhang2019CNNsShiftInvariant} propose to remedy this issue by replacing stride $n$ pooling layers with stride 1 pooling layers (with the same pooling window size), a low-pass filtering, and an $n$-pixel subsampling.
The additional low-pass filtering between the pooling and subsampling operations prevents aliasing effects, which is shown to make the networks sufficiently more stable under translations which are not elements of $(n\Z^d,+)$.

Curved spaces like the 2-sphere $S^2$ do in general not admit regular sampling grids.
A seemingly obvious discretization is in terms of a regular sampling grid in spherical coordinates (Eq.~\eqref{eq:spherical_coords} and Fig.~\ref{fig:spherical_equirectangular_1}), however, as these coordinates are non-isometric, they oversample the signal towards the poles~\cite{zhao2018distortion,tateno2018distortion}.
Approximately uniform sampling grids on $S^2$ are the ``generalized spiral set''~\cite{coors2018spherenet} or the icospherical grid~\cite{jiang2019spherical,kicanaoglu2019gaugeSphere}.
Alternatively, feature fields may be discretized in the spectral domain.
For the sphere, this is done via an expansion in terms of spherical harmonics for scalar fields, spin-weighted spherical harmonics for irrep fields or Wigner D-matrices for general feature fields~\cite{esteves2018zonalSpherical,esteves2020spinweighted,Cohen2018-S2CNN,kondor2018ClebschGordan}.

General surfaces are most commonly represented by triangle meshes; see Section~\ref{sec:surfaces_geom_mesh}.
Feature fields can then be sampled on the mesh vertices, edges or faces~\cite{deGoes2016VectorFieldProcessing}.
A higher resolution of the feature fields can be achieved by encoding them via texture maps~\cite{li2019crossAtlas,huang2019texturenet}.
Alternatively, surfaces may be represented as point clouds~\cite{tatarchenko2018tangent,jin2019NPTCnet}.

\paragraph{\textit{G}-structures \textit{GM} and structure groups \textit{G}:}
The specific choice of $G$-structure to be respected by the network depends on the learning task and the topology of~$M$ (if continuity or smoothness of the convolution is demanded).
In general, $M$ comes equipped with an $\O{d}$-structure, i.e. a bundle of orthonormal reference frames with respect to the given Riemannian metric.
A \emph{lift} to structure groups $G$ with ${\O{d} < G \leq \GL{d}}$ is uniquely determined by $G$-valued gauge transformations of orthonormal frames.
\emph{Reductions} of the structure group to $G < \O{d}$ are, in contrast, not necessarily unique, and encode additional geometric information.
For instance, a reduction to $G=\SO{d}$ requires an orientation on the manifold.%
\footnote{
    For a single, connected manifold, this choice is arbitrary as long as the kernel initialization is symmetric w.r.t. both orientations.
    In this case the network will simply learn reflected kernels for different orientations.
    When considering a dataset consisting of multiple manifolds, their relative orientation is relevant for a correct generalization.
}
The following sections discuss further (mostly implicitly made) choices of $G$-structures found in the literature; see for instance
Figs.~\ref{fig:G_structures_R2_main},
\ref{fig:G_structures_R2_no_origin},
\ref{fig:G_structure_R2_no_origin_O2},
\ref{fig:G_structure_R2_no_origin_logpolar},
\ref{fig:G_structure_R3_no_origin}
\ref{fig:G_structures_S2_main},
or~\ref{fig:G_structures_ico}.
They are either determined by a demand for the equivariance under the isometry group $\IsomGM$, canonically given on the manifold or, specifically for $\{e\}$-structures, algorithmically fixed via some heuristic.
Recall that $\{e\}$-structures are on non-parallelizable manifolds (by definition) necessarily discontinuous.

The most commonly encountered structure groups in the literature are the following:
\begin{itemize}
    \item[{\rule[2.0pt]{2pt}{2pt}}]
        \emph{trivial} group $\{e\}$, corresponding to non-coordinate independent CNNs with unconstrained kernels
    \item[{\rule[2.0pt]{2pt}{2pt}}]
        \emph{reflection} group $\Flip \cong \Z/2\Z$, flipping the first frame axis
    \item[{\rule[2.0pt]{2pt}{2pt}}]
        \emph{special orthogonal} groups $\SO{d}$
        \ \ (continuous rotations)
    \item[{\rule[2.0pt]{2pt}{2pt}}]
        \emph{orthogonal} groups $\O{d}$
        \ \ (continuous rotations and reflections)
    \item[{\rule[2.0pt]{2pt}{2pt}}]
        \emph{scaling} group $\Scale \cong (\R^+,*)$,
\end{itemize}
Since the last three groups are continuous Lie groups, they are in numerical implementations sometimes approximated by finite subgroups.
For instance, $\SO2$ and $\O2$ are often modeled by \emph{cyclic} groups $\CN$ or \emph{dihedral} groups~$\DN$, while three-dimensional rotations and reflections in $\O3$ can be approximated by \emph{polyhedral} groups (symmetry groups of Platonic solids, e.g. the icosahedron).
To reduce the complexity of the classification of models in Table~\ref{tab:network_instantiations} we chose to not distinguish between the continuous symmetries and their approximations by finite subgroups.
We will, however, state such approximations in our detailed discussion of the models in the following sections.

\paragraph{\textit{G}-compatible connections:}
All of the models consider either the canonical \emph{Levi-Civita} connection on $M$ or the unique \emph{trivial connection} which is induced by an $\{e\}$-structure.
The choice of connection becomes irrelevant (thus unspecified) for networks which operate solely on \emph{scalar fields}, whose transport is always trivial.

More specifically, all Euclidean CNNs from Section~\ref{sec:instantiations_euclidean} use Levi-Civita transporters, which transport vectors such that they remain parallel in the usual sense on Euclidean spaces $\Euc_d$; see Fig.~\ref{fig:transport_flat}.
This is possible since the Levi-Civita connection is $G$-compatible with the models' $G$-structures (defined in Eq.~\eqref{eq:G_lifted_G_structure_Rd} and visualized in Fig.~\ref{fig:G_structures_R2_main}).%
\footnote{
    In contrast, the Euclidean $\{e\}$-structure in Fig.~\ref{fig:frame_field_automorphism_2} would be incompatible with the Levi-Civita connection on~$\Euc_2$.
}

The models on the punctured Euclidean spaces $\Euc_d \backslash \{0\}$ from Section~\ref{sec:instantiations_euclidean_polar} are either based on $\{e\}$-structures and/or consider scalar fields.
They utilize therefore trivial connections which differ from the canonical Levi-Civita connection on~$\Euc_d \backslash \{0\}$.

All spherical CNNs that rely on the $\SO2$-structure in Fig.~\ref{fig:G_structure_S2_1} (reviewed in Section~\ref{sec:spherical_CNNs_fully_equivariant}) transport features according to the Levi-Civita connection on $S^2$ (Fig.~\ref{fig:transport_sphere}).
Those which operate on the $\{e\}$-structure in Fig.~\ref{fig:G_structure_S2_2} (reviewed in Section~\ref{sec:spherical_CNNs_azimuthal_equivariant}) are again considering a trivial connection since the spherical Levi-Civita connection is incompatible with this $\{e\}$-structure.
The icosahedral CNN with $\C6$-structure, Fig.~\ref{fig:G_structure_ico_3}, transports features according to the $\C6$-compatible icosahedral Levi-Civita connection.%
\footnote{
    Discrete Levi-Civita connections on meshes are discussed in Section~\ref{sec:surfaces_geom_mesh} and~\cite{craneDiscreteDifferentialGeometry2014,craneTrivialConnectionsDiscrete2010}.
}

All CNNs on general surfaces that are listed in rows~(37-39) of Table~\ref{tab:network_instantiations}
assume oriented surfaces that are equipped with an $\SO2$-structure.
They transport features with the $\SO2$-compatible Levi-Civita connection of the surfaces.
The other surface CNNs are based on $\{e\}$-structures and/or operate on scalar fields -- their feature transport is therefore trivial.

Our M\"obius strip convolutions transport features via the Levi-Civita connection, which is compatible with the assumed $\Flip$-structure.

Recall that the Levi-Civita connection is uniquely determined by the metric, and is therefore generally isometry invariant; cf. footnote~\ref{footnote:LeviCivita_isometry_invariance} in Section~\ref{sec:isom_expmap_transport}.
As trivial $\{e\}$-compatible connections are uniquely specified by the $\{e\}$-structure they share its symmetries, that is, they are invariant under the action of~$\IsomeM$.
This implies by Theorem~\ref{thm:isom_equiv_GM_conv} that the $\GM$-convolutions, which are based on these connections, are $\IsomGM$-equivariant.

\paragraph{Transporter pullbacks and alternative projections to $\boldsymbol{\TpM}$:}
The transporter pullback $\Expspf$, defined in Def.~\ref{dfn:Expf_pullback_field} and Eq.~\eqref{eq:transporter_pullback_in_coords}, represents a feature field $f$ in a geodesic parametrization on the tangent space~$\TpM$.
The transportation part of the operation is determined by the $G$-compatible connection.
Geodesics -- and therefore exponential maps $\exp_p: \TpM\to M$ -- have closed form expressions on Euclidean spaces $\Euc_d$ and the sphere~$S^2$.
Specifically, the exponential maps on $\Euc_d$ reduce in Cartesian coordinates to the vector summation in Eq.~\eqref{eq:exp_map_euclidean}, such that Euclidean $\GM$-convolutions reduce to conventional convolutions on~$\R^d$; see Theorem~\ref{thm:Euclidean_GM_conv_is_conventional_conv}.
Geodesics on $S^2$ are well known to be given by the great circles of the sphere.
If the sphere is viewed as being embedded in $\R^3$, the exponential map is explicitly given by Eq.~\eqref{eq:sphere_expmap_explicit}.
The geodesics on general surface meshes are not described by closed form solutions but are computed numerically; see Section~\ref{sec:surfaces_geom_mesh}.
In contrast to the smooth setting, one needs to distinguishes between ``shortest'' and ``straightest'' geodesics on meshes~\cite{polthier1998straightest}.

The pullback of feature fields into geodesic normal coordinates is not the only way of representing feature fields on the tangent spaces.
In the literature on spherical CNNs it is rather common to use gnomonic projections, which are visualized in Fig.~\ref{fig:gnomonic_proj}.
Theorem~\ref{thm:gnomonic} shows that this projection can be viewed as a special case of our more general geodesic parameterization after applying a radial warp to the kernels.
The corresponding models are therefore exactly identified as $\GM$-convolutions.
Surfaces which are embedded in an ambient space like $\R^3$ might furthermore rely on various projections in the embedding space; see for instance the last three models that are discussed in Section~\ref{sec:e_surface_conv}.
Note that these approaches are truly different from ours, i.e. these three models are not exactly $\GM$-convolutions.

\paragraph{\textit{G}-representations and nonlinearities:}
Almost all models consider either of the trivial representation, irreducible representations or regular representations as field types.
Exceptions are quotient representations, more general induced representations, tensor product representations and, specifically for $G=\SO3$, the quaternion representation.
Infinite-dimensional representations, in particular regular and quotient representations of Lie groups, are in implementations discretized.
This can either happen via Monte Carlo sampling or by falling back to the corresponding representations of finite subgroups as discussed above.

The nonlinearities are required to be equivariant w.r.t. the action of the chosen $G$-representations.
Since scalar fields are $G$-invariant, they are acted on by usual nonlinearities like ReLU.
Feature fields that transform according to permutation representations, most importantly regular representations, are acted on channel-wise.
All other field types require custom-tailored nonlinearities -- we refer to~\cite{Weiler2019_E2CNN} for a discussion of specific choices.

\paragraph{\textit{G}-steerable kernel spaces:}
$\GM$-convolutions map input fields of type $\rhoin$ to output fields of type~$\rhoout$ by convolving them with $G$-steerable kernels~$K\in\KG$.
Since the space $\KG$ of $G$-steerable kernels, Def.~\ref{dfn:G-steerable_kernel_def_43}, is a vector space, it is usually parameterized in terms of a basis $\{K_1,\,\dots,\,K_N\}$ of~$\KG$.
\mbox{Before} computing the convolution, the learned kernel $K = \sum_{i=1}^N w_i K_i$ is expanded in this basis, where $\{w_1,\dots,w_N\}$ are real-valued weights to be optimized.
Provably complete kernel spaces for the groups $G\leq\O2$ were implemented in~\cite{Weiler2019_E2CNN}.%
\footnote{\url{https://quva-lab.github.io/e2cnn/api/e2cnn.kernels.html}}
A generalization of the Wigner-Eckart theorem characterizes the kernel space bases for general compact structure groups~$G$~\cite{lang2020WignerEckart}.

In practice, the majority of authors does not use a representation theoretic formulation of feature fields and steerable kernels, but formulate them based on intuition.
Specifically, most authors assume a given input field type and propose various convolution operations which are engineered such that the resulting output field transforms in an equivariant (or coordinate independent) manner.%
\footnote{
    This is opposed to our approach, which fixes the input and output fields and subsequently asks for the resulting constraint on convolution kernels.
}
While these approaches propose certain $G$-steerable kernels that map between $\rhoin$-fields and $\rhoout$-fields, these kernels do sometimes not span the complete space of possible kernels.
This applies for instance to the \emph{MDGCNNs} and \emph{PFCNNs}, which are discussed in Section~\ref{sec:so2_surface_conv}.

%% file: chapters/tab_instantiations.tex

\setcounter{magicrownumbers}{0}
\small
\renewcommand\arraystretch{1.25}
\setlength{\aboverulesep}{0pt}
\setlength{\belowrulesep}{0pt}
\rowcolors{2}{gray!12.5}{white}
\begin{tabular}{>{\tiny\color{gray}}rccclc}
    \toprule
    & manifold      & structure group           & global symmetry           & representation    & citation \\
    & $M$           & $G$                       & $\AffGM$ or $\IsomGM$     & $\rho$            &          \\
    \bottomrule
\rownumber&
    $\Euc_d$        & $\{e\}$                   & $\Trans_d$                & trivial               & \cite{LeCun1990CNNs,
                                                                                                          zhang2019CNNsShiftInvariant}
                                                                                                    + any conventional CNN \\
    \cmidrule(lr){2-6}
    \cmidrule(lr){2-6}
\rownumber&
    $\Euc_1$        & $\Scale$                  & $\Trans_1 \rtimes \Scale$ & regular         & \cite{romero2020wavelet} \\
    \cmidrule(lr){2-6}
    \cmidrule(lr){2-6}
\rownumber&
                    & $\Flip$                   & $\Trans_2 \rtimes \Flip$  & regular           & \cite{Weiler2019_E2CNN} \\
    \cmidrule(lr){3-6}
    \cmidrule(lr){3-6}
\rownumber&
                    &                           &                           & irreps            & \cite{Worrall2017-HNET,
                                                                                                    Weiler2019_E2CNN,
                                                                                                    walters2020trajectory} \\
\rownumber&
                    &                           &                           & regular           & 
                                                                                            \makecell{
                                                                                              \cite{Dieleman2016-CYC,
                                                                                                    Cohen2016-GCNN,
                                                                                                    zhou2017oriented,
                                                                                                    Cohen2017-STEER,
                                                                                                    Weiler2018SFCNN,
                                                                                                    bekkers2018roto,
                                                                                                    Hoogeboom2018-HEX,
                                                                                                    scaife2021RadioGalaxy}
                                                                                              \\
                                                                                              \cite{Weiler2019_E2CNN,
                                                                                                    graham2020dense,
                                                                                                    lafarge2020rototranslation,
                                                                                                    smets2020pde,
                                                                                                    wang2020incorporating,
                                                                                                    romero2020attentive,
                                                                                                    mohamed2020data}
                                                                                              \\
                                                                                              \cite{shen2020PDOeConvs,
                                                                                                    bekkers2020bspline,
                                                                                                    finzi2020generalizing,
                                                                                                    vanderPol2020MDP2,
                                                                                                    gupta2020rotation,
                                                                                                    mondal2020group,
                                                                                                    walters2020trajectory,
                                                                                                    holderrieth2020steerableCNP}
                                                                                              \\
                                                                                              \cite{dey2020groupGANs,
                                                                                                    sifre2012combined,
                                                                                                    bruna2013invariant,
                                                                                                    Sifre2013-GSCAT,
                                                                                                    sifre2014rigid,
                                                                                                    oyallon2015scattering,
                                                                                                    chavan2021rescaling,
                                                                                                    han2021ReDet}
                                                                                              } \\
\rownumber&
                &                           &                           & quotients         & \cite{Cohen2017-STEER,
                                                                                                    Weiler2019_E2CNN} \\
\rownumber&
                &                           &                           & regular$\xrightarrow{\textup{pool}}$trivial
                                                                                            & \cite{Cohen2016-GCNN,
                                                                                                    marcos2016learning,
                                                                                                    Weiler2019_E2CNN} \\
\rownumber&
                & \multirow{-7.5}{*}{$\SO2$}   & \multirow{-7.5}{*}{$\SE2$} & regular$\xrightarrow{\textup{pool}}$vector
                                                                                            & \cite{Marcos2017-VFN,
                                                                                                    Weiler2019_E2CNN} \\
    \cmidrule(lr){3-6}
    \cmidrule(lr){3-6}
\rownumber&
                &                           &                           & trivial           & \cite{khasanova2018isometric,
                                                                                                    Weiler2019_E2CNN} \\
\rownumber&
                &                           &                           & irreps            & \cite{Weiler2019_E2CNN} \\
\rownumber&
                &                           &                           & regular           & 
                                                                                            \makecell{
                                                                                              \cite{Dieleman2016-CYC,
                                                                                                    Cohen2016-GCNN,
                                                                                                    Hoogeboom2018-HEX,
                                                                                                    Cohen2017-STEER,
                                                                                                    Weiler2019_E2CNN}
                                                                                              \\
                                                                                              \cite{mondal2020group,
                                                                                                    graham2020dense,
                                                                                                    shen2020PDOeConvs}
                                                                                              } \\
\rownumber&
                &                           &                           & quotients         & \cite{Cohen2017-STEER} \\
\rownumber&
                &                           &                           & regular$\xrightarrow{\textup{pool}}$trivial    & \cite{Weiler2019_E2CNN} \\
\rownumber&
                & \multirow{-6.2}{*}{$\O2$}   & \multirow{-6.2}{*}{$\E2$} & induced $\SO2$-irreps \hspace*{-2.ex}          & \cite{Weiler2019_E2CNN} \\
    \cmidrule(lr){3-6}
    \cmidrule(lr){3-6}
\rownumber&
\multirow{-15.35}{*}{$\Euc_2$}
                &                           &                           & regular           & \cite{Worrall2019DeepScaleSpaces,
                                                                                                    Sosnovik2020scale,
                                                                                                    bekkers2020bspline,
                                                                                                    zhu2019scale} \\
\rownumber&
                & \multirow{-2}{*}{$\Scale$}& \multirow{-2}{*}{$\Trans_2\rtimes\Scale$}      & regular$\xrightarrow{\textup{pool}}$trivial    & \cite{ghosh2019scale} \\
    \cmidrule(lr){2-6}
    \cmidrule(lr){2-6}
\rownumber&
                &                           &                           & irreps            & \cite{3d_steerableCNNs,
                                                                                                    Thomas2018-TFN,
                                                                                                    miller2020relevance,
                                                                                                    Kondor2018-NBN,
                                                                                                    anderson2019cormorant,
                                                                                                    batzner2021se3equivariant} \\
\rownumber&
                &                           &                           & quaternion        & \cite{zhang2019quaternion} \\
\rownumber&
                &                           &                           & regular           & \cite{finzi2020generalizing,
                                                                                                    winkels3DGCNNsPulmonary2018,
                                                                                                    Worrall2018-CUBENET} \\
\rownumber&
                & \multirow{-4}{*}{$\SO3$}  & \multirow{-4}{*}{$\SE3$}  & regular$\xrightarrow{\textup{pool}}$trivial
                                                                                            & \cite{andrearczyk2019exploring} \\
    \cmidrule(lr){3-6}
    \cmidrule(lr){3-6}
\rownumber&
                &                           &                           & regular           & \cite{winkels3DGCNNsPulmonary2018} \\
\rownumber&
                &                           &                           & quotient $\O3/\O2$ \hspace*{-2ex}
                                                                                            & \cite{janssen2018design} \\
\rownumber&
                & \multirow{-3}{*}{$\O3$}   & \multirow{-3}{*}{$\E3$}   & irrep$\xrightarrow{\textup{norm}}$trivial \hspace*{-2ex}
                                                                                            & \cite{poulenard2019effective} \\
    \cmidrule(lr){3-6}
    \cmidrule(lr){3-6}
\rownumber&
                & $\C4$                     & $\Trans_3 \rtimes \C4$    & regular           & \cite{su2020dv} \\
    \cmidrule(lr){3-6}
    \cmidrule(lr){3-6}
\rownumber&
\multirow{-7}{*}{$\Euc_3$}
                & $\D4$                     & $\Trans_3 \rtimes \D4$    & regular           & \cite{su2020dv} \\
    \cmidrule(lr){2-6}
    \cmidrule(lr){2-6}
\rownumber&
    $\Euc_{d-1,1}$& $\SO{d\minus1,1}$       & $\Trans_d\rtimes\SO{d\minus1,1}$& irreps    & \cite{shutty2020learning} \\

    \bottomrule

\rownumber&
                &                           & $\SO2$                    &                &    \cite{chidester2019rotation,
                                                                                                  finzi2020generalizing} \\
\rownumber&
\multirow{-2}{*}{$\Euc_2\backslash\{0\}$} & \multirow{-2}{*}{$\{e\}$}
                                          & $\SO2 \times \Scale$      & \multirow{-2}{*}{trivial}
                                                                                       &    \cite{esteves2017polar,
                                                                                                  finzi2020generalizing} \\
    \cmidrule(lr){2-6}
    \cmidrule(lr){2-6}
\rownumber&
                    & $\O2$                 & $\O3$                      & trivial           & \cite{ramasinghe2019representation} \\
\rownumber&
\multirow{-2}{*}{$\Euc_3\backslash\{0\}$}
                & $\{e\}$                   & $\{e\}$                    & trivial           & \cite{boomsma2017spherical} \\
    \bottomrule


\rownumber&
                &                           &                           & irreps            & 
                                                                                              \cite{kondor2018ClebschGordan,
                                                                                                    esteves2020spinweighted} \\
\rownumber&
                & \multirow{-2}{*}{$\SO2$}  & \multirow{-2}{*}{$\SO3$}  & regular           & \cite{Cohen2018-S2CNN,
                                                                                                    kicanaoglu2019gaugeSphere} \\
    \cmidrule(lr){3-6}
    \cmidrule(lr){3-6}
\rownumber&
    \multirow{-3}{*}{$S^2$}
                & $\O2$                     & $\O3$                     & trivial           & \cite{esteves2018zonalSpherical,
                                                                                                    perraudin2018DeepSphere,
                                                                                                    yang2020rotation} \\
\cmidrule(lr){2-6}
\cmidrule(lr){2-6}
\rownumber&
    $S^2 \,\backslash\,$poles & $\{e\}$     & $\SO2$                    & trivial           & 
                                                                                            \makecell{
                                                                                              \cite{coors2018spherenet,
                                                                                                    tateno2018distortion,
                                                                                                    zhao2018distortion,
                                                                                                    martin2020panoramic,
                                                                                                    jiang2019spherical}
                                                                                              \\
                                                                                              \cite{
                                                                                                    su2017spherical,
                                                                                                    su2019kernel,
                                                                                                    eder2019convolutions,
                                                                                                    lee2019spherephd}
                                                                                              } \\
\cmidrule(lr){2-6}
\cmidrule(lr){2-6}
\rownumber&
    icosahedron & $\C6$                     & $\operatorname{I}$ {\color{gray}$(\approx\SO3)$} & regular        & \cite{gaugeIco2019} \\
\cmidrule(lr){2-6}
\cmidrule(lr){2-6}
\rownumber&
    ico$\,\backslash\,$poles & $\{e\}$     & $\C5$ {\color{gray}$(\approx\SO2)$}           & trivial           & \cite{zhang2019orientation,
                                                                                                    liu2018icoAltAz} \\
    \bottomrule


\rownumber&
                &                           &                           & irreps            & \cite{Wiersma2020} \\
\rownumber&
                &                           &                           & regular           & \cite{poulenard2018multi,
                                                                                                    sun2018zernet,
                                                                                                    Yang2020parallelFrameCNN,
                                                                                                    deHaan2020meshCNNs} \\
\rownumber&
                & \multirow{-3}{*}{$\SO2$}  & \multirow{-3}{*}{$\IsomplusM$} & regular$\xrightarrow{\textup{pool}}$trivial & 
                                                                                              \cite{masci2015geodesic,
                                                                                                    masci2015shapenet,
                                                                                                    monti2017geometric,
                                                                                                    sun2018zernet} \\
    \cmidrule(lr){3-6}
    \cmidrule(lr){3-6}
\rownumber&
\multirow{-4}{*}{surface ($d\!=\!2$)}
                & $\D4$                     & $\Isom_{\D4\mkern-5muM}$            & trivial           & \cite{huang2019texturenet} \\
    \cmidrule(lr){3-6}
    \cmidrule(lr){3-6}
\rownumber&
\multirow{-4}{*}{(e.g. meshes)}
                & $\{e\}$                   & $\IsomeM$                & trivial           & \cite{
                                                                                                    monti2017geometric,
                                                                                                    schonsheck2018parallel,
                                                                                                    jin2018learning,
                                                                                                    tatarchenko2018tangent,
                                                                                                    li2019crossAtlas} \\
    \bottomrule

\rownumber&
                   &                       &                            & irreps           &                            \\
\rownumber&
    \multirow{-2}{*}{M\"obius strip} & \multirow{-2}{*}{$\Flip$}               & \multirow{-2}{*}{$\SO2$}   & regular          & \multirow{-2}{*}{Section~\ref{sec:mobius_conv}} \\
    \bottomrule

\end{tabular}

%% file: chapters/90_Euclidean_intro.tex

\section{Euclidean coordinate independent CNNs}
\label{sec:instantiations_euclidean}

\begin{figure}
    \centering
    \includegraphics[width=1.\textwidth]{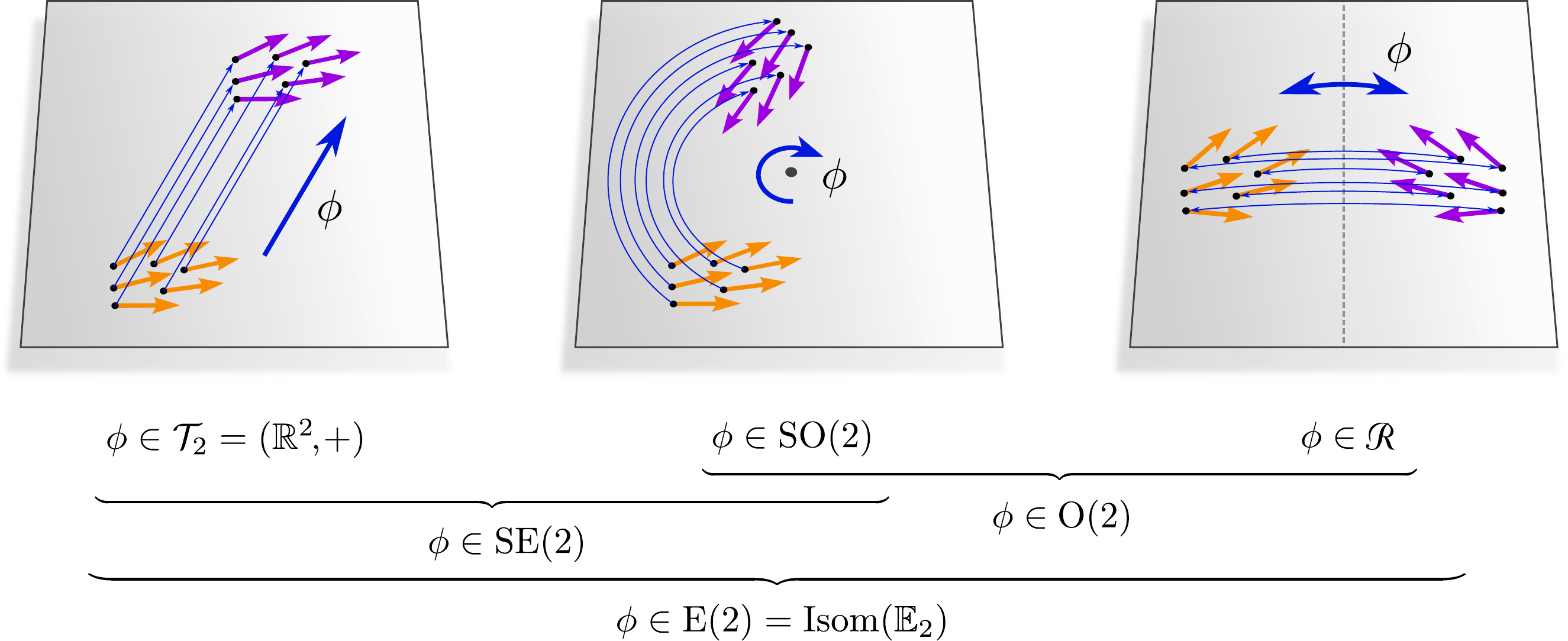}
    \vspace*{.1ex}
    \caption{\small
        Visualization of the full isometry group $\Isom(\Euc_d) = \E{d}$ of Euclidean spaces~$\Euc_d$ for $d=2$.
        It contains subgroups of translations $\Trans_d \mkern-1mu=\! (\R^d,\mkern-2mu+)$, rotations $\SO{d}$ and reflections $\Flip$.
        Rotations and reflections form the orthogonal group ${\O{d} \!=\! \SO{d} \!\rtimes\! \Flip}$ while translations and rotations form the special Euclidean group $\SE{d} = \Trans_d \rtimes \SO{d}$.
     }
    \label{fig:isometries_plane}
\end{figure}

This section considers equivariant convolutions on Euclidean (affine) spaces $M = \Euc_d$, which are undoubtedly of greatest practical relevance.
Convolutional network on Euclidean spaces are applied for analyzing planar and volumetric images, audio signals, videos, physical events in (pseudo-Euclidean) Minkowski spacetime or planar environments in reinforcement learning.
The prototypical convolutional model architecture -- both on Euclidean spaces and in general -- is the conventional CNN on~$\Euc_d$ by \citet{LeCun1990CNNs}.
Its success is to a large extent grounded in its translation equivariance, which allows it to generalize learned inference between different spatial locations.
Motivated by this observation, a lot of effort has been made to generalize the equivariance properties of CNNs to larger \emph{global} symmetry groups of~$\Euc_d$, for instance to the Euclidean isometries in Fig.~\ref{fig:isometries_plane} or more general affine transformations.

Interestingly, most of the \emph{globally} equivariant models in the literature achieve their equivariance by applying some kind of $G$-steerable kernels.
This implies that these models are actually equivariant under \emph{local} gauge transformations as well, despite not explicitly being designed for it.
The reason for this unintentional gauge equivariance is that the global equivariance of the models is usually designed for each layer individually, and therefore applies to the local receptive field of every neuron independently.
Here we explain globally equivariant Euclidean CNNs, listed in rows~(1-26) of Table~\ref{tab:network_instantiations}, from the more general viewpoint of local gauge symmetries and discuss how their global equivariance is induced from their local gauge equivariance.
Theorem~\ref{thm:isom_equiv_GM_conv} asserts thereby that a $\GM$-convolution on $M = \Euc_d$ is $\IsomGM$-equivariant.
However, for the specific case of Euclidean spaces, one can make a stronger statement than mere isometry equivariance:
the convolution with a $G$-steerable kernel on $\Euc_d$ implies the global equivariance of the model under the action of the affine group $\Aff(G) := \Trans_d \rtimes G$, as proven in Theorem~\ref{thm:affine_equivariance_Euclidean_GM_conv} below.
The underlying reason for this result is that the geodesics and Levi-Civita transporters are on Euclidean spaces not only preserved by isometries, but are more generally preserved by any affine transformation.

Convolutions on Euclidean spaces are classically formulated \emph{in coordinates} $\R^d$ of $\Euc_d$.
The advantage of formulating convolutions this way is that $\R^d$ comes with all mathematical structure that is required for the definitions.
However, $\R^d$ is equipped with an excess of structure, for instance a vector space structure (and thus an origin) or its canonical $\{e\}$-structure.
By designing the convolutions to be equivariant, the inference is then (partly) made independent from this structure:
for example, the translation equivariance of convolutions equalizes the particular choice of origin while $\E{d}$ (isometry) equivariance equalizes the particular direction and orientation of the $\{e\}$-structure.
To clarify which mathematical structure is actually being assumed and required, we develop an alternative viewpoint:
we start out with the pure affine and metric structure of the Euclidean space~$\Euc_d$.
If~a $\GM$-convolution on $\Euc_d$ assumes more geometric structure, this structure will subsequently be added by specifying an atlases of (affine) charts $x^A: \Euc_d \to \R^d$, which induce gauges and $G$-structures.

\pagebreak

\etocsettocdepth{3}
\etocsettocstyle{}{} 
\localtableofcontents

To give an overview and a simple introduction, we review the classical formulation of $G$-steerable (affine equivariant) convolutions in coordinates $\R^d$ in Section~\ref{sec:steerable_cnns_in_coords}.
The following Sections~\ref{sec:euclidean_geometry} and~\ref{sec:euclidean_affine_equiv} define affine equivariant convolutions on Euclidean spaces in a coordinate free and coordinate independent setting.
Specifically, Section~\ref{sec:euclidean_geometry} discusses the affine geometry of Euclidean spaces.
Atlases of charts with transition functions in an affine group $\Aff(G)$ (Fig.~\ref{fig:affine_charts}) induce hereby the considered $\Aff(G)$-invariant $G$-structures (Fig.~\ref{fig:G_structures_R2_main}).
Section~\ref{sec:euclidean_affine_equiv} considers $\GM$-convolutions on these $G$-structures and proves their global $\Aff(G)$-equivariance.
The specific instantiations of such models found in the literature, i.e. rows~(1-26) of Table~\ref{tab:network_instantiations}, are discussed in Section~\ref{sec:euclidean_literature}.

The reader may choose to jump over the technical definitions in Sections~\ref{sec:euclidean_geometry} and~\ref{sec:euclidean_affine_equiv}, which are not strictly necessary to understand the models in Section~\ref{sec:euclidean_literature}.

%% file: chapters/91_Euclidean_steerable_Rd.tex

\subsection%
    [Classical formulation of \texorpdfstring{$G$}{G}-steerable CNNs on \texorpdfstring{$\R^d$}{R^d}]%
    {Classical formulation of \textit{G}-steerable CNNs on $\fakebold{\R^d}$}
\label{sec:steerable_cnns_in_coords}

In this section we review the usual notion of convolutions (or cross-correlations%
\footnote{
    In deep learning it became common to use the terms ``convolution'' and ``cross-correlation'' synonymously.
})
on $\R^d$. 
When convolving with a $G$-steerable kernel, the convolutions become equivariant under the action of the affine group $\Aff(G)$.
The following Sections~\ref{sec:euclidean_geometry} and~\ref{sec:euclidean_affine_equiv} will identify these operations on $\R^d$ as coordinate expressions of coordinate free $\GM$-convolutions on Euclidean spaces~$\Euc_d$.

\paragraph{Euclidean steerable CNNs:}
Conventional CNNs consider \emph{feature maps} on $\R^d$, which are functions of the form
\begin{align}
    F: \R^d \to \R^c \,.
\end{align}
A convolution (actually a correlation) with a matrix-valued kernel $K: \R^d \to \R^{\cout\times\cin}$ is then defined as the integral transform
\begin{align}
    \Fout(\mathscr{x}) \,:=\, [K * \Fin] (\mathscr{x}) \,:= \int_{\R^d} K(\mathscr{v})\, \Fin(\mathscr{x} + \mathscr{v})\ d\mathscr{v} \,,
\end{align}
which maps an input feature map with $\cin$ channels to an output feature map with $\cout$ channels.
It can be shown that this operation is the most general linear and translation equivariant mapping between feature maps~\cite{Cohen2019-generaltheory}.%
\footnote{
    For full generality, one would actually have to allow for kernels in the distributional sense (generalized functions).
}

Euclidean steerable CNNs~\cite{Cohen2017-STEER,3d_steerableCNNs,Weiler2019_E2CNN} generalize conventional CNNs to convolutional networks that are equivariant under the action of affine groups on feature fields.
The affine groups of $\R^d$ are hereby defined as semidirect products of the form
\begin{align}\label{eq:AffG_def}
    \Aff(G)\ :=\ \Trans_d \rtimes G \,,
\end{align}
where $G\leq\GL{d}$.
Affine groups include the isometries of $\R^d$, visualized in Fig.~\ref{fig:isometries_plane}, as a special case for $G\leq\O{d}$ but allow for more general point groups (structure groups)~$G$, for instance uniform scaling~$\Scale$.
The following equations give an overview of the most common affine groups in the literature (up to discretizations) and alternative ways of writing them (assuming $\IsomGM$ to be determined by $\Aff(G)$-invariant $G$-structures; see below):
\begin{alignat}{5}
    &\Aff(\{e\}) \ &&=\ \Trans_d                \ &&=\ (\R^d,+)\ &&=\ \IsomeM \notag \\
    &\Aff(\Flip) \ &&=\ \Trans_d \rtimes \Flip    &&           \ &&=\ \IsomRM \notag \\
    &\Aff(\SO{d})\ &&=\ \Trans_d \rtimes \SO{d} \ &&=\ \SE{d}  \ &&=\ \IsomSOM\ &&=\ \Isom_+\!(\R^d) \\
    &\Aff(\O{d}) \ &&=\ \Trans_d \rtimes \O{d}  \ &&=\  \E{d}  \ &&=\ \IsomOM \ &&=\ \Isom(\R^d) \notag \\
    &\Aff(\Scale)\ &&=\ \Trans_d \rtimes \Scale \notag
\end{alignat}
The group $\Aff(\GL{d})$ comprises \emph{all} affine transformations of~$\R^d$.
Since affine groups are defined as semidirect products, any of their elements $tg \in \Aff(G)$ can be uniquely decomposed into a translation $t\in \Trans_d$ and a point group element $g\in G$.
Their (canonical) action on $\R^d$ is given~by
\begin{align}
    \Aff(G)\times\R^d \to \R^d, \quad (tg,\, \mathscr{x})\ \mapsto\ g\mkern1mu \mathscr{x} + t \,.
\end{align}
The action of an inverse group element $(tg)^{-1}$ follows to be given by
\begin{align}
    \big( (tg)^{-1},\, \mathscr{x}\big)\ \mapsto\ g^{-1} (\mathscr{x} - t) \,.
\end{align}

A feature field of type $\rho$ on $\R^d$ transforms according to the \emph{induced representation} $\Ind_G^{\Aff(G)} \rho$ of $\rho$ as specified by
\begin{align}\label{eq:induced_rep_affine}
    (tg) \mkern2mu\rhd_\rho F \ :=\ \big[\Ind_G^{\Aff(G)} \rho\big](tg)\, F \ :=\ \rho(g)\, F\, (tg)^{-1} \,,
\end{align}
which can be seen as the analog of the coordinate free action on sections in Eq.~\eqref{eq:pushforward_section_A}.%
\footnote{
    Induced representations act in a similar way on fields as shown in Fig.~\ref{fig:active_TpM_equivariance}.
    In contrast to the transformation in this Figure, induced representations allow additionally for translations (the transformation law in Section~\ref{sec:gauge_conv} is the restriction $\Res_G^{\Aff(G)}\Ind_G^{\Aff(G)}\!\rho$ of the induced representation back to $G$, i.e. the induced representation without translations).
}
A convolution with a $G$-steerable kernel $K \in \KG$ is equivariant w.r.t. these actions on the input and output field, that is,
\begin{align}\label{eq:Euclidean_conv_equiv_in_coords_Rd}
     K *\, \big(tg \,\rhd_{\rhoin}\! \Fin \big)\ =\ tg \,\rhd_{\rhoout} \big( K*\Fin \big) \qquad \forall\ \ tg \,\in\, \Aff(G) \,.
\end{align}
This is easily checked by an explicit computation,
\begin{align}
     \pig[ K * (tg \rhd_{\rhoin}\! \Fin) \pig] (\mathscr{x})
    \ =&\ \pig[ K * \big(\rhoin(g)\, \Fin\, (tg)^{-1}\big) \pig] (\mathscr{x}) \notag \\
    \ =&\ \int_{\R^d} K(\mathscr{v})\; \rhoin(g)\, \Fin \big((tg)^{-1} (\mathscr{x} + \mathscr{v}) \big)\ d\mathscr{v} \notag \\
    \ =&\ \int_{\R^d} K(\mathscr{v})\; \rhoin(g)\, \Fin \big(g^{-1}( \mathscr{x} + \mathscr{v} - t)\big)\ d\mathscr{v} \notag \\
    \ =&\ \int_{\R^d} K(g \tilde{\mathscr{v}})\; \rhoin(g)\, \Fin \big(g^{-1}(\mathscr{x} - t) + \tilde{\mathscr{v}}\big)\ \detg\ d\tilde{\mathscr{v}} \notag \\
    \ =&\ \int_{\R^d} \rhoout(g)\, K(\tilde{\mathscr{v}})\; \Fin \big(g^{-1}(\mathscr{x} - t) + \tilde{\mathscr{v}}\big)\ d\tilde{\mathscr{v}} \notag \\
    \ =&\ \rhoout(g)\, \big[ K * \Fin \big] \big(g^{-1}(\mathscr{x}-t)\big) \notag \\
    \ =&\ tg \rhd_{\rhoout} \big[K * \Fin \big] (\mathscr{x}) \,,
\end{align}
which used the $G$-steerability of $K$ in the fifth step and which holds for any $\mathscr{x} \in \R^d$ and any $tg\in \Aff(G)$.
As~proven in~\cite{3d_steerableCNNs}, such
\emph{$G$-steerable convolutions are the most general $\Aff(G)$-equivariant linear maps between Euclidean feature fields}.%
\footnote{
    Assuming that the feature fields transform according to the induced representation, Eq.~\eqref{eq:induced_rep_affine}, which is required to end up with a convolution.
}%
\footnote{
    This generalizes the well known statement that conventional Euclidean convolutions are the most general translation equivariant linear maps between functions (or feature maps) on Euclidean spaces, which is recovered for $G=\{e\}$.
}

\paragraph{Relation to Euclidean \textit{GM}-convolutions:}
How do these steerable convolutions on $\R^d$ relate to $\GM$-convolutions on Euclidean spaces ${M = \Euc_d}$?
The fact that steerable convolutions rely on $G$-steerable kernels suggests that they are not only globally $\Aff(G)$-equivariant but (more generally) locally $G$-equivariant.
To draw the connection between classical $G$-steerable CNNs on $\R^d$ and our coordinate free $\GM$-convolutions, we need to identify the geometric structure that is implicitly being considered by the former.

In general, $\R^d$ comes canonically equipped with an $\{e\}$-structure, visualized in Fig.~\ref{fig:G_structure_R2_1}.%
\footnote{
    Formally, the canonical $\{e\}$-structure of $\R^d$ arises as follows:
    the vector space $M=\R^d$ comes itself with a canonical basis, given by the basis vectors $e_i \in \R^d$ with elements $(e_i)_j = \delta_{ij}$.
    The canonical reference frames of tangent spaces $T_p\R^d$ follow from this basis via the canonical isomorphisms
    $\iota_{\R^d,p}: T_p{\R^d} \xrightarrow{\sim} \R^d$ from Eq.~\eqref{eq:canonical_iso_TRk_Rk}.
    Intuitively, the local frames of the tangent spaces $T_p\R^d$ are ``aligned'' with the global frame of $\R^d$.
    This is equivalent to introducing the identity map as global coordinate chart $x = \id_{\R^d} : M=\R^d \to \R^d$ and then defining the canonical $\{e\}$-structure as the field of induced coordinate bases $\big[\frac{\partial}{\partial x_i} \big]_{i=1}^d$.
}
It furthermore comes with a Riemannian metric corresponding to the standard inner product of~$\R^d$.%
\footnote{
    This standard metric $\eta$ is defined as the pullback of the standard inner product
    $\langle\cdot,\cdot\rangle_{\R^d}: \R^d\times\R^d \to \R$
    on~$\R^d$ via the canonical isomorphisms
    $\iota_{\R^d,p}: T_p{\R^d} \xrightarrow{\sim} \R^d$ from Eq.~\eqref{eq:canonical_iso_TRk_Rk}
    to the tangent spaces.
    It is thus for any $v,w\in T_p\R^d$ given by
    ${\eta_p(v,w) := \langle \iota_{\R^d,p}(v) ,\, \iota_{\R^d,p}(w) \rangle_{\R^d}}$.
}
The corresponding Levi-Civita connection gives rise to the parallel transporters in Fig.~\ref{fig:transport_flat}, which keep vectors parallel in the usual sense on Euclidean spaces.
When being expressed relative to the frames of the canonical $\{e\}$-structure, the parallel transporters become trivial and drop therefore out.
The exponential maps reduce to a mere summation (after applying some isomorphisms, see below).

While we are given an $\{e\}$-structure on $\R^d$, $\GM$-convolutions rely on less specific $G$-structures.
These $G$-structures could be seen as the (canonical) $G$-\emph{lifts}
\begin{align}\label{eq:G_lifted_G_structure_Rd}
    \GM\ =\ \eM\lhd G\ :=\ \pig\{\, [e_i]_{i=1}^d \lhd g \;\pig|\; [e_i]_{i=1}^d \in \eM,\ g\in G \,\pig\}
\end{align}
of the canonical $\{e\}$-structure~$\eM$ of~$\R^d$.
Intuitively, these lifted $G$-structures are defined by augmenting every canonical reference frame in $\eM$ with any other $G$-related frame (its right $G$-orbit in $\FM$).
Fig.~\ref{fig:G_structures_R2_main} shows such lifted $G$-structures for different structure groups.
As proven in Theorem~\ref{thm:Aff_GM_in_charts} below, they are invariant under the action of $\Aff(G)$ -- which is in Theorem~\ref{thm:affine_equivariance_Euclidean_GM_conv} shown to explain the $\Aff(G)$-equivariance of the convolutions.
They are furthermore $G$-compatible with the Levi-Civita connection.

The claims made here are more rigorously discussed in the following two Sections.
This formalizations is, however, not strictly necessary to understand our classification of Euclidean CNNs in the literature, such that the reader may skip them and jump immediately to Section~\ref{sec:euclidean_literature}.

%% file: chapters/92_Euclidean_affine_geom.tex

\subsection
    [Affine geometry of Euclidean spaces \texorpdfstring{$\Euc_d$}{}]%
    {Affine geometry of Euclidean spaces $\fakebold{\Euc}_{\boldsymbol{d}}$}
\label{sec:euclidean_geometry}

Before discussing coordinate free convolutions on Euclidean spaces, we need to understand the underlying Euclidean geometry.
Euclidean spaces $\Euc_d$ are by definition \emph{affine} spaces, that is, they come with an associated vector space of dimension $d$, which defines \emph{translations} on~$\Euc_d$.
In addition to being affine spaces, Euclidean spaces are endowed with an \emph{Euclidean metric} (distance function).
This distance functions corresponds to a Riemannian metric $\eta$, i.e. an $\O{d}$-structure $\OM$ on the (Riemannian) manifold $M=\Euc_d$.
This metric has the property that its curvature vanishes everywhere, that is, $\Euc_d$ is globally flat.

\begin{figure}
    \centering
    \includegraphics[width=1.\textwidth]{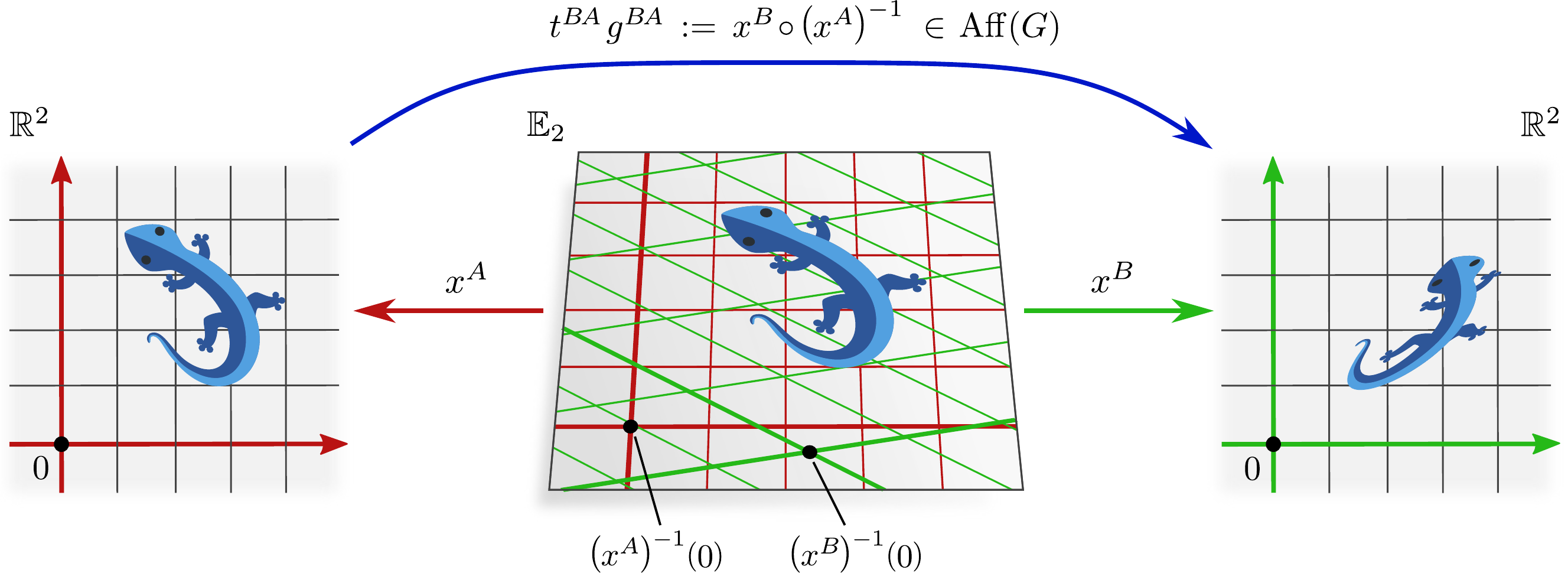}
    \vspace*{2ex}
    \caption{\small
        Visualization of affine charts $x^X: \Euc_d \to \R^d$, which assign global coordinates to Euclidean spaces.
        Both $\Euc_d$ and $\R^d$ are affine spaces, such that one can demand the charts to be affine maps, which preserve collinearity and ratios of distances.
        We define an $\Aff(G)$-atlas $\mathscr{A}^{\Aff(G)}_{\Euc_d}$ as consisting of charts that are related by transition functions $t^{BA} g^{BA} := x^B \circ (x^A)^{-1}$ that are elements in $\Aff(G)$.
        Charts in an $\Aff(G)$-atlas differ at most in their choice of origin $(x^X)^{-1}(0)$ and a $G$-transformation.
        A choice of an $\Aff(G)$-atlas, consisting of charts $x^X$, induces a $G$-atlas $\mathscr{A}^G$ of gauges $\hat{d}x^X$.
        The corresponding $G$-structure $\GM$, which is in Fig.~\ref{fig:G_structures_R2_main} exemplified for different groups $G$, is invariant under the action of $\Aff(G)$.
        Theorem~\ref{thm:affine_equivariance_Euclidean_GM_conv} proves that $\GM$-convolutions on such $G$-structures are $\Aff(G)$-equivariant.
        {\\
        \color{gray}
        \scriptsize
            (Lizards adapted under the Creative Commons Attribution 4.0 International
            \href{https://github.com/twitter/twemoji/blob/gh-pages/LICENSE-GRAPHICS}{\underline{license}}
            by courtesy of Twitter.)
        }
    }
    \label{fig:affine_charts}
\end{figure}

A standard example for Euclidean spaces are the \emph{vector spaces} $\R^d$, however, general Euclidean spaces consider less structure.
In particular, they do not come with a vector space structure and do thus not have a preferred origin.
Furthermore, they are in general not equipped with Cartesian coordinates.
We will therefore start with bare Euclidean spaces $\Euc_d$ and discuss how the relevant geometric structure is added to them.
One could in principle consider any $G$-structure, however, we are specifically interested in those $G$-structures that recover the classical steerable CNNs from the previous section, which explain all models in rows~(1-26) of Table~\ref{tab:network_instantiations}.
Such $\Aff(G)$-invariant $G$-structures are induced from $\Aff(G)$-atlases, consisting of charts of $\Euc_d$ whose transition functions take values in $\Aff(G)$ (Eq.~\eqref{eq:AffG_def}); see Fig.~\ref{fig:affine_charts}.
All statements made in coordinates $\R^d$ can via charts be translated to a coordinate free setting, which we develop here.
More infos about the relation between coordinate charts and gauges can be found in Appendix~\ref{apx:coordinate_bases}.

\subsubsection{Affine charts and Aff(\textit{G})-atlases}
A Euclidean space $\Euc_d$ of dimension $d$ is homeomorphic to $\R^d$, and admits therefore global charts%
\footnote{
    The fact that $\Euc_d$ and $\R^d$ are globally homeomorphic (or even isometric) explains why most related work considers the vector spaces $\R^d$ as models of Euclidean spaces.
    Our approach in this section is a bit more careful as it introduces the minimal structure necessary to define $\GM$-coordinate independent convolutions on Euclidean spaces $M=\Euc_d$.
}
\begin{align}
    x^A: \Euc_d \to \R^d \,.
\end{align}
In the following we will always require these charts to be affine maps, i.e. isomorphisms of affine spaces, which preserve collinearity (i.e. they map straight lines to straight lines) and ratios of distances.
Since compositions of affine maps are affine maps, it follows that the chart transition functions
\begin{align}
    x^B \mkern-1mu\circ\mkern-1mu \big(x^A \big)^{-1}:\ \R^d \to \R^d
\end{align}
are affine transformations of $\R^d$, i.e. elements in $\Aff(\GL{d})$.
The transition functions decompose therefore uniquely into a translation $t^{BA} \in \Trans_d$ and an element $g^{BA} \in \GL{d}$:
\begin{align}
    t^{BA} g^{BA}\ :=\ x^B \mkern-1mu\circ\mkern-1mu \big(x^A \big)^{-1}
\end{align}
The notation $g^{BA}$ is hereby not accidental as these group elements agree with the gauge transformations that are induced by chart transitions, which is proven in Theorem~\ref{thm:AffG_atlas_induced_G_atlas} below.

Given a choice of affine group $\Aff(G)$, we define $\Aff(G)$-atlases of $\Euc_d$ as those atlases of global charts from $\Euc_d$ to $\R^d$, whose chart transition functions take values in $\Aff(G)$:
\begin{dfn}[$\Aff(G)$-atlas of Euclidean space]
\label{dfn:AffG_atlas}
    Let $\mathfrak{X}$ be an index set labeling charts and, for any ${X \mkern-2mu\in \mathfrak{X}}$, let $x^X: \Euc_d \to \R^d$ be a global affine chart of $\Euc_d$.
    The atlas
    \begin{align}
        \mathscr{A}^{\Aff(G)}_{\Euc_d}\ =\ \pig\{ \big( \Euc_d, x^X \big) \;\pig|\ X\in \mathfrak{X} \,\pig\}
    \end{align}
    is said to be an $\Aff(G)$-atlas if all of its chart transition functions take values in $\Aff(G)$, that is, if
    \begin{align}
        x^B \mkern-3mu\circ\mkern-2mu \big(x^A\big)^{-1} \!\in \Aff(G) \quad \forall\ A,B \in \mathfrak{X} \,.
    \end{align}
\end{dfn}
Fig.~\ref{fig:affine_charts} visualizes affine charts and the $\Aff(G)$-valued chart transition maps between them.

\subsubsection{Induced \textit{G}-atlases and \textit{G}-structures}
Any global coordinate chart $x^A: \Euc_d \to \R^d$ induces a global gauge, which is pointwise given by the chart gradients
\begin{align}
    \psiTMp^A := \hat{d}x_p^A :\ \TpM \to \R^d \,,
\end{align}
see Eq.~\eqref{eq:chart_differential_via_gradients} in Appendix~\ref{apx:chart_induced_bases_main} and Table~\ref{tab:coord_charts_gauge_trafos}.
An atlas of charts corresponds therefore an atlas of gauges.
In particular, given that the charts form an $\Aff(G)$-atlas, it is guaranteed that the gauge transformations are $G$-valued, that is, that the induced gauges form a $G$-atlas:
\begin{thm}[$\Aff(G)$-atlases of charts induce $G$-atlases of gauges]
\label{thm:AffG_atlas_induced_G_atlas}
    Let ${\mathscr{A}^{\Aff(G)}_{\Euc_d} \!= \big\{ ( \Euc_d, x^X ) \big| X \mkern-2mu\in\mkern-2mu \mathfrak{X} \big\}}$ be an $\Aff(G)$-atlas of \emph{charts}.
    The induced atlas of \emph{gauges}
    \begin{align}
        \mathscr{A}^G = \pig\{ \big(\Euc_d, \hat{d}x^X \big) \,\pig|\, X \in \mathfrak{X} \pig\}
    \end{align}
    is then guaranteed to be a $G$-atlas.
    In particular, if the chart transition maps are given by ${x^B \circ (x^A)^{-1}} = t^{BA} g^{BA}$, the transition maps between gauges are at any point $p\in \Euc_d$ given by ${g_p^{BA} = g^{BA} \in G}$.
\end{thm}
\begin{proof}
    The transition functions between chart induced gauges coincide by Eq.~\eqref{eq:chart_differential_trafo_law} with the Jacobian of the chart transition maps, that is,
    \begin{align}
        g_p^{BA} \,=\, \hat{d}x^B_p \circ \big(\hat{d}x^A_p \big)^{-1} \,=\, \frac{\partial x^B}{\partial x^A} \Big|_{x^A(p)} \,.
    \end{align}
    The last expression is the usual abuse of notation for Jacobians of chart transition maps, which was in Eq.~\eqref{eq:abuse_of_notation_jacobian} in components defined as
    \begin{align}
        \frac{\partial x^B_\mu}{\partial x^A_\nu} \bigg|_{x^A(p)}
        \,:=\,
        \partial_\nu \pig(x^B_\mu \circ \big(x^A\big)^{-1} \pig) \Big|_{x^A(p)} \,.
    \end{align}
    Using that the chart transition maps are given by $x^B \circ \big(x^A\big)^{-1} = t^{BA} g^{BA}$, this implies
    \begin{align}
        \big( g_p^{BA} \big)_{\mu\nu}
        \ =\ \partial_\nu \pig(x^B_\mu \circ \big(x^A\big)^{-1} \pig) (\mathscr{x}) \big|_{x^A(p)}
        \ =\ \partial_\nu \big( g^{BA} \mathscr{x} + t^{BA} \big)_\mu \big|_{x^A(p)}
        \ =\ g^{BA}_{\mu\nu} \,,
    \end{align}
    that is, that the induced gauge transformations $g_p^{BA}$ are $G$-valued and agree with $g^{BA}$ (which justifies the notation).
    As this argument holds for any $p\in \Euc_d$ and any charts $A,B \in \mathfrak{X}$, this implies that the induced atlas of gauges is a $G$-atlas.
\end{proof}

\begin{figure}
    \centering
    \begin{subfigure}[b]{0.46\textwidth}
        \centering
        \includegraphics[width=.7\textwidth]{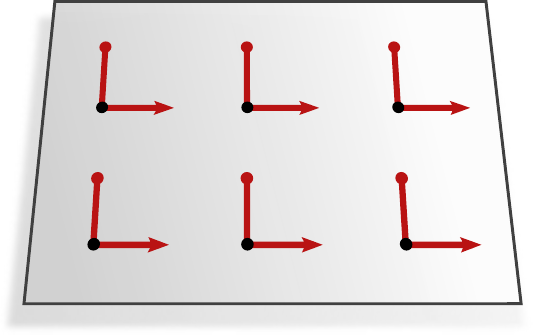}
        \captionsetup{format=hang}
        \caption{\small
            $\Aff(\{e\})$-atlas induced $\{e\}$-structure $\eM$ on ${M=\Euc_2}$, preserved by \emph{translations} in $\Aff(\{e\}) = \IsomeM = \Trans_2$.
        }
        \label{fig:G_structure_R2_1}
    \end{subfigure}
    \hfill
    \begin{subfigure}[b]{0.46\textwidth}
        \centering
        \includegraphics[width=.7\textwidth]{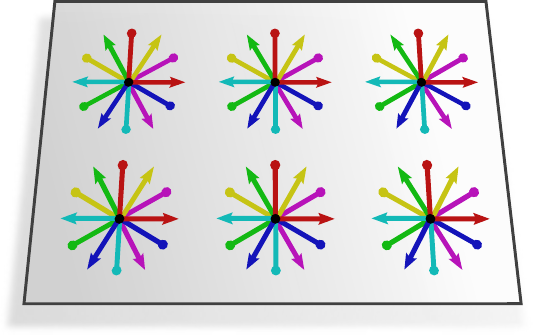}
        \captionsetup{format=hang}
        \caption{\small
            $\Aff(\SO2)$-atlas induced $\SO2$-structure $\SOM$ on $M=\Euc_2$, preserved by \emph{translations} and \emph{rotations} in $\Aff(\SO2) = \IsomSOM = \SE2$.
        }
        \label{fig:G_structure_R2_2}
    \end{subfigure}
    \\[3ex]
    \begin{subfigure}[b]{0.46\textwidth}
        \centering
        \includegraphics[width=.7\textwidth]{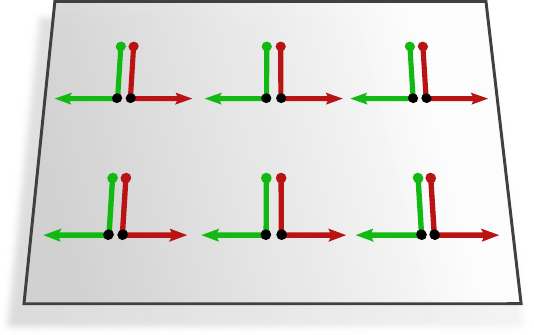}
        \captionsetup{format=hang}
        \caption{\small
            $\Aff(\Flip)$-atlas induced $\Flip$-structure $\RM$ on $M=\Euc_2$, preserved by \emph{translations} and \emph{reflections} in $\Aff(\Flip) = \IsomRM = \Trans_2 \rtimes \Flip$.
            \\
        }
        \label{fig:G_structure_R2_3}
    \end{subfigure}
    \hfill
    \begin{subfigure}[b]{0.46\textwidth}
        \centering
        \includegraphics[width=.7\textwidth]{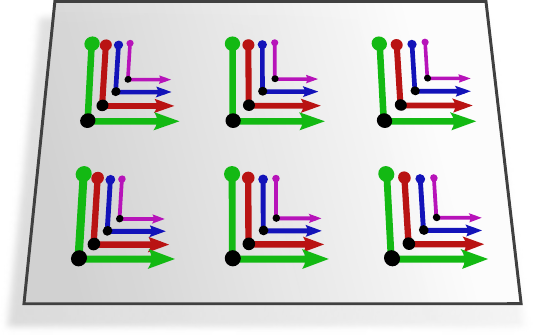}
        \captionsetup{format=hang}
        \caption{\small
            $\Aff(\Scale)$-atlas induced $\Scale$-structure $\SM$ on $M=\Euc_2$, preserved by \emph{translations} and \emph{scalings} in $\Aff(\Scale) = \Trans_2 \rtimes \Scale$.
            Note that $\IsomSM = \Trans_2$ since scalings are non-isometric.
        }
        \label{fig:G_structure_R2_4}
    \end{subfigure}
    \vspace*{0ex}
    \caption{\small
        Visualization of different $G$-structures $\GM$ on Euclidean spaces $M=\Euc_2$ which are induced by an $\Aff(G)$-atlas of charts (Def.~\ref{dfn:AffG_atlas}).
        Fig.~\ref{fig:G_structure_R2_1} shows the translation invariant $\{e\}$-structure $\eM$ that corresponds to conventional Euclidean convolutions.
        The other three $G$-structures correspond to non-trivially $G$-steerable CNNs.
        They generalize locally over all poses that are related by the specific set of reference frames in $\GpM$.
        As the $G$-structures are $\Aff(G)$-invariant (implied by Theorem~\ref{thm:Aff_GM_in_charts}), the $G$-steerable convolutions are globally equivariant w.r.t. $\Aff(G)$ (Theorem~\ref{thm:affine_equivariance_Euclidean_GM_conv}).
        Instead of defining the $G$-structures via an $\Aff(G)$-atlas of charts, one could define them via a $G$-lift $\GM := \eM \lhd G$ of the canonical $\{e\}$-structure of $\R^d$ (Eq.~\eqref{eq:G_lifted_G_structure_Rd}), which augments the frames in the $\{e\}$-structure with all $G$-related frames.
     }
    \label{fig:G_structures_R2_main}
\end{figure}

As discussed in Section~\ref{sec:bundle_trivializations}, any $G$-atlas of gauges implies a $G$-structure $\GM$.
According to Eq.~\eqref{eq:G_atlas_induced_G_structure_GM_def_ptwise}, $\GM$ is pointwise determined by
\begin{align}
    \GpM\ :=\ \big( \psiFMp^A \big)^{-1} (G) \,,
\end{align}
where the particular choice of gauge $A \in \mathfrak{X}$ is arbitrary.
The frames in $\GpM$ are the coordinate bases 
$\big[ \frac{\partial}{\partial x^A_\mu} \big|_p \big]_{\mu=1}^d = \big[\big(\hat{d}x_p^A \big)^{-1} (\epsilon_\mu) \big]_{\mu=1}^d$
and all $G$-transformations of them.
As the maximal $\Aff(G)$-atlas is by definition $\Aff(G)$-invariant, the same holds for the induced $G$-structure (with the action defined via any chart, as clarified and proven below).
Fig.~\ref{fig:G_structures_R2_main} shows such $G$-structures for different affine groups.
In the next section we prove that the corresponding $\GM$-convolutions are equivariant under the action of~$\Aff(G)$.

As it turns out, $\GM \xrightarrow{\piGM} \Euc_d$ is (non-canonically) isomorphic to $\Aff(G) \xrightarrow{\mathscr{q}} \Aff(G)/G \cong \R^d$ as a principal bundle, where
\begin{align}
    \mathscr{q}:\ \Aff(G) \to \R^d,\ \ tg \mapsto t
\end{align}
is the canonical quotient map of the affine group (after identifying cosets $tG$ with translations~$t$).%
\footnote{
    We implicitly employ a canonical isomorphism $\Aff(G)/G \xrightarrow{\sim} \R^d,\ \ tG \mapsto t$, where $t$ denotes a translation group element in $\Trans_d = (\R^d,+)$ on the l.h.s. and a vector in $\R^d$ on the r.h.s.
}
Non-surprisingly, this principal bundle isomorphism depends on the choice of chart.

\begin{thm}[Principal bundle isomorphism between Aff(\textit{G}) and \textit{GM}]
\label{thm:}
    Let $\GM$ be an $\Aff(G)$-atlas induced $G$-structure on $\Euc_d$.
    Then $\GM$ is isomorphic to $\Aff(G) \xrightarrow{\mathscr{q}} \R^d$ as a principal bundle, i.e. there are isomorphisms
    \begin{align}\label{eq:principal_bundle_iso_AffG_GM}
        \alpha^A: \Aff(G) \to \GM,\ \ tg \mapsto \big( \psiGMxAinvt \big)^{-1}(g)
    \end{align}
    and
    \begin{align}
        \big( x^A \big)^{-1}: \R^d \to \Euc_d
    \end{align}
    such that the following diagram commutes:
    \begin{equation}\label{cd:GM_def_embedding}
    \begin{tikzcd}[row sep=2.5em, column sep=7.em]
        \Aff(G) \times G
            \arrow[r, "\alpha^A \times \id_G"]
            \arrow[d, "\cdot\,"']
        & \GM \times G
            \arrow[d, "\,\lhd"]
        \\
        \Aff(G)
            \arrow[r, pos=.5, "\alpha^A"]
            \arrow[d, "\mathscr{q}\,"']
        & \GM
            \arrow[d, "\,\piGM"]
        \\
        \R^d
            \arrow[r, pos=.55, "(x^A)^{-1}"']
        & \Euc_d
    \end{tikzcd}
    \end{equation}
    The inverse of $\alpha^A$ is hereby given by
    \begin{align}
        \big(\alpha^A \big)^{-1}: \GM \to \Aff(G),\ \ [e_i]_{i=1}^d \mapsto tg
        \quad \textup{where}\ \ 
        \begin{cases}
            t = x^A \circ \piGM \big( [e_i]_{i=1}^d \big) \\[1ex]
            g = \psi_{\protect\scalebox{.6}{$G\!M,$}\protect\scalebox{.7}{$\piGM([e_i]_{i=1}^d)$}}^A \big( [e_i]_{i=1}^d \big)
        \end{cases}
    \end{align}
    Note that the isomorphisms are in one-to-one correspondence to the $\Aff(G)$-compatible charts of the considered atlas.
\end{thm}
\begin{proof}
    To prove the statement, we need to show that $\alpha^A$ and $(\alpha^A)^{-1}$ are indeed inverse to each other, that $\alpha^A$ is a bundle map over $(x^A)^{-1}$ and that $\alpha^A$ is right $G$-equivariant.
    That $(\alpha^A)^{-1}$ is both a well defined left and right inverse of $\alpha^A$ is easily checked by the reader.
    That $\alpha^A$ is a bundle map over $(x^A)^{-1}$ means that the bottom square of the diagram commutes.
    This is seen by observing that
    $(x^A)^{-1} \circ \mathscr{q} (tg) = (x^A)^{-1} (t)$ and
    $\piGM \circ \alpha^A (tg) = \piGM \circ \big(\psiGMxAinvt \big)^{-1}(g) = (x^A)^{-1} (t)$
    agree for any $tg \in \Aff(G)$.
    The commutativity of the upper square in the diagram, i.e. the right $G$-equivariance of $\alpha^A$, follows from the fact that
    $\alpha^A (tg \cdot \tilde{g}) = \big(\psiGMxAinvt \big)^{-1}(g \tilde{g}) = \big(\psiGMxAinvt \big)^{-1}(g) \lhd \tilde{g} = \alpha^A(tg) \lhd \tilde{g}$
    holds for any $tg\in \Aff(G)$ and any $\tilde{g} \in G$.
    The second step made use of the fact that $\psiGMxAinvt$ is right $G$-equivariant (Eq.~\eqref{eq:right_equivariance_GM}), which implies the equivariance of its inverse.
    Together, these properties show that $\alpha^A$ is a principal bundle isomorphism.
\end{proof}

\subsubsection{Coordinate free affine transformations}
As we want to prove the equivariance of $\GM$-convolutions under affine transformations in a coordinate free setting, we need to introduce groups of affine transformations of~$\Euc_d$, instead of $\R^d$ as above.
The charts will relate the coordinate free affine groups to the affine groups $\Aff(G)$ of~$\R^d$.

We start with the full group
\begin{align}\label{eq:AffE_def}
    \AffE\ :=\ \big\{ \phi: \Euc_d \to \Euc_d \,\big|\, \textup{$\phi$ is an affine transformation of $\Euc_d$} \big\}
\end{align}
of affine transformations of a Euclidean space $\Euc_d$.
It is easy to prove that $\AffE$ is isomorphic to $\Aff(\GL{d})$, with isomorphisms given by $\phi \mapsto x^A \phi\, (x^A)^{-1}$ for an arbitrary choice of chart $x^A$.
This statement is proven in a more general setting in Theorem~\ref{thm:Aff_GM_in_charts} below.

As in the case of isometries, we define subgroups $\AffGM \leq \AffE$ of $G$-structure preserving affine transformations:
\begin{dfn}[\textit{G}-structure preserving affine transformations]
\label{dfn:AffGM}
    Let $\GM$ be any $G$-structure on $\Euc_d$.
    We define the corresponding subgroup of $G$-structure preserving affine transformations as
    \begin{align}
        \AffGM\ :=\ \big\{ \phi \in \AffE 
        \,\big|\, \dphiFM \GpM = \GphipM\ \ \ \forall p \in \Euc_d \big\} 
        \,\ \leq\ \AffE \,.
    \end{align}
\end{dfn}
Compare this definition to that of $\IsomGM$ in Def.~\ref{dfn:IsomGM}.
As in the case of $\IsomGM$, the gauge transformations that are induced by affine transformations in $\AffGM$ are guaranteed to be $G$-valued.
This statement is formalized by the following theorem, which is essentially analogous to Theorem~\ref{thm:isom_GM_in_coords}:
\begin{thm}[$\AffGM$ in local trivializations]
\label{thm:Aff_GM_in_gauges}
    Let $\phi \in \AffE$ be any isometry of $M=\Euc_d$.
    Then the following three statements are equivalent:
    \begin{enumerate}
        \item $\phi$ is $G$-structure preserving, that is, $\phi \in \AffGM$.
        \item The affine transformation pullback $\psiFMp^{\widetilde{A}}\, \dphiFM^{-1}$ of any gauge $\psiFMp^{\widetilde{A}}$ of the $G$-atlas of $\FM$ that defines~$\GM$ is $G$-compatible with that $G$-atlas.
        \item
        The coordinate expression of $\dphiFM$ relative to any gauges $\psiFMp^{\widetilde{A}}$ and $\psiFMphip^A$ from the $G$-atlas of $\FM$ takes values in the structure group, that is,
        ${g_\phi^{A\widetilde{A}}(p)
           := \psiFMphip^A \,\dphiFM\, \big(\psiFMp^{\widetilde{A}} \big)^{-1}}
           = \hat{d}x^A_{\phi(p)} \,\dphiTM\, \big(\hat{d}x^{\widetilde{A}}_p \big)^{-1}$
        is $G$-valued.
    \end{enumerate}
\end{thm}
\begin{proof}
    The proof is analogous to that of Theorem~\ref{thm:isom_GM_in_coords}.
    More generally, the statement holds for arbitrary $G$-structure preserving \emph{diffeomorphisms}.
\end{proof}

Induced gauge transformations describe the transformation of the coordinate expressions of bundle elements, e.g. tangent or feature vector coefficients.
The action of the affine transformation $\phi$ on the manifold $\Euc_d$ itself can also be described in coordinates~$\R^d$.
This is achieved by a left and right multiplication of $\phi$ with any (affine) chart, which we can w.l.o.g. take to be equal at the source and target location since we are only considering global charts.
The resulting coordinate expression $t_\phi^{AA} g_\phi^{AA}$, defined by the following commutative diagram,
\begin{equation}\label{cd:AffGM_in_chart}
    \begin{tikzcd}[row sep=2.5em, column sep=5em]
        \R^d
            \arrow[rrr, rounded corners, to path={ 
                    |- node[below, pos=.75]{\small$t_\phi^{AA} \mkern1mu g_\phi^{AA}$} ([yshift=-3.ex, xshift=0ex]\tikztotarget.south)
                    -- ([xshift=0ex]\tikztotarget.south)
                    }]
        &
        \Euc_d
            \arrow[l, "x^A"']
            \arrow[r, "\phi"]
        &
        \Euc_d
            \arrow[r, "x^A"]
        &
        \R^c
    \end{tikzcd}
\end{equation}
is guaranteed to take values in $\Aff(G)$ if $\phi$ preserves the $G$-structure.
\begin{thm}[$\AffGM$ in global affine charts]
\label{thm:Aff_GM_in_charts}
    Let $\GM$ be the $G$-structure induced by some $\Aff(G)$-atlas and let $x^A: \Euc_d \to \R^d$ be a chart of this atlas.
    The coordinate expression of an element $\phi \in \AffGM$ relative to $x^A$ is then given by
    \begin{align}\label{eq:AffGM_in_charts_eq}
        x^A \phi\, \big(x^A)^{-1} \,=:\, t_\phi^{AA} \mkern1mu g_\phi^{AA} \ \in\, \Aff(G) \ ,
    \end{align}
    where
    \begin{align}
        t_\phi^{AA} := x^A \,\phi\, \big(x^A \big)^{-1} (0) \ \in\ \Trans_d
        \qquad \textup{and} \qquad
        g_\phi^{AA} := 
        \hat{d}x^A_{\phi(p)} \,\dphiTM\, \big(\hat{d}x^A_p \big)^{-1} \ \in\ G
    \end{align}
    The element $g_\phi^{AA} \in G$ in the coordinate expression coincides hereby with the induced gauge transformation $g_\phi^{AA}(p) \in G$ from Theorem~\ref{thm:Aff_GM_in_gauges} at \emph{any} point $p \in \Euc_d$.

    Furthermore, the coordinatization map
    \begin{align}
        \AffGM \to \Aff(G),\,\ \ \phi \mapsto x^A \,\phi\, \big(x^A \big)^{-1} \,,
    \end{align}
    is a group isomorphism.
\end{thm}
\begin{proof}
    Since $x^A$ and $\phi$ are affine maps, $x^A \,\phi\, \big(x^A)^{-1}: \R^d \to \R^d$ is an affine transformation of $\R^d$, i.e. an element of $\Aff(\GL{d})$ (or some subgroup of it).
    This implies that a first order Taylor expansion of the expression is \emph{exact}.
    The application of the coordinate expression to an arbitrary element $\mathscr{x} \in \R^d$ can therefore be written in terms of the following Taylor expansion around the origin $0$ of $\R^d$:
    \begin{align}
        \big[ x^A \,\phi\, \big(x^A)^{-1} \big] (\mathscr{x})
        \ =&\ \big[ x^A \,\phi\, \big(x^A)^{-1} \big] (0)\ +\ 
            \frac{\partial}{\partial \mathscr{x}'} \big[ x^A \,\phi\, \big(x^A)^{-1} \big]_{\mathscr{x}' = 0} \cdot \mathscr{x} \notag \\
        \ =&\ t_\phi^{AA}\ +\ g_\phi^{AA} \cdot \mathscr{x} \notag \\
        \ =&\ \big( t_\phi^{AA} \mkern2mu g_\phi^{AA} \big) \, \mathscr{x}
    \end{align}
    Here we implicitly defined the translation $t_\phi^{AA} \in \R^d$ and the Jacobian $g_\phi^{AA} \in \R^{d\times d}$ and identified them with group elements, which is possible since all involved morphisms are invertible.

    That the Jacobian $g_\phi^{AA}$ agrees with the induced gauge transformation $g_\phi^{AA}(p)$ at an arbitrary point $p\in M$ is shown by rewriting it via Eq.~\eqref{cd:jacobian_def} in terms of differentials:
    \begin{align}
        g_\phi^{AA}
        \ &=\ \frac{\partial}{\partial \mathscr{x}'} \big[ x^A \,\phi\, \big(x^A)^{-1} \big]_{\mathscr{x}' = x^A(p)} \notag \\
        \ &=\ \iota_{\R^d}\, d\big[ x^A \,\phi\, \big(x^A)^{-1} \big]_{x^A(p)} \, \big(\iota_{\R^d} \big)^{-1} \notag \\
        \ &=\ \iota_{\R^d}\, dx^A_{\phi(p)} \,d\phi_p\, \big(dx^A_p)^{-1}\, \big(\iota_{\R^d} \big)^{-1} \notag \\
        \ &=\ \hat{d}x^A_{\phi(p)} \,\dphiTM\, \big(\hat{d}x^A_p)^{-1} \notag \\
        \ &=\ g_\phi^{AA}(p)
    \end{align}
    In the penultimate step we identified the differential $d\phi$ as an alternative notation for the pushforward $\dphiTM$ and identified the chart gradients $\hat{d}x^A := \iota_{\R^d}\, dx^A$ as defined in Eq.~\eqref{eq:chart_differential_via_gradients}.
    The index $p$ is dropped in the notation $g_\phi^{AA} = g_\phi^{AA}(p)$ due to its arbitrariness.

    That $t_\phi^{AA} g_\phi^{AA}$ is not only an element element of $\Aff(\GL{d})$ but of its subgroup $\Aff(G)$ is clear since Theorem~\ref{thm:Aff_GM_in_gauges} states that $g_\phi^{AA}(p) \in G$ for any $\phi \in \AffGM$.

    To prove that the coordinatization map $C^A: \AffGM \to \Aff(G),\,\ \phi \mapsto x^A \,\phi\, (x^A)^{-1}$ is indeed a group isomorphism, we need to show that it is
    1) a group homomorphism,
    2) injective and
    3) surjective.
    That $C^A$ is a group homomorphism follows immediately from its definition since
    \begin{align}
        C^A(\phi\, \tilde{\phi})
        \ =\ x^A\, \phi\, \tilde{\phi}\, \big(x^A\big)^{-1}
        \ =\ x^A\, \phi\, \big(x^A\big)^{-1}\, x^A\, \tilde{\phi}\, \big(x^A\big)^{-1}
        \ =\ C^A(\phi)\, C^A(\tilde{\phi})
    \end{align}
    holds for any $\phi,\tilde{\phi} \in \AffGM$.
    The injectivity of $C^A$ requires that, for any $\phi,\tilde{\phi} \in \AffGM$, the equality $C^A(\phi) = C^A(\tilde{\phi})$ implies $\phi = \tilde{\phi}$.
    That this is the case is clear since $C^A(\phi) = C^A(\tilde{\phi})$ is equivalent to 
    $x^A\, \phi\, \big(x^A\big)^{-1} = x^A\, \tilde{\phi}\, \big(x^A\big)^{-1}$, which implies the equality of $\phi$ and $\tilde{\phi}$ since $x^A$ is an isomorphism.
    Lastly, $C^A$ is surjective if and only if for any $tg \in \Aff(G)$ there exists some $\phi \in \AffGM$, such that $C^A(\phi) = tg$.
    As an ansatz, let $\phi = \big(x^A\big)^{-1} \,tg\, x^A$, such that $C^A(\phi) = tg$.
    What remains to be shown is that this construction of $\phi$ is indeed an element of $\AffGM$.
    As one can easily check, $g_\phi^{AA} = g \in G$, such that $\phi \in \AffGM$ follows from Theorem~\ref{thm:Aff_GM_in_gauges}, with which surjectivity holds.
    Overall, this proves that $C^A: \AffGM \to \Aff(G)$ is a group isomorphism if $\AffGM$ is induced by an $\Aff(G)$-atlas.
\end{proof}

The isomorphism between $\AffGM$ and $\Aff(G)$ is not unique, as it depends on the particular chart considered.
Different choices are related by the inner automorphisms of $\Aff(G)$ since
\begin{align}
    C^B(\phi) 
    = x^B \,\phi\, (x^B)^{-1}
    = x^B (x^A)^{-1} x^A \,\phi\, (x^A)^{-1} x^A (x^B)^{-1}
    = (t^{BA} g^{BA}) \,C^A(\phi)\, (t^{BA} g^{BA})^{-1}
    \,.
\end{align}
This concludes our analysis of the Euclidean geometry and $\Aff(G)$-invariant $G$-structures that are required for the definition of coordinate free Euclidean convolutions in the next section.

%% file: chapters/93_Euclidean_GM_conv.tex

\subsection
    [Affine group equivariant CNNs on Euclidean spaces \texorpdfstring{$\Euc_d$}{}]%
    {Affine group equivariant CNNs on Euclidean spaces $\fakebold{\Euc}_{\boldsymbol{d}}$}
\label{sec:euclidean_affine_equiv}

We now turn to investigate Euclidean $\GM$-convolutions on $\Aff(G)$-atlas induced $G$-structures.
When being expressed in a chart, these convolutions boil down to classical $G$-steerable convolutions on $\R^d$, as we show next.
Their affine equivariance is in Theorem~\ref{thm:affine_equivariance_Euclidean_GM_conv} proven in a coordinate free setting.

\paragraph{Recovering conventional convolutions on $\pmb{\R^d}$:}

$\GM$-convolutions rely crucially on the transporter pullback $\Expspf$ of feature fields, which in turn depends on parallel transporters and the exponential map.
On Euclidean spaces, these operations take a particularly simple form, which we discuss first.

As stated before, Levi-Civita transporters move tangent vectors such over Euclidean spaces that they remain parallel in the usual sense on Euclidean spaces; see Fig.~\ref{fig:transport_flat}.
Let $x^A: \Euc_d \to \R^d$ be any global chart of an $\Aff(G)$-atlas.
As the induced frame field is ``parallel'', the transporters along \emph{any} path $\gamma$ become trivial when being expressed relative to the induced gauges $\hat{d}x_p^A$:
\begin{alignat}{3}
    g_\gamma^{AA} \,&=\, e&
    \qquad &\textup{for \emph{any} path}\ \gamma
\intertext{
    This implies in particular that the feature vector transporters are in this gauge given by identity maps, i.e.
}
    \rho\big( g_\gamma^{AA} \big) \,&=\, \id_{\R^c}&
    \qquad &\textup{for \emph{any} path}\ \gamma \,.
\end{alignat}
When expressing the exponential map in a chart, it reduces to a summation of the point and vector coordinate expressions:
\begin{align}\label{eq:exp_map_euclidean}
    x^A \big( \exp_p v \big)\ =\ x^A(p) + \hat{d}x_p^A(v)
\end{align}

We furthermore need to express feature fields in coordinates, that is, we pull them via the (global, inverse) chart from $\Euc_d$ back to $\R^d$,
\begin{align}
    F^A\ :=\ f^A \circ \big(x^A \big)^{-1}\, :\,\ \R^d \to \R^c \,,
\end{align}
which is visualized by the following commutative diagram:
\begin{equation}\label{cd:feature_field_chart_pullback_Rd}
    \begin{tikzcd}[row sep=2.5em, column sep=5em]
        \R^d
            \arrow[rr, rounded corners, to path={ 
                    |- node[above, pos=.75]{\small$F^A$} ([yshift=3.ex, xshift=1ex]\tikztotarget.north)
                    -- ([xshift=1ex]\tikztotarget.north)
                    }]
        &
        \Euc_d
            \arrow[l, "x^A"]
            \arrow[r, "f^A"']
        &
        \R^c
    \end{tikzcd}
\end{equation}

With these ingredients at hand, the transporter pullback, Eq.~\eqref{eq:transporter_pullback_in_coords}, of feature fields on Euclidean spaces can in coordinates be expressed as
\begin{align}\label{eq:transporter_pullback_Euclidean}
    \big[\mkern-2mu \Expspf \big]^A (\mathscr{v})
    \ =&\,\ \rho\pig( g^{AA}_{p \,\leftarrow\, \exp_p (\hat{d}x_p^A)^{\shortminus1}(v^A)} \pig) \,
          f^A\, \exp_p \!\pig(\! \big(\hat{d}x_p^A\big)^{\mkern-2mu-1}(\mathscr{v}) \pig) \notag \\
    \ =&\,\ f^A\, \big(x^A\big)^{-1}\, x^A \exp_p \!\pig(\! \big(\hat{d}x_p^A\big)^{\mkern-2mu-1}(\mathscr{v}) \pig) \notag \\
    \ =&\,\ F^A \big( x^A(p) + \mathscr{v} \big) \,.
\end{align}
The coordinate expression of the $\GM$-convolution, Eq.~\eqref{eq:gauge_conv_coord_expression}, becomes therefore
\begin{align}\label{eq:Euclidean_GM_conv_in_coords}
    \fout^A(p)
    \ =\ \big[K \star \fin \big]^A (p)
    \ =\ \int_{\R^d} K(\mathscr{v})\, \big[\mkern-2mu \Expspfin\big]^A (\mathscr{v})\ d\mathscr{v}
    \ =\ \int_{\R^d} K(\mathscr{v})\, \Fin^A\big( x^A(p) + \mathscr{v} \big)\ d\mathscr{v} \,.
\end{align}
This shows that $\GM$-convolutions on Euclidean spaces are conventional convolutions (correlations).
\begin{thm}[$\GM$-convolutions on Euclidean spaces recover convolutions on $\fakebold{\R}^{\boldmath{d}}$]
\label{thm:Euclidean_GM_conv_is_conventional_conv}
    Let $\GM$ be a $G$-structure induced by an $\Aff(G)$-atlas of charts as defined in Section~\ref{sec:euclidean_geometry}.
    When being expressed relative to any global chart $x^A: \Euc_d \to \R^d$ of this $\Aff(G)$-atlas, the $\GM$-convolution takes the form of a conventional convolution (correlation) $*$\,:
    \begin{align}
        \Fout^A(\mathscr{x})
        \ =\ \int_{\R^d} K(\mathscr{v})\, \Fin^A\big( \mathscr{x} + \mathscr{v} \big)\ d\mathscr{v}
        \ =\ \big[ K * \Fin^A \big] (\mathscr{x})
    \end{align}
\end{thm}
\begin{proof}
    The statement follows by evaluating Eq.~\eqref{eq:Euclidean_GM_conv_in_coords} at $p = \big(x^A\big)^{-1}(\mathscr{x})$ and identifying $\Fout^A = \fout^A \circ \big(x^A\big)^{-1}$ on the l.h.s.
\end{proof}

Before proceeding to our proof of the Euclidean $\GM$-convolutions' equivariance in a \emph{coordinate free} setting, we consider its \emph{coordinate independence} -- as we will see, both notions are closely related.
All that is required to demonstrate the coordinate independence is the transformation law of the feature field pullbacks to $\R^d$ via charts.
The transformation law follows directly from transition functions and can from the commutativity of the diagram
\begin{equation}\label{cd:feature_field_chart_pullback_Rd_transitions}
    \begin{tikzcd}[row sep=2.5em, column sep=6em]
        \R^d
            \arrow[rr, "F^A"]
            \arrow[dd, "t^{BA} g^{BA}\ "']
        & &
        \R^c
            \arrow[dd, "\ \rho\big( g^{BA} \big)"]
        \\
        &
        \Euc_d
            \arrow[ul, pos=.4, "x^A"']
            \arrow[dl, pos=.4, "x^B"]
            \arrow[ur, pos=.4, "f^A"]
            \arrow[dr, pos=.4, "f^B"']
        \\
        \R^d
            \arrow[rr, "F^B"']
        & &
        \R^c
    \end{tikzcd}
\end{equation}
be read off to be given by
\begin{align}
    F^B\ =\ \rho\big( g^{BA} \big) \,F^A\, \big( t^{BA} g^{BA} \big)^{-1} \,.
\end{align}
Note that this transformation law is exactly the induced representation $F^B = t^{BA}g^{BA} \rhd_\rho F^A$ as introduced in Eq.~\eqref{eq:induced_rep_affine}.
Leveraging the equivariance of the conventional convolution with $G$-steerable kernels from Eq.~\eqref{eq:Euclidean_conv_equiv_in_coords_Rd}, this implies
\begin{align}\label{eq:Euclidean_conv_coordinate_independence}
    K*\Fin^B
    \ =\ K* \big( t^{BA} g^{BA} \rhd_{\rhoin} \Fin^A \big)
    \ =\ t^{BA} g^{BA} \rhd_{\rhoout} \big( K * \Fin^A \big)
    \ =\ t^{BA} g^{BA} \rhd_{\rhoout} \Fout^A
    \ =\ \Fout^B
\end{align}
The \emph{active $\Aff(G)$-equivariance} of classical $G$-steerable convolutions on $\R^d$ from Section~\ref{sec:steerable_cnns_in_coords} is therefore seen to imply the \emph{passive $\Aff(G)$ coordinate independence} of Euclidean $\GM$-convolutions and vice versa.
The two are two sides of the same coin.
In addition, one can prove the $\Aff(G)$-equivariance of the $\GM$-convolution in the coordinate free setting, which we will do next.

\paragraph{Affine group equivariance}

To prove the affine group equivariance of Euclidean $\GM$-convolutions, we first define the transformation law of coordinate free feature fields $f\in \Gamma(\A)$ under affine transformations $\phi \in \AffGM$ as
\begin{align}\label{eq:affine_action_sections}
    \phi \,\rhd f\ =\ \dphiA\, f\: \phi^{-1} \,,
\end{align}
i.e. as for isometries in Def.~\ref{dfn:isometry_pushforward}.%
\footnote{
    Since the feature vector bundle is defined as a $G$-bundle, i.e. associated to $\GM$, pushforwards can only be defined for the $G$-structure preserving affine transformations in $\AffGM$.
}
The (Levi-Civita) transporter pullback of an affine transformed feature field $\phi \rhd\! f$ is relative to an affine chart $x^A$ given by:
\begin{align}\label{eq:affine_transformed_transporter_pullback}
    &\ \ 
        \big[\mkern-2mu \Expsp (\phi \rhd\! f) \big]^A (\mathscr{v})
    \notag \\[.8ex]
    \ \overset{(1)}{=}&\ \ 
        \big[\mkern-2mu \Expsp (\dphiA f\, \phi^{-1}) \big]^A (\mathscr{v})
    \notag \\[.8ex]
    \ \overset{(2)}{=}&\ \ 
        \underbrace{\rho\big( g^{AA}_{p \,\leftarrow\, \exp_p (\hat{d}x_p^A)^{-1}(\mathscr{v})} \big)}_{=\, \id_{\R^c}} \,
        \psiAp^A\, (\dphiA f\, \phi^{-1})\,
        \exp_p \!\pig(\! \big(\hat{d}x_p^A \big)^{-1}(\mathscr{v}) \pig)
    \notag \\[.8ex]
    \ \overset{(3)}{=}&\ \ 
        \psiAp^A\, \dphiA\,
        \pig[ \big( \psiAphiinvp^A \big)^{-1} \psiAphiinvp^A\pig]\, 
        f\: 
        \pig[ \big(x^A \big)^{-1}\, x^A\pig]\, 
        \phi^{-1} 
        \pig[ \big(x^A \big)^{-1}\, x^A\pig]\, 
        \exp_p \!\pig(\! \big(\hat{d}x_p^A \big)^{-1}(\mathscr{v}) \pig)
    \notag \\[.8ex]
    \ \overset{(4)}{=}&\ \ 
        \pig[\psiAp^A\, \dphiA\, \big(\psiAphiinvp^A \big)^{-1}\pig]
        \pig[\psiAphiinvp^A\, f\, \big(x^A \big)^{-1}\pig]
        \pig[x^A\, \phi^{-1} \big(x^A \big)^{-1}\pig]
        \pig[x^A \exp_p \!\pig(\! \big(\hat{d}x_p^A \big)^{-1}(\mathscr{v}) \pig) \pig]
    \notag \\[.8ex]
    \ \overset{(5)}{=}&\ \ 
        \rho\big( g_\phi^{AA} \big)\, F^A\, \big( t_\phi^{AA} g_\phi^{AA} \big)^{-1} \big(x^A(p) + \mathscr{v} \big)
    \notag \\[.8ex]
    \ \overset{(6)}{=}&\ \ 
         \pig[\big( t_\phi^{AA} g_\phi^{AA} \big) \rhd_\rho F^A \pig]\, \big(x^A(p) + \mathscr{v} \big)
\end{align}
It relates to the transporter pullback of the untransformed field via the induced representation (Eq.~\eqref{eq:induced_rep_affine}), acting with the coordinate expression $t_\phi^{AA} g_\phi^{AA}$ of $\phi$ (Eq.~\eqref{eq:AffGM_in_charts_eq}).
The first two steps make use of Eq.~\eqref{eq:affine_action_sections} and the definition of the transporter pullback in coordinates, where $(\dphiA f\, \phi^{-1})^A := \psiAp^A(\dphiA f\, \phi^{-1})$.
To translate all morphisms into the corresponding coordinate expressions, step three inserts identities $\id_{\R^c} = \big( \psiAphiinvp^A \big)^{-1} \psiAphiinvp^A$ and $\id_{\R^d} = \big( x^A \big)^{-1} x^A$, which are in step four rebracketed to clarify which combinations result in the coordinate expressions after step five.
Recall for step 5 that, by Theorem~\ref{thm:Aff_GM_in_charts}, $g_\phi^{AA}(p) = g_\phi^{AA}$ for any $p$ in $\Euc_d$.
As stated above, the last step identifies the resulting transformation law in coordinates as the action of the induced representation.

With this result we can prove the $\AffGM$-equivariance of Euclidean convolutions in the coordinate free setting.
This generalizes Theorem~\ref{thm:isom_equiv_GM_conv}, proving the isometry equivariance of $\GM$-convolutions for the specific case of Euclidean spaces.
\begin{thm}[Affine equivariance of Euclidean $\GM$-convolutions]
\label{thm:affine_equivariance_Euclidean_GM_conv}
    Let $\GM$ be a $G$-structure that is induced by some $\Aff(G)$-atlas of the Euclidean space~${M = \Euc_d}$ and assume feature vectors to be transported according to the Levi-Civita connection on~$\Euc_d$.
    The corresponding $\GM$-convolutions is then guaranteed to be equivariant under the action of $G$-structure preserving affine transformations $\AffGM \cong \Aff(G)$.
    In equations, we have for arbitrary feature fields $\fin \in \Gamma(\Ain)$ and $G$-steerable kernels $K\in\KG$ that
    \begin{align}
        \big[K \star (\phi\rhd \fin) \big]\ =\ \phi\rhd \big[K \star \fin \big] \qquad \forall\, \phi\in\AffGM \,,
    \end{align}
    i.e. that the following diagram commutes for any $\phi$ in $\AffGM$:
    \begin{equation}\label{cd:}
    \begin{tikzcd}[row sep=3.5em, column sep=5.em]
        \Gamma(\Ain)
            \arrow[r, pos=.5, "\phi \,\rhd"]
            \arrow[d, "K\star\,"']
        & \Gamma(\Ain)
            \arrow[d, "\,K\star"]
        \\
        \Gamma(\Aout)
            \arrow[r, pos=.5, "\phi \,\rhd"']
        & \Gamma(\Aout)
    \end{tikzcd}
    \end{equation}
\end{thm}
\begin{proof}
    Let $x^A: \Euc_d \to \R^d$ be any global chart of the considered $\Aff(G)$-atlas and let $p\in\Euc_d$.
    Our proof of the $\AffGM$-equivariance is then performed by expressing the convolution relative to these coordinates and making use of the $\Aff(G)$-equivariance of classical $G$-steerable convolutions on $\R^d$ from Eq.~\eqref{eq:Euclidean_conv_equiv_in_coords_Rd}:
    \begingroup
    \allowdisplaybreaks
    \begin{align}
        &\ \ 
            \psiAoutp^A\, \big[K \star (\phi\rhd \fin) \big] (p)
            \\[.8ex]
        =&\ \ 
            \int_{\R^d} K(\mathscr{v})\ 
            \big[\mkern-2mu \Expsp (\phi \rhd \fin) \big]^A (\mathscr{v})
            \,\ d\mathscr{v}
        \quad && \big( \text{\small $\GM$-convolution in coordinates, Eq.~\eqref{eq:gauge_conv_coord_expression} } \big) \notag \\[.8ex]
        =&\ \ 
            \int_{\R^d} K(\mathscr{v})\ 
            \pig[\big( t_\phi^{AA} g_\phi^{AA} \big) \rhd_{\rhoin}\! \Fin^A \pig]\, \big(x^A(p) + \mathscr{v} \big)
            \,\ d\mathscr{v}
        \quad && \big( \text{\small transformed transporter pullback, Eq.~\eqref{eq:affine_transformed_transporter_pullback} } \big) \notag \\[.8ex]
        =&\ \ 
            \pig[ K * \big( t_\phi^{AA} g_\phi^{AA} \rhd_{\rhoin}\! \Fin^A \big) \pig] \big( x^A(p) \big)
        \quad && \big( \text{\small identified convolution $*$ on $\R^d$ } \big) \notag \\[.8ex]
        =&\ \ 
            \pig[ t_\phi^{AA} g_\phi^{AA} \rhd_{\rhoout} \big( K * \Fin^A \big) \pig] \big( x^A(p) \big)
        \quad && \big( \text{\small $\Aff(G)$-equivariance on $\R^d$, Eq.~\eqref{eq:Euclidean_conv_equiv_in_coords_Rd} } \big) \notag \\[.8ex]
        =&\ \ 
            \rhoout \big( g_\phi^{AA} \big) \big( K * \Fin^A \big) \pig(\big( t_\phi^{AA} g_\phi^{AA} \big)^{-1} x^A(p) \pig)
        \quad && \big( \text{\small induced representation $\rhd_{\rhoout}$, Eq.~\eqref{eq:induced_rep_affine} } \big) \notag \\[.8ex]
        =&\ \ 
            \rhoout \big( g_\phi^{AA} \big) \big( K * \Fin^A \big) \big( x^A (\phi^{-1}(p)) \big)
        \quad && \big( \text{\small coordinate expression of $\phi$, Eq.~\eqref{eq:AffGM_in_charts_eq} } \big) \notag \\[.8ex]
        =&\ \ 
            \rhoout\big( g_\phi^{AA} \big)
            \int_{\R^d} K( \mathscr{v} )\ 
            \Fin^A \big( x^A\big( \phi^{-1}(p) \big) + \mathscr{v} \big)
            \,\ d\mathscr{v}
        \quad && \big( \text{\small expanded convolution $*$ on $\R^d$ } \big) \notag \\[.8ex]
        =&\ \ 
            \rhoout\big( g_\phi^{AA} \big)
            \int_{\R^d} K( \mathscr{v} )\ 
            \big[ \Expsphiinvpfin \big]^A (\mathscr{v})
            \,\ d\mathscr{v}
        \quad && \big( \text{\small Euclidean transporter pullback, Eq.~\eqref{eq:transporter_pullback_Euclidean} } \big) \notag \\[.8ex]
        =&\ \ 
            \rhoout\big( g_\phi^{AA} \big)
            \psiAoutphiinvp^A \big[K \star \fin \big]\, \phi^{-1} (p)
        \quad && \big( \text{\small $\GM$-convolution in coordinates, Eq.~\eqref{eq:gauge_conv_coord_expression} } \big) \notag \\[.8ex]
        =&\ \ 
            \psiAoutp^A\, \dphiAout
            \big[K \star \fin \big] \phi^{-1} (p)
        \quad && \big( \text{\small pushforward in coordinates, Eq.~\eqref{cd:pushforward_A_coord} } \big) \notag \\[.8ex]
        =&\ \ 
            \psiAoutp^A\, \pig[ \phi\rhd
            \big[K \star \fin \big] \pig] (p)
        \quad && \big( \text{\small $\AffGM$ action on feature fields, Eq.~\eqref{eq:affine_action_sections} } \big) \notag
    \end{align}
    \endgroup
    The statement follows since $\psiAoutp^A$ is an isomorphism.
\end{proof}

In summary, Euclidean $\GM$-convolutions with $\Aff(G)$-atlas induced $G$-structures have the following two properties:
\begin{itemize}[leftmargin=15em]
    \item[\it $\Aff(G)$-coordinate independence:]
        They are guaranteed to produce equivalent results in any chart of the $\Aff(G)$-atlas $\mathscr{A}^{\Aff(G)}_{\Euc_d}$.
        This property was shown in Eq.~\eqref{eq:Euclidean_conv_coordinate_independence} and is in Fig.~\ref{fig:affine_charts} visualized as the transformation \emph{between charts}.
    \item[\it active $\Aff(G)$-equivariance:]
        As proven in Theorem~\ref{thm:affine_equivariance_Euclidean_GM_conv} they are equivariant under active transformations of feature fields by $\AffGM \cong \Aff(G)$.
        In Fig.~\ref{fig:affine_charts}, this would correspond to an transformation of the signal on $\Euc_d$, which would reflect in an active transformation on its representation relative to the \emph{same chart}.
\end{itemize}
The proofs of both properties rely ultimately on the active $\Aff(G)$-equivariance of classical $G$-steerable convolutions on~$\R^d$ in Eq.~\eqref{eq:Euclidean_conv_equiv_in_coords_Rd}.

%% file: chapters/94_Euclidean_literature.tex

\subsection{Euclidean CNNs in the literature}
\label{sec:euclidean_literature}

All of the models in rows~(1-26) of Table~\ref{tab:network_instantiations} are $\Aff(G)$-equivariant $\GM$-convolutions on Euclidean spaces~$\Euc_d$ as discussed in the previous sections.
They differ in the dimensionality~$d$ of the Euclidean space, the structure group~$G$ and thus global symmetry group $\Aff(G)$, the group representations or field types~$\rho$ and choices of discretizations.
This section discusses the models briefly by grouping them by their field types into irrep models, regular representation models (corresponding to group convolutions) and variations of them, quotient representation models and others.
Conventional CNNs, which we review first, fall a bit out of this classification as their trivial structure group leads to feature fields and kernels without any symmetry constraints.

Row~(1) lists Euclidean $\GM$-convolutions on translation invariant $\{e\}$-structures as visualized in Fig.~\ref{fig:G_structure_R2_1}.
Due to the triviality of the structure group $G=\{e\}$ no (non-trivial) gauge transformations exist and the only possible choice of group representation is the trivial representation.
The $G$-steerability constraint becomes therefore trivial, such that the space of admissible convolution kernels remains unrestricted.
When being pulled back to $\R^d$ via a chart, the $\GM$-convolution becomes by Theorem~\ref{thm:Euclidean_GM_conv_is_conventional_conv} a conventional convolution (correlation).
Theorem~\ref{thm:affine_equivariance_Euclidean_GM_conv} asserts its translational equivariance.
The models are therefore seen to correspond to the conventional convolutional networks by~\citet{LeCun1990CNNs}.

All of the other Euclidean models in rows~(2-26) consider non-trivial structure groups.
They can be thought of as conventional convolutions on~$\R^d$ with the additional constraint on the kernels to be $G$-steerable, which guarantees their $\Aff(G)$-equivariance.

\paragraph{Irrep features:}
The networks in rows~(4, 9, 10, 17, 23) and (26) operate on feature fields that transform according to \emph{irreducible representations} (irreps) of~$G$.
For $G=\SO2$, listed in row~(4), this leads to so-called harmonic networks~\cite{Worrall2017-HNET,Weiler2019_E2CNN}.
This name is motivated by the fact that the kernel constraint allows in this case only for spectrally localized circular harmonics of frequency $m-n$ when mapping between fields that transform according to irreps of order~$n$ in the input and order~$m$ in the output.%
\footnote{
    Only frequency $m-n$ is allowed when considering \emph{complex} irreps of~$\SO2$.
    For \emph{real} irreps, the constraint allows for frequencies $|m-n|$ and $m+n$.
    More details can be found in Appendices F.5 of~\cite{Weiler2019_E2CNN} and E.1 and E.2 of~\cite{lang2020WignerEckart}.
}
The additional reflectional constraint for $G=\O2$, listed in row~(10), adds parity selection rules that fix the phase of the circular harmonics, suppressing half of the degrees of freedom as compared to the $G=\SO2$ case~\cite{Weiler2019_E2CNN}.
The models by \cite{3d_steerableCNNs,Thomas2018-TFN,miller2020relevance,Kondor2018-NBN,anderson2019cormorant} in row~(17) consider irreps of $G=\SO3$ and can therefore be seen as the analog to the models in row~(4) in three dimensions.
The space of valid kernels to map between fields that transform according to irreps (Wigner D-matrices) of orders $n$ and $m$ is here spanned by all spherical harmonics of the $2(\min(m,n)+1)$ orders $j$ with $|m-n| \leq j \leq m+n$.
As proven in~\cite{lang2020WignerEckart}, this generalizes to any compact structure group, with the admissible frequencies of the harmonics being determined by the corresponding Clebsch-Gordan coefficients labeled by $m,n$ and~$j$.
A variation of this approach is listed in row~(23) \cite{poulenard2019effective}.
A convolution of an input scalar field with spherical harmonics yields irrep fields of the corresponding order.
However, instead of processing these irrep features further via convolutions, the authors compute their norm.
This results in scalar fields, which are in the next layer processed in the same manner.
The model from~\cite{shutty2020learning} in row~(26) does not assume the standard Euclidean metric but the Minkowski metric.
Its structure group is the Lorentz group $G=\SO{d-1,1}$ and the global symmetry group is the Poincar\'{e} group.
In addition to building the equivariant network, the authors propose an algorithm to compute the irreps of Lie groups from the structure constants of their Lie algebra.

A special case of irreps are \emph{trivial representations}, which describe $G$-invariant feature vectors (scalars).
Due to their invariance, such features can not encode differences between any patterns in $G$-related poses.
The constraint on kernels that map between scalar fields becomes $K(g\mkern1mu \mathscr{v}) = K(\mathscr{v})$ for any $\mathscr{v} \in \R^d$ and any $g\in G$, enforcing kernels that are (in every channel individually) invariant under the action of $G$.
This is for reflections $G=\Flip$ visualized in the upper left entry of Table~\ref{tab:reflection_steerable_kernels}.
Interpreting the pixel grid of an image as a graph and applying a standard graph convolution to it corresponds to a trivially steerable convolution with $\O2$-invariant kernels since standard graph convolutions apply isotropic kernels~\cite{khasanova2018isometric}.

An advantage of irrep features from a practical viewpoint is their low dimensionality and thus memory consumption per feature field.
However, empirical results show that irrep field based steerable convolutions usually achieve a lower performance than other field types, for instance those based on regular representations.
This statement is reflected in our evaluation of M\"{o}bius convolutions in Section~\ref{sec:mobius_evaluation} and the benchmark of isometry equivariant Euclidean convolutions in~\cite{Weiler2019_E2CNN}.

\paragraph{Regular features and group convolutions:}
The probably most prominent class of group representations in equivariant deep learning are \emph{regular representations} of the structure groups.
Regular representations operate on a suitable%
\footnote{
    For instance, for topological groups, the functions are typically required to be continuous.
    For locally compact groups one usually considers the space $L^2(G)$ of square integrable functions on $G$.
}
space of functions $\digamma: G \to \R$ by translating them, that is,
$\big[ \rho_\textup{reg} (\tilde{g}) \digamma \big](g) = \digamma\big( \tilde{g}^{-1}g \big)$.%
\footnote{
    Regular representations over a different field $\mathbb{K}$ than the reals take values in this field, i.e. $\digamma: G \to \mathbb{K}$.
}
For finite groups this implies feature fields with the number of channels $c = |G|$ given by the order of the group.
As non-finite groups imply non-finite regular representations, the corresponding features are in practice discretized, which is mostly done by considering a finite subgroup of the structure group.
Since regular feature fields $f \in \Gamma(\A)$ assign a function $f^A(p): G \to \R$ to each point $p\in M$ (when being expressed relative to any gauge $A$ at $p$), they are equivalent to real-valued functions $\tilde{f}: \GM \to \R$ on the $G$-structure~$\GM$.%
\footnote{
    Theorem~\ref{thm:regular_field_scalar_GM} in Appendix~\ref{apx:regular_field_scalar_GM} proves this isomorphism
    $C^\infty(\GM)\ \cong\ \Gamma(\A_{\rho_\textup{reg}})$
    for the practically relevant case of $G$ being a finite group.
}
For the case of $\GM$ being induced by an $\Aff(G)$-atlas, this is furthermore equivalent to real-valued functions $\accentset{\approx}{f}: \Aff(G) \to \R$ on $\Aff(G) \cong \GM$ (along the isomorphism in Eq.~\eqref{eq:principal_bundle_iso_AffG_GM}).
Equivariant linear maps between functions on the group $\Aff(G)$ are \emph{group convolutions} (see Eq.~\eqref{eq:group_conv_def} in Section~\ref{apx:homogeneous_conv} and Section~7.11 in~\cite{gallier2019harmonicRepr}), which means that affine group convolution based CNNs are covered by our framework~\cite{Cohen2016-GCNN,Kondor2018-GENERAL,bekkers2020bspline}.

$\Aff(G)$-group convolutions are in Table~\ref{tab:network_instantiations} listed in rows (2,3,5,11,15,19,21,24) and (25).
As these models typically process gray-scale or scalar images, they apply an initial convolution from scalar fields to regular fields, followed by group convolutions, i.e. convolutions from regular to regular fields.
As regular representations are permutation representations, they typically apply pointwise nonlinearities like ReLUs to each of the field channels individually.%
\footnote{
    As the action of nonlinear maps depends on the chosen basis, this is what really distinguishes regular (or any other non-irreducible) feature fields from their decomposition into irrep fields; see footnote~\ref{footnote:feature_field_irrep_decomposition} and the discussion in Section~\ref{sec:mobius_representations}.
}
The reflection equivariant CNN on $\Euc_2$ from~\cite{Weiler2019_E2CNN} in 
row (3)
applies $\Flip$-steerable kernels, as derived in Section~\ref{sec:mobius_conv} and visualized in the bottom right entry of Table~\ref{tab:reflection_steerable_kernels}.
As the reflection group is finite with order $|\Flip| = 2$, the regular feature fields have two channels, each of which is associated to one of the two frame orientations of the $\Flip$-structure in Fig.~\ref{fig:G_structure_R2_3}.
The resulting model is globally $\Trans_2 \rtimes \Flip = \Aff(\Flip)$ equivariant.
To construct $\SE2 = \Aff(\SO2)$ equivariant group convolutions one would in theory have to consider the $\SO2$-structure in Fig.~\ref{fig:G_structure_R2_2} with feature fields transforming according to the regular representation of $\SO2$.
In practice, most of the models in 
row (5)
of Table~\ref{tab:network_instantiations}
approximate this via regular representations of the cyclic groups $\CN \leq \SO2$, which are finite subgroups of discrete rotations by multiples of $2\pi/N$.
As the order of these groups is $|\!\CN\!|=N$, the corresponding feature fields are $N$-dimensional.
While the model performance is initially significantly increasing with $N$, it is empirically found to saturate at approximately $8$ to $12$ sampled directions~\cite{Weiler2018SFCNN,Weiler2019_E2CNN,bekkers2020bspline}.
For an intuition on the spaces of $\CN$-steerable kernels we refer to the visualizations in~\cite{Weiler2018SFCNN,bekkers2018roto,bekkers2020bspline}.
The $\E2 = \Aff(\O2)$ equivariant group convolutions in
row (11)
are similarly approximated via dihedral subgroups $\DN \leq \O2$, which consist of $N$ rotations, each in two reflections.
The feature fields are in this case $|\!\DN\!| = 2N$-dimensional.
Simultaneous equivariance under translations and scaling is achieved by the $\Trans_d \rtimes \Scale = \Aff(\Scale)$ group convolutions in
rows (2) and (15).
The scaling group is hereby commonly discretized.
As this would still lead to a (countably) infinite group order, the implementations introduce cutoffs, i.e. minimal and maximal scales as shown by the frames in Fig.~\ref{fig:G_structure_R2_4}.
Note that this leads to similar boundary effects as for conventional convolutions at the border of an image.
The models in
rows (19) and (21)
are equivariant w.r.t. translations, rotations and, for the latter, reflections on three-dimensional Euclidean spaces~$\Euc_3$.
While \citet{finzi2020generalizing} choose a Monte-Carlo discretization of the regular representation,
the models in \cite{Worrall2018-CUBENET,winkels3DGCNNsPulmonary2018} are based on different discrete subgroups of $\SO3$ or $\O3$.
A current limitation of group convolution based rotation and reflection equivariant models in three dimensions is their high memory and compute requirement.
For instance, the symmetry group of the cube, which has still a quite coarse resolution of rotations by $\pi/2$, already consists of $48$ group elements, implying $48$-dimensional feature fields in three-dimensional space.
On the other hand, the large number of symmetries reflects the greatly enhanced data efficiency of such models:
the authors of \cite{winkels3DGCNNsPulmonary2018} report the same performance of an equivariant model in comparison to a non-equivariant ($\{e\}$-steerable) network despite training on a $10$ times smaller dataset.
The models in
rows (24) and (25)
convolve on $\Euc_3$, however, they consider cyclic and dihedral structure groups $\C4$ and $\D4$, i.e. planar rotations and reflections around the (thus defined) $z$-axis.
Their steerable kernels are therefore similar to those of the models from
rows (5) and (11)
but extend additionally in a new $z$-direction.

\paragraph{Regular to scalar and vector pooling:}
A variation of group convolutional networks are the models in
rows (7,8,13,16) and (20),
which are labeled by regular$\xrightarrow{\textup{pool}}$trivial and regular$\xrightarrow{\textup{pool}}$vector.
After applying a convolution to regular feature fields, they perform a $\operatorname{max}$-\emph{pooling} operation over channels, which results in scalar (trivial) fields \cite{Cohen2016-GCNN,marcos2016learning,Weiler2019_E2CNN,ghosh2019scale,andrearczyk2019exploring}, or a $\operatorname{max}$-pooling together with an $\operatorname{argmax}$, from which vector fields can be computed~\cite{Marcos2017-VFN,Weiler2019_E2CNN}.
Subsequent convolutions map from the resulting scalar or vector fields back to regular feature fields.
As the pooling operations reduce the number of channels significantly from $|G|$ to $1$ or $d$, respectively, the models become more memory and compute efficient than conventional group convolutions.
On the downside the pooling is accompanied with a loss of information, which is empirically found to decrease the model performance~\cite{Weiler2019_E2CNN}.

\paragraph{Quotient features:}
Rows (6,12) and (22) list models whose feature fields transform according to \emph{quotient representations} of the structure group, which are permutation representations that are similar to regular representations.
Given a subgroup~$\widehat{G}$ of~$G$, the corresponding quotient representation acts on scalar functions $\digamma: G/\widehat{G} \to \R$ on the quotient space $G/\widehat{G}$ via translation, that is,
$\big[\rho_\textup{quot}^{G/\widehat{G}} (\tilde{g})\mkern2mu \digamma\big] (g\mkern1mu \widehat{G}) = \digamma(\tilde{g}^{-1}g \mkern2mu \widehat{G})$.
The dimensionality of the feature fields is therefore given by the index ${|G:\mkern-2mu\widehat{G}|}$ of $\widehat{G}$ in $G$, which is for finite groups equal to~$|G|/|\widehat{G}|$.
Feature fields that transform under quotient representations can be seen as symmetry-constrained regular feature fields that are forced to take the same value on all group elements in the same coset $g\widehat{G}$ of $\widehat{G}$ in~$G$.
A specific example are the representations in row (22), which are associated with the quotient $\O3/\O2 \cong S^2$.
Instead of allowing for arbitrary convolution kernels, the kernel constraint leads here to kernels which are invariant under rotations around the $z$-axis; see the visualizations in~\cite{janssen2018design}.
More details and a graphical intuition on quotient representation based feature fields can be found in Appendix~C of~\cite{Weiler2019_E2CNN}.
The theory proposed in \cite{Kondor2018-GENERAL} covers quotient fields from an alternative viewpoint of group convolutions on right quotient spaces.

\paragraph{Induced representations:}
A generalization of regular and quotient representations are induced representations like the \emph{induced $\SO2$ irreps} in row (14)
of Table~\ref{tab:network_instantiations}.
Given any $\SO2$-irrep $\rho: \SO2 \to \GL{n}$, the induced representation $\Ind_{\SO2}^{\O2}\rho: \O2 \to \GL{c}$ of~$\O2$ with $c = n \mkern-2mu\cdot\! {|\O2:\SO2|} = 2n$
acts in the following way:
reflections permute two $n$-dimensional, orthogonal subspaces of $\R^{2n}$ which correspond to the two cosets in $\O2/\SO2$ while rotations act on the individual subspaces via~$\rho$.
For $\rho$ being the trivial representation of $\SO2$ this recovers quotient representations as discussed above.
In comparison to $\O2$-irrep feature fields, the induced $\SO2$-irrep fields show a significantly improved performance.
A more detailed description and empirical evaluation of these field types can be found in~\cite{Weiler2019_E2CNN}.

The last type of representation listed in Table~\ref{tab:network_instantiations} is the quaternion representation of three-dimensional rotations in row (18)~\cite{zhang2019quaternion}.
It makes use of the usual representation of rotations via quaternions, which relies the identification of unit quaternions with $\operatorname{SU}(2)$ and the existence of a surjective group homomorphism from $\operatorname{SU}(2)$ to~$\SO3$.
Note that the quaternion representation is actually a projective representation of $\SO3$.

While our theory is formulated on continuous Euclidean spaces, implementations sample feature fields on discrete subsets.
The most common discretization of $\Euc_d$ is in terms of the pixel grid $\Z^d$.
An alternative are hexagonal planar grids on $\Euc_2$ as investigated by~\citet{Hoogeboom2018-HEX}.
If such regular pixel grids are chosen, a basis of $G$-steerable kernels can be precomputed and sampled on this grid.
Data like events in spacetime~\cite{shutty2020learning} or molecules in $\R^3$ \cite{Thomas2018-TFN,Kondor2018-NBN,anderson2019cormorant,miller2020relevance} are instead usually represented by irregular point clouds.
In this case the kernels need to be given analytically, which allows their online sampling during the forward pass.

Finally, we want to mention that there exist \emph{globally} $\Aff(G)$-equivariant models which are \emph{not locally} $G$-equivariant.
An example is TI-pooling (transformation-invariant pooling)~\cite{Laptev_2016_CVPR}, which feeds a set of globally transformed feature fields through a conventional CNN and finally pools the resulting features over these transformations, which results in an invariant descriptor.

%% file: chapters/100_polar_intro.tex

\section{Rotation equivariant CNNs on punctured Euclidean spaces}
\label{sec:instantiations_euclidean_polar}

\begin{figure}
    \centering
    \begin{subfigure}[b]{0.47\textwidth}
        \centering
        \includegraphics[width=.7\textwidth]{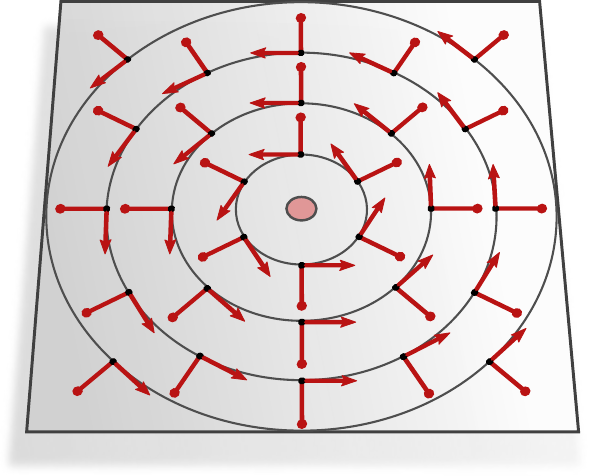}
        \captionsetup{format=hang, width=.82\textwidth}
        \caption{\small
            $\SO2$-invariant $\{e\}$-structure as implicitly assumed by \citet{finzi2020generalizing}.
        }
        \label{fig:G_structure_R2_no_origin_SO2}
    \end{subfigure}
    \hfill
    \begin{subfigure}[b]{0.47\textwidth}
        \centering
        \includegraphics[width=.7\textwidth]{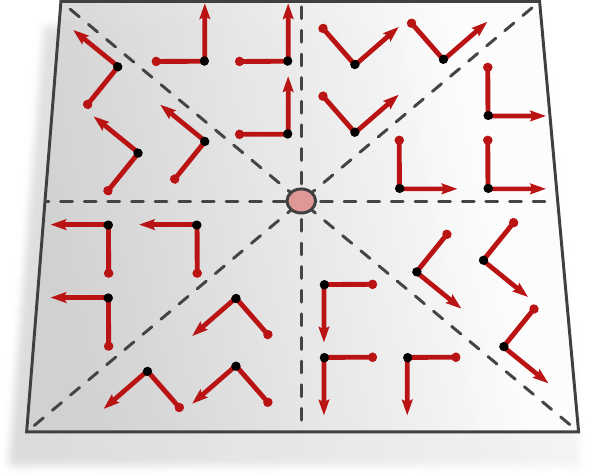}
        \captionsetup{format=hang, width=.78\textwidth}
        \caption{\small
            $\C8$-invariant $\{e\}$-structure as implicitly assumed by \citet{chidester2019rotation}.
        }
        \label{fig:G_structure_R2_no_origin_C8}
    \end{subfigure}
    \caption{\small
        Two examples of $\{e\}$-structures on the punctured plane $\Euc_2 \backslash \{0\}$ which
        1) are invariant under rotations around the origin $\{0\}$ and
        2) consist of orthonormal frames relative to the standard Euclidean metric.
        The corresponding $\GM$-convolutions are rotation equivariant but not translation equivariant (in fact, $\Euc_2 \backslash \{0\}$ does not even admit translations).
    }
    \label{fig:G_structures_R2_no_origin}
\end{figure}

The models in rows (27-30) of Table~\ref{tab:network_instantiations}
provide an interesting alternative for rotation equivariant convolutions on punctured Euclidean spaces $\Euc_d\backslash\{0\}$.
They rely on \emph{$G$-structures that are invariant under rotations around the chosen origin $\{0\}$}, as visualized for instance in Fig.~\ref{fig:G_structures_R2_no_origin}.
By specifying a preferred origin, these models lose the property to be translation equivariant.%
\footnote{
    This issue can be resolved by combining the network with a translation invariant origin predictor network~\cite{esteves2017polar}.
    Note that the rotation equivariance of the combined model is only preserved if this origin predictor is $\SE{d}$-equivariant.
}
However, if the $G$-structure is additionally invariant under scaling, which is for instance the case when it is induced by hyperspherical coordinates with a logarithmic radial component as shown in Fig.~\ref{fig:G_structure_R2_no_origin_logpolar}, the models become equivariant w.r.t. the direct product $\SO{d} \!\times\! \Scale$ of the rotation and scaling group.
Similarly, rotation and reflection invariant $G$-structures, visualized in Fig.~\ref{fig:G_structure_R2_no_origin_O2}, imply the $\O{d}$-equivariance of the corresponding $\GM$-convolutions.

The models relate to spherical CNNs, discussed in Section~\ref{sec:instantiations_spherical} below, in two ways.
Firstly, they assume rotationally invariant $G$-structures on $\Euc_d \backslash \{0\} \,\cong\, S^{d-1} \!\times \R^+$, which can be seen as being composed of multiple rotationally invariant $G$-structures on ${(d -\! 1)}$-dimensional spherical shells $S^{d-1}$ at different radii.
The models can therefore be thought of as (hyper)spherical CNNs with an additional radial dimension~$\R^+$~\cite{ramasinghe2019representation}, which is in Fig.~\ref{fig:G_structure_R3_no_origin} visualized for the case of $d=3$ dimensions.
Secondly, the polar coordinate systems of \cite{esteves2017polar,finzi2020generalizing,chidester2019rotation} (Figs.~\ref{fig:G_structures_R2_no_origin} and~\ref{fig:G_structure_R2_no_origin_logpolar}) induce $G$-structures that exhibit the same type of singularity at their origin like those of the punctured spherical CNNs in Fig.~\ref{fig:G_structure_S2_2} at the poles.
Note that the punctured Euclidean plane $\Euc_2 \backslash \{0\}$ and the punctured sphere $S^2 \backslash \{n,s\}$ (with north and south poles $\{n,s\}$ removed) are both topologically equivalent to a cylinder $S^1 \times \R^+ \cong S^1 \times \R$ and that the cylindrical $\{e\}$-structures visualized in Figs.~\ref{fig:G_structure_R2_no_origin_SO2}, \ref{fig:G_structure_R2_no_origin_logpolar} (left) and~\ref{fig:G_structure_S2_2} are diffeomorphic.

A major difference in comparison to the $\SE{d}$-equivariant networks from the previous section is that the models of the current section are only \emph{globally} $\SO{d}$-equivariant around the origin instead of \emph{locally} $\SO{d}$-equivariant ($\SO{d}$-steerable).
While the globally equivariant models do not require $\SO{d}$-steerable kernels, they still require at least $\SO{d-1}$-steerable kernels.
This is the case since $\SO{d}$ is a $\SO{d-1}$-bundle over the spherical shells $S^{d-1} \cong \SO{d}/\SO{d-1}$ on which the $G$-structure is required to be $\SO{d}$ rotation equivariant.
For $d=2$ this allows for $\{e\}$-structures and non-steerable kernels since $\SO{d-1} = \SO{1} = \{e\}$; see Fig.~\ref{fig:G_structures_R2_no_origin} or~\ref{fig:G_structure_R2_no_origin_logpolar}.
For $d=3$ this requires at least a $\SO{d-1} = \SO{2}$-structure on the individual spherical shells, which is visualized in Fig.~\ref{fig:G_structure_S2_1}.

After these general remarks we will in the following briefly review the individual models on $\Euc_d \backslash \{0\}$ found in the literature from the viewpoint of coordinate independent CNNs.
We start with the models in row~(27) of Table~\ref{tab:network_instantiations}, which are equivariant w.r.t. $\SO2$ rotations around a chosen origin of~$\Euc_2$ and proceed with the models in row~(28), which are additionally scale equivariant.
The network listed in row~(29), which we discuss last, is globally $\O3$-equivariant around the origin of~$\Euc_3$.

%% file: chapters/101_polar_Euc2_rot.tex

\subsubsection*{Global rotation equivariance on $\fakebold{\Euc}_{\boldsymbol{2}} \fakebold{\backslash} \bm{\{0\}}$}
\label{sec:polar_Euc2_rot}

We start with the conceptually simplest models, which are globally rotation equivariant networks that rely on solely rotation invariant $\{e\}$-structures on~$\Euc_2 \backslash \{0\}$~\cite{finzi2020generalizing,chidester2019rotation}.
These models assume the standard Euclidean metric on~$\Euc_2 \backslash \{0\}$, relative to which the frames are orthonormal.
Together, these two requirements imply $\{e\}$-structures as shown in Fig.~\ref{fig:G_structures_R2_no_origin}.

In addition to the considered $G$-structures, the networks depend on the specific implementation of the transporter pullback and thus on the geodesics and parallel transporters.
The geodesics are in both models assumed to be the standard geodesics on Euclidean spaces (i.e. straight lines), corresponding to the Levi-Civita connection of the Euclidean metric.
As $\Euc_2 \backslash \{0\}$ is not geodesically complete, zero-padding has to be used for exponential maps that would end at the origin.
Note that this does not have an impact on the final result as the lost geodesics are of measure zero.

The parallel transport of feature vectors, on the other hand, does \emph{not} correspond to the Levi-Civita connection since the Levi-Civita connection is not compatible with the $\{e\}$-structures.
Instead, the models assume the unique $\{e\}$-compatible \emph{trivial connections} which are implied by the respective $\{e\}$-structures.%
\footnote{
\label{footnote:punctured_Euclidean_transport}
    An animation of the $\{e\}$-compatible transport corresponding to Fig.~\ref{fig:G_structure_R2_no_origin_SO2} can be found on
    \href{https://en.wikipedia.org/wiki/Levi-Civita_connection\#Parallel_transport}{\underline{Wikipedia}}.
}
According to the trivial connections, the numerical coefficients of feature vectors do not transform when being transported, despite the frames being rotated relative to the usual notion of parallelism on Euclidean spaces.
In practice, this just means that the transporters $\rho(g^{A\widetilde{A}}_\gamma) = \id_{\R^c}$ can be ignored in the implementation -- which is the reason that they are not being discussed in the original papers~\cite{finzi2020generalizing,chidester2019rotation}.

As rotations leave the considered $\{e\}$-structures invariant and are at the same time isometries, we have
$\IsomeM = \SO2$ for the model by~\citet{finzi2020generalizing} (Fig.~\ref{fig:G_structure_R2_no_origin_SO2}) and
$\IsomeM = \C8$  for the model by~\citet{chidester2019rotation} (Fig.~\ref{fig:G_structure_R2_no_origin_C8}).
Theorem~\ref{thm:isom_equiv_GM_conv} asserts then that the corresponding $\GM$-convolutions are $\IsomeM$-equivariant, which is in agreement with the statements made by the authors.

\begin{SCfigure}
    \centering
    \includegraphics[width=.34\textwidth]{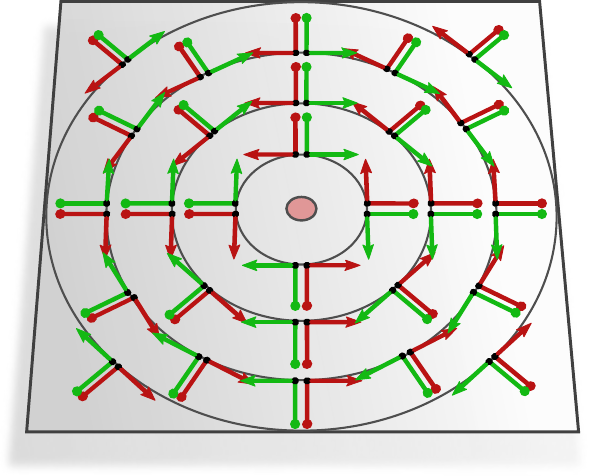}
    \hspace*{2.ex}
    \captionsetup{width=.92\textwidth}
    \caption{\small
        An $\O2$-invariant $\Flip$-structure on $\Euc_2 \backslash \{0\}$, which is constructed by adding a reflected versions to each frame of the $\{e\}$-structure in Fig.~\ref{fig:G_structure_R2_no_origin_SO2}.
        The corresponding $\GM$-convolution is simultaneously equivariant w.r.t. global rotations and reflections in $\IsomRM = \O2$ around the origin.
        \\\protect\rule{0ex}{5.5ex}
    }
    \label{fig:G_structure_R2_no_origin_O2}
\end{SCfigure}

Before going on we want to mention that the $\C8$-invariant $\{e\}$-structure in Fig.~\ref{fig:G_structure_R2_no_origin_C8} is not continuous and does therefore not guarantee a continuous (or smooth) inference.
An advantage of this $\{e\}$-structure from an engineering viewpoint is that it is locally isometric to the canonical $\{e\}$-structure of $\R^2$, which allows to run conventional Euclidean convolution routines on each octant.
The authors discuss the generalization to $\CN$-invariant $\{e\}$-structures, which become in the limit $N\to\infty$ equivalent to the $\SO2$-invariant $\{e\}$-structure in Fig.~\ref{fig:G_structure_R2_no_origin_SO2}.

It is furthermore possible to make the models globally $\O2$-equivariant by using reflection steerable kernels instead of unconstrained kernels.
From a theoretic viewpoint this corresponds to the $\IsomRM = \O2$-invariant $\Flip$-structure $\RM$ on $\Euc_2 \backslash \{0\}$ shown in Fig.~\ref{fig:G_structure_R2_no_origin_O2}.
Note that $\RM$ is a $\Flip$-bundle over $\Euc_2 \backslash \{0\}$, whose restriction to circles of constant radius is as a principal bundle isomorphic to $\O2$, interpreted as $\Flip$-bundle over the quotient space $\O2/\Flip \cong S^1$.

%% file: chapters/102_polar_Euc2_logpolar.tex

\subsubsection*{Global rotation and scale equivariance on $\fakebold{\Euc}_{\boldsymbol{2}} \fakebold{\backslash} \bm{\{0\}}$ via log-polar coordinates}
\label{sec:polar_Euc2_logpolar}

\begin{figure}
    \centering
    \includegraphics[width=.94\textwidth]{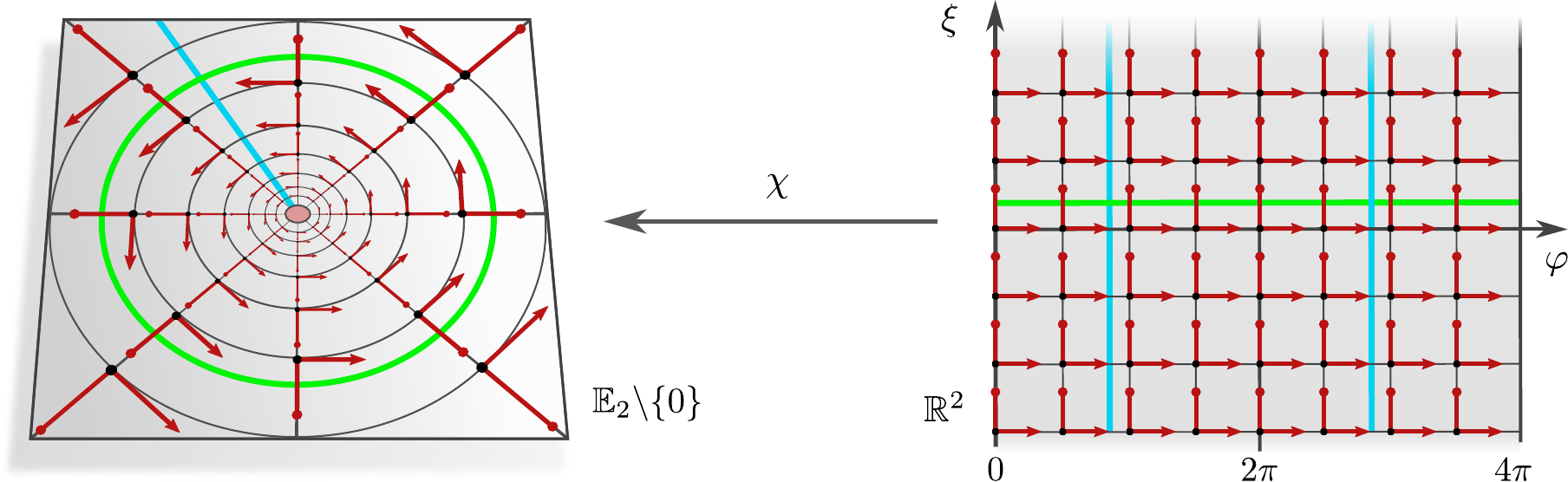}
    \vspace*{2ex}
    \caption{\small
        Log-polar coordinates 
        $\chi: \R^2 \to \R^2 \backslash \{0\} :\, (\varphi,\xi) \mapsto \big( e^{\xi}\cos(\varphi) ,\, e^{\xi}\sin(\varphi) \big)$
        map angles $\varphi \in \R$ and log-radii $\xi = \log\lVert p\rVert \in \R$ to points $p$ in $\R^2 \backslash \{0\}$.
        After choosing Cartesian coordinates of $\Euc_2 \backslash \{0\} \cong \R^2 \backslash \{0\}$, this yields a coordinatization of $\Euc_2 \backslash \{0\}$ by $\R^2$.
        The log-polar coordinates imply an $\{e\}$-structure on $\Euc_2 \backslash \{0\}$, consisting of reference frames
        $\big[ \frac{\partial}{\partial \varphi} ,\, \frac{\partial}{\partial \xi} \big]$ that are aligned with the coordinate grid.
        They furthermore induce a Riemannian metric, which differs from the usual Euclidean metric and relative to which the induced frames are orthonormal.
        $\GM$-convolutions on the $\{e\}$-structure correspond to conventional Euclidean convolutions in the coordinates $\R^2$.
        Translations $(\Delta\varphi,\, \Delta\xi) \in \Trans_2$ on $\R^2$ correspond via $\chi$ to rotations and rescalings of $\Euc_2 \backslash \{0\}$, where the rotation angles and rescaling factors are given by $\Delta\varphi$ and $e^{\Delta\xi}$, respectively.
        The translation equivariance of the convolution in coordinates $\R^2$ implies therefore the ${\SO2 \!\times\! \Scale}$-equivariance of the $\GM$-convolution on $\Euc_2 \backslash \{0\}$.
        This result is in agreement with the isometry equivariance of the $\GM$-convolution since the transformations in $\IsomGM = {\SO2 \!\times\! \Scale}$ are isometries relative to the induced metric.
        \citet{esteves2017polar} implement such $\GM$-convolutions in terms of conventional convolutions on $\R^2$.
     }
    \label{fig:G_structure_R2_no_origin_logpolar}
\end{figure}

By making the rotation invariant $G$-structures from the last section additionally scale invariant, the corresponding $\GM$-convolutions become equivariant w.r.t. the direct product group ${\SO2 \!\times\! \Scale}$.
Such $G$-structures are induced by \emph{log-polar coordinates}, shown in Fig.~\ref{fig:G_structure_R2_no_origin_logpolar}, which allow for a convenient implementation of the $\GM$-convolution in terms of conventional Euclidean convolutions on the coordinate representation $\R^2$.
The translation equivariance of convolutions on $\R^2$ corresponds then to the ${\SO2 \!\times\! \Scale}$-equivariance on $\Euc_2 \backslash \{0\}$.
For clarity, we start by describing the model in terms of log-polar coordinates as proposed by \citet{esteves2017polar}.%
\footnote{
    The idea to implement rotation invariant correlations via log-polar transforms appeared already in the 80's~\cite{saito1983scale,casasent1987real}.
}
Subsequently, we investigate how this model and its properties are explained in our framework.

Log-polar coordinates of the punctured Euclidean vector space $\R^2 \backslash \{0\}$ are defined in terms of the smooth surjection
\begin{align}
    \chi:\, \R^2 \to \R^2 \backslash \{0\} \,,\ \ 
    (\varphi, \xi) \mapsto \big( e^\xi \cos(\varphi) ,\, e^\xi \sin(\varphi) \big) \,,
\end{align}
which assigns points $p = \chi(\varphi,\xi)$ in $\R^2 \backslash \{0\}$ to a given polar angle $\varphi \in \R$ and log-radius ${\xi = \log\lVert p\rVert \in \R}$.
This map is $2\pi$-periodic in the angular coordinate (note the repetition of the blue stripe on the r.h.s. of Fig.~\ref{fig:G_structure_R2_no_origin_logpolar}) and is therefore in particular non-injective.
A restriction to $[0,2\pi) \times \R$ would be bijective and continuous, however, not homeomorphic -- this will require us below to consider at least two charts to cover the punctured plane.
Cartesian coordinates identify $\R^2 \backslash \{0\}$ with $\Euc_2 \backslash \{0\}$, and therefore allow to assign log-polar coordinates to the latter.
As different (right-handed) Cartesian coordinate systems that are centered in the origin of $\Euc_2 \backslash \{0\}$ differ only by rotations, the assignment of log-polar coordinates is ambiguous by a shift in the angular component.

Given a feature map $f: \R^2 \backslash \{0\} \to \R^c$ (an $\eM$-associated feature field, as clarified below), \citet{esteves2017polar} consider its pullback $\tilde{f} := f \circ \chi \ : \R^2 \to \R^c$ via log-polar coordinates, defined by the commutativity of the following diagram:
\begin{equation}
\begin{tikzcd}[row sep=2.5em, column sep=5.em]
    \R^2
        \arrow[r, "\chi"]
        \arrow[rr, rounded corners, to path={ 
                |- node[below, pos=.75]{\small$\tilde{f}$} ([yshift=-2.5ex, xshift=0ex]\tikztotarget.south)
                -- ([xshift=0ex]\tikztotarget.south)
                }]
    & \R^2 \backslash \{0\}
        \arrow[r, "f"]
    & \R^c
\end{tikzcd}
\end{equation}
A rotation and scaling equivariant group convolution of the feature field~$f$ on $\R^2 \backslash \{0\}$ is then defined by
1)~pulling it via $\chi$ back to coordinates $\R^2$
2)~applying a conventional Euclidean convolution there and
3)~mapping the result back to $\R^2 \backslash \{0\}$.
This procedure is well defined since $\chi$ is smooth, such that smooth feature maps (feature fields) $f$ result in smooth and periodic pullbacks $\tilde{f}$.
Since convolutions are position independent, their output feature map will still be periodic and smooth, and corresponds therefore uniquely to a smooth feature map on $\R^2 \backslash \{0\}$.%
\footnote{
    To see this, note that $\chi$ is a quotient map (since its angular part is a quotient map $\R \to S^1 \cong \R/2\pi\Z$).
    For continuous (instead of smooth) feature maps the statement follows from the universal property of quotient spaces; see e.g.
    \href{https://en.wikipedia.org/wiki/Quotient_space_(topology)\#Properties}{\underline{Wikipedia}}.
    As the smoothness of a function is defined as its continuous differentiability, the universal property can be applied recursively to show that the statement holds for smooth feature maps as well.
}

The rotation and scaling equivariance of the implied group convolution on $\R^2 \backslash \{0\}$ follows from the translation equivariance of the coordinate function~$\chi$.%
\footnote{
    That this is possible relies on the fact that there is a group homomorphism
    $\Trans_2 \to \SO2 \!\times\! \Scale,\ (\Delta\varphi,\, \Delta\xi) \mapsto (R_{\Delta\varphi},\, e^{\Delta\xi})$,
    defined by the group isomorphism $\exp: \Trans_1 \to \Scale$ on the second factor and the group homomorphism (quotient map) $R: \Trans_1 \to \SO2 \cong \Trans_1 / 2\pi\Z$ on the second factor, where $R_{\Delta\varphi}$
    \mbox{denotes the rotation matrix by an angle of $\Delta\varphi$.}
}
Let $(\varphi, \xi)$ be any coordinates in $\R^2$ and let $(\Delta\varphi, \Delta\xi)$ be any translation in $\Trans_2$.
The point of $\R^2 \backslash \{0\}$ that corresponds to translated coordinates ${(\varphi \!+\! \Delta\varphi} ,\, {\xi \!+\! \Delta\xi)}$ relates then to the point corresponding to non-translated coordinates $(\varphi, \xi)$ via a scaling by the factor $e^{\Delta\xi}$ and rotation by the angle $\Delta\varphi$:
\begin{align}\label{eq:log_polar_translation_equivariance}
    \chi \big( \varphi+\Delta\varphi ,\, \xi+\Delta\xi \big)
    \ &=\ e^{\xi+\Delta\xi} \begin{pmatrix} \cos(\varphi+\Delta\varphi) \\ \sin(\varphi+\Delta\varphi) \end{pmatrix} \notag \\
    \ &=\ e^{\Delta\xi}
        \begin{pmatrix} \cos(\Delta\varphi)\mkern-8mu & -\sin(\Delta\varphi) \\ \sin(\Delta\varphi)\mkern-8mu & \phantom{-} \cos(\Delta\varphi) \end{pmatrix}
        e^\xi \begin{pmatrix} \cos(\varphi) \\ \sin(\varphi) \end{pmatrix} \notag \\
    \ &=\ e^{\Delta\xi}
        \begin{pmatrix} \cos(\Delta\varphi)\mkern-8mu & -\sin(\Delta\varphi) \\ \sin(\Delta\varphi)\mkern-8mu & \phantom{-} \cos(\Delta\varphi) \end{pmatrix}
        \chi(\varphi,\, \xi) \notag \\
    \ &=:\ (\Delta\varphi,\, \Delta\xi) \,\rhd\, \chi(\varphi,\, \xi)
\end{align}
In terms of a diagram, this means that
\begin{equation}
\begin{tikzcd}[row sep=3.5em, column sep=7.em]
    \R^2
        \arrow[r, "(\Delta\varphi {,}\, \Delta\xi) +"]
        \arrow[d, "\chi\,"']
    & \R^2
        \arrow[d, "\,\chi"]
    \\
    \R^2 \backslash \{0\}
        \arrow[r, "(\Delta\varphi {,}\, \Delta\xi) \rhd"']
    & \R^2 \backslash \{0\}
\end{tikzcd}
\end{equation}
commutes for arbitrary translations.
Together with the translation equivariance of conventional convolutions on $\R^2$, this implies that rotated and scaled input feature maps on $\R^2 \backslash \{0\}$ will lead to rotated and scaled output feature maps on $\R^2 \backslash \{0\}$, i.e. the ${\SO2 \!\times\! \Scale}$-equivariance of the convolution on $\R^2 \backslash \{0\}$.
More details on this viewpoint are found in~\cite{esteves2017polar} and~\cite{blatt1994canonical}.

We will now revisit this convolution operation and its properties from the viewpoint of coordinate free $\GM$-convolutions.
To do so, we consider an atlas of charts that are consistent with the log-polar coordinates, and discuss the induced $\{e\}$-structure, gauges, Riemannian metric, geodesics and parallel transport that it implies.
The claimed ${\SO2 \!\times\! \Scale}$-equivariance follows immediately from the $\IsomeM$-equivariance of $\GM$-convolutions.
For notational convenience, we will again identify $\Euc_2 \backslash \{0\}$ via some choice of Cartesian coordinates with $\R^2 \backslash \{0\}$.

As the restriction $\tilde{\chi}: [0,2\pi) \times \R \to \R^2 \backslash \{0\}$ of the log-polar coordinates $\chi$ to non-redundant angles is bijective and continuous, one might be tempted to take its inverse as a coordinate chart.
This is, however, not possible, since $\tilde{\chi}$ is not a homeomorphism, as required for charts.
Instead, we consider an atlas consisting of two charts that are defined in terms of restrictions of $\chi$ and that cover $\R^2 \backslash \{0\}$.
One particular choice is to define chart codomains as open sets $V^A = (0,2\pi) \times \R$ and $V^B = (-\epsilon,\epsilon) \times \R$ for some $0< \epsilon <\pi$ and, for $X=A,B$, define charts on $U^X = \chi(V^X)$ as $x^X := \big( \chi\big|_{V^X} \big)^{-1} : U^X \to V^X$.
Intuitively, this atlas achieves the same as the naive attempt to define charts as the inverse $\tilde{\chi}$.
The important difference is, however, that the charts are diffeomorphic, which is necessary to assure the smoothness of all operations.

As usual, these charts induce local frame fields and bundle trivializations on $U^A$ and $U^B$, respectively.
It is easy to see that the transition maps $g_p^{BA} = \frac{\partial x^B}{\partial x^A} |_{x^A(p)}$ on $U^A \cap U^B$ are trivial, which implies that the union of the frame fields defines a smooth $\{e\}$-structure $\eM$ on $\R^2 \backslash \{0\}$.
These coordinate bases, which are in the literature often denoted as $\big[ \frac{\partial}{\partial \varphi} ,\, \frac{\partial}{\partial \xi} \big]$, are shown in Fig.~\ref{fig:G_structure_R2_no_origin_logpolar} (left).
Our calculation in Eq.~\eqref{eq:log_polar_translation_equivariance} above implies that the induced $\{e\}$-structure is ${\SO2 \!\times\! \Scale}$-invariant.

The charts induce furthermore a Riemannian metric, which differs from the usual Euclidean metric on $\R^2 \backslash \{0\}$.
It is defined as the pullback of the Euclidean metric $\langle \cdot,\cdot \rangle_{\R^2}$ in the charts' codomains, and is therefore pointwise given by
\begin{align}
    \eta_p(v,w) \,:=\, \big\langle \hat{d}x^X_p(v) \,,\, \hat{d}x^X_p(w) \big\rangle_{\R^2} \,,
\end{align}
where $v,w \in \TpM$ and $X$ denotes either chart with $p\in U^X$.
The chart induced $\{e\}$-structure consists by construction of \emph{frames that are orthonormal w.r.t. this chart induced metric}, even though these \emph{frames grow with the radius when measured relative to the standard Euclidean metric}.
The Levi-Civita connection of the induced metric differs from the usual Euclidean connection and implies therefore alternative geodesics and parallel transporters.
As the metric is pulled back via the charts, the geodesics correspond to straight lines in the charts' codomains -- an example are the coordinate lines on $\R^2 \backslash \{0\}$ in Fig.~\ref{fig:G_structure_R2_no_origin_logpolar}.
The parallel transport corresponds to the usual transport in the charts' codomains as well, which implies that it keeps transported vectors in a fixed angle to the coordinate lines on $\R^2 \backslash \{0\}$; cf. footnote~\ref{footnote:punctured_Euclidean_transport}.
Note that this is the same transport as already discussed above in the models corresponding to Figs.~\ref{fig:G_structures_R2_no_origin}, where it was \emph{not} the transport corresponding to the Levi-Civita connection since these models assumed the standard metric on $\R^2 \backslash \{0\}$ instead of the chart induced metric.

The $\{e\}$-structure preserving isometries $\IsomeM \cong {\SO2 \!\times\! \Scale}$ \emph{relative to the chart induced metric} are given by rotations and rescaling of the $\{e\}$-structure \emph{relative to the usual Euclidean metric}.
Theorem~\ref{thm:isom_equiv_GM_conv} implies the ${\SO2 \!\times\! \Scale}$-equivariance of the corresponding $\GM$-convolution -- which recovers the statement made by \citet{esteves2017polar} in our theory.
As stated above, the fact that the metric is induced via the charts means that all operations reduce to the usual Euclidean operations when being expressed in the chart.
The $\GM$-convolution is therefore best implemented via a conventional convolution on the chart, as proposed by \citet{esteves2017polar}.

Note that the ${\SO2 \!\times\! \Scale}$-equivariance of the $\GM$-convolution is easily extended to ${\O2 \!\times\! \Scale}$-equivariance, which includes reflections.
This is implemented by performing a reflection equivariant convolution in the chart, which corresponds to the $\Flip$-structure shown in Fig.~\ref{fig:G_structure_R2_3}.
On $\R^2 \backslash \{0\}$, this implies a $\Flip$-structure that looks similar to that in Fig.~\ref{fig:G_structure_R2_no_origin_O2} above, with the difference that the $\Flip$-structure is additionally invariant under a global rescaling.

%% file: chapters/103_polar_Euc3.tex

\subsubsection*{Global rotation equivariance on $\fakebold{\Euc}_{\boldsymbol{3}} \fakebold{\backslash} \bm{\{0\}}$}
\label{sec:punctured_euclidean_3dim}

The ideas presented above can be generalized to the three-dimensional setting, i.e. to the punctured Euclidean space $\Euc_3 \backslash \{0\}$.
Globally rotation equivariant $\GM$-convolutions correspond here to $G$-structures that are invariant under $\SO3$ rotations around the origin.
While the radial dependency of such $G$-structures is left unconstrained, the demand for rotational invariance imposes a constraint on their form over spherical shells at fixed radii, which are the orbits of the action of~$\SO3$ on $\Euc_3 \backslash \{0\}$.
The fact that the sphere $S^2 = \SO3/\SO2$ is a homogeneous space of $\SO3$ with stabilizer subgroups isomorphic to $\SO2$ implies that the structure group of an $\SO3$-invariant $G$-structure can not be reduced further than $G=\SO2$; see Fig.~\ref{fig:G_structure_S2_1}.
We are therefore essentially considering spherical CNNs with an additional radial dimension.
For a review on spherical CNNs we refer the reader forward to Section~\ref{sec:instantiations_spherical}.

\begin{figure}
    \centering
    \includegraphics[width=1.\textwidth]{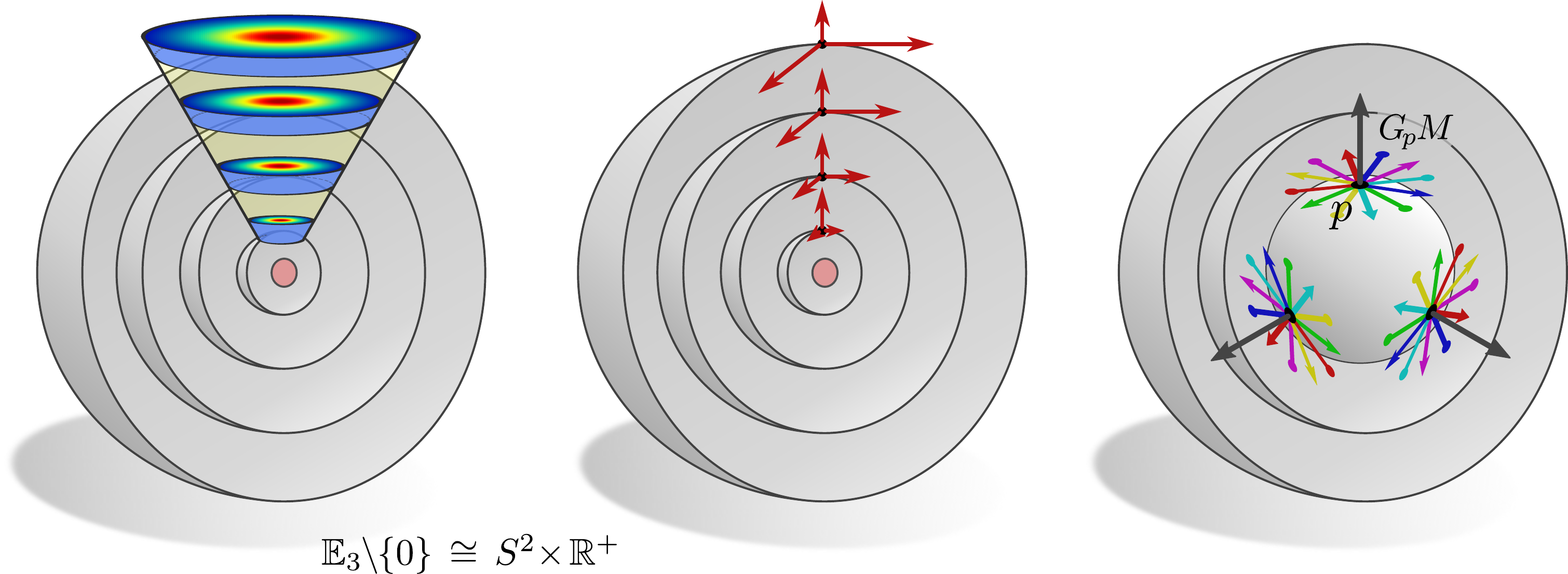}
    \hfill
    \caption{\small
        The $G$-structure that was implicitly assumed by~\citet{ramasinghe2019representation} can be deduced from the weight sharing scheme.
        \emph{Left:}
            Weight sharing of (isotropic) convolution kernels over $\Euc_3 \backslash \{0\} \cong S^2 \times \R^+$ as proposed in~\cite{ramasinghe2019representation}.
            The kernels are defined to cover the same solid angle, independent of the distance from the origin, such that their diameter grows linearly with this distance.
            The kernels' extent in radial direction is independent from the distance from the origin..
        \emph{Middle:}
            In our theory, kernels are shared relative to reference frames of the $G$-structure.
            To recover the proposed weight sharing scheme, $\GM$ needs to consist of frames whose axes in angular direction grow linearly with the radial distance from the origin, while the axes in radial directions need to keep their size fixed (both relative to the standard Euclidean metric).
            Such frames imply an alternative Riemannian metric on $\Euc_3 \backslash \{0\}$.
        \emph{Right:}
            As the resulting $\GM$-convolution should be $\SO3$-equivariant, the $G$-structure is required to be invariant under rotations around the origin.
            This requires (at least) an $\SO2$-structure, whose restriction to one spherical shell is shown in the right part of the figure.
            Compare this to the $\SO3$-invariant $\SO2$-structure of spherical CNNs in Fig.~\ref{fig:G_structure_S2_1}.
    }
    \label{fig:G_structure_R3_no_origin}
\end{figure}

\begin{wrapfigure}[13]{r}{0.25\textwidth}
    \vspace*{-3.ex}
    \centering
    \includegraphics[width=.8\linewidth]{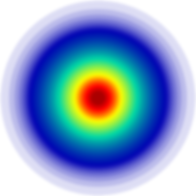}%
    \caption{\small
        A \emph{zonal} (isotropic) kernel is simultaneously $\SO2$- and $\O2$-steerable;
        cf. Eqs.~\eqref{eq:Euc3_punctured_SO2_constraint} and~\eqref{eq:Euc3_punctured_O2_constraint}.
        }
    \label{fig:zonal_kernel}
\end{wrapfigure}

\citet{ramasinghe2019representation} identified this situation and designed $\SO3$-equivariant convolutions on $\Euc_3 \backslash \{0\}$.
Before coming to our classification as $\GM$-convolution, listed in  row (29) of Table~\ref{tab:network_instantiations}, we briefly review the authors' formulation and implementation.
Their implementation is based on spherical CNNs with the addition that
1) kernels extend in the radial direction and
2) are shared over shells at different radii; see Fig.~\ref{fig:G_structure_R3_no_origin} (left).
As commonly done for spherical CNNs, the angular dependency of the kernels is encoded via their Fourier spectrum on~$S^2$, that is, in terms of spherical harmonics expansion coefficients.
The sharing of these expansion coefficients implies that the shared kernels cover the same solid angle for all radii, implying that the \emph{kernels dilate in angular direction linearly with the radius}.%
\footnote{
    The dilation is here measured relative to the standard Euclidean metric of $\Euc_3 \backslash \{0\}$.
}
In the discretized implementation, the spherical shells are located at equidistant radii -- which implies that the \emph{kernels do not dilate in radial direction}.
From these insights we infer the specific $G$-structure that the model assumes below.
The kernels themselves are constrained such that they are invariant under $\SO2$ rotations around the radial axis through their center, which is often referred to as \emph{zonal kernels};
see Fig.~\ref{fig:zonal_kernel} and~\cite{esteves2018zonalSpherical}.
As proven in~\cite{esteves2018zonalSpherical} and~\cite{ramasinghe2019representation}, the convolution with such kernels is $\SO3$-equivariant.
That this is the case is intuitively clear since rotations of the spherical shells have $\SO2$ as stabilizer subgroup, w.r.t. which the zonal kernels are invariant.
As we will argue below, the model is actually $\O3$-equivariant, that is, additionally equivariant under reflections.

To recover this model from the viewpoint of $\GM$-convolutions, we need to determine the corresponding $G$-structure on $M = \Euc_3 \backslash \{0\}$.
As stated above, the $\SO3$-equivariance of the model requires the $G$-structure to be invariant under the action of $\SO3$ but does not constrain their radial variation.
To infer this radial dependency of the $G$-structure, recall that we defined convolutional weight sharing at $p\in M$ as aligning the template kernel $K: \R^3 \to \R^{\cout\times\cin}$ relative to some (arbitrary) frame in $\GpM$ of the tangent spaces $\TpM$.
The kernel sharing considered by \citet{ramasinghe2019representation} lets us therefore draw conclusions about the implicitly considered $G$-structure.
The authors share kernels such that their area tangent to the spherical shells extends with growing distance from the origin (they cover the same solid angle at each radius) while their radial thickness remains constant.
Fig.~\ref{fig:G_structure_R3_no_origin} (left) shows this radial variation of the shared kernels while
Fig.~\ref{fig:G_structure_R3_no_origin} (middle) shows the corresponding scaling of exemplary reference frames.
Together with the required $\SO3$-invariance of the $G$-structure, this implies (at least) an $\SO2$-structure, whose restriction to one spherical shell is visualized in Fig.~\ref{fig:G_structure_R3_no_origin} (right).%
\footnote{
    The two-dimensional analog would look similar to the $G$-structure in Fig.~\ref{fig:G_structure_R2_no_origin_logpolar} but with all frame vectors in radial direction having unit norm (relative to the Euclidean metric).
}
The considered metric follows from this $G$-structure, since its frames define the relevant notion of orthonormality.
Note that this metric differs from the usual Euclidean metric.

By construction, we have rotations $\IsomGM = \SO3$ as $G$-structure preserving isometries.
$\GM$-convolutions defined by this $G$-structure, which may differ in their input and output field type, will therefore (by Theorem~\ref{thm:isom_equiv_GM_conv}) be rotation equivariant.
The specific $\GM$-convolution assumed by \citet{ramasinghe2019representation}, i.e. the assumed field types, can be deduced from the fact that the authors assume zonal kernels:
such kernels arise naturally when considering \emph{scalar fields}, i.e. trivial field representations, since the kernel constraint, Eq.~\eqref{eq:kernel_constraint}, becomes in this case 
\begin{align}\label{eq:Euc3_punctured_SO2_constraint}
    K(g\mkern1mu \mathscr{v}) = K(\mathscr{v}) \quad\ \forall\ \mathscr{v}\in\R^3,\ g\in\SO2 \,,
\end{align}
enforcing isotropic (zonal) kernels.%
\footnote{
    Kernels which map between ``scalar fields'', i.e. fields that transform according to the trivial representation of~$G$, are always $G$-invariant.
    For $G=\SO2$, this implies isotropic (zonal) kernels, while $G=\Flip$ implies the reflection invariant kernels in the upper left entry of Table~\ref{tab:reflection_steerable_kernels}.
}

As a variation of the model, one could consider the $\O2$-structure that follows from the $\SO2$-structure by adding reflected reference frames (reflecting over an arbitrary axis within the planes tangent to the spherical shells, keeping the radial frame vectors still pointing outwards).%
\footnote{
    This $\O2$-structure is the counterpart of the $\O1 = \Flip$-structure in Fig.~\ref{fig:G_structure_R2_no_origin_O2} for $d=3$ instead of $d=2$.
}
In this case one has $G$-structure preserving isometries $\IsomGM = \O3$ that consist of global rotations and reflections around the origin, and therefore $\O3$-equivariant $\GM$-convolutions.
An interesting special case in the current context is that of $\GM$-convolutions that map between scalar fields, for which the kernel constraint reads
\begin{align}\label{eq:Euc3_punctured_O2_constraint}
    K(g\mkern1mu \mathscr{v}) = K(\mathscr{v}) \quad\ \forall\ \ \mathscr{v}\in\R^3,\ g\in\O2 \,.
\end{align}
This seems like a stronger constraint than that in Eq.~\eqref{eq:Euc3_punctured_SO2_constraint} above:
instead of only demanding kernels to be rotationally invariant, it requires them additionally to be invariant under reflections.
However, since rotation invariant kernels are already invariant under reflections, this leads again to zonal kernels, and therefore exactly the same kernel space as for~$\SO2$.%
\footnote{
    More formally, we are searching for kernels that satisfy $K(g\mkern1mu \mathscr{v}) = K(\mathscr{v})\ \ \forall g\in G$, that is, which are invariant on the \emph{orbits} $G.\mathscr{v} = \{g\mkern1mu \mathscr{v} \,|\, g\in G\} \in G \backslash \R^d$ of points $\mathscr{v}$ in $\R^d$.
    As the orbits $\O2.\mathscr{v} = \SO2.\mathscr{v}$ agree for any $\mathscr{v}\in\R^3$, the resulting kernel spaces are the same.
}
This implies that the model by~\citet{ramasinghe2019representation} is actually not only $\SO3$-equivariant, as claimed by the authors, but more generally $\O3$-equivariant, which justifies our classification in row (29) of Table~\ref{tab:network_instantiations},
Note that this is a special case that applies only for scalar fields -- the spaces of $\SO2$- and $\O2$-steerable kernels differ for general group representations.

How does the model by~\citet{ramasinghe2019representation} relate to that of~\citet{esteves2017polar}, which relies on the $G$-structure shown in Fig.~\ref{fig:G_structure_R2_no_origin_logpolar}?
A key difference between the two approaches is that the $G$-structure in Fig.~\ref{fig:G_structure_R2_no_origin_logpolar} consists of frames whose outward pointing axes grow with the radial distance from the origin, which is not the case for the $G$-structure in Fig.~\ref{fig:G_structure_R3_no_origin}.
If we modify the latter to consist of frames whose radial axes grow linearly with the frames' distance from the origin, one would have $\IsomGM = \SO3 \!\times \Scale$ (instead of $\IsomGM = \SO3$).
The corresponding $\GM$-convolution would therefore additionally be scale equivariant.
In an implementation, this could easily be realized by spacing the discrete spherical shells considered by~\citet{ramasinghe2019representation} exponentially instead of uniformly (corresponding to a uniform spacing of the logarithmized radius).

Lastly, we briefly discuss the convolution by~\citet{boomsma2017spherical} that is listed in row (30) of Table~\ref{tab:network_instantiations}.
It~relies on a radial projection of the signal on spherical shells to a circumscribing cube.
To define a convolution on the cube, the authors cut it open at some of its edges and flatten it out; see Fig.~2 in their work.
Subsequently, they perform a conventional two-dimensional convolution on the flattened cube faces.
Extending this operation with a third, radial dimension, defines a convolution on $\Euc_3 \backslash \{0\}$.
As the radial shells are in the discretized implementation again spaced equidistantly, this operation corresponds to a $\GM$-convolution on an $\{e\}$-structure that varies radially as shown in Fig.~\ref{fig:G_structure_R3_no_origin}.
The projection from the spherical shells to the cube implies a distortion of the frames on each of the cubes faces, and thus to a distortion of the metric on the spherical shells.
The $\{e\}$-structure is discontinuous at most of the cuts and does therefore not allow the convolution to preserve the continuity of feature fields.
Since $S^2$ is not parallelizable, this issue can not be resolved without assuming a non-trivial structure group~$G$.
The $\{e\}$-structure as a whole is not preserved by any isometries, implying that the model's global equivariance group $\IsomeM = \{e\}$ is trivial.
However, as the restriction of the $\{e\}$-structure to the four ``vertical'' faces of the cube is invariant under rotations by multiples of $\pi/2$, the model is in practice partially equivariant w.r.t. global $\C4$-rotations around the vertical axis.
For datasets whose samples are centered around the origin $\{0\}$ and are rotationally symmetric in distribution, this property is empirically shown to lead to an improved performance in comparison to conventional convolutions on $\Euc_3$.
The authors are furthermore investigating the effect of different weight sharing schemes over the radial dimension, finding that full weight sharing works best in practice.

%% file: chapters/110_spherical_intro.tex

\section{Spherical coordinate independent CNNs}
\label{sec:instantiations_spherical}

\begin{figure}
    \centering
    \includegraphics[width=1.\textwidth]{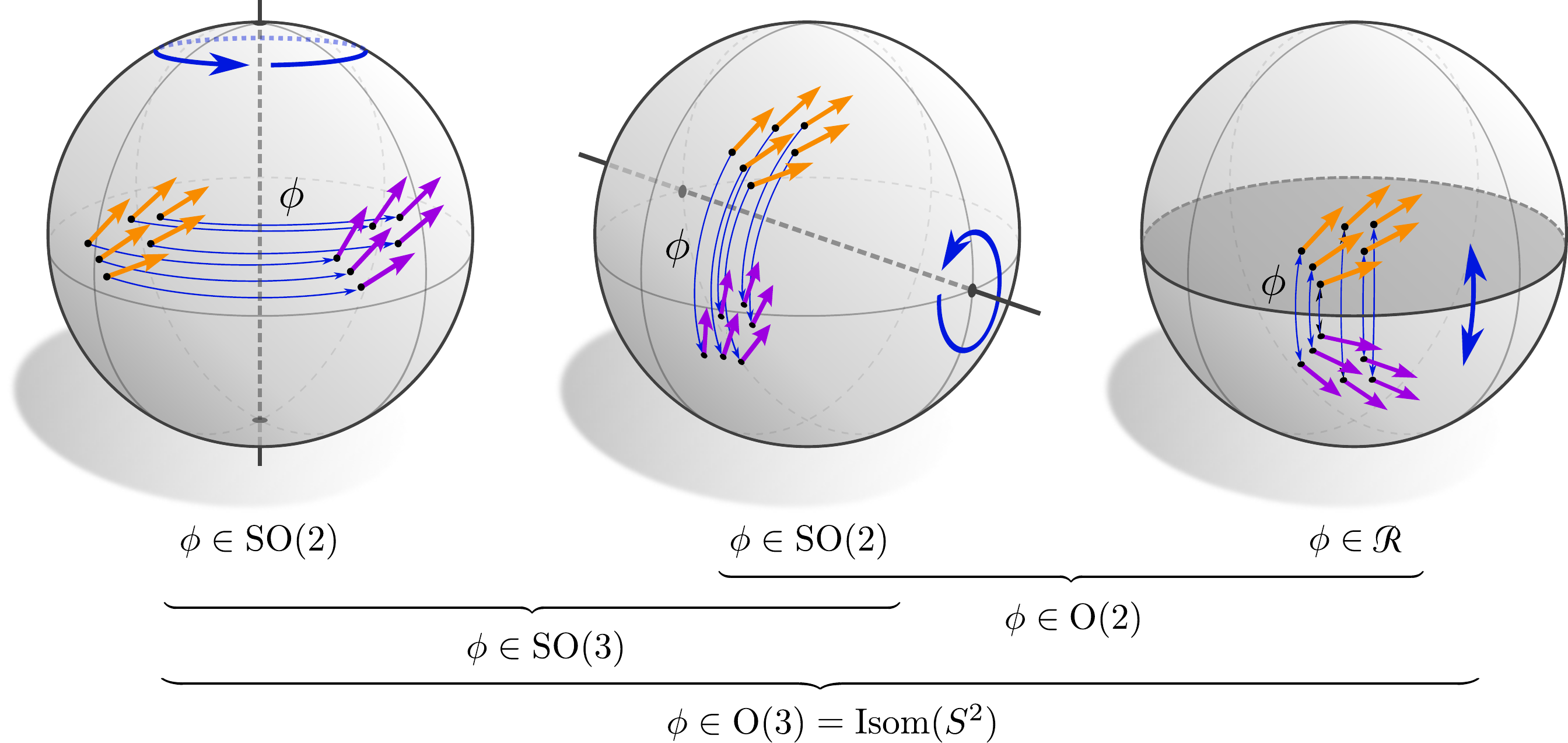}
    \caption{\small
        Visualization of the 2-sphere's isometry group $\Isom(S^2) = \O3$ and its various subgroups.
        The isometry group can be thought of as being composed of the orientation preserving rotations in $\Isom_+(S^2) = \SO3$ and reflections $\Flip$ via the direct product $\O3 = \SO3 \times \Flip$.
        $\SO3$, in turn, is generated by $\SO2$ rotations around any two non-parallel axes, which is used in the Euler angle parametrization.
        See the main text for more relevant subgroups and their relations.
    }
    \label{fig:isometries_sphere}
\end{figure}

Beyond convolutions on Euclidean spaces, convolutions on the 2-sphere $S^2$ are of great practical relevance.
Applications include omnidirectional vision tasks,  global weather forecasting, or the analysis of the cosmic microwave background.
Instead of being translation equivariant, spherical convolutions are typically required to be rotation equivariant.
The isometry group $\Isom(S^2) = \O3$ of the sphere and its decomposition in the most relevant subgroups is visualized in Fig.~\ref{fig:isometries_sphere}.

\begin{figure}
    \centering
    \begin{subfigure}[b]{0.48\textwidth}
        \centering
        \includegraphics[width=.66\textwidth]{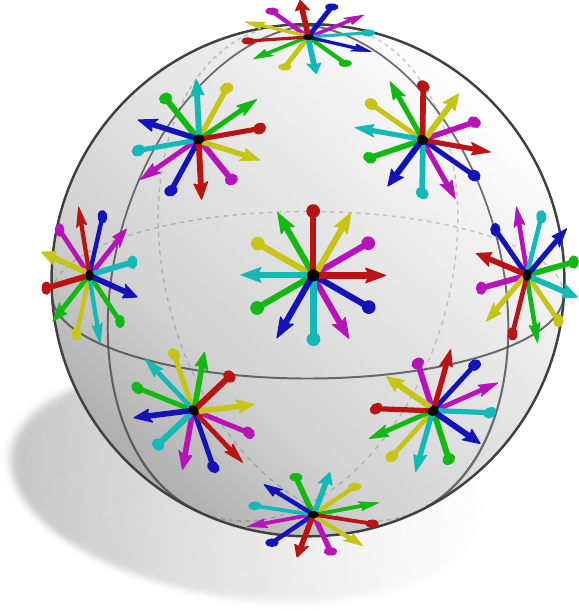}
        \captionsetup{format=hang, width=.9\textwidth}
        \caption{\small
            $\SO2$-structure $\SOM$ on the 2-sphere $M=S^2$, preserved by general three-dimensional rotations in $\IsomSOM = \SO3$.
        }
        \label{fig:G_structure_S2_1}
    \end{subfigure}
    \hfill
    \begin{subfigure}[b]{0.48\textwidth}
        \centering
        \includegraphics[width=.66\textwidth]{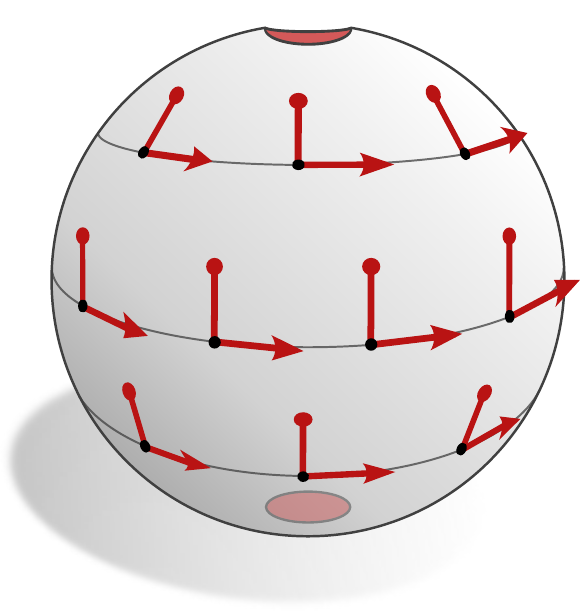}
        \captionsetup{format=hang, width=.9\textwidth}
        \caption{\small
            $\{e\}$-structure $\eM$ on a punctured 2-sphere $M = S^2 \backslash \{n,s\}$, preserved by azimuthal rotations in $\IsomeM = \SO2$.
        }
        \label{fig:G_structure_S2_2}
    \end{subfigure}
    \vspace*{2ex}
    \caption{\small
        Common $G$-structures underlying spherical CNNs.
        Topological obstructions prevent a reduction of the 2-sphere's structure group beyond $G=\SO2$.
        Fig.~\ref{fig:G_structure_S2_1} shows the standard $\SO2$-structure on $S^2$, which is in agreement with the embedding metric (Eq.~\eqref{eq:spherical_embedding_metric_explicit}) induced from the inner product of $\R^3$.
        It is invariant under rotations in $\protect\IsomSOM = \SO3$, implying the rotation equivariance of the corresponding $\GM$-convolution.
        Note that the fibers $\GpM$ and $\GqM$ at different points $p$ and $q$ are isomorphic but not canonically so -- frame colors in the visualization seem to suggest such an isomorphism, however, they are randomly chosen and carry no meaning.
        Fig.~\ref{fig:G_structure_S2_2} shows a sphere that is punctured at two antipodal poles.
        This turns the sphere into a topological cylinder $S^2 \backslash \{n,s\} \cong S^1\times(-\frac{\pi}{2},\frac{\pi}{2})$ with sphere like metric -- which allows for a complete reduction to a trivial structure group.
        The figure shows the most prominent choice of $\{e\}$-structure, which corresponds to orthonormal frames that are aligned with the coordinate grid of spherical coordinates; cf. Fig.~\ref{fig:spherical_equirectangular_1}.
        As this $\{e\}$-structure is invariant under azimuthal rotations around the polar axis, the corresponding $\GM$-convolutions are $\protect\IsomeM = \SO2$-equivariant.
        Note that the puncturing of the sphere is just a means of hiding the models' discontinuity at the poles.
    }
    \label{fig:G_structures_S2_main}
\end{figure}

A major difference between Euclidean spaces $\Euc_d$ and the sphere $S^2$ is that the latter is not parallelizable, i.e. does not allow for a global, continuous frame field.
Reductions of the structure group beyond $G=\SO2$ are topologically obstructed, which means that spherical convolutions require at least $\SO2$-steerable kernels if they should preserve the continuity of feature fields.
The corresponding $\SO2$-structure, which is fully determined by the sphere's metric and orientation, is shown in Fig.~\ref{fig:G_structure_S2_1}.
$\GM$-convolutions on this globally rotation invariant $G$-structure are guaranteed to be $\IsomSOM = \SO3$-equivariant.

Despite the unavoidable topological obstruction, many authors proposed spherical CNNs that do not apply $\SO2$-steerable kernels.
The most prevalent choice of $\{e\}$-structure corresponding to such convolutions is the frame field shown in Fig.~\ref{fig:G_structure_S2_2}, whose orthonormal reference frames (Eq.~\eqref{eq:spherical_e_structure_frames}) are aligned with the coordinate grid of spherical coordinates.
Note that this frame field comes with singularities at the poles, where the convolutions becomes discontinuous.
To reconcile such models with our theory, in particular the smoothness assumption of the $G$-structures, they need to be described as convolutions on a topological cylinder with sphere-like metric.
The isometry group of this punctured sphere $S^2 \backslash \{n,s\}$ without poles $n,s \in S^2$ is the subgroup $\O2$ (Fig.~\ref{fig:isometries_sphere} (middle and right)) of the sphere's full isometry group $\O3$.
The visualized $\{e\}$-structure is preserved by azimuthal rotations in $\IsomeM = \SO2$, i.e. rotations around the axis through the poles.

From an engineering perspective, both approaches have their justification:
fully isotropic applications like the analysis of the cosmic microwave background require fully $\SO3$-equivariant models on~$S^2$.
Learning tasks that come with a preferred rotation axis, which is for instance the case for the earth or panoramic images with a distinguished ``up'' and ``down'' direction, might benefit from the additional geometric information encoded in the $\{e\}$-structure.
Empirical results suggest that it is in such cases often useful to work with a combination of both approaches:
initial layers with fully equivariant convolutions can exploit local symmetries in the data, while subsequent layers with only azimuthal equivariance can learn to discriminate based on the preferred axis; see Section~2.7 in~\cite{3d_steerableCNNs}.

\etocsettocdepth{3}
\etocsettocstyle{}{} 
\localtableofcontents

We start by describing the sphere's geometry in Section~\ref{sec:sphere_geometry}.
Section~\ref{sec:spherical_CNNs_fully_equivariant} discusses fully $\SO3$ and $\O3$-equivariant spherical $\GM$-convolutions, which rely on $\SO2$ or $\O2$-structures as shown in Fig.~\ref{fig:G_structure_S2_1}.
Globally $\SO2$ and $\O2$-equivariant spherical CNNs, corresponding to the $\{e\}$-structure in Fig.~\ref{fig:G_structure_S2_2} or the corresponding $\Flip$-structure, respectively, are reviewed in Section~\ref{sec:spherical_CNNs_azimuthal_equivariant}.
Section~\ref{sec:spherical_CNNs_icosahedral} focuses on icosahedral approximations of spherical convolutions, which allow for compute-efficient implementations since the icosahedron is piecewise flat and admits regular sampling grids; see Fig.~\ref{fig:ico_neighborhoods}.
The $\SO2$-structure and $\{e\}$-structure in Figs.~\ref{fig:G_structure_S2_1} and~\ref{fig:G_structure_S2_2} are hereby approximated by the $\C6$-structure and the $\{e\}$-structures in Figs.~\ref{fig:G_structure_ico_3} and~\ref{fig:G_structure_ico_1} or~\ref{fig:G_structure_ico_2}, respectively.

%% file: chapters/111_spherical_geometry.tex

\subsection%
    [Geometry of the 2-sphere \texorpdfstring{$S^2$}{S2}]%
    {Geometry of the 2-sphere $\boldsymbol{S^2}$}
\label{sec:sphere_geometry}

As a basis for our discussion of spherical CNNs, this section discusses the differential geometry of the (unit) sphere~$M = S^2$.
It is usually defined as the subset
\begin{align}
    S^2 \,:=\, \big\{ p\in\Euc_3 \,\big|\, \lVert p\rVert = 1 \big\}
\end{align}
of those points in Euclidean 3-space $\Euc_3$ that have unit distance from the origin.
As an embedded surface, it inherits a Riemannian metric (first fundamental form) from the embedding space $\Euc_3$.
In the following, we will for simplicity model $\Euc_3$ by the vector space $\R^3$.
When interpreting the tangent spaces $\TpM$ literally as those two-dimensional subspaces of $\R^3$
that contain all tangent vectors at $p \in S^2$, then the metric, exponential maps, parallel transporters, frames and gauges can all be expressed in terms of usual vector space operations in $\R^3$.
Before coming to these concrete expressions, which come handy when implementing spherical CNNs,
we discuss some properties of the sphere from a more abstract angle.

The isometry group of the sphere is given by
\begin{align}
    \Isom(S^2) = \O3 \,,
\end{align}
i.e. three-dimensional rotations and reflections, which are visualized in Fig.~\ref{fig:isometries_sphere}.
The action of any isometry $\phi\in\O3$ coincides with its usual action on $\R^3$ via matrix multiplication, restricted to the embedded sphere $S^2 \subset \R^3$.
Note that this yields indeed a well defined action on $S^2$ since $\O3$ consists by definition of all distance and angle preserving linear maps, and thus preserves the sphere.
As the sphere is orientable, it comes with a subgroup of orientation preserving isometries
\begin{align}
    \Isom_+(S^2) = \SO3 \,,
\end{align}
consisting of all three-dimensional rotations.
Further subgroups that are relevant in the deep learning context are the following:
any choice of rotation axis determines a subgroup of two-dimensional rotations, isomorphic to $\SO2$, and all of these subgroups are conjugated to each other.
Similarly, any choice of two-dimensional subspace of $\R^3$ corresponds to a subgroup of reflections over this plane, which is isomorphic to $\Flip$.
The subgroups of two-dimensional rotations around two non-parallel rotation axes generate $\SO3$, which relates to the Euler angle parametrization of $\SO3$.
A choice of reflection plane and any rotation axis within this plane generates the semidirect product subgroup $\O2 = \SO2 \rtimes \Flip$.
If the rotation axis is instead chosen to be orthogonal to the reflection plane, the two-dimensional rotations and reflections commute, and generate therefore subgroups isomorphic to the direct product $\SO2 \times \Flip$.
$\O3$ has furthermore discrete subgroups, the most practically relevant of which are the symmetry groups of the platonic solids, for instance of the icosahedron, shown in Fig.~\ref{fig:ico_neighborhoods}.%
\footnote{
    An exhaustive list of all finite subgroups of $\SO3$ can be found at \href{https://ncatlab.org/nlab/show/SO\%283\%29\#finite_subgroups}{nLab}.
}

$\O3$ acts transitively on the sphere, that is, for any two points $p$ and $q$ of $S^2$, there exists at least one isometry $\phi\in\O3$ such that $q = \phi(p)$.
The actions of $\O3$ on $S^2$ are not fixed point free:
any point $p\in S^2$ is stabilized by the subgroup $\Stab{p} \cong \O2 < \O3$, consisting of rotations and reflections around the axis spanned by $p$ in $\R^3$.
Together, these two properties imply that the sphere is a homogeneous space of $\O3$ and is algebraically realized as the quotient space
\begin{align}
    \O3/\O2 \,\cong\, S^2 \,,
\end{align}
which consists of cosets of the form ${\phi\mkern1mu.\mkern-4mu\O2}$.
A similar statement holds for $\SO3$, which has stabilizer subgroups $\Stab{p} \cong \SO2 < \SO3$ and thus
\begin{align}
    \SO3/\SO2 \,\cong\, S^2 \,.
\end{align}
With these relations, Theorem~\ref{thm:GM_conv_homogeneous_equivalence} proves that any $\O3$ or $\SO3$-equivariant kernel field transform on $S^2$ is equivalent to a $\GM$-convolution with $G$ being $\O2$ or $\SO2$, respectively.
This result is in line with the classical viewpoint of group equivariant CNNs on homogeneous spaces~\cite{Cohen2019-generaltheory} -- the precise relation between the two is clarified in Theorem~\ref{thm:spherical_conv_GM_conv} below.
Recall that isometries preserve the Riemannian metric by definition.
That $\O3$ acts transitively on $S^2$ with stabilizer $\O2$ implies therefore that the Riemannian geometry of $S^2$ ``looks similar'' at each point and in each direction and orientation -- $S^2$ is a maximally symmetric space.

As a Riemannian manifold, $S^2$ comes by design with an $\O2$-structure.
A restriction to right-handed frames, which is possible since the sphere is orientable, yields the $\SO2$-structure in Fig.~\ref{fig:G_structure_S2_1}, which is preserved by rotations in $\SO3$.
One can show that these two $G$-structures $\OM$ and $\SOM$ are as principal bundles isomorphic to $\O3$ and $\SO3$, respectively.
The specific isomorphism is hereby given by a choice of frame from the $G$-structure, that is to be identified with the identity group element.

The hairy ball theorem states that no continuous vector field exists on $S^2$, which implies in particular that no (continuous) $\{e\}$-structure can exist.
A reduction of the structure group beyond $\SO2$ requires therefore a change in the topology of the manifold.
For example, puncturing the sphere at an arbitrary point $p\in S$ yields a surface that is homeomorphic to the Euclidean plane, and is therefore parallelizable.%
\footnote{
    This process corresponds for instance to the stereographic projection of the sphere.
}
Puncturing the sphere at two arbitrarily chosen antipodal points, as shown in Fig.~\ref{fig:G_structure_S2_2}, turns the topology of the sphere into that of a cylinder and thus allows for $\{e\}$-structures.
The most common choice of $\{e\}$-structure on the punctured sphere $S^2 \backslash \{n,s\}$ is the $\SO2$-invariant $\{e\}$-structure in Figs.~\ref{fig:G_structure_S2_2} and~\ref{fig:spherical_equirectangular_1}.
Its frames
\begin{align}\label{eq:spherical_e_structure_frames}
    \left[ \frac{\partial}{\partial\theta} ,\; \frac{1}{\cos(\theta)} \frac{\partial}{\partial\varphi} \right]
\end{align}
are aligned with the usual spherical coordinates, which are in physics conventions
(i.e. with $\varphi$ and $\theta$ denoting the azimuthal angle and inclination against the $xy$-plane, respectively)
given by the following surjective, $2\pi$-periodic map:
\begin{align}\label{eq:spherical_coords}
    \chi:\, \big( {\textstyle \minus\frac{\pi}{2}, \frac{\pi}{2} }\big) \times \R \:\to\: S^2 \backslash \{n,s\}
    \,, \quad
    (\theta, \varphi) \,\mapsto
    \begin{pmatrix}
        \cos(\theta) \cos{\varphi} \\
        \cos(\theta) \sin{\varphi} \\
        \sin(\theta)
    \end{pmatrix}
\end{align}
Some $\{e\}$-steerable CNNs are implemented by representing feature fields on $S^2 \backslash \{n,s\}$ in spherical coordinates; see Section~\ref{sec:spherical_CNNs_azimuthal_equivariant} below.
As the coordinate map $\chi$ is $\emph{not}$ isometric, these methods require an alternative metric (or $\{e\}$-structure) on the coordinates ${(\minus \pi/2 ,\; \pi/2) \times \R \subset \R^2}$; see the stretched frames in Fig.~\ref{fig:spherical_equirectangular_1} (right).

Since $S^2$ is compact, it is geodesically complete.
The geodesics are given by the great circles of the sphere, i.e. those circles that correspond to the intersection of the sphere with a plane through the origin of $\R^3$.
The exponential maps $\exp_p(v)$ follow these great circles through $p$ in direction of $v$ for a distance of $\lVert v\rVert$.
Logarithmic maps $\log_p(q)$ are therefore for all points $q \in S^2 \backslash \mkern-1mu\minus\mkern1mu p$, which are not antipodal to $p$, given by the unique vector in the shorter direction along the great circle through $p$ and $q$, with $\lVert\log_p(q)\rVert$ given by the arc-length along this path.
Geodesics between antipodal points $p$ and $-p$ are not unique, such that the logarithmic map does not exist.

\subsubsection*{Explicit geometry of $\boldsymbol{S^2}$ as embedded surface in $\boldsymbol{\mathds{R}^3}$}
As stated above, the tangent spaces of $S^2 \subset \R^3$ are in the classical differential geometry of surfaces defined as two-dimensional subspaces of the embedding space~$\R^3$.
A specific tangent space $\TpM$ at $p\in S^2$ is in this interpretation given by
\begin{align}
    \TpM \,=\, \big\{ v\in\R^3 \,\big|\, \langle p,v \rangle = 0 \big\} \ \subset\ \R^3 \,,
\end{align}
i.e. the space of all vectors that are orthogonal to the surface normal at $p$, which coincides for the sphere with $p$ itself.
Note that, despite being expressed relative to the standard frame of $\R^3$, these tangent vectors are coordinate free object in the sense that they are not described by 2-tuples of coefficients $v^A \in \R^2$ relative to some gauge $\psiTMp^A$ of~$\TpM$.
The identification of tangent spaces with subspaces of the embedding space allows to express many of the abstract algebraic relations in terms of vector space operations on~$\R^3$.
In the remainder of this section we will state such expressions for the metric, exponential and logarithmic maps, frames, gauges, Levi-Civita transporters along geodesics and induced gauge transformations.

By definition, $S^2$ inherits its Riemannian metric from the embedding space.
This induced metric is for any $v,w \in \TpM \subset \R^3$ given by
\begin{align}\label{eq:spherical_embedding_metric_explicit}
    \eta_p(v,w) \,:=\, \langle v,w \rangle_{\R^3} \,,
\end{align}
i.e. the standard inner product of $\R^3$, restricted to $\TpM$.
To reduce clutter, we drop the subscript $\R^3$ in the notation $\langle \cdot,\cdot \rangle_{\R^3}$ in the remainder of this section.

The exponential map $\exp_p$ maps vectors $v\in \TpM$ to points $q = \exp_p(v) \in S^2$ at a distance of $\lVert v\rVert$ along the great circle in direction of~$v$.
Lying on the same great circle, $p$ and $q$ relate via a rotation by an angle of $\alpha = \lVert v\rVert / r = \lVert v\rVert$ around the rotation axis $a = \frac{p\times v}{\lVert p \times v\rVert} = \frac{p\times v}{\lVert v\rVert}$,
where the equations simplify since the sphere has unit radius $r = \lVert p\rVert = 1$ and the vectors $p$ and $v$ are orthogonal in $\R^3$.
Using Rodrigues' rotation formula, $q = p \cos(\alpha) + (a\times p) \sin(\alpha) + a\langle a,p\rangle \big(1- \cos(\alpha) \big)$,
together with the orthogonality $\langle a,p\rangle = 0$ and
$a\times p
 = \frac{1}{\lVert v\rVert} (p\times v) \times p
 = \frac{1}{\lVert v\rVert} \big( \langle p,p\rangle v + \langle p,v\rangle p \big)
 = \frac{v}{\lVert v\rVert}$,
this leads to the explicit expression
\begin{align}\label{eq:sphere_expmap_explicit}
    \exp_p:\, \R^3 \supset \TpM \to S^2 \subset \R^3, \quad v \mapsto \exp_p(v) = p\mkern1mu \cos \big(\lVert v\rVert\big) + \frac{v}{\lVert v\rVert} \sin \big(\lVert v\rVert\big)
\end{align}
for the exponential map.

An explicit expression of the logarithmic map is found along the same line of reasoning:
the norm of $\log_p(q)$, where $q\in {S^2 \backslash \mkern-1mu\minus\mkern1mu p}$, coincides with the rotation angle $\alpha = \arccos\!\big( \langle p,q\rangle \big)$.
Its direction is given by the direction tangent to the great circle, which may be expressed in terms of the normalized projection
$\frac{v}{\lVert v\rVert} = \frac{q - \langle p,q\rangle p}{\lVert q - \langle p,q\rangle p\rVert}$
of~$q$ on~$\TpM$.
Overall, the logarithmic map is therefore instantiated as
\begin{align}\label{eq:sphere_logmap_explicit}
    \log_p:\, S^2\backslash \mkern-1mu\minus\mkern1mu p \to B_{\TpM}(0,\pi), \quad
    q \mapsto \log_p(q) = \arccos\!\big( \mkern-1mu\langle p,q\rangle \mkern-1mu\big) \, \frac{q - \langle p,q\rangle p}{\lVert q - \langle p,q\rangle p\rVert} \,,
\end{align}
where $B_{\TpM}(0,\pi) \subset \TpM \subset \R^3$ denotes the open ball of injectivity-radius $\pi$ around the origin of $\TpM$.

Reference frames on $S^2$ are by definition just $2$-tuples of linearly independent tangent vectors.
When expressing the axes of a reference frame explicitly as vectors in the embedding space $\R^3$, this frame can be identified with the $3\times 2$ rank $2$ matrix
\begin{align}\label{eq:embedding_space_R3_frame}
    \big[ e_1^A,\, e_2^A \big]\ =\ 
    \left[\! \begin{array}{cc}
        e^A_{1,1} & e^A_{2,1} \\
        e^A_{1,2} & e^A_{2,2} \\
        e^A_{1,3} & e^A_{2,3}
    \end{array} \!\right]
    \; =: E^A_p
    \ \ \in\, \R^{3\times2} .
\end{align}
It defines the vector space isomorphism
\begin{align}
    E^A_p = \big[ e_1^A, e_2^A \big]:\, \R^2 \to \TpM,\ \ \ v^A \mapsto E^A_p v^A = v^A_1 e^A_1 + v^A_2 e^A_2
\end{align}
from vector coefficients to coordinate free tangent vectors.
The tangent spaces $\TpM$ are therefore exactly the image of $E^A_p$.

The corresponding gauges $\psiTMp^A: \TpM \to \R^2$ are technically just the inverses of the frames, when being interpreted as maps $E^A_p: \R^2 \to \TpM$.
In contrast, when being interpreted as $3\times 2$ matrices that map $\R^2$ non-surjectively to $\R^3$, $E^A_p$ is non-invertible but only admits a pseudo-inverse
\begin{align}
    \big(E^A_p \big)^+ \,:=\, \big( (E^A_p)^\top E^A_p \big)^{-1} (E^A_p)^\top \ \in\, \R^{2\times 3} .
\end{align}
Geometrically, this matrix acts by
1) projecting vectors in $\R^3$ to the image of $E_p^A$, which is just $E_p^A(\R^2) = \TpM \subset \R^3$, and
2) applying the inverse of the isomorphism $E_p^A: \R^2 \to \TpM$ on this subspace.
This means that the pseudo-inverse is indeed the inverse of $E_p^A$ on the tangent space, implying that the gauge map is given by
\begin{align}
    \psiTMp^A: \TpM \to \R^2, \ \ v \mapsto \big(E_p^A \big)^+ v \,.
\end{align}
Written out, the gauge map acts according to
\begin{align}
    \psiTMp^A(v)\ =&\ 
    \Bigg( \mkern-9mu
    \begin{array}{cc}
        \langle e_1^A, e_1^A \rangle    & \langle e_1^A, e_2^A \rangle \\[4pt]
        \langle e_2^A, e_1^A \rangle    & \langle e_2^A, e_2^A \rangle
    \end{array}
    \mkern-9mu \Bigg)^{\!-1}
    \Bigg( \mkern-9mu
    \begin{array}{cc}
        \langle e_1^A, v \rangle \\[4pt]
        \langle e_2^A, v \rangle
    \end{array}
    \mkern-9mu \Bigg)
    \notag \\
    \ =&\ 
    \frac{1}{
          \langle e_1^A, e_1^A \rangle \langle e_2^A, e_2^A \rangle
        - \langle e_1^A, e_2^A \rangle \langle e_2^A, e_1^A \rangle
    }
    \Bigg( \mkern-9mu
    \begin{array}{cc}
        \phantom{\minus}\langle e_2^A, e_2^A \rangle   &          \minus \langle e_1^A, e_2^A \rangle \\[4pt]
                 \minus \langle e_2^A, e_1^A \rangle   & \phantom{\minus}\langle e_1^A, e_1^A \rangle
    \end{array}
    \mkern-9mu \Bigg)
    \Bigg( \mkern-9mu
    \begin{array}{cc}
        \langle e_1^A, v \rangle \\[4pt]
        \langle e_2^A, v \rangle
    \end{array}
    \mkern-9mu \Bigg) \,.
\end{align}
Note that, in general, $\langle e_i^A, v \rangle \neq v^A_i$.
However, if (and only if) $E^A_p$ is an orthonormal frame, i.e. for $G\leq\O2$, the gauge map is simply given by the projection of the tangent vector on the frame axes:
\begin{align}\label{eq:embedding_gauge_map_orthonormal_frame}
    \psiTMp^A(v)
    \ =\ 
    \big(E^A_p \big)^{\!\top} v
    \ =\ 
    \Bigg( \mkern-9mu
    \begin{array}{cc}
        \langle e_1^A, v \rangle \\[4pt]
        \langle e_2^A, v \rangle
    \end{array}
    \mkern-9mu \Bigg)
    \qquad \textup{for \emph{orthonormal} frames, i.e.}\ \langle e_i^A, e_j^A \rangle = \delta_{ij}
\end{align}

The explicit expression for the coordinate free Levi-Civita transporters \emph{along geodesics} is similar to that of the exponential map, with the difference that Rodrigues' rotation formula is not applied to rotate the source to the target point but tangent vectors between source and target.
Let $\gamma$ be the shortest geodesic between $p\in S^2$ and $q\in {S^2 \backslash \mkern-1mu\minus\mkern1mu p}$.
The rotation from $p$ to $q$ along this geodesic is then given by the axis $a = p\times q$ and angle $\alpha = \arccos\!\big( \langle p,q\rangle \big)$.
In terms of these quantities, the Levi-Civita transport of an embedded tangent vector $v\in \TpM \subset \R^3$ along the geodesic $\gamma$ is given by the rotated vector
\begin{align}\label{eq:sphere_transport_embedded}
    \PTMgamma(v)\ =\ v \cos(\alpha) + (a\times v) \sin(\alpha) + \big(a\langle a,v\rangle\big) \big(1- \cos(\alpha) \big)
\end{align}
in $\TqM \subset \R^3$.
Relative to gauges $\psiTMp^A$ and $\psiTMq^{\widetilde{A}}$ at the start point $q$ and end point $q$ of the geodesic, this transporter is expressed by the group element
\begin{align}\label{eq:sphere_transporter_explicit_in_coords}
    g_\gamma^{A\widetilde{A}}
    \ =\ \psiTMp^A \circ \PTMgamma \circ \big(\psiTMq^{\widetilde{A}}\big)^{-1}
    \ =\ \big(E_p^A\big)^+ \circ \PTMgamma \circ E_q^{\widetilde{A}} \,.
\end{align}

Isometry induced gauge transformations are relative to the explicit reference frames similarly given by the following matrix multiplication:
\begin{align}\label{eq:embedded_sphere_isom_induced_gauge_trafo}
    g_\phi^{A\widetilde{A}}(p)
    \ =\ \psiTMphip^A \circ \phi \circ \big(\psiTMp^{\widetilde{A}}\big)^{-1}
    \ =\ \big(E_{\phi(p)}^A \big)^{\!+} \phi\: E_p^{\widetilde{A}}
\end{align}

%% file: chapters/112_spherical_SO3.tex

\subsection{Fully rotation equivariant spherical CNNs}
\label{sec:spherical_CNNs_fully_equivariant}

This section discusses the fully $\SO3$ or $\O3$-equivariant spherical convolutions that are listed in rows (31-33) of Table~\ref{tab:network_instantiations}.
They can all be understood as specific instances of $\GM$-convolutions on either the $\SO2$-structure in Fig.~\ref{fig:G_structure_S2_1} or the corresponding $\O2$-structure, which is additionally closed under frame reflections.

Instead of organizing this discussion in terms of the considered structure groups and group representations, we assort the models by the theoretical frameworks in which they are developed:
\citet{kicanaoglu2019gaugeSphere} define a pixel grid on the sphere and formulate the convolution directly as $\GM$-convolution, that is, in terms of gauges, steerable kernels and feature vector transporters.
An alternative framework is that of graph convolutions on spherical pixel meshes~\cite{perraudin2018DeepSphere,yang2020rotation}.
Such graph convolutions correspond to $\GM$-convolutions with isotropic kernels.
They map therefore between (directionally insensitive) scalar fields.
Lastly, we come to implementations that consider (steerable) convolution kernels on $S^2$ instead of our kernels on the tangent spaces~\cite{esteves2018zonalSpherical,Cohen2018-S2CNN,kondor2018ClebschGordan,esteves2020spinweighted}.
Theorem~\ref{thm:spherical_kernel_space_iso} proves that such spherical steerable kernels can be identified with $G$-steerable kernels on the tangent spaces, when being expressed in geodesic normal coordinates.
Based on this result, we prove in Theorem~\ref{thm:spherical_conv_GM_conv} that convolutions with spherical kernels are equivalent to our $\GM$-convolutions.
For completeness, we need to mention that such models are typically implemented in the spectral domain.
We do not focus on this viewpoint but refer the interested reader to the review by~\citet{esteves2020theoretical}.

\paragraph{Spherical \textit{GM}-convolutions:}

We start with the spherical CNN by \citet{kicanaoglu2019gaugeSphere} since its formulation agrees precisely with our more general theory when being applied to the spherical geometry.
The authors assume the $\SO2$-structure from Fig.~\ref{fig:G_structure_S2_1}, and therefore $\SO2$-steerable feature fields and convolution kernels.
Feature fields are discretized in terms of feature vectors that are assigned to a sampling grid on the sphere.
While the method is in principle independent from the particular sampling scheme, the authors propose to discretize the spherical geometry by an icosphere mesh.
This mesh is constructed by taking an embedded icosahedron, repeatedly subdividing its faces as shown in Fig.~\ref{fig:ico_neighborhoods}, and finally projecting the grid vertices radially to the sphere, i.e. to unit norm.
The sampled feature fields are numerically represented by a set of coefficient vectors $f^A(p) \in \R^c$ at the grid vertices $p$, which are expressed relative to some arbitrarily chosen right-handed orthonormal frames $\big[e_1^A, e_2^A \big]$ at the vertices.%
\footnote{
    This corresponds to an independent choice of gauge $\psiTMp^{A_p}$ on any open neighborhood $U^{A_p}$ of each vertex~$p$.
}
In practice, the frames are represented by a single tangent vector of unit norm, from which the second frame vector follows uniquely since the frames are right-handed.

To compute the coordinate independent convolution $[K \star f](p)$ from Eq.~\eqref{eq:gauge_conv_coord_expression}, \citet{kicanaoglu2019gaugeSphere} need to contract the $\SO2$-steerable kernel $K$ with the transporter pullback $[\Expspf]^A$ of the feature field $f$, Eq.~\eqref{eq:transporter_pullback_in_coords}.
As usual in deep learning, $K$ is hereby assumed to be compactly supported, such that it covers only a few vertices in a one-ring or two-ring neighborhood $\mathcal{N}_p$ around a center vertex $p$.
In the continuous theory, the transporter pullback takes features from all points $\exp_p (\psiTMp^A)^{-1}(\mathscr{v})$ for $\mathscr{v} \in \R^2$ and transports them back to~$p$.
In practice, the feature fields are only sampled at the grid vertices $q$, which correspond to the tangent vector coefficients $v^A_{pq} = \psiTMp^A \log_p(q) \in \R^2$ relative to gauge~$A$ at vertex~$p$.%
\footnote{
    If the exponential map is not restricted to the injectivity radius, each vertex $q$ is represented by multiple tangent vectors.
    This is in practice no issue since the kernel is assumed to be locally supported within the injectivity radius.
}
The logarithmic maps $\log_p(q)$ are thereby computed as defined in Eq.~\eqref{eq:sphere_logmap_explicit}.
The Levi-Civita transporters $\rho\big( g_{p\leftarrow q}^{A\widetilde{A}}\big)$ along the geodesics from $q$ to $p$ are in principle given by Eq.~\eqref{eq:sphere_transporter_explicit_in_coords}.
Since the frames are all right-handed and orthonormal, and since the transport corresponds to the Levi-Civita connection on $S^2$, the group elements $g_{p\leftarrow q}^{A\widetilde{A}}$ are $\SO2$-valued.
They are therefore fully determined by the angle between the transported first frame axis $\PTMgamma\big(e_1^{\widetilde{A}}\big)$ from $q$ and the first frame axis $e_1^A$ at~$p$.
With these ingredients at hand, the authors propose to approximate the continuous convolution integral by the discrete sum
\begin{align}
    \big[K\star f\big]^A(p)
    \ =\ \int_{\R^2} K(\mathscr{v})\, \big[\!\Expspf]^A(\mathscr{v}) \,\ d\mathscr{v}
    \ \approx\ \sum_{q\in\mathcal{N}_p} K\mkern-1.5mu\big(v^A_{pq}\big)\, \rho\big( g^{A\widetilde{A}}_{p\leftarrow q} \big)\, f^{\widetilde{A}}(q)
\end{align}
over neighboring mesh nodes.
The missing normalization factor can be thought of as being absorbed in the learnable parameters $w_i \in\R$ of the $\SO2$-steerable convolution kernel $K = \sum_i w_i K_i$.
As an alternative to this naive approximation, the authors propose an optimized quadrature integration scheme, which is empirically shown to improve the model's $\SO3$ isometry equivariance.

The model is in Table~\ref{tab:network_instantiations} listed as processing feature fields that transform according to the regular representation of $\SO2$.
In their implementation, \citet{kicanaoglu2019gaugeSphere} consider irrep fields of $\SO2$ in the convolutions.
A change of basis before and after the convolutions transforms these feature fields to regular feature fields, which are then acted on by pointwise nonlinearities like e.g. ReLU.
The infinite-dimensional regular representation of $\SO2$ is hereby approximated by regular representations of discrete cyclic subgroups $\CN$, whose irreps are just the irreps of $\SO2$ up to a bandlimiting frequency of $\lfloor N/2 \rfloor$; see e.g. Appendix~F.2 of~\cite{Weiler2019_E2CNN}.
The change of basis between the representations is in this specific case just the usual discrete Fourier transform.

\paragraph{Spherical graph convolutions:}

The spherical CNNs by \citet{perraudin2018DeepSphere} and \citet{yang2020rotation}, which are listed in row (33) of Table~\ref{tab:network_instantiations}, are based on conventional graph convolutions~\cite{kipf2016semi}.
Pixel meshes on the sphere are hereby interpreted as graphs.
The graph convolutional networks process signals on the sphere by multiplying them with degree $\kappa$ polynomials $\sum_{k=0}^\kappa w_k L^k$ of the graph's Laplacian matrix $L$, where $w_k \in \R$ are trainable parameters.
Since the Laplacian matrix has non-zero entries only for adjacent nodes, the $k$-th order term affects only the $k$-hop neighborhood around each node.
On a regular mesh with unweighted graph edges, the contribution of a neighboring node $q$ to the accumulated feature at $p$ depends only on their graph distance (``radius''), but not on the particular neighbor (``direction'').
The graph convolution applies therefore in such cases \emph{isotropic} kernels on the graph.
The considered pixel graph on the sphere satisfy these properties approximately.
As their \emph{embedding} on the sphere is furthermore such, that the nodes are geodesically approximately equidistant, the topological isotropy of the graph convolution kernels corresponds to their metric isotropy on the sphere.

The isometry group $\O3$ of the sphere induces $\O2$-valued gauge transformation, that is, it acts by moving patterns to a new location and in a new orientation.
Due to the convolutional weight sharing and the isotropy of the kernels, the graph convolutions are trivially isometry equivariant.
As already argued in Eq.~\eqref{eq:Euc3_punctured_O2_constraint}, isotropic kernels are in our framework recovered as $\O2$-steerable kernels that map between \emph{scalar fields}.
The $\O3$-equivariance of the convolution is in our theory explained by the $\O3$-invariance of the sphere's $\O2$-structure.

\paragraph{Spherical convolutions with kernels on $\bm{S^2}$:}

As a homogeneous space, the sphere admits group (or quotient space) convolutions~\cite{Kondor2018-GENERAL} and more general steerable convolutions on homogeneous spaces~\cite{Cohen2019-generaltheory}.%
\footnote{
    A more general review of convolutions on homogeneous spaces is found in Appendix~\ref{apx:homogeneous_conv}.
}
Instead of defining the convolution kernels on the tangent spaces or on graph neighborhoods, these approaches define kernels immediately as matrix-valued function on the sphere, that is, as
\begin{align}\label{eq:spherical_kernel}
    \kappa:\, S^2 \to \R^{\cout\times\cin} \,.
\end{align}
\citet{Cohen2019-generaltheory} showed that these kernels are required to satisfy a symmetry constraint in order to guarantee the equivariance of the convolution.
We argue in the following that such kernels on $S^2$ are equivalent to $G$-steerable kernels on the tangent spaces (Theorem~\ref{thm:spherical_kernel_space_iso}), which implies that the spherical CNNs covered in~\cite{Cohen2019-generaltheory} and~\cite{Kondor2018-GENERAL} can be viewed as $\GM$-convolutions (Theorem~\ref{thm:spherical_conv_GM_conv}).
The identification between the two kinds of kernels is hereby made by pulling the spherical kernels via the exponential map back to the tangent spaces.
Before explaining this operation, we briefly discuss the models proposed in \cite{Cohen2018-S2CNN,esteves2018zonalSpherical,esteves2020spinweighted,kondor2018ClebschGordan} as specific instances of spherical convolutions with spherical kernels.
For a more details on these models, specifically on their formulation in Fourier space, we refer the reader to the comprehensive review by~\citet{esteves2020theoretical}.

We start our discussion with the group convolutional spherical CNN by \citet{Cohen2018-S2CNN}, listed in row (32) of Table~\ref{tab:network_instantiations}.
This model processes stacks of $\cin$ \emph{scalar} fields
\begin{align}
    f:\, S^2 \to \R^{\cin}
\end{align}
on the sphere by matching them with spherical kernels, Eq.~\eqref{eq:spherical_kernel}, in any $\SO3$ transformed pose.
In equations, this operation is defined as
\begin{alignat}{3}
\label{eq:spherical_lifting_conv}
    \big[\kappa \star_{\mkern-2mu S^2}\! f\big](\phi)\ &:=\, \int_{S^2} \kappa(\phi^{-1}(p))\, f(p)\ dp \qquad\quad &&\phi\in\SO3 \,.
\intertext{
Note that the resulting feature map is viewed as a stack of $\cout$ scalar functions on the symmetry group $\SO3$.
Such feature maps of the form $f:\SO3 \to \R^{\cin}$ (with the new number of input channels corresponding to the previous layer's output channels) are processed further by group convolutions of the form
}
\label{eq:spherical_group_conv}
    \big[\kappa \star_{\SO3}\! f\big](\phi)\ &:=\, \int_{\SO3}\! \kappa(\phi^{-1}\omega)\, f(\omega)\ d\omega \qquad\quad &&\phi\in\SO3 \,,
\end{alignat}
where $\kappa: \SO3 \to \R^{\cout\times\cin}$ is now a matrix-valued function on $\SO3$ and $d\omega$ is the Haar measure on~$\SO3$.
From the viewpoint of steerable CNNs on homogeneous spaces~\cite{Cohen2019-generaltheory} and $\GM$-convolutions, scalar functions on $\SO3$ are viewed as feature fields on $S^2 \cong \SO3/\SO2$, that transform according to the \emph{regular representation} of the fibers (stabilizer subgroups)~$\SO2$.
The initial convolution in Eq.~\eqref{eq:spherical_lifting_conv} applies in this interpretation $\SO2$-steerable kernels between scalar and regular fields, while the group convolution in Eq.~\eqref{eq:spherical_group_conv} applies $\SO2$-steerable kernels between regular fields on~$S^2$.

\citet{esteves2018zonalSpherical} apply spherical convolutions as in Eq.~\eqref{eq:spherical_lifting_conv} with the additional assumption that the kernels are \emph{zonal}, that is, invariant under $\SO2$ rotations around the polar axis; cf. Fig.~\ref{fig:zonal_kernel}.
While the integral technically still gives responses in $\SO3$, the kernel symmetry implies that these responses are constant on the fibers $\SO2$ of $\SO3$, when being interpreted as bundle over $S^2$.
The resulting feature fields are therefore identified as scalar fields on $S^2$, which allows for a repeated application of this type of convolution.
Note that the zonal symmetry of the kernel is consistent with the steerability kernel constraint between scalar fields (trivial representations) that we encountered before in Eq.~\eqref{eq:Euc3_punctured_SO2_constraint}.
As already discussed in the previous Section~\ref{sec:punctured_euclidean_3dim}, this constraint is equivalent to the $\O2$-steerability constraint between scalar fields in Eq.~\eqref{eq:Euc3_punctured_O2_constraint}, which implies that the model of \citet{esteves2018zonalSpherical} is actually $\O3$-equivariant.
It is in spirit similar to the spherical graph convolutions discussed above, but is derived from a different viewpoint and is discretized differently in the implementation.

\citet{esteves2020spinweighted} generalize this model from scalar fields to general \emph{spin weighted spherical functions}.
These functions depend not only on the position $p\in S^2$ on the sphere, but in addition on the particular choice of right-handed, orthonormal reference frame at that point.
They are associated to the \emph{irreps} $\rho_s$ of $\SO2$, where the integer $s\in\Z$ is denoted as the functions' spin weight.%
\footnote{
    One can generalize this concept to spin representations, labeled by half-integer spin weights.
}
Their values for the different frames $\SOpM$ of the $\SO2$-structure $\SOM$ are constrained such that gauge transformations of the frame by $g\in\SO2$ lead to a transformation of the function value by $\rho_s(g)$.
In equations, they are therefore defined by%
\footnote{
    A real-valued implementation would instead consider spin weighted functions of the form $\prescript{}{s}f: \SOM \to \R^{\dim(\rho_s)}$, where~$\rho_s$ are the irreps of $\SO2$ over the real numbers.
}
\begin{align}
    \prescript{}{s}f: \SOM \to \Cm
    \ \ \ \textup{such that} \ \ \
    \prescript{}{s}f\big( [e_1,e_2] \lhd g \big) = \rho_s(g) \prescript{}{s}f \big( [e_1,e_2] \big)
    \ \ \ \forall\ [e_1,e_2] \in \SOpM,\ g\in\SO2 \,;\!
\end{align}
see~\cite{boyle2016should} for more details and alternative definitions.
Note the similarity of this symmetry constraint to the equivalence relation
\begin{align}
    \big[ [e_i]_{i=1}^2 \lhd g,\, \mathscr{f} \big]\ \sim_{\rho_s}\ \big[ [e_i]_{i=1}^2,\, \rho_s(g) \mathscr{f} \big]
\end{align}
from Eq.~\eqref{eq:equiv_relation_A}, which is underlying the definition of associated bundles.
Spin weighted spherical functions are indeed equivalent to sections of the associated bundles
$(\SOM \times \Cm)/\!\sim_{\rho_s}$;
see for instance Proposition~1.6.3 in~\cite{wendlLectureNotesBundles2008}.
They appear in our theory simply as $\SO2$-irrep fields, including scalar fields for $s=0$ and vector fields for~$s=1$.
The neural networks proposed by~\citet{esteves2020spinweighted} convolve spin weighted features with spin weighted kernels on the sphere.
This operation corresponds to a convolution with $\SO2$-steerable kernels where $\rhoin$ and $\rhoout$ are irreps.

The models in \cite{Cohen2018-S2CNN,esteves2018zonalSpherical,esteves2020spinweighted} are initially formulated in the spatial domain, i.e. as processing functions on $S^2$ as discussed above.
They are, however, implemented in the spectral domain, which is possible thanks to generalized convolution theorems on $S^2$ and on $\SO3$~\cite{makadia2006rotation,Kondor2018-GENERAL,vilenkin2013representation}.
\citet{kondor2018ClebschGordan} generalize these approaches, proposing a model that is based on learned linear combinations of all feature fields' Fourier modes of the same frequency.
The authors argue that this approach covers the full space of $\SO3$-equivariant linear maps between feature fields on the sphere.
On the other hand, \citet{Cohen2019-generaltheory} show that any such map can in the spatial domain be written as a convolution with $\SO2$-steerable spherical kernels.
A notable property of the model proposed by \citet{kondor2018ClebschGordan} is that it operates fully in Fourier space:
instead of transforming back to the spatial domain and applying pointwise nonlinearities like ReLUs there, as done in the previous approaches, the authors compute the tensor product between all feature fields and decompose them subsequently via the Clebsch-Gordan decomposition back into irreducible features (Fourier modes).
This is computationally beneficial, however, comes at the expense of losing the locality of the nonlinearities.
Certain learning tasks, especially in the natural sciences, might benefit from such nonlinearities since physical interactions are often described by tensor products.

As argued in~\cite{Cohen2019-generaltheory,Cohen2018-intertwiners}, all of these models can be viewed as applying steerable kernels on $S^2$ that map between scalar fields~\cite{esteves2018zonalSpherical}, regular feature fields~\cite{Cohen2018-S2CNN} or irrep fields~\cite{esteves2020spinweighted,kondor2018ClebschGordan}.
In the remainder of this section and Appendix~\ref{apx:spherical_conv_main} we show that they can as well be viewed as $\GM$-convolutions.
The claim that spherical convolutions with steerable kernels on $S^2$ are equivalent to $\GM$-convolutions is thereby made precise in Theorem~\ref{thm:spherical_conv_GM_conv}.
This theorem relies crucially on Theorem~\ref{thm:spherical_kernel_space_iso}, which establishes an isomorphism between the spherical steerable kernels and $G$-steerable kernels on the tangent spaces.

Let $\I$ be any transitive isometry group of the sphere, i.e. $\I=\O3$ or $\I=\SO3$.
\citet{Cohen2019-generaltheory} describe $\I$-equivariant spherical convolutions in terms of $\Stab{n}$-steerable spherical kernels ${\kappa: S^2 \to \R^{\cout\times\cin}}$, where $\Stab{n} < \I$ is the stabilizer subgroup of any point $n\in S^2$, e.g. the north pole.
As these kernels are defined on the sphere, which is topologically distinct from $\R^2$, it is not directly possible to define an isomorphism between them and $G$-steerable kernels.
However, as the south pole $-n$ is a set of measure zero, we can replace the integration domain $S^2$ of the spherical convolutions with $S^2\backslash \mkern-1mu\minus\mkern1mu n$ without changing the result.
With this adaptation, the spherical steerable kernels of \citet{Cohen2019-generaltheory} are defined as
\begin{align}\label{eq:spherical_steerable_kernel_space}
    \mathscr{K}^{\Stab{n}}_{\rhoin\mkern-1mu,\rhoout}
    := \pig\{ \kappa: S^2\backslash \mkern-1mu\minus\mkern1mu n \to \R^{\cout\times\cin}
    \,\pig|\ \kappa\big(\xi (p)\big) = \rhoout\big( g_\xi^{NN}(n) \big) \mkern-1.5mu\cdot\mkern-1.5mu \kappa(p) \mkern-1.5mu\cdot\mkern-1.5mu \rhoin\big( g_\xi^{XP}(p) \big)^{-1}
    \qquad \\ \notag
    \forall\,\ p\in S^2\backslash \mkern-1mu\minus\mkern1mu n,\ \ \xi\in\Stab{n} \pig\} \,,
\end{align}
when being translated to our notation.
Since the kernels are globally defined on the sphere, their values in $\R^{\cout\times\cin}$ are expressed relative to potentially different gauges $N$ at $n$, where the kernel is centered, $P$ at $p\in S^2$, where the kernel contracts a feature $f^P(p) \in \R^\cin$ and $X$ at $\xi(p)$, where this feature moves under the action of $\xi \in \Stab{n}$.
This kernel constraint relates all kernel values that lie on the orbits ${\Stab{n}\mkern-4mu.\mkern1mup} = \{ \xi(p) \,|\, \xi\in\Stab{n} \}$ via their isometry induced gauge transformations $g_\xi^{XP}(p)$ and $g_\xi^{NN}(n)$; see Eqs.~\eqref{cd:pushforward_GM_coord_extended} and~\eqref{cd:pushforward_A_coord}.%
\footnote{
    \citet{Cohen2019-generaltheory} denote the isometry induced gauge transformations by $\operatorname{h}(p,\xi)$ instead of $g_\xi^{XP}(p)$, assuming that the gauges $X$ at $\xi(p)$ and $P$ at $p$ are the same.
    Their definition of $\operatorname{h}(p,\xi)$ is similar to our Eq.~\eqref{eq:pushfwd_section_right_action}.
}
Our equivalent $G$-steerable kernels, where $G \cong \Stab{n}$, is given by
\begin{align}\label{eq:G_steer_kernel_space_open_ball_pi}
    \mathscr{K}^{G,B_{\R^2}(0,\pi)}_{\rhoin\mkern-1mu,\rhoout}
    := \Big\{ K\!: B_{\R^2}(0,\pi) \to \R^{\cout\times\cin} \mkern1.5mu\Big|\,
    K(g\mkern1mu \mathscr{v}) =
    \rhoout(g) \mkern-2mu\cdot\mkern-2mu K(\mathscr{v}) \mkern-2mu\cdot\mkern-2mu \rhoin(g)^{-1} \ \ \ \forall\ \mathscr{v}\in B_{\R^2}(0,\pi),\,\ g\in G \Big\} .
\end{align}
The kernel domain is hereby restricted from $\R^2$ to the open ball $B_{\R^2}(0,\pi) := \{ \mathscr{v}\in\R^2 \,|\, \lVert \mathscr{v}\rVert < \pi \}$ of radius $\pi$ around the origin of $\R^2$ -- which can via the exponential map be identified with $S^2\backslash \mkern-1mu\minus\mkern1mu n$.
Note that $\mathscr{K}^{G,B_{\R^2}(0,\pi)}_{\rhoin\mkern-1mu,\rhoout}$ is well defined since $\Stab{n} \cong G$ contains isometries, implying $G=\O2$ or $G=\SO2$, under whose action $B_{\R^2}(0,\pi)$ is closed.
We furthermore dropped the determinant factor from the more general $G$-steerability constraint in Eq.~\eqref{eq:G-steerable_kernel_space} since ${|\!\det g| = 1}$ for $G\leq\O2$.
Our kernel constraint is considerably simpler than that of \citet{Cohen2019-generaltheory} since it describes the kernel locally relative to a single gauge, instead of globally relative to an atlas of gauges.
Note further that we dropped the smoothness assumption on the kernels, since the smoothness or continuity of feature fields is not discussed by~\citet{Cohen2019-generaltheory}.
This property could easily be added by demanding that the $G$-steerable kernels converge to the same value for $\lVert\mathscr{v}\rVert$ going to $\pi$, corresponding via the exponential map to the south pole.

The spaces of $\Stab{n}$-steerable kernels on $S^2\backslash \mkern-1mu\minus\mkern1mu n$ and $G$-steerable kernels on $B_{\R^2}(0,\pi)$ are isomorphic, that is, their kernels are identified by an invertible map $\Omega$ that respects the kernel constraints:
\begin{equation}
    \begin{tikzcd}[row sep=3.5em, column sep=12.em]
        \mathscr{K}^{G,B_{\R^2}\mkern-1mu(0,\pi)}_{\rhoin\mkern-1mu,\rhoout}
            \arrow[r, bend left=8, shift left=2pt, "\Omega"]
        &
        \mathscr{K}^{\Stab{n}}_{\rhoin\mkern-1mu,\rhoout}
            \arrow[l, bend left=8, shift left=2pt, "\Omega^{-1}"]
    \end{tikzcd}
\end{equation}
This isomorphism (or rather its inverse $\Omega^{-1}$) can be viewed as the analog of the \emph{transporter pullback} of feature fields:
it pulls the kernel values from points $\exp_n\! \big(\psiTMn^N\big)^{\!-1} \mathscr{v}$ in $S^2\backslash \mkern-1mu\minus\mkern1mu n$ back to \emph{geodesic normal coordinates} $\mathscr{v}\in B_{\R^2}(0,\pi)$.
To express the kernel values from all points $p\in S^2\backslash \mkern-1mu\minus\mkern1mu n$ relative to the same gauge, it applies Levi-Civita transporters $\rho\big( g_{n\leftarrow p}^{NP} \big)$ from $p$ along the geodesics to the north pole~$n$.
In addition, it rescales the kernel values by the Riemannian volume element
$\sqrt{\big|\eta_p^{\partial\mkern-2mu/\mkern-2mu\partial\mathscr{v}}\big|}
:= {\sqrt{\big| \!\det\!\big( \eta_p\big( \frac{\partial}{\partial \mathscr{v}_i} \mkern-2mu\big|_p, \frac{\partial}{\partial \mathscr{v}_j} \mkern-2mu\big|_p \big)_{ij} \big)\big|} }$
relative to the geodesic normal coordinate system (coordinate chart)
$\mathscr{v}: S^2\backslash \mkern-1mu\minus\mkern1mu n \to B_{\R^2}(0,\pi),\ \ p \mapsto \mathscr{v}(p) := \psiTMn^N \log_n p$.%
\footnote{
    Note that the coordinate bases
    $\big[ \frac{\partial}{\partial\mathscr{v}_1} \mkern-2mu|_p,\, \frac{\partial}{\partial\mathscr{v}_2} \mkern-2mu|_p \big]$
    that are induced by the geodesic normal coordinates
    $\mathscr{v}: S^2\backslash \mkern-1mu\minus\mkern1mu n \to B_{\R^2}(0,\pi)$
    are for $G\leq\O2$ \emph{not} contained in~$\GM$.
    These bases play no role for the $\GM$-convolution but appear only to correct for the Riemannian volume when integrating in geodesic normal coordinates over the sphere.
}
The following theorem defines and proves the kernel space isomorphism formally.
\begin{thm}[Spherical steerable kernels in geodesic coordinates]
\label{thm:spherical_kernel_space_iso}
    Let $\I$ be any transitive isometry group of $S^2$ and let $\Stab{n}$ be its stabilizer subgroup at the north pole $n\in S^2$.
    Given any choice of gauge $\psiTMn^N$ at this pole, let $G\leq \GL{2}$ be the isomorphic structure group that represents $\Stab{n}$ in coordinates according to
    $\Stab{n} \xrightarrow{\sim} G,\ \xi \mapsto \psiTMn^N \circ \dxiTM \circ \big(\psiTMn^N \big)^{-1}$.
    The space $\mathscr{K}^{\Stab{n}}_{\rhoin\mkern-1mu,\rhoout}$ of $\Stab{n}$-steerable kernels on $S^2\backslash \mkern-1mu\minus\mkern1mu n$ by \citet{Cohen2019-generaltheory} (Eq.~\eqref{eq:spherical_steerable_kernel_space}) is then isomorphic to the space $\mathscr{K}^{G,B_{\R^2}\mkern-1mu(0,\pi)}_{\rhoin\mkern-1mu,\rhoout}$ of $G$-steerable kernels on the open ball $B_{\R^2}(0,\pi)$ (Eq.~\eqref{eq:G_steer_kernel_space_open_ball_pi}).
    The kernel space isomorphism 
    \begin{align}
        \Omega:\ 
        \mathscr{K}^{G,B_{\R^2}\mkern-1mu(0,\pi)}_{\rhoin\mkern-1mu,\rhoout}
        \xrightarrow{\,\sim\,}\,
        \mathscr{K}^{\Stab{n}}_{\rhoin\mkern-1mu,\rhoout}
    \end{align}
    is given by
    \begin{alignat}{4}
    \label{eq:spherical_kernel_space_iso_Omega}
        \Omega(K)\! &:&\,\ S^2 \backslash \mkern-1mu\minus\mkern1mu n \,&\to&\, \R^{\cout\times\cin},
        \quad p \,&\mapsto\,
        \big[\Omega(K)\big](p)\ &:=&\ K\big( \psiTMn^N \log_n p \big)\, \rhoin\big( g_{n\leftarrow p}^{NP} \big)\, \sqrt{\big|\eta_p^{\partial\mkern-2mu/\mkern-2mu\partial\mathscr{v}}\big|}^{\,-1}
    \intertext{
        if the kernel is expressed relative to (potentially independent) gauges $N$ at $n$ and $P$ at~$p$.
        Its inverse is given by
    }
        \Omega^{-1}(\kappa)\! &:&\,\ B_{\R^2}\mkern-1mu(0,\pi) &\to& \R^{\cout\times\cin},
        \quad \mathscr{v} \,&\mapsto\,
        \big[\Omega^{-1}(\kappa)\big](\mathscr{v})\ &:=&\ \kappa\big(\! \exp_n\! \big(\psiTMn^N\big)^{\!-1} \mathscr{v} \big)\, \rhoin\big( g_{n\leftarrow p}^{NP} \big)^{\!-1} \sqrt{\big|\eta_p^{\partial\mkern-2mu/\mkern-2mu\partial\mathscr{v}}\big|} ,
    \end{alignat}
    where we abbreviated $p := \exp_n\! \big(\psiTMn^N\big)^{\!-1} \mathscr{v}$.
\end{thm}
\begin{proof}
    By inserting the two expressions, one can easily see that $\Omega^{-1}$ is a well defined inverse of $\Omega$ since
    $\Omega \circ \Omega^{-1} = \id_{\mathscr{K}^{\Stab{n}}_{\rhoin\mkern-1mu,\rhoout}}$
    and
    $\Omega^{-1} \circ \Omega = \id_{\mathscr{K}^{G,B_{\R^2}\mkern-1mu(0,\pi)}_{\rhoin\mkern-1mu,\rhoout}}$.
    The technical part of the proof is to show that the two kernel constraints imply each other, which is done in Appendix~\ref{apx:spherical_conv_kernel_space_iso}.
\end{proof}
Note that the volume scaling factor is not necessary to establish the isomorphism between the kernel spaces but is required to make the spherical convolution integral over $S^2\backslash \mkern-1mu\minus\mkern1mu n$ equivalent to the $\GM$-convolution integral over $B_{\R^2}(0,\pi)$.

\citet{Cohen2018-intertwiners} define the convolution ${[\kappa \star_{\mkern-2mu S^2}\! f]}$ of a feature field $f\in\Gamma(\Ain)$ with spherical steerable kernels $\kappa \in \mathscr{K}^{\Stab{n}}_{\rhoin\mkern-1mu,\rhoout}$ in coordinates.
Given gauges $P$ at $p$ and $q$ at $Q$, let $\phi_p \in \I$ be the unique isometry that moves the north pole to $p$, i.e. $\phi_p(n) = p$, and that maps the frame at $n$ to the frame at $p$, that is, $\dphipGM \sigma^N(n) = \sigma^P(p)$ or, equivalently, $g_{\phi_p}^{PN}(n) = e$.
Let furthermore $X$ be the gauge at $\phi_p^{-1}(q)$.
The spherical convolution is then in~\cite{Cohen2018-intertwiners} relative to these gauges pointwise defined by
\begin{align}\label{eq:spherical_steerable_conv}
    \big[\kappa \star_{\mkern-2mu S^2}\! f\big]^P(p)
    \ :=\ \int\limits_{S^2} \kappa\big(\phi_p^{-1}q)\, \rhoin\big( g_{\phi_p^{-1}}^{XQ}(q) \big)\, f^Q(q)\ dq
    \ = \int\limits_{S^2 \backslash \mkern-2mu -p} \mkern-8mu \kappa\big(\phi_p^{-1}q)\, \rhoin\big( g_{\phi_p^{-1}}^{XQ}(q) \big)\, f^Q(q)\ dq \,,
\end{align}
where we removed the antipodal point $-p$ in the second step without changing the result.%
\footnote{
    This formulation is more general than that in Eq.~\eqref{eq:spherical_lifting_conv}.
    The latter is recovered for kernels that map scalar fields to regular feature fields.
}
Intuitively, this operation computes an output feature at $p$ by
1)~taking both the kernel and the input field,
2)~rotating them via $\phi_p^{-1}$ such that $p$ moves to the north pole (via the induced gauge transformation for the feature vector) and
3)~integrating their product over the sphere.
Instead of sharing weights directly over the tangent spaces, as we do, this operation is therefore sharing weights via the isometry action.
By the definition of $\phi_p$, both definitions of weight sharing orient the kernel at the target location $p$ such that it is aligned with the chosen frame $\sigma^P(p)$ at this location.
The following theorem proves that the $\GM$-convolution with a kernel $K \in \mathscr{K}^{G,B_{\R^2}(0,\pi)}_{\rhoin\mkern-1mu,\rhoout}$ is equivalent to the spherical convolution with the corresponding spherical kernel $\Omega(K) \in \mathscr{K}^{\Stab{n}}_{\rhoin\mkern-1mu,\rhoout}$.
\begin{thm}[Spherical steerable convolutions as \textit{GM}-convolutions]
\label{thm:spherical_conv_GM_conv}
    Let $\Stab{n}$ be the stabilizer subgroup of any transitive isometry group $\I$ of $S^2$ and let $G \leq \GL{2}$ be any isomorphic structure group.
    Let furthermore $K \in \mathscr{K}^{G,B_{\R^2}(0,\pi)}_{\rhoin\mkern-1mu,\rhoout}$ be any $G$-steerable kernel on the open ball $B_{\R^2}(0,\pi)$ of radius $\pi$ (Eq.~\eqref{eq:G_steer_kernel_space_open_ball_pi})
    and let $\Omega(K) \in \mathscr{K}^{\Stab{n}}_{\rhoin\mkern-1mu,\rhoout}$ be its corresponding $\Stab{n}$-steerable kernel on $S^2\backslash \mkern-1mu\minus\mkern1mu n$ (Eqs.~\eqref{eq:spherical_steerable_kernel_space} and~\eqref{eq:spherical_kernel_space_iso_Omega}).
    The $\GM$-convolution (here for clarity denoted by $\star_{\mkern-1mu\scalebox{.64}{$\GM$}}$) with $K$ is then equivalent to the spherical convolution ($\star_{\mkern-2mu S^2}$, Eq.~\eqref{eq:spherical_steerable_conv}) by \citet{Cohen2018-intertwiners} with the spherical kernel $\Omega(K)$, that is,
    \begin{align}
        \Omega(K) \star_{\mkern-2mu S^2}\! f\ =\ K \star_{\mkern-1mu\scalebox{.64}{$\GM$}} f
    \end{align}
    holds for any spherical feature field $f \in \Gamma(\Ain)$.
\end{thm}
\begin{proof}
    The proof is given in Appendix~\ref{apx:spherical_conv_equivalence}.
\end{proof}
This proof justifies our claim that the models from \cite{Cohen2018-S2CNN,esteves2018zonalSpherical,esteves2020spinweighted,kondor2018ClebschGordan}, discussed in this section, are all special cases of $\GM$-convolutions.

%% file: chapters/113_spherical_SO2.tex

\subsection{Azimuthal rotation equivariant spherical CNNs on cylindrical topologies}
\label{sec:spherical_CNNs_azimuthal_equivariant}

\begin{figure}
    \centering
    \includegraphics[width=.90\textwidth]{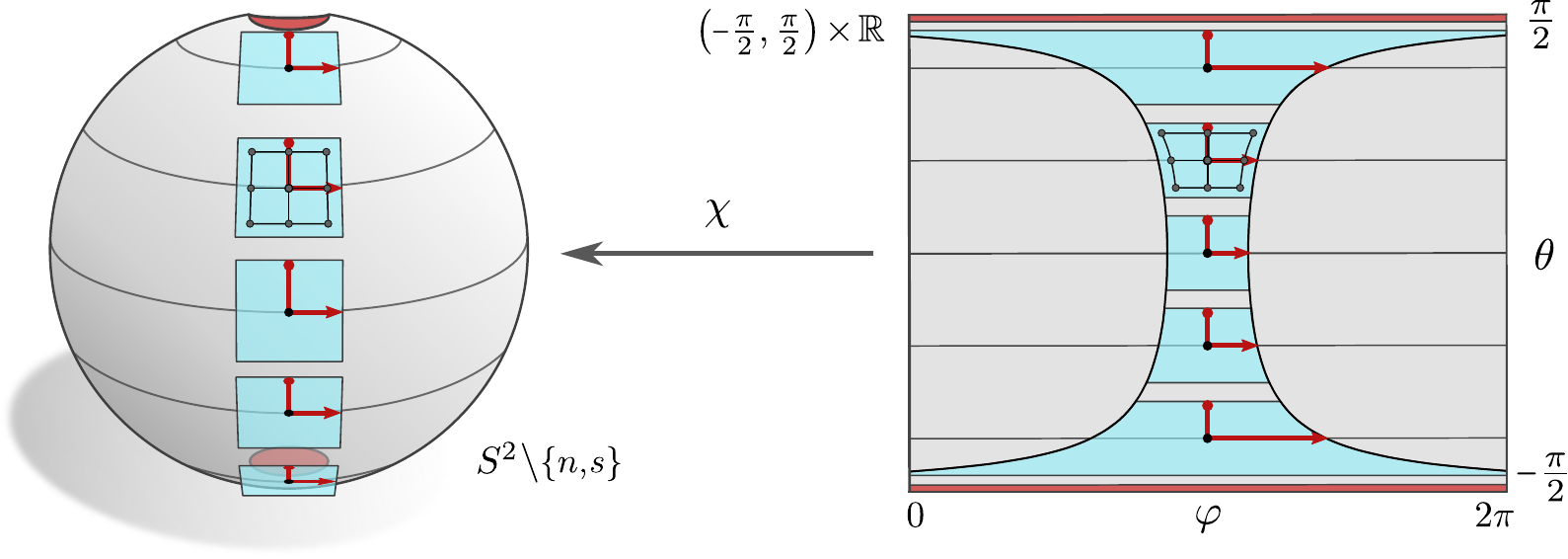}
    \caption{\small
        Visualization of the $\SO2$-invariant $\{e\}$-structure that is considered by most models discussed in Section~\ref{sec:spherical_CNNs_azimuthal_equivariant}.
        All of the frames 
        $\big[ \frac{\partial}{\partial\theta} ,\; \frac{1}{\cos(\theta)} \frac{\partial}{\partial\varphi} \big]$
        are aligned towards the north pole and are orthonormal w.r.t. the embedding metric of the sphere in $\R^3$.
        The spherical coordinate map
        $\chi: (-\frac{\pi}{2}, \frac{\pi}{2}) \times \R \to S^2 \backslash \{n,s\}$ from Eq.~\eqref{eq:spherical_coords}
        allows to pull spherical feature fields back to feature fields on spherical angles $(-\frac{\pi}{2}, \frac{\pi}{2}) \times \R$, which is denoted as equirectangular projection.
        Since~$\chi$ is non-isometric, the spherical $\{e\}$-structure is in coordinates deformed by a latitude dependent factor of $1/\cos(\theta)$, which diverges towards the poles.
        The spherical convolution on this $\{e\}$-structure is in~\cite{coors2018spherenet,eder2019convolutions,martin2020panoramic} implemented by projecting and interpolating a kernel sampling grid on the tangent spaces to a feature field sampling grid on the sphere.
        If the feature fields are instead sampled on the equirectangular projection, the kernel sampling grid is in a second step mapped further from the sphere to a deformed sampling grid on $(-\frac{\pi}{2}, \frac{\pi}{2}) \times \R$ \cite{zhao2018distortion,tateno2018distortion}.
        Note that regular sampling grids on the equirectangular projection oversample the signal (relative to the spherical metric) towards the poles.
    }
    \label{fig:spherical_equirectangular_1}
\end{figure}

\begin{figure}
    \centering
    \includegraphics[width=.90\textwidth]{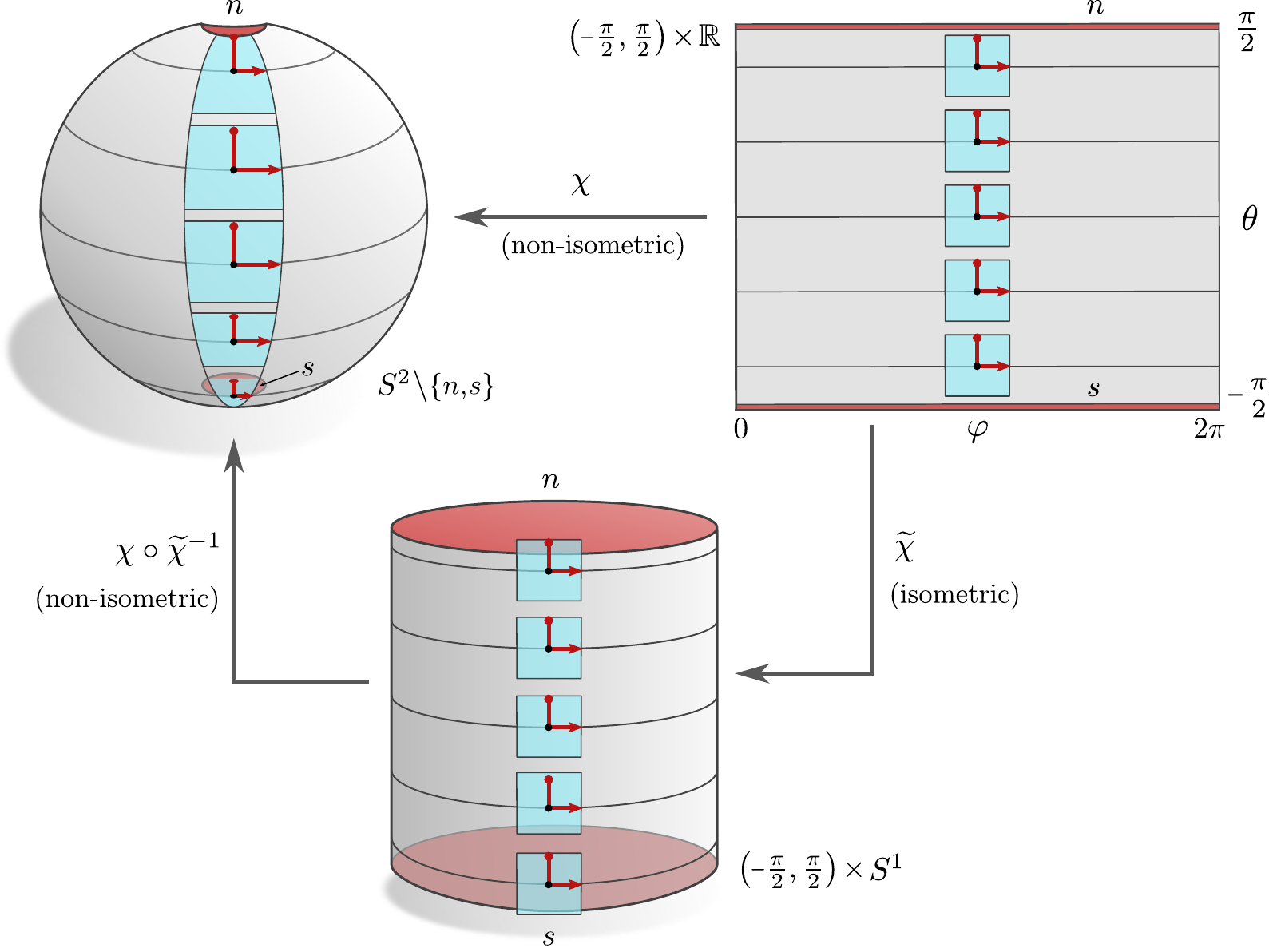}
    \caption{\small
        The spherical coordinate map
        $\chi: \big(\protect\minus\frac{\pi}{2}, \frac{\pi}{2}\big) \times \R \to S^2 \backslash \{n,s\}$, Eq.~\eqref{eq:spherical_coords}, sends angles $(\theta,\phi)$ to points on the sphere.
        It is non-isometric, which means that the pushforward of orthonormal frames
        $\big[ \frac{\partial}{\partial\theta}, \frac{\partial}{\partial\varphi} \big]$
        w.r.t. the \emph{Euclidean metric} on $\big(\protect\minus\frac{\pi}{2}, \frac{\pi}{2}\big) \times \R$ does not yield frames that are orthonormal w.r.t. the \emph{spherical metric}.
        A conventional Euclidean convolution in coordinates $\big(\protect\minus\frac{\pi}{2}, \frac{\pi}{2}\big) \times \R$ does therefore not correspond to a spherical convolution -- its kernels would be contracted by a factor of $\cos(\theta)$ in longitudinal direction.
        Since distances are measured in terms of angles, this operation corresponds rather to a convolution on a cylinder, which is via the isometric map
        $\widetilde{\chi}: \big( {\textstyle \protect\minus\frac{\pi}{2}, \frac{\pi}{2}} \big) \times \R
        \to \big( {\textstyle \protect\minus\frac{\pi}{2}, \frac{\pi}{2}} \big) \times S^1$,
        Eq.~\eqref{eq:cylindrical_coords}, embedded in $\R^3$.
        A spherical convolution requires the $\{e\}$-structure that is shown in Fig.~\ref{fig:spherical_equirectangular_1}.
    }
    \label{fig:spherical_equirectangular_2}
\end{figure}

Besides fully $\SO3$ or $\O3$-equivariant spherical convolutions, many spherical CNNs are designed to be equivariant w.r.t. azimuthal rotations around a specified polar axis.
All of the models discussed in this section rely either on the $\SO2$-invariant $\{e\}$-structure
that is shown in Figs.~\ref{fig:G_structure_S2_2} and \ref{fig:spherical_equirectangular_1}
or, alternatively, that in Fig.~\ref{fig:spherical_equirectangular_2}.
Due to the triviality of the structure group $G=\{e\}$, the kernel spaces remain unconstrained ($\{e\}$-steerable).
Features are transported according to the unique $\{e\}$-compatible trivial connection which differs from the usual spherical Levi-Civita connection.
With this information, and with the explicit exponential maps in Eq.~\eqref{eq:sphere_expmap_explicit}, the spherical $\GM$-convolutions in this section are in theory fully specified.
In practice, the implementations, listed in row (34) of Table~\ref{tab:network_instantiations}, differ in their numerical implementations, which we discuss in the following.

\begin{SCfigure}
    \centering
    \includegraphics[width=.55\textwidth]{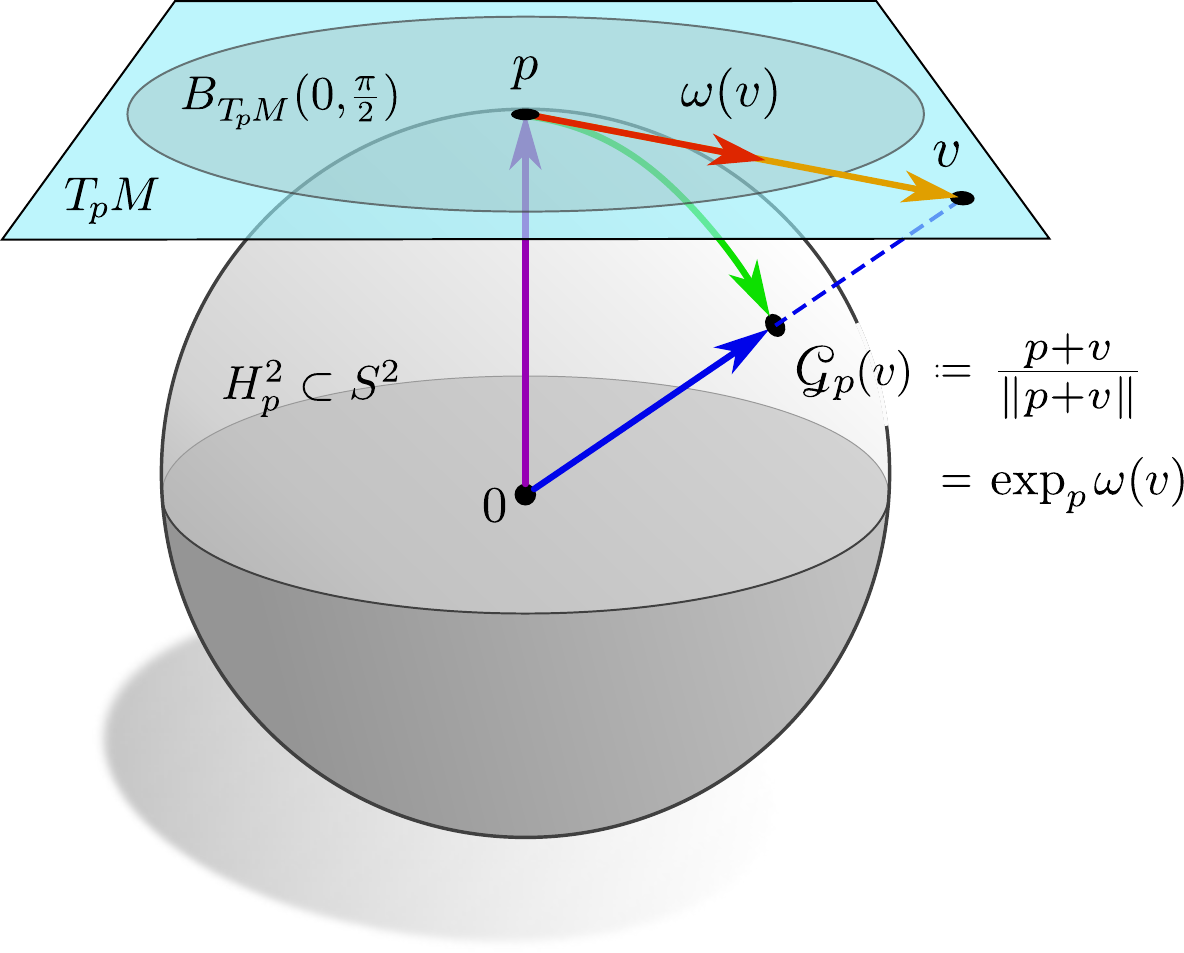}
    \captionsetup{width=1.02\textwidth}
    \hspace{1ex}
    \caption{\small
        Gnomonic projection ${\mathscr{G}_p\!: \TpM \to H_p^2}$ of the tangent space at $p$ to the upper hemisphere $H_p^2 \subset S^2$ around~$p$.
        When interpreting the sphere as being embedded in $\R^3$, the gnomonic projection $\mathscr{G}_p(v)$ (blue) is given by the sum of ${p\in S^2 \subset \R^3}$ (purple) and $v\in\TpM \subset\R^3$ (yellow) in ambient space, followed by a normalization back to the sphere.
        Theorem~\ref{thm:gnomonic} proves that this operation is equivalent to a projection of a radially warped vector $\omega(v) = \arctan(\lVert v\rVert) \frac{v}{\lVert v\rVert} \in B_{\TpM}(0,{\textstyle\frac{\pi}{2}})$ (red) via the exponential map (green).
        The gnomonic projection based spherical convolutions in
        \cite{coors2018spherenet,zhao2018distortion,tateno2018distortion,eder2019convolutions,martin2020panoramic}
        are therefore special cases of spherical $\GM$-convolutions with radially warped kernels.
        $\GM$-convolutions are more general since they allow for kernel projections over the whole sphere instead of the upper hemisphere only.
        \\\protect\rule{0ex}{0.ex}
    }
    \label{fig:gnomonic_proj}
\end{SCfigure}

Concurrent with our definition of convolutional weight sharing, the models in \cite{coors2018spherenet,zhao2018distortion,tateno2018distortion,eder2019convolutions,martin2020panoramic} share a given template kernel over the tangent spaces by orienting it relative to the frames of the considered $\{e\}$-structure in Fig.~\ref{fig:G_structure_S2_2}.
However, contrary to $\GM$-convolutions, the matching of these kernels with the feature field is not done via exponential maps (or transporter pullbacks), but via the gnomonic projection.
This gnomonic projection is at any point $p$ defined by
\begin{align}\label{eq:gnomonic_proj_def}
    \mathscr{G}_p:\, \TpM \to H_p^2, \quad v \mapsto \frac{p+v}{\lVert p+v\rVert} \,,
\end{align}
which is visualized in Fig.~\ref{fig:gnomonic_proj}.
The summation of $p\in S^2 \subset \R^3$ with tangent vectors $v \in \TpM \subset \R^3$ is hereby performed in the embedding space $\R^3$ and the normalization projects the result back to the sphere. 
The codomain of the gnomonic projection is the ``upper'' hemisphere
\begin{align}\label{eq:upper_hemisphere_def}
    H_p^2 := \big\{ q\in S^2 \,\big|\, \langle p,q\rangle_{\R^3} > 0 \big\} \,\subset S^2
\end{align}
centered around $p$.
Given this difference in the kernel projections, it might seems like the models in \cite{coors2018spherenet,zhao2018distortion,tateno2018distortion,eder2019convolutions,martin2020panoramic} are not (or only approximately) explained as $\GM$-convolution.
The following theorem proves, however, that the gnomonic projection is equivalent to a projection via the exponential map after applying a \emph{radial warp}
\begin{align}\label{eq:radial_warp}
    \omega:\ \TpM \to B_{\TpM}(0,{\textstyle\frac{\pi}{2}}), \quad
    v \mapsto \arctan\! \big(\lVert v\rVert\big) \frac{v}{\lVert v\rVert}
\end{align}
to the tangent spaces, which contracts tangent vectors to an open ball of radius $\pi/2$ around the origin:
\begin{thm}[Gnomonic projections as warped exponential maps]
\label{thm:gnomonic}
    The gnomonic projection $\mathscr{G}_p$ of $\TpM$ to the upper hemisphere $H_p^2\subset S^2$, defined in Eq.~\eqref{eq:gnomonic_proj_def}, is equivalent to a projection of its radial warp $\omega(\TpM) = B_{\TpM}(0,{\textstyle\frac{\pi}{2}})$, Eq.~\eqref{eq:radial_warp}, via the exponential map, that is, the following diagram commutes:
    \begin{equation}\label{cd:gnomonic_exp_warp}
    \begin{tikzcd}[column sep=60, row sep=6pt, font=\normalsize]
        \TpM
            \arrow[dr, pos=.45, "\mathscr{G}_p"]
            \arrow[dd, "\omega\,"']
        \\
        & H_p^2 \subset S^2
        \\
        B_{\TpM}(0,{\textstyle\frac{\pi}{2}})
            \arrow[ur, pos=.3, "\exp_p"']
    \end{tikzcd}
    \end{equation}
    In equations,
    \begin{align}
        \mathscr{G}_p(v) \,=\, \exp_p \circ\, \omega(v)
    \end{align}
    holds for any $p\in S^2$ and any $v\in\TpM$.
\end{thm}
\begin{proof}
    The proof is given by the following simple calculation, which holds for any $p\in S^2$ and any $v\in \TpM$:
    \begin{align}
        \exp_p \circ\mkern2mu \omega (v)
        \ &\overset{(1)}{=}\ p\cdot \cos\! \big(\lVert \omega(v)\rVert\big) \,+\, \frac{\omega(v)}{\lVert \omega(v)\rVert}\cdot \sin\! \big(\lVert \omega(v)\rVert\big) \notag \\
        \ &\overset{(2)}{=}\ p\cdot \cos\! \big(\!\arctan(\lVert v\rVert)\big) \,+\, \frac{v}{\lVert v\rVert}\cdot \sin\! \big(\!\arctan(\lVert v\rVert)\big) \notag \\
        \ &\overset{(3)}{=}\ \frac{p+v}{\sqrt{1+\lVert v\rVert^2}} \notag \\
        \ &\overset{(4)}{=}\ \frac{p+v}{\lVert p+v\rVert} \notag \\
        \ &\overset{(5)}{=}\ \mathscr{G}_p(v)
    \end{align}
    The first two steps make use of the explicit definition of the embedded sphere's exponential map, Eq.~\eqref{eq:sphere_expmap_explicit}, and the radial warp, Eq.~\eqref{eq:radial_warp}.
    The third step follows since $\cos \circ \arctan(x) = \frac{1}{\sqrt{1+x^2}}$ and $\sin \circ \arctan(x) = \frac{x}{\sqrt{1+x^2}}$.
    In the fourth step we used that $\lVert p\rVert = 1$ and $\langle p,v\rangle_{\R^3} = 0$, while the last step identified the gnomonic projection, Eq.~\eqref{eq:gnomonic_proj_def}.
\end{proof}
This theorem implies that the gnomonic projection based convolutions in
\cite{coors2018spherenet,zhao2018distortion,tateno2018distortion,eder2019convolutions,martin2020panoramic}
are indeed specific $\GM$-convolutions after identifying the kernels via the radial warp $\omega$.%
\footnote{
    Technically, the equivalence of both convolutions requires furthermore a radially dependent change of the kernel amplitude to account for the change in the volume measure when warping the kernel.
}
Note that this identification holds not only for $\{e\}$-steerable kernels but for any subgroup $G\leq\O2$ since the corresponding $G$-steerability constraints affect only the kernels' angular parts but are independent from the warped radial parts.
We furthermore want to mention that the exponential map based projection of $\GM$-convolutions is insofar more general than the gnomonic kernel projection that it can describe kernels that extend beyond the upper hemisphere $H_p^2$ around~$p$.
Note that both kernel projections become in the practically relevant limit of small kernels even without the radial warp equivalent since $\arctan\big(\lVert v\rVert\big) = \lVert v\rVert + \mathcal{O}\big(\lVert v\rVert^3\big)$.

The implementations in \cite{coors2018spherenet,zhao2018distortion,tateno2018distortion,eder2019convolutions,martin2020panoramic}
are in the continuum all equivalent to each other and to our $\GM$-convolution, however, their numerical discretizations differ.
\citet{coors2018spherenet}, \citet{eder2019convolutions} and \citet{martin2020panoramic} discretize feature fields on (approximately) uniform sampling grids on the sphere.
Specifically, \citet{coors2018spherenet} and \citet{martin2020panoramic} use the ``generalized spiral set on~$S^2$'' from~\cite{saff1997distributing} as sampling points, while \citet{eder2019convolutions} use the vertices of an icosphere.
Since the gnomonic projections of kernel sampling grids on the tangent spaces do not match the spherical sampling grid, the authors interpolate bilinearly between them.
The kernel sampling coefficients can hereby be precomputed in an offline step.
The actual convolution computes then an output feature field by contracting the projected, interpolated kernels at each point with the input field.

\citet{zhao2018distortion} and \citet{tateno2018distortion} discretize their spherical feature fields $f: S^2 \backslash \{n,s\} \to \R^c$ instead in form of a regular pixel grid on an equirectangular projection of the sphere.
Mathematically, the equirectangular projection, visualized in Fig.~\ref{fig:spherical_equirectangular_1}, is formalized as the pullback
$\chi^*f = f \circ \chi: \big(\minus\frac{\pi}{2}, \frac{\pi}{2}\big) \times \R \to \R^c$,
of the image by the spherical coordinate map $\chi$ from Eq.~\eqref{eq:spherical_coords}:
\begin{equation}\label{cd:equirectangular_proj}
\begin{tikzcd}[column sep=50pt, row sep=25pt, font=\normalsize]
    {\big(\minus\frac{\pi}{2}, \frac{\pi}{2}\big)} \times \R
        \arrow[r, "\chi"]
        \arrow[rr, pos=.5, rounded corners, to path={ 
                -- ([yshift=-2.5ex]\tikztostart.south) 
                --node[below]{\small$
                    \chi^*f
                    $} ([yshift=-2.5ex]\tikztotarget.south) 
                -- (\tikztotarget.south)
                }]
    & S^2 \backslash \{n,s\}
        \arrow[r, "f"]
    & \R^c \mkern-12mu\phantom{\big)}
\end{tikzcd}
\end{equation}
As in the previous approaches, the authors project a kernel sampling grid via the gnomonic projection from the tangent spaces to the sphere.
In an additional step, they map it via $\chi$ to the equirectangular projection where they compute interpolation coefficients between the projected kernel sampling grid and the feature field sampling gird.
Since the deformation incurred by the equirectangular projection is independent from the longitude~$\phi \in \R$, it is sufficient to compute it only once for each latitude~$\theta \in {\textstyle \big(\minus\frac{\pi}{2}, \frac{\pi}{2}\big)}$.
The following diagram, which commutes by the definitions of $K^{\textup{sphere}}_p$ and $K^{\textup{equirect}}_p$, gives an overview of the gnomonic projection of a kernel $K: \R^2 \to \R^{\cout\times\cin}$ to the sphere \cite{coors2018spherenet,eder2019convolutions,martin2020panoramic} and to its equirectangular projection \cite{zhao2018distortion,tateno2018distortion} (note that $\mathscr{G}_p$ is invertible on $H_p^2$):
\begin{equation}
\begin{tikzcd}[column sep=50pt, row sep=36pt, font=\normalsize]
    & &[20pt]
    \R^{\cout\times\cin}
    \\
    \R^2
        \arrow[urr, pos=.75, rounded corners, to path={ 
                -- ([yshift=0pt]\tikztostart.north) 
                |-node[above]{\small$
                        K
                    $} ([xshift=0ex]\tikztotarget.west) 
                }]
    & \TpM
        \arrow[l, "\psiTMp^A"]
        \arrow[r, "\mathscr{G}_p = \exp_p\mkern-2mu\circ\mkern2mu\omega"']
    & H_p^2
        \arrow[u, "K_p^\textup{sphere}\,"']
    &[10pt] \underbrace{\chi^{-1}(H_p^2)}_{\subset\ (\protect\minus\frac{\pi}{2}, \frac{\pi}{2}) \times \R}
        \arrow[l, "\chi"]
        \arrow[ul, pos=.72, rounded corners, to path={ 
                -- ([yshift=0pt]\tikztostart.north) 
                |-node[above]{\small$
                    K^{\textup{equirect}}_p
                    $} ([xshift=0ex]\tikztotarget.east) 
                }]
\end{tikzcd}
\end{equation}
A major disadvantage of discretizing spherical feature fields via a regular pixel grid on the equirectangular projection is that this approach oversamples the signal towards the poles.

Further variants of spherical convolutions on the equirectangular projection were proposed by \citet{su2017spherical,su2019kernel}.
Instead of precomputing the deformed kernel sampling pattern, \citet{su2017spherical} untie the weight sharing such that each latitude applies its own, independent kernel in a 1-dimensional Euclidean convolution.
The network is then on each latitude being pretrained to recover the result that would be obtained when convolving with a kernel that is shared over the tangent spaces as discussed above.
If convolutional weight sharing is a suitable inductive bias, this method should optimally converge to the geometry based methods by \citet{zhao2018distortion,tateno2018distortion}.
\citet{su2019kernel} develop this approach further and employ a meta-network that predicts a deformed kernel based on a shared input template kernel and the target latitude.
Both of these approaches share weights over the circular orbits (lines of constant latitude) of the considered isometry group~$\SO2$ of~$S^2 \backslash \{n,s\}$; cf. Fig.~\ref{fig:isom_invariant_kernel_field_multiple_orbits}.
They are therefore identified as kernel field transforms with $\SO2$-invariant kernel fields, which are by Theorem~\ref{thm:isometry_equivariant_kernel_field_trafos} $\SO2$-equivariant.

Given a spherical feature field in equirectangular projection, it might
furthermore
be tempting to process it directly with a conventional Euclidean CNN, skipping the kernel projection from the tangent spaces, as done for instance in~\cite{lai2017semantic,hu2017spherical}.
As discussed in Section~\ref{sec:instantiations_euclidean}, such Euclidean convolutions correspond to $\GM$-convolutions on the canonical $\{e\}$-structure of $\big(\minus\frac{\pi}{2}, \frac{\pi}{2}\big) \times \R \subset \R^2$, visualized in Figs.~\ref{fig:G_structure_R2_1} and~\ref{fig:spherical_equirectangular_2} (top right).
This $\{e\}$-structure consists of frames 
$\big[ \frac{\partial}{\partial\theta} ,\, \frac{\partial}{\partial\varphi} \big]$,
which are orthonormal w.r.t. the \emph{Euclidean metric} of $\big(\minus\frac{\pi}{2}, \frac{\pi}{2}\big) \times \R$.
These frames are, however, not orthonormal w.r.t. the \emph{spherical metric}, Eq.~\eqref{eq:spherical_embedding_metric_explicit}, which is in Fig.~\ref{fig:spherical_equirectangular_2} (top left) reflected in the frame contraction by a factor of $\cos(\theta)$ in longitudinal direction.
A $\GM$-convolution on this $\{e\}$-structure corresponds therefore geometrically \emph{not} to a spherical convolution.
It rather corresponds to a $\GM$-convolution on a cylinder, which is via the \emph{isometric} coordinate map
\begin{align}\label{eq:cylindrical_coords}
    \widetilde{\chi}:\, \big( {\textstyle \minus\frac{\pi}{2}, \frac{\pi}{2}} \big) \!\times \R
    \,\to\, \big( {\textstyle \minus\frac{\pi}{2}, \frac{\pi}{2}} \big) \!\times S^1,
    \quad (\theta,\phi) \mapsto
    \begin{pmatrix}
        \cos{\phi} \\
        \sin{\phi} \\
        \theta
    \end{pmatrix}
\end{align}
embedded in $\R^3$.
In contrast, the $\{e\}$-structure that is shown in Figs.~\ref{fig:G_structure_S2_2} and~\ref{fig:spherical_equirectangular_1} consists of frames
$\big[ \frac{\partial}{\partial\theta} ,\; \frac{1}{\cos(\theta)} \frac{\partial}{\partial\varphi} \big]$,
which are orthonormal w.r.t. the spherical metric.
Note that these frames and the spherical metric are stretched by a factor of $1/\cos(\theta)$ relative to their canonical Euclidean counterparts on $\big( {\textstyle \minus\frac{\pi}{2}, \frac{\pi}{2}} \big) \times \R$.

\citet{jiang2019spherical} propose an alternative approach for spherical convolutions on the $\{e\}$-structure shown in Figs.~\ref{fig:G_structure_S2_2} and~\ref{fig:spherical_equirectangular_1}.
Instead of defining kernels on the tangent spaces, they process the signal via second order partial differential operators of the form $w_{\id} + w_{e^A_1} \partial_1 + w_{e^A_2} \partial_2 + w_\textup{Laplace} (\partial_1^2 + \partial_2^2)$, where $\partial_i$ denotes the partial derivative in the direction of the $i$-th frame axis and the weights $w_{(\cdot)} \in \R^{\cout\times\cin}$ are optimized during training.
That the weights are position independent corresponds to our spatial weight sharing.
Together with the $\SO2$-invariance of the $\{e\}$-structure, along which the differential operators are aligned, this guarantees the $\SO2$-equivariance of the operation.
In the continuous theory, this model corresponds to a $\GM$-convolution in the limit of infinitesimally small kernels.
In practice, \citet{jiang2019spherical} sample the feature field on an icosphere mesh and represent the differential operators in terms of spatially extended stencils on the mesh vertices.
This makes the method equivalent to a $\GM$-convolution with spatially extended kernels.

The model of \citet{lee2019spherephd} operates again on an icosphere, however, with a drastically changed (non-smooth) $\{e\}$-structure:
instead of aligning the reference frames such that they all point towards the north pole, the frames point alternatingly towards the north or south.
This design is motivated by the pixelation of the icosphere, whose triangular faces are facing either north or southwards.
Adjacent pixels can therefore be processed by kernels that are rotated by $180^\circ$ relative to each other.
The authors argue that the training process should make up for this rotation by learning accordingly steerable kernels.
Despite the drastic kernel rotations, the $\{e\}$-structure is invariant under those azimuthal rotations that map northwards pointing frames on themselves, resulting in an approximate $\SO2$-equivariance of the convolution.

The models discussed in this section are easily extended to other solids of revolution ($\SO2$-invariant manifolds) like the cylinder from Fig.~\ref{fig:spherical_equirectangular_2} or the egg from Fig.~\ref{fig:isom_egg_main}.
They are furthermore adapted to be $\O2$-equivariant when considering a lift of the $\{e\}$-structures to $\Flip$-structures, which corresponds to using $\Flip$-steerable kernels as shown in Fig.~\ref{fig:isom_invariant_kernel_field_quotient}.

%% file: chapters/114_spherical_ico.tex

\subsection{Icosahedral approximations of spherical CNNs}
\label{sec:spherical_CNNs_icosahedral}

\begin{figure}
    \centering
    \includegraphics[width=.8\textwidth]{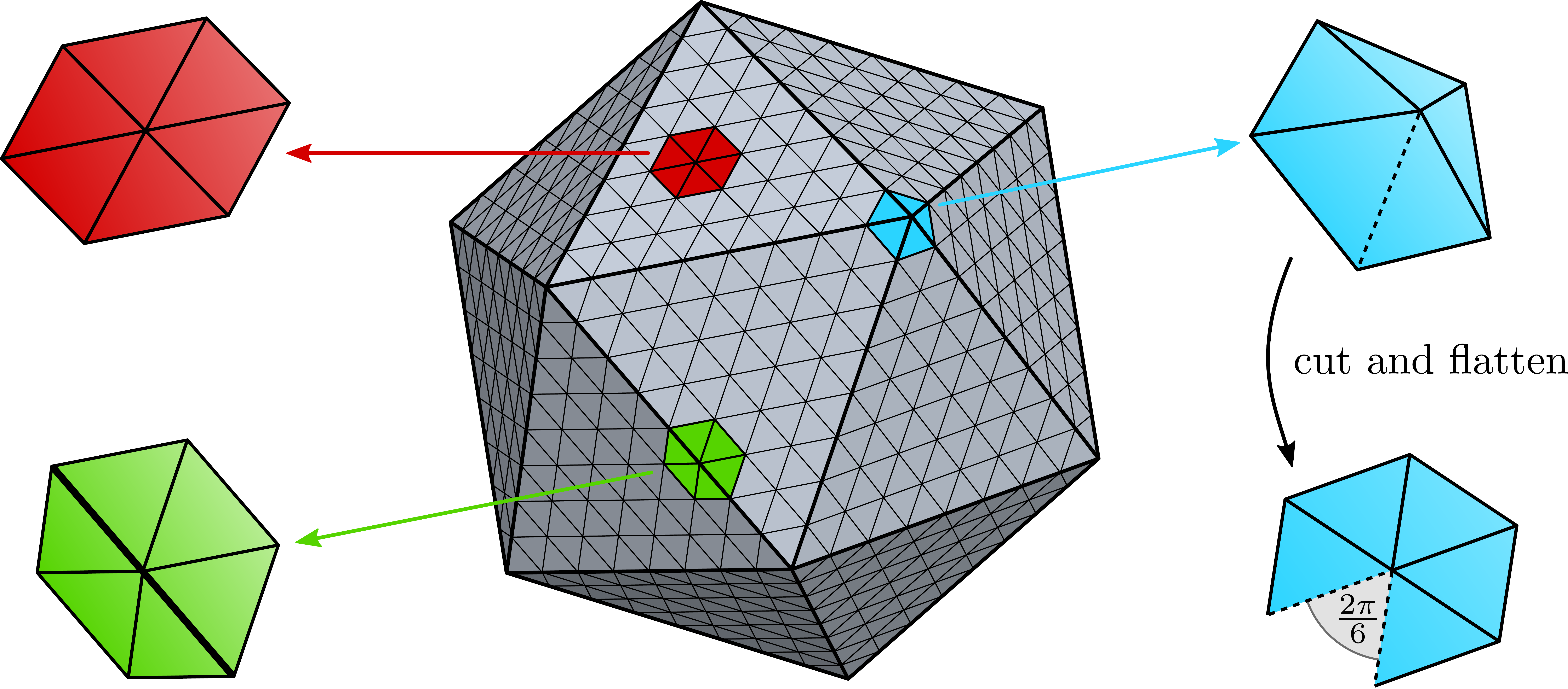}
    \vspace*{1ex}
    \caption{\small
        The icosahedron is a Platonic solid that is in \cite{liu2018icoAltAz,gaugeIco2019,zhang2019orientation} used as a piecewise flat approximation of the spherical geometry.
        It consists of 12~vertices, 20~equilateral triangular faces and 30~edges.
        It admits a regular sampling grid, which is constructed by iteratively subdividing each triangle into four smaller triangles.
        After~$r$ iterations, this procedure results in a grid of ${5\mkern-1mu\cdot\mkern-1mu 2^{2r+1} + 2}$ vertices.
        The three highlighted patches show the qualitatively different geometry of neighborhoods around vertices on faces (red), edges (green) and icosahedron vertices (blue).
        The red neighborhood is obviously flat.
        While the green neighborhood is bent in the embedding space, its intrinsic Gaussian curvature is again vanishing.
        That this is the case reflects in the facts that it can be flattened out isometrically (i.e. without being cut) and, equivalently, that the Levi-Civita transport along a closed path around the central node is the identity map.
        The blue neighborhood needs to be cut along one edge in order to be flattened out.
        The angle defect, i.e. the angle by which the cut is spread when flattening the cusp, equals~$\frac{2\pi}{6}$.
        When parallel transporting a vector once around the central vertex of the neighborhood, it gets rotated by this angle defect.
        Instead of having constant positive Gaussian curvature like the sphere $S^2$, the icosahedron's curvature is concentrated (singular) at its vertices and vanishes everywhere else.
    }
    \label{fig:ico_neighborhoods}
\end{figure}

\begin{figure*}
    \centering
    \begin{subfigure}[b]{0.31\textwidth}
        \centering
        \includegraphics[width=1.\textwidth]{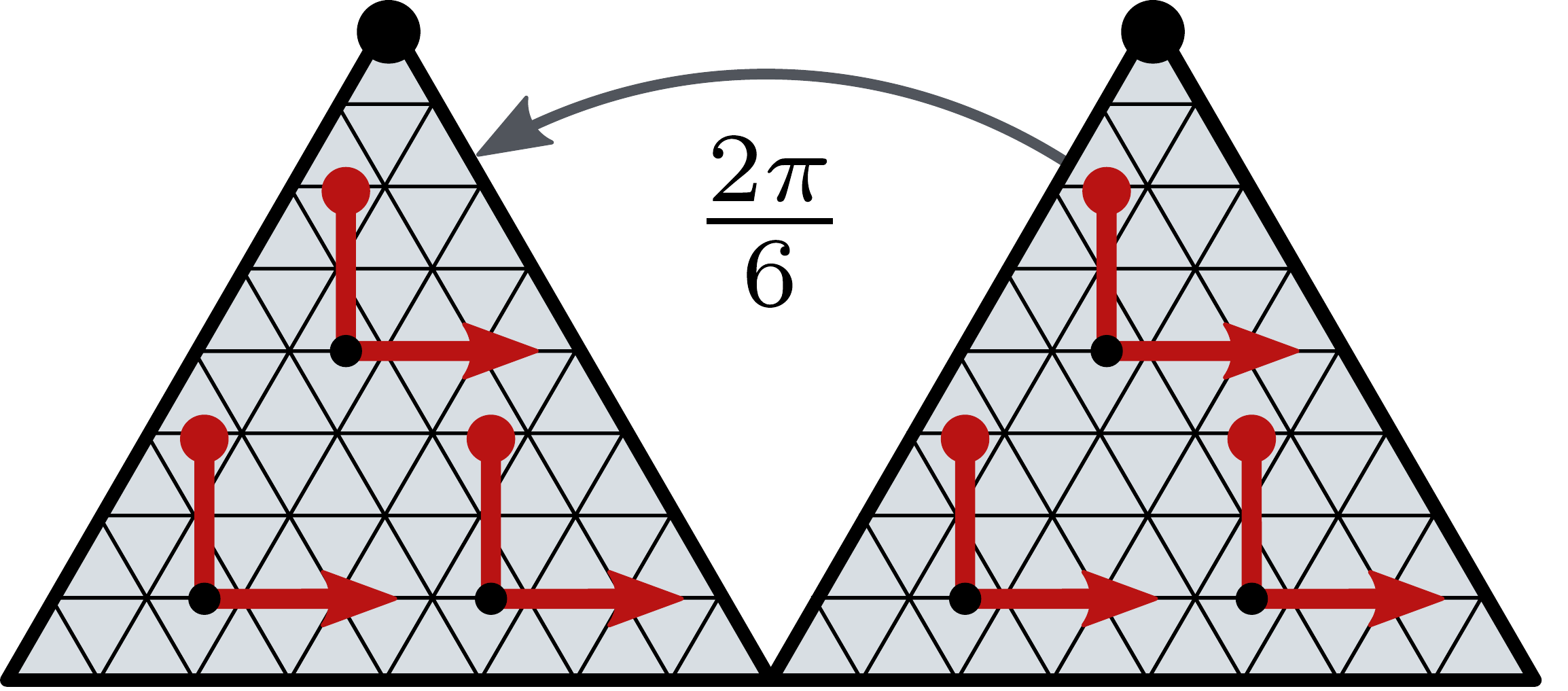}
        \vspace*{-6pt}
        \captionsetup{width=.9\textwidth}
        \caption{\small
            Grid-aligned icosahedral $\{e\}$-structure by \citet{liu2018icoAltAz}.
        }
        \label{fig:G_structure_ico_1}
    \end{subfigure}
    \hfill
    \begin{subfigure}[b]{0.31\textwidth}
        \centering
        \includegraphics[width=1.\textwidth]{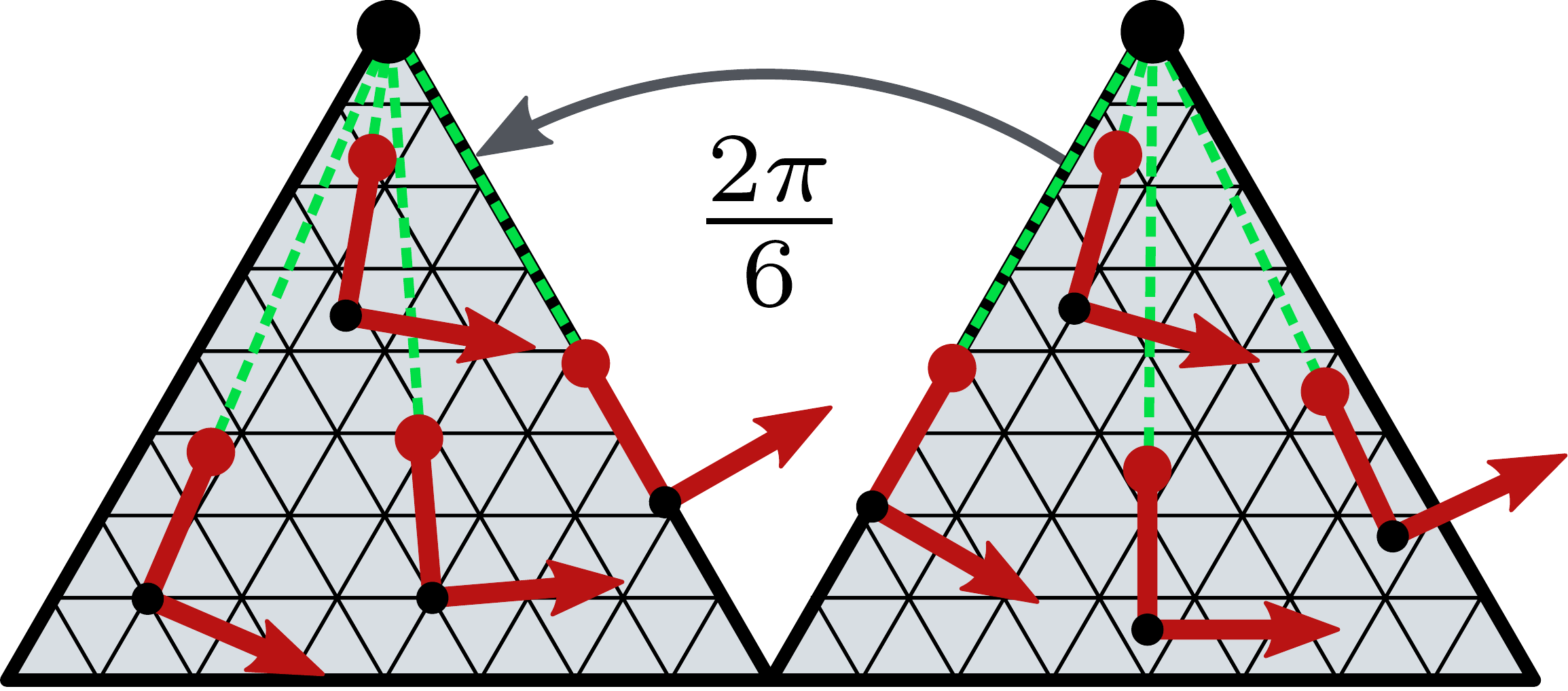}
        \vspace*{-6pt}
        \captionsetup{width=.9\textwidth}
        \caption{\small
            North-aligned icosahedral \mbox{$\{e\}$-structure by \citet{zhang2019orientation}}.
        }
        \label{fig:G_structure_ico_2}
    \end{subfigure}
    \hfill
    \begin{subfigure}[b]{0.31\textwidth}
        \centering
        \includegraphics[width=1.\textwidth]{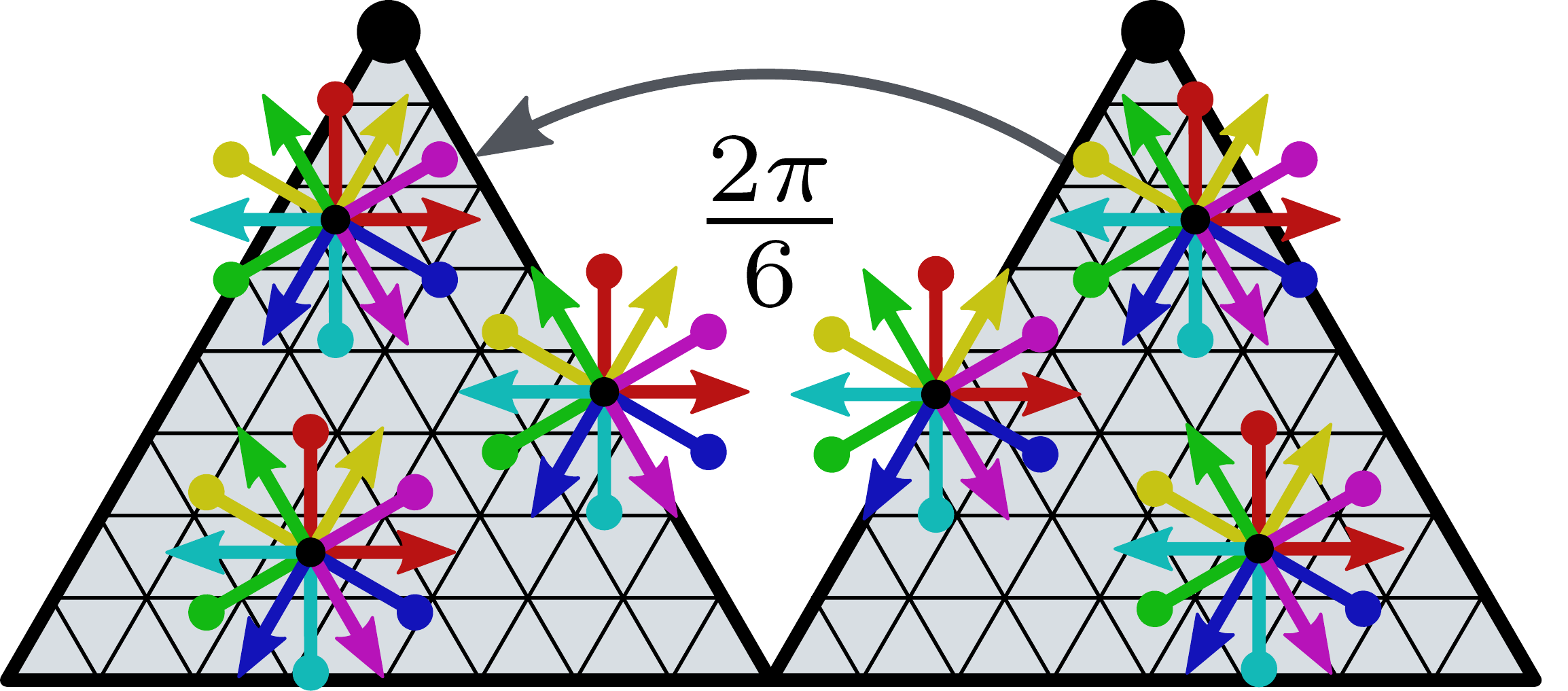}
        \vspace*{-6pt}
        \captionsetup{width=.9\textwidth}
        \caption{\small
            Grid-aligned icosahedral $\C6$-structure by \citet{gaugeIco2019}.
        }
        \label{fig:G_structure_ico_3}
    \end{subfigure}
    \caption{\small
        Conceptual idea of the $G$-structures assumed in~\cite{liu2018icoAltAz,zhang2019orientation,gaugeIco2019}.
        For space reasons, only two adjacent faces next to the north pole of the flattened icosahedron (Fig.~\ref{fig:ico_cutting}) are shown.
        The $\{e\}$-structure in Fig.~\ref{fig:G_structure_ico_1} is defined by aligning all frames along the ``horizontal'' edges of the faces (assuming the polar axis to be vertical).
        Fig.~\ref{fig:G_structure_ico_2} shows an alternative $\{e\}$-structure whose frames are aligned towards the north pole.
        It is in contrast to the previous $\{e\}$-structure continuous since frames on the cut edges agree with each other when gluing the edges back together.
        The $\C6$-structure in Fig.~\ref{fig:G_structure_ico_3} is constructed by adding frames that are rotated by multiples of $\frac{2\pi}{6}$ to the $\{e\}$-structure from Fig.~\ref{fig:G_structure_ico_1}.
        Since this angle agrees with the angle defect at the cut edges, the thus defined $\C6$-structure is smooth (continuous).
        Note that the two $\{e\}$-structures are incompatible with (i.e. not closed under) the Levi-Civita transport but imply an alternative trivial connection.
        The $\C6$-structure, in contrast, is compatible with the Levi-Civita transport.
     }
    \label{fig:G_structures_ico}
\end{figure*}

\begin{figure}
    \centering
    \includegraphics[width=.9\textwidth]{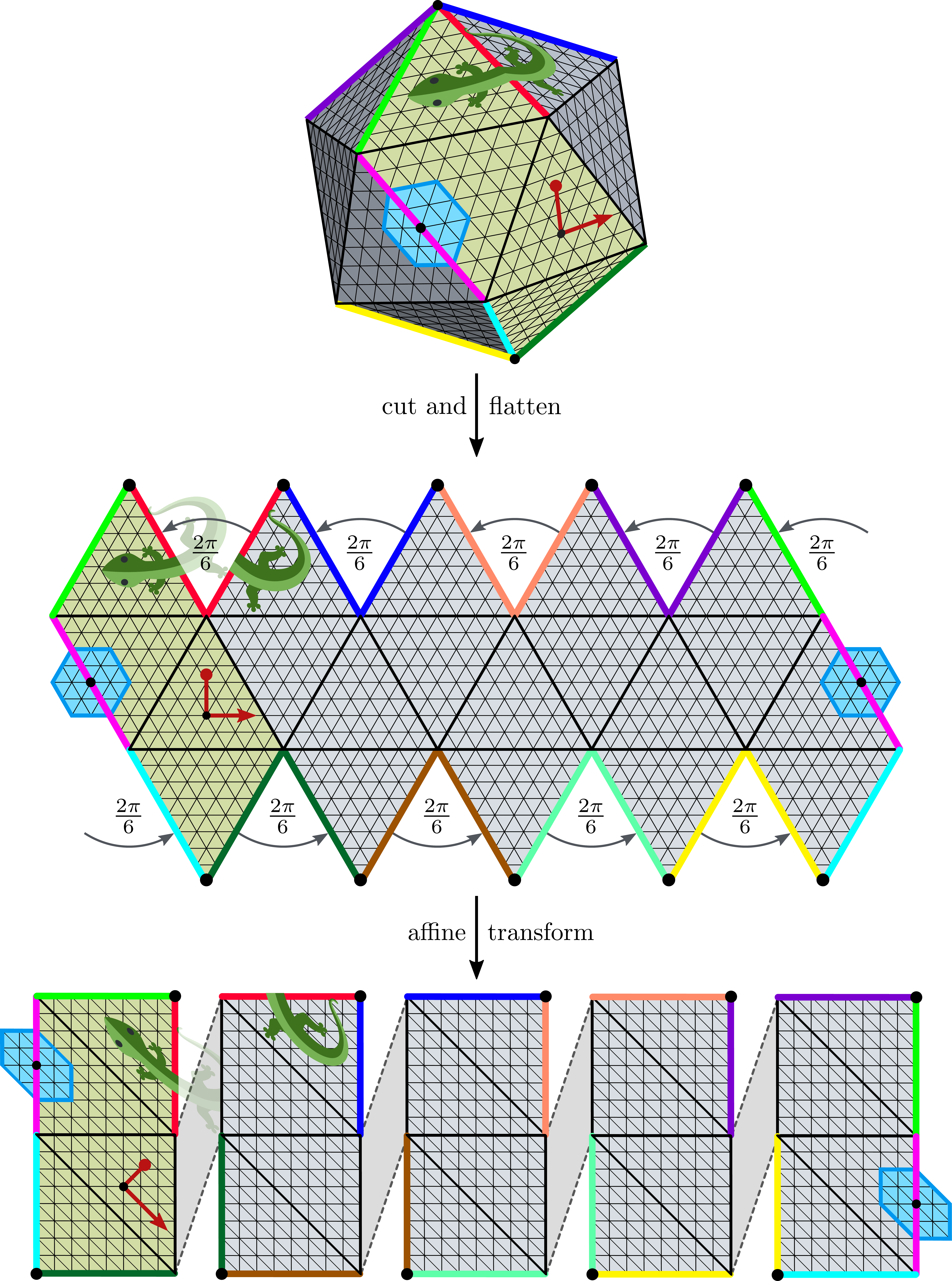}
    \vspace*{1.5ex}
    \caption{\small
        The implementations in \cite{liu2018icoAltAz,gaugeIco2019,zhang2019orientation} represent feature fields relative to an atlas that covers the icosahedron with five charts.
        To construct these charts, one cuts the icosahedron along the colored edges and flattens it out.
        Five regions, consisting of four triangles each, are then sheared to rectangular chart codomains.
        This operation maps the hexagonal grid to a grid of square pixels, such that icosahedral feature fields can be encoded by a set of five rectangular arrays.
        Note that reference frames and kernels are deformed accordingly on the chart codomains.
        The Levi-Civita transport over all colored edges but the magenta one picks up a rotation of $\pm\frac{2\pi}{6}$, with the sign depending on the transport direction.
        This is implemented by transport padding rows of pixels along the cut edges as previously described in Fig.~\ref{fig:mobius_conv_numerical}.
        {
        \\ \color{gray} \scriptsize
            (Lizards adapted under the Creative Commons Attribution 4.0 International
            \href{https://github.com/twitter/twemoji/blob/gh-pages/LICENSE-GRAPHICS}{\underline{license}}
            by courtesy of Twitter.)
        }
     }
    \label{fig:ico_cutting}
\end{figure}

The sphere $S^2$ is in computational sciences commonly approximated by Platonic solids, i.e. regular convex polyhedra.
In the context of deep learning, interest has mostly been focused on the icosahedron, Fig.~\ref{fig:ico_neighborhoods}, which approximates the sphere most closely among the platonic solids~\cite{schroder1995spherical}.
While the Riemannian geometry of the sphere is only approximated, Platonic solids have the advantage to be piecewise flat and admit regular meshes.
These properties allow for the use of planar convolution routines, which are computationally better optimized than the methods from the previous two sections.
This section discusses the icosahedral CNNs from~\cite{liu2018icoAltAz}, \cite{zhang2019orientation} and~\cite{gaugeIco2019}, which rely on the $G$-structures that are shown in Figs.~\ref{fig:G_structure_ico_1}, \ref{fig:G_structure_ico_2} and~\ref{fig:G_structure_ico_3}, respectively.
Before coming to their implementations in terms of the atlas of affine charts in Fig.~\ref{fig:ico_cutting},
we give more details on the icosahedral geometry and the considered $G$-structures.

\paragraph{Icosahedral geometry:}
The icosahedron is a discrete two-dimensional manifold consisting of 20~equilateral triangular faces, 12~vertices and 30~edges.
As done for the 2-sphere, we define the icosahedron as being embedded in $\R^3$, from which it inherits the embedding metric in~Eq.~\eqref{eq:spherical_embedding_metric_explicit}.
The embedded tangent spaces $\TpM \subset \R^3$ on the faces are hereby defined such that their normals coincide with the face normals.
Tangent spaces on the vertices and edges could be defined via the average of the adjacent faces' normals as discussed in the following Section~\ref{sec:instantiations_mesh}.
However, as we consider feature fields as being sampled on the icosahedron faces (which is almost everywhere), we are independent form this choice.
Assuming the Levi-Civita connection, 
the parallel transport of tangent vectors over faces acts such that it keeps them parallel in the embedding space~$\R^3$.
When being parallel transported across an edge, tangent vectors keep the same angle relative to the edge on either side -- this transport may intuitively be thought of as 1) flattening the two adjacent faces out 2) transporting the vector over the edge as usual on a two-dimensional Euclidean space, and 3) bending the two faces back to their original embedding; see Fig.~\ref{fig:transport_mesh} and \cite{craneDiscreteDifferentialGeometry2014}.
Geodesics are therefore piecewise linear in $\R^3$, crossing edges such that their angle of emanation equals their angle of incidence.
Exponential maps $\exp_p(v)$ are thus easily computed by tracing out a piecewise constant path for a distance of $\lVert v\rVert$.
In practice, the authors of~\cite{liu2018icoAltAz,gaugeIco2019,zhang2019orientation} sample feature fields on a regular mesh and consider only those tangent vectors that map to the neighboring mesh vertices.

Fig.~\ref{fig:ico_neighborhoods} shows disc-like neighborhoods around exemplary points on faces (red), edges (green) and vertices (blue) of the icosahedron.
The red neighborhood is fully contained within a face, and is therefore flat.
The green neighborhood is bent in the embedding space, however, its intrinsic (Riemannian or Gaussian) curvature is still zero since the Levi-Civita transport of vectors once around the central vertex preserves them as they are.
That this is the case is equivalent to the fact that the green neighborhood can be flattened out isometrically, i.e. without stretching or cutting it.
This isometric flattening is not possible for the blue type of neighborhoods around vertices, which have to be cut open at one of the edges in order to be flattened out.
Being constructed from five equilateral triangles, the flattened cusp exhibits an angle defect of~$\frac{2\pi}{6}$.
The holonomy of any closed path around any (single) vertex, that is, the angle between an arbitrary vector and its transport once around the loop, is given exactly by this angle defect.
Overall, these results imply that the (discrete) Gaussian curvature of the icosahedron is zero everywhere but at the vertices, where it is singular with holonomy $\frac{2\pi}{6}$.
The simple geometry of the icosahedron allows for it to be cut open and globally flattened out as visualized in Fig.~\ref{fig:ico_cutting}, which was in~\cite{liu2018icoAltAz,gaugeIco2019,zhang2019orientation} used for an efficient implementation of icosahedral $\GM$-convolutions.

The icosahedron's full isometry group $\IsomM = \operatorname{I}_h \leq \O3$ is finite and consists of 120~elements.
It can be thought of as being constructed as the direct product $\operatorname{I} \times \Flip$ of the subgroup $\Flip$ of reflections and the subgroup $\IsomplusM = \operatorname{I} \leq \SO3$ of orientation preserving isometries, containing 60 rotations.
Each vertex~$p$ is stabilized by five discrete rotations around the axis through $p$ and its antipodal vertex, which form the cyclic group~$\C5 \leq \SO2$.
The vertex~$p$ is furthermore stabilized by reflections over the plane defined by the rotation axis and any edge emanating from~$p$, such that its full stabilizer subgroup is given by the dihedral group $\Stab{p} = \D5 \leq \O2$.
The equivariance of icosahedral $\GM$-convolutions w.r.t. isometry groups $\operatorname{I}_h$, $\operatorname{I}$, $\D5$ or $\C5$ was in~\cite{gaugeIco2019} shown to approximate the full $\O3$, $\SO3$, $\O2$ or $\SO2$ equivariance of spherical CNNs reasonably well when continuous rotational data augmentation is used.%
\footnote{
    This was in~\cite{gaugeIco2019} empirically shown for $\operatorname{I} \leq \SO3$.
    That this result generalizes to $\operatorname{I}_h \leq \O3$ is clear since the groups differ only by reflections, w.r.t. which icosahedral $\GM$-convolutions can be made exactly equivariant.
    It holds furthermore for $\D5 \leq \O2$ and $\C5 \leq \SO2$, since these are subgroups of $\operatorname{I}_h \leq \O3$.
}

\paragraph{Icosahedral \textit{G}-structures:}
The icosahedral $\GM$-convolutions by~\citet{liu2018icoAltAz} and~\citet{zhang2019orientation} (implicitly) assume $\{e\}$-structures, while that by~\citet{gaugeIco2019} assumes a $\C6$-structure.
Fig.~\ref{fig:G_structures_ico} visualizes the idea behind these $G$-structures, which we explain in the following three paragraphs in more detail.

The $\{e\}$-structure by~\citet{liu2018icoAltAz}, shown in Fig.~\ref{fig:G_structure_ico_1}, is defined by aligning the first frame axes with the ``horizontal'' edges of the corresponding triangular faces.
When flattening the icosahedron into a plane as shown in Fig.~\ref{fig:ico_cutting}, all frames of this $\{e\}$-structure are parallel in this plane, which greatly simplifies the implementation of the corresponding $\GM$-convolutions.
As usual, the $\{e\}$-structure specifies a unique trivial connection according to which features are transported.
This trivial connection agrees within the faces, on edges which are not cut in Fig.~\ref{fig:ico_cutting} and on the magenta cut edge with the Levi-Civita connection.
However, its transport over the remaining cut edges differs from the Levi-Civita transport since the frames of the $\{e\}$-structure rotate there discontinuously by an angle of~$\frac{2\pi}{6}$.
As the $\{e\}$-structure is preserved by rotations in $\C5$ around the polar axis, its $\GM$-convolutions are approximately $\SO2$-equivariant, i.e. approximate the models from the previous Section~\ref{sec:spherical_CNNs_azimuthal_equivariant}.
However, the $\{e\}$-structure -- and therefore the network inference -- is \emph{non-continuous} over the edges with non-zero angle defect.
Furthermore, the reference frames do not point exactly towards the north pole, as it is the case for the spherical $\{e\}$-structure from Section~\ref{sec:spherical_CNNs_azimuthal_equivariant} and Fig.~\ref{fig:G_structure_S2_2}.

\citet{zhang2019orientation} propose to resolve the latter two issues by working with the $\{e\}$-structure in Fig.~\ref{fig:G_structure_ico_2}.
It is defined such that the frames point exactly along the projection of the polar axis onto the faces, i.e. towards the north pole.
This $\{e\}$-structure is continuous everywhere except at the north and south poles.%
\footnote{
    To see this, imagine to glue the cut edge in Fig.~\ref{fig:G_structure_ico_2} back together:
    the frames on the left and right half of the edge are then being mapped together, which is not the case in Fig.~\ref{fig:G_structure_ico_1}.
}
It is in this sense a better approximation of the spherical $\{e\}$-structure from Fig.~\ref{fig:G_structure_S2_2}.
The $\{e\}$-structure implies again a unique trivial connection.
Its transport agrees with the Levi-Civita transport over edges, however, it differs from it when transporting over faces since it rotates vectors smoothly along with the frames.
As the other $\{e\}$-structure, this frame field is invariant under azimuthal rotations in $\C5$, and approximates thus azimuthal rotation equivariant spherical CNNs.

The $\C6$-structure in Fig.~\ref{fig:G_structure_ico_3} by~\citet{gaugeIco2019} is defined by augmenting the frames of the $\{e\}$-structure from Fig.~\ref{fig:G_structure_ico_1} with those frames that are rotated by multiples of $\frac{2\pi}{6}$.
It is clearly continuous since the angles between the set of preferred frames at each point equal exactly the angle defects at the cut edges.
It is in contrast to the previous two $\{e\}$-structures compatible with the Levi-Civita transport since the structure group $\C6$ agrees with the icosahedron's holonomy group.
The $\C6$-structure is furthermore preserved under the action of the icosahedron's orientation preserving isometries $\operatorname{I}$.
$\GM$-convolutions on this $\C6$-structure approximate therefore the fully $\SO3$ rotation equivariant spherical CNNs from Section~\ref{sec:spherical_CNNs_fully_equivariant}.

\paragraph{Implementations:}
To implement the $\GM$-convolutions on the corresponding $G$-structures,
\citet{liu2018icoAltAz}, \citet{zhang2019orientation} and~\citet{gaugeIco2019}
assume a regular grid on the icosahedron's faces; see Fig.~\ref{fig:ico_neighborhoods}.
This regular hexagonal grid is constructed by iteratively subdividing edges, replacing each triangle with four smaller ones.
At resolution~$r$, this yields a grid with ${5\mkern-1mu\cdot\mkern-1mu 2^{2r+1} + 2}$ vertices.
Note that this grid is by construction exactly symmetric under isometries of the icosahedron, which leads to an exact $\IsomGM$-equivariance of the discretized $\GM$-convolutions.%
\footnote{
    The icosphere grid, used by some of the models from Sections~\ref{sec:spherical_CNNs_fully_equivariant} and~\ref{sec:spherical_CNNs_azimuthal_equivariant}, is defined by projecting the nodes of this grid to unit radial distance from the origin, i.e. to~$S^2$.
    The models in this section do not assume this projection but convolve directly over the piecewise flat icosahedral geometry.
}
\citet{liu2018icoAltAz} proposed to represent icosahedral feature fields relative to the atlas of charts that is shown in Fig.~\ref{fig:ico_cutting}.
The charts have the advantage that they map the hexagonal grids on the icosahedron's faces to common square pixel grids.
Note, however, that orthonormal frames on the icosahedron are in this representation deformed, such that they are not orthonormal relative to the canonical Euclidean metric.
Hexagonal convolution kernels on the icosahedron are deformed accordingly and can be implemented in terms of square kernels which are masked such that two of their corners are filled with zeros.

The $\GM$-convolution by~\citet{liu2018icoAltAz} assumes frames that are all parallel and can therefore in the interior of the charts, where the kernel support does not extend over its boundaries, be implemented via a conventional Euclidean convolution.
At points that are close to an edge between different charts, the kernel accumulates features from beyond the cut.
As already discussed and visualized in Section~\ref{sec:mobius_experiment_main} and Fig.~\ref{fig:mobius_conv_numerical}, this is conveniently implemented via a transport padding operation which pads a border of parallel transported features around the array of square pixels before running the convolution operation.
For the trivial transport implicitly assumed by~\citet{liu2018icoAltAz}, this padding operation just copies a row of features at each edge without transforming them.
Since the authors assume the trivial structure group $G=\{e\}$, the hexagonal kernels remain unconstrained.

The implementation of~\citet{gaugeIco2019} is mostly similar, however, it differs crucially in that it uses Levi-Civita transporters and $\C6$-steerable kernels.
Instead of directly padding rows of pixels across edges, the Levi-Civita transport requires that the features are steered either by $g=e$ for all internal edges and the magenta edge or by an angle of $\pm\frac{2\pi}{6}$ over all edges with angle defect $\frac{2\pi}{6}$, with the sign depending on the transport direction.
\citet{gaugeIco2019} assume the regular representation of~$\C6$ as field type and constrain the convolution kernels to satisfy the corresponding steerability constraint.
After transport padding, their $\GM$-convolution is implemented as a conventional Euclidean convolution with these steerable kernels.
Note that this $\GM$-convolution is within the faces, i.e. except for the transport padding, similar to the HexaConv by \citet{Hoogeboom2018-HEX}.

Since the $\GM$-convolution by~\citet{zhang2019orientation} assumes a trivial structure group $G=\{e\}$, the transport padding is again implemented as a trivial copy of pixels without steering and the kernels are again left unconstrained.
However, as the frames of the $\{e\}$-structure are aligned towards the north pole, they are not longer parallel in the rectangular square pixel representation, which prevents an immediate implementation in terms of conventional convolutions.
Instead, the kernels have to be applied in a different rotation at each grid point.
As the hexagonal kernel can be rotated by $\frac{2\pi}{6}$ without using interpolation, and since the alignments towards the north pole differ at most by this angle from each other, the authors propose the following efficient approximating of this operation:
they convolve twice on each face, once with the original kernel and once with its $\frac{2\pi}{6}$ rotated version.
The two response fields are then linearly combined, with the precomputed interpolation weights depending on the angles of the north-aligned reference frames relative to the two kernel alignments (i.e. relative to the pixel grid).
This implementation is therefore approximately twice as costly as those in~\cite{liu2018icoAltAz,gaugeIco2019}.

An alternative implementation of spherical convolutions on the icosahedron was proposed by~\citet{eder2020tangent}.
The authors project the spherical signal on planes spanned by the 20 faces (denoted as tangent images) and subsequently run a conventional CNN on each of these images.
We did not include this network in our list as it processes these representations independently from each other, that is, it does not transport or pad features between them, and is therefore not exactly described as $\GM$-convolution.

As mentioned before, the empirical results by~\citet{kicanaoglu2019gaugeSphere} suggest that the icosahedral geometry approximates the spherical geometry reasonably well for deep learning applications.
More specifically, the authors compare their spherical CNN on an icosphere grid with the piecewise flat icosahedral CNN by~\citet{gaugeIco2019} and find that both perform similarly despite the deformed geometry of the latter.
The equivariance of icosahedral CNNs under continuous rotations in $\SO3$ is found to be violated significantly, however, this seems to be a mere overfitting effect as it is easily and without loss of model performance counteracted by leveraging $\SO3$ data augmentation.

As always, we want to mention that the $\C5$-equivariant CNNs by \citet{liu2018icoAltAz} and \citet{zhang2019orientation} can easily be made $\D5$-equivariant by considering a $\Flip$-structure and thus reflection-steerable kernels.
Similarly, the $\operatorname{I}$-equivariant CNN by \citet{gaugeIco2019} can be made equivariant under the full isometry group $\operatorname{I}_h$ of the icosahedron by making the kernels $\D6$-steerable instead of $\C6$-steerable.

%% file: chapters/120_mesh_intro.tex

\section{Coordinate independent CNNs on general surfaces}
\label{sec:instantiations_mesh}

Instead of operating on a fixed geometry, the $\GM$-convolutions in the current section are defined on general manifolds.
We restrict our review to surfaces ($d=2$) since we are not aware of implementations on (general) higher-dimensional manifolds.
The signals to be processed could either be directly given by the dataset or are computed from the surfaces' geometries.
Examples for the former would be color textures or physical quantities like temperature fields or the wall stress of a pressurized container.
The latter could for instance be Gaussian and principal curvatures, SHOT descriptors or wave kernel signatures.
Most applications so far focus on classifying the surfaces~\cite{huang2019texturenet,jin2018learning,Wiersma2020}, segmenting parts of them~\cite{poulenard2018multi,huang2019texturenet,Wiersma2020,Yang2020parallelFrameCNN} or finding correspondences between different surfaces~\cite{masci2015geodesic,boscaini2016learning,schonsheck2018parallel,Wiersma2020,deHaan2020meshCNNs}.
Further applications are the prediction of physical quantities like mechanical stress~\cite{sun2018zernet} or the synthesis of color textures~\cite{turk2001texture,ying2001texture} or geometric deformations~\cite{hertz2020GeomTextureSynthesis}.

The design of Euclidean and spherical CNNs is strongly guided by the requirement for global symmetry equivariance.
Since general surfaces come usually with trivial isometry groups this guiding principle falls away, which leaves us with a large freedom in the choice of $G$-structures.
The models that we review in this section can be classified into \emph{rotation-steerable} and \emph{$\{e\}$-steerable} surface convolutions.
Both approaches address the issue of a missing canonical direction on surfaces, however, they do it in a fundamentally different way.
Rotation-steerable models account for the lack of reference direction by their equivariant design, treating all directions equivalently.
Their underlying $\SO2$-structure is -- up to a practically irrelevant choice of orientation%
\footnote{
    The chosen orientation is on a (connected, orientable) manifold arbitrary since the kernels are learned.
    If the opposite orientation was chosen, the training would just result in oppositely oriented kernels.
}
-- fixed by the Riemannian metric.
The rotation steerable models differ therefore mainly in their choice of field types.
The $\{e\}$-steerable models are non-equivariant and are therefore not associated to a (non-trivial) field type.
However, they differ from each other by the specific choice of $\{e\}$-structure that is used to determine the kernel alignments.

\etocsettocdepth{3}
\etocsettocstyle{}{} 
\localtableofcontents

This section is organized as follows:
we start in Section~\ref{sec:surfaces_geom_classical_smooth} with a (very) short introduction to the classical differential geometry of surfaces, discussing in particular the difference between their intrinsic and extrinsic geometry.
In practice, most implementations operate on discretized surfaces.
Section~\ref{sec:surfaces_geom_mesh} gives an overview of the geometry of triangular surface meshes, which are arguably the most common surface discretizations in the deep learning literature.
In Section~\ref{sec:so2_surface_conv} we discuss rotation-steerable surface convolutions.
Heuristics for fixing the frame fields that define $\{e\}$-steerable surface convolutions are reviewed in Section~\ref{sec:e_surface_conv}.

For completeness, we mention in the following paragraph a few alternative approaches to define surface convolutions
before coming to the actual content of this section.

\paragraph{Surface CNNs beyond \textit{GM}-convolutions:}

While quite some surface CNNs can be interpreted as $\GM$-convolutions, many alternative network designs have been proposed.
These methods rely for instance on
graph convolutions on surface meshes,
spectral approaches,
multi view renderings of surface embeddings,
volumetric methods in the embedding space,
differential operators,
or other operators which operate immediately on the mesh data structures.
The following brief review is intended to give an overview on the different directions which have been explored.

One method to classify or segment embedded surfaces is to \emph{render them from multiple viewpoints} and process the renderings with conventional Euclidean CNNs.
The resulting features are then aggregated by
pooling over the viewpoints~\cite{su2015multi,qi2016volumetric}
or via a consensus method~\cite{paulsen2018multi}.
\citet{esteves2019multiView} choose to place the camera viewpoints on a sphere according to a discrete subgroup of $\SO3$, for instance the icosahedral group.
The resulting features are then processed jointly via a discrete group convolution (not a surface convolution).

Instead of projecting the surface by rendering it, it can be projected to $\R^2$ by defining a \emph{chart}.
\citet{sinha2016deep} define approximately authalic (area preserving) global charts on spherical topologies.
These charts are discontinuous and in general not conformal (angle preserving).
A conventional Euclidean CNN is used to process the resulting images.
The discontinuities can be circumvented by pulling the surface features back along toric~\cite{maron2017convolutional} or more general~\cite{haim2018surface,benhamu2018multichart} covering maps.
The subsequent Euclidean convolution on the pullback can not be interpreted as a $\GM$-convolution since the sheets of the covering map induce different, incompatible $\{e\}$-structures on the surface.
\citet{li2019crossAtlas} use an atlas of (approximately) isometric charts -- as discussed at the end of Section~\ref{sec:e_surface_conv}, this corresponds indeed to a $\GM$-convolution.

\emph{Volumetric methods} process embedded surfaces with CNNs in the embedding space~$\R^3$, for instance by interpreting the vertices of a surface mesh as a \emph{point cloud}~\cite{qi2017pointnet,qi2017pointnet++,thomas2019kpconv} or by \emph{voxelizing} the input.
Point cloud based methods are reviewed in~\cite{guo2020deep}.
\citet{mescheder2019occupancyNets} and \citet{peng2020occupancyCNNs} argue that an implicit surface parametrization is more economical and propose networks which model surfaces as decision boundaries.

\emph{Spectral approaches} are inspired by the convolution theorem.
The Fourier basis on a manifold is thereby given by the eigenfunctions of the Laplace-Beltrami operator.
Spectral neural networks process feature maps by manipulating their Fourier spectrum with learned linear operators.
As the Fourier basis is non-localized, \citet{boscaini2015learning} use instead a windowed Fourier transform; an alternative are the localized manifold harmonics of~\citet{melzi2018localized}.
\citet{bruna2013spectral} interpret surface meshes as graphs.
They are therefore applying graph Fourier transforms, which are based on the eigenfunctions of the graph Laplacian.

\citet{sharp2020diffusion} suggest a model which is based on \emph{differential operators}.
Scalar features are propagated via heat diffusion with a learnable diffusion time.
As the Laplacian (occurring in the heat equation) is isotropic, it can not respond selectively to pattern in specific rotations.
The authors are therefore additionally applying a gradient operator, followed by taking scalar products of the resulting tangent vector-valued features.
Note that both operations are gauge invariant.
The networks can be implemented on all data structures which admit partial differential operators, for instance point clouds or meshes.

Quite some networks do not operate on the \emph{Riemannian manifold structure} but rather on the \emph{data structure} which represents the surfaces numerically.
An example are networks which interpret the nodes and edges of a surface mesh as forming a graph and consequently apply \emph{graph networks}.
The isometry equivariance of graph networks was investigated in~\cite{khasanova2018isometric,horie2020isometric}.
\citet{verma2018feastnet} proposed a graph network with dynamic filters, i.e. filters that are during the forward pass predicted from the features.
The model of \citet{milano2020primaldual} operates on the primal and dual graphs of meshes and utilizes attention mechanisms.

Spiral nets process features on meshes via local spiral operators~\cite{lim2018simple,gong2019spiralnet++}.
These operators enumerate features by following a spiral path outwards from the central node.
A response is computed by applying a LSTM to the resulting sequence of features or an MLP to their concatenation.
The choice of first neighbor and spiraling direction corresponds to a choice of $\{e\}$-structure.
\citet{hanocka2019meshcnn} and \citet{hertz2020GeomTextureSynthesis} define convolutions on mesh faces and edges, respectively.
Both models are made invariant to the arbitrariness in the mesh element ordering, which could be generalized to a permutation equivariant design.

For more in-depth reviews of such methods we point the reader to \citet{bronstein2017geometric} and \citet{guo2020deep}.

%% file: chapters/121_mesh_geometry.tex

\subsection{Geometry of embedded surfaces}
\label{sec:surfaces_geom_main}

This section gives a brief introduction to the geometry of surfaces.
Some concepts of the differential geometry of \emph{smooth} embedded surfaces are discussed in Section~\ref{sec:surfaces_geom_classical_smooth}.
Section~\ref{sec:surfaces_geom_mesh} attempts to give an overview of possible ways to \emph{discretize} differential quantities on surface meshes.

For a more in depth treatment of parameterized surfaces, we refer the reader to \cite{gallier2011geomMethods}.
A concise and intuitive introduction to the topic and its relation to computational (discretized) geometry can be found in~\cite{craneDiscreteDifferentialGeometry2014}.

\subsubsection{Classical differential geometry of embedded surfaces}
\label{sec:surfaces_geom_classical_smooth}

Classically, surfaces have been described \emph{extrinsically}, that is, as being immersed (or embedded) in an Euclidean ambient space $\R^3$.
This immersion can be defined in multiple equivalent ways, for instance local parametrizations, Monge patches or implicit functions.
Local surface parametrizations are smooth maps
\begin{align}
    \chi:\, \R^2 \supset V \to M \subset \R^3
\end{align}
which immerse open subsets $V$ of $\R^2$ into the ambient space $\R^3$.
They are required to be regular, that is, their partial derivatives
\begin{align}
    e_i = \frac{\partial\chi}{\partial x_i}\ ,\qquad i=1,2
\end{align}
are required to be linearly independent in $\R^3$.
The derivatives $e_1(x_1,x_2) \in\R^3$ and $e_2(x_1,x_2) \in\R^3$
span the embedded tangent spaces $\TpM \subset \R^3$ at $p=\chi(x_1,x_2)$.%
\footnote{
    The derivative vectors $e_i$ correspond in the intrinsic chart formalism to \emph{coordinate bases}; see Appendix~\ref{apx:coord_basis_def}.
}
Surface normals in the embedding space are therefore well defined and given by $n = \frac{e_1 \times e_2}{\lVert e_1\times e_2\rVert}$.
An atlas of compatible local surface parametrizations allows to describe surfaces that differ topologically from the plane on a global level.

The Riemannian metric of the surface -- in this context often denoted as its \emph{first fundamental form} -- is induced from the embedding space.
In accordance with the analogous definition for the embedded sphere $S^2$ in Eq.~\eqref{eq:spherical_embedding_metric_explicit} we have:
\begin{align}\label{eq:surface_embedding_metric}
    \eta_p(v,w) \,:=\, \langle v,w \rangle_{\R^3} \qquad \forall\ v,w \in \TpM
\end{align}
Let $v = \sum_i \mathscr{v}_i e_i$ and $w = \sum_i \mathscr{w}_i e_i$ be tangent vectors in $\TpM$ that are expressed in terms of their coefficient vectors $\mathscr{v},\mathscr{w}\in\R^2$ relative to the coordinate basis.
The metric is relative to this basis represented by a symmetric coefficient matrix
\begin{align}
    \operatorname{I}
    \ =\ 
    \begin{pmatrix}
           E \mkern-8mu& F \\
           F \mkern-8mu& G
    \end{pmatrix}
\end{align}
with elements%
\footnote{
    In modern notation, the coefficients of a (coordinate free) metric $g$ relative to a given basis are often denoted by $g_{\mu\nu}$.
}
$E = \langle e_1, e_1 \rangle_{\R^3}$,
$F = \langle e_1, e_2 \rangle_{\R^3}
   = \langle e_2, e_1 \rangle_{\R^3}$ and
$G = \langle e_2, e_2 \rangle_{\R^3}$.
This matrix acts on vector coefficients according to $\eta_p(v,w) = \mathscr{v}^\top \operatorname{I}\mathscr{w}$.
The first fundamental form encodes the \emph{intrinsic} geometry of a surface as a two-dimensional Riemannian manifold, i.e. that part of the geometry which is independent of its immersion into the ambient space.

\begin{figure}
    \centering
    \includegraphics[width=1.\textwidth]{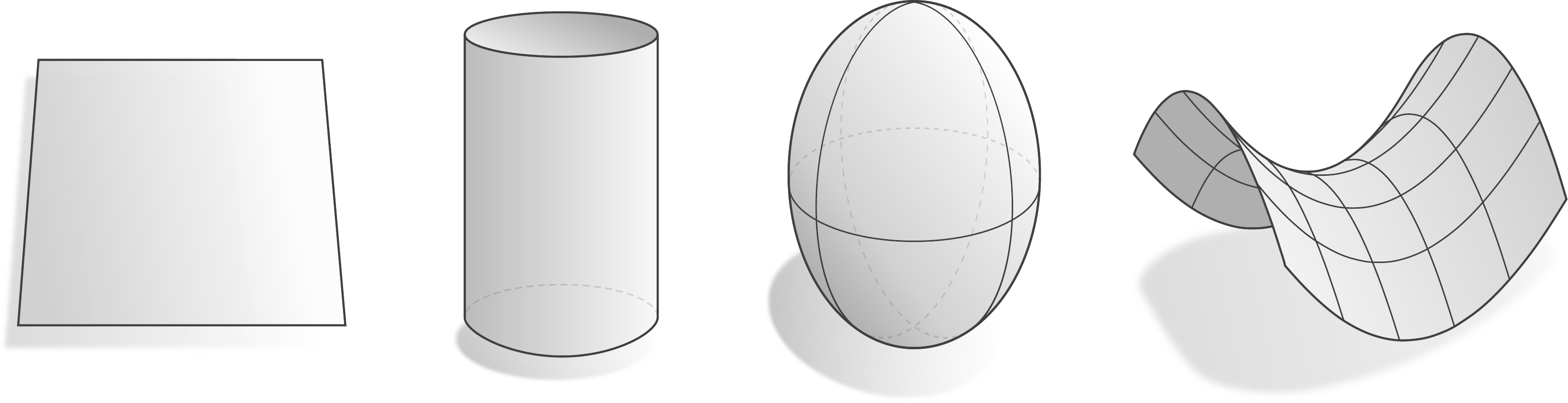}
    \caption{\small
        Embedded surfaces of qualitatively different extrinsic curvatures.
        \textit{Left:}
            The plane is characterized by vanishing principal and Gaussian curvatures
            $\kappa_{\max} = \kappa_{\min} = \kappa_{\textup{Gauss}} = 0$.
        \textit{Middle left:}
            A cylinder has one direction of positive curvature and one of vanishing curvature, i.e. 
            $\kappa_{\max} > 0$ and $\kappa_{\min} = 0$.
            Its Gaussian curvature $\kappa_{\textup{Gauss}} = 0$ is therefore zero as well.
            The plane and the cylinder are locally isometric, that is, their intrinsic geometry is locally indistinguishable.
            Note that the plane can be rolled up (developed) to form a cylinder -- the difference between the two is just the embedding into ambient space.
        \textit{Middle right:}
            An ellipsoid is characterized by its positive principal and Gaussian curvatures $\kappa_{\max} > 0$, $\kappa_{\min} > 0$ and $\kappa_{\textup{Gauss}} > 0$ at every point.
        \textit{Right:}
            The surface of a saddle bends in opposite directions, implying opposite signs of the principal curvatures $\kappa_{\max} > 0$ and $\kappa_{\min} < 0$.
            As a result, the Gaussian curvature $\kappa_{\textup{Gauss}} < 0$ is negative.
     }
    \label{fig:curvature_surfaces}
\end{figure}

A surface's \emph{extrinsic} geometry, i.e. details about its particular immersion into the ambient space, is captured by its \emph{second fundamental form}.
Relative to $e_1$ and $e_2$ this form is represented by the matrix
\begin{align}
    \operatorname{I\!I}
    \ =\ 
    \begin{pmatrix}
           L \mkern-8mu& M \\
           M \mkern-8mu& N
    \end{pmatrix}
\end{align}
with elements
$L = \big\langle n, \frac{\partial^2\chi}{\partial x_1^2} \big\rangle_{\R^3}
   = \big\langle n, \frac{\partial e_1}{\partial x_1} \big\rangle_{\R^3}$,
$M = \big\langle n, \frac{\partial^2\chi}{\partial x_1 \partial x_2} \big\rangle_{\R^3}
   = \big\langle n, \frac{\partial e_1}{\partial x_2} \big\rangle_{\R^3}
   = \big\langle n, \frac{\partial e_2}{\partial x_1} \big\rangle_{\R^3}$
and
$N = \big\langle n, \frac{\partial^2\chi}{\partial x_2^2} \big\rangle_{\R^3}
   = \big\langle n, \frac{\partial e_2}{\partial x_2} \big\rangle_{\R^3}$.
These elements measure essentially how the coordinate bases -- and thus tangent spaces -- bend in ambient space (into the normal direction) when moving along the coordinate lines.
It can for instance be used to determine the \emph{normal curvature}
\begin{align}
    \kappa_n(v)\ =\ \frac{\mathscr{v}^\top \operatorname{I\!I}\mathscr{v}}{\mathscr{v}^\top \operatorname{I}\mathscr{v}}
\end{align}
of the surface at $p$ in direction of $v = \sum_i \mathscr{v}_i e_i \in \TpM$.
Intuitively, this normal curvature can be understood as the curvature of the curve defined by the intersection of the surface with the plane spanned by the direction $v$ and the normal $n$ at that point.
This curvature agrees with the inverse radius $r=1/\kappa_n(v)$ of the osculating circle to the curve at $p$, and therefore measures how the surface bends into the normal direction when moving in the direction of $v$; see~\cite{craneDiscreteDifferentialGeometry2014} for great visualizations of this situation.
Other quantities of interest in the study of immersed surfaces are their principal, mean and Gaussian curvatures, which can be expressed in terms of the normal curvatures and are exemplified in Fig.~\ref{fig:curvature_surfaces}.
The directions (unit vectors in $\TpM$) $v_{\max}$ and $v_{\min}$ in which the normal curvature at a given point $p$ are maximal or minimal are denoted as \emph{principal directions} at~$p$.
The corresponding curvatures
\begin{align}
    \kappa_{\max} = \kappa_n(v_{\max})
    \qquad \textup{and} \qquad
    \kappa_{\min} = \kappa_n(v_{\min})
\end{align}
are the \emph{principal curvatures} at~$p$.
Their mean value
\begin{align}\label{eq:mean_curvature}
    \kappa_{\textup{mean}} \ =\ \frac{\kappa_{\max}+\kappa_{\min}}{2}
\end{align}
is known as \emph{mean curvature}.
The mean curvature is zero at ``saddle-like'' points where $\kappa_{\min} = -\kappa_{\max}$.
Minimal surfaces have zero mean curvature at every point.
The product
\begin{align}
    \kappa_{\textup{Gauss}}\ =\ \kappa_{\max} \cdot \kappa_{\min}
\end{align}
of the principal curvatures is denoted as \emph{Gaussian curvature}.
This curvature is positive if the principal curvatures have the same sign, which is for instance the case for ellipsoids.
In order for the Gaussian curvature to be negative, the signs of the principal curvatures need to differ, as around hyperbolic (saddle-like) regions.
The Gaussian curvature is zero if either (or both) of the principal curvature values is (are) zero, i.e. if the surface has a flat direction.
An example for a manifold with zero Gaussian curvature is the cylinder.
Such surfaces are said to be \emph{developable}, which means that they can be flattened out into a plane without being distorted, or, more rigorously formulated, they are locally isometric to the plane.
Carl Friedrich Gauss proved in his \emph{theorema egregium} that the Gaussian curvature of a surface is actually an intrinsic property, i.e. that it does not depend on how the surface is immersed into ambient space.
It is in one-to-one correspondence with the (intrinsic) Riemannian curvature tensor of a surface (and thus also to its Ricci and scalar curvature).
An important property of the Gaussian curvature is that its integral over a topological disk $D \subset M$ equals the holonomy $\delta_{\partial\mkern-1mu D}$, i.e. the angle by which a vector is rotated when (Levi-Civita) transporting it once around the disk boundary~$\partial\mkern-1mu D$:
\begin{align}\label{eq:gauss_curvature_holonomy_smooth}
    \int_{D} \kappa_{\textup{Gauss}}\, dp\ =\ \delta_{\partial\mkern-1mu D}
\end{align}
As we will see below, this relation can be used to generalize the Gaussian curvature to meshes, where the holonomy $\delta_{\partial\mkern-1mu D}$ agrees with the angle defect of an unfolded loop of faces (like the blue neighborhood in Fig.~\ref{fig:ico_neighborhoods}).

Since $\GM$-convolutions depend only on the intrinsic geometry of a surface, the reader might wonder why we are discussing their extrinsic properties like principal curvatures.
The reason is that $\GM$-convolutions may nonetheless be informed about a surface's extrinsic geometry, for instance by encoding it in feature fields.
The extrinsic geometry may furthermore be used to heuristically align the frames of an $\{e\}$-structure and thus kernels.
For example, \citet{jin2018learning} and \citet{li2019crossAtlas} align the frames along the $z$-axis of the ambient space $\R^3$, while \citet{boscaini2016learning} and \citet{tatarchenko2018tangent} align the frames along the surface's dominant principal curvature direction.
Note that these heuristics are not always well defined:
for instance, the projection of the $z$-axis on a ``horizontal'' tangent space (in the ambient space) is zero,
and the dominant principal curvature direction might not be defined, as it is the case on the sphere.

\subsubsection{Discretized geometry of surface meshes}
\label{sec:surfaces_geom_mesh}

In principle, it would be possible to describe $\GM$-convolutions on local surface parametrizations as described in the last section.
While this approach might be suitable for certain simple or symmetric geometries like ellipsoids, hyperboloids or tori, it seems impractical for more complex geometries.
In practice, surfaces come mostly discretized, for instance in form of triangle meshes, quad meshes, halfedge meshes, subdivision surfaces or point clouds.
Due to their widespread use -- both in general and specifically in the description of the surface $\GM$-convolutions that we review in the next two sections -- we will in the following focus primarily on triangle meshes.
Our goal for the remainder of the current section is therefore to take the quantities and definitions from the smooth theory and discuss their discrete counterparts on triangle meshes.
Unfortunately, these discrete analogues are usually not unique, such that a plethora of inequivalent definitions exists.%
\footnote{
    \citet{meyer2003discrete} describe this situation as follows:
    \emph{``Despite extensive use of triangle meshes in Computer Graphics, there is no consensus on the most appropriate way to estimate simple geometric attributes such as normal vectors and curvatures on discrete surfaces.''.}
    Similarly, \citet{craneDiscreteDifferentialGeometry2014} claims:
    \emph{``There is no one “right” way to discretize a given geometric quantity, but rather many different ways, each suited to a particular purpose.''}
}
We will in the following try to give a general idea about some of the most common approaches of discretizing the smooth geometry of surfaces in terms of triangle meshes.

\paragraph{Topology, geometry and embedding of triangle meshes:}
Triangle meshes $(\mathcal{V},\mathcal{F})$ are commonly encoded in terms of a set
\begin{align}
    \mathcal{V}\subseteq\N
\end{align}
of \emph{vertices} and a set
\begin{align}
    {\mathcal{F} \subseteq \big\{\mkern-2mu \{i,j,k\} \mkern2mu\big|\mkern2mu i\neq j\neq k \in\mathcal{V} \big\}}
\end{align}
of triangular \emph{faces},
satisfying that each vertex is contained in at least one of the faces.%
\footnote{
    Faces may alternatively be defined as ordered 3-tuples of vertices.
    The ordering of the vertices (or rather the equivalence classes of orderings under an even number of permutations) may then be used to encode the faces' orientations.
    We will instead encode face orientations as in our smooth theory by a choice of handedness of reference frames.
}
A set
\begin{align}
    \mathcal{E} = \big\{ \{i,j\} \,\big|\ i\neq j\in \{i',j',k'\} \textup{ for some } \{i',j',k'\} \in \mathcal{F} \big\}
\end{align}
of \emph{edges}, bounding the faces, follows immediately.
In practice, one is often given a set
\begin{align}
    P = \big\{ p_i \in \R^3 \,\big|\, i\in\mathcal{V} \big\}
\end{align}
of vertex positions, specifying an \emph{embedding} of the mesh in the ambient space~$\R^3$.
This embedding implies lengths
\begin{align}
    l_{\{i,j\}} = \lVert p_j-p_i \rVert
\end{align}
of edges $\{i,j\}$ and areas
\begin{align}
    \textup{A}_{\{i,j,k\}} = \frac{1}{2} \big\lVert {(p_j-p_i)} \times (p_k-p_i) \big\rVert
\end{align}
of faces $\{i,j,k\}$.

We are specifically interested in \emph{surface meshes}, which are required to satisfy additional conditions.
In order to formulate these conditions, note that the mesh elements $\{ i_0, \dots, i_n \}$
(where $n=0,1\ \text{or}\ 2$ for vertices, edges or faces)
imply $n$-\emph{simplices}, defined as the convex hulls
\begin{align}
    \operatorname{convex} \!\big(\{ i_0, \dots, i_n \}\big)\ :=\ 
    \Big\{ \sum\nolimits_{j=0}^n \alpha_j\mkern2mu p_{i_j} \,\Big|\, 
    \sum\nolimits_{j=0}^n \alpha_j = 1\,\ \textup{and}\,\ \alpha_j\geq0\,\ \forall\,j=0,\dots,n \Big\}
    \ \ \subset\ \R^3 \,.
\end{align}
The set that comprises all of these simplices (mesh elements) forms a \emph{pure $2$-simplicial complex}~\cite{desbrun2005DiscreteExteriorCalculus,craneDiscreteDifferentialGeometry2014}.
That the 2-simplicial complex is \emph{pure} means that each 0-simplex (vertex) and 1-simplex (edge) is a subset of at least one 2-simplex (face).
In other words, there are no disconnected vertices or edges in the mesh.
The \emph{underlying space}
\begin{align}
    \bigcup_{\{ i_0, \dots, i_n \} \in \mathcal{V} \cup \mathcal{E} \cup \mathcal{F}}
    \mkern-40mu
    \operatorname{convex} \!\big( \{ i_0, \dots, i_n \}\big)
    \ \ \overset{\textup{(pure)}}{=}\,
    \bigcup_{\{i,j,k\} \in \mathcal{F}}
    \mkern-10mu
    \operatorname{convex} \!\big( \{i,j,k\}\big)
    \ \ \ \subset\ \R^3
\end{align}
of the simplicial complex is defined as the union of all of its simplices, equipped with the usual topology as a subset of~$\R^3$.
A mesh is then said to be a surface mesh (manifold mesh) if the underlying space is a topological surface (manifold), optionally with boundary.
Intuitively, this requires
1) that each edge is adjacent to two faces (or one at boundaries) and
2) that the faces around each vertex form a topological disk (or a half-disk at boundaries).

Such defined surface meshes are discrete counterparts to \emph{embedded} Riemannian surfaces.
However, since $\GM$-convolutions are independent from the \emph{extrinsic} geometry of the underlying manifold, it is instructive to briefly discuss their \emph{intrinsic} geometry.
Take therefore the vertex, edge and face sets $\mathcal{V}$, $\mathcal{E}$ and $\mathcal{F}$, but discard the embedding locations $P$ of the vertices.
Together, these sets form an \emph{abstract 2-simplicial complex}
$\mathcal{V} \cup \mathcal{E} \cup \mathcal{F} \,,$
defined as a family of abstract simplices $\{i_0,\dots,i_n\}$ that is closed under taking subsets~\cite{craneDiscreteDifferentialGeometry2014}.
If this (now abstract) 2-simplicial complex is
1) pure and
2) such that the ``Star'' of every vertex (given by the simplices containing that vertex) form a combinatorial disk,
it forms an \emph{abstract simplicial surface} (which is exactly the case if the embedded mesh is a surface mesh).
Abstract simplicial surfaces can be viewed as combinatorial counterparts of topological manifolds.
They admit to compute topological invariants, for instance the Euler characteristic
\begin{align}
    \mathcal{X}_{\textup{Euler}} = |\mathcal{V}|-|\mathcal{E}|+|\mathcal{F}| \,.
\end{align}
As a topological invariant, the Euler characteristic agrees for any two homeomorphic spaces, in particular for a smooth manifold and any of its triangulations.
For instance, the icosahedron from Section~\ref{sec:spherical_CNNs_icosahedral} has $\mathcal{X}_{\textup{Euler}}^\textup{ico} = 12-30+20 = 2$, which agrees with $\mathcal{X}_{\textup{Euler}}^{S^2} = 2$ for the 2-sphere.

To arrive at an intrinsic description of a triangulated surface's \emph{geometry}, one assigns \emph{edge lengths} ${l_{ij} \in \R^+}$ to edges $\{i,j\} \in \mathcal{E}$.
For consistency, these lengths are required to satisfy the triangle inequality $l_{\{i,j\}} + l_{\{j,k\}} > l_{\{k,i\}}$ for any face $\{i,j,k\} \in \mathcal{F}$.
The edge lengths imply Euclidean metrics (distance functions) on the faces, and therefore a piecewise defined Euclidean metric on the whole surface.
It corresponds to a Riemannian metric (or first fundamental form) which is Euclidean away from vertices and ``cone-like'' (singular) on a small neighborhood around the vertices~\cite{Crane2020DiscreteConformalGeometry,desbrun2005DiscreteExteriorCalculus}.

In order to close the circle to our initial, extrinsic definition of triangle meshes, on needs to embed the mesh into ambient space~$\R^3$.
The necessary information on the extrinsic geometry is given by equipping the mesh with a \emph{second fundamental form}.
In the discrete setting, this form can be defined as a choice of dihedral angle (bending angle) between any two adjacent triangles, i.e. one angle per non-boundary edge of the mesh.
Provided that this data is chosen consistently%
\footnote{
    The discrete first and second fundamental forms are required to satisfy an integrability condition,
    similar to the Gauss's equation and the Mainardi-Codazzi equations in the smooth setting~\cite{wang2012surfaceReconstruction}.
},
it is possible to reconstruct the embedding, i.e. vertex positions~$P$, up to rigid motions in~$\E3$~\cite{lipman2005linear,wang2012surfaceReconstruction}.
While an embedding of the surface is not necessary for the intrinsic $\GM$-convolutions, all of the papers listed in rows (37-41) of Table~\ref{tab:network_instantiations} evaluate their models on embedded triangle meshes.

\paragraph{Tangent spaces and vector fields:}
To describe vector fields on meshes, and to equip the meshes with geometric structure like connections, it is necessary to define a notion of tangent spaces that are attached to them.
Multiple incompatible definitions, tailored towards the specific application in mind, occur in the literature.
Since vector fields are commonly sampled at discrete locations, the discrete tangent bundles are often only partially defined, for instance only on faces, edges or vertices.
We briefly review some of these definitions, a more detailed survey can be found in~\cite{deGoes2016VectorFieldProcessing}.

Since the faces (2-simplices) of an embedded mesh are flat, one can naturally define their tangent spaces as those two-dimensional subspaces of~$\R^3$ in which they are contained~\cite{craneTrivialConnectionsDiscrete2010,craneDiscreteDifferentialGeometry2014,wang2012surfaceReconstruction}.
Specifically, given a face $\{i,j,k\} \in \mathcal{F}$, one may define the tangent spaces $\TpM = \operatorname{span}(p_j-p_i,\, p_k-p_i) \subset \R^3$ for every $p\in \operatorname{convex}\!\big(\{i,j,k\}\big)$ as the \emph{linear span of any two edge vectors}.
The alignment of the tangent space in ambient space is often represented in terms of the face normal $n = (p_j-p_i) \times (p_k-p_i)$.
Discrete tangent (or feature) vector fields can in face based representations be defined as being face-wise constant, i.e. represented by a single tangent (or feature) vector per face.
Relative to a choice of reference frame on each of the faces, such tangent and feature vector fields are encoded by $2|\mathcal{F}|$ or $c|\mathcal{F}|$ vector coefficients, respectively.
Note that such vector fields do not extend to vertices or edges.
Due to their discontinuity, the notion of differential operators, acting on such fields, is quite limited~\cite{deGoes2016VectorFieldProcessing} (which is irrelevant for our specific application).
A linear interpolation scheme of face based vector fields was proposed in~\cite{li2006representing}.

As there is no natural normal direction at the vertices of a mesh, there are multiple common definitions of vertex tangent spaces.
\emph{Vertex normals} can for instance be defined as an area weighted average of the adjacent faces' normals~\cite{lipman2005linear,lai2009metric,deHaan2020meshCNNs}.
Besides area weighting, uniform weights or tip angle weights are sometimes used~\cite{craneDiscreteDifferentialGeometry2014}.
Another option is to define normal vectors via a mean curvature normal operator \cite{meyer2003discrete}.
The resulting normal agrees with normals derived via area gradients, but differs from those that are derived via volume gradients or sphere inscribed normals; see~\cite{craneDiscreteDifferentialGeometry2014}.

Alternatively, one can define vertex tangent spaces in an \emph{intrinsic} way, simply by defining them as two-dimensional vector spaces that are attached to the vertex.
Their relation to the mesh geometry in a local neighborhood around the vertex is hereby encoded by representing the one-ring neighborhood in the tangent planes.
The arguably most prominent of such approaches is based on a rescaling of the total angle
\begin{align}\label{eq:mesh_total_incident_angle}
    \Theta_i \,=\! \sum_{\{i,j,k\}\in\mathcal{F}} \theta_{\{i,j,k\}}^i \,,
\end{align}
which is summed from the tip angles
$\theta_{\{i,j,k\}}^i = \arccos \big\langle \frac{p_j-p_i}{\lVert p_j-p_i\rVert} \frac{p_k-p_i}{\lVert p_k-p_i\rVert} \big\rangle$
of all the triangles $\{i,j,k\}$ adjacent to vertex $i\in\mathcal{V}$.
If this angle is exactly $2\pi$, the local neighborhood around the vertex is intrinsically flat; see for instance the red or green neighborhood in Fig.~\ref{fig:ico_neighborhoods}.
An angle $\Theta_i < 2\pi$, as for the blue neighborhood, signals a positive discrete Gaussian curvature (properly defined below) $\kappa_{\textup{Gauss},i} = 2\pi - \Theta_i$, i.e. a cone-like neighborhood.
An angle $\Theta_i > 2\pi$ corresponds similarly to a saddle-like neighborhood with negative Gaussian curvature.
The approach followed in
\cite{polthier1998straightest,zhang2006vectorFieldDesign,Knoppel:2013:GOD,Sharp2019VectorHeatMethod,craneDiscreteDifferentialGeometry2014}
is then to \emph{flatten} the one-ring neighborhood out by isotropically \emph{rescaling polar angles} by a factor of $s_i = \frac{2\pi}{\Theta_i}$ to the total $s_i \Theta_i = 2\pi$ of a Euclidean (tangent) space.
A vector field can is in this setting be represented by one vector per vertex.
A choice of gauge, which are often aligned with one of the edges, allows then to encode tangent or feature vector fields in terms $2|\mathcal{V}|$ or $c|\mathcal{V}|$ coefficients, respectively.
\citet{zhang2006vectorFieldDesign} proposed to interpolate the vectors with a piecewise linear hat function weighting from the vertices to the faces.
The direction of the vectors is thereby determined by a usual Euclidean transport on the flattened tangent spaces.
As pointed out by~\citet{deGoes2016VectorFieldProcessing}, this interpolation is not continuous.
To resolve this issue, the same authors propose in \cite{liu2016discreteConnection} to define a smooth structure on the triangle mesh and to represent the one-ring neighborhoods in \emph{smooth charts}.
A smooth interpolation is then performed by transporting vectors via a smooth simplicial connection on the mesh, which is optimized to be as close as possible to the original embedding space induced Levi-Civita connection.
Note that both approaches are effectively flattening the geometry around the vertices, that is, they do not exactly operate on the triangle mesh.

Yet another approach, rooted in discrete exterior calculus~\cite{desbrun2005DiscreteExteriorCalculus,elcott2005building}, is to define tangent vectors $v = \omega^\sharp \in \TpM$ in terms of \emph{1-forms} $\omega \in \TspM$ by leveraging the (metric-dependent) musical isomorphism $\sharp^\eta: \TsM \to \TM$ (``index raising'').
Since simplicial 1-forms are naturally assigned to \emph{edges} (1-simplices), this leads to an vector fields which are parameterized in terms of one vector per edge and thus $2|\mathcal{E}|$ coefficients after choosing frames.
However, as argued by \citet{deGoes2016VectorFieldProcessing} a piecewise linear interpolation of the vectors over the faces will again lead to discontinuities.
It is furthermore not clear to us how this approach could be generalized to general associated vector bundles and thus feature fields.

Given any of the above constructions of tangent spaces, \emph{local reference frames} are readily defined as 2-tuples of linearly independent tangent vectors.
A common choice is thereby to align the first frame axis with one of the adjacent edges of the current simplex (mesh element).
Specifically for the case of orthonormal, right-handed frames, i.e. whenever $G\leq\SO2$, a choice of (oriented) edge determines a frame completely.
Tangent vectors are then often represented in polar coordinates, with the angle measured relative to the reference edge.
If the tangent spaces are modeled extrinsically, that is, as two-dimensional subspaces of the ambient space $\R^3$, it is most common to represent the frames explicitly as a 2-tuple of vectors in $\TpM \subset \R^3$.
The definitions of frames and gauges are then fully equivalent to those in 
Eqs.~\eqref{eq:embedding_gauge_map_orthonormal_frame} and~\eqref{eq:embedding_space_R3_frame}
in Section~\ref{sec:sphere_geometry}.

$G$-structures are, as usual, defined as bundles of frames, which are in each tangent space related through $G$-valued gauge transformations.
In the computer graphics community, there is a particular interest in \emph{$N$-direction fields} (or unit $N$-RoSy fields), which are there defined as a collection of $N$ unit vector fields, such that the $N$ vectors in each tangent space are spaced by an angle of~$2\pi/N$.
Since any unit vector implies on an oriented manifold a corresponding right-handed, orthonormal frame, $N$-direction fields are seen to be equivalent to $\CN$-structures.
An example is the $\C6$-structure on the icosahedron in Fig.~\ref{fig:G_structure_ico_3}, which effectively assigns $6$ unit directions to each point, except for the poles, where it has singularities of index $\frac{1}{6}$ (or angle $\frac{2\pi}{6}$).
The interactive design of smooth direction fields, with user defined singularities amongst other constraints, is an active field of research in the computer graphics community
\cite{li2006representing,ray2008nSymmDirectionField,lai2009metric,craneTrivialConnectionsDiscrete2010,Knoppel:2013:GOD,liu2016discreteConnection,Sharp2019VectorHeatMethod}.
Some of the surface $\GM$-convolutions that we review in the following section use such algorithms to compute a $\CN$-structure~\cite{huang2019texturenet,Yang2020parallelFrameCNN}.

\paragraph{Riemannian metric and isometries:}
Having a mesh equipped with tangent spaces, one can define a \emph{Riemannian metric} on it.
The most common case that of isometrically embedded meshes with tangent spaces modeled as two-dimensional subspaces of the ambient space~$\R^3$.
As described before in Eqs.~\eqref{eq:spherical_embedding_metric_explicit} and~\eqref{eq:surface_embedding_metric}, the metric is then induced by restricting the standard Euclidean inner product $\langle\cdot,\cdot\rangle_{\R^3}$ of the embedding space to the tangent spaces.

If the tangent spaces are modeled intrinsically, a metric can be fixed by choosing an $\O{d}$-structure, i.e. reference frames that are \emph{defined} to be orthonormal.
Somewhat less tautological, if one is given edge lengths~$l_{\{i,j\}}$, and therefore a piecewise defined Euclidean distance function on the surface as discussed above, the choice of $\O{d}$-structure is required to be compatible with these lengths.
Specifically, the logarithmic map should result in tangent vectors of (Riemannian) norm $|\log_p(q)| = \mathscr{d}$ if the points $p$ and~$q$ are separated by a Euclidean distance $\mathscr{d} \in \R^+$.
Note that this statement requires a consistent definition of Levi-Civita connection on the mesh, which we discuss further below.

\emph{Isometries} are intrinsically defined as usual, that is, as those mappings of the mesh to itself, which preserve the metric.
Extrinsically, the isometry group is comprised of those isometries $\phi \in \E2$ of the embedding space, which leave the mesh invariant.
Most of the papers in
rows (37-41)
of Table~\ref{tab:network_instantiations}
consider datasets whose meshes have a trivial isometry group.
However, local neighborhoods of the meshes are often nonetheless isometric (or approximately isometric) to each other, which was exemplified in Fig.~\ref{fig:suzanne_local_isometry}.
As discussed in Section~\ref{sec:isometry_groups}, the isometry equivariance of $\GM$-convolutions will still hold locally if the kernels' field of view is sufficiently small.

\paragraph{Connections, transporters and geodesics on triangle meshes:}
The last ingredient that we need to implement $\GM$-convolutions on meshes is the transporter pullback $\Expsp$ of feature fields, Eq.~\eqref{eq:transporter_pullback_in_coords}.
We are therefore required to know how to
1) parallel transport feature vectors over meshes and
2) compute geodesics on meshes, specifically the exponential map or, depending on the implementation, the logarithmic map.
All of these mappings depend ultimately on a choice of \emph{connection} on the mesh.
In the smooth setting, a connection is essentially a collection of infinitesimal transporters between adjacent tangent spaces.
One defines discretized connections on meshes therefore usually as transporters between adjacent mesh elements.
The particular choice of tangent bundle discretization, options of which were discussed above, influences the particular definition of connection.
In the following, we review some discretizations of connections found in the literature and explain how they can be used to compute transporters and geodesics.

\begin{figure}
    \centering
    \includegraphics[width=1.\textwidth]{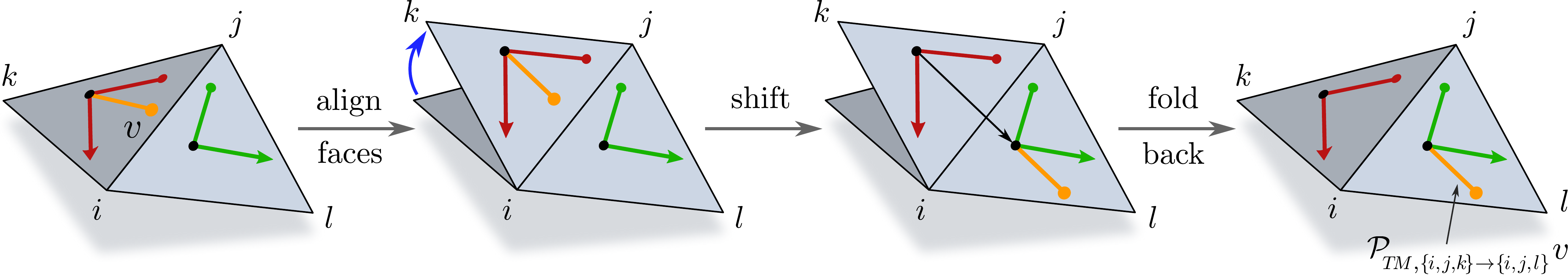}
    \caption{\small
        Parallel transport between mesh faces.
        The local geometry of two adjacent faces is developable, that is, it is intrinsically flat and can be unfolded into a plane.
        The Levi-Civita transport between the faces is therefore given by shifting a vector over the unfolded faces, followed by bending the faces back to their original embedding.
        This parallel transport between adjacent faces can be viewed as the discrete analog of the continuous Levi-Civita connection in the smooth setting~\cite{craneTrivialConnectionsDiscrete2010}.
        Given any choice of reference frames, the transport
        $\mathcal{P}_{\mkern-2mu\overset{}{\protect\scalebox{.62}{$\!T\!M$},\protect\scalebox{.68}{$\{i,j,k\mkern-1mu\}\!\to\!\{i,j,l\}$}}}$
        is represented by a group element
        $g^{A\widetilde{A}}_{\protect\scalebox{.68}{$\{i,j,k\mkern-1mu\}\!\to\!\{i,j,l\}$}} \in \GL{2}$ (or $\SO2$ when considering right-handed, orthonormal frames).
        More general connections apply an additional linear transformation to the coordinate free vector when transitioning between the faces.
        Alternative definitions of discrete connections, for instance for the transport between vertices along edges, are discussed in the main text.
    }
    \label{fig:transport_mesh}
\end{figure}

The simplest case to consider is the transport or connection between two adjacent faces.
Recall that the Levi-Civita transport on a flat plane is defined as shifting a vector such that it stays parallel in the usual Euclidean sense; see Fig.~\ref{fig:transport_flat}.
As connections are inherently intrinsic, they do not depend on the particular embedding of this plane into ambient space, which tells us how to transport on any developable surface.
It tells us in particular how to transport between two adjacent triangles, since they can be unfolded (developed) into a plane as visualized in Fig.~\ref{fig:transport_mesh}.
The Levi-Civita connection between faces can therefore be thought of as
1)~flattening the faces
2)~transporting the vector as usual on the plane and
3)~folding the faces back to their original embedding~\cite{craneTrivialConnectionsDiscrete2010,mitchell1987discrete,craneDiscreteDifferentialGeometry2014}.
The resulting transporter
$\mathcal{P}_{\mkern-2mu\overset{}{\protect\scalebox{.62}{$\!T\!M$},\protect\scalebox{.68}{$\{i,j,k\mkern-1mu\}\!\to\!\{i,j,l\}$}}}$
between the faces $\{i,j,k\}$ and $\{i,j,l\}$ can optionally be expressed in terms of a group element
$g^{A\widetilde{A}}_{\protect\scalebox{.68}{$\{i,j,k\mkern-1mu\}\!\to\!\{i,j,l\}$}}$ in~$\GL{2}$.
Since the Levi-Civita connection is a metric connection, it results for the specific case of orthonormal frames in group elements in~$\O2$, and for oriented orthonormal frames in $\SO2$-elements or rotation angles.

As proposed by \citet{craneTrivialConnectionsDiscrete2010}, it is possible to generalize this construction beyond Levi-Civita connections:
instead of merely shifting vectors between the flattened faces, more general connections apply an additional linear transformation, for instance an additional rotation.
While this additional transformation will be reflected by a corresponding transformation of the transporter's coordinate expression
$g^{A\widetilde{A}}_{\protect\scalebox{.68}{$\{i,j,k\mkern-1mu\}\!\to\!\{i,j,l\}$}}$,
it is conceptually independent from it and can be defined in a purely coordinate free setting.
The authors use this idea to construct smooth trivial connections, which are defined by having a transport of zero holonomy around any possible loop, and which are optimized to be as smooth as possible, except for at some singularities, which are topologically enforced~\cite{craneTrivialConnectionsDiscrete2010}.
They consider furthermore connections which apply (coordinate free) rotations by $\frac{2\pi}{N}$ and can be used to construct $N$-direction fields, corresponding to $\CN$-structures.
In our applications, we will \emph{always} consider Levi-Civita connections to compute geodesics.
The models reviewed in the following Section~\ref{sec:so2_surface_conv} assume a structure group $G=\SO2$ on oriented meshes and utilize Levi-Civita transporters for feature vectors.
In contrast, the models in Section~\ref{sec:e_surface_conv} assume a trivial structure group $G=\{e\}$ and therefore allow only for $\{e\}$-structure compatible trivial connections.
They transport features such that their coefficient vectors relative to the $\{e\}$-structure frames remain invariant, i.e. merely copy their numerical values.

Given embedded tangent spaces $\TpM \subset \R^3$ at other mesh elements like vertices or edges, this approach is naturally generalized to transitions between arbitrary mesh elements~\cite{deHaan2020meshCNNs}:
instead of aligning the faces, one could e.g. align the vertex tangent space with the adjacent face before shifting the vector.
Geometrically, this operation can be thought of as the transport over a mesh whose vertices and edges are cut off in an infinitesimal neighborhood, an are replaced with a polygonal face.

An alternative definition of discrete connections is given in \cite{Knoppel:2013:GOD} and \cite{Sharp2019VectorHeatMethod}.
The authors of both papers model tangent spaces only at the vertices, where they are defined in terms of the rescaling of the total incident angle, Eq.~\eqref{eq:mesh_total_incident_angle}, to $2\pi$, as discussed above.
A connection on the mesh is then given by transporters over all edges $\{i,j\} \in\mathcal{E}$, which link the adjacent vertices' tangent spaces.
Since the geometric notion of unfolding triangles is hereby missing, the edge transporters are encoded via group elements relative to a source and reference frame.
Specifically for the Levi-Civita connection, and orthonormal, right-handed reference frames, these group elements lie in~$\SO2$.
The utility of this construction for the direct transport along arbitrary paths over the manifold is unclear, however, it is useful to solve PDEs that depend on the covariant derivative.
\citet{Sharp2019VectorHeatMethod} showed that a solution of the vector heat equation allows nonetheless to use such connections to (indirectly) compute the parallel transport between arbitrary points on a mesh.
\citet{liu2016discreteConnection} propose yet another construction, namely smooth simplicial connections between and within all mesh elements.
They discuss furthermore how such connections can be optimized to be as close to the (non-smooth) Levi-Civita connection as possible.

A given connection determines the \emph{parallel transport} along a path.
In the smooth setting, where connections are infinitesimal transporters, the finite transport is computed by integrating the connection along the path.
In the discrete setting, the transport is accordingly given by composing the individual transformations that constitute the connection between the mesh elements that are crossed by the path.
For the Levi-Civita connection, this process corresponds to a flattening of all the mesh elements along the path, followed by shifting the vector over it; see Fig.~7 in~\cite{lai2009metric}.
The vector heat equation based method by \citet{Sharp2019VectorHeatMethod} computes the transport of vectors specifically along geodesics.
Since it solves for the transport from a source location to \emph{any} other location on the manifold simultaneously, this approach can be more efficient than integrating the transport for every single path individually.

The \emph{curvature of a connection} is in the smooth setting defined as the holonomy of its transport around an infinitesimally small disk.
The curvature at a vertex is in the discrete setting similarly defined as the holonomy of the transport around this vertex.
For the Levi-Civita connection, this is just the Gaussian curvature, which is given by the angle defect
\begin{align}
    \kappa_{\textup{Gauss},i} \,=\, \delta_i \,=\, 2\pi - \Theta_i \,,
\end{align}
where $\Theta_i$ is the total tip angle from Eq.~\eqref{eq:mesh_total_incident_angle}.
We refer again to the icosahedron as example, which has vanishing curvature everywhere, except for its original twelve vertices, where the angle defect (curvature) equals~$\frac{2\pi}{6}$.
Trivial connections have by construction zero curvature.

Lastly, we need to discuss \emph{geodesics}.
In the smooth setting, geodesics are defined as \emph{straightest paths}, which is formalized by the statement that the covariant derivatives of their tangent vectors along the curve vanish, that is, $\nabla_{\dot{\gamma}} \dot{\gamma} = 0$.
This is equivalent to the requirement that the transport of a tangent vector $\dot{\gamma}(t_0)$ along the geodesic remains tangent to it, i.e.
$\mathcal{P}_{\mkern-2mu\overset{}{\protect\scalebox{.6}{$\!T\!M$}, \gamma(t_1) \leftarrow \gamma(t_0)}}
 \dot{\gamma}(t_0) = \dot{\gamma}(t_1)$
for arbitrary $t_0$ and $t_1$.
Furthermore, the \emph{shortest path} between any two points on a connected manifold is given by a geodesic.
As pointed out by \citet{polthier1998straightest}, this equivalence of shortest and straightest paths does not longer hold on meshes, such that one needs to distinguish between the two concepts.

Recall that the \emph{exponential map} $\exp_p: \TpM \to M$ is defined as mapping vectors $v$ to that point which is reached when walking for a distance of $\lVert v\rVert$ from $p$ along the (unit speed) geodesic in direction of $v$.
This concept is readily generalized to meshes, where one follows the \emph{straightest geodesic} in the direction of $v$ for distance $\lVert v\rVert$.
As in the smooth setting, one may define such straightest geodesics on meshes as those curves that keep their tangent vector parallel to the curve.
This property is naturally satisfied on the planar faces (or along edges), such that the resulting geodesic is \emph{piecewise linear}, with the only nontrivial points being those where the geodesics transitions between adjacent mesh elements.
The outgoing direction of the geodesic after such a transition is thereby determined by the connection, i.e. by the transport of the incoming tangent direction to the next mesh element.
If one considers the Levi-Civita connection, which we always do to compute geodesics, this results in an ordinary straight line after unfolding the mesh elements into a plane.
To implement the discrete exponential map, it is sufficient to trace out such a straightest geodesic until reaching the distance $\lVert v\rVert$.

\emph{Logarithmic maps} $\log_p: M \to \TM$, on the other hand, can be thought of as computing the \emph{shortest geodesics} between points $p$ and $q$.
They return that vector $\log_p(q)$ in $\TpM$ which is tangent to this geodesic at $p$ and whose norm equals the geodesic distance between the points.
A prominent way of computing geodesic distances from a source point (or set) $p$ is to solve the eikonal equation
\begin{align}\label{eq:eikonal_equation}
    |\nabla \tau| = 1
    \quad\textup{subject to}\quad
    \tau(p) = 0 \,,
\end{align}
where $\nabla$ denotes the covariant derivative.
The first part of this PDE enforces the natural requirement that the gradient of the distance function should be one, while the second part fixes the distance at the source to zero.
A Fast Marching algorithm, which solves the eikonal equation on triangle meshes, was proposed by \citet{kimmel1998computingGeodesics}.
Given the distance function $\tau$, the geodesic $\gamma$ between $p$ and any other point $q$ can be traced back by following the distance gradient starting from $q$, i.e. by solving the ODE
\begin{align}
    \dot{\gamma} = -\nabla \tau \,.
\end{align}
With this information, we know that $\lVert\log_p(q)\rVert = \tau(q)$, with the direction of $\log_p(q)$ given by geodesic path at $p$.
The solution by \citet{mitchell1987discrete} generalizes the Dijkstra algorithm for computing distances along edges of a graph to a continuous version, which can cross faces and therefore operate on meshes.
It computes a distance function by propagating a wavefront starting from~$p$.
The heat method by \citet{Crane2017HeatMethodDistance} computes geodesic distances by exploiting Varadhan’s formula, which establishes a connection to the heat kernel.
Their algorithm is essentially solving the heat equation $\dot{u} = \Delta u$ with initial condition $u_0 = \delta(p)$, i.e. it diffuses a ``heat spike'' from the source point $p$.
For short diffusion times, the gradient $\nabla u$ points exactly in the opposite direction of the geodesic distances' gradient.
Since it is known that the geodesic distance gradient has unit magnitude (Eq.~\eqref{eq:eikonal_equation}), one can compute the distance field from this information.
The method is substantially faster than previous algorithms.
\citet{Sharp2019VectorHeatMethod} generalize this method to the vector heat equation, which allows to diffuse vector-valued quantities instead of scalar heat.
The algorithm can be used to transport vectors from a source point (or set) over the whole manifold, but it also suitable for solving with high accuracy for logarithmic maps.

%% file: chapters/122_mesh_SO2.tex

\subsection{Rotation-steerable surface convolutions}
\label{sec:so2_surface_conv}

In this section we review the $\SO2$, $\CN$ and $\DN$-steerable surface convolutions that are listed in  rows (37-40) of Table~\ref{tab:network_instantiations}.
All of these models have in common that they address the ambiguity of reference directions on general surfaces via a locally rotation equivariant (or invariant) design, which distinguishes them from the $\{e\}$-steerable models discussed in the following Section~\ref{sec:e_surface_conv}.
Before discussing the individual models in detail, we start with a higher level overview of common design choices and possible numerical discretizations.

\paragraph{General remarks and overview:}
All of the models that are reviewed in this section operate on \emph{triangle surface meshes} and are rotation-steerable.
The continuous structure group $G=\SO2$ is for all models that assume regular field representations (rows (38) and (39)) discretized by cyclic groups~$\CN$, i.e. $N$ equally spaced directions.
The model by \citet{huang2019texturenet} assumes a more specific structure group $\D4$.
Note that the purely rotation-steerable architectures operate only on \emph{oriented surfaces} without violating the smoothness (continuity) of their inference.
Non-oriented surfaces require additional reflection-steerability, i.e. structure groups $\O2$ or $\DN$.
This requirement is often easily satisfiable with minor adaptations, most importantly by using further restricted kernel spaces.

In accordance with the definition of $\GM$-convolutions, the models parameterize features in the local neighborhood around each sampling point in terms of \emph{geodesic normal coordinates}.
Almost all of the models sample feature fields on the \emph{mesh vertices}; only \citet{huang2019texturenet} samples features densely on the mesh faces.
The continuous convolution integral in Eq.~\eqref{eq:gauge_conv_coord_expression}, which matches the features in geodesic normal coordinates with a steerable kernel, can be discretized in different ways.
The majority of models discretize this integral at a vertex $p\in\mathcal{V}$ as a summation over its neighboring vertices $\mathcal{N}_p \subset \mathcal{V}$.
Features from these vertices $q\in\mathcal{N}_p$ are then matched with the values of the continuous kernel at point $\psiTMp \log_p(q) \in \R^2$, where $\psiTMp^A$ is the gauge corresponding to the chosen reference frame at~$p$.
Together with the transport from $q$ to $p$, this results in the discretization
\begin{align}\label{eq:mesh_conv_neighbor_log_discretization}
    \fout^A(p)\ =\ \sum_{q\in\mathcal{N}_p} \textup{A}_q\, K\big(\psiTMp^A \log_p(q)\big)\, \rho\big( g^{A\widetilde{A}}_{p\leftarrow q} \big)\, \fin^{\widetilde{A}}(q) \,,
\end{align}
where $\textup{A}_q \in \R$ are suitably chosen area weights that sum to the total mesh area, ${\sum_{q\in\mathcal{V}} \textup{A}_q = \int_M 1 dp}$.
Common choices are barycentric area weights of the form
\begin{align}\label{eq:triangle_area_weights}
    w_q = \frac{1}{3} \sum_{\{i,j,q\}\in\mathcal{F}} A_{\{i,j,q\}} \,,
\end{align}
with the sum running over all triangles that are adjacent to vertex~$q$, or Voronoi areas~\cite{vouga2014lectures}.
Since the discretization in Eq.~\eqref{eq:mesh_conv_neighbor_log_discretization} sums over neighboring vertices, the algorithms compute log maps via \emph{shortest geodesics} between $q$ and~$p$; see Section~\ref{sec:surfaces_geom_mesh} and~\cite{polthier1998straightest}.

Instead of computing logarithmic maps of neighboring vertices, one can alternatively discretize the convolution integral on the kernel domain~$\R^2$.
The authors of \cite{masci2015geodesic} use an equiangular and equiradial binning of geodesic polar coordinates.
They compute the exponential map for each sampling point $(r,\varphi)$, that is, they shoot a \emph{straightest geodesic} (\cite{polthier1998straightest}) of length $r$ in direction $\varphi$ relative to the reference frame.
As these geodesics end in general in a face, the feature vectors from adjacent vertices need to be interpolated, 
for instance based on barycentric coordinates.
\citet{Yang2020parallelFrameCNN} approximate the geodesic neighborhood via a ``parallel transport unfolding'' algorithm~\cite{budninskiy2018parallel}.

Table~\ref{tab:network_instantiations} organizes the models by their respective \emph{field types}, i.e. by the group representations $\rho$ that specify their transformation laws under gauge transformations.
The only non-trivial field types used so far are (complex) \emph{irreducible representations} of $\SO2$ \cite{Wiersma2020} and \emph{regular representations} of $\SO2$, discretized by regular representations of a discrete subgroup $\CN$~\cite{poulenard2018multi,sun2018zernet,deHaan2020meshCNNs,Yang2020parallelFrameCNN}.
Regular representations of $\SO2$ act by definition on functions on $L^2(\SO2)$, that is, on features which assign ``one value per direction''.
In the discretized version, we have $L^2(\CN) \cong \R^{|\CN|} = \R^N$, where each of the $N$~dimensions of a regular feature vector corresponds to one of the directions in~$\big\{ k\frac{2\pi}{N} \big|\, k=0,\dots,N-1 \big\}$.
The correspondence to regular representations is in most of these papers implicit -- the network architectures are rather derived from a more intuitive viewpoint.
It turns out that the authors use only a subset of the complete space of steerable kernels that map between $\CN$-regular feature fields.
We substantiate this claim further below when discussing the models in detail.
A construction of the complete kernel space is given in \cite{Weiler2019_E2CNN}, a visualization can be found in Fig.~3 of~\cite{Weiler2018SFCNN}.
The remaining models are based on \emph{trivial representations}, i.e. \emph{scalar fields}.
One approach to compute scalar fields is to apply a kernel in $N$ directions, resulting in an intermediate $\CN$-regular feature field, followed by a pooling operation over the $N$ responses~\cite{masci2015geodesic,monti2017geometric,sun2018zernet}.
Since gauge transformations in $\CN$ will lead to a mere cyclic shift (a permutation) of the feature's direction channels, the pooling operations are \emph{invariant} under gauge transformations, i.e. result in scalar fields.
\citet{huang2019texturenet} uses immediately $\D4$-invariant kernels; see Fig.~\ref{fig:3x3_D4_invariant_kernel}.
As gauge transformation leave such kernels invariant, the resulting feature fields are invariant as well, i.e. scalar fields.

Lastly, we can compare the models by the \emph{feature transporters} that they assume.
All of the convolutional networks in \cite{Wiersma2020,poulenard2018multi,sun2018zernet,deHaan2020meshCNNs} assume the canonical \emph{Levi-Civita} transporters on the mesh.
As all of the models in \cite{masci2015geodesic,monti2017geometric,sun2018zernet,huang2019texturenet} rely on \emph{scalar fields} their parallel transport is trivial.
An alternative approach was followed by \citet{Yang2020parallelFrameCNN} who compute a $\CN$-valued connection on the mesh.
This connection is flat (trivial) everywhere except for at a few singularities with a holonomy of $k\frac{2\pi}{N}$ for some $k=0,\dots,N-1$ and fixed~$N$.
The authors optimize their $\CN$-valued connection such that it approximates the $\SO2$-valued Levi-Civita connection as close as possible; see also~\cite{craneTrivialConnectionsDiscrete2010}.
Note that this approach is similar to the local flattening of spherical CNNs into icosahedral CNNs ($N=6$) from Section~\ref{sec:spherical_CNNs_icosahedral} but applies to general meshes.

With these general remarks in mind, we focus on some more specific design choices that are made in the models.

\paragraph{Harmonic Surface Networks:}
The \emph{Harmonic Surface Networks} by \citet{Wiersma2020}, listed in row (37) of Table~\ref{tab:network_instantiations},
are a prototypical example of $\GM$-convolutions on meshes.
They generalize Harmonic Networks \cite{Worrall2017-HNET} -- whose features transform according to the complex irreps of $G=\SO2$ -- from the Euclidean plane to general curved spaces.
The authors define their convolution as in Eq.~\eqref{eq:mesh_conv_neighbor_log_discretization}, using the barycentric area weights from Eq.~\eqref{eq:triangle_area_weights}.
Levi-Civita transporters and logarithmic maps are computed via the vector heat method~\cite{Sharp2019VectorHeatMethod}, which is not restricted to triangle meshes but allows to apply the model to polygon meshes and point clouds.
The $\SO2$-equivariant nonlinearities used by the models act only on the absolute value of the complex features but leave their argument invariant.

As proven in \cite{lang2020WignerEckart,Weiler2019_E2CNN}, the $\SO2$-steerable kernel spaces that are used by the authors are complete over the complex field.
However, if the complex feature fields are implemented in terms of two channels that contain their real and complex parts, they should rather be viewed as transforming according to the real irreps of~$\SO2$.
The kernel constraint allows in this case for additional steerable kernels; see Appendix~F.5 of~\cite{Weiler2019_E2CNN} for a detailed discussion.
We furthermore want to mention that empirical evidence suggests that networks which are based on irrep fields perform significantly worse that those that are based on regular representations; see e.g. the benchmark in~\cite{Weiler2019_E2CNN}.
Note that Harmonic Surface Networks can easily be turned into networks that operate on regular feature fields by employing the ``regular nonlinearity'' from~\cite{deHaan2020meshCNNs}, which essentially applies a Fourier transformation of a stack of irrep fields to transform them into a regular feature field.

\begin{figure}
    \centering
    \includegraphics[width=1.\textwidth]{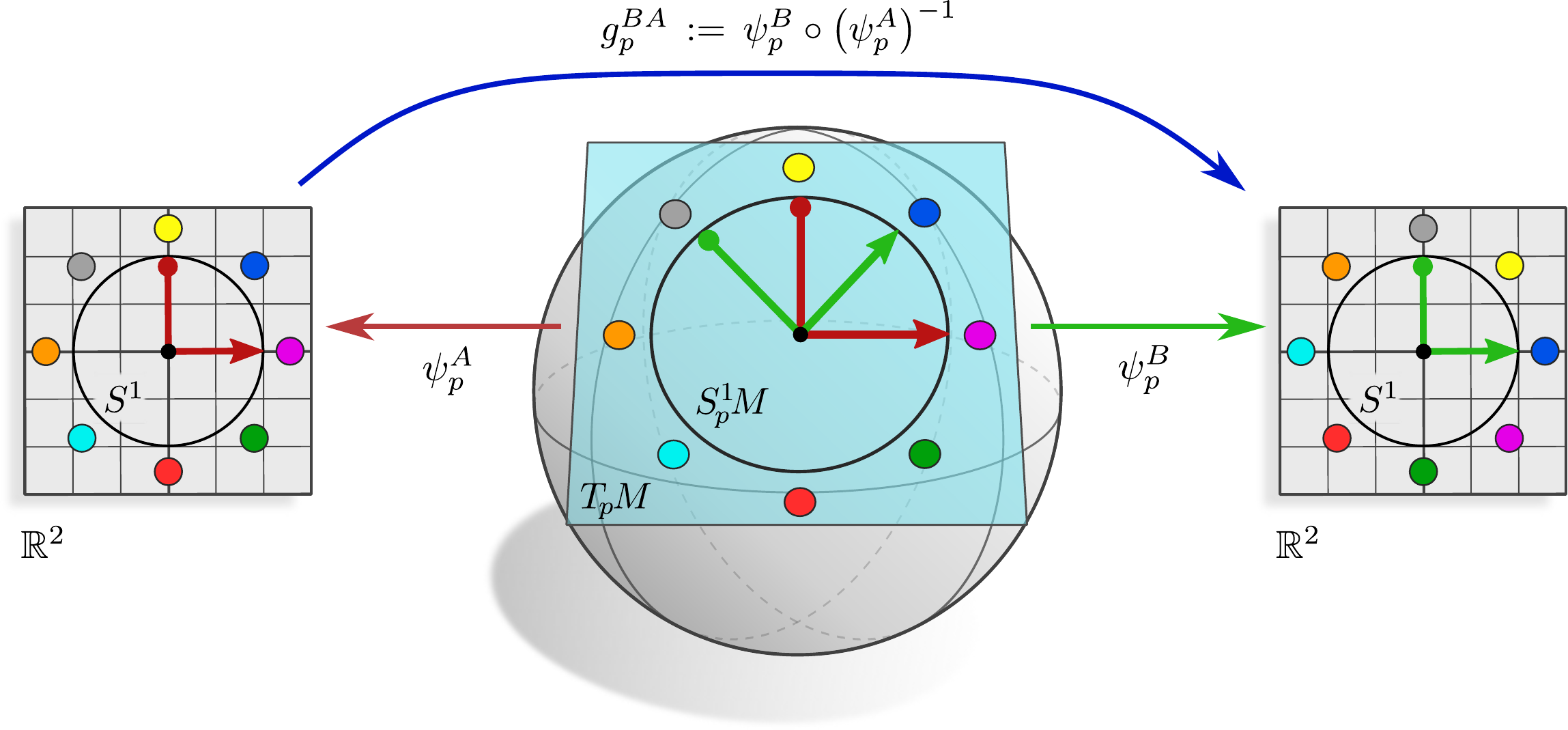}
    \caption{\small
        Visualization of the \emph{directional functions} by \citet{poulenard2018multi}.
        Directional functions assign a real-valued response (colored dots) to each direction (unit vector) in $\SpM \subset \TpM$ (black circle).
        When expressing these functions relative to right-handed, orthonormal reference frames or gauges $\psiTMp^X$, the coordinate representations assign real-valued responses to unit vectors in $S^1 \subset \R^2$.
        The transformation law between these coordinate representations is given by a rotation of the feature values on~$S^1$.
        Mathematically, this transformation law is identified as the action of the \emph{regular representation} of $\SO2$; see Eq.~\eqref{eq:directional_functions_trafo_law_regular}.
        Directional functions are therefore regular feature fields, and the surface CNN of \citet{poulenard2018multi} is based on $\GM$-convolutions between such fields.
        A diagrammatic version of this figure is given in Eq.~\eqref{cd:directional_function_trafo_law}.
    }
    \label{fig:directional_function}
\end{figure}

\paragraph{Multi Directional Geodesic CNNs:}
\citet{poulenard2018multi} proposed \emph{Multi Directional Geodesic CNNs} (MDGCNNs) which operate on so called \emph{directional functions}.
As we argue in the following, directional functions are equivalent to regular feature fields and MDGCNNs are specific $\GM$-convolutions between such features.
The authors define directional functions as real-valued function that depend on points $p\in M$ and unit directions $v\in\TpM,\ \lVert v\rVert=1$.
Denoting the circle of unit directions in $\TpM$ by
\begin{align}
    \SpM\ :=\ \big\{ v\in\TpM \,\big|\, \lVert v\rVert = 1 \big\} \ \ \cong\ \ S^1 \,,
\end{align}
a directional feature at $p$ is defined as a map
\begin{align}
    \digamma: \SpM \to \R
\end{align}
from unit directions in the tangent plane to real-valued responses.%
\footnote{
    The full directional function can then be defined as a map from ${S^1\!M}$, the bundle with fibers $\SpM$, to real values.
}
A choice of right-handed, orthonormal reference frame fixes a reference direction relative to which the directional function can be expressed.
Let~$\psiTMp^A$ be the gauge corresponding to a chosen frame, which maps the unit directions in $\SpM \subset \TpM$ to ``coordinate unit directions'' in $S^1 \subset \R^2$.
The coordinate expression of the directional function is then given by
\begin{align}\label{eq:directional_functions_trafo_law_regular}
    \digamma_p^A\ :=\ \digamma \circ \big(\psiTMp^A \big|_{\SpM} \big)^{-1}\ :\ S^1 \to \R \,,
\end{align}
that is, it assigns real-valued responses to the unit coefficient vectors on $\R^2$.
From the commutativity of the diagram
\begin{equation}\label{cd:directional_function_trafo_law}
\begin{tikzcd}[column sep=60, row sep=32, font=\normalsize]
    \R^2 \supset S^1
        \arrow[rr, rounded corners, to path={ 
                -- ([yshift=5.ex]\tikztostart.north) 
                --node[above, pos=.5]{\small$g_p^{BA} \mkern2mu\cdot$} ([yshift=5.ex]\tikztotarget.north) 
                -- (\tikztotarget.north)
                }]
        \arrow[dr, "\digamma_p^A"']
    & \SpM
        \arrow[d, pos=.4, "\digamma"]
        \arrow[l, pos=.46, "\psiTMp^A \big|_{\SpM}"']
        \arrow[r, pos=.46, "\psiTMp^B \big|_{\SpM}"]
    &  S^1 \subset \R^2
        \arrow[dl, "\digamma_p^B"]
    \\
    & \R
\end{tikzcd}
\end{equation}
one can read off that the coordinate expressions of directional functions obey the following transformation law:
\begin{align}
    \digamma_p^B\ =\ \digamma_p^A \circ \big(g_p^{BA}\big)^{-1}\ =:\ \rho_\textup{reg}\big( g_p^{BA}\big)\, \digamma_p^A
\end{align}
The second equality identified the transformation law between the coordinate expressions as the action of the regular representation, which justifies our statement that directional functions are just regular feature fields.%
\footnote{
    Strictly speaking, the regular representation of $\SO2$ acts on functions $\SO2 \to \R$.
    However, we can canonically identify such functions with functions on $S^1$ by identifying $(1,0)\in S^1$ with $\{e\}\in\SO2$.
}
Fig.~\ref{fig:directional_function} shows a directional function and its coordinate representations relative to different frames.

The multi directional geodesic convolutions by \citet{poulenard2018multi} map in a coordinate independent manner between directional functions by contracting them with equivariant kernels in a geodesic parametrization around each vertex.
This observation implies that these convolutions are specific $\GM$-convolutions between regular feature fields.
A difference in the formulation of multi directional geodesic convolutions is that their transporter pullback does not transport the whole regular feature vector (directional function) back along the geodesics, but only that single response that corresponds to the tangent direction of the geodesic.
Instead of matching the transported features with a matrix-valued kernel, multi directional convolutions match the single transported response with a scalar kernel.
The equivalence of both operations is restored by imposing a corresponding sparsity pattern to our matrix-valued $\SO2$-steerable kernels, effectively zeroing out those responses that are not transported back by MDGCNNs.
While multi directional geodesic convolutions are just $\GM$-convolutions between regular feature fields, they do therefore not use the complete space of $G$-steerable kernels between regular feature fields.
This sparsity makes MDGCNNs computationally efficient, however, the memory cost remains the same and it is unclear how severely this choice limits their expressional capacity.

The infinite number of directions in $\SO2$ (or $\SpM$ or $S^1$) is in practice discretized to the $N$ equally spaced directions in the cyclic group~$\CN$, e.g. the 8~directions that are visualized in Fig.~\ref{fig:directional_function}.
Since the Levi-Civita transport along features is in general $\SO2$-valued instead of $\CN$-valued, the authors use a linear interpolation between the $N$ discrete directions.

As discussed above, MDGCNNs transport only those specific responses of the features back which correspond to the direction of the emanating geodesic relative to the local reference frame at~$p$.
This direction is undefined at the origin $v=0 \in \TpM$, which prevents self-interactions of the vertices.
The authors resolve this issue by applying an additional \onexone\ which adds the missing self-interaction back.
As derived in Section~\ref{sec:gauge_1x1}, the \onexone\ kernels are required to be \emph{intertwiners} in order to preserve the coordinate independence of the model.
This requirement is indeed satisfied by MDGCNNs%
\footnote{
    Personal correspondence with the author.
}
as the \onexone\ matrix is constructed such that it mixes whole regular feature vectors with the same weight instead of linearly combining their channels independently.
This is implemented by representing $m_\textup{in}$ regular $\CN$-features not as a ${c=N \!\cdot\! m_\textup{in}}$-dimensional feature vector but as an array of shape $(N,m_\textup{in})$, and then applying a (shared) matrix of shape $(m_\textup{out},m_\textup{in})$ over the last axis which results in an output array of shape $(N,m_\textup{out})$.

\paragraph{Parallel Frame CNNs:}
The \emph{Parallel Frame CNNs} (PFCNNs) by \citet{Yang2020parallelFrameCNN} rely on \emph{$N$-direction frame fields}, which are just $G$-structures $\GM$ for cyclic structure groups~$G=\CN$.
Recall from our discussion above that these fields encode a connection which is trivial everywhere but at a few singularities and which is optimized to approximate the original Levi-Civita connection.
As this $G$-structure is precomputed in an offline step, we take it in the following as given and focus on the actual PFCNN convolution.
It turns out that this operation is equivalent to a $\GM$-convolution between $\CN$-regular feature fields, however, again assuming specific sparsity pattern in the kernels that is implied by the particular network design.

\begin{SCfigure}[2.4]
    \hspace*{-2ex}
    \includegraphics[width=.24\columnwidth]{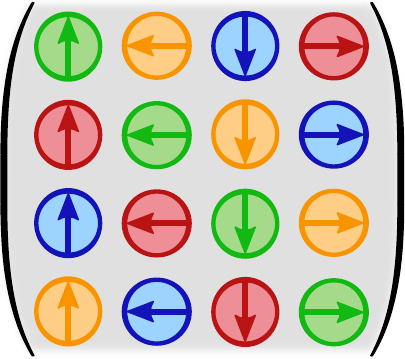}
    \captionsetup{width=1.1\columnwidth}
    \hspace{2ex}
    \caption{\small
        Degrees of freedom of a $\CN$-steerable kernel ${K: \R^2 \to \R^{N\times N}}$ which maps between feature fields that transform according to the regular representation ${\rho_{\textup{reg}}: \CN \to \GL{N}}$ for $N=4$~\cite{Weiler2019_E2CNN}.
        The kernel constraint, Eq.~\eqref{eq:kernel_constraint}, enforces the color-coded weight sharing pattern.
        The \mbox{PFCNNs} of \citet{Yang2020parallelFrameCNN} convolve features on each sheet of their $N$-direction field ($\CN$-structure) with rotated versions of a single scalar-valued kernel but do not include interaction between the different sheets.
        This shared kernel corresponds to the diagonal entries (green) of the complete kernel space.
        Off-diagonal entries, which are implicitly forced to zero, would correspond to interactions between the sheets.
        }
    \label{fig:regular_C4_kernel}
\end{SCfigure}

The feature spaces of PFCNNs are the spaces $C^\infty(\GM)$ of real-valued functions on $\GM$.
Since $\GM\!\xrightarrow{\piGM}\!M$ is for $G=\CN$ a $|G| = N$-fold cover of $M$, such feature fields can analogously be seen as assigning a tuple of $N$ real numbers to each point~$p\in M$.
As the $N$ sheets of the covering space are furthermore identified with $N$ directions (given by the first frame axes), these features are equivalent to the (discretized) directional functions of \citet{poulenard2018multi}.
Theorem~\ref{thm:regular_field_scalar_GM} in Appendix~\ref{apx:regular_field_scalar_GM} proves furthermore that there is an isomorphism
\begin{align}
    C^\infty(\GM)\ \cong\ \Gamma(\A_{\rho_\textup{reg}})
\end{align}
between the features of PFCNNs and our \emph{regular feature fields}.
PFCNNs are therefore performing coordinate independent convolutions between (an equivalent to) regular feature fields, and are thus identified as (specific) regular $\GM$-convolutions.

The formulation of parallel frame convolutions seems at first glance to be quite different from ours:
instead of convolving the full $N$-dimensional regular feature fields with a matrix-valued $\CN$-steerable kernel ${K: \R^2 \to \R^{N\times N}}$,
PFCNNs convolve their scalar functions on each of the $N$ sheets independently with a shared scalar-valued kernel which is aligned with the frame of the respective sheet.
This operation is in our framework interpreted as a convolution with a matrix-valued $\CN$-steerable kernel whose only non-zero values are on its diagonal and are rotated relative to each other, which is visualized by the green entries in Fig.~\ref{fig:regular_C4_kernel}.
The missing coupling between features on different sheets implies that the off-diagonal entries (yellow, blue and red) of the complete steerable kernel space are implicitly set to zero.
As already stated for MDGCNNs, the sparsity pattern of this regular $\GM$-convolution makes it computationally more efficient than a dense $\GM$-convolution but is likely to affect its performance and does not save memory cost.

\begin{figure}
    \centering
    \includegraphics[width=.7\columnwidth]{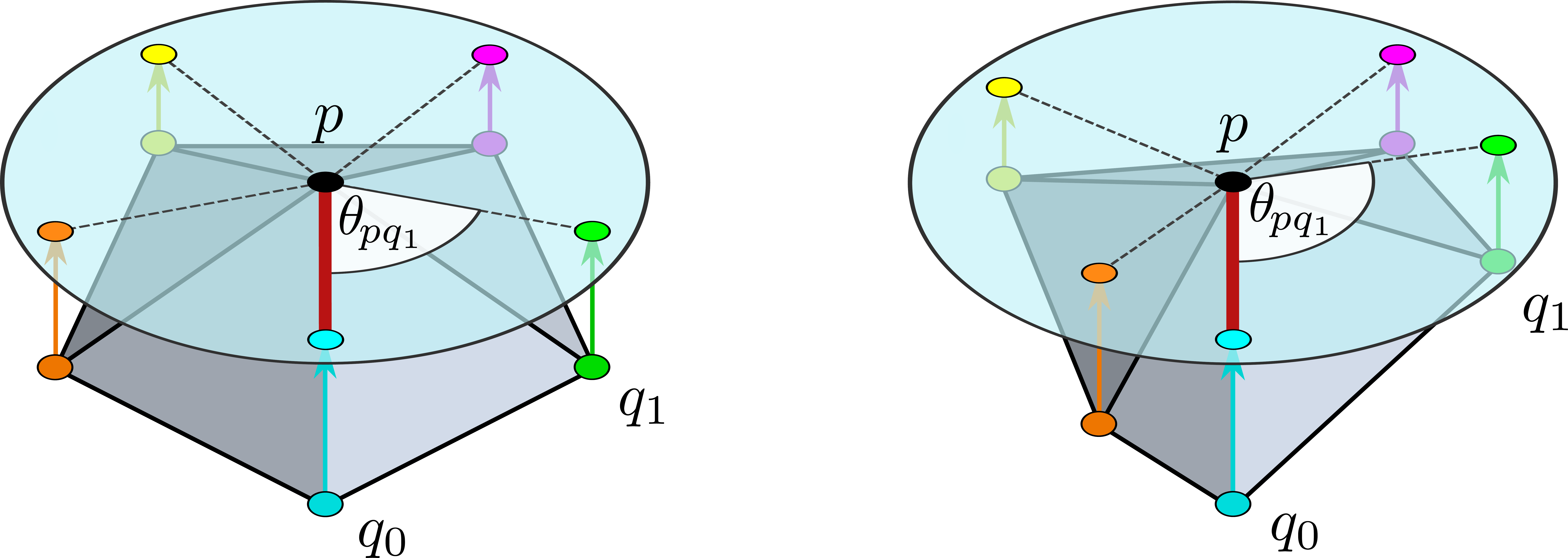}
    \vspace*{1ex}
    \caption{\small
        Two mesh regions which are topologically equivalent but geometrically distinct.
        One approach to define convolutions on meshes is to consider their underlying graph $(\mathcal{V},\mathcal{E})$, which captures the mesh topology, and run a graph neural network on it.
        Lacking information about the mesh geometry, (conventional) graph neural networks can not distinguish between the two visualized neighborhoods.
        Geometrically, they apply \emph{isotropic} kernels.
        The \emph{Gauge Equivariant Mesh CNNs} by \citet{deHaan2020meshCNNs} address this issue by projecting the neighboring vertices $q_i$ on the tangent planes and assigning them angles $\theta_{pq_i}$ relative to some reference edge, i.e. gauge (red).
        Requiring the coordinate independence of the convolutions leads to $G$-steerable kernels.
        While the model can discriminate based on the \emph{direction} of the neighboring nodes, it ignores their \emph{distance}.
        It furthermore deviates from our geodesic parametrization in that its kernel support is based on the local edge connectivity instead of geodesic distances.
        }
    \label{fig:mesh_CNNs_neighborhood}
\end{figure}

\paragraph{Gauge equivariant Mesh CNNs:}
The \emph{Gauge Equivariant Mesh CNN} (GEMCNN) by \citet{deHaan2020meshCNNs} is motivated by the shortcomings of conventional graph neural networks for the processing of feature fields on meshes.
Specifically, vanilla graph neural networks could be used to process vertex-sampled feature fields on meshes by convolving over the graph $(\mathcal{V},\mathcal{E})$ that is induced by the mesh.
The issue with this approach is that the graph encodes only the mesh \emph{topology}, but is not able to capture its \emph{geometry}.
Conventional graph convolutions do accordingly not distinguish between the ordering of edges, which corresponds on meshes to the use of \emph{isotropic kernels} that map between scalar fields.
Fig.~\ref{fig:mesh_CNNs_neighborhood} shows two regions of a mesh with distinct geometry but equivalent topology -- for conventional graph convolutions both neighborhoods look the same.
GEMCNNs address this issue by choosing a reference edge at each vertex $p\in\mathcal{V}$, relative to which the direction of all other edges $\{p,q_i\} \in\mathcal{E}$ to the one-ring of neighbors $q_i\in \mathcal{N}_p \subset\mathcal{V}$ is measured in terms of angles $\theta_{pq_i} \in [0,2\pi)$.
A choice of reference edge corresponds to a choice of orthonormal, right-handed frame.
Different choices are related by gauge transformations in the structure group~$G=\SO2$.

As in our theory, the feature spaces of GEMCNNs are defined as sections of associated vector bundles, i.e. as spaces of $c$-dimensional feature fields whose coefficients transform under gauge transformations according to some group representation $\rho: \SO2 \to \GL{c}$.
Each edge is assigned an $\SO2$-valued Levi-Civita transporter.
The convolution operation is demanded to be independent from the choice of reference edge, which leads to the requirement on the kernels to be $G$-steerable (gauge equivariant).
In contrast to our formulation, the kernels are not directly applied in geodesic normal coordinates but pass messages only from the one-ring neighborhoods $\mathcal{N}_p := \{q\in\mathcal{V} \,|\, \{p,q\}\in\mathcal{E} \}$ to that node~$p$ around which the kernel is centered.%
\footnote{
    For a (sufficiently) regular grid and compactly supported kernel in geodesic coordinates both approaches become equivalent.
}
The kernels are furthermore \emph{radially insensitive} --  in how far this affects the model performance remains an open question.

The authors decided for (real) \emph{irreps} as field types for the convolution, however, they perform a change of basis to \emph{regular representations} to apply ReLU nonlinearities, which is why we list them in row (38) of Table~\ref{tab:network_instantiations} instead of row (37).%
\footnote{
    The equivalence of $\rho$-fields to their \emph{irrep decomposition} was discussed in Section~\ref{sec:mobius_representations} and in Section~2.4 of~\cite{Weiler2019_E2CNN}.
}
Specifically, the authors use the change of basis $Q\in\R^{N\times N}$ that decomposes the regular representation $\rho_{\textup{reg}}: \CN \to \GL{N}$ of $\CN$ into its irrep components to transform a stack of irrep fields into one regular feature field.
For $\CN$, this matrix is just the discrete Fourier transform.
After applying the ReLU nonlinearity to each of the $N$ channels of the regular feature field individually -- which is a $\CN$-equivariant operation since regular representations are permutation representations -- the features are transformed back to a stack of irrep fields for the following convolution operation.
This design has the advantage that the features can be transported exactly with $\SO2$-valued transporters, without having to fall back to an interpolation scheme, as done by~\citet{poulenard2018multi}.
Note, however, that the full network is due to the use of the regular nonlinearities only $\CN$-equivariant.

That the authors use the \emph{real} irreps of $\SO2$ means that their kernel spaces are approximately twice as large as those of the Harmonic Surface Networks by \citet{Wiersma2020}; cf. the discussions in~\cite{Weiler2019_E2CNN,lang2020WignerEckart}.

\paragraph{Geodesic CNNs:}
The earliest work on geodesic convolutions that we are aware of is that of \citet{masci2015geodesic}.
The authors identified the rotational ambiguity of geodesic polar coordinates on an oriented Riemannian manifold and address it via a rotation \emph{invariant} architecture.
Their \emph{Geodesic convolutions} represent a \emph{scalar} field relative to arbitrarily oriented geodesic polar coordinates.
As the field type is trivial, the transporter pullback to geodesic coordinates does not require (non-trivial) transporters.
The feature field in geodesic coordinates is then matched with a scalar kernel, which is applied in $N$ equally spaced rotations by angles $\frac{2\pi}{N}k$ relative to the reference frame, where $k=0,\dots,N-1$.
Since a gauge transformation by $\frac{2\pi}{N}l$ for some $l\in\{0,\dots,N-1\}$ rotates all kernels accordingly, it result in a cyclic permutation of the responses by $l$ steps.
This operation corresponds therefore in our framework to a $\CN$-steerable convolution from scalar fields to $\CN$-regular feature fields.
Instead of processing these fields further via regular group convolutions
-- as done in MDGCNNs~\cite{poulenard2018multi}, PFCNNs~\cite{Yang2020parallelFrameCNN} and GEMCNNs~\cite{deHaan2020meshCNNs} --
the authors apply a $\max$-pooling operation over the $N$ responses.
Since $\CN$-valued gauge transformation result in cyclic shifts of the intermediate regular feature fields, the pooling operation is gauge-invariant, i.e. produces scalar fields.
While this networks design is simple to implement, it prevents features form encoding directional information.
Further variations of this networks design can be found in \cite{masci2015shapenet,monti2017geometric}.

\paragraph{ZerNet:}
Next, we turn to \emph{ZerNet} by \citet{sun2018zernet}.
To avoid confusion, we point out that the authors proposed two models, which we list in rows (38) \emph{and} (39) of Table~\ref{tab:network_instantiations}, respectively.
We describe both models, starting with their common design choices.

The key concept underlying ZerNets is their parameterization of convolution kernels in terms of \emph{Zernike polynomials}, which form an orthogonal basis of functions on the closed unit disk $B_{\R^2}(0,1)$ around the origin of $\R^2$.
In polar coordinates, Zernike polynomials are given by
\begin{alignat}{3}
    \textup{even:}\quad&&
        Z_n^m    &:\ [0,1]\times[0,2\pi) \to [-1,1], \quad (r,\varphi) \mapsto R_n^m(r)\, \cos(m\varphi)
        \qquad&& n\in\N,\ \ 0\leq m\leq n \\
    \textup{odd:}\quad&&
        Z_n^{-m} &:\ [0,1]\times[0,2\pi) \to [-1,1], \quad (r,\varphi) \mapsto R_n^m(r)\, \sin(m\varphi)
        \qquad&& n\in\N,\ \ 1\leq m\leq n \,,
\end{alignat}
where $R_n^m$ are the Zernike radial polynomials.
That (suitably normalized) Zernike polynomials are orthonormal means that they satisfy the orthonormality relations
\begin{align}
    \big\langle Z_n^m,\, Z_k^l \big\rangle_{B_{\R^2}(0,1)}
    \ =\ \int_0^1 \int_0^{2\pi} Z_n^m(r,\varphi)\, Z_k^l(r,\varphi)\ r\, dr\, d\varphi
    \ =\ \delta_{nk}\, \delta_{ml} \,.
\end{align}
A function on the unit disk, for instance a scalar kernel $K: B_{\R^2}(0,1) \to \R$, can be expanded in the Zernike polynomial basis:
\begin{align}
    K(r,\varphi)\ =\ \sum_{n\in\N}\, \sum_{m=-n}^n \widehat{K}_n^m\, Z_n^m(r,\varphi)
\end{align}
To retrieve the expansion coefficients of a given function on the unit disk, one projects it on the Zernike basis:
\begin{align}
    \widehat{K}_n^m
    \ =\ \big\langle K,\, Z_n^m \big\rangle_{B_{\R^2}(0,1)}
    \ =\ \int_0^1 \int_0^{2\pi} K(r,\varphi)\, Z_n^m(r,\varphi)\ r\, dr\, d\varphi
\end{align}
The inner product between two functions $K$ and $\Expspf^A$ on the unit disk can with these relations be expressed in terms of their expansion coefficients:
\begin{align}\label{eq:zernike_kernel_matching}
    \big\langle K,\, \Expspf^A \big\rangle_{B_{\R^2}(0,1)}
    \ =&\ \int_0^1 \int_0^{2\pi} K(r,\varphi)\, \Expspf^A(r,\varphi)\ r\, dr\, d\varphi \notag \\
    \ =&\ \int_0^1 \int_0^{2\pi}
        \sum_{n\in\N}\, \sum_{m=-n}^n \widehat{K}_n^m\, Z_n^m(r,\varphi)\,
        \sum_{k\in\N}\, \sum_{l=-k}^k \widehat{\big[\Expspf^A\big]}_k^l\, Z_k^l(r,\varphi)\ 
        r\, dr\, d\varphi \notag \\
    \ =&\ 
        \sum_{n\in\N}\, \sum_{m=-n}^n 
        \sum_{k\in\N}\, \sum_{l=-k}^k 
        \underbrace{\int_0^1 \int_0^{2\pi}\!
        Z_n^m(r,\varphi)\, Z_k^l(r,\varphi)\ 
        r\, dr\, d\varphi}_{\delta_{nk}\, \delta_{ml}}\ 
        \widehat{K}_n^m\, \widehat{\big[\Expspf^A\big]}_k^l \notag \\
    \ =&\ 
        \sum_{n\in\N}\, \sum_{m=-n}^n 
        \widehat{K}_n^m\, \widehat{\big[\Expspf^A\big]}_n^m
\end{align}
As suggested by the choices $K$ and $\Expspf^A$ for these functions, the authors use this property to match kernels with the pullback of the feature fields to geodesic polar coordinates.
The kernel coefficients $\widehat{K}_n^m$, which are set to zero beyond a user specified threshold, are optimized as learnable parameters of the network.
The expansion coefficients $\widehat{\big[\Expspf^A\big]}_n^m$ of the feature field's transporter pullback are computed by solving a linear system of equations.

An advantage of the kernel parameterization in terms of Zernike polynomials is that they are by definition \emph{$\SO2$-steerable kernels}.
Specifically, the pairs $\big(Z_n^m, Z_n^{-m}\big)^\top$ of kernels for a given $n\in\N$ and $1\leq m\leq n$ form a pair of kernels that are rotated by multiplying them with the $m$-th order real irrep of~$\SO2$,
\begin{align}
    \begin{pmatrix}
        Z_n^m \\ Z_n^{-m}
    \end{pmatrix}
    (r,\varphi + \Delta\varphi)
    \ =\ 
    \begin{pmatrix}
        \cos(m\Delta\varphi) &          -  \sin(m\Delta\varphi) \\
        \sin(m\Delta\varphi) & \phantom{-} \cos(m\Delta\varphi)
    \end{pmatrix}
    \begin{pmatrix}
        Z_n^m \\ Z_n^{-m}
    \end{pmatrix}
    (r,\varphi) \,,
\end{align}
while the kernels $Z_n^0$, i.e. for $m=0$, transform trivially (they are isotropic).
Note that the \emph{expansion coefficients} $\widehat{K}_n^m$ of a kernel $K$ transform \emph{inversely} to the basis.
The authors use this transformation law to rotate kernels analytically in terms of their expansion coefficients.
The rotation steerability of the Zernike polynomials' is independent from their radial parts but relies on the fact that their angular parts are \emph{circular harmonics}, which are the harmonic basis functions in the Peter-Weyl decomposition of $L^2(\SO2)$~\cite{lang2020WignerEckart}.
Due to their steerability properties, circular harmonic bases have been extensively used to parameterize real~\cite{Weiler2018SFCNN,graham2020dense} and complex~\cite{Worrall2017-HNET,Wiersma2020} convolution kernels since at least the '80s~\cite{Hsu1982optical,Rosen1988circularHarmonic,freeman1991design,hel1998canonical}.
In fact, circular harmonics are underlying \emph{any} $\SO2$-steerable kernel~\cite{Weiler2019_E2CNN,lang2020WignerEckart}.

The first and main model design described by \citet{sun2018zernet} is similar to that by~\citet{masci2015geodesic}.
A scalar field is pulled back to geodesic normal coordinates, where it is matched with a scalar kernel that is applied in $N$ discrete rotations, resulting in an intermediate $\CN$-regular feature field.
A subsequent $\max$-pooling operation over the $N$ responses yields then a $\CN$-invariant output, i.e. an output scalar field.
The difference to the implementation by \citet{masci2015geodesic} is that this operation is performed in the Zernike polynomial basis as specified in Eq.~\eqref{eq:zernike_kernel_matching}.
This choice corresponds ultimately to an alternative interpolation scheme.
The second model design, described Section~4.4 of~\cite{sun2018zernet}, is a reimplementation of the MDGCNNs from~\citet{poulenard2018multi} in the Zernike polynomial basis.
The authors observe that this design leads to a significantly improved performance since the regular feature fields are able to encode directional information.

\paragraph{TextureNet:}
The last rotation steerable model that we discuss is the \emph{TextureNet} by \citet{huang2019texturenet}.
\begin{wrapfigure}[13]{r}{0.22\textwidth}
    \centering
    \includegraphics[width=.16\textwidth]{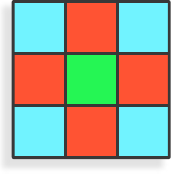}
    \captionsetup{width=.21\textwidth}
    \caption{\small
        \\
        A $\D4$-invariant kernel of $3\times3$ pixels is parameterized by three degrees of freedom.
        }
    \label{fig:3x3_D4_invariant_kernel}
\end{wrapfigure}%
In contrast to the previous models, TextureNets assume a $\D4$-structure, which could easily be generalized to a $\DN$-structure.
This $\D4$-structure is precomputed via QuadriFlow, a 3rd party software package which can be used to compute 4-RoSy fields that are optimized to be smooth and have few singularities~\cite{Huang2018QuadriFlow}.
As the name suggests, TextureNets process input feature fields that are represented as textures, and are of potentially higher resolution than the mesh.
The convolution kernels are applied at a dense set of sampling locations, which are uniformly distributed over the mesh's faces.
At each sampling point the scalar feature field is pulled back into geodesic normal coordinates and represented relative to an arbitrary frame of the $\D4$-structure.
It is then matched with a $\D4$-invariant $3\times3$ kernel.
As visualized in Fig.~\ref{fig:3x3_D4_invariant_kernel}, the 9 pixels of such kernels are described by 3 degrees of freedom.
The convolution is implemented in terms of three \onexones\ whose responses are subsequently binned and aggregated in each of the tangent spaces.
The additional reflection steerability of the kernels implies that TextureNets are well defined on non-orientable surfaces.
However, as the features of TextureNet are scalar fields they can neither encode directions nor orientations.
To overcome this issue, it is necessary to use non-trivial $\DN$ or $\O2$-steerable kernels.

%% file: chapters/123_mesh_e-steer.tex

\subsection{\{\textit{e}\}-steerable surface convolutions}
\label{sec:e_surface_conv}

This section reviews the networks from 
\cite{monti2017geometric,jin2018learning,schonsheck2018parallel,tatarchenko2018tangent,jin2019NPTCnet,li2019crossAtlas},
which have in common that they rely on $\{e\}$-structures on the surfaces.
From the viewpoint of $\GM$-convolutions, these architectures differ mainly in the specific choice of \emph{heuristic that determines the $\{e\}$-structure}.

Assuming a trivial structure group $G=\{e\}$, the models apply $\{e\}$-steerable (i.e. unconstrained) kernels, which are aligned along the frames of the chosen $\{e\}$-structure.
The field types (group representations) are necessarily trivial.
The same holds for all parallel transporters, which are necessarily $\{e\}$-structure compatible.
Transporter pullbacks $\Expspf$ of feature fields $f$ to the tangent spaces reduce therefore to pullbacks $\exp_p^*f$ by the usual exponential map, that is, they don't apply (non-trivial) transporters.
Recall that continuous $\{e\}$-structures exist only on parallelizable manifolds, implying that the networks' inference is inevitably discontinuous on non-parallelizable surfaces.
The heuristics that determine the frame fields are furthermore not always well defined, or are instable under deformations of the surfaces' geometry, as further discussed below.

The models of \citet{monti2017geometric}, \citet{jin2018learning} and \citet{schonsheck2018parallel} operate on triangular meshes and process feature fields that are sampled at the vertices.
\citet{tatarchenko2018tangent} and \citet{jin2019NPTCnet} propose networks that operate on surface point clouds while the architecture of \citet{li2019crossAtlas} defines convolutions on texture atlases of meshes.

\paragraph{Geodesic MoNets:}
The first model family that we discuss are the \emph{MoNets} by \citet{monti2017geometric}.
The authors discuss a variety of models on graphs and manifolds, most of which are not explained as $\GM$-convolutions.
These models have in common that they apply kernels relative to some choice of ``pseudo-coordinates'' on the manifold or graph
-- we are here only interested in those MoNets that rely on geodesic normal coordinates and are therefore identified as $\GM$-convolutions.

As stated above, the main difference between $\{e\}$-steerable surface convolutions is their particular choice of $\{e\}$-structure.
Inspired by previous work of \citet{boscaini2016learning}%
\footnote{
    The \emph{Anisotropic CNNs} by \citet{boscaini2016learning} assume the same principal curvature direction based $\{e\}$-structure.
    However, their kernels are not defined in geodesic normal coordinates but are based on anisotropic heat kernels on the manifold.
    \citet{monti2017geometric} claim that such heat kernels correspond to anisotropic Gaussian kernels in geodesic coordinates -- if this statement is true, Anisotropic CNNs can be viewed as $\GM$-convolutions.
},
the authors choose to align the reference frames of the $\{e\}$-structure with the \emph{principal curvature direction} of the manifold.
Note that this heuristic is not well defined when the principal curvatures $\kappa_{\textup{max}} = \kappa_{\textup{min}}$ agree, i.e. when the principal curvature direction is degenerate.
An extreme example is the 2-sphere $S^2$, where the principal curvature direction is nowhere well defined.
Even when the principal curvatures are unequal, they determine only an undirected line, disambiguating reference frames up to a $\C2$-structure (with the two constituent frames pointing along the two directions along the line).
To make the network independent form the choice of frame, they should therefore actually apply $\C2$-steerable kernels.
Moreover, the principal curvature directions are instable under deformations of the surface.
As an example, imagine the principal curvature direction at the north pole (on the positive $z$-axis) of the 2-sphere $S^2$:
an infinitesimal squeezing of the sphere along the $x$-axis results in a principal curvature direction along the $x$-axis while an infinitesimal stretching along the $x$-axis results in a principal curvature direction along the $y$-axis.
We furthermore want to mention that principal curvatures depend on the embedding of a manifold, that is, the approach is non-intrinsic.

\paragraph{3DMCNN:}
\citet{jin2018learning} proposed a \emph{3D Mesh CNN} (3DMCNN) that convolves over the surfaces of scanned faces to recognize expressions like happiness, anger or surprise.
As the face-masks are topologically planes (with holes at the eyes) they are parallelizable, which allows for smooth $\GM$-convolutions for ${G=\{e\}}$.

The convolution kernel is discretized into one central sampling point and eight other points at a fixed radial distance $R$ and angles $\varphi_k = k\frac{2\pi}{8},\ k=0,\dots,7$ in polar coordinates.
The kernels -- and thus the frames that constitute the $\{e\}$-structure -- are rotated such that they are aligned with the $z$-axis of the embedding space~$\R^3$.
This approach seems reasonable since the face masks are parallelizable and, more importantly, aligned upright.
To match a such oriented kernel with a feature field, geodesics of length $R$ are shot in the eight directions.
Barycentric coordinates are used to interpolate the signal from the surrounding vertices to the end point of the geodesic.

\paragraph{Parallel Transport Convolutions:}
As a last mesh-based $\{e\}$-steerable convolution we discuss the \emph{Parallel Transport Convolutions} (PTCs) by \citet{schonsheck2018parallel}.
The key idea of PTCs is to define the convolution kernel at some ``origin'' $p_0 \in M$ and share it with any other location $p\in M$ by Levi-Civita transporting it along the shortest geodesics between $p_0$ and~$p$.
To formulate this weight sharing procedure in more detail, consider the closed disks $B_{\TpM}(0,R) \subset\TpM$ of radius $R$ around the origins of the tangent spaces, where $R\in\R_+$ is the injectivity radius of the manifold.
Let furthermore $M_{p,R} := \exp_p( B_{\TpM}(0,R)) \subset M$ be the images of these disks under the exponential map, which include all points whose geodesic distance from~$p$ is smaller than or equal to~$R$.
\citet{schonsheck2018parallel} define their (unconstrained) scalar convolution kernels than as real-valued functions
\begin{align}
    \widehat{K}_{p_0} \!: M_{p_0,R} \to \R
\end{align}
on the neighborhood around the origin~$p_0$, i.e. directly on the manifold.
To share the kernel with other locations $p\in M$, the authors compute the shortest geodesics between $p_0$ and the target locations $p$ via Fast Marching.
They parallel transport the kernel then along these geodesics, which is done by pulling them back to the tangent spaces.
In equations, the kernel at $p$ is defined as
\begin{align}\label{eq:PTCs_kernel_transport}
    \widehat{K}_p \!: M_{p,R} \to \R ,\quad
    q \mapsto \widehat{K}_p(q) \,:=\,
    \widehat{K}_{p_0} \circ \exp_{p_0} \circ\,
    \mathcal{P}_{\mkern-2mu\overset{}{\protect\scalebox{.6}{$\!T\!M$}\mkern-2mu,\mkern1mu p_0\to p}}^{-1}
    \circ \log_p (q) \,,
\end{align}
which is visualized by the following commutative diagram:
\begin{equation}\label{cd:PTC_kernel_1}
\begin{tikzcd}[column sep=50, row sep=10, font=\normalsize]
    M_{p_0,R}
        \arrow[drr, rounded corners, to path={ 
                |-node[below, pos=.8]{\small$\widehat{K}_{p_0}$} ([xshift=-5.ex]\tikztotarget.west) 
                -- (\tikztotarget.west)
                }]
    & B_{T_{\mkern-1.mu p_0}\!M}(0,R)
        \arrow[l, "\exp_{p_0}"']
        \arrow[rr, "\mathcal{P}_{\mkern-2mu\overset{}{\protect\scalebox{.6}{$\!T\!M$}\mkern-2mu,\mkern1mu p_0\to p}}"]
    &[-30pt]
    &[-30pt]
      B_{\TpM}(0,R)
        \arrow[r, "\exp_p"]
    & M_{p,R}
        \arrow[dll, rounded corners, to path={ 
                |-node[below, pos=.8]{\small$\widehat{K}_p$} ([xshift=5.ex]\tikztotarget.east) 
                -- (\tikztotarget.east)
                }]
    \\
    & & \R
\end{tikzcd}
\end{equation}
The existence of the logarithmic map is guaranteed since the domain is restricted to points~$q$ within the injectivity radius.
To compute the convolution response at $p$, the transported kernel is matched with the (scalar) feature field on~$M_{p,R}$.

In order to describe PTCs as $\GM$-convolutions, we need to identify the corresponding $\{e\}$-structure and $\{e\}$-steerable kernel on~$\R^2$.
A compatible $\{e\}$-structure is fixed by choosing an arbitrary frame $\big[e^A_i(p_0)\big]_{i=1}^d$ at the \mbox{origin}~$p_0$.
The frames at any other location $p$ are then determined by Levi-Civita transporting this initial frame along the shortest geodesics, that is, they are defined as%
\footnote{
    Since this relation \emph{defines} the $\{e\}$-structure, we need to use the Levi-Civita transporters on the full frame bundle~$\FM$.
}
\begin{align}
    \big[e^A_i(p) \big]_{i=1}^d\ :=\ 
    \mathcal{P}_{\mkern-2mu\overset{}{\protect\scalebox{.6}{$\!F\!M$}\mkern-2mu,\mkern1mu p_0\to p}}
    \big[e^A_i(p_0)\big]_{i=1}^d \,.
\end{align}
Note that this definition implies in particular the following equivalent relation for the corresponding gauges, which is easily seen by applying it to the frame field:
\begin{align}\label{eq:PTC_e-structure_gauges}
    \psiGMp^A
    \ =\ 
    \psi_{\protect\scalebox{.6}{$G\!M,\mkern2mu$}\protect\scalebox{.7}{$p_0$}}^A
    \circ
    \mathcal{P}_{\mkern-2mu\overset{}{\protect\scalebox{.6}{$\!G\!M$}\mkern-2mu,\mkern1mu p_0\to p}}^{-1}
\end{align}
Given the reference frame at~$p_0$, we can express $\widehat{K}_{p_0}$ in geodesic normal coordinates, which gives rise to our usual notion of template kernel on $\R^2$:
\begin{align}\label{eq:PTCs_kernel_lift_R2}
    K: B_{\R^2}(0,R) \to \R,
    \quad \mathscr{v} \mapsto K(\mathscr{v}) :=
    \widehat{K}_{p_0} \circ \exp_{p_0} \circ\mkern2mu
    \big( \psi_{\protect\scalebox{.6}{$T\!M,$}\protect\scalebox{.7}{$p_0$}}^A \big)^{-1} (\mathscr{v})
\end{align}
To show that our weight sharing via the such constructed $\{e\}$-structure is indeed consistent with that by \citet{schonsheck2018parallel}, we reproduce the kernels $\widehat{K}_p$ at $p$ by mapping our template kernel $K$ down to the manifold:
\begin{align}
    K \circ \psiTMp^A \circ \log_p
    \ =&\ \widehat{K}_{p_0} \circ \exp_{p_0} \circ\,
        \big( \psi_{\protect\scalebox{.6}{$T\!M,$}\protect\scalebox{.7}{$p_0$}}^A \big)^{-1}
        \circ \psiTMp^A \circ \log_p \notag \\
    \ =&\ \widehat{K}_{p_0} \circ \exp_{p_0} \circ\,
        \mathcal{P}_{\mkern-2mu\overset{}{\protect\scalebox{.6}{$\!T\!M$}\mkern-2mu,\mkern1mu p_0\to p}}^{-1}
        \circ \log_p \notag \\
    \ =&\ \widehat{K}_p
\end{align}
The second step in this calculation used the equivalent to Eq.~\eqref{eq:PTC_e-structure_gauges} for the tangent bundle transporter and gauges.
All definitions, and their consistency, are concisely summarized by the statement that the following diagram commutes:
\begin{equation}\label{cd:PTC_kernel_2}
\begin{tikzcd}[column sep=50, row sep=24, font=\normalsize]
    M_{p_0,R}
        \arrow[ddrr, rounded corners, to path={ 
                |-node[left, pos=.36]{\small$\widehat{K}_{p_0}$} ([xshift=-5.ex]\tikztotarget.west) 
                -- (\tikztotarget.west)
                }]
    & B_{T_{\mkern-1.mu p_0}\!M}(0,R)
        \arrow[l, "\exp_{p_0}"']
        \arrow[rr, "\mathcal{P}_{\mkern-2mu\overset{}{\protect\scalebox{.6}{$\!T\!M$}\mkern-2mu,\mkern1mu p_0\to p}}"]
        \arrow[dr, "\psi_{\protect\scalebox{.6}{$T\!M,$}\protect\scalebox{.7}{$p_0$}}^A"']
    &[-45pt]
    &[-45pt]
      B_{\TpM}(0,R)
        \arrow[dl, "\psiTMp^A"]
        \arrow[r, "\exp_p"]
    & M_{p,R}
        \arrow[ddll, rounded corners, to path={ 
                |-node[right, pos=.36]{\small$\widehat{K}_p$} ([xshift=5.ex]\tikztotarget.east) 
                -- (\tikztotarget.east)
                }]
    \\
    & & B_{\R^2}(0,R)
        \arrow[d, "K"]
    \\
    & & \R
\end{tikzcd}
\end{equation}
Since we constructed our $\{e\}$-structure by choosing an initial frame at $p_0$, the reader might wonder about the implications of this choice.
A different choice of initial frame will result in a corresponding transformation of the geodesic normal coordinates at $p_0$, and therefore of the template kernel~$K$ (Eq.~\eqref{eq:PTCs_kernel_lift_R2}).
However, since the $\{e\}$-structure is constructed by transporting the initial frame, all of its reference frames will transform accordingly.
The transformation of the template kernel will then cancel out with the transformation of the $\{e\}$-structure such that all choices of initial frames are ultimately equivalent.

The $\{e\}$-structures underlying PTCs depend crucially on the choice of origin $p_0$ from which the frame field is constructed
-- different choices of origins can lead to very different $\{e\}$-structures.
As most manifolds do not come with a canonical notion of origin, the proposed heuristic seems somewhat arbitrary.
Transport based $\{e\}$-structures, and thus PTCs, are furthermore discontinuous at the cut locus.
This implies in particular that they are close to the cut locus unstable under deformations of the surfaces' geometry since such deformations may shift the cut locus.
In contrast to the heuristics of the previous models, the heuristic of PTCs depends solely on the intrinsic geometry of the surface, that is, it is not based on its embedding in ambient space.

To avoid confusion, we need to mention that \citet{schonsheck2018parallel} construct in their implementation (Section~3.2) another frame field, which should not be confused the $\{e\}$-structure that we described above.
This frame field is required for the numerical computation of the Levi-Civita connection on the mesh, according to which the kernels are then transported.
Our analysis above is purely based on their coordinate free definition of the model, most importantly the definition of weight sharing in (our) Eq.~\eqref{eq:PTCs_kernel_transport}.

Note furthermore that the implicitly assumed feature vector transporters in the transporter pullback rely necessarily on the $\{e\}$-compatible trivial connection that is implied by the $\{e\}$-structure.
The feature transport agrees along the geodesics emanating from $p_0$, based on which the $\{e\}$-structure was constructed, with Levi-Civita transporters.
Transporters along any other path differ in general from the Levi-Civita transport.

\paragraph{Tangent convolutions:}
The \emph{tangent convolutions} by \citet{tatarchenko2018tangent} operate on \emph{point clouds} ${P\subset \R^3}$ whose points are assumed to lie on a surface.
Tangent spaces at the sampling points are computed via a \emph{local principal component analysis} (LPCA).
The LPCA at~$p\in P$ is essentially computing the \mbox{eigenvectors} ${e_i \!\in \R^3}$, ${i=\!1,2,3}$, of the covariance matrix of all points within a spherical neighborhood ${\mathcal{N}_p =} {\{q\in P | \lVert q-p\rVert<R \}}$ of radius~$R$ around~$p$.
As the point cloud is sampled from a surface, one of the eigenvalues should be close to zero.
The corresponding eigenvector $e_3$ is taken as the normal vector of the embedded tangent plane $\TpM \subset \R^3$ at~$p$.
The two other eigenvectors span an orthonormal frame $[e_1,e_2]$ on the tangent plane, such that the collection of LPCA eigenvectors implies an $\{e\}$-structure on the point cloud.
Note that the eigenvector with the largest eigenvalue points in the direction of minimal principal curvature, that is, one has $\kappa_n(e_1) = \kappa_{\min}$ and $\kappa_n(e_2) = \kappa_{\max}$.
The considered $\{e\}$-structure is therefore similar to that of \citet{boscaini2015learning} and \citet{monti2017geometric}, however, the frames are rotated by $\pi/2$ since they are aligned with the minimal instead of maximal curvature direction.%
\footnote{
    Since all reference frames are rotated by the same angle, this difference is irrelevant if the kernels are learned.
}
Since the sign of the eigenvectors is arbitrary, this heuristic fixes frames actually only up to rotations by $\pi$.
To address this ambiguity, tangent convolutions would either have to disambiguate between the two directions or fall back to $\C2$-steerable kernels.

Instead of representing the feature field in geodesic normal coordinates, tangent convolutions project the features along the normal direction on the tangent plane.%
\footnote{
    This choice makes tangent convolutions (and NPTC-nets) different from $\GM$-convolutions.
    In the limit of small kernels relative to the curvature of the surface both projections of feature fields to the tangent spaces become equivalent.
}
They are then interpolated to a regular grid of ${N\times N}$ pixels.
As this grid is aligned with the reference frame, it can be viewed as a discretization of the tangent space coordinatization $\psiTMp^A(\TpM) = \R^2$.
The convolution computes features then by taking the inner product with a ${N\times N}$ pixel kernel.

\paragraph{NPTC-net:}
\citet{jin2019NPTCnet} proposed \emph{NPTC-nets} on surface point clouds~${P\subset\R^3}$.
Like tangent convolutions, NPTC-nets compute tangent planes via a local principal component analysis,
however, their $\{e\}$-structure is independent from the LPCA.
The $\{e\}$-structure that is underlying NPTC-nets is rather aligned with the gradient of the geodesic distance function from some initial point~$p_0 \in P$.
To solve for the distance function, \citet{jin2019NPTCnet} solve the Eikonal equation via a Fast Marching algorithm.
Instead of operating directly on the point cloud as done for instance in~\cite{Crane2017HeatMethodDistance}, the authors propose to use a sparse voxel grid whose voxels lie in a narrow band around the point cloud.
Having computed the distance function on the voxel grid, which should produce approximately geodesic distances, its gradient is computed and projected on the tangent planes.
The projected vector determines the first frame axes of the $\{e\}$-structure.
Note that such defined frame fields are singular at~$p_0$.

\citet{jin2019NPTCnet} observe that this $\{e\}$-structure implies a trivial connection on the surface (defined such that the frame field is closed under this transport).
The frame field (or convolution kernels) can be understood as being transported according to this trivial connection, which motivates the ``PTC'' (parallel transport convolution) in the model name.
Note, however, that NPTC-nets rely in contrast to the PTCs of \citet{schonsheck2018parallel} not on the Levi-Civita transport.
Moreover, this statement can be made for \emph{any} $\{e\}$-structure and corresponding trivial connection.

Like tangent convolutions, NPTC-nets project the features in the ambient space to the tangent plane.
Instead of using a projection along the normal direction, the authors use a nearest neighbor interpolation with distances measured in ambient space.
The convolution kernel is then oriented along the frames of the $\{e\}$-structure and matched with the interpolated feature field.
Given a convolution kernel $K: \R^2 \to \R$, the authors formulate its assignment to that tangent spaces as $K \circ \psiTMp^A: \TpM \to \R$ where $\psiTMp^A := (\langle e_1^A,\,v\rangle,\, \langle e_2^A,v\rangle )^\top$.
This procedure matches our definition of weight sharing and gauges (Eq.~\eqref{eq:embedding_gauge_map_orthonormal_frame}) exactly.

\paragraph{Cross-atlas convolutions:}
An entirely different approach was followed by \citet{li2019crossAtlas}.
Their \emph{cross-atlas convolutions} compute a texture atlas whose charts are optimized to be approximately isometric.
The convolution operation is then performed on the texture atlas, with pixel offset maps modeling the transition maps between charts.

Before running the actual convolutions, an atlas of charts is computed.
From an abstract viewpoint, the charts map patches of the surface to $\R^2$, such that the whole surface is covered.
Concretely, they map patches of a $c$-channel input feature field (texture) in a non-overlapping way to an array of dimensions $(X,Y,c)$.
Since the patches in the array should approximately represent geodesic neighborhoods on the surface, the charts should be approximately isometric, i.e. minimize distortions.
To satisfy this requirement, the surface is cut such into patches that the mutual angles between all triangle normals within a patch stay below a user specified threshold -- note that this approach is based on the surfaces' extrinsic geometry.
After optimizing the patches on the surface, the feature field is on each patch projected along a dominant projection direction.
A bin-packing algorithm packs the projected patches densely into the texture map of shape $(X,Y,c)$.
To resolve the directional ambiguity of the patches they are required to be \emph{rotation aligned}.
This is achieved by demanding that the projections of the ambient space's $z$-axis to each patch are all aligned in the texture map.

The convolution operates directly on the texture map.
It groups the pixels into three different categories which are processed in a different manner.
Pixels which are in the interior of a patch, such that the kernel does not range out of the patch, are convolved via conventional Euclidean convolutions.
Since the charts are approximately isometric, this corresponds approximately to a geodesic convolution on the patch interior regions on the surface.
Pixels that are outside of the patches are not processed, their value is fixed to zero.
The interesting case is that of pixels which are close to the boundary of the patches.
As the convolution kernel ranges for such pixels out of the current patch, it requires transition maps which query features from a neighboring patch on the surface.
The query location is computed by
1) finding the original point on the surface that corresponds to the current kernel location,
2) shooting a geodesic to find the kernel sampling location on the surface and
3) mapping this location to the corresponding pixel in the texture map.
Using these transition maps the patches are stitched together according to the surface geometry and the convolution on the texture map corresponds approximately to a geodesic convolution on the surface.
In the limit of the normal angle threshold going to zero, the approximating converges to an exact geodesic convolution.
However, the patches shrink then down to individual faces, leading to more non-trivial transition maps.

Cross-atlas convolutions correspond in this limit to $\GM$-convolutions whose $\{e\}$-structure is induced from the charts.
The $\{e\}$-structure is at the boundaries between adjacent patches discontinuous, however, the jumps should due to the rotation alignment of the patches in the texture map in most cases be minimized.
The discontinuities are expected to be large at patches of the surface which are approximately horizontal.

For completeness, we want to point to the atlas based methods by \citet{sinha2016deep} and \citet{maron2017convolutional}.
Both consider \emph{non-isometric} projections of the surface to a planar domain, which implies that the subsequent Euclidean convolutions do not correspond to geodesic convolutions on the surface.

%% file: chapters/130_conclusions.tex

\mypart{Conclusion}

We conclude with a summary of our work and an outlook on potential future developments.
\vspace*{-1ex}

\paragraph{Summary:}

In this work we investigated the design of convolutional networks on Riemannian manifolds.
In contrast to Euclidean vector spaces~$\R^d$, manifolds do in general not come with a canonical choice of reference frames relative to which features and kernels could be expressed.
Given that no specific choice of frames may be geometrically preferred, it is natural to demand geometric quantities like feature vectors to be coordinate independent.
They are therefore associated with some transformation law (field type) which determines how their coordinate expressions in different gauges relate to each other.
The layers of coordinate independent networks are required to respect the features' transformation laws in the sense that the action of a transformed layer on a transformed input feature results in a correspondingly transformed output feature.
As argued and exemplified at multiple examples, this does in general not imply a constraint on the network's connectivity.
However, local template functions like convolution kernels, biases or nonlinearities may only then be shared in a coordinate independent manner if they are $G$-steerable, i.e. equivariant w.r.t. gauge transformations.
In a nutshell, coordinate independent convolutional networks are models that apply a shared $G$-steerable kernel in geodesic normal coordinates at each point of the manifold.

While the construction of coordinate independent CNNs focuses solely on their local gauge equivariance, the models are automatically equivariant w.r.t. the action of isometries on feature fields.
The relevant subgroup $\IsomGM \leq \IsomM$ of isometries consists thereby of those symmetries of the Riemannian manifold that are simultaneously symmetries of the $G$-structure~$\GM$
-- this connection allows to reduce the design of isometry equivariant convolutions to the design of invariant $G$-structures.
Investigating isometry equivariant networks beyond convolutions we find that weights need not necessarily be shared over the whole manifold but just over the isometry orbits.
For the specific case of homogeneous spaces there is one single orbit which agrees with the manifold itself
-- weight sharing becomes in this case global and implies a convolution.
This finding recovers the results from previous work on convolutions on homogeneous spaces by
\citet{Kondor2018-GENERAL}, \citet{Cohen2018-intertwiners}\cite{Cohen2019-generaltheory} and \citet{bekkers2020bspline};
see Appendix~\ref{apx:homogeneous_conv} for a detailed comparison.

To clarify the practical application of our theory and to evaluate it empirically, we implemented orientation independent convolutions on the M\"obius strip.%
\footnote{
    The code is publicly available at \url{https://github.com/mauriceweiler/MobiusCNNs}.
}
This required us to solve the equivariance constraints on reflection steerable kernels, biases and nonlinearities for the different field types under consideration.
Since the M\"obius strip is locally flat, the network is implemented via conventional Euclidean convolutions on a chart codomain.
Parallel transporters are either trivial or incur a reflection -- they are in practice implemented via a reflect-padding operation at the line where the strip was cut open.
Orientation-independent convolutions were empirically shown to outperform a naive coordinate dependent baseline model.
They were furthermore shown to be equivariant under the action of the M\"obius strip's isometry group.

As our differential geometric theory of convolutional networks allows for arbitrary Riemannian manifolds, $G$-structures, connections and field types, it describes a wide range of related work in a unified framework.%
\footnote{
    Any of the models is -- up to discretization -- fully specified by the design choices listed in the intro of Part~\ref{part:literature_review}.
}
We substantiated this claim in an extensive literature review which covers affine group equivariant Euclidean CNNs, rotation equivariant CNNs on punctured Euclidean spaces, spherical and icosahedral CNNs and CNNs on more general surfaces;
see Table~\ref{tab:network_instantiations}.
Besides discussing the specific design choices of each network architecture in detail, our review gives an introduction to the underlying manifolds' geometries from the viewpoint of our gauge formalism.
We hope that this facilitates the future design of network architectures.

\paragraph{Outlook:}

Our work suggests a number of future research directions, ranging from applications and empirical investigations over extensions of the theory to the transfer of insights and developed techniques to related work.

While we developed the theory of coordinate independent CNNs in quite some depth, a \emph{systematic empirical study of their practical aspects and design choices} is still pending.
Our literature review in Part~\ref{part:literature_review} covers many choices of manifolds, $G$-structures and field types, however, the models are trained on different learning tasks or in different setups which prevents a direct comparison of their performances.
The work of \citet{Weiler2019_E2CNN} made a first step towards a systematic empirical evaluation of field types and equivariant nonlinearities.
However, their benchmark covers only $\Aff(G)$-invariant $G$-structures for $G\leq\O2$ on two-dimensional flat spaces~$M=\Euc_2$.
Future work should extend this study to further manifolds of different dimensionalities and to further $G$-structures.

An interesting extension of our theoretical framework would be to replace spatially extended kernels with \emph{learned partial differential operators}.
In contrast to kernels, which are applied in geodesic normal coordinates, partial differential operators act locally and allow therefore to dispense with the manifold's Riemannian structure altogether.
As a consequence, coordinate independent neural PDEs would not only be isometry equivariant but more generally \emph{diffeomorphism equivariant}.%
\footnote{
    The relevant diffeomorphism group $\DiffGM\leq\DiffM$, defined in Eq.~\eqref{eq:DiffGM_def_part1}, would consist of those diffeomorphisms that are symmetries of the $G$-structure.
}
Specifically for $G=\GL{d}$, this would result in generally covariant neural PDEs, which might be relevant for the application of convolutional networks in numerical general relativity.
A differential formulation of neural networks is expected to show profound similarities to theories in physics and other natural sciences -- this might provide new pathways to connect both fields and transfer knowledge between them.
Specifically, the results by \citet{lang2020WignerEckart} suggest that $G$-steerable partial differential operators might just be representation operators as described by the Wigner-Eckart theorem from quantum mechanics.
First steps towards steerable neural PDEs have been presented in prior work, however, a general formulation is still missing:
the formulation in \cite{jiang2019spherical} is non-equivariant, while \cite{shen2020PDOeConvs,shen2021PDOeSpherical} consider only regular representations as field types and \cite{smets2020pde,sharp2020diffusion} restrict their attention to a specific subset of equivariant differential operators.
Note that the majority of the related work in our literature review in Part~\ref{part:literature_review} would \emph{not} be explained by neural PDEs since their models are explicitly assuming spatially extended kernels.%
\footnote{
    If partial differential operators are discretized in terms of a stencil of finite size, this approximation may result in a $G$-steerable kernel.
}

Further extensions of the theory could investigate structure groups that are not subgroups of~$\GL{d}$.
An interesting application from quantum field theory would be networks that rely on \emph{spin structures} ($G={\operatorname{Spin}(d)}$) and operate on spinor bundles.
One could furthermore try to \emph{learn a $G$-structure} on a manifold instead of fixing it.
\citet{sommer2019horizontal} investigated an alternative approach to accumulate feature vectors, replacing their parallel transport along geodesics with a diffusion process.

Coordinate independent neural networks could furthermore be combined with orthogonal advances in equivariant deep learning.
Examples include equivariant \emph{attention mechanisms} \cite{hutchinson2020lietransformer,romero2020attentive,romero2020selfAttention,Romero2020CoAttentive,fuchs2020se3transformers,fuchs2021iterative},
equivariant \emph{capsule networks} \cite{lenssen2018groupCapsule,zhao2019quaternion,venkataraman2020equivCapsule}
or \emph{probabilistic} equivariant models~\cite{bloem2019probabilistic},
including in particular equivariant \emph{flows}~\cite{kohler2020equivariant,rezende2019equivariant,li2020exchangeable} or equivariant \emph{neural processes}~\cite{finzi2020probabilistic,holderrieth2020steerableCNP,kawano2021GCNN_CNP}.
Another topic of interest is the \emph{universality} of equivariant networks~\cite{yarotsky2018universal,maron2019universality,sannai2019universalPermutation,keriven2019universal,segol2020universalSet,ravanbakhsh2020universal,kumagai2020universal,dym2020universality}.

While coordinate independent CNNs operate on Riemannian manifolds as base spaces, their features (at any point $p\in M$) are vector-valued, i.e. live in a Euclidean vector space.
An independent line of research investigates neural networks with \emph{manifold-valued features}
like SPD matrices~\cite{huang2017riemannian},
Grassmannians~\cite{huang2018building},
hyperbolic spaces~\cite{ganea2018hyperbolic,peng2021hyperbolic,chami2019hyperbolic,liu2019hyperbolic,gulcehre2018hyperbolic,shimizu2020hyperbolic}
or more general manifolds~\cite{chakraborty2018manifoldnet,banerjee2020volterranet,pfau2020disentangling}.
Related are autoencoder networks whose latent space is manifold-valued~\cite{davidson2018hyperspherical,falorsi2018explorations,de2018topological,schonsheck2019chart}.
We believe that the design of such models should be coordinate independent as well whenever the choice of coordinates is non-canonical.

In a broader sense, we believe that neural networks should generally be constrained such that they \emph{preserve the mathematical structure of the learning task};
see also the review paper by \cite{celledoni2020structure}.
While our coordinate independent CNNs respect the manifold's $G$-structure,
\citet{hoffmann2020algebranets} and \citet{alej2020algebraic} propose networks that respect algebraic structures of features,
\citet{greydanus2019hamiltonian} construct energy conserving Hamiltonian networks that preserve a symplectic structure
and \citet{hernandez2021structure} design models that enforce the metriplectic structure of dissipative Hamiltonian systems, ensuring that they comply with the first and second principle of thermodynamics.
In general, any mathematical structure exhibits certain symmetries, formalized by their automorphism group.
\citet{bronstein2021geometric5Gs} recently proposed an ``Erlangen Programme of Deep Learning''
targeted at a systematic classification of network architectures with a focus on the underlying mathematical structures and their symmetries.
Developing such a classification for $\GM$-coordinate independent CNNs on Riemannian manifolds,
our work reached a first milestone in this programme.

\subsection*{Acknowledgement}
We would like to thank Leon Lang and Gabriele Cesa for plenty of valuable discussions and for their feedback on the manuscript.
We also thank Erik Bekkers for discussions on scale equivariant convolutions and convolutions on homogeneous spaces.

\subsection*{List of theorems and definitions}
\theoremlisttype{allname}
\listtheorems{thm,cor,lem,dfn}

%% file: chapters/apx01_coordinate_bases.tex

\section{Relation to the coordinate chart formalism of differential geometry}
\label{apx:coordinate_bases}

This appendix serves the purpose of drawing connections between the \emph{bundle formalism}, underlying the theory of coordinate independent CNNs, and the \emph{coordinate chart formalism}, which one likely encounters in a first study of differential geometry.
The main difference between both is that the bundle formalism refers to points~$p$ of the base space $M$ in a \emph{coordinate free} way.
If required, coordinates are directly assigned to the fibers (e.g. tangent spaces) via local bundle trivializations.
In contrast, the chart formalism relies on \emph{coordinate charts} (diffeomorphisms)
\begin{align}
    x: M \supseteq U \to V \subseteq \R^d \,,
\end{align}
which assign coordinates to local patches $U$ of the manifold.
Local bundle trivializations and gauge transformations between them are \emph{induced as differentials of charts and chart transition functions}.
In this section we work out the connection between both formalisms.
An overview of the results is given in Table~\ref{tab:coord_charts_gauge_trafos}.

\etocsettocdepth{3}
\etocsettocstyle{}{} 
\localtableofcontents

We start in Section~\ref{apx:tangent_cotangent_dual_bases} by briefly introducing tangent spaces $\TpM$ as spaces of directional derivative operators, from which the cotangent spaces $\TspM$ follow as dual spaces.
Section~\ref{apx:differentials_gradients_jacobians} defines general differentials and the more specific gradients and Jacobians.
Based on these preparations, we will in Section~\ref{apx:coord_basis_def} define \emph{coordinate bases} (holonomic bases)
${\pig[ \frac{\partial}{\partial x_1}\big|_p, \ \dots,\ \frac{\partial}{\partial x_d}\big|_p \pig]\ \in \FpM}$
of the tangent spaces $\TpM$, which are spanned by directional derivative operators along the coordinate grid that is pulled by the chart from $V$ to~$U$.
The dual bases
${\pig[ \hat{d}x_\mu|_p, \ \dots,\ \hat{d}x_\mu|_p \pig]}$
of the cotangent spaces $\TspM$ are given by the gradients of the chart components $x_\mu$.
Transition maps between charts induce covariant and contravariant gauge transformations between the corresponding bases, which are derived in Section~\ref{apx:chart_transition_induced_gauge_trafos}.
Section~\ref{apx:correspondences_bundle_trivializations} interprets the coordinate bases as local bundle trivializations and makes the connection between the bundle formalism and the chart formalism precise.
The bases and trivializations induced from coordinate charts do not cover all possible trivializations, such that one distinguishes between coordinate bases and non-coordinate bases (the bundle formalism allows with general non-coordinate bases).
In the physics literature, non-coordinate bases are usually introduced via \emph{vielbein fields}.
Section~\ref{apx:vielbein_fields} argues that these vielbein fields are just $\GL{d}$-valued gauge transformations from general frames in $\FM$ into a given $G$-structure~$\GM$, within which one can subsequently apply $G$-valued gauge transformations that preserve the $G$-structure.

Comprehensive introductions to the chart formalism are given in~\cite{nakahara2003geometry,schullerGeometricalAnatomy2016,carroll2004spacetime}.
A more rigorous exposition is found in~\cite{schullerGeometricalAnatomy2016}.

We want to remind the reader that we are \emph{not} making use of covariant and contravariant indices.
Indices will always appear as subscripts, with Greek letters $\mu,\nu,\dots$ signaling coordinate chart related indices and Latin letters $i,j,\dots$ signaling indices of general gauges.
Superscripts $A,B,\dots$ are preserved for labeling different charts or gauges.

\subsection{Tangent spaces, cotangent spaces and dual bases}
\label{apx:tangent_cotangent_dual_bases}

\subsubsection{Tangent spaces in terms of directional derivatives}
A common definition of the tangent spaces $\TpM$ of a manifold $M$ is as vector spaces of directional derivative operators at $p \in M$, which we will briefly motivate here.
Let $f\in C^\infty(M)$, that is, $f:M\to\R$ is a smooth map, and, for some interval $I\subseteq\R$ containing $0$, let $\gamma: I \to M$ be a smooth curve which passes at time $t = 0$ through~$p$, i.e. satisfies $\gamma(0) = p$.
One then defines the \emph{directional derivative operator} at~$p$ along~$\gamma$ as the linear operator
\begin{align}\label{eq:tangent_vector_directional_derivative}
    v_\gamma: C^\infty(M) \to \R,\ \ \ f \mapsto \big(f \circ \gamma \big)'(0) \,.
\end{align}
As the derivative is taken along the direction of $\gamma$, that is, tangential to it, $v_\gamma$ is called \emph{tangent vector}.
It can be thought of as the velocity of a particle with trajectory $\gamma$ at time $t=0$.
For later reference we give the following simple commutative diagram, which shows the pullback $f\circ\gamma$ of $f$ from $M$ to $\R$ via $\gamma$, in terms of which the directional derivative is defined:
\begin{equation}\label{cd:directional_derivative}
    \begin{tikzcd}[row sep=2.5em, column sep=4em]
          \R \supset I
                \arrow[r, "\gamma"]
                \arrow[rr, rounded corners, to path={ 
                        |- node[below, pos=.75]{\small$f \circ \gamma$} ([yshift=-3.5ex]\tikztotarget.south)
                        -- (\tikztotarget.south)
                        }]
        & M     \arrow[r, "f"]
        & \R
    \end{tikzcd}
\end{equation}

One can show that the space of all tangent vectors to curves at $p$ forms a $d$-dimensional vector space
\begin{align}
    \TpM\ :=\ \big\{ v_\gamma \,\big|\, \gamma\ \text{is a smooth curve through}\ p \big\} \,,
\end{align}
known as the tangent space at~$p$.
For more details on the definition of tangent vectors and the vector space structure of the tangent spaces we refer to~\cite{schullerGeometricalAnatomy2016}.

Having defined the tangent spaces as vector spaces, one might choose to treat tangent vectors as abstract geometric vectors, thereby ``forgetting'' about their definition via directional derivatives (or any alternative definition made).
We do this at most places, but refer back to the definition via directional derivatives in the following sections to derive differentials of smooth maps and coordinate bases.

\subsubsection{Cotangent spaces}
\label{apx:cotangent_spaces}
As real vector spaces, the tangent spaces $\TpM$ have corresponding \emph{dual spaces} $\TspM := (\TpM)^*$, the \emph{cotangent spaces}.
By the definition of dual spaces, they consist of linear functionals
\begin{align}
    \omega: \TpM \to \R \,,
\end{align}
which are in differential geometry usually called \emph{covectors} or \emph{1-forms}.
Together with the (co)vector addition
$(\omega + \widetilde{\omega})(v) = \omega(v) + \widetilde{\omega}(v)$
and scalar multiplication 
$(\lambda \cdot \omega)(v) = \lambda\cdot (\omega(v))$,
the cotangent spaces are vector spaces themselves.

As finite-dimensional duals of each other, $\TpM$  and $\TspM$ are isomorphic and are thus in particular of the same dimensionality $d = \dim(M) = \dim(\TpM) = \dim(\TspM)$.
The isomorphism between both is, however, not canonical.
A vector space isomorphism can be specified via a (non-degenerate) bilinear form $\eta_p: \TpM \times \TpM \to \R$ on $\TpM$, for instance a Riemannian metric, via
\begin{align}
    \widehat{\eta}_p:\ \TpM \to \TspM,\ \ v \mapsto \eta_p(v, \cdot) \,,
\end{align}
which determines the linear functional
$\widehat{\eta}_p(v):\ \TpM \to \R,\ \ w \mapsto \eta_p(v, w)$.

\subsubsection{Dual bases}
Any basis $\big[e_i\big]_{i=1}^d$ of $\TpM$ canonically induces a \emph{dual basis} $\big[e_i^*\big]_{i=1}^d$ of $\TspM$, defined to satisfy the relations
\begin{align}
    e^*_i e_j = \delta_{ij} \quad \textup{for any }\ i,j \in 1,\dots,d \,.
\end{align}
Let $\big[e^A_i\big]_{i=1}^d$ and $\big[e^B_i\big]_{i=1}^d = \big[e^A_i\big]_{i=1}^d \lhd \big(g^{BA}\big)^{-1}$ be two bases of $\TpM$, which are related by the right action~$\lhd$ of the (inverse) structure group element $\big(g^{BA}\big)^{-1} \in \GL{d}$ in Eq.~\eqref{eq:frame_rightaction}, that is, for $j=1,\dots,d$\ :
\begin{align}\label{eq:general_tangent_basis_gauge_trafo}
    e^B_j\ =\ \sum_l e^A_l \big(g^{BA}\big)^{-1}_{lj}
\end{align}
The dual basis $\big[e^{A,*}_i\big]_{i=1}^d$ transforms accordingly under that left action which sends $e_i^{A,*}$ to
\begin{align}\label{eq:general_cotangent_basis_gauge_trafo}
    e_i^{B,*}\ =\ \sum_k g^{BA}_{ik} e_k^{A,*} \,.
\end{align}
This is affirmed by pairing
\begin{align}
    e_i^{B,*} e_j^B
    \ &=\ \sum_{k,l} g^{BA}_{ik} e_k^{A,*} e_l^A \big(g^{BA}\big)^{-1}_{lj}  \notag \\
    \ &=\ \sum_{k,l} g^{BA}_{ik} \delta_{kl} \big(g^{BA}\big)^{-1}_{lj} \notag \\
    \ &=\ \sum_{k} g^{BA}_{ik} \big(g^{BA}\big)^{-1}_{kj} \notag \\
    \ &=\ \delta_{ij} \,.
\end{align}
The inverse transformation behavior of bases and dual bases is usually referred to as \emph{covariant} and \emph{contravariant} transformation.
Note the similarity of the dual basis transformation to the contravariant transformations $\psi^B = g^{BA} \psi^A$ of gauges in Eq.~\eqref{eq:gauge_trafo_local_def_21} and $v^B = g^{BA} v^A$ of vector components in \eqref{eq:components_leftaction}.
Indeed, gauges are just choices of a cotangent basis as further discussed below.

\subsection{Differentials, gradients and Jacobians}
\label{apx:differentials_gradients_jacobians}

In vector calculus one considers functions $\phi: \R^m \to \R^n$, which can at any point $p\in \R^m$ be linearly approximated by their Jacobian matrix (or total derivative or differential) $d\phi_p = \big(\frac{\partial\phi_i}{\partial x_j} \big|_p \big)_{ij}$.
Here we introduce the generalization of this concept to differentials of smooth functions between smooth manifolds.

\paragraph{Differentials in general:}
Let $\phi: M \to N$ be a smooth map between smooth manifolds $M$ and $N$.
At any point $p\in M$, such a map induces a differential (or pushforward)
\begin{align}
    d\phi_p : \TpM \to \TphipN,\ \ v \mapsto d\phi_p(v)
\end{align}
which linearly maps tangent vectors at~$p$ to tangent vectors at $\phi(p)$.
For the definition of tangent spaces in terms of directional derivatives in Eq.~\eqref{eq:tangent_vector_directional_derivative}, the pushforward of $v \in \TpM$ along $\phi$ is explicitly given by
\begin{align}
    d\phi_p(v): C^\infty(N) \to \R,\ \ \ f \mapsto \big( d\phi_p(v) \big)(f)\ :=\ v(f \circ \phi) \,,
\end{align}
that is, by the application of~$v$ on the pullback $f \circ \phi: M \to \R$ of $f: N\to \R$ via $\phi$.
These definitions are clarified by the following two commutative diagrams:
\begin{equation}
\begin{tikzcd}[column sep=70pt, row sep=30, font=\normalsize]
    M
        \arrow[r, "\phi"]
        \arrow[dr, "f \circ \phi"']
    &
    N
        \arrow[d, "\ f"]
    \\
    & \R
\end{tikzcd}
\qquad\qquad
\begin{tikzcd}[column sep=60pt, row sep=30, font=\normalsize]
    C^\infty(M)
        \arrow[d, "v\ "']
    &
    C^\infty(N)
        \arrow[l, "(\,\cdot\,) \circ \phi"']
        \arrow[dl, "d\phi(v)"]
    \\
    \R
\end{tikzcd}
\end{equation}

From this definition it follows immediately that the differential of the composition of smooth maps equals the composition of their individual differentials, which is just the chain rule:
\begin{align}
    d(\phi \circ \psi)_p\ =\ d\phi_{\psi(p)} \circ d\psi_p
\end{align}
If $\phi$ is invertible (a diffeomorphism) it furthermore follows that its differential is a vector space isomorphism whose inverse equals the differential of $\phi^{-1}$, that is,
\begin{align}\label{eq:differential_inverse}
    \big( d\phi_{p} \big)^{-1}\ =\ d\big( \phi^{-1} \big)_{\phi(p)} \,.
\end{align}

Together, the differentials $d\phi_p$ at individual points $p\in M$ imply a vector bundle morphism (a fiber-wise linear bundle map, see Sections~\ref{sec:fiber_bundles_general}) between the tangent bundles of $M$ and $N$:
\begin{equation}
\begin{tikzcd}[column sep=60pt, row sep=35, font=\normalsize]
    TM
        \arrow[r, "d\phi"]
        \arrow[d, "\piTM"']
    &
    TN
        \arrow[d, "\piTM"]
    \\
    M
        \arrow[r, "\phi"']
    &
    N
\end{tikzcd}
\end{equation}

Note that we are in this appendix using a different notation, namely $d\phi$, than in the main paper, where we instead write $\dphiTM$.
We decided for the former to connect to the usual notation $dx_\mu$ for the chart induced bases of cotangent spaces.
The latter is used in the main text to emphasize the similarity to the bundle maps $\dphiFM$, $\dphiGM$ and $\dphiA$, which are induced on the associated bundles $\FM$, $\GM$ and $\A$.

\paragraph{Gradients:}
In the case of smooth real-valued functions $\phi: M \to \R$, i.e. $\phi \in C^\infty(M)$, the differential $d\phi_p: \TpM \to T_{\phi(p)}\R$ pushes vectors $v$ in $\TpM$ to vectors $d\phi(v): C^\infty(\R) \to \R,\ f \mapsto v(f \circ \phi)$ in $T_{\phi(p)}\R$.
By leveraging the canonical isomorphism
\begin{align}\label{eq:canon_isom_TR_R}
    \iota_{\R}: T_{\phi(p)}\R \xrightarrow{\sim} \R,\ \ v \mapsto v(\id_{\R})
\end{align}
one defines the \emph{gradient operator}
\begin{align}
    \hat{d}_p: C^\infty(M) \to \TspM,\ \ \phi \mapsto  \hat{d}\phi_p := \iota_{\R} \circ d\phi_p = \big( d\phi_p(\,\cdot\,) \big)(\id_{\R}) \,,
\end{align}
which sends smooth functions $\phi$ to covectors%
\footnote{
    The gradient field is often defined as a \emph{vector} field $\nabla f := (\hat{d}f)^{\sharp^\eta}$ which is computed from the \emph{covector} field $\hat{d}f$ via the musical isomorphism $\sharp^\eta: \TsM \to \TM$ corresponding to the metric (``raising indices'').
}
$\hat{d}\phi$, which in turn act on vectors as
\begin{align}\label{eq:gradient_vector_action}
    \hat{d}\phi_p: \TpM \to \R,\ \ v \mapsto \hat{d}\phi_p(v) = \big( d\phi_p(v) \big)(\id_{\R}) = v(\id_{\R} \circ \phi) = v(\phi) \,.
\end{align}
By an \emph{abuse of notation} one usually drops the ``hat'' on $\hat{d}$ and immediately defines $d\phi_p(v) := v(\phi)$.
While this notation is very common, we stick in the following with the ``hat'' to make the requirement for the canonical isomorphism $\iota_{\R}$ explicit.

In Section~\ref{apx:coord_basis_def} below we will see that the bases of $\TspM$ which are dual to coordinate bases of $\TpM$ are given by the gradient 1-forms $\hat{d}x_\mu|_p$, where $x_\mu$ are the components of the coordinate chart.

\paragraph{Jacobians:}
Specifically for functions $\phi: \R^n \to \R^m$ between (subsets of) Euclidean spaces the differential
$d\phi_{x_0}: {T_{x_0}\R^n \to T_{\phi(x_0)}\R^m}$
is easily seen to coincide with the \emph{Jacobian} $\frac{\partial \phi}{\partial x} \big|_{x_0}: \R^n \to \R^m$
after canonically identifying $T_p\R^k \cong \R^k$ in both the domain and codomain.
The canonical isomorphism is here given by
\begin{align}\label{eq:canonical_iso_TRk_Rk}
    \iota_{\R^k}: v \mapsto \big( v(\proj_1), \dots, v(\proj_k) \big) \,,
\end{align}
which generalizes $\iota_{\R}$ from Eq.~\eqref{eq:canon_isom_TR_R} to multiple dimensions.
As the calculation is mostly similar as in the case of gradients, we will not repeat it here but visualize the idea via a commutative diagram:
\begin{equation}\label{cd:jacobian_def}
    \begin{tikzcd}[row sep=2.5em, column sep=4em]
          \R^n
                \arrow[rrr, rounded corners, to path={ 
                        |- node[below, pos=.75]{$\frac{\partial \phi}{\partial x} \Big|_{x_0}$} ([yshift=-3.5ex]\tikztotarget.south)
                        -- (\tikztotarget.south)
                        }]
        & T_{x_0}\R^n
                \arrow[l, "\iota_{\R^n}"']
                \arrow[r, "d\phi |_{x_0}"]
        & T_{\phi(x_0)}\R^m
                \arrow[r, "\iota_{\R^m}"]
        & \R^m
    \end{tikzcd}
\end{equation}

If $\phi$ is invertible, the identity in Eq.~\eqref{eq:differential_inverse} becomes
\begin{align}\label{eq:inv_fct_thm_jacobian}
    \frac{\partial \phi}{\partial x} \bigg|_{x_0}^{-1} \ =\ 
    \frac{\partial \phi^{-1}}{\partial x} \bigg|_{\phi(x_0)} \,,
\end{align}
which is just the inverse function theorem.
We will use this identity later on to invert gauge transformations between different coordinate bases which are induced as Jacobians of chart transition maps.

\subsection{Chart induced coordinate bases}
\label{apx:chart_induced_bases_main}

In this Section we consider \emph{coordinate charts} of the form
\begin{align}
  x: U \to V \,,
\end{align}
which diffeomorphically assign coordinates $x(p) \in V \subseteq \R^d$ to each point $p \in U \subseteq M$.
Any such chart induces a natural choice of bases for the tangent spaces $\TpM$ over~$U$, known as \emph{coordinate bases}.
The dual spaces $\TspM$ of the tangent spaces over~$U$ are accordingly endowed with dual coordinate bases of cotangent vectors.
Transition maps between the coordinates of two charts induce gauge transformations which translate between the corresponding coordinate bases.
These gauge transformations are given by the Jacobians of the transition maps.

\subsubsection{Charts and induced coordinate bases}
\label{apx:coord_basis_def}

\paragraph{Coordinate bases for $\bm\TpM$:}
To motivate the definition of coordinate bases, observe that~$x$ implies a \mbox{``coordinate grid''} on~$U$ by pulling the canonical coordinate grid on $V$ back to the manifold.
The coordinate basis at a specific point $p \in U$ can then be thought of as consisting of those $d$ many \emph{directional derivative operators} which are going \emph{along the coordinate grid lines of~$x$ on~$U$}.

To make this more precise, consider first the curves
\begin{align}
    \widetilde{\gamma}_\mu: I \to V,\ \ \ t \mapsto x(p) + t \epsilon_\mu \quad\qquad \mu = 1,\dots,d
\end{align}
which pass at time $t=0$ with unit velocity in $\mu$-direction through~$x(p) \in V$.
Mapping those $\widetilde{\gamma}_\mu$ via the chart to~$U$ defines the above mentioned curves
\begin{align}
    \gamma_\mu: I \to U,\ \ \ t \mapsto\,
    x^{-1}\mkern-2mu \circ \widetilde{\gamma}_\mu (t) \ =\ 
    x^{-1} \big( x(p) + t \epsilon_\mu \big)
\end{align}
which pass at time $t=0$ along the coordinate grid of~$x$ on~$U$ through~$p$.
The $d$-dimensional coordinate basis of~$\TpM$ induced by~$x$ is then given by the directional derivative operators in Eq.~\eqref{eq:tangent_vector_directional_derivative} along the paths~$\gamma_\mu$.
Denoting the $\mu$-th basis vector by the usual abuse of notation as $\frac{\partial}{\partial x_\mu}\big|_p$ one therefore defines:
\begin{align}\label{eq:coord_basis_def}
    \frac{\partial}{\partial x_\mu} \bigg|_p \!:\ \ f\, \mapsto\,
    \frac{\partial}{\partial x_\mu} \bigg|_p f
    \ \ :=&\ \ \big(f \circ \gamma_\mu \big)'(0) \notag \\
    \ \  =&\ \ \big(f \circ x^{-1} \circ \widetilde{\gamma}_\mu \big)'(0) \notag \\
    \ \  =&\ \ \big(f \circ x^{-1}\big( x(p) + t\epsilon_\mu \big) \big)'(0) \notag \\
    \ \  =&\ \ \pig[\mkern1.5mu \partial_\mu \big(f \circ x^{-1} \big)\pig] \big(x(p)\big)
\end{align}
In the last step we identified the usual $\mu$-th partial derivative of the pullback $f\circ x^{-1}: V \to \R$, which motivates the notation $\frac{\partial}{\partial x_\mu}\big|_p$.
These definitions are visualized in the following commutative diagram which extends the diagram in Eq.~\eqref{cd:directional_derivative}:
\begin{equation}
    \begin{tikzcd}[row sep=4.em, column sep=6.em]
        & V     \arrow[rd, "f\circ x^{-1}"]
        \\
          \R \supset I
                \arrow[r, "\gamma_\mu"]
                \arrow[ru, "\widetilde{\gamma}_\mu"]
                \arrow[rr, rounded corners, to path={ 
                        |- node[below, pos=.75]{\small$f \circ \gamma_\mu$} ([yshift=-3.5ex]\tikztotarget.south)
                        -- (\tikztotarget.south)
                        }]
        & U     \arrow[r, pos=.4, "f"]
                \arrow[u, pos=.4, "x"]
        &[1.4em] \R
    \end{tikzcd}
\end{equation}

\paragraph{Dual coordinate bases for $\bm\TspM$:}

As stated in Section~\ref{apx:tangent_cotangent_dual_bases}, any basis of $\TpM$ induces a \emph{dual basis} of $\TspM$.
Specifically for coordinate bases, spanned by vectors $\frac{\partial}{\partial x_\mu} \big|_p$, the dual basis elements are given by the \emph{gradients $\hat{d}x_\mu|_p = \hat{d}(x_\mu)_p \in \TspM$ of the chart components} $x_\mu = \proj_\mu \circ x: U \to \R$.
That these gradients do indeed make up the dual basis, is easily seen by acting on the basis vectors as defined in Eq.~\eqref{eq:gradient_vector_action}:
\begin{align}
    \hat{d}x_\mu \big|_p\ \frac{\partial}{\partial x_\nu} \bigg|_p
    \ &=\ \frac{\partial}{\partial x_\nu} \bigg|_p x_\mu \notag \\
    \ &=\ \pig[ \partial_\nu \big( x_\mu \circ x^{-1} \big) \pig] \big(x(p)\big) \notag \\
    \ &=\ \pig[ \partial_\nu \big( \proj_\mu \big) \pig] \big(x(p)\big) \notag \\
    \ &=\ \delta_{\mu\nu} \,.
\end{align}

\paragraph{Chart differentials as canonical local trivialization:}

Given that the chart maps from $U \subseteq M$ to $V \subseteq \R^d$, its differentials at $p\in U$ are maps of the form
\begin{align}
    dx_p: \TpM \to T_{x(p)}\R^d \,.
\end{align}
Employing the canonical isomorphism $\iota_{\R^d}$ from $T_{x(p)}\R^d$ to $\R^d$ from Eq.~\eqref{eq:canonical_iso_TRk_Rk} once again, we obtain a map
\begin{align}\label{eq:chart_differential_via_gradients}
    \qquad
    \hat{d}x_p: \TpM \to \R^d,\ \ \ v\ \mapsto\ \hat{d}x_p (v)
    :=\ &\iota_{\R^d} \circ dx_p (v) \notag \\
     =\ & \Big( \big(dx_p(v) \big)(\proj_1) \,,\,\dots,\, \big(dx_p(v) \big)(\proj_d) \Big)^\top \notag \\
     =\ & \Big( v\big(\proj_1 \circ x \circ x^{-1}\big)(x(p)) \,,\,\dots,\, v\big(\proj_1 \circ x \circ x^{-1}\big)(x(p)) \Big)^\top \notag \\
     =\ & \Big( v(x_1(p)) \,,\,\dots,\, v(x_d(p)) \Big)^\top \notag \\
     =\ & \Big( \hat{d}x_1 |_p(v) \,,\,\dots,\, \hat{d}x_d |_p(v) \Big)^\top
\end{align}
after identifying the individual chart component gradients in the last step.
Note that the action of this chart differential on the $\mu$-th coordinate basis yields
\begin{align}
    \hat{d}x_p \: \frac{\partial}{\partial x_\mu} \bigg|_p
     \ &=\ \bigg( \hat{d}x_1|_p \: \frac{\partial}{\partial x_\mu} \bigg|_p \,,\,\dots,\, \hat{d}x_d|_p \: \frac{\partial}{\partial x_\mu} \bigg|_p \bigg)^\top \notag \\
     \ &=\ \big( \delta_{\mu1} \,,\,\dots,\, \delta_{\mu d} \big)^\top \notag \\
     \ &=\ \epsilon_\mu \,,
\end{align}
that is, the $\mu$-th unit vector $\epsilon_\mu$ of $\R^d$.
This implies that $\hat{d}x_p: \TpM \to \R^d$ plays the role of a \emph{gauge} $\psi_p$ at~$p$.
One could therefore equally well have started by defining a cotangent basis and setting
\begin{align}\label{eq:coord_basis_vector_via_chart_differential}
    \frac{\partial}{\partial x_\mu} \bigg|_{x(p)}\ =\ \hat{d}x_p^{-1} (\epsilon_\mu) \,,
\end{align}
which is the analog of Eq.~\eqref{eq:framefield_gauge_equivalence} in the chart formalism.

\subsubsection{Chart transition maps and induced gauge transformations}
\label{apx:chart_transition_induced_gauge_trafos}

Different charts induce different coordinate bases.
Chart transitions therefore induce gauge transformations, i.e. transformations of bases and vector coefficients, which we derive in this section.

In the following we consider two arbitrary, overlapping charts $x^A: U^A \to V^A$ and $x^B: U^B\to V^B$.
The different coordinates which they assign to the overlap $U^A \cap U^B \neq \varnothing$ are then related via \emph{chart transition maps}
\begin{align}\label{eq:chart_transition_fct}
  x^B\circ\left(x^A\right)^{-1} \!:\ x^A\big(U^A\cap U^B\big)\to x^B\big(U^A\cap U^B\big) \,.
\end{align}

\paragraph{Transformation of tangent coordinate bases:}
The coordinate bases of $\TpM$ which are induced by the two charts are according to the last line of Eq.~\eqref{eq:coord_basis_def} by their action on $f \in C^\infty(M)$ defined as
\begin{align}
    \frac{\partial}{\partial x^A_\mu} \bigg|_p f
    \ =\ \Big[\mkern1.5mu \partial_\mu \pig(f \circ \big(x^A\big)^{-1} \pig)\Big] \big(x^A(p)\big)
    \qquad \text{and} \qquad
    \frac{\partial}{\partial x^B_\mu} \bigg|_p f
    \ =\ \Big[\mkern1.5mu \partial_\mu \pig(f \circ \big(x^B\big)^{-1} \pig)\Big] \big(x^B(p)\big) \ ,
    \quad
\end{align}
which is visualized by the following commutative diagram:
\begin{equation}\label{cd:scalar_field_chart_expressions}
    \begin{tikzcd}[row sep=4.em, column sep=6.em] 
        V^A \supset x^A \big( U^A \cap U^B \big)
                \arrow[rd, "
                    f\circ \big(x^A\big)^{-1}
                    "]
                \arrow[dd, rounded corners, to path={ 
                        -| node[left, pos=.75]{\small$x^B \circ \big(x^A\big)^{-1}$} ([xshift=-3.5ex]\tikztotarget.west)
                        -- (\tikztotarget.west)
                        }]
        \\
        U^A \cap U^B
                \arrow[r, pos=.4, "f"]
                \arrow[u, "x^A"]
                \arrow[d, "x^B"']
        &
        \R
        \\
        V^B \supset x^B \big( U^A \cap U^B \big)
                \arrow[ru, "
                    f\circ \big(x^B\big)^{-1}
                    "']
    \end{tikzcd}
\end{equation}

Via the chart transition maps, the different coordinate bases relate by
\begin{align}\label{eq:coord_basis_trafo_action_f}
    \frac{\partial}{\partial x^B_\mu} \bigg|_p f
    \ &=\ \Big[\mkern1.5mu \partial_\mu \pig(f \circ \big(x^B\big)^{-1} \pig) \Big] \big(x^B(p)\big) \notag \\
    \ &=\ \Big[\mkern1.5mu \partial_\mu \pig(f \circ \big(x^A\big)^{-1} \circ x^A \circ \big(x^B\big)^{-1} \pig) \Big] \big(x^B(p)\big) \,, \notag
\intertext{
which, making use of the multivariate chain rule, further leads to:
}
    \frac{\partial}{\partial x^B_\mu} \bigg|_p f
    \ &=\ \sum_{\nu=1}^d
        \Big[\mkern1.5mu \partial_\nu \pig(f \circ \big(x^A\big)^{-1} \pig)\Big] \big(x^A(p)\big) \cdot
        \Big[\mkern1.5mu \partial_\mu \pig(x^A_\nu \circ \big(x^B\big)^{-1} \pig)\Big] \big(x^B(p)\big) \notag \\
    \ &=\ \sum_{\nu=1}^d \,
        \frac{\partial f}{\partial x^A_\nu} \bigg|_p \ 
        \frac{\partial x^A_\nu}{\partial x^B_\mu} \bigg|_{x^B(p)}
\end{align}
In the last step we introduced the usual abuse of notation%
\footnote{
    The ``abuse'' is that $x^A$ is interpreted as a function of $x^B(p)$, and should therefore rather be written $x^A \circ \big(x^B\big)^{-1}$ as made precise on the right-hand side.
}
\begin{align}\label{eq:abuse_of_notation_jacobian}
    \frac{\partial x^A_\nu}{\partial x^B_\mu} \bigg|_{x^B(p)}
    :=\ \partial_\mu \pig(x^A_\nu \circ \big(x^B\big)^{-1} \pig) \big( x^B(p) \big)
\end{align}
for the components of the \emph{Jacobian}
\begin{align}
    \frac{\partial x^A}{\partial x^B} \bigg|_{x^B(p)}
    =\ \hat{d}x^A_p \circ \hat{d}(x^B_p)^{-1}
\end{align}
\emph{of the transition maps}.
Dropping~$f$ from Eq.~\eqref{eq:coord_basis_trafo_action_f}, we identify the transformation law
\begin{align}\label{eq:coord_bases_trafo_law}
    \frac{\partial}{\partial x^B_\mu} \bigg|_p
    \ =\ \sum_{\nu=1}^d \,
        \frac{\partial}{\partial x^A_\nu} \bigg|_p \ 
        \frac{\partial x^A_\nu}{\partial x^B_\mu} \bigg|_{x^B(p)}
\end{align}
of tangent coordinate bases.
We did hereby choose to write the Jacobian on the right of the basis vector to emphasize that the change of basis is to be understood as a \emph{right action}.
Doing so, we need to warn the reader that $\frac{\partial}{\partial x_\nu}\big|_p$ is just an abuse of notation for the basis vector but does not imply an action of a differential operator on the Jacobian on the right.

\paragraph{Transformation of cotangent coordinate bases:}
The contravariant transformation law of cotangent space coordinate bases follows from the inverse transformation of dual bases in Eq.~\eqref{eq:general_cotangent_basis_gauge_trafo} relative to~\eqref{eq:general_tangent_basis_gauge_trafo}.
To apply this relation, we first adapt Eq.~\eqref{eq:coord_bases_trafo_law} to our convention that bases transform according to a right action with an \emph{inverse} group element.
This is achieved by applying Eq.~\eqref{eq:inv_fct_thm_jacobian} to invert the Jacobian (remember the abuse of notation)
\begin{align}
    \frac{\partial x^A}{\partial x^B} \bigg|_{x^B(p)} \ =\ 
    \frac{\partial x^B}{\partial x^A} \bigg|_{x^A(p)}^{-1}
\end{align}
which implies:
\begin{align}\label{eq:coord_bases_trafo_law_with_inv}
    \frac{\partial}{\partial x^B_\mu} \bigg|_p
    \ =\ \sum_{\nu=1}^d \,
        \frac{\partial}{\partial x^A_\nu} \bigg|_p \ 
        \frac{\partial x^A_\nu}{\partial x^B_\mu} \bigg|_{x^B(p)}
    \ =\ \sum_{\nu=1}^d \,
        \frac{\partial}{\partial x^A_\nu} \bigg|_p \ 
        \bigg( \frac{\partial x^A}{\partial x^B} \bigg|_{x^B(p)} \bigg)_{\nu\mu}
    \ =\ \sum_{\nu=1}^d \,
        \frac{\partial}{\partial x^A_\nu} \bigg|_p \ 
        \bigg( \frac{\partial x^B}{\partial x^A} \bigg|_{x^A(p)}^{-1} \bigg)_{\nu\mu}
\end{align}
The cotangent basis elements therefore transform according to Eqs.~\eqref{eq:general_tangent_basis_gauge_trafo} and ~\eqref{eq:general_cotangent_basis_gauge_trafo} like
\begin{align}\label{eq:chart_component_gradient_trafo_law}
    \hat{d}x^B_\mu|_p \ =\ 
    \sum_{\nu=1}^d\ 
        \frac{\partial x^B_\mu}{\partial x^A_\nu} \bigg|_{x^A(p)}
        \hat{d}x^A_\nu|_p \,.
\end{align}

\paragraph{Transformation of chart differentials:}
The expression of chart differentials $\hat{d}x^A|_p$ in terms of chart component gradients $\hat{d}x^A_\mu|_p$ in Eq.~\eqref{eq:chart_differential_via_gradients} allows to deduce their transformation law from that in Eq.~\eqref{eq:chart_component_gradient_trafo_law}.
Alternatively, one obtains the transformation law by right multiplying with the identity in the form $\id_{\TpM} = \hat{d}x^A|_p \circ \big( \hat{d}x^A|_p \big)^{-1}$ and identify a left multiplication with the Jacobian of the chart transition maps:
\begin{align}\label{eq:chart_differential_trafo_law}
    \hat{d}x^B|_p
    \ &=\ \hat{d}x^B|_p \circ \big( \hat{d}x^A|_p \big)^{-1} \circ \hat{d}x^A|_p \notag \\
    \ &=\ \frac{\partial x^B}{\partial x^A} \bigg|_{x^A(p)} \hat{d}x^A|_p
\end{align}
Note that this result is simply the matrix expression of Eq.~\eqref{eq:chart_component_gradient_trafo_law}.

\paragraph{Transformation of vector coefficients:}
Vectors $v \in \TpM$ are relative to a coordinate basis
$\big[\frac{\partial}{\partial x^B_\mu} \big|_p \big]_{\mu=1}^d$
expressed by coefficients $v^A \in \R^d$:
\begin{align}
    v\ =\
    \sum_{\mu=1}^d \,
    v_\mu^A \frac{\partial}{\partial x^A_\mu} \bigg|_p
\end{align}
The individual coefficients are recovered by the action of the cotangent basis:
\begin{align}
    \hat{d}x^A_\mu \big|_p(v)
    \ =\ \hat{d}x^A_\mu \big|_p\ \sum_{\nu=1}^d \, v_\nu^A \frac{\partial}{\partial x^A_\nu} \bigg|_p
    \ =\ \sum_{\nu=1}^d \, v_\nu^A \delta_{\mu\nu}
    \ =\ v^A_\mu
\end{align}
This implies that the coefficients transform contravariantly, just as the cotangent coordinate basis:
\begin{align}
    v^B_\mu
    \ =\ \hat{d}x^B_\mu \big|_p (v)
    \ =\ \sum_{\nu=1}^d\ 
        \frac{\partial x^B_\mu}{\partial x^A_\nu} \bigg|_{x^A(p)}
        \hat{d}x^A_\nu|_p (v)
    \ =\ \sum_{\nu=1}^d\ 
        \frac{\partial x^B_\mu}{\partial x^A_\nu} \bigg|_{x^A(p)}
        v^A_\nu
\end{align}
It is easily asserted that this transformation law does indeed lead to a coordinate independent representation of coordinate free vectors $v\in\TpM$:
\begin{align}
    \sum_\mu \frac{\partial}{\partial x^B_\mu} \bigg|_p v_\mu^B
    \ =\ \sum_{\mu,\nu,\rho} \frac{\partial}{\partial x^A_\nu} \bigg|_p \,
        \frac{\partial x^A_\nu}{\partial x^B_\mu} \bigg|_{x^B(p)} \,
        \frac{\partial x^B_\mu}{\partial x^A_\rho} \bigg|_{x^A(p)} \,
        v^A_\rho
    \ =\ \sum_{\nu,\rho} \frac{\partial}{\partial x^A_\nu} \bigg|_p \,
        \delta_{\nu\rho} \,
        v^A_\rho
    \ =\ \sum_\nu \frac{\partial}{\partial x^A_\nu} \bigg|_p v_\nu^A
\end{align}

\subsection{Coordinate bases as local bundle trivializations}
\label{apx:correspondences_bundle_trivializations}

The chart transition map induced transformation laws in Section~\ref{apx:chart_transition_induced_gauge_trafos} coincide the gauge transformations as formulated in Section~\ref{sec:21_main} when identifying the Jacobians
$\frac{\partial x^B}{\partial x^A} \big|_{x^A(p)}$ with $g_p^{BA}$.
In Section~\ref{apx:correspondences_chart_gauge_ptwise} we make these connections precise by listing all correspondences.
Section~\ref{apx:correspondences_chart_gauge_local} extends these results by deriving expressions for chart induced bundle trivializations on extended domains~$U \subseteq M$ as introduced in Section~\ref{sec:bundles_fields}.
A dictionary which summarizes the correspondences is given in Table~\ref{tab:coord_charts_gauge_trafos}.

\subsubsection[Correspondences to pointwise trivializations of \texorpdfstring{$   \TpM$}{TpM}]%
              {Correspondences to pointwise trivializations of \texorpdfstring{$\bm\TpM$}{TpM}}
\label{apx:correspondences_chart_gauge_ptwise}

\paragraph{Gauges and chart differentials:}
The bundle formalism relies on the definition of gauges (Eq.~\eqref{eq:gauge_definition})
\begin{align}
    \psiTMp^A: \TpM \to \R^d \,,
\end{align}
which are vector bundle isomorphisms, assigning coordinates to tangent spaces with $p\in U^A$.
In the chart formalism, gauges over $U^A$ are \emph{induced} as chart differentials (Eq.~\eqref{eq:chart_differential_via_gradients}):
\begin{align}
    \hat{d}x^A_p: \TpM \to \R^d
\end{align}
Different gauges are related by gauge transformations (Eq.~\eqref{eq:gauge_trafo_local_def_21})
\begin{align}
    \psiTMp^B\ =\ g^{BA}_p\, \psiTMp^A \,\ \qquad
    \qquad &\textup{with} \qquad
    g^{BA}_p\ :=\ \psiTMp^B \circ \big(\psiTMp^A\big)^{-1} \ \ \in\ G \,.
\intertext{
The same definition holds for the chart induced gauges, where gauge transformations turn out to coincide with the Jacobian of the chart transition maps (Eq.~\eqref{eq:chart_differential_trafo_law}):
}
    \hat{d}x^B_p\ =\ \frac{\partial x^B}{\partial x^A} \bigg|_{x^A(p)} \hat{d}x^A_p
    \qquad &\textup{with} \qquad
    \frac{\partial x^B}{\partial x^A} \bigg|_{x^A(p)}
    \!=\ \hat{d}x^B_p \circ \big( \hat{d}x^A_p \big)^{-1}
    \ \  \in\ \GL{d}
\end{align}

\paragraph{Vector components:}
As vector components $v^A = \psiTMp^A(v)$ or $v^A = \hat{d}x^A|_p(v)$ are given by the action of gauges, they show the same covariant transformation behavior
\begin{align}
    v^B = g_p^{BA} v^A
    \qquad &\textup{and} \qquad
    v^B = \frac{\partial x^B}{\partial x^A} \bigg|_{x^A(p)} v^A \,.
\intertext{In terms of components, these relations are written as}
    v^B_i = \sum_{j=1}^d \big(g_p^{BA}\big)_{ij}\, v^A_j
    \qquad &\textup{and} \qquad
    v^B_\mu = \sum_{\nu=1}^d \frac{\partial x^B_\mu}{\partial x^A_\nu} \bigg|_{x^A(p)} v^A_\nu \,.
\end{align}

\paragraph{Induced reference frames:}

Reference frames are in the bundle formalism induced by mapping the vectors~$\epsilon_i$ of the standard frame $e\in G$ of $\R^d$ through the gauge map back to $\TpM$ (Eq.~\ref{eq:framefield_gauge_equivalence}):
\begin{align}
    \big[ e_i^A \big]_{i=1}^d\ =\ \Big[ \big(\psiTMp^A \big)^{-1} (\epsilon_i) \Big]_{i=1}^d
\end{align}
The corresponding relation in the chart formalism is according to Eq.~\eqref{eq:coord_basis_vector_via_chart_differential} given by 
\begin{align}
    \bigg[ \frac{\partial}{\partial x^A_\mu} \bigg|_p \bigg]_{\mu=1}^d\ =\ \Big[ \big(\hat{d}x_p^A \big)^{-1} (\epsilon_\mu) \Big]_{\mu=1}^d
\end{align}
Eq.~\eqref{eq:frame_rightaction} shows that the transformation laws of reference frames is given by the right action
\begin{align}\label{eq:trafo_law_comparison_basis_gauge}
    \left[e_{i}^B\right]_{i=1}^d
    \  =\ \left[ e_{i}^A \right]_{i=1}^d \!\lhd \left(g_p^{BA}\right)^{-1}
    \ :=\ \left[ \sum\nolimits_{j=1}^d e_{j}^A\, \big(g_p^{BA}\big)^{-1}_{ji} \right]_{i=1}^d
    \ =\ \left[ \sum\nolimits_{j=1}^d e_{j}^A\, \big(g_p^{AB}\big)_{ji} \right]_{i=1}^d \,.
\end{align}
In analogy, the transformation law of coordinate bases is from Eq.~\eqref{eq:coord_bases_trafo_law_with_inv} seen to be given by
\begin{align}\label{eq:trafo_law_comparison_basis_chart}
    \bigg[\frac{\partial}{\partial x^B_\mu} \bigg|_p \bigg]_{\mu=1}^d
    =\ \bigg[\frac{\partial}{\partial x^A_\mu} \bigg|_p \bigg]_{\mu=1}^d \mkern-6mu\lhd \frac{\partial x^B}{\partial x^A} \bigg|_{x^{\mkern-1muA}\mkern-1mu(p)}^{\;-1}
    &=\ \Bigg[ \sum_{\nu=1}^d \,
            \frac{\partial        }{\partial x^A_\nu} \bigg|_p \ 
            \bigg( \frac{\partial x^B}{\partial x^A} \bigg|_{x^{\mkern-1muA}\mkern-1mu(p)} \bigg)^{-1}_{\nu\mu}
          \Bigg]_{\mu=1}^d 
    \notag \\
    &=\ \Bigg[ \sum_{\nu=1}^d \,
            \frac{\partial        }{\partial x^A_\nu} \bigg|_p \ 
            \frac{\partial x^A_\nu}{\partial x^B_\mu} \bigg|_{x^{\mkern-1muB}\mkern-1mu(p)}
          \Bigg]_{\mu=1}^d 
\end{align}

\subsubsection[Chart induced local trivializations of \texorpdfstring{$    \pi_{\TM}^{-1}(U) $}{TU}]%
              {Chart induced local trivializations of \texorpdfstring{$\bm{\pi_{\TM}^{-1}(U)}$}{TU}}
\label{apx:correspondences_chart_gauge_local}

The correspondences laid out in the last section were relating \emph{pointwise} trivializations $\psiTMp$ of $\TpM$ to chart differentials $\hat{d}x_p$.
In order to complete this picture, this section adds expressions for local trivializations
\begin{align}
    \PsiTM: \piTM^{-1}(U) \to U \times \R^d
\end{align}
which are induced by charts.

A good candidate to construct $\PsiTM$ from is the chart differential
\begin{align}
    dx: \piTM^{-1}(U) \to TV
\end{align}
which is a vector bundle isomorphism that differs from the vector space isomorphisms $dx_p$ by not being restricted to a single point $p \in U$.
To proceed, we generalize the canonical isomorphism $\iota_{\R^d}$ in Eq.~\eqref{eq:canonical_iso_TRk_Rk} from a single point to all the tangent spaces $T_xV \cong \R^d$ over $V \subseteq \R^d$, resulting in the following \emph{canonical local trivialization} of $TV$:
\begin{align}
    \iota_{TV}: TV \to V\times\R^d,\ \ v \mapsto \big( \piTV(v),\, \iota_{\R^d}(v) \big) \,.
\end{align}
This allows to generalize $\hat{d}x_p$ from a single point to a map
\begin{align}
    \hat{d}x\, :=\, \iota_{V\times\R^d} \circ dx \,:\ \piTM^{-1}(U) \to V \times \R^d \,,
\end{align}
which is, however, still not the local trivialization sought for.
By mapping the first factor via the inverse chart from $V$ to $U$, we obtain the \emph{chart induced local bundle trivialization}:
\begin{align}
    \PsiTM\ :=\ \big(x^{-1} \times \id \big) \circ \hat{d}x
\end{align}
As usual, we visualize the definitions made in a commutative diagram:
\begin{equation}\label{cd:coordinate_basis_bundle_trivialization}
    \begin{tikzcd}[row sep=3.5em, column sep=6em]
        V \times \R^d
            \arrow[rrr, rounded corners, to path={ 
                    ([xshift=-1ex]\tikztostart.north)
                    |- node[above, pos=.75]{\small$\big( x^{-1} \times \id \big)$} ([yshift=10ex]\tikztotarget.north)
                    -- (\tikztotarget.north)
                    }]
            \arrow[dr, "\proj_1"']
        &
        TV  \arrow[d, "\piTV"]
            \arrow[l, "\iota_{V \times \R^d}"']
        &
        \piTM^{-1}(U)
            \arrow[d, "\piTM"']
            \arrow[r, "\PsiTM"]
            \arrow[l, "dx"']
            \arrow[ll, rounded corners, to path={ 
                    |- node[above, pos=.75]{\small$\hat{d}x$} ([yshift=4ex, xshift=1ex]\tikztotarget.north)
                    -- ([xshift=1ex]\tikztotarget.north)
                    }]
        &
        U \times \R^d
            \arrow[ld, "\proj_1"]
        \\
        &
        V
        &
        U
            \arrow[l, "x"]
    \end{tikzcd}
\end{equation}

Considering two overlapping charts $x^A: U^A \to V^A$ and $x^B: U^B \to V^B$ and denoting $U^{AB} = U^A \cap U^B$, one obtains transition maps
\begin{align}
    \hat{d}x^B \circ \big( \hat{d}x^A \big)^{-1} \,=\,
    \left( x^B \mkern-5mu\circ\mkern-3mu (x^A)^{-1} \times \frac{\partial x^B}{\partial x^A} \right)
    \,:\ x^A\big( U^{AB}\big) \times \R^d \to x^A\big( U^{AB}\big) \times \R^d
\end{align}
and
\begin{align}
    \PsiTM^B \circ \big( \PsiTM^A \big)^{-1} \,=\,
    \left( \id \times \frac{\partial x^B}{\partial x^A} \right)
    \,:\ U^{AB} \times \R^d \to U^{AB} \times \R^d \,.
\end{align}
These definitions and their mutual relation is shown in the following commutative diagram:
\begin{equation}\label{cd:coordinate_basis_bundle_trivialization_transition}
    \begin{tikzcd}[row sep=3.5em, column sep=4em]
        x^B\big(U^{AB}\big) \times \R^d
            \arrow[rr, "\big( (x^B)^{-1} \times \id \,\big)\ "]
        & &
        U^{AB} \times \R^d
        \\
        &
        \piTM^{-1}\big(U^{AB}\big)
            \arrow[ul, "\hat{d}x^B"]
            \arrow[dl, "\hat{d}x^A"']
            \arrow[ur, "\PsiTM^B"']
            \arrow[dr, "\PsiTM^A"]
        \\
        x^A\big(U^{AB}\big) \times \R^d
            \arrow[rr, "\big( (x^A)^{-1} \times \id \,\big)"']
            \arrow[uu, "\pig( x^B \mkern-5mu\circ\mkern-3mu (x^A)^{-1} \times \frac{\partial x^B}{\partial x^A} \pig)\ "]
        &&
        U^{AB} \times \R^d
            \arrow[uu, "
                    \hspace*{10pt}
                    $\big(\id \times g^{BA} \big)$
                    \\ \rule{0pt}{16pt}
                    $ = \pig( \id \times \frac{\partial x^B}{\partial x^A} \pig)$
                    "' align = left]
    \end{tikzcd}
\end{equation}

\subsection{\textit{G}-structures and vielbein fields}
\label{apx:vielbein_fields}

As discussed in Sections~\ref{sec:G_associated_bundles} and~\ref{sec:bundle_trivializations}, any $G$-atlas $\{(\PsiTM^X, U^X)\}$ of local tangent bundle trivializations specifies a corresponding $G$-structure, that is, a subbundle~$\GM$ of distinguished reference frames which respect (or define) some geometric structure on~$M$.
By definition, the transition maps $g^{BA}$ of associated $G$-bundles take values in a reduced structure group $G\leq\GL{d}$.
This raises the question whether one can similarly find ``$G$-atlases of charts'' $\{(x^X, U^X)\}$, whose Jacobians $\frac{\partial x^B}{\partial x^A}$ take values in a reduced structure group $G \leq \GL{d}$ and therefore encode a $G$-structure.
For some structure groups this is certainly possible; for instance, an orientation of an orientable manifold can always be fixed by specifying some $\operatorname{GL}^+(d)$-atlas of positively oriented charts, whose transition Jacobians take values in $\operatorname{GL}^+(d)$.
In general, it is, however, impossible to find coordinate charts which induce coordinate bases that lie in a given $G$-structure.
One therefore resorts to \emph{explicit gauge transformation from coordinate bases into the $G$-structure}, known as \emph{vielbein fields}~\cite{yepez2011einstein, zhou2016gauge, nakahara2003geometry, carroll2004spacetime}.
After initially transforming from coordinate bases to the $G$-structure, the gauge freedom within the $G$-structure allows for further $G$-valued gauge transformations.

An important example in physics are $\O{d}$-structures (or $\O{1,\, d-1}$-structures for spacetimes), which consist of orthonormal reference frames relative to the (pseudo) Riemannian metric $\eta$ of~$M$.%
\footnote{
    The symbol $\eta$ is in the physics literature commonly preserved for the Minkowski metric $\operatorname{diag}(+1,\, -1,\, \dots,\, -1)$ while the (pseudo) Riemannian metric of~$M$ is denoted by~$g$.
    In contrast, we are writing group elements in the structure group as $g\in G$ and thus use $\eta$ for the (pseudo) Riemannian metric of~$M$.
}
Such orthonormal frames represent the possible laboratory frames of an inertial observer.
They are for instance used to formulate relativistic quantum field theories, specifically the Dirac equation, in curved spacetimes.
Recall that a given $G$-structure is to be respected by local bundle trivializations, which means that the gauge maps $\psiGMp$ need to map the $G$-structure $\GpM$ at~$p\in M$ to the canonical standard $G$-structure $G$ of $\R^d$.
For the specific case of $\O{d}$-structures this is equivalent to the requirement on bundle trivializations to preserve the metric, i.e. $\eta_p(v,w) = \langle \psiTMp(v), \psiTMp(w) \rangle$ for any $p\in M$ and $v,w \in \TpM$, which is accomplished without problems in the bundle formalism.
Given a coordinate chart $x: U \to V$, the induced gauges on $p\in U$ were in the previous sections shown to be given by $\psiTMp = \hat{d}x_p: \TpM \to \R^d$.
The requirement on them to preserve the metric therefore becomes
\begin{align}
    \eta_p(v,w) = \pig\langle \hat{d}x_p(v) \,,\, \hat{d}x_p(w) \pig\rangle \,,
\end{align}
which is exactly the defining property for~$x$ being an \emph{isometry}.
This result implies that \emph{coordinate bases only define an $\O{d}$-structures if $U$ and $V$ are isometric} -- which is only the case if $M$ is locally flat on $U$.
For any non-flat region of $M$ it is therefore impossible to describe an $\O{d}$-structures via coordinate bases directly.
This incompatibility expresses itself for instance in the fact that the components~$\eta_{\mu\nu}$ of the Riemannian metric on~$M$ relative to the chosen coordinate basis differ from $\delta_{\mu\nu}$ (or $\operatorname{diag}(+1,-1,\dots,-1)_{\mu\nu}$).

As mentioned before, the orthonormal frames of an $\O{d}$-structure $\OM$ are in the physics literature typically defined via a gauge transformation relative to some chart induced frame field $\big[ \frac{\partial}{\partial x_\mu} \big]_{\mu=1}^d$.
Denoting this gauge transformation, which is called \emph{vielbein field}, by
\begin{align}
    \mathfrak{e}^A: U \to \GL{d} \,,
\end{align}
the orthonormal frame field is defined by%
\footnote{
    In the physics literature this relation is expressed as
    $e^A_i = (\mathfrak{e}^A)^{\mu}_{\,\ i} \frac{\partial}{\partial x^\mu}$
    The inverse is here merely signaled by the opposite position of the indices
    $(\mathfrak{e}^A)^\mu_{\,\ i} := (\mathfrak{e}^A)^{-1}_{\mu i}$ 
    in comparison to
    $(\mathfrak{e}^A)_\mu^{\,\ i} := \mathfrak{e}^A_{\mu i}$.
}
\begin{align}
    \big[e^A_i\big]_{i=1}^d
    \ :=\ \bigg[ \frac{\partial}{\partial x_\mu} \bigg]_{i=1}^d \lhd \big(\mathfrak{e}^A\big)^{-1}
    \  =\ \bigg[ \sum_\mu \frac{\partial}{\partial x_\mu} \big(\mathfrak{e}^A\big)^{-1}_{\mu i} \bigg]_{i=1}^d
    \ \ \in\ \Gamma(U, \OM) \,.
\end{align}
The orthonormality of the resulting frame field is usually expressed as%
\footnote{
    In the physics literature this relation is usually written
    $\eta_{\mu\nu}\ (\mathfrak{e}^A)^\mu_{\,\ i}\ (\mathfrak{e}^A)^\nu_{\,\ j}\, =\, \delta_{ij}$.
}
\begin{align}
    \delta_{ij}
    \ &=\ \eta\big( e_i^A,\, e_j^A \big) \notag \\
    \ &=\ \eta\bigg( \sum_\mu \frac{\partial}{\partial x_\mu} \big(\mathfrak{e}^A\big)^{-1}_{\mu i}  \,,\; \sum_\nu \frac{\partial}{\partial x_\nu} \big(\mathfrak{e}^A\big)^{-1}_{\nu j} \bigg) \notag \\
    \ &=\ \sum_{\mu\nu} \eta\bigg( \frac{\partial}{\partial x_\mu} \,,\, \frac{\partial}{\partial x_\nu} \bigg)\, \big(\mathfrak{e}^A\big)^{-1}_{\mu i}\, \big(\mathfrak{e}^A\big)^{-1}_{\nu j} \notag \\
    \ &=\ \sum_{\mu\nu} \eta_{\mu\nu}\, \big(\mathfrak{e}^A\big)^{-1}_{\mu i}\, \big(\mathfrak{e}^A\big)^{-1}_{\nu j} \,,
\end{align}
which explains why the vielbein field is sometimes called ``square root of the metric''.
As usual, vector components are translated via the non-inverted gauge transformation, that is:%
\footnote{
    Again, in the usual notation in physics this relation reads $(v^A)^i\, =\, (\mathfrak{e}^A)_\mu^{\,\ i} v^\mu$.
}
\begin{align}
    v^A_i\ =\ \sum_\mu \mathfrak{e}^A_{i\mu}\, v_\mu
\end{align}

A simple dimension counting argument illustrates the gauge freedom in the $\O{d}$-structure:%
\footnote{
    In physics, one rather considers local Lorentz transformations $\Lambda \in \O{1,3}$, which describe rotations and boosts of local reference frames.
}
Being an element of the general linear group, a vielbein $\mathfrak{e}^A(p) \in \GL{d}$ has $d^2$ degrees of freedom, while the metric $\eta$, as a symmetric, bilinear form, has $d(d+1)/2$ degrees of freedom.
The missing $d(d-1)/2$ degrees of freedom correspond exactly to gauge transformations by structure group elements $g^{BA} \in \O{d}$.
Alternatively, from the viewpoint of $G$-structures, $\FpM \cong \GL{d}$ has $d^2$ degrees of freedom while $\OpM \cong \O{d}$ has $d(d-1)/2$ degrees of freedom, fixing $d(d+1)/2$ degrees of freedom which correspond to the choice of metric.

All constructions are obviously generalized to arbitrary $G$-structures with $\GL{d}$-valued vielbein fields mapping coordinate bases into $\GM$ and the freedom to apply $G$-valued gauge transformation afterwards.

\begin{landscape}
\begin{table}[h!]
    \vspace*{8ex}
    \centering%
    \scalebox{1.}{%
        \input{chapters/tab_coord_basis_comparison.tex}
    }%
    \vspace*{4ex}%
    \captionsetup{width=.9\linewidth}
    \caption{
        An overview of different types of coordinatizations on manifolds.
        The bundle formalism (3rd column), which is used in this work, directly assigns coordinates to the tangent spaces, while referring to the points~$p$ of the base space~$M$ in a coordinate free fashion.
        In contrast, the chart formalism (4th column) assigns coordinates to local subsets $U^X \subseteq M$ of the manifold.
        Local trivializations of the tangent bundle and bundle transition maps between them are induced as differentials of the charts and their transition maps, the latter usually referred to as Jacobians.
        The second last row gives expressions for the reference frames which are induced as identity sections of local trivializations of $\TM$ (3rd column) or as chart induced coordinate bases (4th column).
        Similarly, the last row compares definitions of $G$-structures -- for instance orthonormal frames -- via an $G$-atlas for $\TM$ (3rd column) and via vielbein fields as gauge transformations relative to coordinate bases (4th column).
        As usual, we abbreviate $U^A \cap U^B$ by $U^{AB}$ and assume $p\in U^{AB}$.
    }
    \label{tab:coord_charts_gauge_trafos}
\end{table}
\end{landscape}

%% file: chapters/tab_coord_basis_comparison.tex

\def\arraystretch{2.75}
\setlength\tabcolsep{2.8ex}
\small
\begin{tabular}{ @{\ \ } l r@{\,}r@{\ }c@{\ }l cc @{\ \ } } 
    \toprule 
    \\[-8.0ex]
    & \multicolumn{4}{c}{isomorphism}
    & bundle formalism
    & chart formalism \\
    \midrule[0.07em] 
    chart
        & $x^A :\ $
        & $U^A$
        & $\xrightarrow{\sim}$
        & $V^A$
        & ---
        & any diffeomorphism
    \\
    transition map
        & $x^B \!\circ\! \big(x^A\big)^{-1} :\ $
        & $x^B \mkern-1mu\big( U^{\mkern-2muA\mkern-2muB} \big)$
        & $\xrightarrow{\sim}$
        & $x^A \mkern-1mu\big( U^{\mkern-2muA\mkern-2muB} \big)$
        & ---
        & implied by charts
    \\
    \midrule[0.04em] 
    pointwise trivialization \hspace*{-3ex}
        & $\psiTMp^A :\ $
        & $\TpM$
        & $\xrightarrow{\sim}$
        & $\R^d$
        & linear isomorphism from $G$-atlas
        & $\hat{d}x^A_p = \big( \hat{d}x^A_1|_p,\, \dots,\, \hat{d}x^A_d|_p\, \big)^{\!\top}$
    \\
    transition map
        & $\psiTMp^B \mkern-2mu\circ\! \big(\psiTMp^A\big)^{\mkern-2mu-1} \!:\ $
        & $\R^d$
        & $\xrightarrow{\sim}$
        & $\R^d$
        & structure group element $g_p^{BA} \in G$
        & $\hat{d}x^B_p \circ \big(\hat{d}x^A_p\big)^{-1}
           =\, \displaystyle \frac{\partial x^B}{\partial x^A} \bigg|_{\mkern-1mu x^{\mkern-1mu A}\mkern-2mu(p)} $
    \\
    \midrule[0.04em] 
    local trivialization
        & $\PsiTM^A :\ $
        & $\piTM^{-1} \big(U^A\big)$
        & $\xrightarrow{\sim}$
        & $U^A \times \R^d$
        & $v \mapsto \big( \piTM\mkern-1mu(v),\ \psiTMpiv(v) \big)$
        & $\big((x^A)^{-1} \times \id\big) \circ \hat{d}x^A$
    \\
    transition map
        & $\PsiTM^B \mkern-2mu\circ\! \big(\PsiTM^A\big)^{\mkern-2mu-1} :\ $
        & $U^{\mkern-2muA\mkern-2muB} \mkern-4mu\times\! \R^d$
        & $\xrightarrow{\sim}$
        & $U^{\mkern-2muA\mkern-2muB} \mkern-4mu\times\! \R^d$
        & $\big(\id \times g^{BA} \big)$
        & $\displaystyle \bigg( \id \times \frac{\partial x^B}{\partial x^A} \bigg)$
    \\
    \midrule[0.07em] 
    general frame
        & \multicolumn{4}{c}{$\big[ e^A_i \big]_{i=1}^d \in \FpM$}
        & $\Big[ \big(\psiTMp^A\big)^{-1} (\epsilon_i) \Big]_{i=1}^d$\ from $\GL{d}$-atlas
        & $\displaystyle \bigg[\frac{\partial}{\partial x^A_\mu} \bigg|_p \,\bigg]_{\mu=1}^d = \Big[ \big(\hat{d}x^A_p \big)^{-1} (\epsilon_i) \Big]_{\mu=1}^d$
    \\
    $G$-structure frame
        & \multicolumn{4}{c}{$\big[ e^A_i \big]_{i=1}^d \in \GpM$}
        & $\Big[ \big(\psiTMp^A\big)^{-1} (\epsilon_i) \Big]_{i=1}^d$\ from $G$-atlas \kern16pt
        & $\displaystyle \bigg[\sum\nolimits_{\mu}\, \frac{\partial}{\partial x_\mu} \bigg|_p\, \big( \mathfrak{e}^A \big)^{-1}_{\!\mu i}\, \bigg]_{i=1}^d$
    \\
    \bottomrule
\end{tabular}

%% file: chapters/apx02_kernel_figures.tex

\section{Coordinate independent weight sharing and \textit{G}-steerable kernels}
\label{apx:coord_indep_weight_sharing}

\begin{figure}
    \centering
    \includegraphics[width=1.\columnwidth]{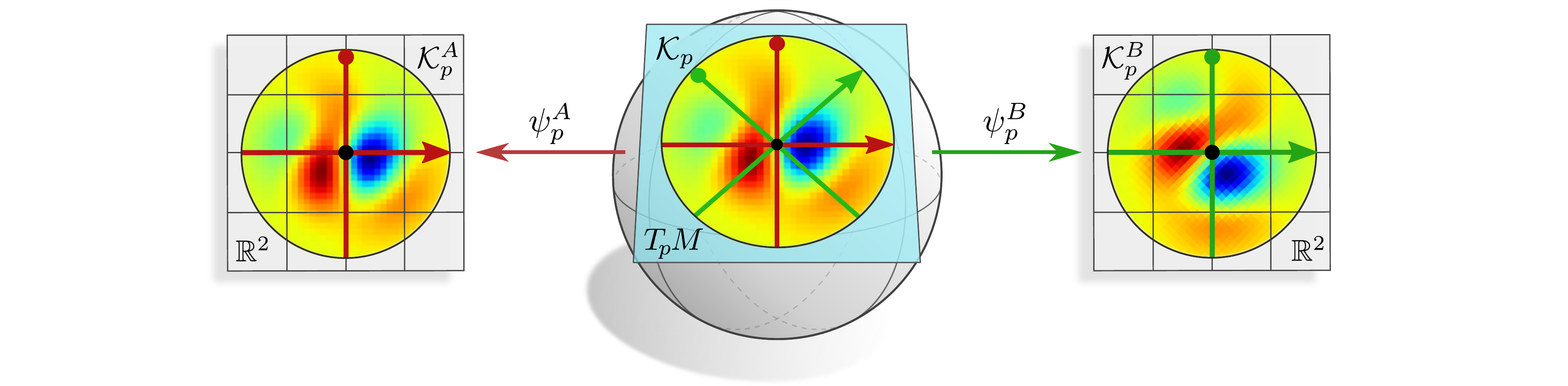}
    \vspace*{-3.5ex}
    \caption{\small
        A \emph{given} coordinate free kernel $\Kp$ on the tangent space $\TpM$ may be represented in arbitrary gauges $\psi_p^A$ or $\psi_p^B$.
        Its coordinate expressions $\Kp^A$ and $\Kp^B$ on $\R^d$ differ in general from each other.
        $G$-steerable kernels have the property to take exactly the same form in all gauges, that is, they satisfy $\Kp^A = \Kp^B = K$ (not visualized).
        }
    \label{fig:kernel_apx_coordinatization}
\end{figure}

\begin{figure}
    \centering
    \includegraphics[width=1.\columnwidth]{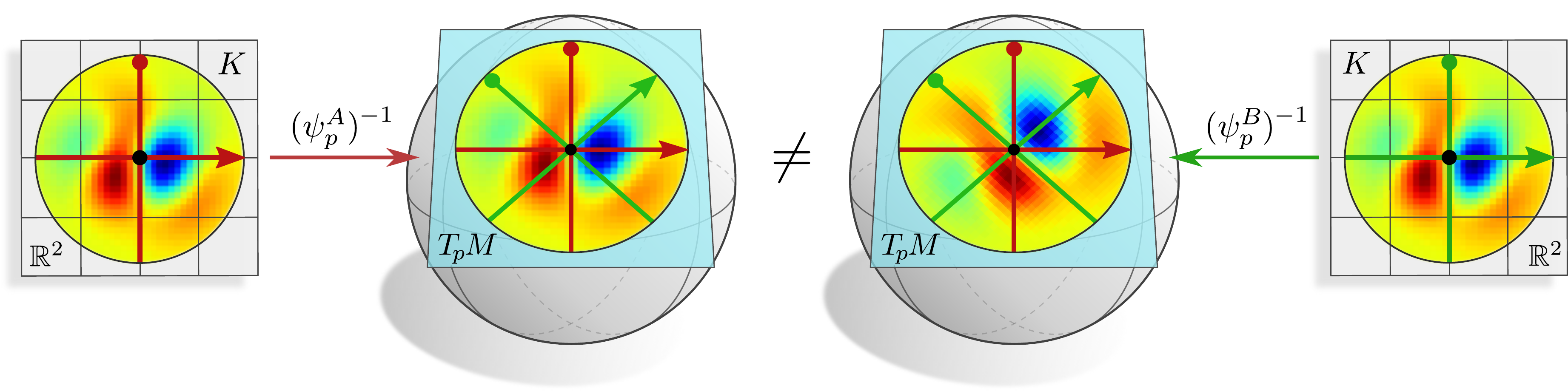}
    \vspace*{-3.5ex}
    \caption{\small
        A coordinate free kernel may be \emph{defined by sharing} a given kernel $K$ on $\R^d$ relative to some reference frame.
        Different choices of frames result in a different coordinate free kernel.
        $G$-steerable kernels have the property to produce exactly the same coordinate free kernel, independent from the chosen reference frame along which they are shared (not visualized).
        This allows for a coordinate independent weight sharing.
        }
    \label{fig:kernel_apx_sharing}
\end{figure}

A fundamental assumption in the design of $\GM$-convolutions is that kernels $K$ on $\R^d$ are shared relative to some choice of reference frame as visualized in Fig.~\ref{fig:kernel_apx_sharing}.
For general kernels, different choices of frames will lead to different alignments of the resulting coordinate free kernel on the tangent space~$\TpM$
-- the weight sharing process is therefore not coordinate independent.
Fig.~\ref{fig:kernel_apx_coordinatization} shows a different situation:
here we assume a coordinate free kernel $\Kp$ that is already given on~$\TpM$ and express it in different gauges on~$\R^d$.
The coordinate representations $\Kp^A$ and $\Kp^B$ do in general not agree with each other but the construction is coordinate independent.

$G$-steerable kernels are constrained exactly such that they guarantee the coordinate independence of the weight sharing process.
Sharing them relative to different frames results in the same coordinate independent kernel on the tangent space,
that is, there will be no difference between the two kernels in the middle of Fig.~\ref{fig:kernel_apx_sharing}.
Equivalently, the resulting coordinate free kernel on $\TpM$ will take the same form $\Kp^A = \Kp^B = K$ when being expressed in different gauges,
that is, the left and the right kernel in Fig.~\ref{fig:kernel_apx_coordinatization} would agree.

Note that this does not necessarily require the kernel to be invariant in the sense that $K(g\mathscr{v}) = K(\mathscr{v})$ for any $g\in G$ and any~$\mathscr{v}\in\R^d$, as the visual intuition might suggest.
This is indeed a special case for kernels that map between scalar fields, i.e. for which both $\rhoin$ and $\rhoout$ are trivial representations (see e.g. Fig.~\ref{fig:intro_steerable_kernel} for $G=\Flip$ or Fig.~\ref{fig:zonal_kernel} for $G=\O2$).
For more general field types the kernels need to be gauge equivariant, i.e. need to satisfy the $G$-steerability constraint
$K(g\mkern1.5mu \mathscr{v}) = \detg^{-1} \rhoout(g)\: K(\mathscr{v})\: \rhoin(g)^{-1}$
which allows for a steering of the ${\cout\times\cin}$ kernel channels (not visualized).
Of course, $G$-steerable kernels can be interpreted as being gauge \emph{invariant} in the sense that
$K(\mathscr{v}) = \detg \rhoout(g)^{-1}\: K(g\mkern1.5mu \mathscr{v})\: \rhoin(g)$
for any $g\in G$ and any~$\mathscr{v}\in\R^d$.
This notion of gauge invariance allows the coordinate independent sharing of $G$-steerable kernels.

A more detailed discussion of coordinate free kernels and their coordinate expressions is found in Section~\ref{sec:kernel_field_trafos}.
The $G$-steerability constraint is derived in Section~\ref{sec:gauge_conv}.

%% file: chapters/apx03_tangent_integral.tex

\section{Integration over tangent spaces}
\label{apx:tangent_integral}

On a Riemannian \emph{manifold} $(M,\eta)$ the volume density%
\footnote{
    In contrast to a volume \emph{form} $\omega$, volume \emph{densities} $|\omega|$ assign a positive volume to any frame.
    They exist both on oriented and non-oriented manifolds.
}
$dp$ on $M$ is uniquely specified by demanding that \emph{orthonormal frames} $\big[ e_1^O, \,\dots,\, e_d^O \big]$ with respect to the metric $\eta$ are assigned \emph{unit volume}:
\begin{align}
    dp\big(e_1^O, \,\dots,\, e_d^O \big) \ =\ 1
    \mkern36mu &\textup{for any \emph{orthonormal} frame
    $\ \big[ e_1^O, \,\dots,\, e_d^O \big]\ $ of $\ \TpM$}
\intertext{
Similarly, a volume density $dv$ on the \emph{tangent spaces} $\TpM$ of a Riemannian manifold is uniquely defined by assigning unit volume to its orthonormal frames w.r.t. $\eta_p$:
}
    dv\big(\mathfrak{e}_1^O, \,\dots,\, \mathfrak{e}_d^O \big) \ =\ 1
    \mkern36mu &\textup{for any \emph{orthonormal} frame
    $\ \big[ \mathfrak{e}_1^O, \,\dots,\, \mathfrak{e}_d^O \big]\ $ of $\ \TvTpM$}
\end{align}

To avoid an unnecessarily complicated discussion of the double tangent bundle $\TTM$, we define the integration over $\TpM$ equivalently by pulling it via some \emph{isometric} (and thus volume preserving) gauge back to~$\R^d$.
Let $\psiTMp^O$ be such an isometric gauge from an $\O{d}$-atlas, which identifies orthonormal frames in $\TpM$ with orthonormal frames in $\R^d$.
The integral of a function $f: \TpM \to \R$ is then defined via its pullback
\begin{align}
    \int_{\TpM} f(v)\, dv
    \ :=&\ \int_{\R^d} f \mkern-2mu\circ\mkern-2mu \big(\psiTMp^O \big)^{\!-1} (v^O)\,\ dv^O \notag \\
    \ =&\ \int_{\R^d} f^O(v^O)\, dv^O \,,
\end{align}
where we defined the coordinate expression $f^O := f \circ \big(\psiTMp^O \big)^{-1} : \R^d \to \R$ of $f$ as usual.
The fact that $\psiTMp^O$ is isometric ensures hereby that $dv$ does indeed assign unit volume to orthonormal frames if $dv^O$ does.
Since the latter is just the standard Lebesgue measure on $\R^d$, this is the case.

Let now $\psiTMp^A$ be \emph{any} gauge at $p$, relative to which one might want to express the integration.
The transition map between both coordinatizations is simply given by the gauge transformation
$v^O = \psi^O \circ (\psi^A )^{-1} (v^A) = g^{OA}_p (v^A)$.
By the standard rules for changes of variables in multidimensional integrals, the differentials are required to transform according to the Jacobian determinant of this transformation in order for the volume to be preserved.
As the transformation is liner, the Jacobian is given by $g_p^{OA}$ itself, such that we obtain
\begin{align}\label{eq:integral_gOA}
    \int_{\TpM} f(v)\, dv
    \ &=\ \int_{\R^d} f^A(v^A)\; \pig|\mkern-2mu \det \!\big(g_p^{OA} \big)\mkern-1mu\pig|\; dv^A \,.
\end{align}

Through the gauge transformation, this expression still depends on the arbitrary choice of isometric gauge~$\psiTMp^O$.
This dependency can be purged by expressing the integration measure directly in terms of the metric~as
\begin{align}\label{eq:integral_etaA}
    \int_{\TpM} f(v)\, dv
    \ &=\ \int_{\R^d} f^A(v^A)\; \sqrt{|\eta_p^A|}\ dv^A \,,
\end{align}
where the factor
\begin{align}\label{eq:volume_element_def}
    \sqrt{|\eta_p^A|}\ :=\ \sqrt{\mkern2mu \pig|\det\!\pig( \big[ \eta_p(e_i^A, e_j^A) \big]_{ij} \pig)\pig| \,}
\end{align}
measures the (absolute) volume of the reference frame $[e_i^A]_{i=1}^d$ relative to the metric $\eta$.
To assert the equality of the right-hand sides of Eqs.~\eqref{eq:integral_gOA} and~\eqref{eq:integral_etaA}, we express the metric $\eta_p$ of $\TpM$ in terms of the standard inner product $\langle\cdot,\, \cdot\rangle$ of~$\R^d$, which is once again done by using the isometric gauge $\psiTMp^O$ from the $\O{d}$-atlas:
\begin{align}
    \eta_p\big( e_i^A,\, e_j^A \big)
    \ &=\ \pig\langle \psiTMp^O\big( e_i^A\big),\; \psiTMp^O\big( e_j^A\big) \pig\rangle \notag \\
    \ &=\ \pig\langle \psiTMp^O \circ \big(\psiTMp^A \big)^{-1} (\epsilon_i),\; \psiTMp^O \circ \big(\psiTMp^A \big)^{-1} (\epsilon_j) \pig\rangle \notag \\
    \ &=\ \pig\langle g_p^{OA} \epsilon_i,\; g_p^{OA} \epsilon_j \pig\rangle \notag \\
    \ &=\ \epsilon_i^\top \big(g_p^{OA} \big)^\top \, g_p^{OA} \epsilon_j \notag \\
    \ &=\ \Big( \big(g_p^{OA} \big)^\top \, g_p^{OA} \Big)_{ij}
\end{align}
The absolute value of the determinant in Eq.~\eqref{eq:volume_element_def} is therefore given by
\begin{align}
    \pig|\det\!\pig( \big[ \eta_p(e_i^A, e_j^A) \big]_{ij} \pig)\pig|
    \ &=\ \pig| \det\pig( \big(g_p^{OA} \big)^\top \, g_p^{OA} \pig) \pig| \notag \\
    \ &=\ \pig| \det\pig( \big(g_p^{OA} \big)^\top \pig) \, \det\pig( g_p^{OA} \pig) \pig| \notag \\
    \ &=\ \pig| \det\big( g_p^{OA} \big) \big|^2 \,,
\end{align}
from which the equality of the right-hand-sides of Eqs.~\eqref{eq:integral_gOA} and~\eqref{eq:integral_etaA} follows by taking the square root.

Since the factors $\sqrt{|\eta_p^A|}$ and $\sqrt{|\eta_p^B|}$ measure the volumes of their respective frames, one can easily show that they are related by the \emph{inverse} change of volume $\big|\mkern-2mu \det g_p^{BA} \big|$:
\begin{alignat}{3}
    \sqrt{|\eta_p^B|}\ &=\ \frac{1}{\big|\mkern-2mu \det g_p^{BA} \big|}\, \sqrt{|\eta_p^A|}
    && \qquad\quad \big(\Rightarrow \quad & -1 & \textup{-density} \,\big)
\intertext{
Together with the usual change of variables formula
}
    dv^B\ &=\ \big|\mkern-2mu \det g_p^{BA} \big|\ dv^A
    && \qquad\quad \big(\Rightarrow \quad  & +1 & \textup{-density} \,\big) \,,
\intertext{
this implies that the coordinatizations of the Riemannian volume element $dv$ are by design invariant under gauge transformations, that is,
}
    \qquad\qquad
    \sqrt{|\eta_p^B|}\ dv^B\ &=\ \sqrt{|\eta_p^A|}\ dv^A
    && \qquad\quad \big(\Rightarrow \quad  & 0 & \textup{-density} \,\big) \,.
\end{alignat}
This relation assures that the integration in Eq.~\eqref{eq:integral_etaA} is well defined, i.e. coordinate independent.

%% file: chapters/apx04_homogeneous_conv.tex

\section{Equivariant convolutions on homogeneous spaces}
\label{apx:homogeneous_conv}

The works by \citet{Kondor2018-GENERAL}, \citet{Cohen2018-intertwiners}\cite{Cohen2019-generaltheory} and \citet{bekkers2020bspline} are in spirit quite similar to ours in that they are defining group equivariant convolutions in a fairly general setting.
These papers have in common that they operate on \emph{feature maps on homogeneous spaces}~$\I/H$ of a \emph{global symmetry group}~$\I$, where $H\leq\I$.%
\footnote{
    We use $\I$ here to denote arbitrary global symmetries, not necessarily isometries.
}%
\footnote{
    \cite{Kondor2018-GENERAL,Cohen2019-generaltheory,bekkers2020bspline} use $G$ instead of $\I$ to refer to global symmetries.
    We use $\I$ since we reserve $G$ for the structure group.
}
They differ in the types of groups $\I$ which they cover and in the definition of their feature spaces, specifically the linear group actions on them.
The main theorems of the papers assert that \emph{the most general equivariant linear maps between such feature spaces are convolutions} (or correlations) with \emph{symmetry constrained kernels}.
The specific details on these generalized convolutions depend on the particular feature spaces and group actions which the models consider.

This appendix examines these theories and their relation to our coordinate independent convolutions.
The most important similarities and differences are summarized in the following list:
\begin{itemize}
    \item[{\rule[2.2pt]{2pt}{2pt}}]
        Not any \emph{homogeneous space} is a \emph{Riemannian manifold} and not any Riemannian manifold is a homogeneous space of its isometry group.%
        \footnote{
            For instance, $\I/H = \O2/\SO2 \cong \{\pm1\}$ is a set but not a Riemannian manifold.
            Another example are $(\Z^d,+)$ group convolutions on the discrete pixel grid~$\Z^d$.
        }
        There is, however, a significant overlap, for instance for Euclidean CNNs on $\Euc_d \cong \E{d}/\O{d}$ or spherical CNNs on $S^d \cong \O{d+1}/\O{d}$.
    \item[{\rule[2.2pt]{2pt}{2pt}}]
        The authors consider \emph{compact}~\cite{Kondor2018-GENERAL}, \emph{locally compact, unimodular}~\cite{Cohen2018-intertwiners}\cite{Cohen2019-generaltheory}, and \emph{Lie} groups~\cite{bekkers2020bspline}, respectively.
        The global symmetry groups in our theory are \emph{isometries} of~$M$ or, specifically for Euclidean spaces, \emph{affine} groups~$\Aff(G)$.
        Note that affine groups are not compact and only for $G\leq\O{d}$ unimodular -- general affine groups are therefore not covered in the respective theories.
    \item[{\rule[2.2pt]{2pt}{2pt}}]
        Coordinate independent CNNs shift the focus from \emph{global} to \emph{local symmetries}.
        On homogeneous spaces $\I/H$ these local symmetries correspond to the stabilizer subgroups $\Stab{p} \cong H$ of~$\I$.
        Our Section~\ref{sec:isometry_intro} works out the relations between global and local symmetries in detail --
        the models' local equivariance induces their global equivariance.
    \item[{\rule[2.2pt]{2pt}{2pt}}]
        The models assume different \emph{types} of feature fields and \emph{group actions} on them:
        \citet{Kondor2018-GENERAL} and \citet{bekkers2020bspline} assume scalar fields on homogeneous spaces, i.e. real-valued functions ${f: \I/H \to \R}$ which transform according to
        $\phi.f (\zeta.H) = f\big( \phi^{-1} \zeta.H \big)$.%
        \footnote{
            Multi-channel feature maps are constructed by stacking multiple such functions.
            In contrast to the case of feature fields, the individual channels of such feature maps transform independently from each other.
        }
        \citet{Cohen2018-intertwiners}\cite{Cohen2019-generaltheory} consider feature fields of more general types $\rho$
        which are defined as sections of $H$-associated feature vector bundles.
        Their transformation laws are given by induced representations $\Ind_H^\I \rho$.
        This setting covers the real-valued functions from \cite{Kondor2018-GENERAL,bekkers2020bspline} as a special case when choosing trivial field representations
        (or, as made precise below, more general quotient representations $\rho_\textup{quot}^{G/H}$ where $H\leq G\leq\I$).
        Our theory models feature fields as sections of associated bundles as well.
        Their transformation is given by pushforwards $\phi \rhd f := \dphiA \circ f \circ \phi^{-1}$, which generalize induced representations.
    \item[{\rule[2.2pt]{2pt}{2pt}}]
        The works by \citet{Kondor2018-GENERAL}, \citet{Cohen2018-intertwiners}\cite{Cohen2019-generaltheory} and \citet{bekkers2020bspline} derive \emph{convolutional weight sharing} from the requirement on the models to be globally equivariant.
        Our $\GM$-convolutions, on the other hand, share weights by definition over the $G$-structure.
        We adopted the idea of deriving weight sharing from global symmetries (isometries) in Section~\ref{sec:quotient_kernel_fields}.
        The requirement for isometry equivariance implies weight sharing over the isometry orbits and a stabilizer constraint on the kernels; see e.g. Fig.~\ref{fig:isom_invariant_kernel_field_quotient}.
        Theorem~\ref{thm:GM_conv_homogeneous_equivalence} asserts that \emph{isometry equivariant kernel field transforms on homogeneous spaces are $\GM$-convolutions} -- this result mirrors those of \citet{Kondor2018-GENERAL}, \citet{Cohen2018-intertwiners}\cite{Cohen2019-generaltheory} and \citet{bekkers2020bspline}
        closely.
    \item[{\rule[2.2pt]{2pt}{2pt}}]
        All of the theories derive some \emph{linear symmetry constraint on the kernel spaces}.
        In the case of \citet{Kondor2018-GENERAL} and \citet{bekkers2020bspline}, the kernels are essentially scalar functions on double quotient spaces $\Hout\backslash \I/\Hin$ (assuming correlations, for convolutions $\Hin$ and $\Hout$ are swapped; see below).
        The kernels of \citet{Cohen2018-intertwiners}\cite{Cohen2019-generaltheory} and in our theory are satisfying a steerability constraint which depends on the particular choice of field types $\rhoin$ and~$\rhoout$.
        Note that the determinant factor is missing in the $G$-steerability constraint of \citet{Cohen2018-intertwiners}\cite{Cohen2019-generaltheory} since the authors restrict to unimodular groups.
        The factor does appear in the kernel constraint by \citet{bekkers2020bspline}.
    \item[{\rule[2.2pt]{2pt}{2pt}}]
        While \citet{Kondor2018-GENERAL} and \citet{Cohen2018-intertwiners}\cite{Cohen2019-generaltheory} describe kernels immediately on the group or homogeneous space,
        \citet{bekkers2020bspline} and our $\GM$-convolutions define kernels on the tangent spaces and project them subsequently via the exponential map.
        These approaches are in general inequivalent, for instance since the exponential map is on a non-connected manifold non-injective.
        On Euclidean spaces both approaches are obviously equivalent since the exponential map becomes trivial; see Section~\ref{sec:euclidean_affine_equiv}.
        Our Theorem~\ref{thm:spherical_kernel_space_iso} in Section~\ref{sec:spherical_CNNs_fully_equivariant} bridges this gap furthermore for spherical kernels by providing an isomorphism between kernels of the two approaches.
        In practice, the general incompatibility is irrelevant since kernels of convolutional networks are usually compactly supported within the injectivity radius of the exponential map.
\end{itemize}

We will in the following elaborate on the theories of
\citet{Kondor2018-GENERAL}, \citet{bekkers2020bspline} and \citet{Cohen2018-intertwiners}\cite{Cohen2019-generaltheory}
in more detail.
As a preparation, we will first discuss homogeneous spaces, group convolutions and group correlations.
For an alternative review of the topic we refer the reader to~\citet{esteves2020theoretical}.
We furthermore want to point to the work by \citet{chakraborty2018H-CNNs}, which also defines convolutions on homogeneous spaces.
It is not covered in more detail in this section since their models assume $\Hout=\{e\}$, that is, their convolution kernels are unconstrained and always lift the input signal to a scalar field on~$\I$.

\toclesslab\subsection{General remarks on homogeneous spaces, group convolutions and group correlations}{apx:homogeneous_preliminaries}

\paragraph{Homogeneous spaces:}
Let $\I$ be some group which acts on some space~$X$.
The space is said to be \emph{homogeneous} if the group action is \emph{transitive}, i.e. if any two points $p,q\in X$ are related by the $\I$-action.
In equations, $X$ is homogeneous if and only if for any $p,q\in X$ there exists an element $\phi\in\I$ such that~$q=\phi(p)$.
Note that the action on~$X$ is not required to be fixed point free, that is, each point $p\in X$ has a potentially non-trivial stabilizer subgroup $\Stab{p} = {\{\xi\in\I \,|\, \xi(p)=p \}} \leq\I$.
It can be shown that the homogeneous space can be identified with the quotient space $\I/H$ where $H = \Stab{p}$ for some $p\in X$.%
\footnote{
    Other choices of points yield other realizations of the non-canonical isomorphism $\I/H \cong X$.
    Any choice is equally valid since $\Stab{p} \cong \Stab{q}$ for homogeneous spaces.
}

Since any homogeneous space arises as a quotient, we consider in the following always some subgroup $H$ of~$\I$.
This subgroup has \emph{left cosets}, i.e. subsets of the form
\begin{align}
    \phi.H\ =\ \big\{ \phi h \,\big|\, h\in H \big\}
\end{align}
which are elements of the (homogeneous) quotient space
\begin{align}
    \I/H\ =\ \big\{ \phi.H \,\big|\, \phi\in \I \big\} \,.
\end{align}
A natural left action of $\I$ on $\I/H$ is given by
\begin{align}
    \I \times \I/H \to \I/H,\ \ \ \big(\widetilde{\phi},\, \phi.H\big) \mapsto \widetilde{\phi}\phi.H \,.
\end{align}
This action is easily seen to be transitive, making $\I/H$ a homogeneous space of~$\I$.
The canonical quotient map
\begin{align}
    \mathscr{q}^{\I}_{\I/H}:\ \I\to \I/H,\ \ \phi \mapsto \phi.H
\end{align}
turns $\I$ into a principal $H$-bundle over~$\I/H$.
Analogous definitions can be made for \emph{right cosets}
\begin{align}
    H.\phi\ \ &\in\ \ H\backslash\I \,.
\intertext{and \emph{double cosets}}
    \widetilde{H}.\phi.H\ \ &\in\ \ \widetilde{H} \backslash\I/H
\end{align}
and their respective quotient spaces.

An universal property of the quotient maps $\mathscr{q}^{\I}_{\I/H}$, which will become important in our discussion below, is the following.
Let $f^\uparrow: \I \to \R$ be a continuous, right $H$-invariant function, i.e. a function which satisfies $f^\uparrow(\phi h) = f^\uparrow(\phi)$ for any $\phi\in\I$ and~$h\in H$.
Then there exists a unique continuous function $f: \I/H \to \R$ such that $f^\uparrow = f \circ \mathscr{q}^{\I}_{\I/H}$.
Conversely, one may lift any continuous map $f: \I/H \to \R$ uniquely to a right $H$-invariant map $f^\uparrow: \I \to \R$, which is used by \citet{Kondor2018-GENERAL} to generalize group convolutions to homogeneous spaces.
The relation between both functions is visualized in the following commutative diagram:
\begin{equation}
\label{cd:left_cosets_lift}
\begin{tikzcd}[column sep=55pt, row sep=35pt, font=\normalsize]
    \I
        \arrow[rd, "\!f^\uparrow := f\circ \mathscr{q}^{\I}_{\overset{}{\I/H}}"]
        \arrow[d, "\mathscr{q}^{\I}_{\overset{}{\I/H}}\,"']
    \\
    \I/H
        \arrow[r, "f"']
    & \R
\end{tikzcd}
\end{equation}
An analogous construction can obviously be made for right quotient spaces $H\backslash\I$ and left $H$-invariant maps.
The following commutative diagram visualizes the case of double quotient spaces $\widetilde{H} \backslash \I/H$ and maps $f^\uparrow$ which are simultaneously left $\widetilde{H}$-invariant and right $H$-invariant, i.e. which satisfy ${f^\uparrow\big(\widetilde{h}\phi h\big) = f^\uparrow(\phi)}$ for any ${\phi\in\I}$, \ ${\widetilde{h}\in\widetilde{H}}$, and ${h\in H}$:
\begin{equation}
\label{cd:double_cosets_lift}
\begin{tikzcd}[column sep=55pt, row sep=35pt, font=\normalsize]
    \I
        \arrow[rd, "\!f^\uparrow := f\circ \mathscr{q}^{\I}_{\overset{}{\widetilde{H} \backslash \I/H}}"]
        \arrow[d, "\mathscr{q}^{\I}_{\overset{}{\widetilde{H} \backslash \I/H}}\,"']
    \\
    \widetilde{H} \backslash \I/H
        \arrow[r, "f"']
    & \R
\end{tikzcd}
\end{equation}

\paragraph{Group convolutions and group correlations:}
Convolutions are naturally generalized from Euclidean spaces (or translation groups) to arbitrary locally compact groups.
Let $\I$ be a locally compact group and let $d\zeta$ be a left Haar measure on~$\I$.
The \emph{group convolution} $(f \ast_{\overset{}{\protect\scalebox{.64}{$\mkern-.5mu \I$}}} \kappa): \I \to \R$ of two integrable functions $f:\I\to\R$ and $\kappa:\I\to\R$ is then defined by the following equivalent expressions, taken from \cite{gallier2019harmonicRepr}:
\begin{align}\label{eq:group_conv_def}
    \big(f \ast_{\overset{}{\protect\scalebox{.64}{$\mkern-.5mu \I$}}} \kappa \big)(\phi)
    \ :&=\ \int_{\I} f(\zeta)\, \kappa\big( \zeta^{-1} \phi \big) \; d\zeta \notag \\
    \ &=\ \int_{\I} f(\phi\, \zeta)\, \kappa\big( \zeta^{-1}\big) \; d\zeta \notag \\
    \ &=\ \int_{\I} f\big(\zeta^{-1} \big)\, \kappa( \zeta\phi)\: \Delta(\zeta^{-1}) \; d\zeta \notag \\
    \ &=\ \int_{\I} f\big(\phi\, \zeta^{-1} \big)\, \kappa( \zeta)\: \Delta(\zeta^{-1}) \; d\zeta \,,
\end{align}
The group homomorphism $\Delta: \I \to (\R^+_{>0},*)$, appearing in the last two expressions, is the modular function of~$\I$.
\citet{Kondor2018-GENERAL} define group convolutions as in the last line, however, without the modular function.
This is valid since the authors assume compact groups, which are unimodular, i.e. satisfy $\Delta(\phi) = 1$ for any $\phi \in\I$.

Closely related to group convolutions are \emph{group correlations}
\begin{align}\label{eq:group_corr_def}
    \big(f \star_{\overset{}{\protect\scalebox{.64}{$\mkern-.5mu \I$}}} \kappa \big)(\phi)
    \ :=\ \big\langle f,\, \phi.\kappa \big\rangle_{L^1(\I)}
    \ =\ \int_{\I} f(\zeta)\, \kappa\big( \phi^{-1} \zeta \big) \; d\zeta\ ,
\end{align}
which are defined as the inner product of a function~$f$ with a shifted kernel~$\phi.\kappa$.
A comparison with Eq.~\eqref{eq:group_conv_def} reveals that group convolutions and group correlations are equivalent up to an inversion of the kernel argument, that is,
\begin{align}\label{eq:group_conv_corr_kernel_inversion}
    \big(f \star_{\overset{}{\protect\scalebox{.64}{$\mkern-.5mu \I$}}} \kappa \big)
    \ =\ \big(f \ast_{\overset{}{\protect\scalebox{.64}{$\mkern-.5mu \I$}}} \big[\kappa \circ (\,\cdot\,)^{-1} \big] \big) \,.
\end{align}
While \citet{Kondor2018-GENERAL} consider (generalized) group convolutions, \citet{bekkers2020bspline} and \citet{Cohen2019-generaltheory} assume correlations -- to reconcile the theories one has to invert the kernel arguments.

Group convolutions and group correlations are by definition \emph{equivariant} w.r.t. left actions $\alpha.f(\phi) = f\big( \alpha^{-1}\phi \big)$ of group elements $\alpha\in\I$ on the first factor.
For the case of convolutions, this is shown by
\begin{align}\label{eq:group_conv_equivariance}
    \big( [\alpha.f] \ast_{\overset{}{\protect\scalebox{.64}{$\mkern-.5mu \I$}}} \kappa \big)(\phi)
    \ &=\ \int_{\I} \big[\alpha.f \big](\zeta)\, \kappa\big( \zeta^{-1} \phi \big) \; d\zeta \notag \\
    \ &=\ \int_{\I} f\big( \alpha^{-1} \zeta\big)\, \kappa\big( \zeta^{-1} \phi \big) \; d\zeta \notag \\
    \ &=\ \int_{\I} f\big(\widetilde{\zeta} \big)\, \kappa\big( \widetilde{\zeta}^{-1} \alpha^{-1} \phi \big) \; d\big( \alpha \widetilde{\zeta} \big) \notag \\
    \ &=\ \big( f \ast_{\overset{}{\protect\scalebox{.64}{$\mkern-.5mu \I$}}} \kappa \big) (\alpha^{-1} \phi) \notag \\
    \ &=\ \big[\alpha.( f \ast_{\overset{}{\protect\scalebox{.64}{$\mkern-.5mu \I$}}} \kappa )\big] (\phi) \,,
\end{align}
where we substituted $\widetilde{\zeta} = \alpha^{-1}\zeta$ in the third step and made use of the fact that $d\widetilde{\zeta}$ is a left Haar measure, i.e. satisfies $d\big(\alpha \widetilde{\zeta}\big) = d\widetilde{\zeta}$.
The case of correlations follows trivially by Eq.~\eqref{eq:group_conv_corr_kernel_inversion}.

The majority of equivariant CNNs rely on group convolutions or group correlations.
In particular, the models in rows~(1-3), (5), (11), (15), (19), (21), (24), (25) and (32) of Table~\ref{tab:network_instantiations}, all of which are (or could equivalently be) labeled by regular representations, are group convolutional CNNs.
Prior to their use in equivariant CNNs, group convolutions have been widely applied in robotics~\cite{chirikjian1998numerical} or for image analysis~\cite{mallat2012group,sifre2012combined,Sifre2013-GSCAT,bruna2013invariant,sifre2014rigid,oyallon2015scattering}.
\citet{Cohen2016-GCNN} showed that group convolutions (or rather correlations) naturally generalize conventional CNNs.
Since the feature maps of convolutional networks comprise multiple channels, they are not given by real-valued functions on $\I$ but by vector-valued functions $f: \I \to \R^c$.
Kernels are accordingly defined to be (unconstrained) matrix-valued functions on the group, i.e. $\kappa: \I \to \R^{\cout\times\cin}$.
The works of \citet{Kondor2018-GENERAL}, \citet{bekkers2020bspline} and \citet{Cohen2018-intertwiners}\cite{Cohen2019-generaltheory}, which we review in the following, generalize such group convolutional networks to arbitrary homogeneous spaces.

\toclesslab\subsection{Scalar field convolutions on homogeneous spaces}{apx:homogeneous_scalar_field_convs}

We start with the $\I$-equivariant convolutional (or correlational) networks on homogeneous spaces by \citet{Kondor2018-GENERAL} and \citet{bekkers2020bspline}.
Both theories define feature maps as \emph{scalar fields on homogeneous spaces}, that is, each channel is given by a real-valued function
\begin{align}\label{eq:scalar_field_homogeneous_space}
    f: \I/H \to \R \,.
\end{align}
Individual channels transform independently under the action of the global symmetry group $\I$ as specified by
\begin{align}\label{eq:group_action_homogeneous_space_scalar_field}
    \big[\mkern1mu \widetilde{\phi}.f\big] (\phi.H)\ :=\ f\big( \widetilde{\phi}^{-1} \phi.H \big)
    \qquad\ \widetilde{\phi}\in H,\ \ \ \phi.H\in \I/H \,.
\end{align}

Each layer $l = 1,\dots, L$ may be assigned a different subgroup $H_l \leq\I$ and thus homogeneous space $\I/H_l$ on which its feature maps live.
This allows for instance to model lifting convolutions from the sphere $S^2 \cong \SO3/\SO2$ to the group $\SO3 \cong \SO{3}/\{e\}$ when choosing subgroups $\SO2$ and $\{e\}$, respectively.
The choices of subgroups correspond in some sense to the choices of group representations in our theory, which we will explain further below.

The results of the two papers are to large parts equivalent, however,
\citet{Kondor2018-GENERAL} consider compact groups $\I$ and convolutions
while \citet{bekkers2020bspline} assume $\I$ to be a Lie group and use correlations.

\paragraph{\citet{Kondor2018-GENERAL} :}

In a nutshell, \citet{Kondor2018-GENERAL} investigate the most general $\I$-equivariant linear maps between scalar field features on homogeneous spaces $\I/\Hin$ and $\I/\Hout$, assuming the transformation law in Eq.~\eqref{eq:group_action_homogeneous_space_scalar_field}.
They prove that this operation is given by a generalized group convolution with a kernel
\begin{align}
    \kappa: \Hin\backslash \I/\Hout \to \R
\end{align}
on the double quotient space specified by $\Hin$ and $\Hout$.
Formulated for finite groups, as done by the authors, this generalized convolution operation is shown to be given by
\begin{align}\label{eq:quotient_space_conv_Kondor}
    \big(f \ast_{\overset{}{\scalebox{.64}{$\I/\Hin$}}} \kappa \big) (\phi.\Hout)
    \ \ :=\ \ |\Hin| \sum_{\Hin.\zeta \,\in\, \Hin\mkern-2.5mu\backslash\mkern-.5mu\I}
        f\big( \phi\mkern2mu \zeta^{-1}\!. \Hin\big)\ \kappa\big( \Hin.\zeta.\Hout \big) \,.
\end{align}
A comparison with the last line of Eq.~\eqref{eq:group_conv_def} suggests that this operation is indeed closely related to group convolutions -- the modular function $\Delta$ drops out since $\I$ is compact and therefore unimodular.
The generalized convolution is in fact equivalent to a group convolution
\begin{align}
    \big(f \ast_{\overset{}{\scalebox{.64}{$\I/\Hin$}}} \kappa \big) (\phi.\Hout)
    \ =\ \big(f^\uparrow \ast_{\overset{}{\protect\scalebox{.64}{$\mkern-.5mu \I$}}} \kappa^\uparrow \big)(\phi)
\end{align}
with features and kernels that are lifted according to the diagrams in Eqs.~\eqref{cd:left_cosets_lift} and~\eqref{cd:double_cosets_lift}.
Note that the convolution kernel on $\Hin\backslash \I/\Hout$ corresponds to a correlation kernel on $\Hout\backslash \I/\Hin$ since convolutions and correlations are according to Eq.~\eqref{eq:group_conv_corr_kernel_inversion} related by an inversion of the kernel argument.
One could therefore view the kernels by \citet{Kondor2018-GENERAL} as left $\Hout$-invariant correlation kernels on the input space~$\I/\Hin$.

To give an intuition on these results, we come back to our spherical CNN example from above.
Let therefore $\I=\SO3$, $\Hin=\SO2$ and, for now, $\Hout=\{e\}$.
This setting describes lifting convolutions from the 2-sphere $\I/\Hin = \SO3/\SO2 \cong S^2$ to the rotation group manifold $\I/\Hout = \SO3/\{e\} \cong \SO3$.
Considering correlations instead of convolutions, the kernels are real-valued functions on $\Hout\backslash \I/\Hin = \{e\}\backslash \SO3/\SO2 \cong S^2$.
If we let instead $\Hout=\SO2$, the convolution maps from scalar fields on the 2-sphere to scalar fields on the 2-sphere $\I/\Hout = \SO3/\SO2$.
In this case the correlation kernels are given by real-valued functions on $\SO2\backslash \SO3/\SO2$.
Equivalently, the correlation kernels are given by left $\SO2$-invariant functions on~$S^2$, i.e. zonal kernels as visualized in Fig.~\ref{fig:zonal_kernel}.
When assuming $\Hin=\Hout=\{e\}$, one has $\I/\Hin = \I/\Hout \cong \SO3$ and unconstrained kernels on $\Hout\backslash \I/\Hin \cong \SO3$, corresponding to conventional group convolutions (or correlations).
These results are in line with our discussion in Section~\ref{sec:spherical_CNNs_fully_equivariant}.

For completeness, we mention that \citet{Kondor2018-GENERAL} explain their results additionally from a representation theoretic perspective, i.e. with features and kernels in Fourier space.
The fact that features and kernels live on quotient spaces is in this formulation reflected in sparsity patterns of the Fourier coefficients.

\paragraph{\citet{bekkers2020bspline} :}

Instead of considering compact groups, \citet{bekkers2020bspline} assumes $\I$ to be a general Lie group.
The feature maps of layer~$l$ are defined as real-valued square integrable functions in $L^2(\I/H_l)$
which transform according to Eq.~\eqref{eq:group_action_homogeneous_space_scalar_field} when being acted on by~$\I$.

\citet{bekkers2020bspline} models the layers of his convolutional (or rather correlational) networks as linear bounded operators
\begin{align}
    \mathfrak{K}:\ L^2(\I/\Hin) \to L^2(\I/\Hout)
\end{align}
between feature maps on homogeneous spaces $\I/\Hin$ and~$\I/\Hout$.
Such operators are in general given by integral operators of the form
\begin{align}
    \big[\mathfrak{K}f\big] (\phi.\Hout)\ =\ 
    \int_{\I/\Hin} \widehat{\kappa} \big(\phi.\Hout,\, \zeta.\Hin \big)\ f(\zeta.\Hin)\,\ \dmuIHin \,,
\end{align}
where $\dmuIHin$ is some Radon measure on $\I/\Hin$ and
\begin{align}
    \widehat{\kappa}:\ \I/\Hout \times \I/\Hin \to \R
\end{align}
is an integrable 2-argument kernel.

The requirement on the operator to be equivariant, that is,
\begin{align}
    \mathfrak{K}\big( \phi.f \big)\ =\ \phi \mkern1mu.\mkern1.5mu \mathfrak{K}(f)
    \qquad \forall\,\ \phi\in\I,\ \ f\in L^2(\I/\Hin) \ ,
\end{align}
is shown to imply that the 2-argument kernel reduces to a single argument kernel
\begin{align}
    \widehat{\kappa} \big(\phi.\Hout,\, \zeta.\Hin \big)
    \ =\ \frac{\dmuIHin(\phi^{-1} \zeta.\Hin)}{\dmuIHin(\zeta.\Hin)}\ \kappa\big(\phi^{-1} \zeta.\Hin \big) \,.
\end{align}
The group element $\phi\in \phi.\Hout \subset \I$ is hereby an arbitrary representative of the coset in which it is contained.
This 1-argument kernel is -- up to a measure dependent scale factor -- constrained to be left $\Hout$-invariant:
\begin{align}
    \kappa(\zeta.\Hin)\ =\ \frac{\dmuIHin(\xi^{-1} \zeta.\Hin)}{\dmuIHin(\zeta.\Hin)}\ \kappa\big(\xi^{-1} \zeta.\Hin \big)
    \qquad \forall\ \ \zeta.\Hin \in \I/\Hin,\ \ \xi\in\Hout
\end{align}
Note that this result is very similar to that of \citet{Kondor2018-GENERAL} since a left $\Hout$-invariant kernel on $\I/\Hin$ is equivalent to an element of $\Hout\backslash \I/\Hin$ (again assuming correlation kernels instead of convolution kernels).
The main difference is the additional scale factor, which appears since the Radon measure $\dmuIHin$ is not necessarily left $\I$-invariant.

One of the practically relevant cases is that of group correlations, for which $\Hin = \{e\}$ and $\I/\{e\} = \I$.
In this case $d\mu_{\overset{}{\I}}$ is a left (invariant) Haar measure on $\I$, such that the scale factor drops out.
A second relevant case is that of affine equivariant convolutions on Euclidean spaces, i.e. the choices $\I = \Aff(G)$ and $\Hin=G$, for which $\I/\Hin \cong \R^d$.
Assuming $\dmuIHin$ to be the Lebesgue measure on $\R^d$ and denoting $\phi=tg \in \I$, \citet{bekkers2020bspline} prove that the scale factor is in this case given by:
\begin{align}
    \frac{\dmuIHin\!( (tg)^{-1}x )}{\dmuIHin\!(x)}\ =\ \frac{1}{\detg} \qquad \forall\ x\in\R^d
\end{align}
This is exactly the determinant factor which appears in our $G$-steerability kernel constraint, Eq.~\eqref{eq:kernel_constraint}, as well.

Since $\SO3$ is a Lie group, the spherical CNN examples that we gave after discussing the theory by \citet{Kondor2018-GENERAL} apply without changes (assuming the standard left-invariant measure on $S^2$).

\citet{bekkers2020bspline} defines kernels in close analogy to our $\GM$-convolutions on the tangent spaces and projects them via exponential maps to the homogeneous spaces.
The kernels on the tangent spaces are hereby modeled via B-splines.
A difference is that \citet{bekkers2020bspline} does not need to consider parallel transporters since he is assuming scalar feature maps on the homogeneous spaces.

\paragraph{Relation to \emph{GM}-convolutions:}

Due to the quite different formulation it is not immediately obvious how the results of \citet{Kondor2018-GENERAL} and \citet{bekkers2020bspline} relate to our theory.
Instead of considering different quotient spaces $\I/H_l$ in each layer~$l$, we consider a fixed manifold~$M$.
To see how both approaches connect, assume another subgroup~$G$ to be given such that $H_l \leq G \leq \I$ for all layers~$l=1,\dots,L$ and satisfying that $M := \I/G$ is a manifold.
The scalar features on $\I/H_l$ can in this case be viewed as $G$-associated feature fields on $M$ which transform according to \emph{quotient representations}~$\rho_\textup{quot}^{G/H_l}$.
To see this, note that the group action in Eq.~\eqref{eq:group_action_homogeneous_space_scalar_field} is nothing but the induced representation $\Ind_{H_l}^{\I} \rho_\textup{triv}^{H_l} = \rho_\textup{quot}^{\I/H_l}$ from the trivial representation of $H_l$, which describes the transformation law of scalar fields on~$\I/H_l$.
This representation can via induction in stages (see~\cite{ceccherini2009induced}) be decomposed into
\begin{align}
    \Ind_{H_l}^{\I} \rho_\textup{triv}^{H_l}
    \ =\ \Ind_G^{\I} \Ind_{H_l}^G \rho_\textup{triv}^{H_l}
    \ =\ \Ind_G^{\I} \rho_\textup{quot}^{G/H_l} \,,
\end{align}
that is, into the induction of the quotient representation $\rho_\textup{quot}^{G/H_l}$ from~$G$ to~$\I$.
The real-valued functions on $\I/H_l$ are therefore equivalent to $\rho_\textup{quot}^{G/H_l}$-fields on~$M=\I/G$.

Interesting special cases are $G=H_l$ and $G=\{e\}$.
For the former one has $\rho_\textup{quot}^{G/H_l} = \rho_\textup{triv}^G$, describing scalar fields on~$M = \I/G = \I/H_l$.
For the latter, $\rho_\textup{quot}^{G/H_l} = \rho_\textup{reg}^G$ is the regular representation, corresponding to conventional group convolutions.

These insights imply that the theory of \citet{Kondor2018-GENERAL} explains all models in Table~\ref{tab:network_instantiations} which operate on homogeneous spaces of compact groups $\I$ and are labeled by either trivial, regular or more general quotient representations -- these are essentially the spherical CNNs in rows~(32) and~(33).
A minor generalization of the theory to locally compact, unimodular groups would additionally describe some of the isometry equivariant Euclidean CNNs.
As \citet{bekkers2020bspline} is assuming arbitrary Lie groups, his models additionally describe the $\Aff(G)$-equivariant CNNs in Table~\ref{tab:network_instantiations} which are labeled by trivial, regular or more general quotient representations.
They cover in particular scale equivariant Euclidean CNNs ($G=\Scale$) for which the determinant factor $\detg$ is non-trivial.

Other types of feature fields and non-homogeneous spaces like punctured Euclidean spaces $\Euc_d\backslash\{0\}$ and spheres $S^2\backslash\{n,s\}$, the icosahedron, general surfaces and the M\"obius strip are not covered.

\toclesslab\subsection{Steerable CNNs on homogeneous spaces}{apx:homogeneous_steerable_convs}

Motivated by Kondor and Trivedi's \cite{Kondor2018-GENERAL} generalization of group convolutions to homogeneous spaces, \citet{Cohen2018-intertwiners}\cite{Cohen2019-generaltheory} generalized steerable CNNs to homogeneous spaces of locally compact unimodular groups.%
\footnote{
    Note that there are a preprint version~\cite{Cohen2018-intertwiners} and a conference version~\cite{Cohen2019-generaltheory} of this paper.
}
Instead of restricting to scalar fields, \citet{Cohen2018-intertwiners}\cite{Cohen2019-generaltheory} assume more general \emph{$H_l$-associated feature fields} on $I/H_l$ which transform according to \emph{induced representations}~$\Ind_{H_l}^\I \rho_l$ of~$\I$.
The network layers implement linear equivariant maps between such fields, i.e. they are \emph{intertwiners between induced representations}.
As~expected, these layers are parameterized by -- and are thus isomorphic to -- spaces of steerable kernels.
\citet{Cohen2018-intertwiners}\cite{Cohen2019-generaltheory} show that these kernels can be described on $\I$, on $\I/\Hin$ or on $\Hout\backslash \I/\Hin$, in each case still satisfying a linear steerability constraint.%
\footnote{
    As we will argue below, the constructions on $\I/\Hin$ and $\Hout\backslash \I/\Hin$ depend on \emph{local} sections and are therefore only possible for trivial bundles.
    We adapt the former to nontrivial bundles by defining kernels on an open cover of~$\I/\Hin$.
}
The following three paragraphs will
1) introduce feature fields and their transformation laws on a global and local level,
2) review the spaces of intertwiners and steerable kernels which map between such fields, and
3) discuss how these results relate to ours.

Our formulation and notation in this section is adapted to be more similar to that which was chosen to develop our theory.
It differs therefore slightly from that of \citet{Cohen2018-intertwiners}\cite{Cohen2019-generaltheory}.
Most notably, we do not assume a single local trivialization (section) which is defined almost everywhere on $\I/H_l$ but consider an atlas of local trivializations which cover the homogeneous space.%
\footnote{
    This is only necessary if the homogeneous space is a (non-trivial) manifold.
    If it is discrete, one may always choose a global section $\I/H \to \I$ which selects coset representatives.
    One would in this case usually not talk about ``atlases'' and ``local trivializations'', however, we will do so for simplicity.
}
The notation of local, coordinatized quantities is therefore augmented with gauge labels $A,B,\dots$ \,.

\paragraph{Feature fields and induced representations:}

Let $\I$ be a locally compact unimodular group and let $H_l\leq\I$ be any subgroup of it.
As stated above, the quotient map
\begin{align}
    \mathscr{q}^{\I}_{\overset{}{\I/H_l}}: \I \to \I/H_l, \quad \phi \mapsto \phi.H_l
\end{align}
implies a \emph{principal $H_l$-bundle}; see Section~\ref{sec:fiber_bundles_general}.
The right $H_l$-action on the total space $\I$ is given by the usual right multiplication
\begin{align}
    \I \times \I/H_l \to \I/H_l, \quad (\phi,h) \mapsto \phi h
\end{align}
of group elements.
It preserves the fibers $\I_{\phi.H_l} = \big( \mathscr{q}^{\I}_{\overset{}{\I/H_l}} \big)^{-1} (\phi.H_l) \,\subset\, \I$ since it satisfies
\begin{align}
    \mathscr{q}^{\I}_{\overset{}{\I/H_l}} (\phi h)
    \ =\ \phi h.H_l\ =\ \phi.H_l
    \ =\ \mathscr{q}^{\I}_{\overset{}{\I/H_l}} (\phi)
\end{align}
for any $\phi\in\I$ and $h\in H_l$ and is easily seen to be both transitive and free.
Abbreviating $U := U^A \cap U^B$, local trivializations $\PsiI^A,\ \PsiI^B$ of this bundle and the transition maps $h^{BA}$ between them are defined via the following commutative diagram:
\begin{equation}
\begin{tikzcd}[row sep=4.em, column sep=5.5em]
    &[-6.5em]
    & U\times H_l
    \\
      \I\ \: \supseteq
    & \big(\mathscr{q}^{\I}_{\overset{}{\I/H_l}} \big)^{\mkern-2mu-1}(U)
                    \arrow[d, swap, "\mathscr{q}^{\I}_{\overset{}{\I/H_l}}"]
                    \arrow[r, "\PsiI^A"]
                    \arrow[ru, "\PsiI^B"]
    & U\times H_l   \arrow[u, swap, "(\id\times h^{BA}\cdot)"]
                    \arrow[ld, "\proj_1"]
    \\
      \I/H_l\ \ \supseteq \mkern30mu
    & U
\end{tikzcd}
\qquad
\end{equation}
As usual, the principal bundle trivializations imply local identity sections
\begin{align}
    \sigma^A: U^A \to \big(\mathscr{q}^{\I}_{\overset{}{\I/H_l}} \big)^{\mkern-2mu-1}(U^A) \,, \quad 
    \phi.H_l \,\mapsto\, \sigma^A (\phi.H_l) \,:=\, \big(\PsiI^A \big)^{-1} (\phi.H_l,\, e) \,,
\end{align}
which were introduced in Section~\ref{sec:bundle_trivializations}.
The identity sections labeled by $\widetilde{A}$ at $\zeta.H_l$ and $A$ at $\phi\,\zeta.H_l$ are related by
\begin{align}\label{eq:homogeneous_induce_gauge_trafo}
    \phi\; \sigma^{\widetilde{A}} (\zeta.H_l) \ =\ \sigma^A (\phi\,\zeta.H_l)\,\ h_\phi^{A\widetilde{A}} (\zeta.H_l) \,,
\end{align}
which defines the $\I$-induced gauge transformations $h_\phi^{A\widetilde{A}} (\zeta.H_l) \in H_l$; see Eq.~\eqref{eq:pushfwd_section_right_action}.%
\footnote{
    To avoid confusion, note that \citet{Cohen2018-intertwiners}\cite{Cohen2019-generaltheory} denote $h_\phi^{A\widetilde{A}} (\zeta.H_l)$ by $\operatorname{h} (\zeta.H_l,\, \phi)$, omitting the gauge labels.
}

The feature fields of steerable CNNs on homogeneous spaces are defined as sections $f\in \Gamma(\A_l)$ of associated $H_l$-bundles
\begin{align}\label{eq:equiv_rel_rhol}
    \A_l\ :=\ (\I\times\R^{c_l})/\!\sim_{\!\rho_l} \,,
\end{align}
which were introduced in Section~\ref{sec:G_associated_bundles}.
The equivalence relation
\begin{align}
    (\phi,\, \mathscr{f})\ \sim_{\!\rho_l}\, (\phi h^{-1},\, \rho_l(h) \mathscr{f})
\end{align}
is determined by a choice of field representation
\begin{align}
    \rho_l: H_l \to \R^{c_l}
\end{align}
of the layer's subgroup; compare this to our analogous definition in Eq.~\eqref{eq:equiv_relation_A}.
Being an associated $H_l$-bundle, the local feature vector bundle trivializations transform covariantly with those of the corresponding principal bundle:
\begin{equation}
\begin{tikzcd}[row sep=3.5em, column sep=5.em]
    & U\times \R^{c_l}
    \\
      \piAl^{-1}(U)  \arrow[d, swap, "\piAl"]
                    \arrow[r, "\PsiAl^A"]
                    \arrow[ru, "\PsiAl^B"]
    & U\times \R^ {c_l} \arrow[u, swap, "\big(\id\times \rho_l\big(h^{BA}\big)\cdot\big)"]
                    \arrow[ld, "\proj_1"]
    \\
    U
\end{tikzcd}
\end{equation}
The precise construction of associated bundle trivializations from principal bundle trivializations was given in Eq.~\eqref{eq:trivialization_A}.

\citet{Cohen2018-intertwiners}\cite{Cohen2019-generaltheory} use two different approaches to describe feature fields.
Globally, feature fields are represented as functions
\begin{align}\label{eq:mackey_functions}
    F: \I \to \R^{c_l}
    \quad \textup{such that} \quad
    F(\phi h^{-1}) = \rho_l(h) F(\phi)
    \quad \forall\ \phi\in\I,\ h\in H_l \,,
\end{align}
whose definition is consistent with the equivalence relation from Eq.~\eqref{eq:equiv_rel_rhol}.
On trivializing neighborhoods $U^A \subseteq \I/H_l$, the fields are furthermore given by feature vector coefficient fields
\begin{align}
    f^A: U^A \to \R^c
\end{align}
relative to some gauge $\PsiAl^A$.
While the former is more convenient for algebraic manipulations, the latter is non-redundant, and therefore more suitable for numerical implementations.
The local field representation may at any time be computed from the global one by setting
\begin{align}
    f^A( \phi.H_l)\ =\ F\big( \sigma^A (\phi.H_l) \big)
    \qquad \textup{for}\ \ \phi.H_l \in U^A \,.
\end{align}
Here $\sigma^A: U^A \to \I$ is that local section of the principal $H_l$ bundle which corresponds to the chosen trivialization $\PsiI^A$ (``identity section'') and is analogously defined to Eq.~\eqref{eq:GM_section_psi_inverse_def}.
Note that the global field representation can in general not be recovered from a (single) local one.
It is, however, locally over $U^A$ given by
\begin{align}
    \quad
    F(\phi)\ =\ \rho_l\big( \psiI{\phi.H_l}^A(\phi) \big)^{-1}\, f^A(\phi.H_l)
    \qquad \textup{for}\ \ \phi \in \big(\mathscr{q}^{\I}_{\overset{}{\I/H_l}} \big)^{\mkern-2mu-1} (U^A)\ \subseteq\,\I \,,
\end{align}
which is closely related to Eq.~\eqref{eq:trivialization_A_p}.

The global, active transformations of feature fields are formalized by \emph{induced representations} $\Ind_{H_l}^{\I}\mkern-2mu \rho_l$ of~$\I$, which are conceptually similar to our isometry pushforwards from Def.~\ref{dfn:isometry_pushforward}.
For the global field representations, this action is simply defined as a shift on~$\I$:
\begin{align}
    \pig[ \big[ \Ind_{H_l}^{\I}\mkern-2mu \rho_l\big] (\zeta)\, F \pig] (\phi)\ =\ F\big( \zeta^{-1} \phi\big)
\end{align}
Since the $\I$-action is global, it is more difficult to describe for local field representations.
Let $U^A$ be a trivializing neighborhood around $\phi.H_l$ and $U^{\widetilde{A}}$ around $\zeta^{-1}\phi.H_l$.
The action of the induced representation is relative to gauges on these neighborhoods given by
\begin{align}
    \pig[ \big[\Ind_{H_l}^{\I}\mkern-2mu \rho_l\big] (\zeta)\, f\pig]^{\raisebox{2.5pt}{$\scriptstyle A$}} (\phi.H_l)
    \ =\ \rho_l\big( h^{A\widetilde{A}}_{\zeta} \big) f^{\widetilde{A}} \big( \zeta^{-1} \phi.H_l \big) \,,
\end{align}
where $h^{A\widetilde{A}}_{\zeta}$ is the $\zeta$-induced gauge transformation, which is analogously defined to that in Eq.~\eqref{cd:pushforward_GM_coord_extended}.
Note the similarity of this definition to our isometry pushforward of feature fields in coordinates from Eq.~\eqref{eq:feature_field_trafo_in_coords}.
We furthermore identify the transformation law of scalar fields on homogeneous spaces from Eq.~\eqref{eq:group_action_homogeneous_space_scalar_field} as a special case for trivial representations~$\rho_l$.
Steerable CNNs on homogeneous spaces cover therefore the homogeneous scalar field convolutions of \citet{Kondor2018-GENERAL} and \citet{bekkers2020bspline} as a special case (ignoring the different assumptions made on the type of group $\I$).

\paragraph{Intertwiners between induced representations and steerable kernels:}
The main endeavor of \citet{Cohen2018-intertwiners}\cite{Cohen2019-generaltheory} is to characterize the space
\begin{align}
    \Hom_{\I}\! \big( \Gamma(\Ain), \Gamma(\Aout) \big)
    \ :=\ \big\{ \mathfrak{K}: \Gamma(\Ain) \to \Gamma(\Aout)\,\ \textup{linear} \ \big|\ 
        \mathfrak{K}\circ \Ind_{\Hin}^{\I}(\phi) = \Ind_{\Hout}^{\I}(\phi) \circ\mathfrak{K}\ \ \ \forall\ \phi\in\I \big\}
\end{align}
of intertwiners between induced representations, i.e. the space of linear equivariant maps between feature fields.
Diagrammatically, this space consists of those linear maps $\mathfrak{K}$ which let the following diagram commute for any $\phi\in\I$:
\begin{equation}
\begin{tikzcd}[column sep=60pt, row sep=35, font=\normalsize]
    \Gamma(\Ain)
        \arrow[r, "\mathfrak{K}"]
        \arrow[d, "\Ind_{\Hin}^{\I} \rhoin(\phi)"']
    &
    \Gamma(\Aout)
        \arrow[d, "\Ind_{\Hout}^{\I} \rhoout(\phi)"]
    \\
    \Gamma(\Ain)
        \arrow[r, "\mathfrak{K}"']
    &
    \Gamma(\Aout)
\end{tikzcd}
\end{equation}
These maps are the analog to our isometry equivariant kernel field transforms, which were defined in Def.~\ref{dfn:isometry_equivariance}.
\citet{Cohen2018-intertwiners}\cite{Cohen2019-generaltheory} prove that these maps are given by correlations with steerable kernels.
We will in the following briefly review these results for both global and local field representations.

When working with the global field representation from Eq.~\eqref{eq:mackey_functions}, \citet{Cohen2018-intertwiners}\cite{Cohen2019-generaltheory} start with a general bounded linear operator $\mathfrak{K}$ of the form
\begin{align}\label{eq:general_linear_map_mackey}
    \big[\mathfrak{K} F \big](\phi) \ =\ 
    \int_{\I} \widehat{\kappa} (\phi,\zeta)\, F(\zeta)\ d\zeta
\end{align}
where $d\zeta$ is a left Haar measure on $\I$ and
\begin{align}
    \widehat{\kappa}: \I\times\I \to \R^{\cout\times\cin}
\end{align}
is a matrix-valued 2-argument kernel.
The equivariance constraint is shown to require the kernels to satisfy the relation
$\widehat{\kappa} (\widetilde{\phi}\phi, \widetilde{\phi}\zeta) = \widehat{\kappa} (\phi, \zeta)$
for any choice of group elements $\widetilde{\phi},\, \phi,\, \zeta \in \I$.
This result is resembling our Theorem~\ref{thm:isometry_equivariant_kernel_field_trafos}, which states that isometry equivariant kernel field transforms imply kernel fields which are invariant under the action of isometries, Def.~\ref{dfn:isometry_invariant_kernel_fields}.
Given this constraint, the 2-argument kernel can be replaced by a 1-argument kernel which is defined as
\begin{align}\label{eq:one_arg_kernel_global}
    \kappa: \I \to \R^{\cout\times\cin},\ \ \ \phi \mapsto \kappa(\phi) := \widehat{\kappa}(e,\phi) \,.
\end{align}
We therefore have
$\widehat{\kappa}(\phi,\zeta) = \widehat{\kappa}(\phi^{-1} \phi,\, \phi^{-1}\zeta) = \kappa\big( \phi^{-1} \zeta\big)$,
implying that the linear operator is given by a \emph{group correlation} (Eq.~\eqref{eq:group_corr_def}), that is:
\begin{align}
    \big[\mathfrak{K} F \big](\phi)
    \ =\ \int_{\I} \kappa\big( \phi^{-1}\zeta \big)\, F(\zeta)\ d\zeta
    \ =\ \big(F \star_{\overset{}{\protect\scalebox{.64}{$\mkern-.5mu \I$}}} \kappa \big)(\phi)
\end{align}
The correlation kernel is furthermore required to satisfy the linear $\Hout$-$\Hin$-steerability constraint
\begin{align}\label{eq:double_steerability}
    \kappa(h_\textup{out}\, \phi\, h_\textup{in})
    \ =\ \rhoout(h_\textup{out})\, \kappa(\phi)\, \rhoin(h_\textup{in})
    \qquad \forall\,\ \phi\in\I,\ h_\textup{in}\in\Hin,\ h_\textup{out}\in\Hout \,.
\end{align}
This constraint is reminiscent of that found by \citet{Kondor2018-GENERAL} and \citet{bekkers2020bspline}.
Instead of enforcing kernels to be left $\Hout$- and right $\Hin$-\emph{invariant},
which would correspond to trivial representations $\rhoout=\rho_\textup{triv}^{\Hout}$ and $\rhoin=\rho_\textup{triv}^{\Hin}$,
the constraint of \citet{Cohen2018-intertwiners}\cite{Cohen2019-generaltheory} allows for more general steerable kernels.
The vector space $\mathscr{K}^{\I}_{\rhoin\mkern-1mu,\rhoout}$ of such steerable correlation kernels is argued to be isomorphic to the intertwiner space $\Hom_{\I}\! \big( \Gamma(\Ain), \Gamma(\Aout) \big)$.

Since the global field representations $F$ on $\I$ are redundant they are not the best choice for numerical implementations.
\citet{Cohen2018-intertwiners}\cite{Cohen2019-generaltheory} are therefore additionally investigating intertwiners which operate on local field representations.
The authors approach this problem by assuming one \emph{single local trivialization} to be given, which is \emph{defined almost everywhere} on the homogeneous space~$\I/\Hin$.
They are therefore effectively operating on a trivial bundle.
Our following review adapts their results slightly to the more general case of a set of field representations relative to an \emph{atlas of local trivializations}.
The formulation of \citet{Cohen2018-intertwiners}\cite{Cohen2019-generaltheory} is retrieved by restricting the integration to one single trivialization.
We explicitly write out all gauge labels to make the coordinate dependencies transparent.
To give an overview on the local trivializations that will play a role in the following, we mention that we will need to consider
trivializing neighborhoods $U^A, U^{\widetilde{A}}, U^H \subseteq \I/\Hin$ such that
\begin{align}
    \zeta.\Hin                   \in U^A \,, \qquad
    \widetilde{\phi}\zeta.\Hin   \in U^{\widetilde{A}} \quad \textup{and}\ \ \
    h\zeta.\Hin                  \in U^H
\end{align}
and trivializing neighborhoods $U^P, U^{\widetilde{P}}, U^E \subseteq \I/\Hout$ such that
\begin{align}
    \phi.\Hout                   \in U^P \,, \qquad
    \widetilde{\phi}\phi.\Hout   \in U^{\widetilde{P}} \quad \textup{and}\ \ \
    e.\Hout                  \in U^E \,.
\end{align}
We will furthermore assume any partition of unity
$\{ \mathscr{P}_{U^X} \}_{X\in\mathfrak{X}}$
subordinate to the open cover
underlying the atlas
$\mathscr{A}_\textup{in} = \{( U^X, \Psi^X )\}_{X\in\mathfrak{X}}$
of local trivializations on $\I/\Hin$.
This means that we are given maps $\mathscr{P}_{U^X}: \I/\Hin \to [0,1]$ with the properties
\begin{align}
    \supp\big( \mathscr{P}_{\!\overset{}{U^X}} \big) \subseteq U^X
    \qquad \textup{and} \qquad
    \sum_{U^X\in \mathscr{A}_\textup{in}} \mathscr{P}_{\!\overset{}{U^X}} (\phi.\Hin) = 1
    \quad \forall\ \phi.\Hin \in \I/\Hin \,.
\end{align}

Eq.~\eqref{eq:general_linear_map_mackey} stated the general form of a bounded linear operator between global field representations~$F$.
Its local analog, which makes use of the partition of unity, is given by
\begin{align}
    \big[ \mathfrak{K} f \big]^P (\phi.\Hout)\ =\ 
    \sum_{U^{\!A} \in \mathscr{A}_\textup{in}} \int_{U^{\!A}} \mathscr{P}_{\!\overset{}{U^{\!A}}} (\zeta.\Hin)\ \ 
        \widehat{\overleftarrow{\kappa}} \rule{0pt}{0pt}^{\mkern-1mu PA}(\phi.\Hout,\, \zeta.\Hin)\,\ 
        f^A (\zeta.\Hin)\ \ d(\zeta.\Hin) \,,
\end{align}
where $P$ and $A$ label local trivializations as stated above and $d(\zeta.\Hin)$ is a measure on~$\I/\Hin$.
We furthermore have 2-argument kernels
\begin{align}
    \widehat{\overleftarrow{\kappa}} \rule{0pt}{0pt}^{\mkern-1mu PA}:\, U^P \!\times U^A \to \R^{\cout\times\cin}
    ,\qquad (\phi.\Hout,\, \zeta.\Hin) \,\mapsto\,
    \widehat{\kappa} \big( \sigma^P(\phi.\Hout),\, \sigma^A(\zeta.\Hin) \big)
\end{align}
which are inherently \emph{locally defined} on ${U^P \!\times U^A} \,\subseteq\, \I/\Hout \times \I/\Hin$.
The global 2-argument kernel can be recovered from a set of local kernels on the open covers.
\citet{Cohen2018-intertwiners}\cite{Cohen2019-generaltheory} prove that these local kernels are required to satisfy
\begin{align}\label{eq:intertwiner_constraint_local}
    \widehat{\overleftarrow{\kappa}} \rule{0pt}{0pt}^{\mkern-1mu PA}(\phi.\Hout,\, \zeta.\Hin)
    \ =\ 
    \rhoout\big( h_{\widetilde{\phi}}^{\widetilde{P}P} (\phi.\Hout) \big)^{\!-1}\,\ 
    \widehat{\overleftarrow{\kappa}} \rule{0pt}{0pt}^{\mkern-1mu \widetilde{P}\widetilde{A}}( \widetilde{\phi}\phi.\Hout,\, \widetilde{\phi}\zeta.\Hin)
    \,\ \rhoin\big( h_{\widetilde{\phi}}^{\widetilde{A}A} (\zeta.\Hin) \big)
\end{align}
for any $\widetilde{\phi} \in\I$.
Note that $h_{\widetilde{\phi}}^{\widetilde{P}P} (\phi.\Hout)$ is hereby an induced gauge transformation on $\I/\Hout$ while $h_{\widetilde{\phi}}^{\widetilde{A}A} (\zeta.\Hin)$ is an induced gauge transformation on $\I/\Hin$.
In order to reduce these local 2-argument kernels to local 1-argument kernels, \citet{Cohen2018-intertwiners}\cite{Cohen2019-generaltheory} consider the unique group element $\widetilde{\phi} \in\I$ which satisfies
1) $\widetilde{\phi} \phi.\Hout = e.\Hout$ and
2) $\widetilde{\phi} \sigma^P(\phi.\Hout) = \sigma^E(e.\Hout) = e$,
where the last equality fixes a specific gauge at the ``origin'' $e.\Hout$, which is always possible.
The first point allows us to identify the gauges $\widetilde{P}$ and~$E$ without loss of generality.
The relations imply furthermore $\widetilde{\phi} = \sigma^P (e.\Hout)^{-1}$ and, by Eq.~\eqref{eq:homogeneous_induce_gauge_trafo}, $h_{\widetilde{\phi}}^{EP} (\phi.\Hout) = e$.
Plugging these choices into Eq.~\eqref{eq:intertwiner_constraint_local} yields
\begin{align}
    \widehat{\overleftarrow{\kappa}} \rule{0pt}{0pt}^{\mkern-1mu PA}(\phi.\Hout,\, \zeta.\Hin)
    \ =\ 
    \id_{\R^\cout}^{PE}\ 
    \underbrace{ \vphantom{\Big(}
    \widehat{\overleftarrow{\kappa}} \rule{0pt}{0pt}^{\mkern-1mu E\widetilde{A}}\big( e.\Hout,\, \sigma^P \mkern-2mu (\phi.\Hout)^{-1} \zeta.\Hin\big)}_{ \rule{0pt}{14pt} \displaystyle
    =: \overleftarrow{\kappa}^{E\widetilde{A}} \big( \sigma^P \mkern-2mu (\phi.\Hout)^{-1} \zeta.\Hin \big) }
    \,\ \rhoin\big( h_{\sigma^P \mkern-2mu (\phi.\Hout)^{-1}}^{\widetilde{A}A} (\zeta.\Hin) \big) \,,
\end{align}
where the identity map is kept explicit to explain the gauge labels.
We furthermore introduced the local 1-argument kernels
\begin{align}\label{eq:local_one_arg_kernel}
    \overleftarrow{\kappa}^{E\widetilde{A}} :\, U^{\widetilde{A}} \to \R^{\cout\times\cin} \,,
\end{align}
whose responses are always given in the specific gauge $E$ at $e.\Hout$.
These kernels are still required to satisfy the $\Hout$-steerability constraints
\begin{align}\label{eq:single_steerability}
    \overleftarrow{\kappa}^{EH} (h_\textup{out} \zeta.\Hin)
    \ =\ \rhoout(h_\textup{out})\, \overleftarrow{\kappa} (\zeta.\Hin)^{EA} \,
    \rhoin\big( h_{h_\textup{out}}^{HA} (\zeta.\Hin) \big)^{-1}
    \qquad \forall\,\ \zeta.\Hin\in \I/\Hin,\ h_\textup{out}\in\Hout \,.
\end{align}
Putting everything together, the equivariant correlation becomes
\begin{align}
    & \big[ \mathfrak{K} f \big]^P (\phi.\Hout) \ = \\
    & \id_{\R^\cout}^{PE} \mkern-6mu
    \sum_{U^{\!A} \in \mathscr{A}_\textup{in}} \int_{U^{\!A}} \mathscr{P}_{\!\overset{}{U^{\!A}}} (\zeta.\Hin)\ \ 
        \overleftarrow{\kappa} \rule{0pt}{0pt}^{\mkern-1mu E\widetilde{A}} \pig( \sigma^P \mkern-2mu (\phi.\Hout)^{-1} \zeta.\Hin \pig)\,\ 
        \rhoin\pig( h_{\sigma^P \mkern-2mu (\phi.\Hout)^{-1}}^{\widetilde{A}A} (\zeta.\Hin) \pig)\,\ 
        f^A (\zeta.\Hin)\ \ d(\zeta.\Hin) \,. \notag
\end{align}
Adding the assumption that a single gauge $A = \widetilde{A}$ covers $\I/\Hin$ almost everywhere, we can drop the partition of unity and retrieve the formulation of \citet{Cohen2018-intertwiners}\cite{Cohen2019-generaltheory}:
\begin{align}
    \big[ \mathfrak{K} f \big]^P (\phi.\Hout)
    \ =\ \id_{\R^\cout}^{PE} \int_{U^{\!A}}
        \overleftarrow{\kappa} \rule{0pt}{0pt}^{\mkern-1mu EA} \pig( \sigma^P \mkern-2mu (\phi.\Hout)^{-1} \zeta.\Hin \pig)\,\ 
        \rhoin\pig( h_{\sigma^P \mkern-2mu (\phi.\Hout)^{-1}}^{AA} (\zeta.\Hin) \pig)\,\ 
        f^A (\zeta.\Hin)\ \ d(\zeta.\Hin)
\end{align}
We comment on the relation of this operation to our $\GM$-convolutions further below.

Instead of defining the \emph{local} 1-argument kernels in coordinates from Eq.~\eqref{eq:local_one_arg_kernel} on local subsets $U^{\widetilde{A}}$, \citet{Cohen2018-intertwiners}\cite{Cohen2019-generaltheory} define them \emph{globally} on~$\I/\Hin$.
Since their construction relies on a continuous section, this is only possible if the bundles are trivial.
Our adaptation to local kernel representations on an open covering is bridging this gap.

\citet{Cohen2018-intertwiners}\cite{Cohen2019-generaltheory} claim an isomorphism between the global kernels on $\I$,
satisfying the steerability constraint in Eq.~\eqref{eq:double_steerability},
and their kernels on $\I/\Hin$,
satisfying the steerability constraint in Eq.~\eqref{eq:single_steerability}.
Note that this isomorphism can only hold if
either the bundle is trivial
or the continuity assumption on the sections (and therefore network inference) is dropped.
It should, however, be possible to prove an isomorphism between the global kernel and a collection of local kernels on a covering of $\I/\Hin$, satisfying the relations in Eq.~\eqref{eq:single_steerability}.

The authors furthermore claim that the steerable kernels can be described on the double quotient space $\Hout\backslash \I/\Hin$, still satisfying a steerability constraint.

\paragraph{Relation to \textit{GM}-convolutions:}

The steerable CNNs on homogeneous spaces by \citet{Cohen2018-intertwiners}\cite{Cohen2019-generaltheory} are conceptually quite similar to our $\GM$-convolutions on Riemannian manifolds, however, there are some important differences which we discuss in the following.
Most importantly, the theories differ in
1) being based on different spaces $\I/H_l$ in each layer $l$ vs. assuming a fixed manifold $M$,
2) modeling kernels on the space $\I/\Hin$ itself or on tangent spaces $\TpM$ of it,
3) the way of how weights are shared, and
4) the types of global symmetry group $\I$ and spaces $\I/H_l$ or $M$, which they cover.
Despite these differences, many of the results of \citet{Cohen2018-intertwiners}\cite{Cohen2019-generaltheory} have analogs in our theory.

Both theories share the idea to define feature fields as sections of associated vector bundles.
While \citet{Cohen2018-intertwiners}\cite{Cohen2019-generaltheory} consider a global symmetry group $\I$ as a set of multiple principal $H_l$-bundles over homogeneous spaces $\I/H_l$, we work with some $G$-structure $\GM$ over a fixed Riemannian manifold~$M$.
All of our feature vector bundles are defined as $G$-bundles and are associated to each other, while the feature bundles of \citet{Cohen2018-intertwiners}\cite{Cohen2019-generaltheory} may not be associated to each other if their structure groups $H_l$ do not agree.
As already claimed at the end of the last Section~\ref{apx:homogeneous_scalar_field_convs}, these differences can be mitigated if a structure group $G$ can be chosen such that $H_l \leq G \leq \I$ for every layer $l$ and $M := \I/G$ is a Riemannian manifold.
One can then replace all homogeneous spaces $\I/H_l$ with $M$ and all $H_l$-representations $\rho_l$ with induced $G$-representations
\begin{align}
    \rho_l^G\ :=\ \Ind_{H_l}^G \rho_l \,.
\end{align}
The global field transformation laws are preserved by this reinterpretation since
\begin{align}
    \Ind_{H_l}^{\I} \rho_l
    \ =\ \Ind_G^{\I} \Ind_{H_l}^G \rho_l
    \ =\ \Ind_G^{\I} \rho_l^G
\end{align}
holds by induction in stages~\cite{ceccherini2009induced}.

Another main difference lies in the definition of convolution kernels and weight sharing.
On the global, coordinate free level and prior to the isometry assumption, \citet{Cohen2018-intertwiners}\cite{Cohen2019-generaltheory} start in Eq.~\eqref{eq:general_linear_map_mackey} with a bounded linear operator which is parameterized by an unconstrained kernel
\begin{align}
    \widehat{\kappa}: \I\times\I \to \R^{\cout\times\cin} \,.
\end{align}
This operator corresponds in our theory to a general kernel field transform, Def.~\ref{dfn:kernel_field_trafo}, which is parameterized by an unconstrained kernel field
\begin{align}
    \K: \TM \to \Hom(\Ain,\Aout) \,,
\end{align}
see Def.~\eqref{dfn:kernel_field_general}.
The 2-argument kernels $\widehat{\kappa}$ can be thought of as representing a kernel field as well.
Their two arguments are thereby thought of as addressing
1)~a specific (1-argument) kernel, yielding a response at the corresponding point in the output bundle $\I \to \I/\Hout$ and
2)~the spatial dependency of this 1-argument kernel on the input bundle $\I \to \I/\Hin$.
The analog in our kernel fields $\K$ is that elements $v \in \TM$ encode
1)~the location $p = \piTM(v)$ of the kernel and
2)~its spatial dependency via $v\in\TpM$.

When requiring the bounded linear operator to be $\I$-equivariant, the 2-argument kernel $\widehat{\kappa}$ becomes constrained to satisfy
\begin{align}
    \widehat{\kappa} \big( \widetilde{\phi}\phi, \widetilde{\phi}\zeta \big) = \widehat{\kappa} (\phi, \zeta)
    \qquad \forall\ \widetilde{\phi} \in\I \,.
\end{align}
Isometry equivariant kernel field transforms were in Theorem~\ref{thm:isometry_equivariant_kernel_field_trafos} shown to require the isometry invariance of the kernel field, i.e.
\begin{align}
    \widetilde{\phi}_{\overset{}{\mkern-2mu*\mkern-1mu\scalebox{.55}{$,\mkern-2mu\mathscr{K}\mkern2mu$}}} \K = \K
    \qquad \forall\ \widetilde{\phi} \in\I \,;
\end{align}
see Def.~\ref{dfn:isometry_invariant_kernel_fields} and Fig.~\ref{fig:isom_invariant_kernel_field_multiple_orbits}.

The invariance constraint on 2-argument kernels $\widehat{\kappa}$ allows to replace them with 1-argument kernels
\begin{align}
    \kappa: \I \to \R^{\cout\times\cin},
\end{align}
defined in Eq.~\eqref{eq:one_arg_kernel_global}.
They are still required to satisfy the steerability constraint in Eq.~\eqref{eq:double_steerability}.
Our isometry invariant kernel fields were in Theorem~\ref{thm:manifold_quotient_repr_kernel_fields} shown to be equivalent to a field of kernels
\begin{align}
    \Qhat \mkern-2mu: \piTM^{-1}\big( \rM(\IM) \big) \to \piHom^{-1}\big( \rM(\IM) \big)
\end{align}
whose support is restricted to the tangent spaces over representatives $\rM(\IM) \subseteq M$ of the quotient~$\IM$.%
\footnote{
    Theorem~\ref{thm:tangent_quotient_repr_kernel_fields} proves another isomorphism to a space of kernels $\Q \mkern-2mu: \rTM(\ITM) \to \rHom(\IHom)$ whose support is even further restricted to representatives of the tangent bundle quotient~$\ITM$.
}
These kernels are required to satisfy a stabilizer subgroup steerability constraint as well.
For the specific case that $M$ is a homogeneous space of its isometry group, the quotient $\IM$ reduces to a single element.
Theorem~\ref{thm:manifold_quotient_repr_kernel_fields} implies in this case a single (1-argument) kernel
\begin{align}\label{eq:Qhat_homogeneous_rel_work_section}
    \Qhat \mkern-2mu: \TpM \to \Hom(\Ainp, \Aoutp)
\end{align}
at $p = \rM(\IM)$, which is the direct analog to the 1-argument kernel of \citet{Cohen2018-intertwiners}\cite{Cohen2019-generaltheory}.

Note that the full kernel fields can via the action of $\I$ be reconstructed from the single 1-argument kernels.
The theories derive therefore both a form of convolutional weight sharing from the requirement of global symmetry equivariance.
While kernels can for transitive symmetries be shared over the whole homogeneous space, they can in general only be shared over the orbits of the symmetry group.
If the manifold is asymmetric in such a way that the orbits are single points no weights can be shared with this definition.
As this is the default case for Riemannian manifolds, $\GM$-convolutions resort to the sharing of $G$-steerable kernels by placing them relative to frames of the $G$-structure.
This definition does not have a counterpart in steerable CNNs on homogeneous spaces.
Our Theorem~\ref{thm:GM_conv_homogeneous_equivalence} shows, however, that the global symmetry induced weight sharing is for the specific case of homogeneous spaces equivalent to our process of sharing $G$-steerable kernels.
In other words, isometry equivariant kernel field transforms on homogeneous spaces are necessarily convolutions
-- this mirrors the central results of \citet{Kondor2018-GENERAL}, \citet{bekkers2020bspline} and \citet{Cohen2018-intertwiners}\cite{Cohen2019-generaltheory}.

After investigating the analogies for the global, coordinate free kernels of both theories, we compare the definition of their coordinate representations relative to local trivializations.
Given some choice of trivializing neighborhoods $U^P \subseteq \I/\Hout$ and $U^A \subseteq \I/\Hin$, the unconstrained global 2-argument kernels of \citet{Cohen2018-intertwiners}\cite{Cohen2019-generaltheory} are locally represented by unconstrained functions
\begin{align}
    \widehat{\overleftarrow{\kappa}}:\, U^P \times U^A \to \R^{\cout\times\cin} \,.
\end{align}
In our theory, we instead have a single trivializing neighborhood $U^P = U^A \subseteq M$ relative to which a kernel field is given by an unconstrained map
\begin{align}
    \K^A:\, U^A \times \R^d \to \R^{\cout\times\cin} \,.
\end{align}
Investigating the \emph{global} $\I$-equivariance of the operator $\mathfrak{K}$ based on \emph{local} kernels is on non-trivial bundles necessarily difficult as it involves multiple trivializations.
The equivariance requirement implies for steerable CNNs on homogeneous spaces the constraints between different local kernels in Eq.~\eqref{eq:intertwiner_constraint_local}.
They leads to the 1-argument kernels
\begin{align}
    \overleftarrow{\kappa}^{EA} :\, U^A \to \R^{\cout\times\cin} \,,
\end{align}
from Eq.~\eqref{eq:local_one_arg_kernel}, which are still subject to the steerability constraint in Eq.~\eqref{eq:single_steerability}.
Single kernels $\Kp: \TpM \to \Hom(\Ainp,\Aoutp)$ (like e.g. $\Qhat$ from Eq.~\eqref{eq:Qhat_homogeneous_rel_work_section}) are according to Eq.~\eqref{eq:kernel_field_general_coord_expression} in coordinates given by functions
\begin{align}
    \Kp^A:\, \R^d \to \R^{\cout\times\cin} \,,
\end{align}
whose domains are tangent space coordinates $\R^d$ instead of of a open subset $U^A$ of the manifold.
A particular important example are $G$-steerable kernels, which correspond to the $\GM$-convolutional kernel fields from Def.~\ref{dfn:conv_kernel_field}.

While our kernels are globally defined in a single gauge $\psiTMp^A$ of $\TpM$, the local 1-argument kernels of \citet{Cohen2018-intertwiners}\cite{Cohen2019-generaltheory} need to be defined on an open cover of $\I/\Hin$.
As this is significantly more complicated, they propose therefore to represent the kernels on a single gauge which is defined almost everywhere.%
\footnote{
    In practice, one might anyways work with compactly supported kernels on a single trivializing neighborhood, which would render this choice unproblematic.
}
Note that this still requires that this single trivializing neighborhood is closed under the left action of $\Hout$ in order for the constraint in Eq.~\eqref{eq:single_steerability} to make sense.
We investigated this approach in Section~\ref{sec:spherical_CNNs_fully_equivariant} for the specific example of spherical CNNs, defining kernels on the trivializing neighborhood $U^A = S^2\backslash -n$.
Theorem~\ref{thm:spherical_kernel_space_iso} proved that Cohen \mbox{et al.'s}~\cite{Cohen2018-intertwiners}\cite{Cohen2019-generaltheory} steerable kernels on $S^2\backslash -n \subset S^2$ are in this case isomorphic to our $G$-steerable kernels on $B_{\R^2}(0,\pi) \subset \R^2$.
The equivalence of the corresponding convolutions was established in Theorem~\ref{thm:spherical_conv_GM_conv}.

Finally, we discuss which class of models steerable CNNs on homogeneous spaces cover.
Obviously, the theory does not describe convolutions on non-homogeneous spaces like punctured Euclidean spaces $\Euc_d\backslash\{0\}$, the sphere without poles $S^2 \backslash \{n,s\}$, whose isometries $\O2$ are non-transitive, the icosahedron, general surfaces or the M\"obius strip.
However, in contrast to $\GM$-convolutions, the base spaces $\I/H_l$ are not required to be Riemannian manifolds.
While \citet{Kondor2018-GENERAL} and \citet{bekkers2020bspline} cover only those convolutions whose feature fields transform according to scalar fields on $\I/H_l$, the associated bundle formulation of \citet{Cohen2018-intertwiners}\cite{Cohen2019-generaltheory} allows for general field representations~$\rho_l$.
Restricting to unimodular groups, steerable CNNs on homogeneous spaces do, however, only include those $\Aff(G)$-equivariant Euclidean convolutions for which the structure groups are subgroups of~$\O{d}$.
This reflects in the fact that the steerability constraints of \citet{Cohen2018-intertwiners}\cite{Cohen2019-generaltheory} do not include the determinant factor in the constraint of \citet{bekkers2020bspline} and of our $G$-steerable kernels.

%% file: chapters/apx05_lifting_iso_proof.tex

\section{Quotient representative kernel fields -- proofs}

In this appendix we give proofs for Theorems~\ref{thm:tangent_quotient_repr_kernel_fields} and~\ref{thm:GM_conv_homogeneous_equivalence}.

\toclesslab\subsection{Proof of Theorem~\ref{thm:tangent_quotient_repr_kernel_fields} -- Isomorphism between isometry invariant and quotient representative kernel fields}{apx:lifting_iso_proof}

Theorem~\ref{thm:tangent_quotient_repr_kernel_fields} claims that the spaces $\KIfull$ of isometry invariant kernel fields in Eq.~\eqref{eq:KIfull_def} and $\KIquot$ of quotient representative kernel fields in Eq.~\eqref{eq:KIquot_def} are isomorphic to each other and that the isomorphism is given by the lift $\Lambda$ whose inverse $\Lambda^{-1}$ is the restriction to $\rTM(\ITM)$.
Here we present a proof for this statement which consists of showing that
\textit{1)} $\Lambda^{-1}$ is indeed an inverse of $\Lambda$,
\textit{2)} the defining properties of $\KIfull$ and $\KIquot$ are satisfied after lifting and restricting and
\textit{3)} the constructions do not depend on arbitrary choices.

\begin{itemize}[leftmargin=0cm]

    \item[] {\textit{1)} $\Lambda^{\!-1}$ in Eq.~\eqref{eq:lifting_isomorphism_lambda_inv} is a well defined inverse of $\Lambda$ in Eq.~\eqref{eq:lifting_isomorphism_lambda} : }

    \begin{itemize}[leftmargin=1.1cm]
    \setlength\itemsep{2ex}

        \item[\it 1\hspace{1pt}a)]
            $\Lambda \circ \Lambda^{-1} = \id_{\KIfull}$,
            that is, $\Lambda^{-1}$ is a right inverse of $\Lambda$ :

            This claim follows for any $\K\in \KIfull$ and any $v \in TM$ from
            \begin{align}
                \big[\Lambda \circ \Lambda^{-1} (\K) \big](v)
                \ =&\ \big[ \Lambda(\Krestr) \big](v) \notag \\
                \ =&\ \dPhirHom{v} \, \Krestr     \, \rTM \QTM (v) \notag \\
                \ =&\ \dPhirHom{v} \, \K          \, \rTM \QTM (v) \notag \\
                \ =&\ \K           \, \dPhirTM{v} \, \rTM \QTM (v) \notag \\
                \ =&\ \K(v) \,,
            \end{align}
            where the invariance (equivariance) of the kernel field in Eq.~\eqref{eq:kernel_constraint_isom_full_1} allowed to swap the order of the isometry action and the evaluation of the kernel field in the penultimate step.

        \item[\it 1\hspace{1pt}b)]
            $\Lambda^{-1} \circ \Lambda = \id_{\KIquot}$,
            that is, $\Lambda^{-1}$ is a left inverse of $\Lambda$ :

            Let $\Q\in \KIquot$ and $w \in \rTM(\ITM)$.
            Note that $\rTM\QTM(w) = w$ since $w$ is an orbit representative.
            Furthermore, since $w = \dPhirTM{w}\, \rTM\QTM(w) = \dPhirTM{w} (w)$ it follows that $\Phir{w} \in \Stab{w}$ such that, by the constraint in Eq.~\eqref{eq:KIquot_def}, $\dPhirHom{w} \Q(w) = \Q(w)$.
            Together, this proves the claim:
            \begin{align}
                \big[\Lambda^{-1} \circ \Lambda (\Q) \big](w)
                \ =&\ \Lambda(\Q) \big|_{\rTM(\ITM)}(w) \notag \\
                \ =&\ \Lambda(\Q)(w) \notag \\
                \ =&\ \dPhirHom{w} \Q\, \rTM \QTM(w) \notag \\
                \ =&\ \dPhirHom{w} \Q(w) \notag \\
                \ =&\ \Q(w)
            \end{align}

    \end{itemize}

    \item[] {\emph{2)} The defining properties of $\KIfull$ and $\KIquot$ are satisfied after lifting and restricting : }

    \begin{itemize}[leftmargin=1.1cm]
    \setlength\itemsep{2ex}
        \item[\it 2\hspace{1pt}a)]
            $\piHom \mkern-5mu\circ\mkern-2mu \Lambda(\Q) = \piTM$ for any $\Q \in \KIquot$,
            that is, the lift $\Lambda(\Q)$ is a bundle $M$-morphism :

            For any $\Q\in \KIquot$ and for any $v\in TM$ this claim follows from
            \begin{align}
                \big[ \piHom \Lambda(\Q) \big](v)
                \ =&\ \piHom \dPhirHom{v} \Q\, \rTM \QTM(v) \notag \\
                \ =&\ \Phir{v} \piHom \Q\, \rTM \QTM(v) \notag \\
                \ =&\ \Phir{v} \piTM\, \rTM \QTM(v) \notag \\
                \ =&\ \Phir{v} \rM \piITM\, \QTM(v) \notag \\
                \ =&\ \Phir{v} \rM \QM \piTM(v) \notag \\
                \ =&\ \piTM(v) \,,
            \end{align}
            where the last step made use of Eq.~\eqref{eq:reconstruction_isometry_basespace}.

        \item[\it 2\hspace{1pt}b)]
            ${\piHom \mkern-5mu\circ\mkern-2mu \Lambda^{-1}(\K) = \piTM}$ for any $\K \in \KIfull$,
            that is, $\Lambda^{-1}(\K)$ is a bundle $\rM(\IM)$-morphism :

            This property follows immediately from the corresponding property of $\K$ after restricting to $\rTM(\ITM) \subseteq \piTM^{-1}\big(\rM(\IM)\big)$.
            For any $w \in \rTM(\ITM)$:
            \begin{align}
                \piHom \big[ \Lambda^{-1}(\K) \big](w)
                \ =&\ \piHom \Krestr(w) \notag \\
                \ =&\ \piHom \K(w) \notag \\
                \ =&\ \piTM(w) \notag \\
            \end{align}

        \item[\it 2\hspace{1pt}c)]
            $\dphiHom \Lambda(\Q)\, \dphiTMinv = \Lambda(\Q)\ \ \forall \phi \in \I$,
            that is, $\Lambda(\Q)$ satisfies the full isometry invariance constraint :

            Let $v\in TM$ and $\phi \in \I$.
            Due to the invariance of the quotient map $\QTM$ under isometries we have $\QTM(\dphiTMinv v) = \QTM(v)$.
            Note further that
            \begin{align}
                & \big[\Phir{v}^{-1}\, \phi\; \Phir{\dphiTMinv v}\big]_{*,\scalebox{.58}{$TM$}} \rTM \QTM(v) \notag \\
                \ =\ & \big[\Phir{v}^{-1}\, \phi\; \Phir{\dphiTMinv v}\big]_{*,\scalebox{.58}{$TM$}} \rTM \QTM\big( \dphiTMinv v\big) \notag \\
                \ =\ & \big[\Phir{v}^{-1}\, \phi \big]_{*,\scalebox{.58}{$TM$}}\, \dphiTMinv\, v \notag \\
                \ =\ & \dPhirTM{v}^{-1}\, v \notag \\
                \ =\ & \rTM \QTM(v)
            \end{align}
            implies
            \begin{align}
                \big[\Phir{v}^{-1}\, \phi\; \Phir{\dphiTMinv v}\big]\ \in\ \Stab{\rTM\QTM(v)} \,,
            \end{align}
            which, via the stabilizer constraint in Eq.~\eqref{eq:KIquot_def}, leads to
            \begin{align}
                \big[\Phir{v}^{-1}\, \phi\; \Phir{\dphiTMinv v}\big]_{*,\scalebox{.58}{$\Hom$}} \Q\; \rTM \QTM(v)
                \ =\ \Q\; \rTM \QTM(v) \,.
            \end{align}
            Putting these observations together proves the claim:
            \begin{align}
                \dphiHom \Lambda(\Q)\, \dphiTMinv(v)
                \ =&\ \dphiHom \dPhirHom{\dphiTMinv v} \Q\; \rTM \QTM \big(\dphiTMinv v\big) \notag \\
                \ =&\ \dphiHom \dPhirHom{\dphiTMinv v} \Q\; \rTM \QTM(v) \notag \\
                \ =&\ \big[\Phir{v}\, \Phir{v}^{-1}\big]_{*,\scalebox{.58}{$\Hom$}} \dphiHom\, \dPhirHom{\dphiTMinv v} \Q\; \rTM \QTM(v) \notag \\
                \ =&\ \dPhirHom{v}\, \big[\Phir{v}^{-1}\, \phi\; \Phir{\dphiTMinv v}\big]_{*,\scalebox{.58}{$\Hom$}} \Q\; \rTM \QTM(v) \notag \\
                \ =&\ \dPhirHom{v}\, \Q\; \rTM \QTM(v) \notag \\
                \ =&\ \Lambda(\Q)
            \end{align}

        \item[\it 2\hspace{1pt}d)]
            $\dxiHom \big[\Lambda^{-1}(\K)\big](w) = \big[\Lambda^{-1}(\K)\big](w) \ \ \
               \forall\; w \mkern-2mu\in\mkern-1mu \rTM(\ITM),\ \xi \mkern-1mu\in\mkern-1mu \Stab{w}$,\ 
            that is, $\Lambda^{-1}(\K)$ satisfies the stabilizer constraint :

            This statement is easily proven since the invariance (equivariance) properties of $\K$ carry over to its restriction $\Lambda^{-1}(\K)$.
            We obtain for arbitrary $w\in \rTM(\ITM)$ and $\xi\in \Stab{w}$, that:
            \begin{align}
                \dxiHom \big[\Lambda^{-1}(\K)\big](w)
                \ =&\ \dxiHom \Krestr (w) \notag \\
                \ =&\ \dxiHom \K(w) \notag \\
                \ =&\ \K \big(\dxiTM w\big) \notag \\
                \ =&\ \K(w) \notag \\
                \ =&\ \Krestr(w) \notag \\
                \ =&\ \big[\Lambda^{-1}(\K)\big](w)
            \end{align}

    \end{itemize}

    \item[] {\emph{3)} All constructions and proofs are independent from the particular choice of $\PhirNoArg$ : }
    \begin{itemize}[leftmargin=1.1cm]
    \setlength\itemsep{2ex}
        \item[]%
            The definition
            \begin{align}
                \PhirNoArg: TM \to \I \quad \textup{such that}\quad \dPhirTM{v} \rTM \QTM(v) = v
            \end{align}
            from Eq.~\eqref{eq:reconstruction_isometry} is unique up to right multiplication of $\PhirNoArg$ with \emph{any}
            \begin{align}
                \xirNoArg: TM \to \I \quad \textup{such that}\quad \xir{v} \in \Stab{\rTM\QTM(v)}
            \end{align}
            since, obviously, $\dPhirTM{v}\, \dxirTM{v}\, \rTM\QTM(v)\ =\ \dPhirTM{v}\, \rTM\QTM(v)\ =\, v\,$ for any $v\in TM$.
            As argued in footnote~\ref{footnote:ambiguity_reconstruction_isometry}, this covers all degrees of freedom in the definition of reconstruction isometries.
            From the stabilizer constraint in Eq.~\eqref{eq:KIquot_def} it follows that $\dxirHom{v} \Q\, \rTM\QTM(v) = \Q\, \rTM\QTM(v)$ such that the lift $\Lambda$ is seen to be invariant w.r.t. the ambiguity of $\PhirNoArg$:
            \begin{align}
                \Lambda(\Q)
                \ =&\ \dPhirHom{v} \Q\, \rTM\QTM(v) \notag \\
                \ =&\ \dPhirHom{v}\, \dxirHom{v} \Q\, \rTM\QTM(v) \notag \\
            \end{align}
            Except from the definition of the lifting isomorphism, $\PhirNoArg$ is only used (in a slightly different context) in step \textit{2\,c)}, where the ambiguity is seen to drop out by similar arguments.

    \end{itemize}

\end{itemize}

\noindent Together, these steps prove that $\Lambda: \KIquot \to \KIfull$ is an isomorphism.
\hfill$\Box$

\toclesslab\subsection{Proof of Theorem~\ref{thm:GM_conv_homogeneous_equivalence} -- Equivalence of equivariant kernel field transforms and convolutions on homogeneous spaces}{apx:homogeneous_equivalence_proof}

To keep a better overview, we split the proof in two parts, proving the claims made in the first and second statement of Theorem~\ref{thm:GM_conv_homogeneous_equivalence}, respectively.

\paragraph{Part 1) -- Constructing \textit{H}, \textit{HM} and Isom\textsubscript{\textit{HM}}:}
Let $r\in M$ be any representative point and, without loss of generality, let $\psiGMr^{\widetilde{A}}$ be any isometric gauge at $r$.
We set
\begin{align}
    H\ :=\ \psiGMr^{\widetilde{A}} \,\Stab{r} \big(\psiGMr^{\widetilde{A}} \big)^{-1} \,,
\end{align}
which is just a particular representation of $\Stab{r}$ relative to the chosen coordinatization.
Since the gauge maps are isomorphisms, we get an isomorphism between the two groups:
\begin{align}\label{eq:stabr_H_iso}
    \alpha: \Stab{r} \to H,\ \ \ \xi \to \psiGMr^{\widetilde{A}} \;\dxiGM\, \big(\psiGMr^{\widetilde{A}} \big)^{-1} =: h_\xi^{\widetilde{A}\widetilde{A}}(r)
\end{align}
Since $\Stab{r} \leq \I \leq \IsomGM$, Theorem~\ref{thm:isom_GM_in_coords} assures that $h_\xi^{\widetilde{A}\widetilde{A}}(r)$ is for any $\xi \in \Stab{p}$ an element of $G$ and thus that $H \leq G$.
We furthermore have that $H\leq\O{d}$, which is seen by the following calculation, which holds for any $\mathscr{v},\mathscr{w} \in \R^d$:
\begin{align}
    \pig\langle h_\xi^{\widetilde{A}\widetilde{A}}(r) \cdot\mathscr{v} \,,\,\ h_\xi^{\widetilde{A}\widetilde{A}}(r) \cdot\mathscr{w} \pig\rangle
    \ \overset{(1)}{=}&\ \ \pig\langle \pig( \psiGMr^{\widetilde{A}} \;\dxiGM\, \big(\psiGMr^{\widetilde{A}} \big)^{-1} \pig) \cdot \mathscr{v} \,,\,\ \pig( \psiGMr^{\widetilde{A}} \;\dxiGM\, \big(\psiGMr^{\widetilde{A}} \big)^{-1} \pig) \cdot \mathscr{w} \pig\rangle \notag \\
    \ \overset{(2)}{=}&\ \ \pig\langle \psiTMr^{\widetilde{A}} \;\dxiTM\, \big(\psiTMr^{\widetilde{A}} \big)^{-1} \,\mathscr{v} \,,\,\ \psiTMr^{\widetilde{A}} \;\dxiTM\, \big(\psiTMr^{\widetilde{A}} \big)^{-1} \,\mathscr{w} \pig\rangle \notag \\
    \ \overset{(3)}{=}&\ \ \eta_r\pig( \dxiTM\, \big(\psiTMr^{\widetilde{A}} \big)^{-1} \,\mathscr{v} \,,\,\ \dxiTM\, \big(\psiTMr^{\widetilde{A}} \big)^{-1} \,\mathscr{w} \pig) \notag \\
    \ \overset{(4)}{=}&\ \ \eta_r\pig( \big(\psiTMr^{\widetilde{A}} \big)^{-1} \,\mathscr{v} \,,\,\ \big(\psiTMr^{\widetilde{A}} \big)^{-1} \,\mathscr{w} \pig) \notag \\
    \ \overset{(5)}{=}&\ \ \langle \mathscr{v} \,, \mathscr{w} \rangle
\end{align}
Step~$(1)$ made use of Eq.~\eqref{eq:stabr_H_iso}.
In step~$(2)$ we identified the expression of $h_\xi^{\widetilde{A}\widetilde{A}}(r)$ via $\psiGMr^{\widetilde{A}}$ with its expression via $\psiTMr^{\widetilde{A}}$, which is justified by the commutativity of the diagrams in Eqs.~\eqref{cd:pushforward_GM_coord_extended} and~\eqref{cd:pushforward_TM_coord}.
As we assumed~$\psiTMr^{\widetilde{A}}$ w.l.o.g. to be isometric, we can identify the inner product $\langle\,\cdot,\cdot\,\rangle$ on~$\R^d$ in step~$(3)$ with the Riemannian metric~$\eta_r$.
Step~$(4)$ uses that $\xi \in \Stab{r} \leq \I$ is an isometry, which preserves the metric by definition; see Eq.~\eqref{eq:isometry_def}.
Lastly, we pull the metric in step~$(5)$ via the isometric gauge back to the inner product on~$\R^d$.
The equality of the initial and final expression shows that $h_\xi^{\widetilde{A}\widetilde{A}}(r)$ preserves the inner product on $\R^d$ -- this is exactly the requirement that \emph{defines} the orthogonal group.
We therefore have that $H\leq \O{d}$, and, together with $H\leq G$, that
\begin{align}
    H \,\leq\, G \cap \O{d} \,.
\end{align}
This proves the first statement of part 1) of Theorem~\ref{thm:GM_conv_homogeneous_equivalence}.
We move on to the second statement of part 1), the construction of $\HM$ and $\IsomHM$.

Given that $\Stab{r}$ is a subgroup of $\I$, we have the canonical quotient map
\begin{align}
    \mathscr{q}: \I \to \I/\Stab{r},\ \ \ \phi \to \phi.\Stab{r}
\end{align}
which sends group elements $\phi \in \I$ to the left coset $\phi.\Stab{r} := \{\phi\,\xi \,|\, \xi\in\Stab{r} \}$ of $\Stab{r}$.
It is well known that this quotient map makes $\I$ to a principal $\Stab{r}$-bundle over the base space $\I/\Stab{r}$, with the right action given by the right multiplication $\blacktriangleleft \,: \I \times \Stab{r} \to \I,\ (\phi,\xi) \mapsto \phi\,\xi$ with stabilizer elements~\cite{gallier2019diffgeom2,neeb2010differential}.
Furthermore, $\I/\Stab{r}$ is isomorphic to the homogeneous space~$M$.
The isomorphism is given by
\begin{align}
    \beta: \I/\Stab{r} \to M,\ \ \ \phi.\Stab{r} \mapsto \phi(r) \,,
\end{align}
which is obviously independent of the choice of coset representative since different representatives differ by group elements that stabilize~$r$.
Note that we could equally well view $\mathscr{q}: \I \to \I/\Stab{r}$ as a principal $H$-bundle since the typical fiber is only defined up to isomorphism.

With these preparations we define the $H$-structure $\HM$ as an embedding of the principal $H$-bundle $\I$ into $\GM$ (and therefore into $\FM$).
We define the embedding map as
\begin{align}
    \mathscr{E}: \I \to \GM,\ \ \ \phi \mapsto \dphiGM\, \sigma^{\widetilde{A}}(r) \,,
\end{align}
which depends once again on our choice of gauge since $\sigma^{\widetilde{A}}(r) = \big(\psiGMr^{\widetilde{A}}\big)^{-1}(e)$.
It can be thought of as tracing out an embedded copy of $\I$ in $\GM$ by pushing around the frame $\sigma^{\widetilde{A}}(r) \in \GrM$.
That this gives indeed a valid embedding is guaranteed since the action of $\I$ on frames is fixed point free.
The embedding $\mathscr{E}$ is a bundle map over $\beta$, that is, $\beta\circ\mathscr{q} = \piGM \circ \mathscr{E}$.
To show this, it is sufficient to apply both sides on an arbitrary element $\phi\in\I$, which gives the same result:
$\beta \circ \mathscr{q} (\phi) = \beta\big( \phi.\Stab{r} \big) = \phi(r)$ and
$\piGM \circ \mathscr{E} (\phi) = \piGM\, \dphiGM\, \sigma^{\widetilde{A}}(r) = \phi\; \piGM\, \sigma^{\widetilde{A}}(r) = \phi(r)$.
The embedding map is furthermore right equivariant:
For any $\xi \in \Stab{r}$ and any $\phi \in \I$ one has
\begin{align}
    \mathscr{E}(\phi\,\xi)
    \ =&\ \ \dphiGM\, \dxiGM \sigma^{\widetilde{A}}(r) \notag \\
    \ =&\ \ \dphiGM\, \dxiGM \big(\psiGMr^{\widetilde{A}}\big)^{-1}(e) \notag \\
    \ =&\ \ \dphiGM\, \big(\psiGMr^{\widetilde{A}}\big)^{-1} \psiGMr^{\widetilde{A}}\, \dxiGM \big(\psiGMr^{\widetilde{A}}\big)^{-1}(e) \notag \\
    \ =&\ \ \dphiGM\, \big(\psiGMr^{\widetilde{A}}\big)^{-1} \big( h_\xi^{\widetilde{A}\widetilde{A}}(r) \big) \notag \\
    \ =&\ \ \dphiGM\, \big(\psiGMr^{\widetilde{A}}\big)^{-1}(e) \lhd h_\xi^{\widetilde{A}\widetilde{A}}(r) \notag \\
    \ =&\ \ \mathscr{E}(\phi) \lhd h_\xi^{\widetilde{A}\widetilde{A}}(r) \,,
\end{align}
where we used the right $G$ (and thus $H$) equivariance of $\psiGMr^{\widetilde{A}}$ (and thus $\big(\psiGMr^{\widetilde{A}}\big)^{-1}$) in the penultimate step.
Together, these properties show that $\mathscr{E}$ is a principal bundle map that makes the following diagram commutative:
\begin{equation}\label{cd:GM_def_embedding}
\begin{tikzcd}[row sep=3.em, column sep=6.5em]
    \I \times \Stab{r}
        \arrow[r, "\mathscr{E}\times\alpha", hook]
        \arrow[d, "\blacktriangleleft\,"']
    & \GM \times H
        \arrow[d, "\,\lhd"]
    \\
    \I
        \arrow[r, pos=.55, "\mathscr{E}", hook]
        \arrow[d, "\mathscr{q}"']
    & \GM
        \arrow[d, "\piGM"]
    \\
    \I/\Stab{r}
        \arrow[r, pos=.45, "\beta"']
    & M
\end{tikzcd}
\end{equation}
The claimed $H$-structure is then defined as the image
\begin{align}
    \HM\ :=\ \mathscr{E}(\I)\ =\ \big\{ \dphiGM\, \sigma^{\widetilde{A}}(r) \,\big|\, \phi \in \I \big\}
\end{align}
of $\mathscr{E}$ together with the restricted right action and projection map of $\GM$.
Since embeddings are necessarily injective, we have in particular that $\I$ and $\HM$ are isomorphic as principal bundles.

As a last point we argue that $\I$ and $\IsomHM = \{ \theta \in \IsomM \,|\, \dthetaGM \HM = \HM \}$ coincide.
The equality $\dthetaGM \HM = \HM$ holds for a given $\theta \in \IsomM$ if $\dthetaGM \HM$ is at the same time a subset and a superset of~$\HM$.
The first case, $\dthetaGM \HM \subseteq \HM$, requires that for any element $\dthetaGM\, \dphiGM \sigma^{\widetilde{A}}(r) \in \dthetaGM \HM$, there exists some $\dphiGM' \sigma^{\widetilde{A}}(r) \in \HM$ such that $\dthetaGM\, \dphiGM \sigma^{\widetilde{A}}(r) = \dphiGM' \sigma^{\widetilde{A}}(r)$.
Since the action of isometries on the frame bundle is free, this requires $\dthetaGM = \dphiGM' \dphiGM^{-1}$, which in turn implies $\theta = \phi' \phi^{-1}$.
As one can easily check, the second case results in the same requirement.
Both $\phi'$ and $\phi$ are elements of $\I$ such that $\theta$ is required to be an element of $\I$.
This proves the claim
\begin{align}\label{eq:IsomHM_I}
    \IsomHM = \I \,.
\end{align}

\paragraph{Part 2) -- Equivalence of $\bm\I$-equivariant kernel field transforms and \textit{HM}-convolutions:}

To prove the second statement of the theorem, we construct an $\I$-equivariant kernel field transform on $M$ and show that it is equivalent to a $\HM$-convolution.
Theorem~\ref{thm:isometry_equivariant_kernel_field_trafos} proved that $\I$-equivariant kernel field transforms require $\I$-invariant kernel fields, which can, according to Theorem~\ref{thm:manifold_quotient_repr_kernel_fields}, be equivalently encoded in terms of a field of representative kernels $\Qhat: \piTM^{-1}(\rM(\IM)) \to \piHom^{-1}(\rM(\IM))$.
For the case of a homogeneous space $M$, the quotient space $\IM$ consists of a single element, which we represent by $r=\rM(\IM) \in M$.
The full invariant kernel field is therefore described by a single kernel $\Qhat|_r = \Qhat: \TrM \to \Hom(\Ainr,\Aoutr)$.
This kernel is required to satisfy the stabilizer constraint
$\dxiHom \Qhat\; \dxiTM^{-1} = \Qhat \quad \forall\ \xi \in \Stab{r}$
and is shared over $M$ via the lifting isomorphism
$\widehat{\Lambda}(\Qhat)(v) = \dPhirHom{v} \Qhat\ \rTM \QTM(v) = \dPhirHom{v} \Qhat\ \dPhirTM{v}^{-1}(v)$.
As shown below, the single $\Stab{r}$-constrained representative kernel corresponds exactly to an $H$-steerable template kernel, while the weight sharing via the lifting isomorphism $\widehat{\Lambda}$ from Theorem~\ref{thm:manifold_quotient_repr_kernel_fields} corresponds exactly to the convolutional weight sharing in Def.~\ref{dfn:conv_kernel_field}.

To make the equivalence of the kernel constraints explicit, we express the kernel $\Qhat$ via Eq.~\eqref{eq:conv_kernel_field_def_ptwise} relative to the same gauge~$\widetilde{A}$ as considered before as
$K := \psiHomr^{\widetilde{A}}\, \Qhat\, \big(\psiTMr^{\widetilde{A}} \big)^{-1}$.
The frame volume factor $\sqrt{|\eta_r^{\widetilde{A}}|}$ drops hereby out since we assumed the gauge w.l.o.g. to be isometric.
The stabilizer constraint relative to this gauge then leads to
\begin{alignat}{3}
    K
    \ &=&\ \ \psiHomr^{\widetilde{A}}\, &\Qhat\ \big(\psiTMr^{\widetilde{A}} \big)^{-1} \notag \\
    \ &=&\ \ \psiHomr^{\widetilde{A}}\, \dxiHom\, &\Qhat\,\ \dxiTM^{-1}\, \big(\psiTMr^{\widetilde{A}} \big)^{-1} \notag \\
    \ &=&\ \ \psiHomr^{\widetilde{A}}\, \dxiHom\, \big(\psiHomr^{\widetilde{A}})^{-1}\, &K\ \psiTMr^{\widetilde{A}}\, \dxiTM^{-1}\, \big(\psiTMr^{\widetilde{A}} \big)^{-1} \notag \\
    \ &=&\ \ \rhoHom\big( h_\xi^{\widetilde{A}\widetilde{A}}(r) \big)\, &K\ \big( h_\xi^{\widetilde{A}\widetilde{A}}(r) \big)^{-1} \notag \\
    \ &=&\ \ \frac{1}{\big|\mkern-2mu \det h_\xi^{\widetilde{A}\widetilde{A}}(r) \big|} \;
        \rhoHom\big( h_\xi^{\widetilde{A}\widetilde{A}}(r) \big)\, &K\ \big( h_\xi^{\widetilde{A}\widetilde{A}}(r) \big)^{-1}
\end{alignat}
for any $\xi$ in $\Stab{r}$.
Note that we can include the determinant factor in the last step since $h_\xi^{\widetilde{A}\widetilde{A}}(r) \in \O{d}$, as shown above.
The isomorphism between $\Stab{r}$ and $H$ in Eq.~\eqref{eq:stabr_H_iso} thus allows us to rewrite the stabilizer constraint as the $H$-steerability constraint
\begin{align}
    K\ =\ \frac{1}{|\mkern-2mu \det h \mkern1mu|}\; \rhoHom(h) \circ K \circ h^{-1} \qquad \forall\ h \in H \,.
\end{align}
on template kernels $K$ of a $\HM$-convolution.%
\footnote{
    Since $h\in H \leq G\cap\O{d}$, the determinant factor always drops out and could therefore be omitted.
}

What remains to be shown is the equivalence of the two ways of sharing weights.
The weight sharing via $\widehat{\Lambda}$, expressed via gauge $\widetilde{A}$ in terms of $K$, reads
\begin{alignat}{3}\label{eq:weight_sharing_lifting_homogeneous}
    \widehat{\Lambda}(\Qhat)(v)
    \ &=&\ \dPhirHom{v}\; &\Qhat\,\ \rTM\, \QTM (v) \notag \\
    \ &=&\ \dPhirHom{v}\; &\Qhat\,\ \dPhirTM{v}^{-1}(v) \notag \\
    \ &=&\ \dPhirHom{v}\, \big(\psiHomr^{\widetilde{A}}\big)^{-1}\, \psiHomr^{\widetilde{A}}\; &\Qhat\,\ \big(\psiTMr^{\widetilde{A}}\big)^{-1}\, \psiTMr^{\widetilde{A}}\; \dPhirTM{v}^{-1}(v) \notag \\
    \ &=&\ \Big( \psiHomr^{\widetilde{A}}\, \dPhirHom{v}^{-1} \Big)^{-1}\, &K\, \Big( \psiTMr^{\widetilde{A}}\, \dPhirTM{v}^{-1} \Big)(v) \,.
\end{alignat}
The last line already looks quite similar to the definition of $\HM$-convolutional kernel fields in Def.~\ref{dfn:conv_kernel_field}.
To prove their equivalence, we need to show
1) that the isometry induced gauges 
$\psiTMr^{\widetilde{A}}\, \dPhirTM{v}^{-1}$ and $\psiHomr^{\widetilde{A}}\, \dPhirHom{v}^{-1}$ at~$\piTM(v)$
are $H$-compatible with the original gauges $\psiTMr^{\widetilde{A}}$ and $\psiHomr^{\widetilde{A}}$ and
2) that the induced gauges correspond to reference frames of unit volume (to explain the missing frame volume factor in Eq.~\eqref{eq:weight_sharing_lifting_homogeneous}).
For the first point, note that the codomain of the reconstruction isometry $\PhirNoArg: \TM \to \I$ coincides by Eq.~\eqref{eq:IsomHM_I} with~$\IsomHM$.
Theorem~\ref{thm:isom_GM_in_coords} therefore asserts that these induced gauges are compatible with any $H$-atlas of~$\HM$.
The second point follows immediately since $H \leq \O{d}$ (or since $\Phir{v}$ is an isometry and $\widetilde{A}$ is isometric).
The weight sharing of $\Qhat$ via the lifting isomorphism in Eq.~\ref{eq:weight_sharing_lifting_homogeneous} is therefore seen to coincide with the $\HM$-convolutional weight sharing of the $H$-steerable kernel~$K$ in Def.~\ref{dfn:conv_kernel_field}.
Together with the result that the stabilizer kernel constraint results in the $H$-steerability constraint, this implies that the lifted kernel field is equivalent to a $\HM$-convolutional kernel field, which proves part 2) of the theorem.

A different choice of gauge $\widetilde{A}$ might for $G<\O{d}$ result in a conjugate subgroup $\overline{H}$ to $H$ and an embedding $\overline{H}\mkern-2muM$ of $\I$ that differs from $\HM$.
As one can easily check, the $\overline{H}$-steerability constraint allows to describe the same kernel relative to $\overline{H}\mkern-2muM$ like the $H$-steerability constraint in relation to $\HM$, since the transformation falls out.

%% file: chapters/apx06_spherical_kernels.tex

\section{Spherical steerable convolutions as \textit{GM}-convolutions -- proofs}
\label{apx:spherical_conv_main}

This Section presents the proofs of Theorems~\ref{thm:spherical_kernel_space_iso} and~\ref{thm:spherical_conv_GM_conv} from Section~\ref{sec:spherical_CNNs_fully_equivariant} in the following two subsections.
Together, these theorems assert that the $\Stab{n}$-steerable spherical convolution kernels by \citet{Cohen2019-generaltheory} are equivalent to certain $\Stab{n} \cong G$-steerable kernels, and that the $\I$-equivariant spherical convolutions with these kernels are equivalent to our corresponding $\GM$-convolutions.

\toclesslab\subsection{Proof of Theorem~\ref{thm:spherical_kernel_space_iso} -- Kernel space isomorphism}{apx:spherical_conv_kernel_space_iso}

Theorem~\ref{thm:spherical_kernel_space_iso} establishes an isomorphism
\begin{align}
  \Omega:\ 
  \mathscr{K}^{G,B_{\R^2}\mkern-1mu(0,\pi)}_{\rhoin\mkern-1mu,\rhoout}
  \xrightarrow{\,\sim\,}\,
  \mathscr{K}^{\Stab{n}}_{\rhoin\mkern-1mu,\rhoout}
\end{align}
between the space $\mathscr{K}^{G,B_{\R^2}\mkern-1mu(0,\pi)}_{\rhoin\mkern-1mu,\rhoout}$ of $G$-steerable kernels on the open ball $B_{\R^2}\mkern-1mu(0,\pi) \subset \R^2$ and the space $\mathscr{K}^{\Stab{n}}_{\rhoin\mkern-1mu,\rhoout}$ of $G\cong \Stab{n}$-steerable kernels on $S^2 \backslash \mkern-1mu\minus\mkern1mu n$, which are defined in Eqs.~\eqref{eq:G_steer_kernel_space_open_ball_pi} and~\eqref{eq:spherical_steerable_kernel_space}.
Given arbitrary gauges $N$ at the north pole $n$, around which the kernel is centered, and gauges $P$ at any other point $p$, this isomorphism is given by
\begin{alignat}{4}
    \Omega(K)\! &:&\,\ S^2 \backslash \mkern-1mu\minus\mkern1mu n \,&\to&\, \R^{\cout\times\cin},
    \quad p \,&\mapsto\,
    \big[\Omega(K)\big](p)\ &:=&\ K\big( \psiTMn^N \log_n p \big)\, \rhoin\big( g_{n\leftarrow p}^{NP} \big)\, \sqrt{\big|\eta_p^{\partial\mkern-2mu/\mkern-2mu\partial\mathscr{v}}\big|}^{\,-1} \,.
\intertext{
    Abbreviating $p := \exp_n\! \big(\psiTMn^N\big)^{\!-1} \mathscr{v}$, its inverse is given by
}
    \Omega^{-1}(\kappa)\! &:&\,\ B_{\R^2}\mkern-1mu(0,\pi) &\to& \R^{\cout\times\cin},
    \quad \mathscr{v} \,&\mapsto\,
    \big[\Omega^{-1}(\kappa)\big](\mathscr{v})\ &:=&\ \kappa\big(\! \exp_n\! \big(\psiTMn^N\big)^{\!-1} \mathscr{v} \big)\, \rhoin\big( g_{n\leftarrow p}^{NP} \big)^{\!-1} \sqrt{\big|\eta_p^{\partial\mkern-2mu/\mkern-2mu\partial\mathscr{v}}\big|} \!,
\end{alignat}

\begin{proof}
    That $\Omega^{-1}$ is a well defined inverse of $\Omega$ is easily shown by inserting their expressions and verifying that
    \begin{align}
        \Omega \circ \Omega^{-1} = \id_{\mathscr{K}^{\Stab{n}}_{\rhoin\mkern-1mu,\rhoout}}
        \qquad \textup{and} \qquad
        \Omega^{-1} \circ \Omega = \id_{\mathscr{K}^{G,B_{\R^2}\mkern-1mu(0,\pi)}_{\rhoin\mkern-1mu,\rhoout}}
    \end{align}
    hold.
    To see this, note that gauges $\psiTMn^N$, the transporters $\rhoin\big( g_{n\leftarrow p}^{NP} \big)$ and the (non-zero) volume factor $\sqrt{\big|\eta_p^{\partial\mkern-2mu/\mkern-2mu\partial\mathscr{v}}\big|}$ are always invertible and the latter two commute since the volume scaling factor is a scalar.
    The exponential map $\exp_n: B_{\R^2}\mkern-1mu(0,\pi) \to S^2 \backslash \mkern-1mu\minus\mkern1mu n$ on $B_{\R^2}\mkern-1mu(0,\pi)$ is inverted by $\log_n: S^2 \backslash \mkern-1mu\minus\mkern1mu n \to B_{\R^2}\mkern-1mu(0,\pi)$.

    The kernel constraints of the two kernel spaces furthermore imply each other.
    Given any $G$-steerable kernel $K \in \mathscr{K}^{G,B_{\R^2}\mkern-1mu(0,\pi)}_{\rhoin\mkern-1mu,\rhoout}$, the kernel $\Omega(K) \in \mathscr{K}^{\Stab{n}}_{\rhoin\mkern-1mu,\rhoout}$ satisfies the $\Stab{n}$-steerability constraint from Eq.~\eqref{eq:spherical_steerable_kernel_space}.
    This is for any $p\in S^2 \backslash \mkern-1mu\minus\mkern1mu n$, any $\xi \in \Stab{n}$ and any gauge $X$ at $\xi(p)$ shown by:
    \begin{align}
        \big[\Omega(K)\big]\big( \xi(p) \big)
        \,&\overset{(1)}{=}\, K\big( \psiTMn^N \log_n \xi(p) \big) \cdot \rhoin\big( g_{n\leftarrow\xi(p)}^{NX} \big) 
               \ \sqrt{\big|\eta_{\xi(p)}^{\partial\mkern-2mu/\mkern-2mu\partial\mathscr{v}}\big|}^{\,-1} \notag \\
        \,&\overset{(2)}{=}\, K\big( \psiTMn^N \dxiTM \log_n p \big) \cdot \rhoin\big( g_{n\leftarrow\xi(p)}^{NX} \big) 
               \ \sqrt{\big|\eta_{\xi(p)}^{\partial\mkern-2mu/\mkern-2mu\partial\mathscr{v}}\big|}^{\,-1} \notag \\
        \,&\overset{(3)}{=}\, K\big( g_\xi^{NN}(n) \psiTMn^N \log_n p \big) \cdot \rhoin\big( g_{n\leftarrow\xi(p)}^{NX} \big) 
               \ \sqrt{\big|\eta_{\xi(p)}^{\partial\mkern-2mu/\mkern-2mu\partial\mathscr{v}}\big|}^{\,-1} \notag \\
        \,&\overset{(4)}{=}\, \rhoout\big( g_\xi^{NN}(n) \big) \cdot K\big(\psiTMn^N \log_n p \big) \cdot \rhoin\big( g_\xi^{NN}(n) \big)^{-1} \rhoin\big( g_{n\leftarrow\xi(p)}^{NX} \big) 
               \ \sqrt{\big|\eta_{\xi(p)}^{\partial\mkern-2mu/\mkern-2mu\partial\mathscr{v}}\big|}^{\,-1} \notag \\
        \,&\overset{(5)}{=}\, \rhoout\big( g_\xi^{NN}(n) \big) \cdot K\big(\psiTMn^N \log_n p \big) \cdot \rhoin\big( g_\xi^{NN}(n) \big)^{-1} \rhoin\big( g_{n\leftarrow\xi(p)}^{NX} \big)
               \ \sqrt{\big|\eta_p^{\partial\mkern-2mu/\mkern-2mu\partial\mathscr{v}}\big|}^{\,-1} \notag \\
        \,&\overset{(6)}{=}\, \rhoout\big( g_\xi^{NN}(n) \big) \cdot K\big(\psiTMn^N \log_n p \big) \cdot \rhoin\big( g_{n\leftarrow p}^{NP} \big) \rhoin\big( g_\xi^{XP}(p) \big)^{-1} 
               \ \sqrt{\big|\eta_p^{\partial\mkern-2mu/\mkern-2mu\partial\mathscr{v}}\big|}^{\,-1} \notag \\
        \,&\overset{(7)}{=}\, \rhoout\big( g_\xi^{NN}(n) \big) \cdot \big[\Omega(K)\big](p) \cdot \rhoin\big( g_\xi^{XP}(p) \big)^{-1}
    \end{align}
    The first step just expanded $\Omega(K)$, while the second step used $\log_n\xi(p) = \dxiTM \log_{\xi^{-1}(n)} p$, which follows from Eq.~\eqref{eq:exp_isom_commutation}, together with $\xi^{-1}(n) = n$ since $\xi\in\Stab{n}$.
    In the third step, we used the definition of isometry induced gauge transformations in Eq.~\eqref{eq:pushforward_TM_coord}.
    Step four used the $G$-steerability constraint from Eq.~\eqref{eq:G_steer_kernel_space_open_ball_pi}.
    The firth step replaced the volume element 
    $\sqrt{\big|\eta_{\xi(p)}^{\partial\mkern-2mu/\mkern-2mu\partial\mathscr{v}}\big|}$
    with that at
    $\sqrt{\big|\eta_p^{\partial\mkern-2mu/\mkern-2mu\partial\mathscr{v}}\big|}$,
    which is possible since the whole Riemannian geometry of the sphere, including the metric and exponential map and therefore the volume factors of the geodesic normal coordinates, is invariant under the action of~$\Stab{n}$.
    Before identifying $\Omega(K)$ in the last step, step six used the identity
    \begin{align}
        \rhoin\big( g_\xi^{NN}(n) \big)^{-1}\, \rhoin\big( g_{n\leftarrow\xi(p)}^{NX} \big)
        \ =&\ \pig[ \psiAinn^N\, \dxiAin^{-1}\, \big(\psiAinn^N \big)^{-1} \pig] \pig[ \psiAinn^N\, \PAinnxip\, \big(\psiAinxip^X \big)^{-1} \pig] \notag \\
        \ =&\ \psiAinn^N\; \dxiAin^{-1}\, \PAinnxip\, \big(\psiAinxip^X \big)^{-1} \notag \\
        \ =&\ \psiAinn^N\; \PAinnp\, \dxiAin^{-1}\ \big(\psiAinxip^X \big)^{-1} \notag \\
        \ =&\ \pig[ \psiAinn^N\, \PAinnp\, \big(\psiAinp^P \big)^{-1} \pig] \pig[ \psiAinp^P\, \dxiAin^{-1}\, \big(\psiAinxip^X \big)^{-1} \pig] \notag \\
        \ =&\ \rhoin\big( g_{n\leftarrow p}^{NP} \big)\ \rhoin\big( g_\xi^{XP}(p) \big)^{-1} \,,
    \end{align}
    which relies crucially on the commutativity of transporters and isometry pushforwards from Eq.~\eqref{eq:transport_isom_commutation}.

    For the opposite direction, assume a $\Stab{n}$-steerable kernel $\kappa\in \mathscr{K}^{\Stab{n}}_{\rhoin\mkern-1mu,\rhoout}$ to be given.
    The corresponding kernel $\Omega^{-1}(\kappa)$ satisfies then the $G$-steerability constraint from Eq.~\eqref{eq:G_steer_kernel_space_open_ball_pi}.
    To show this, let $\mathscr{v} \in B_{\R^2}\mkern-1mu(0,\pi)$, let $g\in G$ and let $\xi\in\Stab{n}$ be the unique stabilizer element such that $g_\xi^{NN}(n) = \psiTMn^N\, \dxiTM \big(\psiTMn^N\big)^{-1} = g$.
    For brevity, we abbreviate $p := \exp_n \big(\psiTMn^N\big)^{-1} \mathscr{v}$ and thus $\xi(p) = \exp_n \big(\psiTMn^N\big)^{-1} g\mathscr{v}$, which is as justified by steps 1-3 below.
    We then find:
    \begin{align}
        \big[\Omega^{-1}(\kappa)\big](g\mathscr{v})
        \,&\overset{(1)}{=}\, \kappa\big(\! \exp_n\! \big(\psiTMn^N\big)^{-1} (g\mathscr{v}) \big) \cdot \rhoin\big( g_{n\leftarrow\xi(p)}^{NX} \big)^{-1}
               \ \sqrt{\big|\eta_{\xi(p)}^{\partial\mkern-2mu/\mkern-2mu\partial\mathscr{v}}\big|} \notag \\
        \,&\overset{(2)}{=}\, \kappa\big(\! \exp_n\! \dxiTM \big(\psiTMn^N\big)^{-1} \mathscr{v} \big) \cdot \rhoin\big( g_{n\leftarrow\xi(p)}^{NX} \big)^{-1}
               \ \sqrt{\big|\eta_{\xi(p)}^{\partial\mkern-2mu/\mkern-2mu\partial\mathscr{v}}\big|} \notag \\
        \,&\overset{(3)}{=}\, \kappa\big(\xi \exp_n\! \big(\psiTMn^N\big)^{-1} \mathscr{v} \big) \cdot \rhoin\big( g_{n\leftarrow\xi(p)}^{NX} \big)^{-1}
               \ \sqrt{\big|\eta_{\xi(p)}^{\partial\mkern-2mu/\mkern-2mu\partial\mathscr{v}}\big|} \notag \\
        \,&\overset{(4)}{=}\, \rhoout\big( g_\xi^{NN}(n) \big) \cdot \kappa\big(\exp_n\! \big(\psiTMn^N\big)^{-1} \mathscr{v} \big) \cdot \rhoin\big( g_\xi^{XP}(p) \big)^{-1}\, \rhoin\big( g_{n\leftarrow\xi(p)}^{NX} \big)^{-1}
               \ \sqrt{\big|\eta_{\xi(p)}^{\partial\mkern-2mu/\mkern-2mu\partial\mathscr{v}}\big|} \notag \\
        \,&\overset{(5)}{=}\, \rhoout\big( g_\xi^{NN}(n) \big) \cdot \kappa\big(\exp_n\! \big(\psiTMn^N\big)^{-1} \mathscr{v} \big) \cdot \rhoin\big( g_\xi^{XP}(p) \big)^{-1}\, \rhoin\big( g_{n\leftarrow\xi(p)}^{NX} \big)^{-1}
               \ \sqrt{\big|\eta_p^{\partial\mkern-2mu/\mkern-2mu\partial\mathscr{v}}\big|} \notag \\
        \,&\overset{(6)}{=}\, \rhoout\big( g_\xi^{NN}(n) \big) \cdot \kappa\big(\exp_n\! \big(\psiTMn^N\big)^{-1} \mathscr{v} \big) \cdot \rhoin\big( g_{n\leftarrow p}^{NP} \big)^{-1}\, \rhoin\big( g_\xi^{NN}(n) \big)^{-1}
               \ \sqrt{\big|\eta_p^{\partial\mkern-2mu/\mkern-2mu\partial\mathscr{v}}\big|} \notag \\
        \,&\overset{(7)}{=}\, \rhoout\big( g_\xi^{NN}(n) \big) \cdot \big[\Omega^{-1}(\kappa)\big](\mathscr{v}) \cdot \rhoin\big( g_\xi^{NN}(n) \big)^{-1} \notag \\
        \,&\overset{(8)}{=}\, \rhoout(g) \cdot \big[\Omega^{-1}(\kappa)\big](\mathscr{v}) \cdot \rhoin(g)^{-1}
    \end{align}
    The first three steps expanded $\Omega^{-1}(\kappa)$, used the definition of $\xi$ in terms of $g$ and the commutativity of exponential maps with isometry pushforwards, Eq.~\eqref{eq:exp_isom_commutation}.
    In the fourth step, the $\Stab{n}$-steerability constraint of $\kappa$ from Eq.~\eqref{eq:spherical_steerable_kernel_space} is used.
    Step five replaced again the Riemannian volume element at $\xi(p)$ with that at $p$ since they are equal.
    The sixth step used the relation
    \begin{align}
        \rhoin\big( g_\xi^{XP}(p) \big)^{-1}\, \rhoin\big( g_{n\leftarrow\xi(p)}^{NX} \big)^{-1}
        \ =&\ \rhoin\big( g_{n\leftarrow\xi(p)}^{NX}\, g_\xi^{XP}(p) \big)^{-1} \notag \\
        \ =&\ \Big( \pig[ \psiAinn^N\, \PAinnxip\, \big(\psiAinxip^X \big)^{-1} \pig] \pig[ \psiAinxip^X\, \dxiAin\, \big(\psiAinp^P \big)^{-1} \pig] \Big)^{-1} \notag \\
        \ =&\ \Big( \psiAinn^N\, \PAinnxip\, \dxiAin\, \big(\psiAinp^P \big)^{-1} \Big)^{-1} \notag \\
        \ =&\ \Big( \psiAinn^N\, \dxiAin\, \PAinnp\, \big(\psiAinp^P \big)^{-1} \Big)^{-1} \notag \\
        \ =&\ \Big( \psiAinn^N\, \dxiAin\, \big(\psiTMn^N\big)^{-1}\, \psiTMn^N\, \PAinnp\, \big(\psiAinp^P \big)^{-1} \Big)^{-1} \notag \\
        \ =&\ \rhoin\big( g_{n\leftarrow p}^{NP} \big)^{-1}\, \rhoin\big( g_\xi^{NN}(n) \big)^{-1} \,,
    \end{align}
    which relies again on the commutativity of transporters and isometry pushforwards from Eq.~\eqref{eq:transport_isom_commutation}.
    The last two steps identify $\Omega^{-1}(\kappa)$ and, by definition of $\xi$, that $g_\xi^{NN}(n) = g$.

    Together, these arguments that $\Omega$ is indeed an isomorphism between the kernel spaces.
\end{proof}

\toclesslab\subsection{Proof of Theorem~\ref{thm:spherical_conv_GM_conv} -- Equivalence of steerable spherical and \textit{GM}-convolutions}{apx:spherical_conv_equivalence}

Theorem~\ref{thm:spherical_conv_GM_conv} claims that $\GM$-convolutions with a $G$-steerable kernel $K \in \mathscr{K}^{G,B_{\R^2}(0,\pi)}_{\rhoin\mkern-1mu,\rhoout}$ are equivalent to the spherical convolution with the $\Stab{n}$-steerable kernel $\Omega(K) \in \mathscr{K}^{\Stab{n}}_{\rhoin\mkern-1mu,\rhoout}$.
The spherical convolution with a $\Stab{n}$-steerable kernel $\kappa \in \mathscr{K}^{\Stab{n}}_{\rhoin\mkern-1mu,\rhoout}$ from \citet{Cohen2018-intertwiners,Cohen2019-generaltheory} was hereby in Eq.~\eqref{eq:spherical_steerable_conv} pointwise defined as
\begin{align}
    \big[\kappa \star_{\mkern-2mu S^2}\! f\big]^P(p)
    \ = \int\limits_{S^2 \backslash \mkern-2mu -p} \mkern-8mu \kappa\big(\phi_p^{-1}q)\, \rhoin\big( g_{\phi_p^{-1}}^{XQ}(q) \big)\, f^Q(q)\ dq \,,
\end{align}
where $P$, $Q$ and $X$ denote arbitrary gauges at $p$, $q$ and $\phi_p^{-1}(q)$, respectively.
The isometry $\phi_p \in \I$ is uniquely specified by demanding that $\dphipGM \sigma^N(n) = \sigma^P(p)$.
Note that this implies in particular that
\begin{align}\label{eq:phipn_sphere_action}
    \phi_p(n) \ =\ p
\end{align}
and, using the definition of sections of $\GM$ (frame fields) in terms of inverse gauges from Eq.~\eqref{eq:GM_section_psi_inverse_def}, that
\begin{align}\label{eq:phipn_gauges}
    \psiTMn^N \circ \dphipGM^{-1} \ =\ \psiTMp^P \,,
\end{align}
both of which we will use below.
With these preparations, we turn to the proof of Theorem~\ref{thm:spherical_conv_GM_conv}, i.e. the equivalence
\begin{align}
    \Omega(K) \star_{\mkern-2mu S^2}\! f\ =\ K \star_{\mkern-1mu\scalebox{.64}{$\GM$}} f
\end{align}
of the convolutions.

\begin{proof}
    Since $\Omega(\kappa)$ is defined on $S^2 \backslash \mkern-1mu\minus\mkern1mu n$, the transformed kernel $\Omega(\kappa) \circ \phi_p^{-1}$ is defined on $S^2 \backslash \mkern-2mu -p$.
    Inserting $\Omega(\kappa)$ in the pointwise definition of the spherical convolution in Eq.~\eqref{eq:spherical_steerable_conv} leads therefore to
    \begin{align}
        \big[\Omega(K) \star_{\mkern-2mu S^2}\! f\big]^P(p)
        \ &= \int\limits_{S^2 \backslash \mkern-2mu -p} \mkern-8mu \big[\Omega(\kappa)\big] \big(\phi_p^{-1}q)\; \rhoin\big( g_{\phi_p^{-1}}^{XQ}(q) \big)\; f^Q(q)\ dq \\
        \ &= \int\limits_{S^2 \backslash \mkern-2mu -p} \mkern-8mu K\big( \psiTMn^N \log_n \phi_p^{-1}q)\; \rhoin\big( g_{n\leftarrow \phi_p^{-1}(q)}^{NX} \big) \; \rhoin\big( g_{\phi_p^{-1}}^{XQ}(q) \big)\; f^Q(q)\; \sqrt{\big|\eta_{\phi_p^{-1}(q)}^{\partial\mkern-2mu/\mkern-2mu\partial\mathscr{v}}\big|}^{\,-1} dq \,, \notag
    \end{align}
    where the second step follows by expanding $\Omega(K)$ as defined in Eq.~\eqref{eq:spherical_kernel_space_iso_Omega}.
    To simplify this expression, note that 
    \begin{align}
        \psiTMn^N\, \log_n\, \phi_p^{-1} (q)
        \ =\ \psiTMn^N\, \dphipTM^{-1}\, \log_{\phi_p(n)} (q)
        \ =\ \psiTMp^P\, \log_p (q) \,,
    \end{align}
    which follows from Eq.~\eqref{eq:exp_isom_commutation} in the first step and Eqs.~\eqref{eq:phipn_sphere_action} and~\eqref{eq:phipn_gauges} in the second step.
    Note furthermore, that
    \begin{alignat}{3}
        \qquad
           &\ \rho\big( g_{n\leftarrow \phi_p^{-1}(q)}^{NX} \big) \; \rho\big( g_{\phi_p^{-1}}^{XQ}(q) \big) \notag \\
        \ =&\ \pig[ \psiAn^N\, \mathcal{P}_{\A, n\leftarrow \phi_p^{-1}(q)}\, \big(\psiAphipinvq^X\big)^{-1} \pig]
              \pig[ \psiAphipinvq^X\, \dphipA^{-1}\, \big(\psiAq^Q\big)^{-1} \pig]
            \qquad && \big( \text{\small Eqs.~\eqref{eq:transporter_gauge_A} and~\eqref{cd:pushforward_A_coord} } \big) \notag\\
        \ =&\ \psiAn^N\, \mathcal{P}_{\A, n\leftarrow \phi_p^{-1}(q)}\, \dphipA^{-1}\, \big(\psiAq^Q\big)^{-1}
            \qquad && \big( \text{\small canceled inverse gauges } \big) \notag\\
        \ =&\ \psiAn^N\, \dphipA^{-1}\, \mathcal{P}_{\A, \phi_p(n)\leftarrow q}\, \big(\psiAq^Q\big)^{-1}
            \qquad && \big( \text{\small Eq.~\eqref{eq:transport_isom_commutation} } \big) \notag\\
        \ =&\ \psiAp^P\, \mathcal{P}_{\A, p\leftarrow q}\, \big(\psiAq^Q\big)^{-1}
            \qquad && \big( \text{\small Eq.~\eqref{eq:phipn_gauges} } \big) \notag\\
        \ =&\ \rho\big( g_{p\leftarrow q}^{PQ} \big) \,.
            \qquad && \big( \text{\small Eq.~\eqref{eq:transporter_gauge_A} } \big) \notag
    \end{alignat}
    Inserting these two identities, we obtain
    \begin{align}
        \big[\Omega(K) \star_{\mkern-2mu S^2}\! f\big]^P(p)
        \ &= \int\limits_{S^2 \backslash \mkern-2mu -p} \mkern-8mu K\big( \psiTMp^P \log_p q)\; \rhoin\big( g_{p\leftarrow q}^{PQ} \big)\; f^Q(q)\; \sqrt{\big|\eta_{\phi_p^{-1}(q)}^{\partial\mkern-2mu/\mkern-2mu\partial\mathscr{v}}\big|}^{\,-1} dq \,, \notag
    \end{align}
    To proceed, we express the integral in geodesic normal coordinates 
    $\mathscr{v}: S^2\backslash \mkern-1mu\minus\mkern1mu p \to B_{\R^2}(0,\pi),\ \ q \mapsto \mathscr{v}(q) := \psiTMp^P \log_p q$
    of $S^2 \backslash \mkern-2mu -p$, which are centered at point $p$.
    This cancels the Riemannian volume factor
    $\sqrt{\big|\eta_{\phi_p^{-1}(q)}^{\partial\mkern-2mu/\mkern-2mu\partial\mathscr{v}}\big|}$
    (and thus justifies its appearance in the definition of $\Omega$),
    such that the spherical convolution becomes
    \begin{align}
        \big[\Omega(K) \star_{\mkern-2mu S^2}\! f\big]^P(p)
        \ =& \int\limits_{B_{\R^2}\mkern-1mu(0,\pi)} \mkern-8mu K(\mathscr{v})\; \rhoin\pig( g_{p\leftarrow \exp_p (\psiTMp^P)^{-1} \mathscr{v}}^{PQ} \pig)\; f^Q\big( \exp_p (\psiTMp^P)^{-1} \mathscr{v} \big)\ d\mathscr{v} \,, \notag \\
        \ =& \int\limits_{B_{\R^2}\mkern-1mu(0,\pi)} \mkern-8mu K(\mathscr{v})\; \big[\Expspf\big]^P(\mathscr{v})\ d\mathscr{v} \,, \notag \\
       =&\ \big[K \star_{\mkern-1mu\scalebox{.64}{$\GM$}}\! f\big]^P(p) \,.
    \end{align}
    Since all arguments are independent form the chosen point $p$ and the chosen gauges, this implies
    \begin{align}
        \Omega(K) \star_{\mkern-2mu S^2}\! f\ =\ K \star_{\mkern-1mu\scalebox{.64}{$\GM$}} f \,,
    \end{align}
    in a coordinate free setting, which proves the theorem.
\end{proof}

%% file: chapters/apx07_smoothness_kernel_field_trafo.tex

\filbreak

\section{Existence and smoothness of kernel field transforms}
\label{apx:smoothness_kernel_field_trafo}

In Def.~\ref{dfn:kernel_field_trafo} we proposed \emph{kernel field transforms} $\TK$ as smooth integral transforms
\begin{align}
    \TK: \Gamma(\Ain)\to \Gamma(\Aout)
\end{align}
which are parameterized by some kernel field $\K$ (Def.~\ref{dfn:kernel_field_general}) and are pointwise given by
\begin{align}\label{eq:APX_kernel_field_trafo_def_ptwise}
    \big[ \TK (f)\big] (p)
    \ \ :=\ 
    \int\limits_{\TpM}\!\!
    \K(v) \,
    \Expspf(v)
    \ dv
    \ \ =\ 
    \int\limits_{\TpM}\!\!
    \K(v) \ 
    \PAinexppv \,
    f(\exp_p\!v)
    \ dv \,.
\end{align}
Kernel field transforms include $\GM$-convolutions from Def.~\ref{dfn:coord_free_conv} as a special cases for $\GM$-convolutional kernel fields.

Here we briefly discuss the well-definedness of kernel field transforms.
It is clear that the integrand of Eq.~\eqref{eq:APX_kernel_field_trafo_def_ptwise} lies for any $p\in M$ and $v \in \TpM$ in $\Aoutp$.
What remains to be shown is the \emph{existence} of the integral and the \emph{smoothness} of the resulting feature field.
In the following we will first give some general remarks on how to approach these questions.
We will then prove Theorem~\ref{thm:existence_kernel_field_trafo_compact_kernels}, i.e. the well-definedness of kernel field transforms for the specific case of fields of kernels which are compactly supported on a ball of fixed radius around the origin.

\paragraph{Existence:}
The existence and smoothness of kernel field transforms requires a suitable choice of kernel field~$\K$.
Similar to the case of conventional convolutions on $M=\R$, the requirements on $\K$ in order for the kernel field transform to exist depend on the specific properties of the input feature field $f\in\Gamma(\Ain)$.%
\footnote{
    See the discussion at \url{https://en.wikipedia.org/wiki/Convolution\#Domain_of_definition}.
}
In~general, $\K$ needs to decay sufficiently rapidly in order to make the integrand in Eq.~\eqref{eq:APX_kernel_field_trafo_def_ptwise} integrable.

A special case of great practical importance is that of kernels $\Kp: \TpM \to \Hom(\Ainp,\Aoutp)$ which are at any $p\in M$ \emph{compactly supported}.
In this case the integral is always guaranteed to exist.
To see this, note that (input) feature fields and kernel fields are defined to be smooth.
The smoothness of the metric further implies that the
Riemannian volume density, the exponential map and the parallel transport are smooth~\cite{gallier2019diffgeom1}.
In combination, the whole integrand in Eq.~\eqref{eq:APX_kernel_field_trafo_def_ptwise} is seen to be a smooth and thus continuous function from~$TM$ to~$\Aout$.
If~$\Kp$ is in addition compactly supported, the integrand becomes continuous and compactly supported on $\TpM$, which, by a generalization of the extreme value theorem, implies that its image is compact (and in a local trivialization $\R^{c}$ of $\Aoutp$ bounded)~\cite{rudin1976analysis}.
This guarantees the existence of the integral~\cite{forster2012analysis3,spivak2019calculus}.

Depending on the application, the requirement on the support of $\K$ might be relaxed.
For instance, images on $M=\R^d$ are usually compactly supported themselves, such that no additional properties of $\K$ except for its smoothness are required.

\paragraph{Smoothness:}

We turn to discuss the smoothness of kernel field transforms, that is, their property to map smooth input fields $\fin\in\Gamma(\Ain)$ to smooth output fields $\fout := \TK(\fin) \in \Gamma(\Aout)$.
By definition, a map $\fout: M\to \Aout$ \emph{between manifolds} $M$ and $\Aout$ is said to be smooth if its coordinate representations are smooth.
In equations, $\fout$ is smooth if for any $p\in M$ there exist smooth charts $(U,\phi)$ about $p$ in $M$ and $(\widetilde{U}, \widetilde{\phi})$ about $\fout(p)$ in $\Aout$ with $\fout(U) \subseteq \widetilde{U}$ such that $\widetilde{\phi} \circ \fout \circ \phi^{-1}: \phi(U) \to \widetilde{\phi}(\widetilde{U})$ is smooth as a map between (subsets of) Euclidean spaces.
Given $(U,\phi)$, a convenient choice for $(\widetilde{U}, \widetilde{\phi})$ would be
$ \widetilde{\phi} := \big(\phi \times \id\big) \circ \PsiAout\!:\ 
  \piAout^{-1}(U) \mapsto \phi(U) \times \R^c\, \subseteq\, \R^d \times \R^c \,, $
however, the following discussion is independent from this choice.%
\footnote{
    Note that $\piAout^{-1}(U)$ is guaranteed to be trivializable given that $(\phi,U)$ is a chart of $M$.
    This is clear since the coordinate bases $\big[ \frac{\partial}{\partial \phi_\mu} \big]_{\mu=1}^d$ of $(\phi,U)$ yields a trivialization of $\piTM^{-1}(U)$ (see Appendix~\ref{apx:coordinate_bases}) and since the local trivializations of $FM$ and $\A$ were in Section~\ref{sec:bundle_trivializations} induced from those of $TM$.
}
A map between (subsets of) Euclidean spaces is smooth if it is smooth in each component of its image, here in each of the $d+c$ dimensions of $\phi(U)\times\R^c$.
We are therefore interested in the smoothness of the maps
\begin{align}
    F_i: \phi(U) \to \R,\ \ x \mapsto \Big[\, \widetilde{\phi} \circ \fout \circ \phi^{-1} \Big]_i
\end{align}
for any $i = 1,\dots,d+c$.
By writing out $\fout$ and expressing the integral over $\TpM$ by an integral over $\R^d$ as discussed in Appendix~\ref{apx:correspondences_bundle_trivializations}, the $F_i$ are seen to be of the form
\begin{align}\label{eq:integral_smoothness_component}
    F_i(x) = \int_{\R^d} I_i(\mathscr{v},x)\,\ d\mathscr{v} \,.
\end{align}
The coordinate expressions of the integrands $I_i$ are hereby for any $i = 1,\dots,d+c$ given by
\begin{align}\label{eq:integrand_smoothness_full}
    I_i\!: \R^d \mkern-2mu\times\mkern-2mu \phi(U) &\to \R, \\
    (\mathscr{v},x) &\mapsto
    \bigg[ \widetilde{\phi} \circ
    \K\pig(\mkern-1mu \psi_{\protect\scalebox{.65}{$T\!M\mkern-1mu,\mkern2mu$}\protect\scalebox{.75}{$\phi^{\mkern-1mu\shortminus1}\mkern-1mu(x)$}}^{-1} \mkern-2mu(\mathscr{v}) \mkern-1mu\pig) \circ
    \mathcal{P}_{\mkern-4mu\protect\scalebox{.8}{$\!\Ain$},\protect\scalebox{.83}{$\,\phi^{\mkern-1mu\shortminus1}(x)\!\leftarrow\!\exp \circ\mkern1mu \psi_{\protect\scalebox{.7}{$T\!M\mkern-1mu,\mkern1mu$}\protect\scalebox{.9}{$\phi^{\mkern-1mu\shortminus1}(x)$}}\!(\mathscr{v}) $}} \mkern-1mu\circ
    \fin \circ \exp \circ \,
    \psi_{\protect\scalebox{.65}{$T\!M\mkern-1mu,$}\protect\scalebox{.7}{$\phi^{\mkern-1mu\shortminus1}(x)$}}^{-1} \mkern-2mu(\mathscr{v})
    \bigg]_i
    \nonumber \,,
\end{align}
where we assumed, for convenience and without loss of generality, that
$\psi_{\protect\scalebox{.65}{$T\!M\mkern-1mu,\mkern2mu$}\protect\scalebox{.75}{$\phi^{\mkern-1mu\shortminus1}\mkern-1mu(x)$}}$
is an \emph{isometric} gauge of $T_{\phi^{\mkern-1mu\shortminus1}\mkern-1mu(x)}M$, such that the volume scaling factor $\sqrt{|\! \det\eta_p|} = 1$ drops out.
Note that the integrands $I_i$ are composed of smooth maps and are therefore smooth as well.

From the previous discussion it is clear that the smoothness of $\fout$ holds if all $F_i$ are smooth, i.e. infinitely often partially differentiable.
To prove the smoothness of the $F_i$, it is sufficient to show that the partial differentiations and the integration in Eq.~\eqref{eq:integral_smoothness_component} commute -- which is not always the case.
If they do commute, partial derivatives of arbitrary orders $(n_1,\dots,n_d) \in \N^d$ are given by
\begin{align}\label{eq:APX_smoothness_partial_diff_swapping}
    \Big[ \partial_{x_1}^{n_1} \dots \partial_{x_d}^{n_d}\: F_i \Big](x)
    \ =\ \int_{\R^d} \Big[ \partial_{x_1}^{n_1} \dots \partial_{x_d}^{n_d}\: I_i\Big] (\mathscr{v},x)\,\ d\mathscr{v}
\end{align}
where
1) the partial derivatives $\pig[ \partial_{x_1}^{n_1} \dots \partial_{x_d}^{n_d}\: I_i\pig]$ of the integrand exist (due to the smoothness of $I_i$ their existence is guaranteed) and
2) their integral exists.
Whether or not the differentiations commute with the integral can be investigated by making use of the following lemma
from~\cite{forster2012analysis3},%
\footnote{
    Similar versions of this lemma in English language can be found in \cite{klenke2006probability} or at
    \url{https://en.wikipedia.org/wiki/Leibniz_integral_rule\#Measure_theory_statement}.
    In contrast to those versions, the version from~\cite{forster2012analysis3} allows for $T$ being any non-degenerate interval, including closed intervals, which saves us some additional steps below.
}
which is a consequence of the dominated convergence theorem.
\begin{thm}[Differentiation lemma~\cite{forster2012analysis3}]
\label{thm:differentiation_lemma}
    Let $\mathscr{V}$ be a measure space, let $T\subset\R$ be a non-degenerate interval and let $I: \mathscr{V} \times T \to \R$ be a map with the following properties:
    \begin{itemize}
        \item[(i)]  For any fixed $t\in T$ the map $\mathscr{v} \mapsto I(\mathscr{v},t)$ is Lebesgue integrable on $\mathscr{V}$
        \item[(ii)] For any fixed $\mathscr{v} \in \mathscr{V}$ the map $t\mapsto I(\mathscr{v},t)$ is differentiable in $T$
        \item[(iii)] There exists a Lebesgue integrable function $\mathscr{B}: \mathscr{V} \to \R$ such that
                    $\big| \frac{\partial}{\partial t} I(\mathscr{v},t) \big| \leq \mathscr{B}(\mathscr{v})$
                    for any $(\mathscr{v},t) \in \mathscr{V} \times T$
    \end{itemize}
     Then the function $F: T \to \R,\ t \mapsto \int_\mathscr{V} I(\mathscr{v},t)\, d\mathscr{v}$ is differentiable
     with derivative
    \begin{align*}
        \frac{\partial}{\partial t} F(t)\ =\ \int_\mathscr{V} \frac{\partial}{\partial t} I(\mathscr{v},t)\; d\mathscr{v} \,.
    \end{align*}
\end{thm}
The applicability of this lemma (repeatedly for every single partial differentiation) depends on the properties of the integrand, which in turn depends on the specific properties of the kernel field $\K$ and the input feature field $\fin$.
For the case of a kernel field which is compactly supported on balls of fixed radius around the origin of each tangent space, the lemma applies.
Based on this, we give a proof of Theorem~\ref{thm:existence_kernel_field_trafo_compact_kernels} in the remainder of this appendix.

\toclesslab\subsection{Proof of Theorem~\ref{thm:existence_kernel_field_trafo_compact_kernels} (Sufficiency of compact kernel support on tangent balls)}{apx:proof_sufficiency_ball_kernel_support}

Denote by
$B_{\TpM}^{\mkern1.5mu\textup{closed}}(0,R) := \big\{ v\in\TpM \,\big|\, \lVert v\rVert \leq R \big\}$
the closed ball of radius $R>0$ around the origin of $\TpM$ and by
$B_{\R^d}^{\mkern1.5mu\textup{closed}}(0,R) := \big\{ v\in\R^d \,\big|\, \lVert v\rVert \leq R \big\}$
the corresponding ball around the origin of $\R^d$.
Note that any isometric gauge satisfies $\psiTMp\big(B_{\TpM}^{\mkern1.5mu\textup{closed}}(0,R)\big) = B_{\R^d}^{\mkern1.5mu\textup{closed}}(0,R)$.
Let $\widetilde{\K}$ be a kernel field whose support falls within balls of the same radius $R$ in each tangent space, i.e. which satisfies
\begin{align}
    \supp\!\big(\widetilde{\K}_p\big) \subseteq B_{\TpM}^{\mkern1.5mu\textup{closed}}(0,R) \quad \forall p\in M
\end{align}
and thus, for any isometric gauge $\psiTMp$:
\begin{align}
    \supp\! \pig(\widetilde{\K}_p \circ \big(\psiTMp\big)^{-1} \pig) \subseteq B_{\R^d}^{\mkern1.5mu\textup{closed}}(0,R) \quad \forall p\in M
\end{align}
According to Theorem~\ref{thm:existence_kernel_field_trafo_compact_kernels} this property is sufficient to guarantee that the corresponding kernel field transform $\mathscr{T}_{\mathcal{K}_{\mkern-1muR}}$ is well defined.
A proof of this statement is given in the following.

\begin{proof}
    As already stated in the beginning of this appendix, the existence of the integral is guaranteed given that the kernel supports are compact:
    The compactness of the kernels carries over to the integrands of the kernel field transform.
    Their smoothness further implies their continuity and integrals of compactly supported continuous functions always exists.

    To prove the smoothness of the resulting output feature field $\fout$, we proceed with the discussion earlier in this section.
    We aim to apply the differentiation lemma~\ref{thm:differentiation_lemma} to swap partial derivatives $\frac{\partial}{\partial x_\mu}$ for any $\mu=1,\dots,d$ in Eq.~\eqref{eq:APX_smoothness_partial_diff_swapping} at any $x_0 \in \phi(U)$ with the integration over $\R^d$.
    For this purpose, we introduce the auxiliary functions
    \begin{align}
        I_{i,x_0,\mu}: \R^d \times [-\varepsilon,\varepsilon] \to \R,\ \ (\mathscr{v},t) \mapsto I(\mathscr{v}, x_0 + t\epsilon_\mu)
    \end{align}
    and
    \begin{align}
        F_{i,x_0,\mu}: [-\varepsilon,\varepsilon] \to \R,\ \ t \mapsto F(x_0 + t\epsilon_\mu) = \int_{\R^d} I_{i,x_0,\mu}(\mathscr{v},t)\ d\mathscr{v} \,,
    \end{align}
    where $\epsilon_\mu \in \R^d$ is the unit vector in $\mu$-direction and
    $\varepsilon > 0$ is chosen such that $\big\{ x_0 + t\epsilon_\mu \,\big|\, t\in [-\varepsilon,\varepsilon] \big\} \subset \phi(U)$, which is always possible since $\phi(U)$ is open.
    Then $I_{i,x_0,\mu}$ is with the identifications $\mathscr{V}=\R^d$ and $T=[-\varepsilon,\varepsilon]$ of the form required by lemma~\ref{thm:differentiation_lemma}.
    It satisfies property $(i)$ by the assumption that the kernel field transform exists as discussed earlier.
    Property $(ii)$ holds due to the smoothness of the full integrand in Eq.~\eqref{eq:integrand_smoothness_full}.
    For property $(iii)$, observe that both $I_{i,x_0,\mu}$ and its derivative are smooth such that the absolute value $\big| \frac{\partial}{\partial t} I_{i,x_0,\mu} \big|$ is continuous.
    Since it is in addition compactly supported on $B_{\R^d}^{\mkern1.5mu\textup{closed}}(0,R) \times [-\varepsilon,\varepsilon]$, it is by (a generalization of) the extreme value theorem bounded by some number $b\geq0$.
    We therefore set $\mathscr{B}(\mathscr{v}) = b\cdot \mathbb{I}_{B_{\R^d}^{\mkern1.5mu\textup{closed}}(0,R) \times [-\varepsilon,\varepsilon]}$ where $\mathbb{I}$ is the indicator function.
    This choice satisfies 
    $\big| \frac{\partial}{\partial t} I(\mathscr{v},t) \big| \leq \mathscr{B}(\mathscr{v})$
    for any $(\mathscr{v},t) \in \mathscr{V} \times T$
    and is integrable such that property $(iii)$ is fulfilled as well.
    We can therefore swap the order of differentiation and integration for arbitrary choices of $x_0$ and $\mu$, which we use to pull arbitrary partial derivatives into the integral:
    \begin{align}
        \bigg[ \frac{\partial}{\partial x_\mu} F_i \bigg](x_0)
        \,=\, \bigg[ \frac{\partial}{\partial t} F_{i,x_0,\mu} \bigg](0)
        \,=\, \int_{\R^d} \frac{\partial}{\partial t} I_{i,x_0,\mu}(\mathscr{v},t) \Big|_{t=0}\ d\mathscr{v}
        \,=\, \int_{\R^d} \frac{\partial}{\partial x_\mu} I_i(\mathscr{v},x) \Big|_{x=x_0}\ d\mathscr{v}
    \end{align}

    Due to the smoothness and compact support of the integrand $I_i$, its partial derivatives $\frac{\partial}{\partial x_\mu} I$ are smooth and compactly supported as well.
    They do therefore satisfy properties $(i)$, $(ii)$ and $(iii)$ as well (with a potentially adapted bound $b$).
    It is thus possible to repeat the partial differentiation of $F_i$ infinitely often, which proves its smoothness.
    Since the derivations were independent from the particular choices for the point $p\in M$, charts $(U,\phi)$ and $(\widetilde{U},\widetilde{\phi})$, points $x_0\in\phi(U)$ and indices $i$ and $\mu$, this result proves the smoothness of the whole output feature field $\fout = \TK(\fin)$.

\NoEndMark
$~\hfill\Box$
\end{proof}

%% file: chapters/apx08_regular_fields_as_GM_functions.tex

\filbreak

\section{Regular feature fields as scalar functions on \textit{G}-structure}
\label{apx:regular_field_scalar_GM}

Real-valued functions $\digamma: \GM \to \R$ on the $G$-structure are equivalent to regular feature fields $f: M \to \A_{\textup{reg}}$ on the manifold, that is, that there is an isomorphism
\begin{align}
    C^\infty(\GM)\ \cong\ \Gamma(\A_{\textup{reg}}) \,.
\end{align}
This appendix presents a proof of this claim for the case of finite structure groups~$G$.
We start with the usual definition of (real) regular representations of finite structure groups, which act on the (free) vector spaces $\R^{|G|}$.
One defines a basis $\big\{ \epsilon_g \in \R^{|G|} \,\big|\, g\in G \big\}$ of $\R^{|G|}$, which is labeled by the group elements~$g\in G$.
The (left) regular representation's action on $\R^{|G|}$ is then defined in terms of its action on these basis vectors, which is given by left translation.
Specifically, for any $h,g\in G$, the regular representation acts as follows:
\begin{align}\label{eq:finite_regular_rep_action}
    \rho_{\textup{reg}}(h)\, \epsilon_g\ :=\ \epsilon_{hg} \,.
\end{align}
Note that the action on \emph{coefficients} of a vector is inverse
\begin{align}\label{eq:finite_regular_rep_action}
    \rho_{\textup{reg}}(h)\, \sum_{g\in G} \mathscr{f}_g\, \epsilon_g
    \ =\ \sum_{g\in G} \mathscr{f}_g\, \epsilon_{hg}
    \ =\ \sum_{{\widetilde{g}}\in G} \mathscr{f}_{h^{-1}\widetilde{g}}\ \epsilon_{\widetilde{g}} \,,
\end{align}
which is useful to know, however, we won't need this property in the following.
As the regular representation permutes the basis vectors of $\R^{|G|}$, it is a \emph{permutation representation}.
Some visualizations for the cyclic group $G=\C4$ are found in Appendix~B of~\cite{Weiler2019_E2CNN}.
Regular feature fields are defined as smooth sections of the associated $G$-bundle
\begin{align}
    \A_{\textup{reg}}\ =\ \big(\GM \times \R^{|G|}) /\! \sim_{\rho_\textup{reg}} \,.
\end{align}

The isomorphism $C^\infty(\GM) \cong \Gamma(\A_{\textup{reg}})$ substantiates our claim in Section~\ref{sec:instantiations_mesh} that the \emph{Parallel Frame CNNs} by~\citet{Yang2020parallelFrameCNN} are specific $\GM$-convolutions between regular feature fields.
It furthermore establishes the link between \emph{group convolutions} (see Section~\ref{apx:homogeneous_preliminaries}) and \emph{regular $\GM$-convolutions} that was claimed in Section~\ref{sec:euclidean_literature} and~\cite{Weiler2019_E2CNN}.

With these preparations and remarks we are ready to formulate the theorem:
\begin{thm}[Regular feature fields as scalar functions on $G$-structure]
\label{thm:regular_field_scalar_GM}
    Let $G \leq \GL{d}$ be a finite structure group, let $\GM$ be a $G$-structure over $M$ and let $\A_{\textup{reg}}$ be the bundle that is associated by the action of the regular representation $\rho_\textup{reg}$ of $G$.
    Regular feature fields are then identical to smooth, real-valued functions on the $G$-structure, that is, there is an isomorphism
    \begin{align}\label{eq:regular_field_associated_bundle_def}
        \Lambda: C^\infty(\GM) \xrightarrow{\sim} \Gamma(\A_{\textup{reg}}) \,.
    \end{align}
    This isomorphism is defined by
    \begin{align}
    \label{eq:regular_field_scalar_GM_iso_lambda}
        \big[\Lambda\digamma\big](p)
        \ &=\ \Big[ [e_i]_{i=1}^d \,,\, \sum\nolimits_g \digamma\big( [e_i]_{i=1}^d\lhd g\big)\, \epsilon_g \Big] \,,
    \intertext{
    where $[e_i]_{i=1}^d \in\GpM$ is an arbitrarily chosen representative frame at~$p$.
    Its inverse is given by
    }
    \label{eq:regular_field_scalar_GM_iso_lambda_inv}
        \big[\Lambda^{-1}f\big] \big([e_i]_{i=1}^d\big)
        \ &=\ \Big\langle \epsilon_e \,,\, \psiAp^{[e_i]_{i=1}^d} f(p) \Big\rangle \,,
    \end{align}
    where we abbreviated $p = \piGM(E)$ and denote by $\psiAp^{[e_i]_{i=1}^d}$ that (unique) gauge that corresponds to the frame ${[e_i]_{i=1}^d}$, i.e. which satisfies $\psiAp^{[e_i]_{i=1}^d}\big( [e_i]_{i=1}^d\big) = e$.
\end{thm}
\begin{proof}
    To prove this statement, we need to show that
    \textit{1)} the isomorphism preserves the smoothness of the maps,
    \textit{2)} that the choice of representative frame $[e_i]_{i=1}^d \in\GpM$ in the definition of $\Lambda$ is indeed arbitrary and
    \textit{3)} that $\Lambda^{-1}$ is indeed a left and right inverse of $\Lambda$.

    \item[] {\textit{1)} smoothness : }
    \begin{itemize}[leftmargin=1.1cm]
    \setlength\itemsep{2ex}
        \item[]
        That the isomorphism preserves the smoothness of the equivalent field representations is clear since all involved morphisms (right action, gauge map, inner product) are smooth.
    \end{itemize}

    \item[] {\textit{2)} independence of the definition of $\Lambda$, Eq.~\eqref{eq:regular_field_scalar_GM_iso_lambda}, from the choice of representative frame ${[e_i]_{i=1}^d \in \GpM}$: }
    \begin{itemize}[leftmargin=1.1cm]
    \setlength\itemsep{2ex}
        \item[]
        Suppose that we used \emph{any} other frame $[e_i]_{i=1}^d \lhd h$ for an arbitrary $h\in G$.
        This arbitrary gauge transformation drops then out by making use of the equivalence relation $\sim_{\rho_\textup{reg}}$ that is underlying the associated bundle construction, Eq.~\eqref{eq:regular_field_associated_bundle_def}:
        \begin{alignat}{3}
            \big[\Lambda\digamma\big](p)
            \ =&\ \Big[ [e_i]_{i=1}^d \lhd h \,,\, \sum\nolimits_g \digamma\big( [e_i]_{i=1}^d \lhd hg)\, \epsilon_g \Big]
                \qquad && \big( \text{\small def. of $\Lambda$, Eq.~\eqref{eq:regular_field_scalar_GM_iso_lambda} } \big) \notag\\
            \ =&\ \Big[ [e_i]_{i=1}^d \,,\, \rho_\textup{reg}(h) \sum\nolimits_g \digamma\big( [e_i]_{i=1}^d \lhd hg)\, \epsilon_g \Big]
                \qquad && \big( \text{\small equivalence relation $\sim_{\rho_\textup{reg}}$, Eq.~\eqref{eq:equiv_relation_A} } \big) \notag\\
            \ =&\ \Big[ [e_i]_{i=1}^d \,,\, \sum\nolimits_g \digamma\big( [e_i]_{i=1}^d \lhd hg)\, \epsilon_{hg} \Big]
                \qquad && \big( \text{\small $\rho_{\textup{reg}}$ action on basis $\epsilon_g$, Eq.~\eqref{eq:finite_regular_rep_action} } \big) \notag\\
            \ =&\ \Big[ [e_i]_{i=1}^d \,,\, \sum\nolimits_{\widetilde{g}} \digamma\big( [e_i]_{i=1}^d \lhd \widetilde{g})\, \epsilon_{\widetilde{g}} \Big]
                \qquad && \big( \text{\small substitution $\widetilde{g} = hg$ } \big)
        \end{alignat}
    \end{itemize}

    \item[] {\textit{3)} $\Lambda^{\!-1}$ in Eq.~\eqref{eq:regular_field_scalar_GM_iso_lambda_inv} is a well defined inverse of $\Lambda$ in Eq.~\eqref{eq:regular_field_scalar_GM_iso_lambda} : }

    \begin{itemize}[leftmargin=1.1cm]
    \setlength\itemsep{2ex}

        \item[\it 3\hspace{1pt}a)]
            $\Lambda^{-1} \!\circ\! \Lambda = \id_{C^\infty(\GM)}$,
            that is, $\Lambda^{-1}$ is a left inverse of $\Lambda$ :

            For any $\digamma\in C^{\infty}(\GM)$ and any $[e_i]_{i=1}^d$ this is shown as follows:
            \begin{align}
                &\ \big[\Lambda^{-1} \!\circ\! \Lambda\, \digamma \big] \big( [e_i]_{i=1}^d \big) \notag \\
                \ =&\ \Big\langle \epsilon_e \,,\, \psiAp^{[e_i]_{i=1}^d} [\Lambda\digamma](p) \Big\rangle
                    && \big( \text{\small def. of $\Lambda^{-1}$, Eq.~\eqref{eq:regular_field_scalar_GM_iso_lambda_inv} } \big) \notag\\
                \ =&\ \Big\langle \epsilon_e \,,\, \psiAp^{[e_i]_{i=1}^d} \big[ [e_i]_{i=1}^d \,,\, \sum\nolimits_g \digamma\big( [e_i]_{i=1}^d \lhd g \big)\, \epsilon_g \big] \Big\rangle
                    && \big( \text{\small def. of $\Lambda$, Eq.~\eqref{eq:regular_field_scalar_GM_iso_lambda} } \big) \notag\\
                \ =&\ \Big\langle \epsilon_e \,,\, \sum\nolimits_g \digamma\big( [e_i]_{i=1}^d \lhd g \big)\, \epsilon_g \Big\rangle
                    && \big( \text{\small def. of $\psiAp$, Eq.~\eqref{eq:trivialization_A_p} } \big) \notag\\
                \ =&\ \sum\nolimits_g \digamma\big( [e_i]_{i=1}^d \lhd g \big)\,
                    \langle \epsilon_e, \epsilon_g \rangle
                    && \big( \text{\small pull inner product into sum\,} \big) \notag\\
                \ =&\ \digamma\big( [e_i]_{i=1}^d \big)
                    && \big( \text{\small Kronecker delta $\delta_{e,g} = \langle \epsilon_e, \epsilon_g \rangle$} \big) \notag\\
            \end{align}

        \item[\it 3\hspace{1pt}b)]
            $\Lambda \!\circ\! \Lambda^{-1} = \id_{\Gamma(\A_{\textup{reg}})}$,
            that is, $\Lambda^{-1}$ is a right inverse of $\Lambda$ :

            Let $f\in\Gamma(\A_{\textup{reg}})$ and $p\in M$, then:
            \begin{align}
                &\ \big[\Lambda \!\circ\! \Lambda^{-1} f \big] (p) \notag \\
                \ =&\ \Big[ [e_i]_{i=1}^d \,,\, \sum\nolimits_g \big[\Lambda^{-1}f]\big( [e_i]_{i=1}^d\lhd g\big)\, \epsilon_g \Big]
                    && \big( \text{\small def. of $\Lambda$, Eq.~\eqref{eq:regular_field_scalar_GM_iso_lambda} } \big) \notag\\
                \ =&\ \Big[ [e_i]_{i=1}^d \,,\, \sum\nolimits_g \Big\langle \epsilon_e \,,\, \psiAp^{[e_i]_{i=1}^d\lhd g} f(p) \Big\rangle\, \epsilon_g \Big]
                    && \big( \text{\small def. of $\Lambda^{-1}$, Eq.~\eqref{eq:regular_field_scalar_GM_iso_lambda_inv} } \big) \notag\\
                \ =&\ \Big[ [e_i]_{i=1}^d \,,\, \sum\nolimits_g \Big\langle \epsilon_e \,,\, \rho_{\textup{reg}}(g)^{-1} \psiAp^{[e_i]_{i=1}^d} f(p) \Big\rangle\, \epsilon_g \Big]
                    && \big( \text{\small gauge transformation, Eq.~\eqref{eq:transition_fct_A}} \big) \notag\\
                \ =&\ \Big[ [e_i]_{i=1}^d \,,\, \sum\nolimits_g \Big\langle \rho_{\textup{reg}}(g) \epsilon_e \,,\, \psiAp^{[e_i]_{i=1}^d} f(p) \Big\rangle\, \epsilon_g \Big]
                    && \big( \text{\small unitarity of $\rho_{\textup{reg}}$} \big) \notag\\
                \ =&\ \Big[ [e_i]_{i=1}^d \,,\, \sum\nolimits_g \Big\langle \epsilon_g \,,\, \psiAp^{[e_i]_{i=1}^d} f(p) \Big\rangle\, \epsilon_g \Big]
                    && \big( \text{\small $\rho_{\textup{reg}}$ action on basis $\epsilon_e$, Eq.~\eqref{eq:finite_regular_rep_action}} \big) \notag\\
                \ =&\ \Big[ [e_i]_{i=1}^d \,,\, \psiAp^{[e_i]_{i=1}^d} f(p) \Big]
                    && \big( \text{\small remove expansion in basis $\epsilon_g$} \big) \notag\\
                \ =&\ f(p)
                    && \big( \text{\small def. of $\psiAp$, Eq.~\eqref{eq:trivialization_A_p}} \big)
            \end{align}

    \end{itemize}

    This concludes our prove of the equivalence of $C^\infty(\GM)$ and $\Gamma(\A_\textup{reg})$.
\end{proof}